\documentclass[screen, authorversion, acmsmall, pagebackref]{acmart}
\usepackage{amsmath}
\usepackage{graphicx}
\usepackage{hyperref}
\usepackage[utf8]{inputenc}
\usepackage{xcolor}
\usepackage{ifthen}
\usepackage[normalem]{ulem}
\usepackage{svg}
\usepackage{subfiles}
\usepackage{xspace}
\usepackage{mathtools}
\usepackage{enumitem}
\usepackage{tikz-cd}

\usepackage{amssymb}

\newif\ifshowname

\shownamefalse

\usepackage[noabbrev, capitalise]{cleveref} 
\usepackage{subfig}
\usepackage{longtable}

\usepackage{wrapfig}
\usepackage{dialogue}

\usepackage[most]{tcolorbox}
\usepackage{tabularx}
\usepackage{comment}
\usepackage{bm}
\usepackage{braket}
\usepackage{multirow}
\usepackage{makecell}
\usepackage{todonotes}
\usepackage{nameref}
\usepackage[version=4]{mhchem}  

\usepackage{xcolor}
\NewDocumentCommand{\heng}
{ mO{} }{\textcolor{red}{\textsuperscript{\textit{Heng}}\textsf{\textbf{\small[#1]}}}}

\NewDocumentCommand{\ray}
{ mO{} }{\textcolor{cyan}{\textsuperscript{\textit{Rui Wang}}\textsf{\textbf{\small[#1]}}}}

\NewDocumentCommand{\rose}
{ mO{} }{\textcolor{magenta}{\textsuperscript{\textit{Rose Yu}}\textsf{\textbf{\small[#1]}}}}

\NewDocumentCommand{\shuiwang}
{ mO{} }{\textcolor{blue}{\textsuperscript{\textit{Shuiwang Ji}}\textsf{\textbf{\small[#1]}}}}

\NewDocumentCommand{\michael}
{ mO{} }{\textcolor{blue}{\textsuperscript{\textit{Michael Bronstein}}\textsf{\textbf{\small[#1]}}}}

\NewDocumentCommand{\yi}
{ mO{} }{\textcolor{purple}{\textsuperscript{\textit{Yi Liu}}\textsf{\textbf{\small[#1]}}}}

\NewDocumentCommand{\xuan}
{ mO{} }{\textcolor{orange}{\textsuperscript{\textit{Xuan Zhang}}\textsf{\textbf{\small[#1]}}}}

\NewDocumentCommand{\yuchao}
{ mO{} }{\textcolor{teal}{\textsuperscript{\textit{Yuchao Lin}}\textsf{\textbf{\small[#1]}}}}

\NewDocumentCommand{\limei}
{ mO{} }{\textcolor{olive}{\textsuperscript{\textit{Limei}}\textsf{\textbf{\small[#1]}}}}

\newcommand{\revisionOne}[1]{#1}

\newcommand{\revisionTwo}[1]{#1}

\usepackage{amssymb}
\usepackage{pifont}
\newcommand{\cmark}{\ding{51}}
\newcommand{\xmark}{\ding{55}}

\newcommand\R{\ensuremath{\mathbb{R}}} 

\ifthenelse{\isundefined{\definition}}{\newtheorem{definition}{Definition}}{}
\ifthenelse{\isundefined{\assumption}}{\newtheorem{assumption}{Assumption}}{}
\ifthenelse{\isundefined{\hypothesis}}{\newtheorem{hypothesis}{Hypothesis}}{}
\ifthenelse{\isundefined{\proposition}}{\newtheorem{proposition}{Proposition}}{}
\ifthenelse{\isundefined{\theorem}}{\newtheorem{theorem}{Theorem}}{}
\ifthenelse{\isundefined{\lemma}}{\newtheorem{lemma}{Lemma}}{}
\ifthenelse{\isundefined{\corollary}}{\newtheorem{corollary}{Corollary}}{}
\ifthenelse{\isundefined{\alg}}{\newtheorem{alg}{Algorithm}}{}
\ifthenelse{\isundefined{\example}}{\newtheorem{example}{Example}}{}

\def\stable{1}

\if\stable1
\else
\fi

\if\stable1

  \newcommand\status[1]{}
  \newcommand\defend[1]{}
\else

  \newcommand\status[1]{\textcolor{gray}{\bf [Status: #1]}}
  \newcommand\defend[1]{\textcolor{magenta}{[Cite: #1]}}
\fi

\if\stable1

\newcommand\pl[1]{}
\newcommand\rb[1]{}
\newcommand\ilevent[1]{}
\newcommand\mx[1]{}
\newcommand\FT[1]{}
\newcommand\tnote[1]{}
\newcommand\dor[1]{}
\newcommand\pw[1]{}
\newcommand\at[1]{}
\newcommand\dora[1]{}

\else

\newcommand\pl[1]{\textcolor{purple}{[Percy: #1]}}
\newcommand\rb[1]{\textcolor{magenta}{[Rishi: #1]}}

\newcommand{\tnote}[1]{{\color{blue}{[TM: #1]}}}

\definecolor{CMpurple}{rgb}{0.6,0.18,0.64}
\definecolor{atcolor}{rgb}{0.83,0.28,0.06}
\definecolor{ballblue}{rgb}{0.13, 0.67, 0.8}

\newcommand\cms{\bgroup\markoverwith{\textcolor{CMpurple}{\rule[.4ex]{2pt}{0.8pt}}}\ULon}

\definecolor{stanford}{rgb}{0.54,0.08,0.08}

\newcommand\mx[1]{\textcolor{orange}{[Michael: #1]}}

\definecolor{mypurple}{rgb}{0.8, 0.18, 1}
\newcommand\dor[1]{\textcolor{mypurple}{[Dor: #1]}}

\newcommand\pw[1]{\textcolor{CMpurple}{[PW: #1]}}
\newcommand\at[1]{\textcolor{atcolor}{[AT: #1]}}

\newcommand\FT[1]{\textcolor{blue}{[Florian: #1]}}

\definecolor{tb12}{rgb}{1.0, 0.13, 0.32}

\newcommand\dora[1]{\textcolor{olive}{[Dora: #1]}}

\definecolor{eamorange}{rgb}{.8,.33,0}

\newcommand\ilevent[1]{\textcolor{orange}{[Isabelle L.: #1]}}

\fi

\usepackage{authblk}

\makeatletter
\renewcommand\maketitle{
{\raggedright
\begin{center}
{\Huge \bfseries \sffamily \@title }\\[2ex]
{\@author}\\[2ex]
\end{center}}}
\makeatother

\renewenvironment{abstract}{%
    \newline
    \itshape
    }
{}

\begin{document}

\title{Artificial Intelligence for Science in Quantum, Atomistic, and Continuum Systems}

\renewcommand{\shorttitle}{Artificial Intelligence for Science in Quantum, Atomistic, and Continuum Systems}


\author[1,*]{Xuan Zhang}
\author[1,*]{Limei Wang}
\author[1,*]{Jacob Helwig}
\author[1,*]{Youzhi Luo}
\author[1,*]{Cong Fu}
\author[1,*]{Yaochen Xie}
\author[1]{Meng Liu}
\author[1]{Yuchao Lin}
\author[1]{Zhao Xu}
\author[1]{Keqiang Yan}
\author[2]{Keir Adams}
\author[3]{Maurice Weiler}
\author[1]{Xiner Li}
\author[4]{Tianfan Fu}
\author[5]{Yucheng Wang}
\author[7]{Alex Strasser}
\author[1]{Haiyang Yu}
\author[6]{YuQing Xie}
\author[6]{Xiang Fu}
\author[8]{Shenglong Xu}
\author[9,10]{Yi Liu}
\author[11]{Yuanqi Du}
\author[1]{Alexandra Saxton}
\author[1]{Hongyi Ling}
\author[6]{Hannah Lawrence}
\author[6]{Hannes Stärk}
\author[1]{Shurui Gui}
\author[4]{Carl Edwards}
\author[12]{Nicholas Gao}
\author[6]{Adriana Ladera}
\author[13]{Tailin Wu}
\author[6]{Elyssa F. Hofgard}
\author[6]{Aria Mansouri Tehrani}
\author[14]{Rui Wang}
\author[6]{Ameya Daigavane}
\author[1]{Montgomery Bohde}
\author[1]{Jerry Kurtin}
\author[13]{Qian Huang}
\author[6]{Tuong Phung}
\author[13]{Minkai Xu}
\author[15]{Chaitanya K. Joshi}
\author[15]{Simon V. Mathis}
\author[16]{Kamyar Azizzadenesheli}
\author[17]{Ada Fang}
\author[18,19]{Alán Aspuru-Guzik}
\author[3]{Erik Bekkers}
\author[20,21]{Michael Bronstein}
\author[22]{Marinka Zitnik}
\author[16,23]{Anima Anandkumar}
\author[13]{Stefano Ermon}
\author[15]{Pietro Liò}
\author[14]{Rose Yu}
\author[12]{Stephan Günnemann}
\author[13]{Jure Leskovec}
\author[4]{Heng Ji}
\author[4]{Jimeng Sun}
\author[6]{Regina Barzilay}
\author[6]{Tommi Jaakkola}
\author[2,6]{Connor W. Coley}
\author[1,5,24]{Xiaoning Qian}
\author[5,7,8]{Xiaofeng Qian}
\author[6]{Tess Smidt}
\author[1,+]{Shuiwang Ji}
\affil[1]{Department of Computer Science \& Engineering, Texas A\&M University, College Station, TX}
\affil[2]{Department of Chemical Engineering, Massachusetts Institute of Technology, Cambridge, MA}
\affil[3]{AMLab, University of Amsterdam, Amsterdam, Netherlands}
\affil[4]{Department of Computer Science, University of Illinois Urbana-Champaign, Urbana, IL}
\affil[5]{Department of Electrical \& Computer Engineering, Texas A\&M University, College Station, TX}
\affil[6]{Department of Electrical Engineering and Computer Science, Massachusetts Institute of Technology, Cambridge, MA}
\affil[7]{Department of Materials Science \& Engineering, Texas A\&M University, College Station, TX}
\affil[8]{Department of Physics \& Astronomy, Texas A\&M University, College Station, TX}
\affil[9]{Department of Applied Mathematics \& Statistics, Stony Brook University, Stony Brook, NY}
\affil[10]{Department of Computer Science, Stony Brook University, Stony Brook, NY}
\affil[11]{Department of Computer Science, Cornell University, Ithaca, NY}
\affil[12]{Department of Computer Science, Technical University of Munich, München, Germany}
\affil[13]{Department of Computer Science, Stanford University, Stanford, CA}
\affil[14]{Department of Computer Science \& Engineering, University of California San Diego, La Jolla, CA}
\affil[15]{Department of Computer Science  \& Technology, University of Cambridge, Cambridge, UK}
\affil[16]{Nvidia, Santa Clara, CA}
\affil[17]{Department of Chemistry and Chemical Biology, Harvard University, Cambridge, MA}
\affil[18]{Department of Chemistry, University of Toronto, Toronto, Canada}
\affil[19]{Department of Computer Science, University of Toronto, Toronto, Canada}
\affil[20]{Department of Computer Science, University of Oxford, Oxford, UK}
\affil[21]{AITHYRA, Vienna, Austria}

\affil[22]{Department of Biomedical Informatics, Harvard University, Boston, MA}
\affil[23]{Department of Computing \& Mathematical Sciences, California Institute of Technology, Pasadena, CA}
\affil[24]{Computing and Data Sciences, Brookhaven National Laboratory, Upton, NY}


\affil[*]{Equal contribution}
\affil[+]{Corresponding author: Shuiwang Ji (\href{mailto:sji@tamu.edu}{sji@tamu.edu})}


\renewcommand{\shortauthors}{}

\maketitle
\noindent 
\vspace{-0.2in}



\leavevmode
\begin{abstract}
Advances in artificial intelligence (AI) are fueling a new paradigm of discoveries in natural sciences.
Today, AI
has started to advance natural
sciences by improving, accelerating, and enabling our understanding of natural phenomena at a
wide range of spatial and temporal scales, giving rise to a new area of research known as AI for
science (AI4Science). Being an emerging research paradigm, AI4Science is unique in that it is an enormous and highly interdisciplinary area. Thus, a unified and technical treatment of this field is needed yet challenging. 
This work aims to provide a technically thorough account of a subarea of AI4Science; namely, AI for quantum, atomistic, and continuum systems. These areas aim at understanding the physical world from the subatomic (wavefunctions and electron density), atomic (molecules, proteins, materials, and interactions), to macro (fluids, climate, and subsurface) scales and form an important subarea of AI4Science.
A unique advantage of focusing on these areas is that they largely share a common set of challenges, thereby allowing a unified and foundational treatment. A key common challenge is how to capture physics first principles, especially symmetries, in natural systems by deep learning methods. We provide an in-depth yet intuitive account of techniques to achieve equivariance to symmetry transformations. We also discuss other common technical challenges, including explainability, out-of-distribution generalization, knowledge transfer with foundation and large language models, and uncertainty quantification. To facilitate learning and education, we provide categorized lists of resources that we found to be useful. We strive to be thorough and unified and hope this initial effort may trigger more community interests and efforts to further advance AI4Science.
\end{abstract}

\clearpage
\tableofcontents

\clearpage

\hypertarget{introduction}{\section{Introduction}}
\label{sec:introduction}

Decades of artificial intelligence (AI) research has culminated in the renaissance of neural networks~\cite{lecun1998gradient} under the name of deep learning. Since AlexNet~\cite{AlexNet}, a decade of intensive research has led to many breakthroughs in deep learning, including, for example, ResNet~\cite{he2016deep}, diffusion and score-based models~\cite{ho2020denoising,song2020score}, attention, transformers~\cite{vaswani2017attention}, and recently large language models (LLM) and ChatGPT~\cite{openai2023gpt4}, \emph{etc.} These developments have led to
continuously improved performance for deep models. When coupled with growing computing  power and large-scale datasets, deep learning methods are becoming dominant approaches in various fields, such as computer vision and natural language processing.
Propelled by these advances, AI has started to advance natural sciences by improving, accelerating, and enabling our understanding of natural phenomena at a wide range of spatial and temporal scales,
giving rise to a new area of research, known as AI for science. It is our belief that AI for science opens a door for a new paradigm of scientific discovery and represents one of the most exciting areas of interdisciplinary research and innovation.

Historically, the importance of computing in accelerating discoveries in natural sciences has been noted. Almost one hundred years ago in 1929,
the quantum physicist Paul Dirac stated that \emph{``The underlying physical laws necessary for the
mathematical theory of a large part of physics and the whole of
chemistry are thus completely known, and the difficulty is only that
the exact application of these laws leads to equations much too
complicated to be soluble.''}~\citep{dirac1929quantum}
In quantum physics, it is known that the Schr{\"o}dinger’s equation provides precise descriptions of behaviors of quantum systems, but solving such an equation is only possible for very small systems due to its exponential complexity. In fluid mechanics, the Navier-Stokes equations describe spatiotemporal dynamics of fluid flows, but solving these equations of practically useful sizes is highly demanding, especially when computing efficiency is also required. Similar to these two examples, the underlying physics of many natural science problems are known and can be described by a set of 
mathematical equations. The key difficulty lies in how to solve these equations accurately and efficiently. Recent studies have shown that deep learning methods can accelerate the computing of solutions for these equations.
For example, deep learning methods have been used to compute the solutions of Schr{\"o}dinger’s equation in quantum physics~\cite{carleo2017solving,pfau2020ab,hermann2020deep,hermann2022ab} and Navier-Stokes equations in fluid mechanics~\cite{kochkov2021machine,brunton2020machine}. In these areas, simulators are employed to compute solutions of mathematical equations, and the results are used as data to train deep learning models. Once trained, these models can make predictions at a speed that is much faster than simulators. In addition to improved efficiency, deep learning models have been shown to exhibit better out-of-distribution (OOD) generalization, with scope extended to much wider practical settings, where training and unseen data usually follow different distributions.

In other areas such as biology, the underlying biophysical process is not completely understood and may not ultimately be described by mathematical equations. In these cases, experimentally generated data can be used to train deep learning models in order to model the underlying biophysical process.
For example, in biology, AI systems, such as AlphaFold~\cite{jumper2021highly}, RoseTTAFold~\cite{baek2021accurate}, and ESMFold~\cite{lin2023evolutionary}, trained on experimentally acquired 3D structures, enable the computational prediction of protein 3D structures at an accuracy comparable to experimental results. In addition to technical challenges, a key element in these areas is the availability of large amounts of experimentally generated data. For example, the success of AlphaFold, RoseTTAFold, and ESMFold highly relies on the large amount of protein 3D structure data generated using experiments and deposited into databases, such as the Protein Data Bank. 

\begin{figure}[t]
    \centering
    \includegraphics[width=\textwidth]{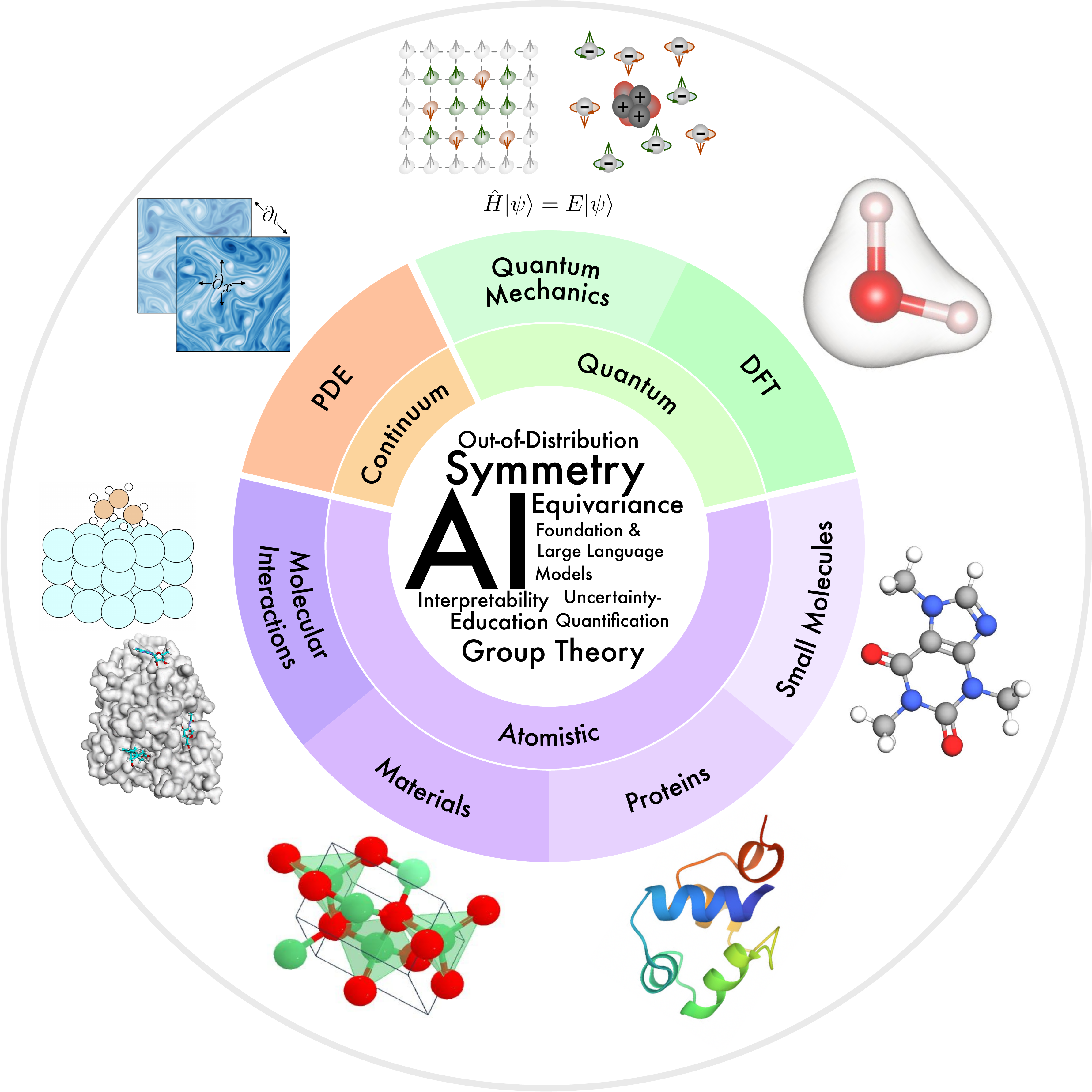}
    \caption{\ifshowname\textcolor{red}{Allie, Meng}\else\fi An integrative overview of the selected research areas in AI for science. As described in Section~\ref{subsec:sci_areas}, we focus on AI for \emph{quantum mechanics}, \emph{DFT}, \emph{small molecules}, \emph{proteins}, \emph{materials}, \emph{molecular interactions}, and \emph{PDE}. We visually depict these diverse areas in the outermost circle. These areas are arranged by their respective spatial and temporal scales of physical world modeling, highlighting \emph{quantum}, \emph{atomistic}, and \emph{continuum} systems. Notably, as summarized in Section~\ref{subsec:tech_areas}, a set of common technical considerations and challenges, such as \emph{symmetry}, \emph{interpretability}, and \emph{out-of-distribution generalization}, exist across these multiple AI for science research areas. We show these technical areas in the innermost circle.}\label{fig:AIRS_Overview}
\end{figure}

\subsection{Scientific Areas}
\label{subsec:sci_areas}

In this work, we provide a technical and unified review of several research areas in AI for science that researchers have been working on during the past several years. We organize different areas of AI for science by the spatial and temporal scales at which the physical world is modeled. An overview of scientific areas we focus in this work is given in Figure~\ref{fig:AIRS_Overview}.

\vspace{0.1cm}\noindent \textbf{Quantum Mechanics} studies physical phenomena at the smallest length scales using wavefunctions, which describe the complete dynamics of quantum systems. In quantum physics, wavefunctions are obtained by solving the Schr{\"o}dinger equation, which incurs exponential complexity. In this work, we provide technical reviews on how to design
advanced deep learning methods for learning neural wavefunctions efficiently.
For a comprehensive review of machine learning in quantum science, one may refer to~\citep{dawid2022modern}.


\vspace{0.1cm}\noindent \textbf{Density Functional Theory (DFT)} and \textit{ab initio} quantum chemistry approaches are first-principles methods widely used in practice to calculate electronic structures and physical properties of molecules and materials. However, these methods are still computationally expensive, limiting their use in small systems ($\sim$1,000 atoms). In this work, we present technical reviews on deep learning methods for accurately predicting quantum tensors, which in turn can be used to derive many other physical and chemical properties, including, electronic, mechanical, optical, magnetic, and catalytic properties of molecules and solids. We also touch on machine learning methods for density functional learning.

\vspace{0.1cm}\noindent \textbf{Small Molecules}, also known as micromolecules, typically have tens to hundreds of atoms and play important regulatory and signaling roles in many chemical and biological processes. For example, 90\% of approved drugs are small molecules, which can interact with target macromolecules (like proteins), altering the activity or function of the target. In recent years, significant progress has been made in using machine learning methods to accelerate scientific discoveries on small molecules at the atomistic level. In this work, we present in-depth technical reviews on small molecule representation learning, molecular generation, simulation, and dynamics.

\vspace{0.1cm}\noindent \textbf{Proteins} are macromolecules that consist of one or more chains of amino acids. It is commonly believed that amino acid sequences determine protein structures, which in turn determines their functions. Proteins perform most of the biological functions, which include structural, catalytic, reproductive, metabolic, and transporting roles, \emph{etc.} Recently, machine learning approaches have led to dramatic advances in protein structure prediction~\cite{jumper2021highly,baek2021accurate,lin2023evolutionary}.
In this work, we provide technical reviews on how to learn representations from protein 3D structures, and how to generate and design novel proteins.

\vspace{0.1cm}\noindent \textbf{Materials Science} studies the relationship of processing, structure, properties, and performance of materials. The intrinsic structure of materials from atomistic, to micro and continuum scale determine their quantum, electronic, catalytic, mechanical, optical, magnetic, and other properties through interplay with external stimuli/environment. Recently, machine learning methods have been developed to predict the properties of crystal materials and design novel crystal structures. In this work, we provide technical reviews on the property prediction and structure generation of crystal materials.

\vspace{0.1cm}\noindent \textbf{Molecular Interactions} study how molecules interact with each other to carry out many of the physical and biological functions. Recent advances in machine learning have spurred the renaissance in modeling various molecular interactions, such as ligand-receptor and molecule-material interactions.
In this work, we present in-depth and comprehensive reviews on such advances.

\vspace{0.1cm}\noindent \textbf{Continuum Mechanics} models physical processes that evolve in time and space at the macroscopic level using partial differential equations (PDEs), including
fluid flows, heat transfer, and electromagnetic waves, \emph{etc.} However, solving PDEs using classic solvers suffers from several limitations, including low efficiency, difficulties in out-of-distribution generalization and multi-resolution analysis. In this work, we provide reviews on recent deep learning methods for surrogate modeling that addresses these limitations.

\begin{figure}[htbp]
    \centering
    \includegraphics[width=.982\textwidth]{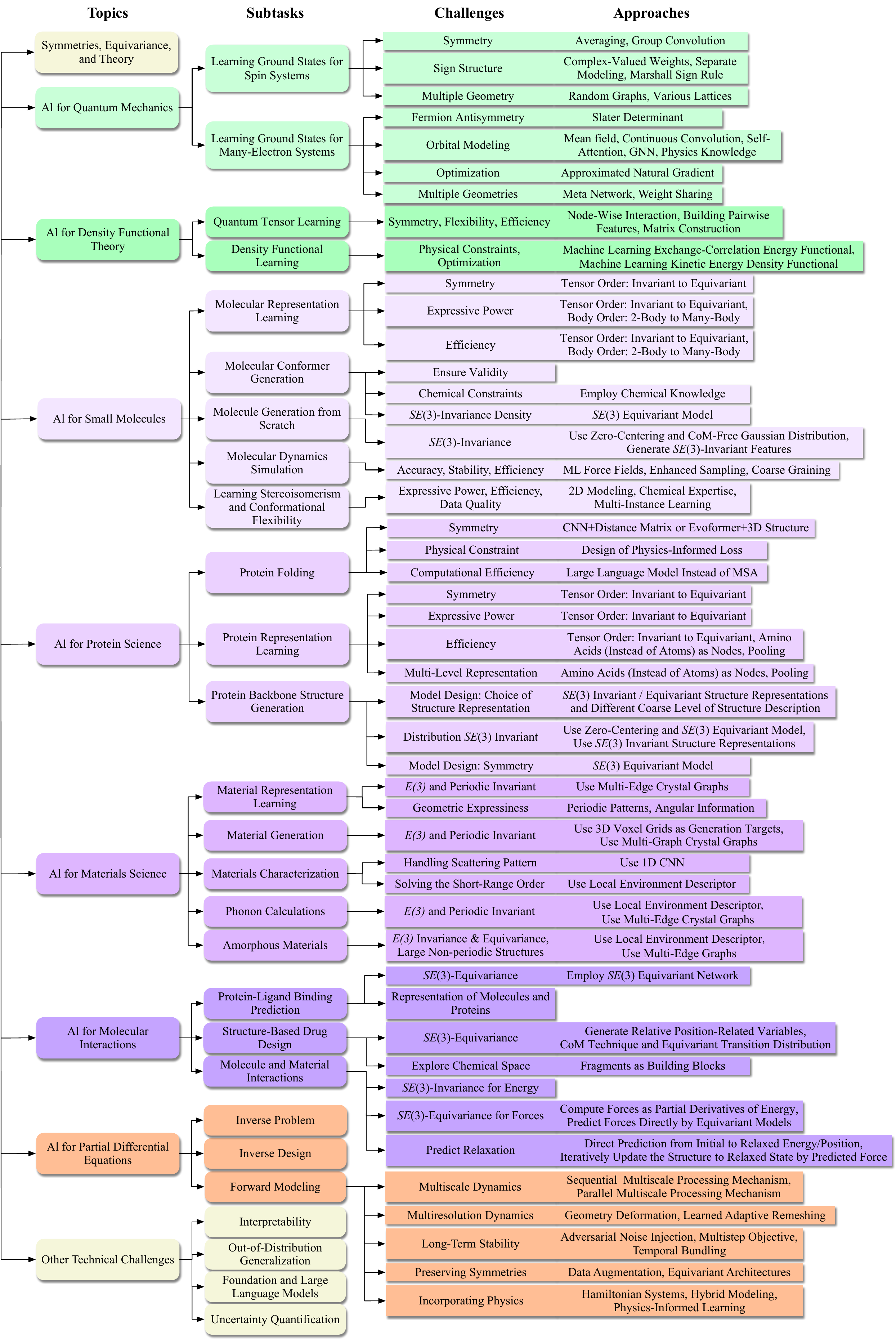}\vspace{-0.3cm}
    \caption{{\ifshowname\textcolor{red}{Xiner}\else\fi} The overall taxonomic structure of this work.
    We outline the areas of AI for science included in this work and summarize selected problems, central challenges, and the major approaches.
    }\label{fig:overall}
\end{figure}

In each area, we provide a precise problem setup and discuss the key challenges of using AI to solve such problems.
We then provide a survey of major approaches that have been developed. 
We also describe datasets and benchmarks that have been used to evaluate machine learning methods. Finally, we summarize the remaining challenges and propose several future directions in each research area.
When applicable, we include the recommended prerequisite sections at the beginning of each subsection to indicate inter-section dependencies. The overall taxonomic structure is summarized as Figure~\ref{fig:overall}. This work presents a comprehensive taxonomy, anchored by the shared mathematical and physical principles of symmetry, equivariance, and group theory, delving into seven specific domains within the realm of AI for science, and discussing common technical challenges existing in multiple areas. This enables a comprehensive and structured exploration of AI for science. 

\subsection{Technical Areas of AI}
\label{subsec:tech_areas}

We have observed that a set of common technical challenges exist in multiple areas of AI for science.

\vspace{0.1cm}\noindent \textbf{Symmetry:} A common and recurring observation from many scientific problems is that objects or systems of interests usually
contain geometric structures. In many cases, these geometric structures imply certain symmetries that the underlying physics obeys. For example, in molecular dynamics, molecules are represented as graphs in 3D space, and translating or rotating a molecule may not change its properties.
Then the symmetry here is named translational or rotational invariance.
Formally, a symmetry is defined as a transformation that, when applied on an object of interest, leaves certain properties of the object unchanged (invariant) or changed in a deterministic way (equivariant)~\cite{bronstein2021geometric}. 
Symmetries are very strong inductive biases, as
P. Anderson (1972) stated that ``\emph{It is only slightly overstating the case to say that physics is the study of symmetry.}'' ~\cite{Anderson1972MoreIsDifferent}. Thus, a key challenge of AI for science is how to effectively integrate symmetries in AI models. We use symmetry as the main common thread to connect many of the topics in this work. The required symmetries for each area are also summarized in Figure~\ref{fig:scale}.



\vspace{0.1cm}\noindent \textbf{Interpretability:} Science aims at understanding the governing rules of the physical worlds. Thus, the aims of AI for science are to (1) design models capable of modeling the physical world accurately, and (2) interpret models to verify or discover the governing physics~\cite{han2020integrating}. Thus, interpretability is essential in AI for science.

\vspace{0.1cm}\noindent \textbf{Out-of-Distribution (OOD) Generalization and Causality:} Traditional machine learning methods assume training and test data follow the same distribution. In reality, different distribution shifts may exist between training and test data, raising the need to identify causal factors capable of OOD generalization. OOD generalization is particularly relevant in scientific simulations as this avoids the need to generate training data for every different settings.

\vspace{0.1cm}\noindent \textbf{Foundation and Large Language Models:} When labeled training data are not readily available, the capability to perform unsupervised or few-shot learning becomes important. Recently, foundation models~\cite{Bommasani2021FoundationModels} have demonstrated promising performance on natural language processing tasks.
Typically, foundation models are large-scale models pre-trained
under self-supervision or generalizable supervision, allowing a wide range of downstream tasks to
be performed in few-shot or zero-shot manners. This paradigm is becoming increasingly popular due to the recent developments of large language models (LLM) such as GPT-4. We provide our perspectives on how such a paradigm could accelerate discoveries in AI for science.

\vspace{0.1cm}\noindent \textbf{Uncertainty Quantification (UQ)} studies how to guarantee robust decision-making under data and model uncertainty, and is a critical part of AI for science. UQ has been studied in various disciplines of applied mathematics,
computational and information sciences, including scientific computation, statistic modeling, and
more recently machine learning. We provide an up-to-date reviews of UQ in the context of scientific discoveries.

\vspace{0.1cm}\noindent \textbf{Education:} AI for science is an emerging and rapidly developing area of research with many useful resources developed physically or online. To facilitate learning and education, we have compiled categorized lists of resources that we find to be useful. We also provide our perspectives on how the community can do better to facilitate the integration of AI with science and education.

\subsection{Integrative Multi-Scale Analysis}

\begin{figure}[t]
    \centering
    \includegraphics[width=\textwidth]{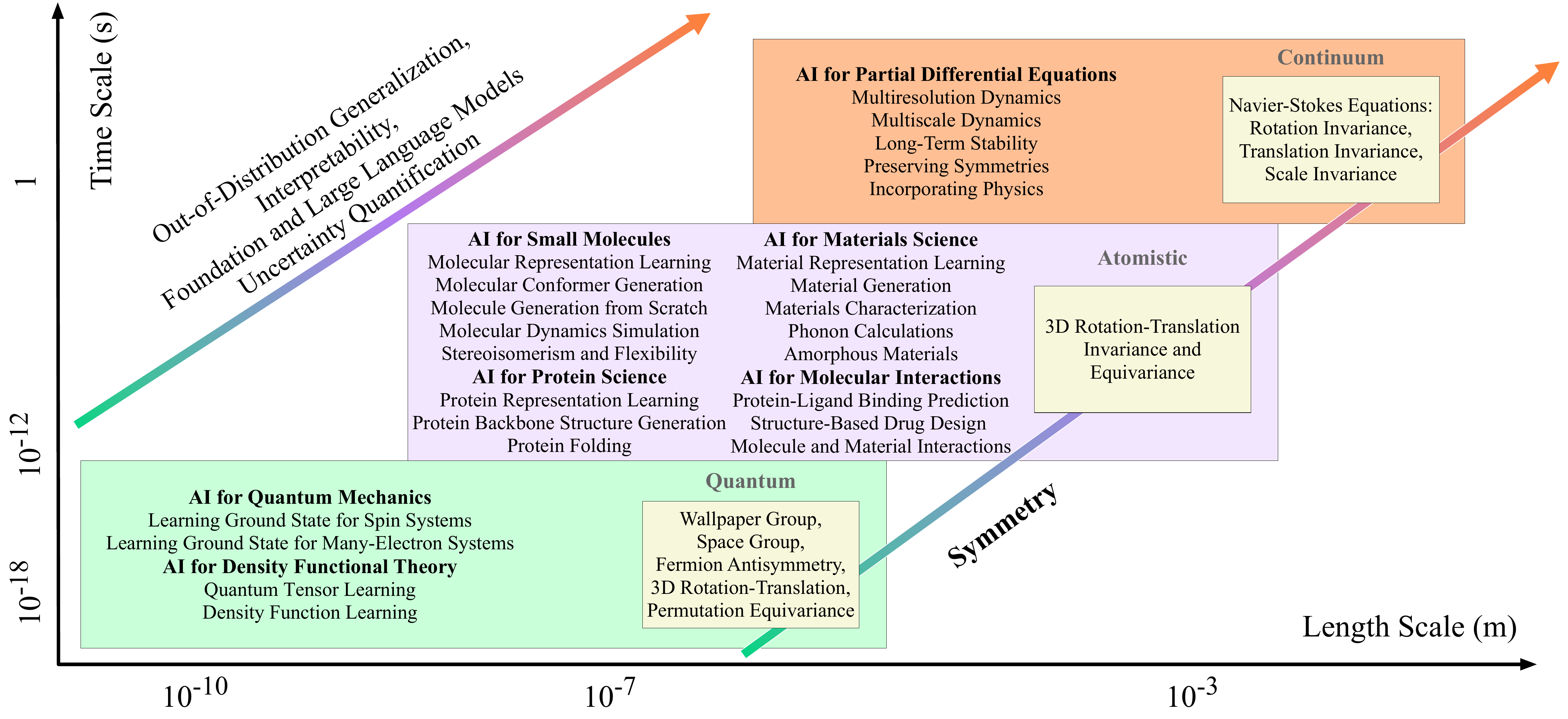}
    \caption{{\ifshowname\textcolor{red}{Xiner}\else\fi} Spatial and temporal scale of scientific areas. We explore the intersection of AI and various scientific disciplines within a continuum of spatial and temporal scales. This framework accommodates a diverse range of areas and problems, unified by their distinct symmetries and shared technical challenges. Symmetries, inherent to the structure of natural science and governed by mathematical and physical laws, manifest in numerous patterns across various scientific fields. This cross-disciplinary perspective provides a fresh lens through which we can address and investigate complex scientific problems with AI methods.}\label{fig:scale}
\end{figure}

{\ifshowname\textcolor{red}{Revised by Xiaofeng}\else\fi} In this survey, we conduct analysis at different levels, including quantum physics, density functional theory (DFT), molecular dynamics (MD), and continuum dynamics. There are notable differences in terms of the level of approximations and the scales they are dealing with. 
Specifically, quantum physics deals with the behavior and interactions of particles such as electrons, protons, and neutrons, as well as their quantum mechanical properties by solving the Schr{\"o}dinger’s equation for many-body interacting system. The spatial scale in quantum physics is typically on the order of the atomic and subatomic level, ranging from picometer ($10^{-12}$ meters) to nanometer ($10^{-9}$ meters) scale, depending on specific problems.
DFT solves the Schr{\"o}dinger’s equation for electrons and ions using an alternative approach by mapping many-body interacting system to many-body non-interacting system, which therefore allows to provide insights into the electronic structure of realistic materials such as atoms, molecules, and solids ranging from angstroms ($10^{-10}$ meters) to hundreds of angstroms. MD simulations operate at a larger scale, typically ranging from the nanometer ($10^{-9}$ meter) to micrometer ($10^{-6}$ meter) scale using empirical/semi-empirical force fields as well as the rising machine learning force fields. MD focuses on the motion and interactions of atoms and molecules over time under various thermodynamic ensembles, allowing for the investigation of dynamic behavior, structural changes, kinetic, and thermodynamic properties. 
In comparison, quantum physics aims to solve many-body wavefunctions and Hamiltonian for many-body interacting system; DFT takes an alternative approach with practical applications for molecules and materials; MD simulations operate at a much larger spatial scale and longer time scale without explicitly dealing with spatial and spinor components of electronic wavefunctions. 
To address even larger scales and eliminate the discrete characteristics of particles, partial differential equations (PDE) are used to study the continuum system behaviors in scales ranging from micrometers ($10^{-6}$ meter, such as the Kolmogorov microscale) in fluid dynamics to kilometers ($10^{3}$ meters) in climate dynamics.
We compare the spatial and temporal scales of different systems
in Figure~\ref{fig:scale}. Accordingly, the focus areas in this work are clustered into quantum, atomistic, and continuum systems. 
The choice of the theoretical levels depends on the phenomena of interest and the computational complexity required for the study. Different analyses can benefit each other and lead to integrative analysis. 


\subsection{Online Resources}

AI for science is an emerging and rapidly developing area of research. To enable continuous updates of this work, we have created an online portal (\url{https://air4.science/}), which will be maintained and updated regularly. The online portal contains our assets including a mindmap, which is designed to visualize the taxonomic structure of the various areas covered in our work. This mindmap serves as a comprehensive overview allowing users to navigate and will be updated regularly after the publication of this work to include new topics and significant advancements in the field. In addition, we include a feedback form (\url{https://air4.science/feedback}) on the portal. This form serves as a channel for individuals to contribute their thoughts, suggestions, and comments regarding this work. We highly value input from the wider community to improve our work.

This work is accompanied by a software library and benchmarks under the project repository ``AIRS: AI Research for Science'' (\url{https://github.com/divelab/AIRS/}), that we have developed as part of our scientific pursuits in these areas. A set of software libraries have been included and will be added continuously as our research progresses. We also maintain a curated list of literature and resources pertaining to each AI for science topics in the project repository. We welcome contributions from the wider community to both the library and literature via pull requests.

\subsection{Scope and Feedback}

AI research for science is an enormous and emerging field, and our focus in this work is on AI for quantum, atomistic, and continuum systems. Thus, our work is by no means comprehensive and only includes selected areas of AI for science related to physics, chemistry, biology, material science, molecular simulation and dynamics, and partial differential equations, \emph{etc.} Given the evolving nature of this area, our work is by no means conclusive in any sense. We expect to continuously include more methods and benchmarks as the area develops. AI for science is highly interdisciplinary, and there is no doubt that we have missed relevant work in the literature, for which we must apologize. We welcome any feedback and comments from the community to improve our work. Readers are encouraged to submit their feedback to us via the above online portal.

\subsection{Contributions and Authorship}

This work was initiated and conceptualized by Shuiwang Ji, who also leads the distributed writing process and provides scientific and administrative support throughout the project.
Each of the individual
sections was written by a subset of authors, and authorship is given in each section. Given that all these sections are related, there have been extensive discussions across sections.
Authorship is based on the amounts of direct contributions to each section, including texts, equations, figures, tables, discussions, and feedback, \emph{etc.} Contributions are approximately quantified based on the number of pages to which each author contributes in the final work, slightly adjusted based on levels of difficulties and thus discussions required. Many authors have provided constructive discussions and feedback, which have also been considered.
When multiple authors work on a part collaboratively, percentage of contributions from each author is estimated and used in the calculation.
Authorship for the entire work is determined based on the cumulative contributions made to all sections.
All authors have made significant contributions to this work, and their orders should be interpreted only in an approximate sense.

\begin{table}[ht]
    \caption{Summary of key notations. Notations used in a single area are individually defined in the table and in each section.}\label{tab:sum_notations}
    \centering
    \begin{tabularx}{\textwidth}{l|X}\toprule
        Sections & Key notations \\\midrule
        Sec.~\ref{sec:group} & Input signal $X\in\mathbb{R}^{s\times s}$, convolution kernel $W\in\mathbb{R}^{k\times k}$, convolution operator $\ast$. Spherical harmonics functions $\bm{Y}^\ell(\cdot):\mathbb{R}^3\to\mathbb{R}^{2\ell+1}$, node feature $\bm{h}^{\ell_1}_i\in\mathbb{R}^{2\ell_1+1}$, message $\bm{m}^{\ell_3}_i\in\mathbb{R}^{2\ell_3+1}$, CG matrix $\revisionTwo{\mathscr{C}}^{\ell_3}_{\ell_1,\ell_2}\in\mathbb{R}^{(2\ell_3+1)\times (2\ell_1+1)(2\ell_2+1)}$, Widger-D matrix $D^\ell(R)\in\mathbb{R}^{(2\ell+1)\times(2\ell+1)}$.\\\hline
        Sec.~\ref{sec:qt} &  Wavefunction $\psi$ or $\ket{\psi}$, a spin configuration $\ket{\bm{\sigma}^{(i)}}$, number of spins $N$, number of electrons of a certain spin $N^\uparrow$, $N^\downarrow$. Electron coordinates $\bm{r}=[\bm{r}_1,\dots, \bm{r}_{N^\uparrow+N^\downarrow}]$. Set of possible molecules $\mathbb{M}=\{M=\{\bm{c}_i,z_i\}_{i=1}^{\vert M\vert}, \mathbf{\bm{c}_i}\in \mathbb{R}^3, z_i\in\mathbb{Z}\}$. Electron orbital network $\bm\phi^\uparrow_\theta$, $\bm\phi^\downarrow_\theta$, determinants: $\det\begin{bmatrix}...\end{bmatrix}$, local energy $E_{loc}$, Hamiltonian matrix for spin systems $H\in \mathbb{C}^{2^N \times 2^N}$, Hamiltonian operator $\hat{H}$, potential energy $V$. \vspace{5pt}\\\hline
        Sec.~\ref{sec:dft} &  Wavefunction $\psi$ or $\ket{\psi}$, number of orbitals $N_o$, $\bm{r}_i$ electron position, electronic wavefunction coefficients matrix $\bm{C}_e \in \mathbb{R}^{N_o \times N_o}$ or $\mathbb{C}^{N_o \times N_o}$ (depending on the nature of physical systems),  Hamiltonian matrix $\bm{H} := \bm{H}_{\bf{DFT}} \in \mathbb{R}^{N_o \times N_o}$ or $\mathbb{C}^{N_o \times N_o}$ (depending on the nature of physical systems), overlap matrix $\bm{S}\in \mathbb{R}^{N_o \times N_o}$, eigen energy diagonal matrix $\bm{\epsilon} \in \mathbb{R}^{N_o \times N_o}$, electron density $\rho$, energy $E[\rho]$ which is a function of electron density, and external potential $V_{ext}(\bm{r})$.
        \vspace{5pt}\\\hline
        Sec.~\ref{sec:mol} & 3D molecule $\mathcal{M}=(\bm{z},C)$, where $\bm{z}$ denotes atom types and $C$ represents coordinates. Distance $d_{ij}$ between two atoms  $i, j$. Scalar feature $\bm{s}\in \mathbb{R} ^ {d}$, vector feature $\bm{v} \in \mathbb{R}^{d\times 3}$, order-$\ell$ feature $\bm{h}_{icm}^{\ell}$, for node $i$, channel $c$, and representation index $-\ell \le m \le \ell$.

        2D molecule (for conformer generation) $\mathcal{G}=(\bm{z}, E)$, where $\bm{z}$ denotes atom types and $e_{ij}\in\mathbb{Z}$ denotes the edge type between node $i$ and $j$. Generative model $f_G$, predictive model $f_P$, equilibrium ground-state geometry $C_{eq}$. 
        \vspace{5pt}\\\hline
        Sec.~\ref{sec:prot} & Alpha-carbon $C_{\alpha}$, coordinate matrix $\mathcal{C}$, protein backbone structure $\mathcal{P}_{\text{base}} = (\bm{z},\mathcal{C}^{C_{\alpha}})$ or $\mathcal{P}_{\text{bb}} = (\bm{z},\mathcal{C}^{C_{\alpha}}, \mathcal{C}^{N}, \mathcal{C}^{C})$, where $a_i \in \{k | 1 \le k \le 20, k \in \mathbb{Z}\}$ denotes the type of the $i$-th amino acid and $\mathcal{C}^{C_{\alpha}}, \mathcal{C}^{N}, \mathcal{C}^{C} \in\mathbb{R}^{3\times n}$ are backbone atom coordinates.
        \vspace{5pt}\\\hline
        Sec.~\ref{sec:mat} & Material $\mathcal{M}=(\bm{z}, C, L)$ with lattice matrix $L=[\bm{\ell}_1,\bm{\ell}_2,\bm{\ell}_3]\in\mathbb{R}^{3\times 3}$, property prediction function $f: M\mapsto y$, material distribution $p$, periodic transformation $C'=C+LK$. \vspace{5pt}\\\hline
        Sec.~\ref{sec:dock} & 
        Molecule $\mathcal{M}=(A, E, C)$, where $A$ refers to the atomic properties, $E$ denotes edge features, $C$ denotes coordinates, and protein $\mathcal{P}=(B, S)$, where $B$ refers to node types, $S$ denotes coordinates, and binding pose prediction function $f\textsubscript{pose}:(B, S, A, E)\mapsto  [C_1,\dots, C_k]$, binding strength prediction function $f\textsubscript{strength}:(A, E, C, B, S)\mapsto  q$.
        
        Molecule-material pair $\mathcal{S}=(\bm{z}, C)$ as an integrated system. Energy prediction function $f_{E}: \mathcal{S}\mapsto e$, force prediction function $f_{F}: \mathcal{S}\mapsto F$, relaxed energy prediction function $f_{RE}: \mathcal{S}_{init}\mapsto e_{rel}$, relaxed structure prediction function $f_{RS}: \mathcal{S}_{init}\mapsto C_{rel}$. \\ \hline
        Sec.~\ref{sec:pde} & Function $u: U\to\mathbb R^m$ of space and time to be solved, partial derivative with respect to space $\partial_x$ and time $\partial_t$, differential operators $\mathcal B$ and $\mathcal D$, spatial domain $\mathbb X$ and its boundary $\partial\mathbb X$, temporal domain $\mathbb T$. Group action of group $G$ on function $f$ is denoted by $L_gf(x):=f(g^{-1}x)$.
        \\\bottomrule
    \end{tabularx}
\end{table}

\subsection{Notations}

We adopt standard mathematical notation in this work. Scalars are denoted by lowercase letters, such as $a$, while boldface lowercase letters, such as $\bm{a}$, are used to denote vectors. Matrices are denoted by uppercase letters, such as $A$, with their $ij$-th entry denoted as $a_{ij}$ and their $k$-th column denoted as $\bm{a}_k$. Tuples or sets are denoted by calligraphic uppercase letters, such as $\mathcal{A}$. The rules hold for all notations except for those with special meanings, in which case we use their conventional forms. For example, the Hamiltonian matrix is denoted by $\bm{H}$, the coefficient matrix in DFT by $\bm{C}$, and energy scalars by $E$ and $V$. We provide a summary of common notations shared by multiple sections followed by key notations for individual directions.

\vspace{0.1cm}\noindent\textbf{Notation of Particle Systems:} We denote an $n$-body particles system, such as a molecule, material, and a protein, by a tuple of matrices $\mathcal{M}=(A, C)$, where $A$ denotes the particle attributes and $C=[\bm{c}_1,...,\bm{c}_n]\in\mathbb{R}^{3\times n}$ represents the Cartesian coordinates of particles in the system. Specifically, when only particle types are used as the attributes, we denote the system by $\mathcal{M}=(\bm{z}, C)$, where $\bm z\in\mathbb{Z}^n$ is a vector representing the types, such as atom charges. Additional attributes of a system can be included in the tuple, such as a material $\mathcal{M}=(\bm{z}, C, L)$ with a lattice matrix $L$. 

\vspace{0.1cm}\noindent\textbf{Notation of Transformations:} We denote the \revisionOne{planar} rotation transformation \revisionOne{of a 2D scalar field} by $R_\alpha: \mathbb{R}^{n\times n}\to\mathbb{R}^{n\times n}$ with an angle $\alpha$.
\revisionOne{A 3D rotation}  can be represented by a rotation matrix $R\in\mathbb{R}^{3\times 3}$, \revisionOne{and} the corresponding order-$\ell$ Wigner-D matrix is denoted by $D^\ell(R)$. We represent the translation transformation by a vector $\bm t\in\mathbb{R}^3$. Consequently, an $E(3)$-transformation on $C$ is denoted as $RC+\bm{t}\bm{1}^T$.

\ifshowname
\textcolor{red}{Shenglong}\else\fi

\vspace{0.1cm}\noindent\textbf{Dirac Notation:} 
Dirac notation, named after Paul Dirac, is commonly used in quantum physics to represent quantum states. In this notation, a quantum state is denoted by a ket vector, written as $\ket{\psi}$, a column vector in a complex vector space. The conjugate transpose of a ket vector is represented by a bra vector, written as $\bra{\psi}$, which is a row vector. The inner product between a bra and a ket is denoted as $\braket{\phi|\psi}$, yielding a complex number. The outer product of a ket and a bra is represented as $\ket{\psi}\bra{\phi}$, resulting in a complex matrix. Operators can be applied to quantum states by writing them to the left of the ket vector, such as $\hat{O}\ket{\psi}$, representing a matrix-vector multiplication. 

\vspace{0.1cm}\noindent\textbf{Key Notations in Individual Sections:} Other notations are defined individually for each area. We summarize the key notations in each direction in Table~\ref{tab:sum_notations}.














\clearpage
\hypertarget{Symmetries, Equivariance, and Theory}{\section{Symmetries, Equivariance, and Theory}} \label{sec:group}

\begin{figure}[t]
    \centering
    \includegraphics[width=\textwidth]{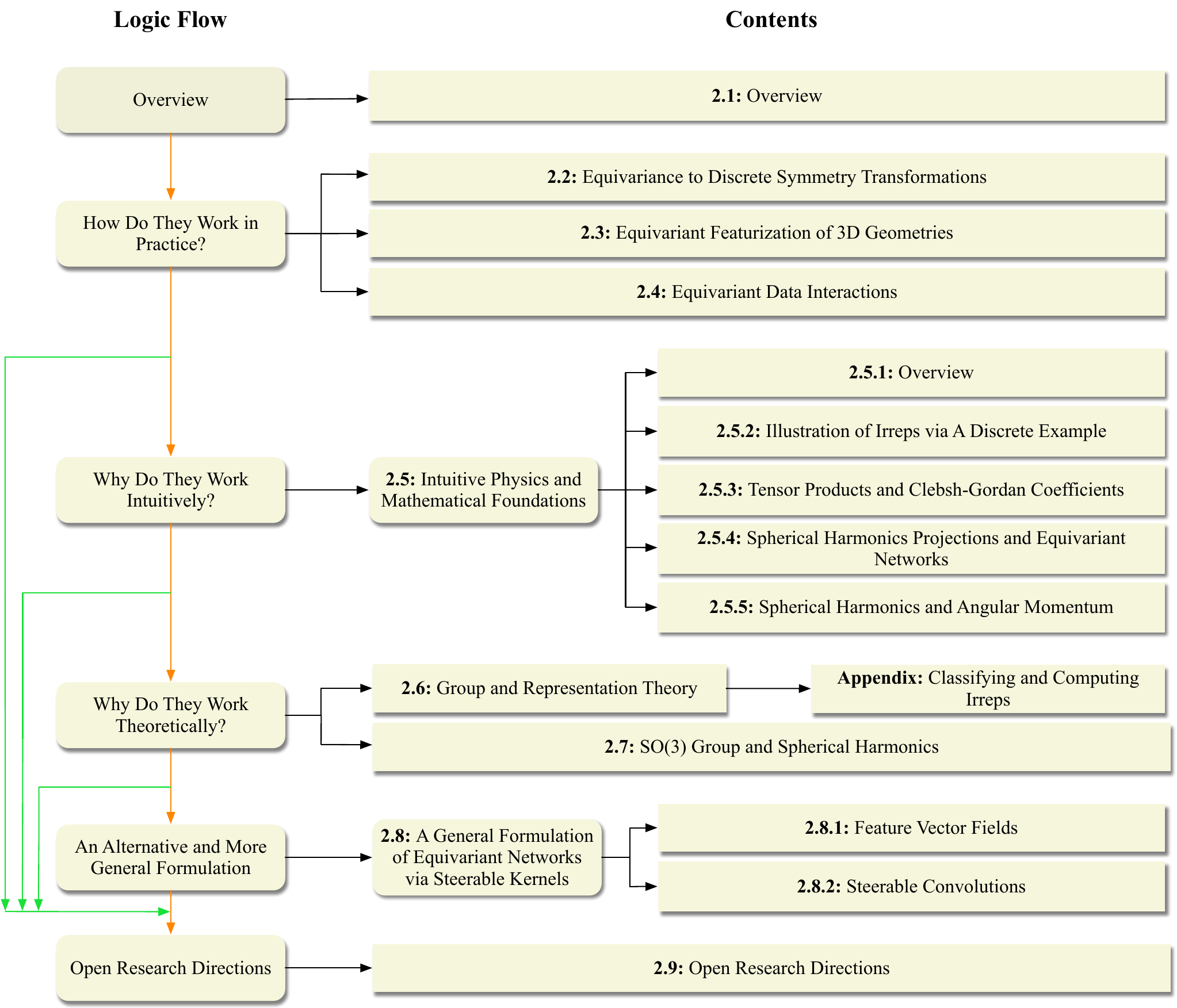}
    \caption{{\ifshowname\textcolor{red}{Allie, Yi}\else\fi}
    The overall logic flow and the associated subsections for \cref{sec:group}. Note the arrows in the logic flow show the dependencies among different subsections, and especially, the green skip connections indicate that some subsections can be skipped.   
    The black arrows show the associations between the logic flow and the subsections, as well as the relationships between each subsection and its child sections.
    Note the purpose of this figure is for readers to quickly navigate to certain contents based on background and interest, \emph{e.g.}, certain readers may skip the subsections associated with ``Why Do They Work Theoretically?''.}
    \label{fig:sec2_logic}
\end{figure}

\ifshowname\textcolor{red}{(Youzhi)}\else\fi
In many scientific problems, the objects of interest normally reside in 3D physical space. Any mathematical representation of these objects invariably relies on a reference coordinate frame, making representations coordinate-dependent. However, nature does not have a coordinate system, and so coordinate-independent representations are desired. Thus, one of the key challenges of AI for science is how to achieve invariance or equivariance. In this section, we provide a detailed review of the mathematical and physical foundations for achieving equivariance.
\ifshowname\textcolor{red}{(Yi)}\else\fi To make the content friendly to readers, we organize this section by a progressive increase in complication, with the logic flow shown in \cref{fig:sec2_logic}.
First, in \cref{subsec:dis_equi}, \cref{subsec:cont_feat}, and \cref{subsec:cont_equi}, we provide motivating examples for equivariance to discrete and continuous symmetry transformations, and describe how the tensor product is used in practice. 
After that, in \cref{subsec:phys}, through concrete and intuitive examples, we try to elucidate the physical and mathematical foundations for the underlying theory, such as symmetry groups, irreducible representations, tensor products, spherical harmonics, \emph{etc.}
Then in \cref{sec:group_theory_overview} and \cref{sec:spherical_harmonics}, we further lay out the detailed and formal theory, which can be skipped for certain readers.
We provide a more general formulation of equivariant networks in \cref{sec:steerable_CNNs}.
Finally, we point out several open research directions that are worth exploring in the field in \cref{sec:open_direct}.

\subsection{Overview}

\noindent{\emph{Authors: Youzhi Luo, Yi Liu, Simon V. Mathis, Alexandra Saxton, Pietro Liò, Shuiwang Ji}}\newline

\ifshowname\textcolor{red}{(Simon)}\else\fi
Describing physical data necessitates making choices, such as establishing a reference frame. While these choices facilitate the numerical representation of physical phenomena within data, the resulting data now mirrors \emph{both} the phenomenon under investigation \emph{as well as such choices}. As choices for description, like the frame of reference, are essentially arbitrary, the represented phenomena should not be influenced by these selections. This concept is referred to as \emph{symmetry}. Symmetries refer to aspects of physical phenomena that remain unchanged, or invariant, under transformations such as the change of reference frame. Understanding how to treat symmetries in data is therefore essential to artificial intelligence in science if we aspire to gain insight into the intrinsic, objective properties of the physical world, independent of our observational or representational biases.

\ifshowname\textcolor{red}{(Youzhi)}\else\fi 
If certain symmetries are present in the system, the predicted targets are naturally invariant or equivariant to the corresponding symmetry transformations. For instance, when predicting energies of 3D molecular structures, the predicted target remains unchanged even if the input 3D molecule is translated or rotated in 3D space. One possible strategy to achieve symmetry-aware learning is adopting data augmentation when training supervised learning models. Specifically, random symmetry transformations are applied to input data samples and labels to force the model to output \textit{approximately} equivariant predictions.
However, there are several drawbacks with data augmentation.
First, to account for the additional degree of freedom from choosing a reference frame, more model capacity would be needed to represent patterns that would be relatively simple in a fixed reference frame.
Second, many symmetry transformations, such as translation, can produce an infinite number of equivalent data samples, making it difficult for finite data augmentation operations to completely reflect the symmetries in data. 
Third, in some scenarios, we need to build a very deep model by stacking multiple layers to achieve good prediction performance. However, it would pose much more challenges to force the deep model to output approximately equivariant predictions by data augmentation if the model does not maintain equivariance at every layer.
Last but not least, in some scientific problems such as molecular modeling, it is important to provide provably robust predictions under these transformations so that users can employ machine learning models in a reliable way.

\ifshowname\textcolor{red}{(Youzhi)}\else\fi Given the drawbacks of using data augmentation, an increasing number of studies focus on developing symmetry-adapted machine learning models that are designed to meet the underlying symmetry constraints. With symmetry-adapted architecture, no data augmentation is required for symmetry-aware learning, and models can focus solely on learning the target prediction task. Recently, such symmetry-adapted models have shown significant success in scientific problems for a variety of different systems, including molecules (see Section~\ref{sec:mol}), proteins (see Section~\ref{sec:prot}), and crystalline materials (see Section~\ref{sec:mat}). In the following sections, we will elaborate on the symmetry transformations considered in the scientific problems discussed in this work, and the equivariant operations in designing symmetry-adapted models for these symmetry transformations.

\subsection{Equivariance to Discrete Symmetry Transformations}
\label{subsec:dis_equi}

\noindent{\emph{Authors: Youzhi Luo, Xuan Zhang, Jerry Kurtin, Erik Bekkers, Shuiwang Ji}}\newline

In certain scientific problems, the prediction targets are internally equivariant to a finite set of discrete symmetry transformations. To be concrete and simple, we consider the case where the inputs are 2D scalar fields,
and the symmetry transformations consist of rotating by the angles of $90^\circ$, $180^\circ$ and $270^\circ$~\cite{cohen2016group}. An example of these problems is simulating the dynamics of the fluid field (\emph{e.g.}, scalar vorticity or density) in a 2D square plane where we learn a mapping between the fluid field at the current time step to the fluid field at the next time step.
The simulated fluid fields should rotate accordingly if the input 2D fluid field rotates by $90^\circ$, $180^\circ$ or $270^\circ$ in certain scenarios (see Section~\ref{sec:pde} for details). Formally, let $X \in \mathbb{R}^{s\times s}$ be the input signals defined on a $s\times s$ grid, and the function $f:\mathbb{R}^{s\times s}\to \mathbb{R}^{s\times s}$ maps $X$ to the predicted field. We define the rotation by the angle of $\alpha$ as $R_\alpha:\mathbb{R}^{s\times s}\to\mathbb{R}^{s\times s}$.
The set of all discrete symmetry transformations is $\{R_\alpha\}_{\alpha\in\mathcal{A}}$, where $\mathcal{A}=\{0^\circ, 90^\circ, 180^\circ, 270^\circ\}$. 
Specifically, $R_{0^\circ}$ is the identity mapping. 
$R_{90^\circ}$ rotates the input matrix by $90^\circ$, \emph{i.e.}, $A'=R_{90^\circ}(A)$ satisfies $A'_{i,j}=A_{j, n-i}$ for any $A\in \mathbb{R}^{n\times n}$ and $0\le i,j \le n-1$ (zero-based index). 
$R_{180^\circ}$ and $R_{270^\circ}$ are compositions of two and three $90^\circ$ rotations, respectively. In other words, $R_{180^\circ}=R_{90^\circ}\circ R_{90^\circ}$ and $R_{270^\circ}=R_{90^\circ}\circ R_{90^\circ}\circ R_{90^\circ}$.
The equivariance to discrete symmetry transformations requires $f$ to satisfy
\begin{equation}
	\label{eqn:dis_equi_prop}
	f\left(R_\beta\left(X\right)\right)=R_\beta\left(f\left(X\right)\right),\ \forall \beta\in\mathcal{A}.
\end{equation}

To motivate the idea of achieving equivariance to discrete symmetry transformations in $\{R_\alpha\}_{\alpha\in\mathcal{A}}$, we first consider a minimal example of an equivariant \emph{group convolutional neural networks} (G-CNNs) ~\citep{cohen2016group}.
 Our example consists of a so-called lifting convolution \cite{bekkers2018roto} which performs convolutions with kernels rotated by every angle in $\mathcal{A}$ and then it applies a pooling operation over the newly introduced rotation axis.
 First, let us reconsider standard convolution. Given the input feature map $X\in\mathbb{R}^{s\times s}$ and a learnable convolution kernel $W\in\mathbb{R}^{k\times k}$, the standard convolution $X\ast W$ computes a $s\times s$ feature map, where the feature value at the $i$-th row, $j$-th column is computed as
\begin{equation}
	\label{eqn:conv}
	(X\ast W)_{ij} = \sum_{p=0}^{k-1}\sum_{q=0}^{k-1} W_{pq} X_{i+p, j+q}, \quad 0\le i,j\le s-k.
\end{equation}
Here we omit paddings for simplicity (the actual output size in~\cref{eqn:conv} is $(s-k+1)\times (s-k+1)$). 

Now consider the group equivariant lifting convolution, it consists of four standard convolutions with kernels rotated by angle $\alpha$. This creates the stack $\{F_\alpha \}_{\alpha \in \mathcal{A}}$ of feature maps $F_\alpha = X\ast R_\alpha\left(W\right)$ in which the new $\alpha$ axis indexes the filter response for each rotation $\alpha$. The output can thus be considered as a field of "rotation response vectors", which is a particular instance of a feature field with fibers that transform via the regular representation of the rotation group \cite{cesa2021ENsteerable}. A discussion of feature fields is beyond the scope of this section, but will be picked up in Section \ref{sec:steerable_CNNs}. The main point here is that the output is not the standard scalar field which we would like when modeling e.g. scalar vorticity or density. As such, our simple network follows the lifting convolution with a max pooling over $\alpha$-axis, i.e., we pool over the rotation responses. The simple architecture is then described as\begin{equation}
	\label{eqn:equi_conv}
	\mbox{GCNN}(X; W)=\mbox{Pool}\left(\{F_\alpha\}_{\alpha\in\mathcal{A}}\right)= \mbox{Pool}\left(\{X\ast R_\alpha\left(W\right)\}_{\alpha\in\mathcal{A}}\right),
\end{equation}
noting that $\mbox{Pool}(\cdot)$ pools over the rotation axis, \revisionOne{which can be defined as $\mbox{Pool}\left(\{F_\alpha\}_{\alpha\in\mathcal{A}}\right)_{ij} = \mbox{max}_{\alpha\in\mathcal{A}} \{\left(F_\alpha\right)_{ij}\}$.} 

The simultaneous use of four convolution operations with rotated kernels in combination with the pooling ensures that the overall G-CNN is equivariant, meaning
\begin{equation}
	\label{eqn:ec_equi_prop}	\mbox{GCNN}\left(R_\beta\left(X\right);W\right)=R_\beta\left(\mbox{GCNN}(X;W)\right),\quad \forall \beta\in\mathcal{A}.
\end{equation}
First, as shown in Equation~(\ref{eqn:equi_conv}), the four convolution operations rotate the kernel $W$ by $0^\circ$, $90^\circ$, $180^\circ$ and $270^\circ$, separately, and produce four feature maps $F_{0^\circ}, F_{90^\circ}, F_{180^\circ}, F_{270^\circ}$ by performing convolution operations on $X$ with these four kernels. From the calculation process of convolution in Equation~(\ref{eqn:conv}), we can show that if the input $X$ is rotated by any $\beta\in\mathcal{A}$, the four output feature maps $F_{0^\circ}, F_{90^\circ}, F_{180^\circ}, F_{270^\circ}$ will be rotated by the same angle $\beta$ and change their permutation order, \emph{i.e.},
\begin{align}
\{R_\beta\left(X\right)\ast R_\alpha(W)\}_{\alpha\in\mathcal{A}} &=\{R_\beta (X\ast R_\beta^{-1}\left(R_\alpha(W)\right))\}_{\alpha\in\mathcal{A}} \\ 
&=\{R_\beta (X\ast \left(R_{\alpha-\beta}(W)\right))\}_{\alpha\in\mathcal{A}} \\
&=\{R_\beta\left(F_{(\alpha-\beta)\text{ mod } 360^\circ}\right)\}_{\alpha\in\mathcal{A}}.
\label{eqn:ec_equi1}
\end{align}
Second, the pooling operation $\mbox{Pool}(\cdot)$ over the rotation axis is invariant to permutations within this axis and it preserves rotation equivariance over the spatial axes. We thus have
\begin{equation}
	\label{eqn:ec_equi2}
	\mbox{Pool}\left(\{R_\beta\left(F_\alpha\right)\}_{\alpha\in\mathcal{A}}\right)=R_\beta\left(\mbox{Pool}\left(\{F_\alpha\}_{\alpha\in\mathcal{A}}\right)\right),\quad \forall \beta\in\mathcal{A}.
\end{equation}
When Equations~(\ref{eqn:ec_equi1}) and~(\ref{eqn:ec_equi2}) hold, equivariance property in Equation~(\ref{eqn:ec_equi_prop}) will always be true.

\begin{figure}[t]
    \centering
    \includegraphics[width=\textwidth]{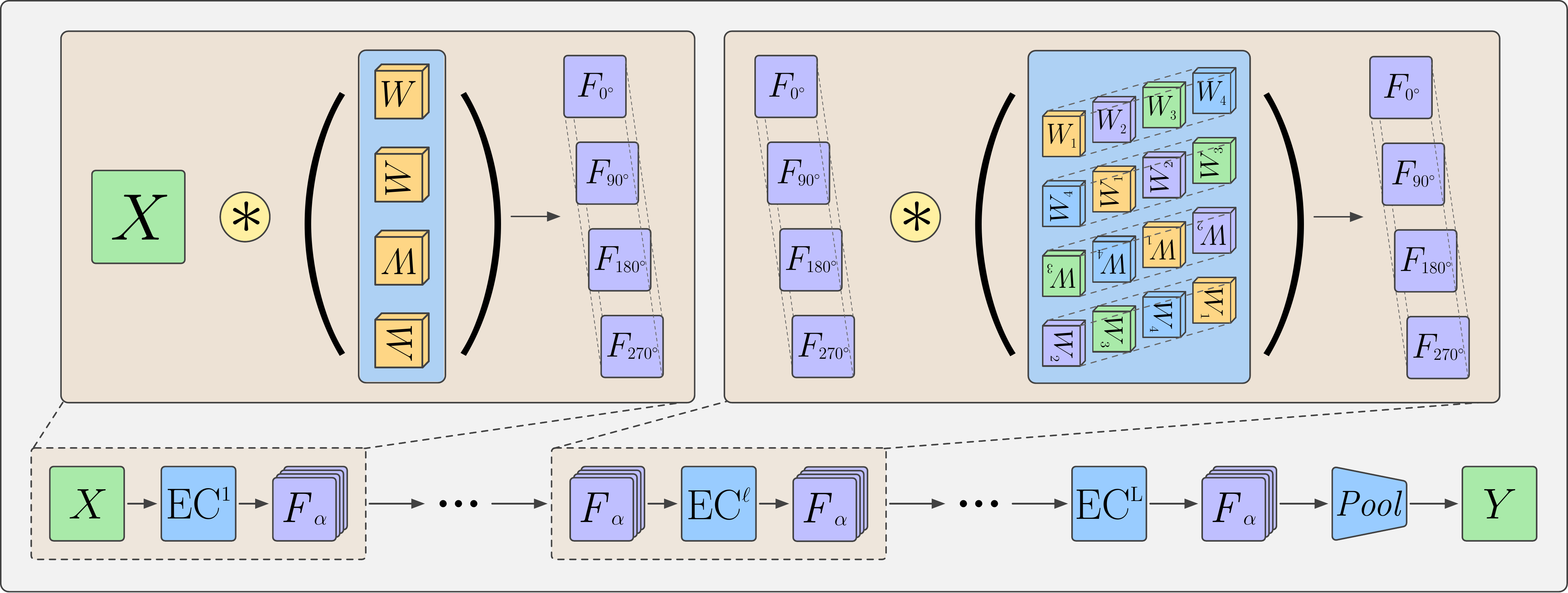}
    \caption{\ifshowname\textcolor{red}{(Jerry)}\else\fi
    Illustration of equivariant convolutional neural networks as in G-CNNs~\citep{cohen2016group}. The network passes input through an equivariant convolution layer (EC), resulting in a set of four output feature maps. If the input $X$ is rotated by $90^\circ$, $180^\circ$, or $270^\circ$, these feature maps will be rotated by $\beta$ and their ordering permuted. Following this, $L$ group convolution layers are applied, and a final pooling layer is added to account for the feature map permutations. This network is equivariant to rotations by $90^\circ$, $180^\circ$, or $270^\circ$.}\label{fig:gc_network}
\end{figure}

The above simple G-CNN creates locally rotation invariant feature fields, and can be used to build deep equivariant networks with \cite{andrearczyk2019exploring}. However, it's intermediate features would not carry any directional information because of the rotation-axis pooling. Instead, full group equivariant convolutional networks (G-CNNs) \cite{cohen2016group} typically start with a lifting convolution, which, as explained above, adds an extra rotation axis to the feature maps (hence often named lifting convolution), followed by group convolution layers that \textit{maintain the extra rotation axis} in the feature maps in order to be able to detect advanced patterns of features in terms of their relative positions and orientations, in which sense the kernels represent part-whole hierarchies \cite{Bekkers2020B-Spline}. The typical architecture then starts with a lifting convolution, followed by multiple equivariant group convolution layers before ending with a pooling layer over the $\alpha$-axis (see~\cref{fig:gc_network} for model illustrations). In each of these intermediate layers, four convolution kernels $W_1, W_2, W_3, W_4$ are used jointly to map the four input feature maps $F^{\text{in}}_{0^\circ}, F^{\text{in}}_{90^\circ}, F^{\text{in}}_{180^\circ}, F^{\text{in}}_{270^\circ}$ to the four output feature maps $F^{\text{out}}_{0^\circ}, F^{\text{out}}_{90^\circ}, F^{\text{out}}_{180^\circ}, F^{\text{out}}_{270^\circ}$ as
\begin{equation}
	\label{eqn:interm_equi}
	F^{\text{out}}_\alpha = \sum_{i=1}^{4}  F^{\text{in}}_{(\alpha+i*90^\circ)\text{ mod }  360^\circ}\ast R_\alpha \left(W_i\right),\quad \alpha\in\mathcal{A}.
\end{equation}

It can be shown that if the model uses the lifting convolution in the first layer, and full group convolutions as in~\cref{eqn:interm_equi}, the output feature maps at each layer are always equivariant to rotations. Additionally, due to the use of pooling over the rotation axis at the output end of the model, the prediction output of the model is ensured to have the equivariance property in~\cref{eqn:dis_equi_prop}. It can further be shown that a linear operator is equivariant \textit{if and only if} it is a group convolution \cite[Thm. 1]{Bekkers2020B-Spline}. It shows the importance of group convolutions as the essential building blocks for building equivariant G-CNNs; as such, in the work \cite[Thm. 3.1]{Cohen2019-generaltheory} the theorem is stated as (group) \textit{convolution is all you need}!%
\footnote{
    While regular group convolutions contain any linear $G$-equivariant maps,
    it is in high-dimensional settings more efficient to operate in their irreducible subspaces.
    This point is in more detail discussed in in \cite[Section 4.5]{weiler2023EquivariantAndCoordinateIndependentCNNs}.
}

\subsection{Equivariant Featurization of 3D Geometries}
\label{subsec:cont_feat}

\noindent{\emph{Authors: Youzhi Luo, Shuiwang Ji}}\newline

In other scientific problems, the symmetry transformations to be considered are not discrete but \emph{continuous}. Particularly, for many science problems discussed in this work, we focus on continuous $SE(3)$ transformations in 3D structures of chemical compounds, including translations and 3D rotations, where $SE(3)$ stands for the special Euclidean group in 3D space. In these problems, we aim to predict certain target properties from chemical compounds. A 3D point cloud is used to represent a chemical compound, where every basic unit of the chemical compound (\emph{e.g.}, every atom in the molecule) corresponds to a point in the 3D point cloud, and each point is associated with a 3D Cartesian coordinate. The target properties are usually constrained to be equivariant to $SE(3)$ transformations, \emph{i.e.}, rotations and translations. Note that different from the discrete rotations discussed in Section~\ref{subsec:dis_equi}, rotations in $SE(3)$ transformations are continuous, meaning that the 3D point cloud can rotate by any angle in 3D space. Formally, let $C=[\bm{c}_1,...,\bm{c}_n]\in\mathbb{R}^{3\times n}$ be the coordinate matrix of a 3D point cloud with $n$ nodes where $\bm{c}_i$ is the coordinate of the $i$-th point, $\bm{f}:\mathbb{R}^{3\times n}\to\mathbb{R}^{2\ell+1}$ be a function mapping coordinate matrices to $(2\ell+1)$-dimensional property vector that is $SE(3)$ equivariant with order $\ell$. 
The reason of involving an odd dimensionality of $2\ell+1$ in $f$ is related to irreducible representations and will be detailed in Section~\ref{subsec:phys}.
Here, order-$\ell$ equivariance requires $\bm{f}$ to satisfy
\begin{equation}
	\label{eqn:se3_property}
	\bm{f}\left(RC+\bm{t}\bm{1}^T\right)=D^\ell(R)\bm{f}(C),
\end{equation}
where $\bm{t}\in\mathbb{R}^3$ is the translation vector and $\bm{1}\in\mathbb{R}^{n}$ is a vector whose elements are all equal to one, which broadcasts the vector $\bm{t}$ to all $n$ input coordinates so that $\bm{t}\bm{1}^T\in \mathbb{R}^{3\times n}$.
$R\in\mathbb{R}^{3\times 3}$ is the rotation matrix satisfying $R^TR=I$ and $|R|=1$. $D^\ell(R)\in\mathbb{R}^{(2\ell+1)\times(2\ell+1)}$ is the (real) Wigner-D matrix of $R$. 
Here we assume $\bm{f}$ to be translation-invariant since most physics properties of a system only depend on the relative positions of its components instead of their absolute positions. For example, the energy of a molecule can be completely determined from its interatomic distances.
Wigner-D matrices are high-order rotation matrices for 3D rotation transformation in physics. When $\ell=0$, $D^\ell(R)=[1]$, and $\bm{f}$ corresponds to the properties that are invariant to $SE(3)$ transformations, such as total energy, Hamiltonian eigenvalues, band gap, \emph{etc.} When $\ell=1$, $D^\ell(R)=R$, and $\bm{f}$ corresponds to the properties that will rotate accordingly in 3D space if $C$ is rotated, such as force fields. When $\ell > 1, \ell \in \mathbb{N}_{+}$, $D^{\ell}(R) \in \mathbb{R}^{(2 \ell + 1) \times (2 \ell + 1)}$, $\bm{f}$ corresponds to properties to be rotated in space with a higher dimension beyond 3D space if $C$ is rotated, such as spherical harmonics functions with degree $\ell > 1$ and Hamiltonian matrix blocks.

To develop machine learning models for predicting such $SE(3)$-equivariant properties, we need advanced methods to encode geometric information in $C$ into $SE(3)$-equivariant features. A commonly used $SE(3)$-equivariant geometric feature encoding in physics and many existing machine learning methods is the spherical harmonics function. Generally, (real) spherical harmonics function $\bm{Y}^\ell(\cdot): \mathbb{R}^3\to\mathbb{R}^{2\ell+1}$ maps an input 3D vector to a $(2\ell+1)$-dimensional vector representing the coefficients of order-$\ell$ spherical harmonics bases (see Section~\ref{subsec:phys} for an introduction about the physical meaning of spherical harmonics bases). A nice property of the spherical harmonics function is that it is equivariant to order-$\ell$ rotations, or so-called order-$\ell$ $SO(3)$ transformations:
\begin{equation}
	\label{eqn:sphe_harm}
	\bm{Y}^\ell(R\bm{c})=D^\ell(R)\bm{Y}^\ell(\bm{c}),
\end{equation}
where $D^\ell(R)$ is the same Wigner-D matrix as in~\cref{eqn:se3_property}. Given the coordinates $\bm{c}_i, \bm{c}_j$ of two points $i,j$ in a 3D point cloud, spherical harmonics function can be used to encode their relative position $\bm{c}_i-\bm{c}_j$ to an order-$\ell$ $SE(3)$-equivariant feature vector.

\subsection{Equivariant Data Interactions}
\label{subsec:cont_equi}

\noindent{\emph{Authors: Youzhi Luo, Haiyang Yu, Hongyi Ling, Zhao Xu, Shuiwang Ji}}\newline

Recently, many $SE(3)$-equivariant operations based on spherical harmonics function have been proposed and applied to machine learning models, where spherical harmonics are used to \emph{featurize} 3D geometries into higher dimensions such that they can directly interact with high dimensional features that reside on the geometries (\emph{e.g.}, node features in a graph). In this section, we review methods of data interactions and operations that preserve equivariance.

\subsubsection{Equivariant Data Interactions via Tensor Product}

\begin{figure}[t]
    \centering
    \includegraphics[width=\textwidth]{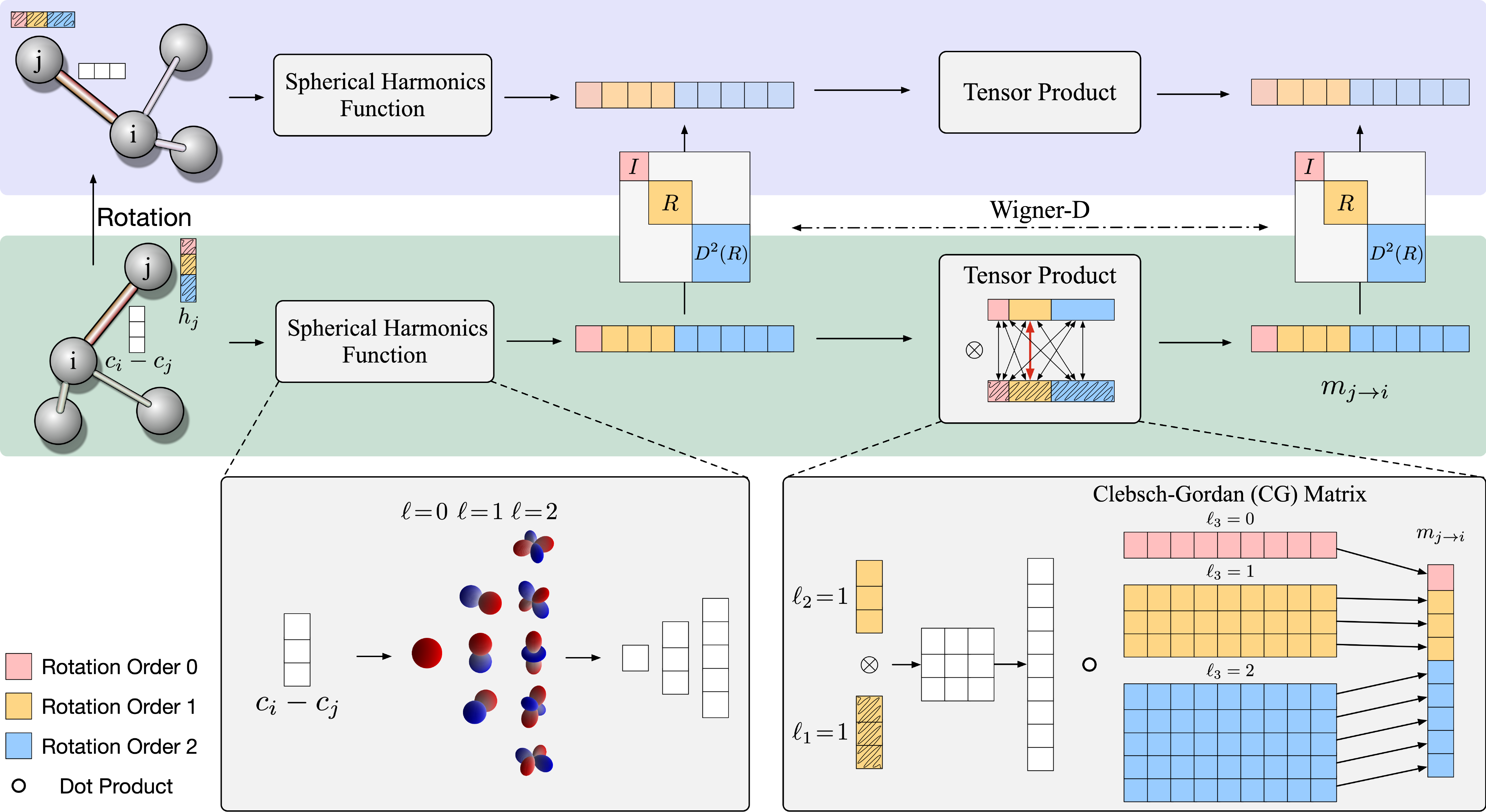}
    \caption{\ifshowname\textcolor{red}{(Hongyi)}\else\fi
    Illustrations of tensor product operations~\cite{thomas2018tensor,3d_steerableCNNs}. Here we show how to compute a message $\bm{m}_{j\to i}$ from node $j$ to node $i$, assuming the rotation orders are up to 2. Given the coordinates $\bm{c}_i$ and $\bm{c}_j$ of node $i$ and node $j$ in a 3D point cloud, their relational position $\bm{c}_i-\bm{c}_j$ is first encoded into an $SE(3)$-equivariant feature vector using spherical harmonics functions. A tensor product is then performed between the computed feature vector and $SE(3)$-equivariant node features $\bm{h}_j$ of node $j$ to compute the message $\bm{m}_{j\to i}$.  The resulting message $\bm{m}_{j\to i}$ is $SE(3)$-equivariant, rotating with the 3D point cloud via the corresponding Wigner-D matrix. }
    \label{fig:group}
\end{figure}

\ifshowname\textcolor{red}{(Youzhi)}\else\fi There are many different ways to featurize local geometry via spherical harmonic related operations. One widely used operation is message passing~\cite{gilmer2017neural} based on tensor product (TP) operations~\cite{thomas2018tensor,3d_steerableCNNs}. For an $n$-node point cloud with coordinates $C=[\bm{c}_1, \dots,\bm{c}_n]$, we assume that each node $i$ is associated with an order-$\ell_1$ $SE(3)$-equivariant node features $\bm{h}^{\ell_1}_i\in\mathbb{R}^{2\ell_1+1}$. The TP based message passing first computes a message $\bm{m}^{\ell_3}_i\in\mathbb{R}^{2\ell_3+1}$, then update $\bm{h}_i^{\ell_1}$ to new node feature $\bm{h}_i^{\prime\ell_1}$. This process can be formally described as
\begin{equation}
\label{eqn:tp_mpnn}
\begin{split}
    \bm{m}_i^{\ell_3}&=\sum_{j\in\mathcal{N}(i)}\bm{m}_{j\to i}^{\ell_3}=\sum_{j\in\mathcal{N}(i)}\mbox{TP}_{\ell_1,\ell_2}^{\ell_3}\left(\bm{c}_i-\bm{c}_j, \bm{h}^{\ell_1}_j\right),\\ \bm{h}_i^{\prime\ell_1}&=\bm{U}(\bm{h}_i^{\ell_1}, \bm{m}_i^{\ell_3}),
\end{split}
\end{equation}
where $\mbox{TP}_{\ell_1,\ell_2}^{\ell_3}(\cdot,\cdot)$ is the TP operation, $\mathcal{N}(i)$ is the neighboring node set of the node $i$, $\bm{U}(\cdot,\cdot)$ is the node feature updating function. $\mathcal{N}(i)$ is commonly defined as the set of nodes whose distances to $i$ are smaller than a radius cutoff $r$, \emph{i.e.}, $\mathcal{N}(i)=\{j: \Vert\bm{c}_i-\bm{c}_j\Vert_2\le r\}$. The TP operation in Equation~(\ref{eqn:tp_mpnn}) uses order-$\ell_2$ spherical harmonics function as the kernel to compute the message $\bm{m}^{\ell_3}_{j\to i}$ propagated from every node $j$ in $\mathcal{N}(i)$ to the node $i$. The detailed calculation process can be described as
\begin{equation}
	\label{eqn:tensor_product}
	\mbox{TP}_{\ell_1,\ell_2}^{\ell_3}\left(\bm{c}_i-\bm{c}_j, \bm{h}^{\ell_1}_j\right)= \revisionTwo{\mathscr{C}}_{\ell_1,\ell_2}^{\ell_3}\mbox{vec}\left(F\left(d_{ij}\right)\bm{Y}^{\ell_2}\left(\bm{r}_{ij}\right)\otimes \bm{h}^{\ell_1}_j\right).
\end{equation}
Here, $F(d_{ij})$ is a multi-layer perceptron (MLP) model that takes the distance $d_{ij}=\Vert\bm{c}_i-\bm{c}_j\Vert_2$ as input, $\bm{r}_{ij}=\frac{\bm{c}_i-\bm{c}_j}{d_{ij}}$, $\otimes$ is the vector outer product operation, \emph{i.e.}, $\bm{a}\otimes\bm{b}=\bm{a}\bm{b}^T$, $\mbox{vec}(\cdot)$ is the operation that flattens a matrix to a vector, and $\mathscr{C}_{\ell_1,\ell_2}^{\ell_3}$ is Clebsch-Gordan (CG) matrix with $2\ell_3+1$ rows and $(2\ell_1+1)(2\ell_2+1)$ columns. Particularly, CG matrix is widely used in physics to ensure that for $|\ell_1-\ell_2|\le \ell_3\le \ell_1+\ell_2$, the $\mbox{TP}_{\ell_1,\ell_2}^{\ell_3}(\cdot,\cdot)$ operation is always $SE(3)$-equivariant as
\begin{equation}
	\mbox{TP}_{\ell_1,\ell_2}^{\ell_3}\left(R\bm{c}_i-R\bm{c}_j,D^{\ell_1}(R)\bm{h}^{\ell_1}_i\right)=D^{\ell_3}(R)\mbox{TP}_{\ell_1,\ell_2}^{\ell_3}\left(\bm{c}_i-\bm{c}_j,\bm{h}^{\ell_1}_i\right).
\end{equation}
Hence, the message $\bm{m}^{\ell_3}_i$ is naturally $SE(3)$-equivariant. 
We refer to \emph{e.g.} Appendix A.5 of~\citet{brandstetter2022geometric} for derivation and \cref{TP_example} for an intuitive example.
Also, for the node feature update function $\bm{U}(\cdot,\cdot)$ in Equation~(\ref{eqn:tp_mpnn}), a linear operation or another TP operation can be used to maintain $SE(3)$-equivariance of the new node feature $\bm{h}_i^{\prime\ell_1}$. Since all calculations of TP-based message passing are $SE(3)$-equivariant, we can develop a powerful $SE(3)$-equivariant model by stacking multiple such message passing layers. Note that in the discussed message passing operation here, both the input node feature and output message have a single rotation order. In practice, a complete node feature $\bm{h}_i$ is composed of $SE(3)$-equivariant features with multiple rotation orders. Multiple messages with different rotation orders are computed by TP operations and concatenated to the message $\bm{m}_{j\to i}$ from the node $j$ to $i$ and the aggregated message $\bm{m}_i$. Then, $\bm{m}_i$ is used to update $\bm{h}_i$ to new node feature $\bm{h}'_i$. We illustrate the tensor product operations of calculating $\bm{m}_{j\to i}$ with rotation orders up to 2 in Figure~\ref{fig:group}.

That spherical harmonics based tensor product operations are not only \emph{sufficient} but strictly \emph{necessary} for $SE(3)$-equivariance was proven in \cite{3d_steerableCNNs}; see also Section~\ref{sec:steerable_CNNs} below.

\subsubsection{Approximately Equivariant Data Interactions via Spherical Channel Networks} \label{sec:equivariant_data_interactions}

\ifshowname\textcolor{red}{(Haiyang)}\else\fi In addition to linear or TP operations, the node feature $\bm{h}_i$ can also be updated on the spherical surface in a nonlinear way by spherical channel networks (SCNs)~\cite{zitnick2022spherical,passaro2023reducing} to achieve equivariance. An SCN considers all feature values in $\bm{h}_i$ as coefficients of spherical harmonics bases, and $\bm{h}_i$ represents a spherical function that maps a unit vector on the spherical surface to a real value. 
This spherical function can be described as a linear combination of spherical harmonics bases $f(\theta, \varphi) = \sum_{m, \ell} h_{i,m}^{\ell} Y_{m}^{\ell}(\theta, \varphi)$, where $\ell$ traverses the $SE(3)$-equivariant feature vectors with different rotation orders in $\bm{h}_i$, and $-\ell\le m\le\ell$ traverses the elements in an order-$\ell$ $SE(3)$-equivariant feature vector. Here, $\bm{Y}^{\ell}$ is the same spherical harmonics function defined in Section~\ref{subsec:cont_feat}, but its input vector is defined by the polar angle $\theta$ and the azimuthal angle $\varphi$ in the spherical coordinate system.
With $f(\theta,\varphi)$, an operation $\bm{G}(\bm{h}_i)$ samples multiple $(\theta, \varphi)$ pairs on the spherical surface and produces a feature map from their corresponding function values $f(\theta,\varphi)$. 
This feature map can be used as a representation of spherical functions.
In SCNs, a similar feature map is constructed from the message $\bm{m}_i$ by $\bm{G}$, and the node feature $\bm{h}_i$ is updated to $\bm{h'}_i$ by the point-wise convolution $\bm{F}_c$ on $\bm{G}(\bm{m}_i)$, $\bm{G}(\bm{h}_i)$ and the inverse operation of $\bm{G}$ as
\begin{equation}
	\bm{h'}_i = \bm{h}_i + \bm{G}^{-1} \left( \bm{F}_c \left( \bm{G}(\bm{m}_i), \bm{G}(\bm{h}_i) \right) \right).
\end{equation}
Here, the inverse operation $\bm{G}^{-1}$ transfers feature map values to coefficients of spherical harmonics bases by performing a dot-product between feature values and spherical harmonics bases.

Following SCN, the equivariant spherical channel network (eSCN) \cite{passaro2023reducing} proposes a novel equivariant convolution that efficiently reduces the complexity of tensor products. For each edge $\bm{r}$, a specific rotation matrix $R$ is applied to rotate the primary axis,
thereby aligning \revisionTwo{z} axis with the direction of the edge shown as \revisionTwo{$R \cdot \bm{r} = (0, 0, 1)$}. 
As a result, the spherical harmonic bases, denoted as $Y_{m}^{\ell}(R \cdot \bm{r})$, are equal to 1 when $m=0$, and $0$ otherwise. Thus, a significant computational cost reduction can be obtained since the calculation for $m\neq 0$ can be omitted in tensor product. Subsequently, 
an inverse of Wigner-D matrix is applied to the message to transform it back to original coordinate system, maintaining the equivariance.
To further improve efficiency of tensor product,  eSCN only considers non-zero entries in the large but sparse Clebsch-Gordan matrix by implementing an $SO(2)$ convolution comprised of two linear layers. Then, a point-wise non-linearity on the spherical surface is performed to obtain the message for each edge. Lastly, the eSCN adopts the same message aggregation as the SCN to update the node feature $\bm{h}_i$.
\revisionTwo{Note that in the eSCN paper~\cite{passaro2023reducing} the edges are aligned with the y axis due to the use of a different convention for spherical harmonics.}

Note that in both SCN and eSCN, the aggregation operation is
 not strictly but \textit{approximately} equivariant. 
Equivariance can only be maintained if the input node features are rotated by the angles that are exactly sampled in constructing the spherical grid. However, due to the continuous nature of the rotation, achieving this ideal condition is not always feasible.

\revisionTwo{
Besides eSCN, another popular approach to reduce computational cost of equivariant feature interaction is through scalarization~\citep{frank2024euclidean, cen2024high, aykentgotennet, yu2023efficient}, where messages between high-order features are first computed as scalars using operations such as inner product, and then return to high-order representations through attention or gating during message aggregation. Such methods have shown promising results in the balance between efficiency and expressiveness of TP-based networks.
Finally, as tensor products between high-order features can incur high computational cost,
research efforts have been made to understand the importance and effectiveness of TP-based operations with high-order features by examining the expressiveness and explainability~\citep{cen2024high,lingexplaining}.}

\subsection{Intuitive Physics and Mathematical Foundations}
\label{subsec:phys}

\noindent{\emph{Authors: Xuan Zhang, Yuchao Lin, Shenglong Xu, Tess Smidt, Yi Liu, Xiaofeng Qian, Shuiwang Ji}}\newline

\ifshowname\textcolor{red}{(Yi)}\else\fi In the above \cref{subsec:dis_equi}, \cref{subsec:cont_feat}, and \cref{subsec:cont_equi}, we provide applications of equivariance to discrete and continuous symmetry transformations in recent research, and describe how the tensor product is used in practice. In this section, we expect that, through some simple and intuitive examples, readers would understand the underlying theory in a reasonably short time. 
Specifically, in \cref{group_theory_example}, we provide a sketch of group and representation theory, \emph{i.e.}, the introduction of irreducible representations (irreps), and how equivariant neural network produce irreps through tensor product; we try to explain the intuitions of symmetry groups and irreps through a simple and discrete case, a square with four nodes, in \cref{symmetry_group_example};
in \cref{TP_example}, we provide an effortless example for readers to understand tensor products and Clebsh-Gordan coefficients; 
in \cref{spherical_harmonics_projection}, we introduce spherical harmonics projection, a concrete application of spherical harmonics;
we further manifest the idea of spherical harmonics functions from the angular momentum perspective in \cref{spherical_harmonics_example}.

\subsubsection{Overview} \label{group_theory_example}



\ifshowname\textcolor{red}{(Tess85\%, Yi15\%-substantial re-org and add some texts)}\else\fi
In this section, we give concrete examples to elucidate the fundamentals of group representation theory. 
Consider the vector space of polynomials $Z$, spanned by $(x^2, xy, xz, y^2, yz, z^2) $ from the direct product $(x, y, z) \times (x, y, z)$. 
When the original space $(x,y,z)$ is transformed by a $3\times 3$ rotation matrix, the vector space $Z$ will be transformed by a $6 \times 6$ matrix.
If we look at random $SO(3)$ rotations on this vector space, the $6 \times 6$ rotation matrices are dense; they do not look like they have independent vector spaces. However, if we perform a change of basis to $(x^2 + y^2 + z^2, xy, yz, 2z^2 - x^2 - y^2, xz, x^2 - y^2)$, then the rotation matrices take on a striking pattern. Factually, the original space can be decomposed into two independent subspaces $L_0 = (x^2 + y^2 + z^2)$ which is invariant (the group representations for all elements take the form of $I=\left[1\right]$) and $L_2 = (xy, yz, 2z^2 - x^2 - y^2, xz, x^2 - y^2)$. This actually describes how to decompose a reducible representation into irreducible representations (irreps). To further elucidate this, we give an example in \cref{symmetry_group_example} for a discrete case, which might be easier for starters to understand.

This transformation is significant, as it means any vector space can be described as a concatenation of these fundamental vector spaces. In principle, it requires that,
when conducting ``representation'' learning with machine learning, if the vector space we learn changes predictably under group action, \emph{e.g.}, rotations, then our ``learned'' vector space must be comprised of irreps, no matter how complex it may be. 
In the equivariant neural network literature, the term tensor product is used to define a tensor product plus decomposition operation, \emph{i.e.}, direct product two representations (reducible or irreducible) to produce (generally) a reducible representation and then decompose the reducible representation into irreps.
A more detailed description of tensor product and such decomposition is provided in~\cref{TP_example}, where we show in a more general setting for the tensor product of two different 3D vectors.

Additionally, the abstract structure from polynomials can be directly extended to geometrical concepts.
In fact, the vector space $L_2$ may look familiar to some readers as in fact, this is the vector space spanned by the angular frequency $\ell=2$ real spherical harmonics (modulo normalization factors), which form a vector space of functions that transform as the irreps of $SO(3)$. Similarly, the $L_0$ vector space is proportional to the $\ell=0$ spherical harmonic, which is a constant for all points on the sphere, $s \in \mathbb{S}^2$.
We will introduce an easy-to-understand application to manifest spherical harmonics
in \cref{spherical_harmonics_projection}, and also provide a detailed description in~\cref{sec:spherical_harmonics}. 
Just briefly and intuitively, spherical harmonics form an orthogonal basis for functions on the sphere. This means that any function in 3D space with a unique origin can be separated into radial and angular degrees of freedom because these degrees of freedom are orthogonal under 3D rotation. In fact, spherical harmonics are the basis functions for performing a Fourier transform on the sphere, which must have integer frequencies due to periodic boundary conditions (analogous to Fourier transforms over periodic spatial domains). As a result, spherical harmonics have a wide range of uses, from lighting in computer graphics, signal processing of sound waves, and description of physical systems, \emph{e.g.}, analyzing the cosmic microwave background and describing atomic orbitals.


\subsubsection{Illustration of Irreducible Representations via A Discrete Example} \label{symmetry_group_example}


In \cref{group_theory_example}, we provide a sketch of group and representation theory, \emph{i.e.}, the introduction of irreps, and how equivariant neural networks produce irreps through tensor product.
In this section, we explain symmetry groups and irreducible representations through a simple example. We further elucidate the motivation of equivariant neural networks to incorporate these symmetries for effective learning.
Note \cref{group_theory_example} gives a continuous form, which could be more generalizable. However, we believe it's easier for readers to understand the concepts through discrete group transformations as follows\ifshowname\textcolor{red}{(Yi up here)}\else\fi.

\ifshowname\textcolor{red}{(Shenglong)}\else\fi
\revisionOne{For simplicity, let us focus on the four-fold rotational symmetry of a square. Each node of the square has a scalar feature $a_1\sim a_4$. Under a 90 degree rotation, they transform as follows:  $ a_1 \rightarrow a_2 $,  
$ a_2 \rightarrow a_3 $,  
$ a_3 \rightarrow a_4$ ,  
$a_4 \rightarrow a_1$.
Consider the following linear-independent combinations of the four features:
\begin{equation}
    A = a_1 + a_2 + a_3 + a_4, \ B = a_1 - a_2 + a_3 - a_4, \ \vec E = (E_x, E_y) =(a_1-a_3, a_2 - a_4).
\end{equation}
Here, \( A \) remains unchanged under the four-fold rotation and is therefore invariant. On the other hand, \( B \) changes sign under the rotation. Furthermore, under the rotation, \( E_x \rightarrow E_y \) and \( E_y \rightarrow -E_x \), meaning that \( \vec{E} \) behaves as a 2D vector that rotates with the square. Most importantly $A$, $B$ and $\vec E$ do not mix under the rotation. In group theory which we introduce in~\cref{sec:group_theory_overview}, $A$, $B$ and $E$ are called irreducible representation of the symmetry group of the square.}

\revisionOne{
To ensure equivariance, the learning outcome must also transform properly under the given symmetry. This imposes strong constraints on how the learning outcome depends on the features. 
For example, let us consider a learning outcome that is linear in the features:  
$
f = w_1 a_1 + w_2 a_2 + w_3 a_3 + w_4 a_4
$
where $ w_i $ are the weights to be learned. If the learning outcome is known to be invariant under rotation, such as energy, then it must be proportional to $ A $ and take the form:
$
f = w (a_1 + a_2 + a_3 + a_4)
$
leaving only one weight to be learned.  
On the other hand, if the learning outcome is a vector—such as force—that rotates with the square, then it must be proportional to $\vec{E}$ and take the form:
$
\vec{f} = w (a_1 - a_3, a_2 - a_4)
$
ensuring that the output transforms correctly under rotation.
}

\revisionOne{
The reduction in the number of weights becomes much more significant when $f$ involves higher-order polynomials of the features. For example, a quadratic polynomial in $a_i$ generally contains 10 independent weights. However, if $f$ is a vector, symmetry constraints drastically reduce the number of possible terms it can take. As a result, the only allowed quadratic form for a vector-valued outcome is:
\begin{equation}
\vec{f} = w_1 A \cdot (E_x, E_y) + w_2 B \cdot (E_y, E_x),    
\end{equation}
which involves the tensor product of the linear irreps.
where $w_1$ and $w_2$ are the only two independent weights to be learned. This dramatic reduction in the number of learnable parameters highlights the power of enforcing equivariance in machine learning models, making them more efficient and interpretable while ensuring proper symmetry constraints.
}

\ifshowname\textcolor{red}{(Yi)}\else\fi
More generally, in practical cases like equivariant neural networks, tensor product takes two irreps and produces a reducible representation, which is further decomposed to irreps as inputs to the next layer, as mentioned in \cref{group_theory_example}. Essentially, this lays the foundation of achieving equivariance in modern equivariant neural networks.

This example illustrates how the group structure imposes significant constraints on the functions that map input data to the desired learning outcomes, based on their irreps. Equivariant neural networks aim to incorporate these constraints into the network architecture explicitly. By doing so, equivariant neural networks can leverage the inherent symmetries and transformations present in the data, leading to more effective and efficient learning.

\subsubsection{Tensor Products and Clebsh-Gordan Coefficients} \label{TP_example}

\ifshowname\textcolor{red}{(Xuan)}\else\fi 
Mathematically, the tensor product is defined to represent bilinear maps, which generalizes the scalar multiplication to vectors (tensors). Let us consider two 3D vectors $x,y\in\mathbb{R}^3$. Let $f\colon \mathbb{R}^3 \times \mathbb{R}^3 \to \mathbb{R}$ be a map taking two 3D vectors as inputs, being bilinear means when fixing one input, the restricted map $f(\cdot, \bm{y})$ or $f(\bm{x}, \cdot)$ is linear w.r.t. the other input. All such bilinear maps can be written as $f(\bm{x}, \bm{y})=\sum_{ij} c_{ij} x_i y_j$, where $x_i$ and $y_j$ are elements in $\bm{x}$ and $\bm{y}$, and $c_{ij}$ are the coefficients defining different maps. The tensor product between $\bm{x}$ and $\bm{y}$ is defined as $\bm{x}\otimes \bm{y} = [x_1 y_1, x_1 y_2, x_1 y_3, x_2 y_1, x_2 y_2, x_2 y_3, x_3 y_1, x_3 y_2, x_3 y_3]^T\in \mathbb{R}^9$. If we define a coefficient vector $\bm{c} = [c_{11}, c_{12}, c_{13}, c_{21}, c_{22}, c_{23}, c_{31}, c_{32}, c_{33}]^T\in\mathbb{R}^9$, then any bilinear map can be expressed as $f(\bm{x},\bm{y})=\bm{c}^T (\bm{x}\otimes \bm{y})$. Consequently, $f$  is uniquely represented by its coefficient vector $\bm{c}$. Thus, $f$ lives in a 9-dimensional vector space whose basis can be defined through tensor product. Concretely, the basis can be defined as $\{\bm{e}_i \otimes \bm{e}_j\}_{i,j\in\{1,2,3\}}$ where $\bm{e}_i$ and $\bm{e}_j$ are the canonical basis vectors of the original 3D space, \emph{i.e.}, $\bm{e}_1 = [1,0,0]^T$, \revisionTwo{$\bm{e}_2 = [0,1,0]^T$ and $\bm{e}_3 = [0,0,1]^T$}. Since $\bm{e}_i \otimes \bm{e}_j$ are vectors with 1 at the $(3i+j)$-th position and 0 elsewhere, they are orthogonal to each other.

An important property of tensor product is its equivariance. When $\bm{x}$ and $\bm{y}$ undergo a global rotation defined by a rotation matrix $R\in\mathbb{R}^{3\times 3}$, 
each element in the tensor product $R\bm{x}\otimes R\bm{y}$ is in the form of $\sum_{ij} c_{ij} x_i y_j$, where $c_{ij}$ is computed from elements in $R$. Thus, the tensor product $\bm{x}\otimes \bm{y}$ is also transformed by a matrix.
Let that matrix be $R^\otimes$, we have $R^\otimes (\bm{x}\otimes \bm{y}) = R\bm{x} \otimes R\bm{y}$. $R^\otimes$ then defines how the rotation transforms in the tensor product space. Note that $R^\otimes$ is a $9\times 9$ matrix and the dimension expands quickly with the dimensions of input spaces. We thus wish to identify smaller building blocks to efficiently describe how $\bm{x}\otimes \bm{y}$ transforms under rotations. Fortunately, this can be achieved for 3D rotations. 
For example, we know that when applying a global rotation, the dot product of two vectors is not changed. The dot product is a bilinear map defined as $f_\text{dot}(\bm{x}, \bm{y}) = x_1y_1 + x_2 y_2+x_3 y_3$. Expressed with the tensor product basis, the dot product can be defined by the coefficient vector $\bm{c}_\text{dot}=[1,0,0,0,1,0,0,0,1]^T$.
The rotation invariance of dot product gives $\bm{c}_\text{dot}^T  (R\bm{x} \otimes R\bm{y}) = \bm{c}_\text{dot}^T R^\otimes  (\bm{x}\otimes \bm{y}) = \bm{c}_\text{dot}^T  (\bm{x} \otimes \bm{y})$. Since this holds for all pairs of $\bm{x}$ and $\bm{y}$ (\emph{e.g.}, $\bm{x}\otimes\bm{y}$ can be any basis vector $\bm{e}_i \otimes \bm{e}_j$), we have $\bm{c}_\text{dot}^T R^\otimes = \bm{c}_\text{dot}^T$. Hence, the space spanned by the dot product (\emph{i.e.}, $\lambda \bm{c}_\text{dot}$, where $\lambda\in \mathbb{R}$) defines a 1-dimensional stable subspace for $R^\otimes$. 

Another stable subspace is the space spanned by the cross product. From the geometric interpretation,  we know that the cross product is equivariant to rotation. The cross product can be expressed as a stack of 3 bilinear maps (vector output) as 
\begin{equation}
    \bm{f}_\text{cross}(\bm{x}, \bm{y}) = \begin{bmatrix}x_2 y_3 - x_3 y_2 \\ x_3 y_1 - x_1 y_3 \\ x_1 y_2 - x_2 y_1\end{bmatrix},
\end{equation}
which can be expressed as the coefficient matrix 
\begin{equation}
    C_\text{cross} = \left[\begin{smallmatrix}0& 0& 0& 0& 0& 1& 0& -1& 0 \\ 0& 0& -1& 0& 0& 0& 1& 0& 0 \\ 0& 1& 0& -1& 0& 0& 0& 0& 0\end{smallmatrix}\right]^T,
\end{equation}
which is $9\times 3$ for 3 output dimensions. 
The equivariance of cross product gives $\bm{f}_\text{cross}(R\bm{x}, R\bm{y}) = R \bm{f}_\text{cross}(\bm{x}, \bm{y})$, which writes in the tensor product basis as $C_\text{cross}^T (R\bm{x}\otimes R\bm{y}) = C_\text{cross}^T R^\otimes (\bm{x} \otimes \bm{y}) = R C_\text{cross}^T (\bm{x} \otimes \bm{y})$, which holds for all pairs of $\bm{x}$ and $\bm{y}$. Thus we have 
\begin{equation}\label{eq:c_cross_rot}
    C_\text{cross}^T R^\otimes = R C_\text{cross}^T.
\end{equation}
We can show the space spanned by the cross product defines a 3-dimensional stable subspace for $R^\otimes$. 
Let $\bm{v}= C_\text{cross} \bm\lambda \in \mathbb{R}^{9}$ be
a linear combination of the columns in $C_\text{cross}$, where $\bm\lambda\in\mathbb{R}^3$. \revisionTwo{Using~\cref{eq:c_cross_rot},} we have $\bm{v}^T R^\otimes = \bm\lambda^T C_\text{cross}^T R^\otimes = \bm\lambda^T R C_\text{cross}^T \vcentcolon = \bm{u}^T$. \revisionTwo{We can see that} $\bm{u}=C_\text{cross} (R^T \bm\lambda)\in \mathbb{R}^9$ is still a linear combination of the columns in $C_\text{cross}$. Hence, we have proven that the 3-dimensional space spanned by the columns in $C_\text{cross}$ is stable to $R^\otimes$.

To have a complete view of this decomposition, we can wrap the coefficient vector $\bm{c}$ for a bilinear map into a matrix as
$$\hat C=\begin{bmatrix}
    c_1 & c_2 & c_3 \\ c_4 & c_5 & c_6 \\ c_7 & c_8 & c_9
\end{bmatrix}.$$
Then the coefficient space spanned by the dot product can be written as $\lambda_1 \hat{C}_\text{dot} = \lambda_1 \left[\begin{smallmatrix} 1 & 0 & 0 \\ 0 & 1 & 0 \\ 0 & 0 & 1\end{smallmatrix}\right]$, $\forall \lambda_1 \in \mathbb{R}$. The coefficient space spanned by the cross product can be written as
$$\lambda_2 \left[\begin{smallmatrix} 0 & 0 & 0 \\ 0 & 0 & 1 \\ 0 & -1 & 0\end{smallmatrix}\right] + \lambda_3 \left[\begin{smallmatrix} 0 & 0 & -1 \\ 0 & 0 & 0 \\ 1 & 0 & 0\end{smallmatrix}\right] + \lambda_4 \left[\begin{smallmatrix} 0 & 1 & 0 \\ -1 & 0 & 0 \\ 0 & 0 & 0\end{smallmatrix}\right], \forall \lambda_2, \lambda_3, \lambda_4 \in \mathbb{R}.$$ 
When projecting any coefficient $\hat C$ onto the space spanned by the dot product, the trace of $\hat C$ is extracted. One can verify that the space spanned by the cross product represents the space of all antisymmetric matrices, \emph{i.e.}, $A^T = -A$. The remaining degrees of freedom in the 9-dimensional space of $\hat C$ results in the space of all symmetric matrices with trace equal to 0, \emph{i.e.}, $A^T = A$, $\sum_i A_{ii}=0$, which is a 5-dimensional space. To summarize, we \revisionTwo{can} rewrite any $\hat C$ as the summation
\begin{equation}\label{eq:tp_decompose}
    \hat{C} =  \underbrace{\begin{bmatrix}\lambda_1 & 0 & 0 \\ 0 & \lambda_1 & 0 \\ 0 & 0 & \lambda_1\end{bmatrix}}_{\text{Trace}} + 
    \underbrace{\begin{bmatrix}0 & \lambda_4 & -\lambda_3 \\ -\lambda_4 & 0 & \lambda_2 \\ \lambda_3 & -\lambda_2 & 0\end{bmatrix}}_{\text{Antisymmetric}} + 
    \underbrace{\begin{bmatrix}\lambda_5 & \lambda_6 & \lambda_7 \\ \lambda_6 & -\lambda_5 - \lambda_9 & \lambda_8 \\ \lambda_7 & \lambda_8 & \lambda_9\end{bmatrix}}_{\text{Symmetric traceless}}.
\end{equation}


To show the symmetric traceless part is indeed 5-dimensional, we can expand the basis and write it as
$$\lambda_5 \left[\begin{smallmatrix} 1 & 0 & 0 \\ 0 & -1 & 0 \\ 0 & 0 & 0\end{smallmatrix}\right]+ 
\lambda_6 \left[\begin{smallmatrix} 0 & 1 & 0 \\ 1 & 0 & 0 \\ 0 & 0 & 0\end{smallmatrix}\right] + 
\lambda_7 \left[\begin{smallmatrix} 0 & 0 & 1 \\ 0 & 0 & 0 \\ 1 & 0 & 0\end{smallmatrix}\right] + 
\lambda_8 \left[\begin{smallmatrix} 0 & 0 & 0 \\ 0 & 0 & 1 \\ 0 & 1 & 0\end{smallmatrix}\right] +
\lambda_9 \left[\begin{smallmatrix} 0 & 0 & 0 \\ 0 & -1 & 0 \\ 0 & 0 & 1\end{smallmatrix}\right], \forall \lambda_{5-9} \in \mathbb{R}.$$ 
Translating to the tensor product basis, we can derive the function $f_{5D}$ as
\begin{equation}
    \bm{f}_\text{5D}(\bm{x}, \bm{y}) = \begin{bmatrix}x_1 y_1 - x_2 y_2 \\ x_1 y_2 + x_2 y_1 \\ x_1 y_3 + x_3 y_1 \\ x_2 y_3 + x_3 y_2 \\ -x_2 y_2 + x_3 y_3 \end{bmatrix}.
\end{equation}

Importantly, \cref{eq:tp_decompose} means the 9-dimensional coefficient space can be viewed a direct sum of a 1D, a 3D and a 5D vector spaces and each of them is stable to arbitrary global rotations. The decomposition can be conceptually written as 
\revisionTwo{$$V^1\otimes V^1=\underbrace{V^0}_{1D}\oplus \underbrace{V^1}_{3D} \oplus \underbrace{V^2}_{5D},$$
where $V^0$, $V^1$ and $V^2$ denote the three vector spaces, which have orders $\ell=$ 0, 1 and 2, respectively.}
It is worth noting the in general such decomposition depends on the choice of the transformation. Here the transformation is the 3D rotation ($SO(3)$ group). The decomposition would be different if we choose another transformation. For example, for the trivial transformation (group $G=\{e\}$), the decomposition would result in 9 1-dimensional trivial subspaces. 

One important property of these subspaces is that they cannot be further decomposed and remain stable to global rotations (\emph{i.e.}, they are irreducible). 
The 1D subspace spanned by dot product transforms under rotation as scalar and is irreducible by definition. 
The 3D subspace spanned by cross product transforms as vector and we can prove that it is also irreducible. Concretely, by~\cref{eq:c_cross_rot}, the 3-dimensional space spanned by $C_\text{cross}$ is transformed by the same rotation matrix $R$ under rotations. Since any 3D vector (under any basis) can be transformed to be colinear with any other 3D vector with some 3D rotation, there is no smaller subspace in the cross product space that is stable under arbitrary rotations.
For the 5D subspace, an intuitive proof for its irreducibility requires more advanced theories such as the angular momentum in physics, or the character theory in mathematics.
\ifshowname\textcolor{red}{(Shenglong)}\else\fi 
Nevertheless, we can gain some intuition about the behaviour of $f_\text{5D}$ by noticing that one of its component $x_1 y_1 - x_2 y_2$ changes sign under $90^\circ$ degree rotation around the $z$ axis, \emph{i.e.}, $(x_1, x_2)\leftarrow (-x_2, x_1)$ and $(y_1, y_2) \leftarrow (-y_2, y_1)$. 
More generally, $f_\text{5D}$ corresponds to a representation with $\ell=2$. Intuitively, an $\ell=2$ object is something that returns to itself after a $180^\circ$ rotation.

\ifshowname\textcolor{red}{(Xuan)}\else\fi 
Generally,  for any input dimensions, we can identify all such stable subspaces so that when the inputs undergo a global rotation, the subspaces in tensor product space will not mix with each other. By changing to the direct sum basis of these stable subspaces, one can efficiently express $R^\otimes$ in a block diagonal form. The matrices for performing such a change of basis are the Clebsh-Gordan (CG) coefficients. In summary, tensor products define a basis for all bilinear maps between two vector spaces, which is particularly suitable for studying equivariance when a global transformation is applied, since equivariance essentially describes maps between transformations in an input-independent way.

\subsubsection{Spherical Harmonics Projections and Equivariant Networks} \label{spherical_harmonics_projection}




\ifshowname\textcolor{red}{(Yuchao)}\else\fi
(Real) spherical harmonics $Y_m^\ell\colon \mathbb{S}^2\rightarrow\mathbb{R} $ are special functions defined on the surface of a unit sphere $\mathbb{S}^2$. They form a set of complete orthogonal bases for functions defined on $\mathbb{S}^2$. Thus every function on $\mathbb{S}^2$ can be expanded as a linear combination of those spherical harmonics.
This expansion is reminiscent of Fourier expansion of $\bm{v}\in V$ based on a set of complete orthogonal bases $\{\bm{u}_1, \dots, \bm{u}_n\}$ of vector space $V$ as
\begin{equation}
    \bm{v} = \sum_{i=1}^n\langle\bm{u}_i, \bm{v}\rangle \bm{u}_i.
\end{equation}
Similarly, a spherical function $f(\cdot):\mathbb{S}^2\rightarrow \mathbb{R}$ can be expanded by spherical harmonics such that
\begin{equation}
    f (\Omega) = \sum_{\ell, m} a_{\ell, m} Y^{\ell}_{m} (\Omega),
\end{equation}
where $a_{\ell,m} = \langle Y_m^\ell, f\rangle = \int Y^\ell_m (\Omega) f(\Omega) d\Omega$.


\revisionOne{To illustrate the idea of spherical harmonics expansion, we use the spherical Dirac delta $\delta$, which "picks out" the value of a function at a given point on the sphere. For any $\Omega, \Omega'\in\mathbb{S}^2$ and every test function $f\in C^\infty(\mathbb{S}^2)$, we have $\int_{\mathbb{S}^2}f(\Omega)\delta(\Omega - \Omega') d\Omega = f(\Omega')$. 
We can then obtain the spherical harmonics expansion of the Dirac delta as
\begin{equation}
    \delta(\Omega - \Omega') = \sum_{\ell,m} a_{\ell,m}Y_m^\ell(\Omega) = \sum_{\ell,m}\int Y^\ell_m (\Omega) \delta(\Omega - \Omega') d\Omega Y_m^\ell(\Omega)=\sum_{\ell,m}Y^\ell_m (\Omega') Y_m^\ell(\Omega).
\end{equation}
}

The above Dirac delta spherical expansion is the basis of the spherical harmonics projection, which can be used in local equivariant descriptors and convolution operations in equivariant neural networks. Specifically, to project a geometry vector to spherical harmonics, it contains two parts: a radial basis function to embed the magnitude of the vector; the spherical harmonics expansion of Dirac delta to embed the direction of the vector. Let a set of geometry vectors $\{\bm{r}_i\in\mathbb{R}^3\}_{i=1}^n$ and spherical harmonics functional input $\bm{x}\in \mathbb{R}^3, \Vert \bm{x}\Vert_2 = 1$. The spherical harmonics projection is given as 
\begin{equation}
    \sum_{i=1}^n\phi(\Vert \bm{r}_i\Vert_2) \sum_{\ell,m} Y_m^\ell \left(\frac{\bm{r}_i}{\Vert \bm{r}_i \Vert_2}\right) Y_m^\ell (\bm{x}),
\end{equation}
where $\phi(\cdot): \mathbb{R}\rightarrow [0,\infty )$ is the radial basis function providing scaling of projection. Since $\ell$ is often defined within a maximum degree $\ell\textsubscript{max}$ instead of over the whole non-negative integers due to computational efficiency, the summation $$\sum_{0\le\ell\le \ell\textsubscript{max},-\ell\le m\le \ell} Y_m^\ell \left(\frac{\bm{r}_i}{\Vert \bm{r}_i \Vert_2}\right) Y_m^\ell (\bm{x})$$ approximates the Dirac delta rather than exactly evaluates it. 
\begin{figure}[t]
    \centering
    \begin{center}
$\begin{array}{c@{\hspace{0.05in}}c@{\hspace{0.05in}}c@{\hspace{0.05in}}c@{\hspace{0.05in}}c@{\hspace{0.05in}}c}
\includegraphics[height=2.1cm, width=2.1cm]{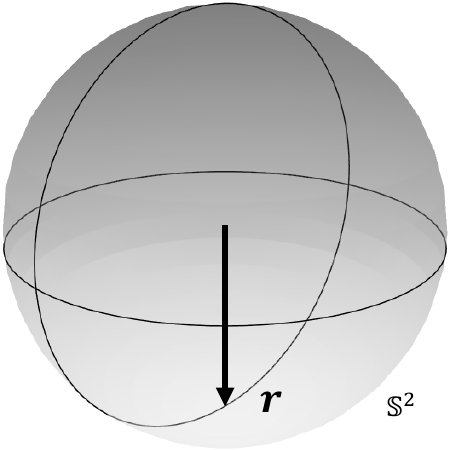}&
\includegraphics[height=2.1cm, width=2.1cm]{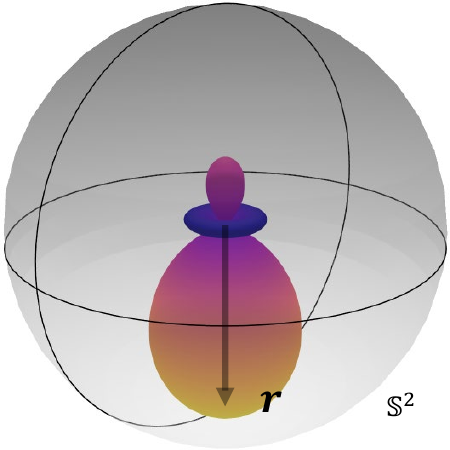}&
\includegraphics[height=2.1cm, width=2.1cm]{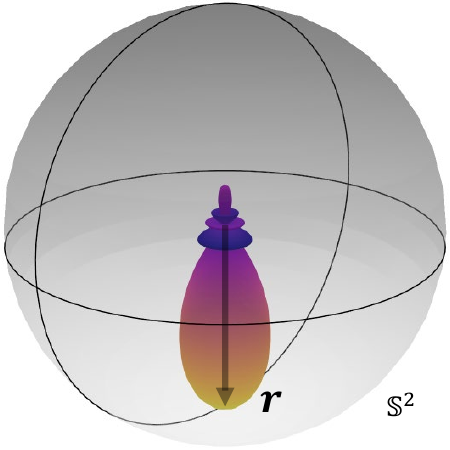}&
\includegraphics[height=2.1cm, width=2.1cm]{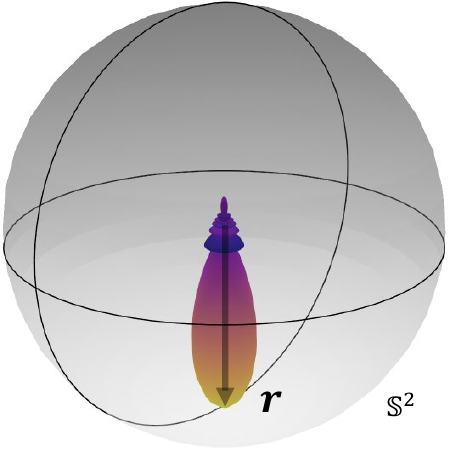}&
\includegraphics[height=2.1cm, width=2.1cm]{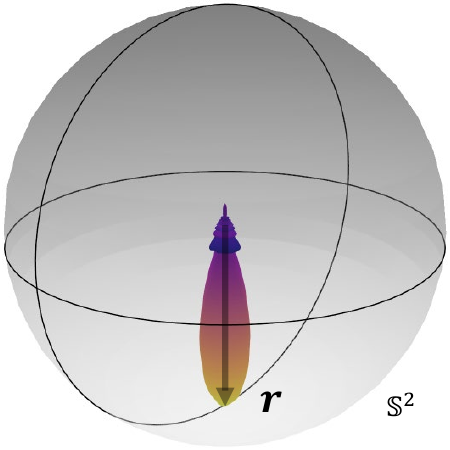}&
\includegraphics[height=2.1cm, width=2.1cm]{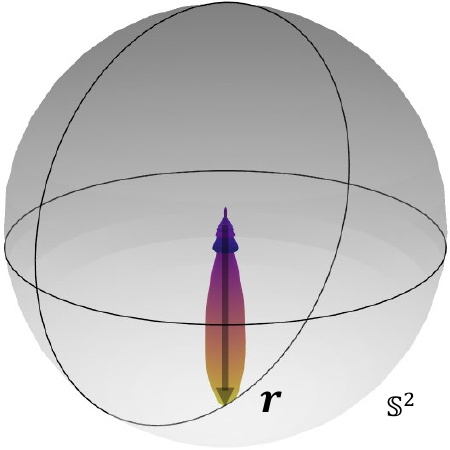}\\
\mbox{\scriptsize{}}&\mbox{\scriptsize{$\ell\textsubscript{max} = 2$}}&\mbox{\scriptsize{$\ell\textsubscript{max} = 4$}} &\mbox{\scriptsize{$\ell\textsubscript{max} = 6$}}&\mbox{\scriptsize{$\ell\textsubscript{max} = 8$}}&\mbox{\scriptsize{$\ell\textsubscript{max} = 10$}}
\end{array}$
\end{center}
    \caption{\ifshowname\textcolor{red}{(Yuchao, Xuan)}\else\fi Illustration of spherical harmonics projection for a single unit vector. From left to right, a unit vector $\bm{r}$ and its reconstruction from the spherical harmonics projection are plotted. A unit vector represents the Dirac delta on the unit sphere $\mathbb{S}^2$, with a non-zero value only in the direction it points to. Spherical harmonics projection gives the spherical expansion of the Dirac delta on $\mathbb{S}^2$ as a linear combination of spherical harmonics. Since the spherical harmonics projection contains an infinite number of terms, in the figure reconstructions are only considered within finite truncated terms for simplicity. Specifically, each sub-figure above from left to right corresponds to a finite subset of terms from the spherical harmonics projection, for $\ell\textsubscript{max} = 2$, $\ell\textsubscript{max} = 4$, $\ell\textsubscript{max} = 6$, $\ell\textsubscript{max} = 8$, and $\ell\textsubscript{max} = 10$, respectively. To visualize a function on the sphere clearly, the reconstruction is plotted as a 3D blob around the vector $\bm{r}$ where the distance to the origin represents the function magnitude on the sphere along that direction. Additionally, the maximum amplitude in the reconstruction is scaled to one for visualization. As illustrated by the above sub-figures, increasing the value of $\ell_{\text{max}}$ leads to a progressively thinner 3D blob, approximating the Dirac delta on the sphere.
    This example is adapted from lecture notes by Tess Smidt with permission.
    }\label{fig:spherical_harmonics_projection}
\end{figure}
Assume the maximum degree $\ell\textsubscript{max} \le 10$ and a vector $\bm{r}$ with $\Vert \bm{r}\Vert_2 = 1$, the spherical harmonics projection forms a blob around the vector $\bm{r}$, as illustrated in Figure~\ref{fig:spherical_harmonics_projection}. Specifically, when $\bm{x}$ is closer to $\bm{r}$, the projection value is larger and the distance to the sphere center is longer. In addition, as $\ell\textsubscript{max}$ increases, the 3D blob becomes progressively thinner, approximating the Dirac delta on the sphere.

\subsubsection{Spherical Harmonics Functions and Angular Momentum} \label{spherical_harmonics_example}

\ifshowname\textcolor{red}{(Xiaofeng)}\else\fi
The aforementioned spherical harmonics-based feature encoding and TP operation are actually tightly related to physical science, particularly quantum mechanics. In physics, spherical harmonics bases are commonly used in solving partial differential equations. Specifically, for single-electron hydrogenic atoms such as Hydrogen, the eigen wavefunctions of the electron are a set of analytic solutions of the Schr{\"o}dinger equation, given by the product of the radial part $R_{nl}(r)$ and complex spherical harmonics. More details of spherical harmonics can be found in Section~\ref{sec:spherical_harmonics}. The latter can be transformed into real spherical harmonics $Y_{m}^{\ell}(\theta, \varphi)$ which are often used in first-principles DFT, quantum chemistry, and recent deep learning models. The set of real spherical harmonics, denoted by $Y^{\ell}_m(\theta, \varphi): \mathbb{S}^2\to\mathbb{R}$ where $\ell \in \mathbb{N}$ is the orbital angular momentum quantum number and $m \in \mathbb{Z}, -\ell \leq m \leq \ell$ is the magnetic quantum number, forms a complete orthogonal basis set that can be used to expand any spherical functions. Additionally, in physical systems, the TP operation is  usually used in angular momentum coupling. Specifically, when we consider two electrons in the system with Coulomb forces, the angular momentum of the coupled wavefunctions can be deduced from the TP of the separate angular momentum. Besides the use of spherical harmonics for feature representations, they are also demonstrated for quantum tensor learning in Section~\ref{sec:dft}, such as quantum Hamiltonian learning.

\subsection{Group and Representation Theory} \label{sec:group_theory_overview}

\noindent{\emph{Authors: Maurice Weiler, YuQing Xie, Tess Smidt, Erik Bekkers}}\newline



Equivariant neural networks are formulated in the language of group and representation theory, the basics of which are briefly introduced in this section.
After some elementary definitions in Section~\ref{sec:group_theory},
we explain in Section~\ref{sec:group_actions} how groups can act on other objects and define invariant and equivariant functions w.r.t. such actions.
In the context of deep learning, symmetry groups act on data and features and the network layers are constrained to be invariant or equivariant.
The networks' feature spaces are usually vector spaces.
Group actions on vector spaces are described by group representation theory, which is discussed in Section~\ref{sec:group_representations}.
A more comprehensive introduction to group and representation theory in the context of equivariant neural networks is given in \cite[Appendix A]{weiler2023EquivariantAndCoordinateIndependentCNNs}.

\subsubsection{Symmetry Groups}
\label{sec:group_theory}

\ifshowname\textcolor{red}{(Maurice)}\else\fi
\emph{Groups} are algebraic objects which formalize symmetry transformations like, \emph{e.g.}, translations, rotations or permutations.
To motivate their formal definition, note first that we can always combine any two transformations into a single transformation.
This composition of transformations is naturally obeying certain properties which characterize the algebraic structure of groups.
Consider, for instance, the case of planar rotations.
Each rotation can be identified with a rotation angle,
and any two rotations by $\alpha$ and $\beta$ are composed to a combined rotation by $\alpha+\beta$ (modulo $2\pi$).
Note that a rotation by any angle $\alpha$ can be undone by another rotation by the negated angle $-\alpha$.
There is furthermore a trivial ``identity'' transformation, the rotation by $\alpha=0$ degrees, which does not do anything.
Finally, given rotations by three angles $\alpha,\beta$ and $\gamma$, the order of composition of the rotations is irrelevant, that is, $(\alpha+\beta)+\gamma = \alpha+(\beta+\gamma)$.
Symmetry groups are defined exactly as sets of transformations whose composition satisfies these three properties.
\begin{definition}[Group]
    Let $G$ be a set and $\ \bullet: G\times G\to G$ be a binary operation that takes two elements from $G$ and maps them to another element.
    If $(G,\bullet)$ satisfy the following three axioms, they form a group that is:
    \begin{itemize}
        \item[] (1) {\it Inverse:}
            for any $g\in G$ there exists an inverse element $g^{-1}\in G$ satisfying $g\bullet g^{-1} = g^{-1}\bullet g = e$;
        \item[] (2) {\it Identity:}
            there exists an identity element $e\in G$ which satisfies $e\bullet g = g \bullet e = g$ for any $g\in G$;
        \item[] (3) {\it Associativity:}
            $(g\bullet h)\bullet k = g\bullet(h\bullet k)$ for any $g,h,k\in G$.
    \end{itemize}
\end{definition} 
For brevity, one often refers to the set $G$ instead of $(G,\bullet)$ as group
and drops the composition in the notation, writing $gh$ for $g\bullet h$.
We will in the following make use of these abbreviations whenever the meaning is unambiguous.

The composition of planar rotations obeys actually yet another property:
for any two angles $\alpha$ and $\beta$,
the order of composition is irrelevant, since $\alpha+\beta = \beta+\alpha$.
This commutativity of group elements is not included in the definition above since it does not apply to all symmetry groups.
For instance, non-planar rotations in 3D do not commute with each other, rotations to not commute with translations or reflections, and permutations do in general not commute.
Groups like planar rotations, whose elements do commute, are called \emph{abelian}.
\begin{definition}[Abelian group]
    Let $G$ be a group.
    If all of its elements commute, that is, if $gh=hg$ for any $g,h\in G$, the group is called abelian.
\end{definition} 

An important class of groups are matrix groups, which are sets of square matrices that are composed via matrix multiplications and satisfy the three group axioms.
Associativity holds hereby by the definition of matrix multiplications;
the identity element is given by the identity matrix;
and the set is required to be closed under matrix inversion.
To give an example, consider the set of all invertible $n\times n$ matrices
$GL(\mathbb{R}^n) := \{ g\in\mathbb{R}^{n\times n} \,|\, \det(g)\neq0 \}$,
which is called \emph{general linear group} and is geometrically interpreted as the group of all possible basis changes of $\mathbb{R}^n$.
As matrix multiplications are in general not commutative, this group is not abelian.

Groups may contain subsets which are themselves forming groups.
They are therefore called subgroups.
\begin{definition}[Subgroup]
    Let $G$ be a group and $H\subseteq G$ be a subset of transformations.
    If $H$ is still forming a group, it is called a subgroup of $G$.

    One can show that it is sufficient to check that $H$ is closed under compositions;
    that is, $gh\in H$ for any $g,h\in H$,
    and under taking inverses, \emph{i.e.}, $g^{-1}\in H$ for any $g\in H$.
\end{definition} 

As an example, we consider the matrix subgroup
$SO(n) := \{ g\in\mathbb{R}^{n\times n} \,|\, \det(g)=1 \revisionOne{,\ g^T\!g=1} \}$
of $GL(\mathbb{R}^n)$.
It does not contain all $n\times n$ matrices with non-zero determinant, but only the subset of those with unit determinant.
That it is indeed a subgroup of $GL(\mathbb{R}^n)$ is clear since it is closed under composition, $\det(gh)=\det(g)\det(h)=1$ for $g,h\in SO(n)$,
and under inversion, $\det(g^{-1})=\det(g)^{-1}=1$.
The groups $SO(n)$ are called \emph{special orthogonal groups} since they consist of rotation matrices which transform between orthogonal bases of $\mathbb{R}^n$.
There are larger (non-special) orthogonal subgroups
$O(n) := \{ g\in\mathbb{R}^{n\times n} \,|\, \det(g)=\pm 1\revisionOne{,\ g^T\!g=1} \}$
of $GL(\mathbb{R}^n)$ which contain not only rotations but also reflections.

\subsubsection{Group Actions and Equivariant Maps}
\label{sec:group_actions}

\ifshowname\textcolor{red}{(Maurice)}\else\fi
The abstract definition of symmetry groups above captures their algebraic properties under composition,
but does not yet allow to describe \emph{transformations of other objects}.
One and the same group can, indeed, act on various different objects, for instance, different feature spaces.
Consider, for instance, the rotation group $SO(2)$ in two dimensions.
It acts naturally on 2-dimensional vectors in $\mathbb{R}^2$ via matrix multiplication,
but 2-dimensional rotations may also act on $\mathbb{R}^3$ by rotating around different axes,
or may even transform images or point clouds by rotating them in space.

Besides having a symmetry group $G$, we therefore also need to consider a set or space $X$ and need to specify how the group acts on it.
This action should certainly satisfy that a consecutive transformations by two group elements $g$ and $h$ should equal a single transformation by the composed group element $gh$.
It is furthermore desirable that the identity group element $e$ leaves any object that it acts on invariant.
These observations give rise to the following definition.
\begin{definition}[Group Action]
\label{def:group_action}
    Assume some group $G$ and denote by $X$ a set to be acted on.
    A (left) group action is then defined as a map
    \begin{align}
        \rhd:\ G\times X \to X,\quad (g,x)\mapsto g\rhd x
    \end{align}
    which satisfies the following two conditions:
    \begin{itemize}
        \item[] (1) Associativity:
            for any $g,h\in G$ and $x\in X$, the combined action decomposes as $(gh)\rhd x = g\rhd(h\rhd x)$; and
        \item[] (2) Identity:
            for any $x\in X$, the identity element $e\in G$ acts trivially, that is, $e\rhd x = x$.
    \end{itemize}
\end{definition} 
A set (or space) $X$ that is equipped with a $G$-action is called $G$-set (or $G$-space).

In general, a function $f\colon X\to Y$ maps between sets $X$ and $Y$.
Invariant and equivariant functions map more specifically between $G$-sets
and respect their group actions in the sense that they commute with them.
In the case of invariant functions, the output does not change at all when the input is transformed.
\begin{definition}[Invariant map]
    Let $f:X\to Y$ be a function whose domain $X$ is acted on by a $G$-action $\rhd_{\overset{}{X}}$.
    This function is called $G$-invariant if its output does not change under transformations of its input; that is, when
    \begin{align}
        f( g\rhd_{\overset{}{X}} x) = f(x)
        \qquad \text{for any}\ g\in G,\ x\in X.
    \end{align}
    This definition is captured graphically by demanding that the following diagram commutes for any $g\in G$,
    which means that following the top arrow yields the same result as following the bottom path:
    \begin{equation}
    \begin{tikzcd}[column sep=36pt, row sep=4pt, font=\normalsize]
            X   \arrow[rd, "f"]
                \arrow[dd, "g\,\rhd_{\overset{}{X}}"']
        \\ &Y
        \\  X   \arrow[ru, "f"']
    \end{tikzcd}
    \end{equation}
\end{definition}
Many objects in deep learning should be group invariant.
For instance, image classification should often be translation invariant,
or the ionization energy of a molecule should by invariant under rotations and reflections of the molecule.

Equivariance generalizes this definition by allowing for the output to co-transform with the input:
any $G$-transformation of the function's input leads to a corresponding $G$-transformation of the output.
\begin{definition}[Equivariant map]
    Let $f:X\to Y$ be a function whose domain $X$ and codomain $Y$ are acted on by $G$-actions $\rhd_X$ and $\rhd_Y$, respectively.
    This function is called $G$-equivariant if its output transforms according to transformations of its input; that is, when
    \begin{align}
        f( g\rhd_{\overset{}{X}} x)\ =\ g\rhd_{\overset{}{Y}} f(x)
        \qquad \text{for any}\ g\in G,\ x\in X.
    \end{align}
    The corresponding commutative diagram is given by:
    \begin{equation}
    \begin{tikzcd}[column sep=36pt, row sep=22pt, font=\normalsize]
           X    \arrow[r, "f"]
                \arrow[d, "g\,\rhd_{\overset{}{X}}"']
        &  Y    \arrow[d, "\ g\,\rhd_{\overset{}{Y}}"]
        \\ X    \arrow[r, "f"']
        &  Y
    \end{tikzcd}
    \end{equation}
\end{definition}
As an example of an equivariant map in deep learning, consider a neural network that predicts a magnetic moment of a molecule.
Since the underlying laws of physics are rotation invariant, a rotation of the molecule should result in a corresponding rotation of the predicted magnetic moment, that is, the mapping is required to be rotation equivariant.
The standard example of an equivariant network layer is the convolution layer:
as is easily checked, translations of their input feature map result in corresponding translations of output feature maps.
$G$-steerable convolutions generalize this behavior to more general symmetry groups~\cite{3d_steerableCNNs,thomas2018tensor,Weiler2019_E2CNN}.

\subsubsection{Group Representations}
\label{sec:group_representations}


\ifshowname\textcolor{red}{(Tess)}\else\fi
Group representation theory describes specifically how symmetry groups act on \emph{vector spaces}.
A group representation $\rho_X(g)$ can be thought of as a set of matrices parameterized by group elements $g \in G$ that act on vector space $X$ via matrix multiplication, $\rho_X(g): X \rightarrow X$.%
\footnote{
    More generally, $\rho_X(g)$ can be a linear operator acting on a vector space.
    If $X$ is finite-dimensional one can always express such operators in terms of matrices relative to some choice of basis.
}
For example, for a vector space of a single 3D Cartesian vector, commonly referred to as $(x, y, z)$, the representation of 3D rotations $SO(3)$ takes the familiar form of $3 \times 3$ matrices, which themselves can be parameterized in many ways, \emph{e.g.}, axis-angle, Euler angles, or quaternions are all valid parameterizations of $g \in SO(3)$.
Confusingly, group representation colloquially can refer to the matrix representation of the group $G$ on a specific vector space $\rho_X(g)$, the vector space $X$ that the group acts on, or the pair ($\rho_X$, $X$).

\ifshowname\textcolor{red}{(Tess/Maurice)}\else\fi
The definition of a group puts specific constraints on these matrix representations:
they must be invertible with $\rho_X(g^{-1}) = \rho_X(g)^{-1}$,
any multiplication of two elements of the representation must also be a representation of the group,
and a group representation will always contain the identity matrix $\mathbb{I}$,
the representation of what is commonly referred to as the group element $e$ in group theory literature. 
\begin{definition}[Group representation]
    Consider a group $G$ and a vector space $X$.
    A group representation of $G$ on $X$ is a pair $(\rho_X,X)$ where
    \begin{align}
        \rho_X: G \to GL(X)
    \end{align}
    is a group homomorphism from $G$ to the general linear group $GL(X)$ of $X$, \emph{i.e.}, to the group of invertible linear maps from $X$ to itself.
    That $\rho_X$ is a homomorphism means that
    \begin{align}
        \rho_X(gh)\ =\ \rho_X(g)\rho_X(h) \qquad \forall g,h\in G,
    \end{align}
    which ensures that the group composition on the l.h.s. is compatible with the matrix multiplication on the r.h.s.
\end{definition}
It is easy to show that $\rho_X(g^{-1}) = \rho_X(g)^{-1}$ and $\rho_X(e) = \mathbb{I}$ follow from this definition.

\subsubsection{Irreducible Representations}


\ifshowname\textcolor{red}{(Tess)}\else\fi
Group representations are not unique, and we have the following definition:
\ifshowname\textcolor{red}{(YuQing)}\else\fi
\begin{definition}[Isomorphic representations]
    Let $\rho_X$ and $\rho_Y$ be representations of group $G$ which act on vector spaces $X$ and $Y$ respectively. Then $\rho_X$ and $\rho_Y$ are said to be isomorphic if there exists a vector space isomorphism $Q:X\to Y$ such that for all $g\in G$
    \begin{equation}
        Q\rho_X(g)=\rho_Y(g)Q.
    \end{equation}
\end{definition}
If $Q$ is invertible such that $\rho_Y(g) = Q^{-1} \rho_X(g) Q$, then this can be thought of as a change of basis.
If $Q$ is unitary, then this is simply a ``rotation'' of the vector space basis. 

One of the most powerful results from group representation theory is that there are reducible and irreducible representations~ (irreps).
A reducible representation contains multiple independent irreps.
The vector spaces spanned by different irreps do not mix under group action, \emph{i.e.}, they are independent.

\begin{definition}[Reducible and irreducible representations]
    A representation $\rho_X$ of group $G$ is said to be reducible if it contains a nontrivial $G$-invariant subspace. In other words, there exists $V\subset X$ where $V\neq0$ such that $\rho_X(g)V=V$ for all $g\in G$.

    If no such subspace exists then the representation is said to be irreducible (commonly abbreviated as an irrep).
\end{definition}
In most cases, when a representation $\revisionOne{\rho_X}$ is reducible, then there exists a similarity transform $Q \rho_Y(g) = \rho_X(g) Q$ \revisionOne{for all $g\in G$} such that $\rho_Y(g)$ is block diagonal. 

\ifshowname\textcolor{red}{(YuQing)}\else\fi
In equivariant neural networks, the symmetry group considered usually acts in some well-defined way on our data. For example, the coordinates of atoms on a molecule would transform under rotation matrices. Hence, our input data \revisionOne{already naturally lives} in the vector space of some representation. Since the representations can be broken up into a direct sum of irreducible ones for most groups, we can specify the way our data transforms as a list of these irreps. In other words, the irreps are \revisionOne{a} natural data type in \revisionOne{many} equivariant neural networks. \revisionOne{Note however that there are many other architectures which do not use irreps such as group convolution. Nonetheless, even in these cases it is often useful to consider an irrep decomposition as Schur's lemma makes solving for equivariance constraints trivial given such a decomposition. See \nameref{representation_theory} in Appendix for an overview of Schur's lemma and finding irrep decompositions.}

However, there can be multiple representations of the same group which are isomorphic (equivalent). Hence, we have to make a choice when specifying the irreps of our group. Further, we would like a way to label our irreps which is independent of our specific choice of matrices. For the finite groups, one can do so using characters. This is essentially the trace of the matrices in our irreps and is why character tables are used extensively (though there are usually other naming conventions for the irreps of point groups). More details about characters and finding irreps of finite groups can be found in the finite groups part of \nameref{representation_theory} in Appendix.

In the case of infinite groups, using characters is infeasible since there are infinite group elements. Instead, there is well-understood theory on the irreps of semisimple Lie groups we can use. For the case of $SO(3)$ (and $SU(2)$), this essentially gives rise to the degree or angular momentum quantum number $\ell$. In general, the irreps are labelled by what are called dominant integral weights. This is the result of a very important theorem called the theorem of highest weights. A brief introduction of the representation theory of semisimple Lie groups can be found in the semisimple Lie groups part of \nameref{representation_theory} in Appendix.


\subsection{$SO(3)$ Group and Spherical Harmonics} \label{sec:spherical_harmonics}

\noindent{\emph{Author: Shenglong Xu}}\newline

The discrete group $C_{4v}$ \revisionOne{discussed in~\cref{symmetry_group_example}} is one of the simplest non-abelian point groups and has a finite number of irreps. On the other hand, the 3D rotation group $SO(3)$, which is relevant to many scientific domains, is continuous. It has an infinite number of irreps labeled by a positive integer, $\ell=0, 1, 2, 3, \ldots$, which are known as angular momentum, with each irreps having a dimension of $2\ell+1$. 

The irreps of $SO(3)$ are expressed using spherical harmonics, denoted as $Y^\ell_m(\theta, \phi)$ or $Y^\ell_m(\hat{r})$, where $-\ell \leq m \leq \ell$. For a fixed angular momentum $\ell$, the $2\ell+1$ spherical harmonics form the corresponding irreps of the $SO(3)$ group. These spherical harmonics are obtained by solving the 3D Laplace equation in spherical coordinates. The Laplace equation can be written as:
\begin{equation}
\vec{\nabla}^2 f(\vec{r}) = 0.
\end{equation}
In spherical coordinates, it takes the form:
\begin{equation}
\frac{1}{r^2} \frac{\partial}{\partial r}\left(r^2 \frac{\partial f}{\partial r}\right) - \frac{1}{r^2}\hat{\ell}^2 f = 0,
\end{equation}
where $\hat{\ell}^2$ represents the angular part:
\begin{equation}
\hat{\ell}^2 f = -\frac{1}{\sin\theta}\frac{\partial}{\partial \theta}\left(\sin\theta \frac{\partial f}{\partial \theta}\right) - \frac{1}{\sin^2\theta} \frac{\partial^2 f}{\partial \phi^2}.
\end{equation}
Since the Laplace equation is invariant under rotations in $SO(3)$ as well as the radial variable $r$, the operator $\hat{\ell}^2$, which depends only on the angular variables, is also rotationally invariant. By employing the method of separation of variables, we can separate the solution $f(\vec{r})$ into the radial part $G(r)$ and the angular part $Y(\theta, \phi)$ such that $f(\vec{r}) = G(r)Y(\theta, \phi)$. Substituting this into the Laplace equation, we obtain two equations:
\begin{equation}
\frac{1}{r^2}\frac{\partial}{\partial r}\left(r^2\frac{\partial G(r)}{\partial r}\right) - \frac{1}{r^2}\lambda G(r) = 0, \quad \hat{\ell}^2 Y(\theta, \phi) = \lambda Y(\theta, \phi).
\end{equation}
Let us focus on the second equation, which solely depends on the angular variables. It represents the eigenvalue equation of $\hat{\ell}^2$. Due to boundary conditions, the eigenvalue can only be $\ell(\ell+1)$, where $\ell$ takes values $\ell = 0, 1, 2, 3, \cdots$. It turns out that for a given $\ell$, there exist $2\ell+1$ linearly independent solutions, which are the spherical harmonics, denoted as $Y^\ell_m(\hat{r})$, where $m$ is an integer from $-\ell$ to $\ell$ that labels the $2\ell+1$ solutions.

The eigenvalue equation of $\hat{\ell}^2$ is rotationally invariant. Consequently, the solutions transform equivariantly under rotations.  If $Y^\ell_m(\hat{r})$ is a solution with the eigenvalue $\ell(\ell+1)$, the rotated function $Y^\ell_m(R\hat{r})$ is also a solution with the same eigenvalue. Therefore the rotated function can be expressed as a linear combination of different $m$ values, while keeping the same $\ell$ value. In other words, we have:
\begin{equation}
Y^\ell_m(R\hat{r}) = \sum\limits_{m'=-\ell}^\ell D_{m m'}(R) Y^\ell_{m'}(\hat{r}),
\end{equation}
where $D_{m m'}(R)$ is a matrix that depends on the rotation $R$. This matrix represents the transformation of the vector space spanned by the $2\ell+1$ solutions under 3D rotations.
Therefore, the $2\ell+1$ solutions, which are characterized by the same $\ell$ but different $m$ values, form a vector space that is closed under 3D rotations. This vector space corresponds to an irrep of the $SO(3)$ group.

The spherical harmonics are important to many scientific domains as they are the angular part of the solutions to arbitrary rotationally invariant partial differential equations. For instance, adding a potential term $V(r)$ to the Laplace equation only affects the radial equation but not the angular eigenequation. The spherical harmonics are also the solution to the Schrödinger equation of atoms, providing a quantum mechanical description of electrons' wavefunction. In this context, the values of the angular momentum quantum number $\ell$ correspond to the different types of atomic orbitals: $s(\ell=0)$, $p(\ell=1$), $d(\ell=2)$, and $f(\ell=3)$ orbitals. 


\revisionOne{Solving the eigenvalue equation for $\hat{\ell}^2$ reveals the explicit form of the spherical harmonics:}
\begin{equation}
    Y^\ell_m (\theta, \phi) = \sqrt{\frac{2\ell+1}{4\pi}\frac{(\ell-m)!}{(\ell+m)!}}P^m_\ell(\cos \theta) e^{i m \phi}
\end{equation}
where $P^m_\ell$ is a polynomial function called the associated Legendre function. It satisfies the relation $P_{\ell}^{-m}(\cos \theta) = (-1)^m (\ell-m)!/(\ell+m)!P_\ell^m(\cos \theta)$. The spherical harmonics form a complete orthogonal basis for functions defined on a sphere, and any spherical function can be expanded using this basis
\begin{equation}
    f(\theta, \phi) = \sum \limits_{\ell,m}a_{\ell,m} Y^\ell_m(\theta, \phi)
\end{equation}
similar to the Fourier series. 
Following the orthonormal condition, 
\begin{equation}
    \int Y^\ell_m{}^* (\theta, \phi) Y^{\ell'}_{m'} (\theta, \phi) d\cos \theta d\phi = \delta_{\ell,\ell'} \delta_{m, m'},
\end{equation}
The coefficient $a_{\ell,m}$ is $\int Y^\ell_m{}^* (\theta, \phi) f (\theta, \phi) d\cos \theta d\phi$. The finer details of $f(\theta, \phi)$ are captured by higher-order spherical harmonics. The spherical harmonics of $\ell=0, 1, 2$ are listed:
\begin{equation}
\begin{aligned}
    &Y^0_0 = \sqrt \frac{1}{4\pi} \\
    &Y^1_{-1} = \sqrt \frac{4}{8\pi}\sin \theta e^{-i\phi}, \ Y^1_{0} = \sqrt \frac{4}{8\pi}\cos \theta, \ Y^1_1 = -\sqrt \frac{4}{8\pi}\sin \theta e^{i\phi}\\
    &Y^2_{-2} = \frac{1}{4}\sqrt\frac{15}{2\pi } \sin ^2 \theta  e^{-2i \phi}, \ Y^2_{-1}= \frac{1}{2}\sqrt\frac{15}{2\pi } \sin \theta \cos \theta  e^{-i \phi}, Y^2_0 = \frac{1}{4}\sqrt \frac{5}{\pi}(3 \cos^2 \theta-1), \\
    &Y^{2}_1 = - Y^{2}_{-1}{}^*, \ Y^{2}_2 = Y^{2}_{-2}{}^*
\end{aligned}
\end{equation}

The complex spherical harmonics are convenient to use as it only gains a phase under rotation around the $z$ axis, \emph{i.e.}, $Y^\ell_m(\theta, \phi+\gamma)=e^{im\gamma} Y^\ell_m(\theta, \phi)$. In Cartesian coordinates, it is sometimes more intuitive to consider the real spherical harmonics $\mathcal{Y}^l_m$ which is a linear combination of the complex ones. Notice that $Y^\ell_m(\theta, \phi) = (-1)^m Y_{-m}^{\ell}{}^* (\theta, \phi)$ from the property of the associated Legendre function. The real spherical harmonics are constructed as 
\begin{equation}
    \mathcal{Y}_m^l \equiv\left\{\begin{array}{ll}
\revisionTwo{\frac{(-1)^m}{i \sqrt{2}}\left(Y_{|m|}^\ell-Y_{|m|}^\ell{ }^{*}\right)} & \revisionTwo{m<0} \\
Y_0^\ell &m=0 \\
\revisionTwo{\frac{(-1)^m}{\sqrt{2}}\left(Y_m^\ell+Y_m^\ell{ }^{*}\right)} \quad &\revisionTwo{m>0}.
\end{array}\right.
\end{equation}
The $\ell=0$ real spherical harmonics is the same as the complex one,
\begin{equation}
    \mathcal{Y}^0_0 = \sqrt \frac{1}{4\pi},
\end{equation}
which is a uniform function on the sphere. This is also called the $s$ orbital in atomic physics. 
The $\ell=1$ real spherical harmonics are
\revisionTwo{
\begin{equation*}
    \mathcal{Y}^{1}_{-1} = \sqrt \frac{3}{4\pi} \sin \theta \sin \phi = \sqrt \frac{3}{4\pi} \frac{y}{r},\
    \mathcal{Y}^{1}_0 = \sqrt \frac{3}{4\pi} \cos \theta  = \sqrt \frac{3}{4\pi} \frac{z}{r},\
    \mathcal{Y}^{1}_{1} = \sqrt \frac{3}{4\pi} \sin \theta \cos \phi = \sqrt \frac{3}{4\pi} \frac{x}{r}        
\end{equation*}
}
Since $(\mathcal{Y}^{1}_1, \mathcal{Y}^{1}_{-1}, \mathcal{Y}^{1}_{0}) \propto (x,y,z)$, it is clear that $\ell=1$ real spherical harmonics transform as a 3D vector under rotations. This is one of the advantages of using real spherical harmonics instead of complex ones. In atomic physics, these are called $p$ orbitals. For completeness, the $\ell=2$ real spherical harmonics are provided below:
\revisionTwo{
\begin{equation*}
    \mathcal{Y}^2_{-2} = \sqrt \frac{15}{4\pi} \frac{xy}{r^2}, \     \mathcal{Y}^2_{-1} = \sqrt \frac{15}{4\pi} \frac{yz}{r^2}, \        \mathcal{Y}^2_{0} = \sqrt \frac{5}{16\pi} \frac{2z^2 - x^2-y^2}{r^2}, \ 
    \mathcal{Y}^2_{1} = \sqrt \frac{15}{4\pi} \frac{xz}{r^2}, \  
    \mathcal{Y}^2_{2} = \sqrt \frac{5}{16\pi} \frac{x^2-y^2}{r^2}.
\end{equation*}}
In atomic physics, these are also called $d$ orbitals. In the rest of this work, we mostly employ the real spherical harmonics and simply refer to them as $Y^\ell_m$, instead of $\mathcal Y^\ell_m $ for the sake of convenience.

\subsection{A General Formulation of Equivariant Networks via Steerable Kernels}
\label{sec:steerable_CNNs}

\noindent{\emph{Author: Maurice Weiler, Alexandra Saxton}}\newline

\ifshowname\textcolor{red}{(Maurice)}\else\fi
All of the equivariant convolution operations discussed above can be unified in a comprehensive representation theoretic language, the theory of \emph{steerable CNNs}~\cite{Cohen2017-STEER,3d_steerableCNNs,Weiler2019_E2CNN,Cohen2019-generaltheory,lang2020WignerEckart,jenner2021steerablePDO,cesa2021ENsteerable,weiler2021coordinateIndependent,zhdanov2022implicitSteerableKernels}.
The feature spaces are in this formulation explained as spaces of \emph{feature vector fields},
whose transformation laws are prescribed by some choice of group representation~$\rho$.
The central result is that \emph{any} equivariant linear map between such feature maps is given by conventional convolutions, however, with symmetry constrained \emph{``steerable kernels''}.
An implementation of such steerable convolutions for any isometry groups in two and three dimensions is available in the PyTorch library \texttt{escnn} \cite{escnnLibrary}.
For a comprehensive review of steerable CNNs, we refer to \cite{weiler2023EquivariantAndCoordinateIndependentCNNs}.

\subsubsection{Feature Vector Fields}

Instead of focusing on a single symmetry group, for instance $E(3) = (\mathbb{R}^3,+)\rtimes O(3)$,
steerable CNNs consider any group $\mathrm{Aff}(G) = (\mathbb{R}^d,+)\rtimes G$ of affine transformations of $d$-dimensional Euclidean space $\mathbb{R}^d$.%
\footnote{
    The operation $\rtimes$ is a \emph{semidirect product}, here combining the translation group with transformations in $G$ (\emph{e.g.}, rotations).
}
They always contain translations in $(\mathbb{R}^d,+)$, which can be shown to necessitate convolution operations. 
$G\leq GL(\mathbb{R}^d)$ is any (sub)group of ${d\!\times\!d}$ matrices, including, \emph{e.g.}, rotations, reflections, scaling or shearing.
Affine group elements can always be written $tg$, where $t\in(\mathbb{R}^d,+)$ is a translation and $g\in G$ is a matrix group element.

\begin{figure}[t]
    \centering
    \includegraphics[width=.75\textwidth]{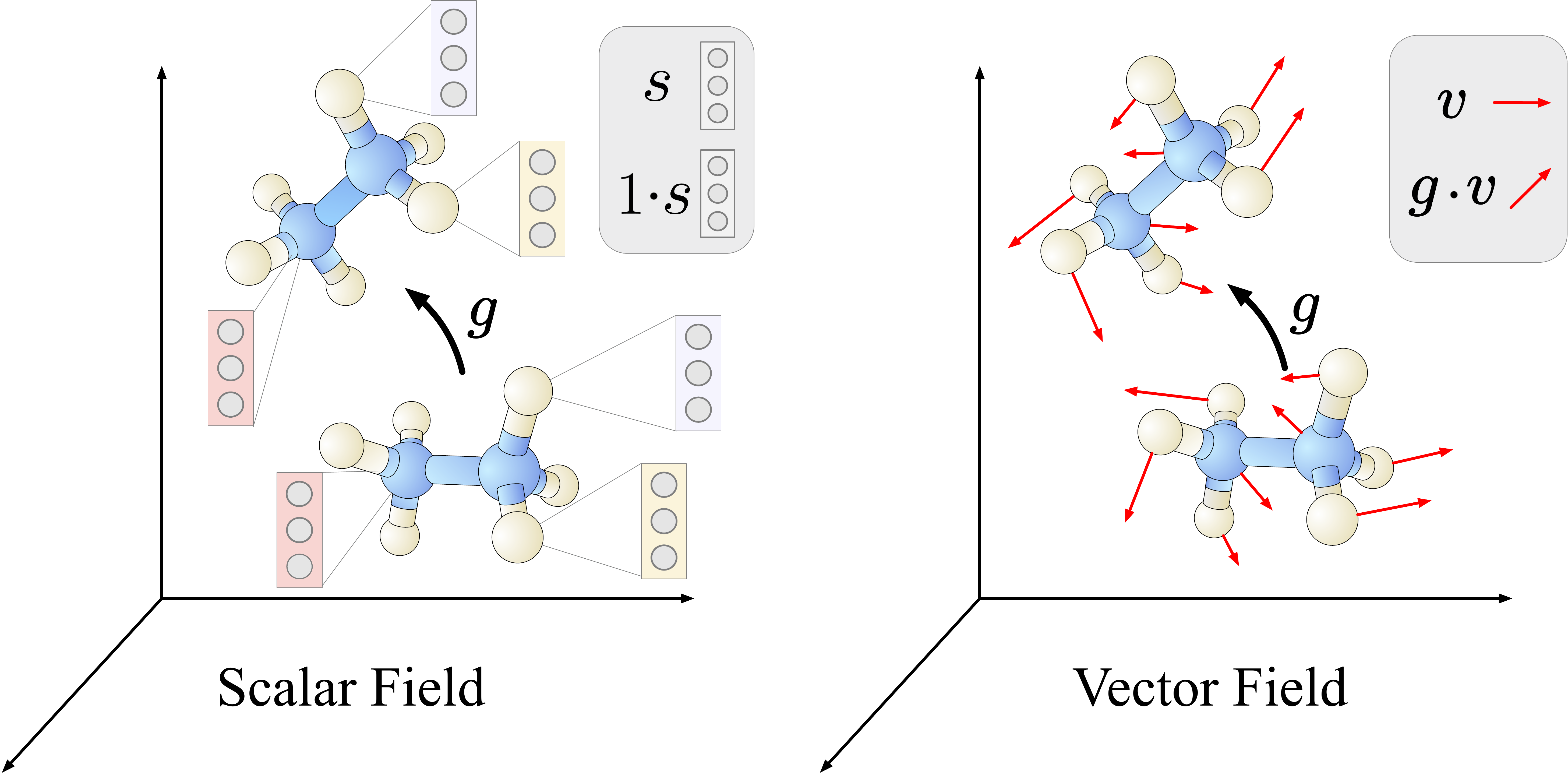}
    \caption{
        \ifshowname\textcolor{red}{(Allie re-draw)}\else\fi
        Scalar and vector fields as simple examples of feature vector fields.
        Affine groups act on such fields by
        (1) moving features across space (black arrow), and
        (2) transforming the features themselves via some group representation $\rho$.
        For the trivial representation $\rho(g)=1$, this explains scalar fields,
        while $\rho(g)=g$ describes vector fields.
        All of the feature spaces in the example above correspond to some choice of $G$-representation,
        \emph{e.g.}, Wigner-D matrices $\rho=D^\ell$ for tensor field networks and $G=SO(3)$.
        Steerable CNNs are build from layers which map in an equivariant way between feature fields,
        for instance from scalar to vector fields or vice versa.
        Linear equivariant maps are necessarily convolutions, however, with additionally symmetry constrained ``steerable kernels''.
        This figure is adapted from \citet{Weiler2019_E2CNN} with permission.
    }
    \label{fig:feature_fields}
\end{figure}

As mentioned above, steerable CNNs operate on spaces of feature vector fields, which are functions
\begin{align}\label{eqn:feature_field}
    f: \mathbb{R}^d \to \mathbb{R}^c
\end{align}
that assign $c$-dimensional feature vectors $f(x)\in\mathbb{R}^c$ to any point of Euclidean space $x\in\mathbb{R}^d$.
This definition is made in continuous space, however it can ultimately be discretized, \emph{e.g.}, on pixel grids or point clouds. 

Recall that equivariant network layers are by definition commuting with group actions
-- feature fields are therefore not yet fully specified by Equation \eqref{eqn:feature_field},
but are additionally equipped with actions of $\mathrm{Aff}(G)$,
examples of which are visualized in Figure~\ref{fig:feature_fields}.
The details of these actions are specified by a choice of \emph{field type}.
Before stating the general definition of such actions, let's look at some simple examples:
\begin{itemize}
\item
    \emph{Scalar fields} $s:\mathbb{R}^d\to\mathbb{R}$ consist of $c=1$ dimensional features, \emph{i.e.}, scalars.
    Under pure translations $t\in(\mathbb{R}^d,+)$ they transform like
    $[t\rhd s](x) := s(t^{-1}x) = s(x-t)$, \emph{i.e.}, the scalar values are shifted across space.%
    \footnote{
        We use $\rhd$ to denote group actions on fields.
        See Definition~\ref{def:group_action} for a general definition of group actions.
    }
    This is the transformation behavior of the feature maps of conventional translation equivariant CNNs.

\item
    More general affine group elements $tg$ act according to
    $[tg\rhd s](x) := s((tg)^{-1}x) = s(g^{-1}x-t)$ on scalar fields.
    This adds, for instance, spatial rotations or reflections $g\in O(d)$ of the scalar field, see Figure~\ref{fig:feature_fields} (left).

\item
    \emph{Tangent vector fields} are functions $v:\mathbb{R}^d\to\mathbb{R}^d$.
    As visualized in Figure~\ref{fig:feature_fields} (right), the transformations $g\in G$ do not only move the vectors to new spatial locations, but act on the individual vectors themselves, for example by rotating them when $G=SO(d)$.
    Mathematically, this action is given by $[tg\rhd v](x) := g\cdot v((tg)^{-1}x)$.

\end{itemize}

In general, the \emph{field type} is specified by any $G$-representation $\rho: G\to GL(\mathbb{R}^c)$,
which explains the action of $G$ on individual feature vectors in $\mathbb{R}^c$.
The corresponding action on the feature field as a whole becomes%
\footnote{
    This action is known as \emph{induced representation}.
    Specifically, $\rho$ is a $G$-representation acting on feature vectors
    and induces an $\mathrm{Aff}(G)$-representation which acts on feature fields as a whole.
}
\begin{align}
    [tg\rhd f](x)\ :=\ \rho(g) f \big( (tg)^{-1}x \big) \,.
\end{align}
Note how scalar and tangent vector fields are recovered when choosing the trivial representation $\rho(g)=1$ or the defining representation $\rho(g)=g$, respectively.
Other examples are tensor product representations $\rho(g) = (g^{-\top})^{\otimes r} \otimes g^{\otimes s}$, which correspond to order $(r,s)$ tensor fields, or irreducible representations, explaining \emph{e.g.}, the ${2\ell\!+\!1}$-dimensional features of tensor field networks when $G=SO(3)$.
The group convolutions from Section~\ref{subsec:dis_equi} correspond to the regular representation of the cyclic group $G=C_4$ (consisting of $90^\circ$ rotations).
It is given by permutation matrices that shift the field's four channels in a cyclic fashion:
\begin{align}
\label{eqn:regular_C4_rep}
\setlength\arraycolsep{2.9pt}
\def\arraystretch{0.8}
    \rho(0^\circ) = 
    \scalebox{.9}{$\begin{pmatrix}
        1 & 0 & 0 & 0 \\
        0 & 1 & 0 & 0 \\
        0 & 0 & 1 & 0 \\
        0 & 0 & 0 & 1
    \end{pmatrix}$}\,,\quad
    \rho(90^\circ) = 
    \scalebox{.9}{$\begin{pmatrix}
        0 & 0 & 0 & 1 \\
        1 & 0 & 0 & 0 \\
        0 & 1 & 0 & 0 \\
        0 & 0 & 1 & 0
    \end{pmatrix}$}\,,\quad
    \rho(180^\circ) = 
    \scalebox{.9}{$\begin{pmatrix}
        0 & 0 & 1 & 0 \\
        0 & 0 & 0 & 1 \\
        1 & 0 & 0 & 0 \\
        0 & 1 & 0 & 0
    \end{pmatrix}$}\,,\quad
    \rho(270^\circ) = 
    \scalebox{.9}{$\begin{pmatrix}
        0 & 1 & 0 & 0 \\
        0 & 0 & 1 & 0 \\
        0 & 0 & 0 & 1 \\
        1 & 0 & 0 & 0
    \end{pmatrix}$}\,.
\end{align}
It can be shown that group convolutions are generally explained by regular $G$-representations.

\subsubsection{Steerable Convolutions}

So far we only described the feature spaces and their group actions, but not the equivariant layers that map between them.
Specifically, for linear layers, \citet[Thm. 4.3.1]{weiler2023EquivariantAndCoordinateIndependentCNNs} show that the most general linear equivariant maps from input fields $f_\textup{in}$ of type $\rho_\textup{in}$ and output fields $f_\textup{out}$ of type $\rho_\textup{out}$ are given by \emph{convolutions}
\begin{align}
    f_\textup{out}(x)
    \ =\ [K \ast f_\textup{in}](x)
    \ =\ \int_{\mathbb{R}^d} K(x-y)\, f_\textup{in}(y)\ \mathrm{d}y
\end{align}
with convolution kernels
\begin{align}
    K: \mathbb{R}^d \to \mathbb{R}^{c_\textup{out}\times c_\textup{in}},
\end{align}
that are additionally required to be \emph{$G$-steerable}, \emph{i.e.}, need to satisfy the symmetry constraint
\begin{align}
    K(gx)\ =\ \frac{1}{|\det g|} \rho_\textup{out}(g) \,K(x)\, \rho_\textup{in}(g)^{-1}
    \qquad \forall\ g\in G,\ x\in\mathbb{R}^d \,.
\end{align}
Intuitively, the convolution operation ensures translational equivariance,
while $G$-steerability adds equivariance under $G$-actions,
thereby ensuring that the operation is mapping between the specified field types $\rho_\textup{in}$ and $\rho_\textup{out}$.
Note that a scalar convolution kernel would assign a single scalar to each point of $\mathbb{R}^d$,
however, as we are mapping between fields of $c_\textup{in}$ and $c_\textup{out}$-dimensional feature vectors, the kernels are ${c_\textup{out}\!\times\! c_\textup{in}}$ matrix valued.%
\footnote{
    This is also the case in non-equivariant convolutions.
    For example, discretized implementations on planar pixel grids in $d=2$ dimensions represent kernels as arrays of shape $(s_1,s_2,c_\textup{out},c_\textup{in})$, where the first two and the last two axes model the domain and codomain of the continuous kernel $K:\mathbb{R}^2\to\mathbb{R}^{c_\textup{out}\times c_\textup{in}}$, respectively.
}

Performant implementations of convolution operations are readily available,
such that the main difficulty in implementing equivariant convolutions is to parameterize the steerable kernels.
To this end, observe that kernels form a vector space and that the kernel constraint is \emph{linear}
-- steerable kernels live therefore in a vector subspace, and it is sufficient to solve for a basis in terms of which steerable kernels are expanded with learnable coefficients.
Such bases were derived for
$SO(3)$ irreps \cite{3d_steerableCNNs},
general representations of any $G\leq O(2)$ \cite{Weiler2019_E2CNN},
and, later, all representations of arbitrary compact groups $G$ (\emph{i.e.}, $G\leq O(d)$) \cite{lang2020WignerEckart,cesa2021ENsteerable}.
They are implemented in the \texttt{escnn} library, which is available for PyTorch and jax \cite{escnnLibrary}.

To clarify the kernel constraint
and to demonstrate how steerable CNNs relate to the equivariant models in the previous sections,
we turn to explicit examples.
\begin{itemize}
    \item
        The simplest example is when $\rho_\textup{in}$ and $\rho_\textup{out}$ are trivial representations, that is, when the kernel maps between scalar fields.
        Then $K:\mathbb{R}^d\to \mathbb{R}^{1\times1} = \mathbb{R}$ is a scalar kernel satisfying $K(gx) = \frac{1}{|\det g|} K(x)$.
        For orthogonal group $G\leq O(d)$, \emph{i.e.}, rotations and reflections, the volume scaling factor drops out, and the constraint requires that the kernel is $G$-invariant (\emph{e.g.}, rotation or reflection invariant).

    \item
        For $d=2$, $\rho_\textup{in}$ being trivial and $\rho_\textup{out}$ being the regular representation of $C_4$ as defined in Equation \eqref{eqn:regular_C4_rep}, the kernel has the signature $K: \mathbb{R}^2 \to \mathbb{R}^{4\times1}$.
        The constraint becomes $K(gx) = \rho_\textup{out}(g) K(x)$ which means that the $G$-rotated kernel on the left hand side should agree with the original kernel after shifting its four channels in a cyclic fashion.
        This is exactly the construction of kernels from Section \ref{subsec:dis_equi}, visualized in the left part of Figure \ref{fig:gc_network}.

    \item 
        We adapt the last example, now requiring both $\rho_\textup{in} = \rho_\textup{out}$ to be given by the regular $C_4$-representation.
        The kernel $K: \mathbb{R}^2 \to \mathbb{R}^{4\times4}$ should then satisfy $K(gx) = \rho_\textup{out}(g) K(x) \rho_\textup{in}(g)^{-1}$, which means that a spatial rotation equals a simultaneous shift of its rows and columns.
        The corresponding operation is a regular group convolution, whose kernel is shown in the right part of Figure \ref{fig:gc_network}.

    \item
        Let now $\rho_\textup{in}$ be trivial and $\rho_\textup{out} = D^\ell$ be an irrep of $SO(3)$.
        The corresponding kernels $K:\mathbb{R}^3\to \mathbb{R}^{(2\ell+1)\times1}$ need to satisfy $K(gx) = D^{\ell}(g) K(x)$,
        which is solved by kernels whose angular parts are spherical harmonics and whose radial parts are freely learnable.
        This explains those TP operations in Equation \eqref{eqn:tensor_product} where the input features $\bm{h}^{\ell_1}_j := f_\textup{in}(x_j)$ are of scalar order $\ell_1=0$ (trivial) and $\ell_2=\ell_3:=\ell$.

    \item
        If $\rho_\textup{in} = D^{\ell_1}$ and $\rho_\textup{out} = D^{\ell_3}$ are both irreps of $SO(3)$ we get
        $K:\mathbb{R}^3\to \mathbb{R}^{(2\ell_3+1)\times(2\ell_1+1)}$.
        The constraint $K(gx) = D^{\ell_3}(g) K(x) D^{\ell_1}(g)^{-1}$ is then equivalent to
        $\mathrm{vec} K(gx) = \big(D^{\ell_1} \otimes D^{\ell_3}\big)(g) \mathrm{vec}K(x)$.
        Using a Clebsch-Gordan decomposition of the irrep tensor product it is easy to show that such steerable kernels correspond exactly to the general TP operation in Equation \eqref{eqn:tensor_product}; see \cite{3d_steerableCNNs} or \cite{lang2020WignerEckart} for details.

    \item
        The $SO(3)$-equivariant spherical channel networks (SCNs) from Section \ref{sec:equivariant_data_interactions} operate on infinite-dimensional feature vectors that are functions on the 2-sphere $\mathbb{S}^2$.
        From a representation theoretic viewpoint, these are just \emph{quotient representations} as described in \cite{Weiler2019_E2CNN} and \cite{cesa2021ENsteerable}.
        As an extension to standard steerable CNNs, the steerable kernels used in SCNs are themselves computed from the data via messages.

\end{itemize}

What is the advantage of the formulation in terms of steerable CNNs?
\begin{enumerate}
\item
    It explains equivariant convolutions in a general setting, independent from specific choices of spaces, symmetry groups or group representations.
    It clarifies thereby how the different approaches in the previous sections relate.

\item
    The previous approaches were introduced by proposing certain operations,
    which were subsequently shown to be equivariant w.r.t. specific group actions.
    Steerable CNNs are, conversely, fixing the group actions and subsequently deriving equivariant linear maps between them.
    While \emph{e.g.}, the TP operations of tensor field networks turn out to be in one-to-one relation to steerable kernel solutions, the kernel constraint formulation allows to prove the \emph{completeness} of these solutions.
    In many other cases it could be shown that the authors were only using a subset of all admissible kernels, thus unnecessarily restricting the networks' expressive power \cite{weiler2021coordinateIndependent}.

\item
    The approaches above describe only a single field type per model (or class of field types, like irreps).
    Steerable CNNs allow to build hybrid models whose feature spaces operate simultaneously on feature vectors of regular, irrep, quotient or any other field type.

\end{enumerate}

The abstract representation theoretic formulation suggests natural generalizations to further spaces.
Specifically, \citet{Cohen2019-generaltheory} extended steerable CNNs to \emph{homogeneous spaces}, including \emph{e.g.}, spherical convolutions.
\citet{weiler2021coordinateIndependent,weiler2023EquivariantAndCoordinateIndependentCNNs} showed that coordinate independent convolutions on \emph{Riemannian manifolds} are similarly requiring $G$-steerable kernels.
This formulation is actually a \emph{gauge field theory},
which proves in particular that the equivariant networks in this section are not only equivariant under global transformations but also under more general local gauge transformations.

Steerable kernels have an interesting connection to the scalar, vector or spherical tensor operators appearing in quantum mechanics.
Both are formalized as so-called representation operators, which are described by the famous \emph{Wigner Eckart theorem} \cite{jeevanjee2011reprOp,wigner1931gruppentheorie}.
\citet{lang2020WignerEckart} proved this connection and showed how it allows to solve the kernel constraint in general.

\citet{jenner2021steerablePDO} extended steerable CNNs to the Schwartz distributional setting.
This covers in particular \emph{steerable partial differential operators} (PDOs),
which explains how the PDOs that appear ubiquitously in the physical sciences respect symmetries.

\subsection{Open Research Directions} \label{sec:open_direct}

\noindent{\emph{Authors: Hannah Lawrence, YuQing Xie, Tess Smidt}}\newline

\ifshowname\textcolor{red}{(Yi)}\else\fi In addition to the aforementioned areas, in this section, we highlight several research directions that are among the most cutting-edge and exciting categories. As the field is growing rapidly, we expect to enrich each of the mentioned directions as well as include more topics in the future.

\subsubsection{Symmetry Breaking}
\ifshowname\textcolor{red}{(YuQing)}\else\fi 
Spontaneous symmetry breaking is crucial for explaining many natural phenomena such as magnetism, superconductivity, and even the Higgs mechanism \cite{beekman2019introduction, strocchi2005symmetry}, and has been related to neural network training~\citep{ziyin2022exact}. In such cases, we have a highly symmetric input and desire to predict a lower symmetry output. It is desirable for equivariant networks to deal with this behavior, however, they are fundamentally limited.

Suppose our equivariant model is the function $f:X\to Y$. For an input $x$, suppose it is symmetric under a group $G$. Then for any $g\in G$, $f(gx)=f(x)$. This means the output must also be invariant under $G$, so it must have the same or higher symmetry. This means we can never predict a single lower symmetry output in an equivariant way. If we try to, the model will just average out all the degenerate outcomes, which might be useless. 

There are two perspectives one can take in resolving the symmetry breaking problem for equivariant models. The first is that there is actually one particular degenerate solution we want to predict. In this case, we know from symmetry that we are missing information to perform the task. It turns out by using the gradient of the loss function, we can infer what type of additional input is required to break the symmetry \cite{smidt2021finding}.

The second perspective is that all of the lower symmetry outputs are equally valid. In this case, we would like to represent all outputs simultaneously and/or randomly sample from them with equal probability. Treating this case properly is still an open problem.

\subsubsection{Empirical Benefits and Expense of Equivariance versus Invariance}\label{subsec:empirical_benefits_and_expense}


\ifshowname\textcolor{red}{(Tess)}\else\fi
Equivariance has been observed to give measurable benefit over invariance. Increasing the order of features (\emph{i.e.}, the maximum spherical harmonic degree) in $SE(3)$ and $E(3)$ equivariant models has been demonstrated to improve performance \cite{batzner20223,musaelian2023learning, owen2023complexity,yu2023efficient,yu2023qh9}.

The computational expense of equivariance is dominated by the tensor product (including decomposition into irreps), which involves the contraction of two inputs with the three index Clebsch-Gordan tensor $\revisionTwo{\mathscr{C}}^{(l_3, m_3)}_{(l_1, m_1) (l_2, m_2)} X_{(l_1, m_1)} Y_{(l_1, m_1)} = Z_{(l_3, m_3)}$. In voxel models, this contraction can be precomputed for ``traditional'' convolutional filters, which reduces the computational cost. Otherwise, it must be computed explicitly, \emph{e.g.}, ``traditional'' point wise convolutions and direct tensor product of features.

It is likely these expenses can be overcome through algorithmic workarounds (\emph{e.g.}, eSCN-like operations \citep{passaro2023reducing} as mentioned in \cref{sec:equivariant_data_interactions}) and optimization of tensor product operation, whether that be via optimized kernels, domain-specific compilers, or more tailored hardware. 

\subsubsection{Universality of Equivariant Neural Architectures} \label{subsec:universality_of_equiv_arch}

\ifshowname\textcolor{red}{(Hannah)}\else\fi
The previous sections discussed in detail how to tailor neural architectures such that they can only represent invariant or equivariant functions, no matter what weights are learned. Although the fundamental goal of this endeavor is to advantageously \emph{restrict} the family of learnable functions to a subfamily known to contain the ground-truth solution, it is important to understand just how expressive a given architecture is within the family of equivariant functions. For instance, is the ground-truth solution still contained in the set of equivariant functions expressible by the architecture family? Clearly, this is an important sanity check.

Informally, an equivariant architecture family is said to be universal if, for any continuous equivariant function and error threshold $\epsilon$, there exists a network in the family, typically that is ``large" enough in some sense (\emph{e.g.}, sufficiently many channels, layers, or orders), that approximates that function within error $\epsilon$, according to some functional norm. Happily, prior work has established that many equivariant architectures are universal. In brief, \citep{yarotsky2018universal} first proved that equivariant networks based on polynomial invariants and equivariants are universal, while \citet{bogatskiy2022symmetry} showed that most architectures based on tensor products of irreducible representations of a Lie group are universal. \citet{dym2021universality} also demonstrated that $SE(3)$-transformers and tensor field networks, two popular architectures operating on point cloud inputs, are universal. 
However, characterizing the expressivity of graph neural networks is an active research area, and they are in general not universal. Foundational work \citep{xu2018powerful} connected the expressivity of message-passing architectures to the Weisfeiler-Lehman hierarchy of graph isomorphism tests, and \citet{joshi2023expressive} recently began extending this work to \emph{geometric} graph networks (\emph{i.e.}, graphs embedded in 3D space, which is often how point clouds are processed after connecting each point to its nearest neighbors). Such analyses of universality are not sufficient for predicting the relative performances of different equivariant architectures, but are a worthwhile criterion to evaluate when selecting an appropriate equivariant learning method for a given scientific task.





\subsubsection{Frame Averaging as an Alternative for Equivariance} \label{subsec:frame_averaging}

\ifshowname\textcolor{red}{(Hannah)}\else\fi
As described in the previous section, most equivariant architectures are therefore expressive within the class of continuous equivariant functions. However, a key drawback of current tensor-based architectures, such as those discussed in \cref{subsec:cont_equi} and \cref{subsec:empirical_benefits_and_expense}, is their scalability. For example, a point cloud architecture following the template of tensor field networks \citep{thomas2018tensor} naively takes time $O(L^6)$ for a single forward pass, where $L$ is the maximum spherical harmonic index \citep{passaro2023reducing}. Recently, frame averaging has emerged as a lightweight alternative to constrained architectures for enforcing equivariance in a learning pipeline. 

Formally, a \emph{moving frame} was first defined in 1937 by mathematician Élie Cartan as a smooth, equivariant map $\rho: \mathcal{M} \rightarrow G$, where $\mathcal{M}$ is a manifold on which the Lie group $G$ acts smoothly \citep{cartan1937frame}. The equivariance property ensures that $\rho(gm) = g\rho(m)$ $\forall m \in \mathcal{M}$. Although Cartan defined these objects for the purpose of studying invariants of submanifolds, 
they provide an intuitive method for enforcing equivariance in a learning pipeline.

First, suppose we are given a function $f:\mathcal{M} \rightarrow \mathcal{Y}$, where $\mathcal{Y}$ is some target space, and a moving frame $\rho$. We can use $\rho$ to make $f$ invariant (known as the invariantization of $f$) as
\begin{align*}
    f'(m) := f(\rho(m)^{-1}m) \: \: \forall m \in \mathcal{M}.
\end{align*}
It is easy to check that $f'$ is invariant if $$f'(hm) = f(\rho(hm)^{-1}hm) = f((h\rho(m))^{-1}hm) = f(\rho(m)^{-1}m) = f'(m).$$
Quite similarly, we can use $\rho$ to make $f$ equivariant as 
\begin{align*}
    f''(m) := \rho(m)f(\rho(m)^{-1}m).
\end{align*}
One can again check that $f''$ is equivariant if
\begin{align*}
    f''(hm) = \rho(hm)f(\rho(hm)^{-1}hm) = \rho(hm)f(\rho(m)^{-1}m) = h\rho(m)f(\rho(m)^{-1}m) = h f''(m).
\end{align*}
Here, note that the input and output group actions are the same. To make $f$ equivariant with respect to a different group action on $\mathcal{Y}$, we simply need another moving frame $\rho'$ that is equivariant with respect to that group action, and can define $f''(m):=\rho'(m)f(\rho(m)^{-1}m)$ instead. Moreover, although moving frames were initially defined for Lie groups acting on manifolds, the straightforward reasoning above applies to any group acting on any space $\mathcal{M}$. 

A straightforward method for equivariant machine learning is therefore to learn the function $f$ using an \emph{arbitrary} architecture, and make it invariant or equivariant using the moving frame constructions above. One must backpropagate through the moving frame, necessitating a degree of smoothness, but the end-to-end framework produces an equivariant function while (1) not requiring any specialization to the group $G$ besides the fixed moving frame, and (2) allowing for an efficient, standard architecture $f$. 
Intuitively, the frames method turns the arbitrary function $f$ into an equivariant function by only relying on its behavior at a fixed point on each orbit. \citet{puny2021frame} generalize this framework to allow for \emph{averaging} over an equivariant set of points on each orbit instead. Concretely, they define a frame $\mathcal{F}$ more generally as a set-valued function, $\mathcal{F}:\mathcal{M} \rightarrow 2^G \backslash \emptyset$, which is equivariant: $\mathcal{F}(gm) = g\mathcal{F}(m)$,
where the equality is between sets. 
It is then easy to check that the following ``frame-averaged'' function is equivariant if
\begin{align*}
    \langle f \rangle_{\mathcal{F}}(x) := \frac{1}{|\mathcal{F}(x)|}\sum_{g \in \mathcal{F}(x)} f(g^{-1}x).
\end{align*}
We note that, when the frame maps to $G \in 2^G$ for all elements of the input space $\mathcal{M}$, then frame-averaging reduces to the well-known Reynolds operator for group-averaging functions (which projects a given function to the closest equivariant function in an $L_2$ sense, see \emph{e.g.}, \citep{elesedy2021provably}). Moreover, this formulation recovers the classical frame perspective when $\mathcal{F}$ always maps to a set containing exactly one group element. Regardless of the particular choice of $\mathcal{F}$, it is worth noting that the resultant equivariant pipelines is capable of resulting any equivariant function, so long as the generic architecture is itself universal.

The trade-offs between frames and equivariant architectures remain an active area of research. For example, \citet{pozdnyakov2023smooth} motivate frame-averaging as superior to choosing a single frame for rotational equivariance, by observing that methods which canonicalize point clouds to a single coordinate system are often not \emph{smooth}, in the sense that, adding or removing one point, or changing its position slightly, may drastically change the choice of coordinate system. Instead, they propose computing a weighted average over the frames defined by all pairs of neighbors of one central point, where the weights are specifically chosen to ensure smoothness. However, this procedure is computationally intensive. \citet{duval2023faenet} address the computational challenge of averaging over a smaller set of coordinate systems defined by principal component analysis, opting to randomly sample a coordinate system at each forward pass during training, sacrificing guaranteed train-time equivariance for efficiency. They demonstrate promising performance-time tradeoffs on materials science tasks. 
In light of the difficulty established by these two papers of finding a ``good'' coordinate frame, or set of frames, over which to average, one promising direction proposed by \citet{kaba2022learned} is to \emph{learn} the coordinate frame using a very lightweight equivariant architecture. \revisionTwo{\citet{lin2024equivariance} introduces minimal frame averaging, which lowers the frame cardinality $|\mathcal F(x)|$ to the minimum required, thereby enabling efficient integration of equivariance into existing model architectures.} Finally, several diverse and recent architectures can be interpreted as establishing local coordinate frames \citep{passaro2023reducing, pozdnyakov2023smooth}, including the structure module of AlphaFold2 \citep{jumper2021highly}, and applying the frame-based method for equivariance to local neighborhoods is a promising direction (as it encodes an inductive bias towards not just global, but also local, equivariance). 
\revisionTwo{Despite these studies over frames and equivariant architectures, frame averaging has its inherent limitations. For example, in symmetric graphs~\citep{cen2024high}, PCA-based frames~\citep{puny2021frame} break permutation equivariance and invariance, and in some domains the resulting frame for rotation or permutation group is discontinuous \citep{dym2024equivariant}. Nevertheless, model-agnostic frame averaging remains a general strategy for constructing efficient equivariant networks.} Going forward, frames may provide an appealing alternative to equivariant architectures in applications for which computational efficiency is paramount.






\subsubsection{Approximate Equivariance}\label{subsec:approx_equivariance}

\ifshowname\textcolor{red}{(Hannah)}\else\fi
Sometimes physical problems do not adhere exactly to group symmetries, but nonetheless symmetries provide a helpful approximation (\emph{e.g.}, if the ground-truth function is still \emph{close} to an equivariant function). Such so-called “approximate symmetries” can arise for a variety of reasons, including boundary effects, discretization error, or something more inherent to the problem, like a partial equivariance or symmetry-breaking property. For example, digit classification is invariant to small-angle rotations, but rotating a “6” yields a “9”, so the problem is not truly rotation-invariant. As a more scientific example, variations in the diffusion coefficient of plate may break the rotational isotropy of heat diffusion \cite{wang2022approximately}. In such problems, an inductive bias towards even approximate symmetry can still advantageously reduce the search space of neural nets. 

Residual Pathway Priors \citep{finzi2021residual} first suggested relaxing exact equivariance constraints by parametrizing the learnable weight matrices as sums of equivariant and unconstrained matrices, where the loss function ensures that equivariance is favored. On tasks in vision, synthetic dynamical systems, and reinforcement learning, they demonstrate that their approach is superior in settings with approximate symmetry, yet does not significantly degrade in cases with exact or no symmetry.

More recently, \citet{wang2022approximately} proposes a generalization of group CNNs (as well as steerable CNNs, the details of which we omit here but are analogous to the G-CNN case), as shown below: 
\begin{align*}
    &\text{Ordinary group-convolution: } (f \ast_G \phi)(g) = \sum_{h \in G} f(h) \phi(g^{-1}h). \\
    &\text{Relaxed group-convolution: }  (f \: \widetilde{\ast}_G \: \phi)(g) = \sum_{h \in G} f(h) \phi(g, h), \text{ where }\phi(g,h) := \sum_{l=1}^L w_l(h)\phi_l(g^{-1}h).\\
\end{align*}
Above, $f$ and $\phi$ are functions from $G$ to $\mathbb{R}^{c_{in}}$ and $\mathbb{R}^{c_{in} \times c_{out}}$, respectively. Intuitively, such formulations allow the convolutional filter to be location-dependent. The choice of parameter $L$, the number of filter banks, influences the extent to which the learned function can stray from full symmetry. To encourage symmetry, the network is initialized to ordinary group-convolution (which is a special case of relaxed group convolution), and a term in the loss function discourages variation in each $w_l$. 
They demonstrate superior performance on synthetic smoke plume and experimental jet flow datasets, relative to both perfectly equivariant and generic (not at all symmetric) architectures. Note that these tradeoffs are also justified theoretically in recent work \citep{petrache2023approximation}.

Others works have presented  alternative relaxations of group convolution. \citet{ouderaa2022relaxing} instead relax $\phi(g^{-1}h)$ very generally to $\phi(g^{-1}h, h)$, which they parameterize using a few tricks (the group's Lie algebra and Fourier features). 
\citet{romero2022partial} instead relax group convolutions by learning a non-uniform measure over the group. 

The previous pipelines were all motivated by, and tested on, data that only approximately adhered to a group symmetry. It is still an open question, however, whether these approximately equivariant networks will offer any long-term advantage over perfectly equivariant networks in tasks with a \emph{genuine} group symmetry. Spherical channel networks for point cloud data (discussed in Section \ref{subsec:cont_equi}), for example, achieved state of the art results on the Open Catalyst dataset at the time of their release, despite not having perfect rotation equivariance. However, they have since been surpassed by fully equivariant networks \citep{passaro2023reducing}. Nonetheless, for real-world data with noise, approximate symmetry, or even slightly misspecified symmetry, these approaches interpolate advantageously between strictly symmetric and unconstrained architectures.





\clearpage
\hypertarget{AI for Quantum Mechanics}{\section{AI for Quantum
Mechanics}} \label{sec:qt}

\begin{figure}[t]
    \centering
\includegraphics[width=\textwidth]{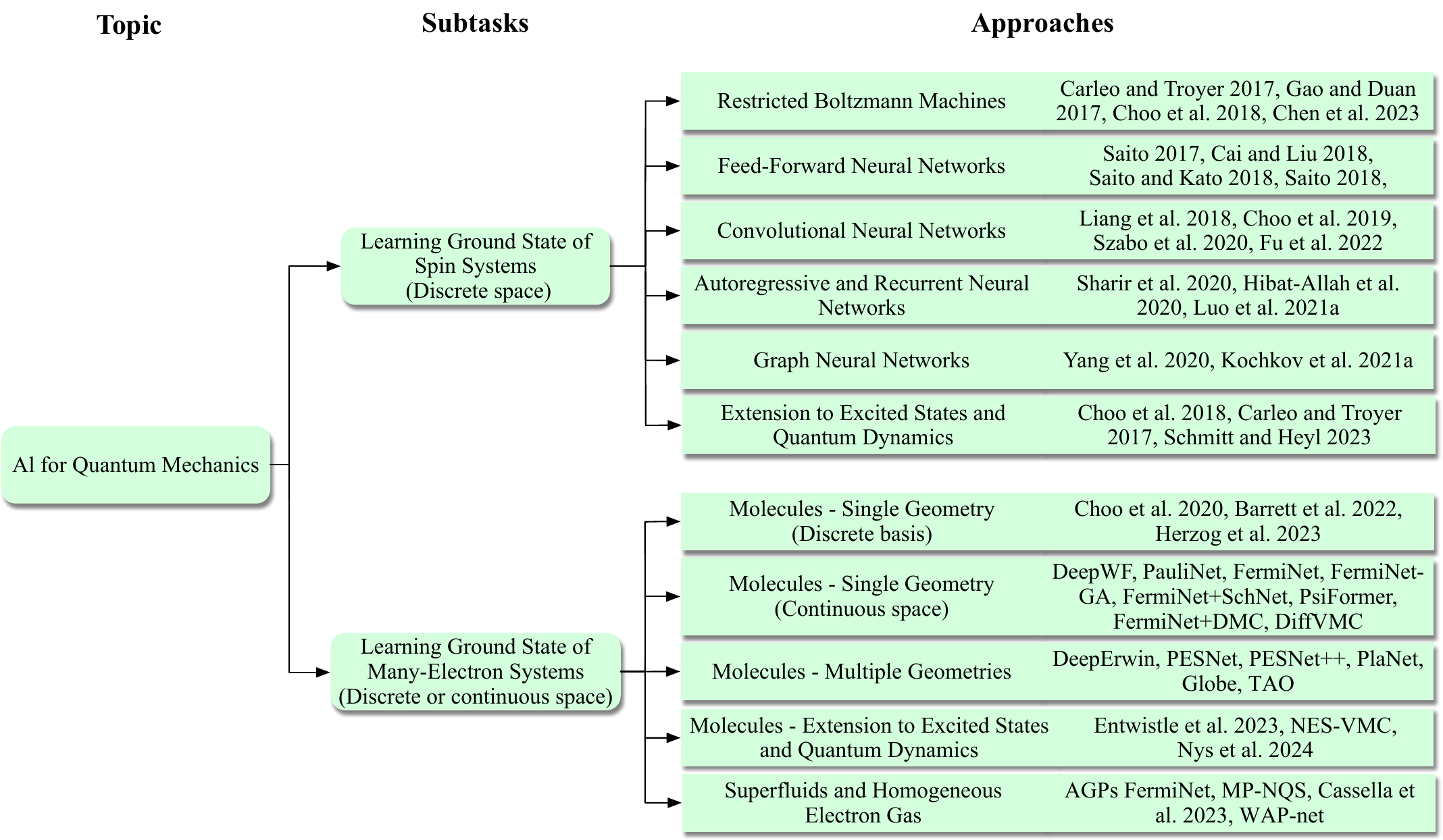}
    \caption{An overview of the tasks and methods in AI for quantum mechanics. In this section, we focus
on two subtasks including learning ground states of spin systems and learning ground states of many-electron systems. The methods for learning ground states of spin systems are grouped in terms of the category of the network they use to represent the quantum state. Specifically,~\citet{carleo2017solving},~\citet{gao2017efficient},~\citet{choo2018symmetries}, and~\citet{chen2023systematic} use restricted Boltzmann machines.~\citet{cai2018approximating},~\citet{saito2018machine},~\citet{saito2018method}, and~\citet{saito2017solving} use feed-forward neural networks.~\citet{liang2018solving},~\citet{choo2019two},~\citet{szabo2020neural}, and~\citet{fu2022lattice} use convolutional neural networks.~\citet{sharir2020deep},~\citet{hibat2020recurrent}, and~\citet{luo2021gauge} use autoregressive and recurrent neural networks.~\citet{yang2020scalable}, and~\citet{kochkov2021learning} use graph neural networks. 
For learning ground states of many-electron systems, one important application is molecules. One category of methods, including 
\citet{choo2020fermionic, barrett2022autoregressive,  herzog2023solving} aim to optimize single geometry of a molecule using discrete basis.
DeepWF~\citep{han2019solving} PauliNet~\citep{hermann2020deep}, FermiNet~\citep{pfau2020ab}, FermiNet-GA~\cite{lin2023explicitly}, FermiNet+SchNet~\citep{gerard2022goldstandard},
PsiFormer~\citep{glehn2023a}, FermiNet+DMC~\citep{ren2022towards, wilson2021simulations}, and DiffVMC~\citep{zhang2023score}, aim to optimize single geometry of a molecule in continuous space. Another category of methods, including DeepErwin~\citep{scherbela2022solving}, PESNet~\citep{gao2021ab}, PESNet++ \& PlaNet~\citep{gao2023samplingfree}, Globe~\citep{gao2023generalizing} ,and TAO~\citep{scherbela2023towards}, aim to optimize multiple geometries of the same molecule or even among different molecules simultaneously.
Beyond molecules, AGPs FermiNet~\citep{lou2023neural} is developed for superfluids. MP-NQS~\citep{pescia2023message},~\citet{cassella2023discovering}, and WAP-net~\citep{wilson2022wave} are developed for homogeneous electron gas.
\revisionOne{Beyond ground states, 
\citet{choo2018symmetries} study excited states for spins and bosons.
\citet{entwistle2023electronic, pfau2023natural} study excited states for molecules.
\citet{carleo2017solving, schmitt2020quantum} study quantum dynamics for spin systems, and \citet{nys2024ab} study quantum dynamics for molecules.
}
}
\label{fig:QM_overview}
\end{figure}

In this section, we provide technical reviews on how to design advanced deep learning methods to efficiently learn neural wavefunctions. In~\cref{sec:QM_Overview}, we give an overview of the definition and how to solve quantum many-body problems in general. In~\cref{sec:QM_spin}, we introduce methods of learning ground states for quantum spin systems. In~\cref{sec:QM_electron}, we introduce methods of learning ground states for many-electron systems. An overview of the tasks and representative methods is shown in~\cref{fig:QM_overview}. 

\subsection{Overview}
\label{sec:QM_Overview}
\noindent{\emph{Authors: Cong Fu, Xuan Zhang, Shenglong Xu, Shuiwang Ji}}\newline

\label{sec:QM problem}
Quantum mechanics is the branch of physics that describes the laws governing atoms and subatomic particles~\citep{feynman1965feynman}. It is of fundamental importance in explaining the physical phenomena of quantum systems in the microscopic domain, ranging from a single particle to molecules and materials~\cite{feynman2011,griffiths2018introduction,sakurai2020modern}. A quantum state contains all the information about a quantum system and is represented as a wavefunction $\ket{\psi}$. Given a set of variables describing the system, such as the position and momentum of its particles, as inputs, the wavefunction $\ket{\psi}$ outputs a complex number that represents the probability amplitude for each possible outcome of a measurement of the system. The wavefunction $\ket{\psi}$ is a high dimensional function that requires an exponential amount of information to fully define. Obtaining the wavefunction of a quantum system is a challenging problem known as the quantum many-body problem. The wavefunction $\ket{\psi}$ is governed by the Schrödinger equation
\begin{align}
    \hat{H} |\psi\rangle = E |\psi\rangle,
    \label{sec:qt eq:sch}
\end{align}
where $\hat{H}$ is the Hamiltonian operator that describes the motion and interaction of particles in the quantum system, and $E$ is the total energy of that system. In the discrete case, the Hamiltonian operator $\hat{H}$ can be represented as a Hamiltonian matrix $H$. In principle, all eigenvalues and eigenvectors of $H$ can be obtained through eigenvalue decomposition. Then, the smallest eigenvalue is the ground-state energy of the system, and the corresponding eigenvector is known as the ground state, which is the lowest-energy stationary state. At zero temperature, the ground state fully determines all the properties of the quantum system. Therefore, we focus on how to obtain the ground state of a given quantum system.

The dimension of the Hamiltonian matrix grows exponentially with the size of the quantum systems, such as the number of particles in the system. For instance, the Hamiltonian matrix has a size of $2^N \times 2^N$ for a spin \revisionOne{1/2} system with size $N$. Therefore, it is not feasible to obtain the ground state through direct eigendecomposition, even for relatively small systems. An alternative way to \textit{approximately} obtain the ground state and its energy is the variational principle. Consider a parameterized function $\ket{\psi(\bm{\theta})}$ that represents a quantum state, where $\bm{\theta}$ are learnable parameters. According to the variational principle, the energy of $\ket{\psi(\bm{\theta})}$ must be larger or equal to the ground state energy, which is the smallest eigenvalue of the Hamiltonian matrix. 
Consequently, to approximate the ground state by $\bm{\theta
}$, one can optimize the variational parameters $\bm{\theta}$ by minimizing the energy of the state. 
Formally, the expectation value of the energy can be written as
\begin{align}
    E(\theta) = \frac{\bra{\psi(\bm{\theta})}\hat{H}\ket{\psi(\bm{\theta})}}{\langle\psi(\bm{\theta})|\psi(\bm{\theta})\rangle}
    = \frac{\int |\psi(\bm{s} ;\bm{\theta})|^2 \frac{\hat{H}\psi(\bm{s};\bm{\theta})}{\psi(\bm{s};\bm{\theta})} d\bm{s}}{\int |\psi(\bm{s} ;\bm{\theta})|^2 d\bm{s}} \geq E_0,
    \label{eq: vmc}
\end{align}
where $E_0$ is the ground state energy and $E$ is the energy associate with the quantum state $\ket{\psi(\bm{\theta})}$. $\bra{\psi(\bm{\theta})}$ is the conjugate transpose of $\ket{\psi(\bm{\theta})}$, and $\langle\psi(\bm{\theta})|\psi(\bm{\theta})\rangle$ denotes the dot product of these two vectors.
The expectation value of the energy is the mean value of the quantity $\revisionOne{E_{loc}(\bm{s};\bm{\theta}) = }\hat H \psi(\bm{s};\bm{\theta})/\psi(\bm{s};\bm{\theta})$, denoted as the local energy, with respect to a probability distribution $p(\bm{s}) = \frac{|\psi(\bm{s} ;\bm{\theta})|^2 }{\int |\psi(\bm{s} ;\bm{\theta})|^2 d\bm{s}}$. 

\begin{figure}[t]
    \centering
    \includegraphics[width=0.85\textwidth]{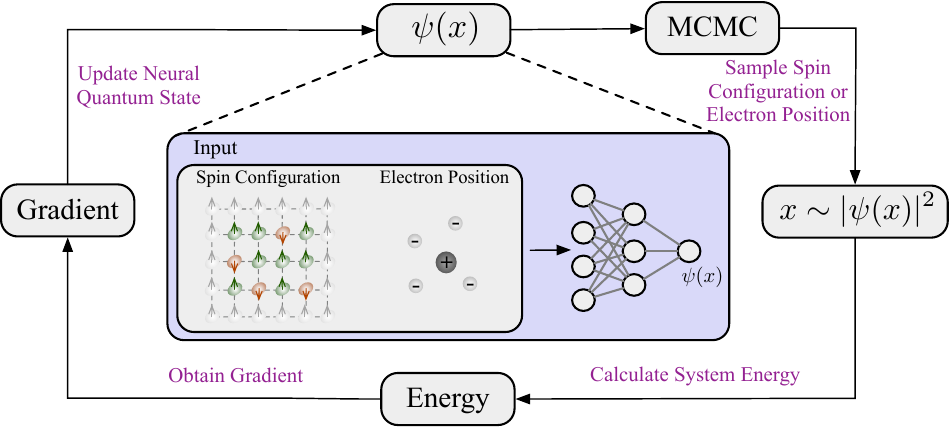}
    \caption{Pipeline of variational Monte Carlo (VMC). The neural quantum state takes as input a spin configuration or electron positions and outputs the wavefunction value. In VMC, spin configurations or electron positions are sampled using Markov chain Monte Carlo (MCMC) according to the probability distribution determined by the wavefunction. And then, energy is calculated from these samples, and the neural quantum state is updated by the gradient of the energy.}
    \label{fig:quantum_vmc}
\end{figure}

The mean value of $E_{loc}$ cannot be obtained exactly due to the high dimensionality of the probability distribution. Instead, one can approximate it by sampling the probability distribution using the Monte Carlo method. In addition, the gradient $\partial E /\partial \bm{\theta}$ can also be obtained through sampling and is used to optimize the parameters $\bm{\theta}$ to decrease the energy. This method combining the variational principle and Monte Carlo sampling is called variational Monte Carlo~(VMC), outlined in Figure~\ref{fig:quantum_vmc}.       

To sample input configurations according to the probability distribution $p(\bm{s})$, Metropolis-Hastings (MH) algorithm is used to create a Markov Chain of input configurations that converges to the stationary distribution $p$. Specifically, with an input configuration $\bm{s}$ on the Markov Chain, a new input configuration $\bm{s}'$ is proposed according to the proposal distribution $g(\bm{s}'|\bm{s})$. And then, $\bm{s}'$ is accepted or rejected according to the acceptance distribution $A(\bm{s}', \bm{s})$. Formally,
\begin{align}
    A(\bm{s}', \bm{s}) = \mathrm{min} \Bigl\{ 1, \frac{p(\bm{s}')g(\bm{s}|\bm{s}')}{p(\bm{s})g(\bm{s}'|\bm{s})} \Bigr\}.
\end{align}
If $\bm{s}'$ is rejected, the next input configuration on the Markov Chain is still $\bm{s}$. Once the Markov Chain converges to the stationary distribution, samples can be drawn from the Markov Chain, and they are ensured to satisfy the desired distribution.

After input configurations are sampled, we can approximate the system energy as the average of local energy, shown as below:
\begin{align}
    \label{eq:quantum_local_energy}
    E \approx \frac{1}{N} \sum_{i=1}^{N} E_{loc}^{(i)}.
\end{align}
Then we can 
optimize the variational parameters $\bm{\theta}$ to make the system energy as low as possible. Then, the optimized function $\ket{\psi(\bm{\theta})}$ with the lowest energy can be seen as a good approximation of the ground state. In~\cref{sec:QM_spin,sec:QM_electron}, we review methods that use neural networks to represent quantum states for learning ground states of quantum spin systems and many-electron systems. Even though we focus on reviewing methods of learning ground state, it is notable to mention that the variational principle can also be applied to the quantum field theory. For instance,~\citet{martyn2023variational} proposes the first neural network quantum field state for continuum quantum field theory.

\revisionOne{
Further, methods for learning ground states can also be extended to excited states and quantum dynamics.
For excited states we aim to find eigen wavefunctions of the Schrödinger equation with higher energies than ground states. 
For spin and boson systems, it has been studied in~\citet{choo2018symmetries} by leveraging the symmetry or projecting the wavefunction to be approximately orthogonal to the ground state wavefunction.
For molecules, it has been studied in~\citep{entwistle2023electronic} by adding a penalty term to the loss function which guides the optimization to converge to states that are orthogonal to ground states.
\citet{pfau2023natural} transform the problem of finding excited states into finding the ground state for an expanded system, where similar methodologies for finding ground states can be employed without explicitly enforcing the orthogonality.
Other related methods for molecular excited states are reviewed in~\citep{feldt2020excited}.
On the other hand, in quantum dynamics, we are interested in modeling the time evolution of wavefunctions.
Instead of targeting on finding the stationary solutions of the Schrödinger equation as in learning ground states or excited states, quantum dynamics aims to solve the more general time-dependent Schrödinger equation, where the learnt parameters of a neural network wavefunction can be updated in time through the time-dependent variational principle.
For spin problems, \citet{carleo2017solving} study variational dynamics with neural network quantum states for 1-dimensional systems using RBMs. \citet{schmitt2020quantum} study 2-dimensional quantum dynamics with CNNs.
In continuous space, quantum dynamics was first studied for rotors in ~\citet{medvidovic2023variational}, and more recently for molecules in~\citet{nys2024ab}.}


\subsection{Learning Ground States for Quantum Spin Systems}
\label{sec:QM_spin}
\noindent{\emph{Authors: Cong Fu, Shuiwang Ji}}\newline

A quantum spin model is a many-body model that describes interacting spins on a lattice resulting from spins of electrons tightly bound to atoms. These spin interactions can result in various magnetic ground states of the system, 
such as ferromagnetism, anti-ferromagnetism, and even spin liquid, which is an exotic magnetic state that holds promise for topological quantum computing. Understanding the ground state of the quantum spin model provides valuable insight into magnetic materials that are integral to modern technology.

\subsubsection{Problem Setup} In a quantum spin system, each spin can be in two states, spin-up $\uparrow$, spin-down $\downarrow$, or their superposition. Any quantum state of $N$ spins can be expressed as a superposition of $2^N$ spin configurations. All the combinations of spins constitute a computational basis. Specifically, a quantum state can be written as
\begin{align}
    |\psi\rangle = \sum_i^{2^N} \psi(\bm{\sigma}^{(i)}) |\bm{\sigma}^{(i)}\rangle,
\end{align}
where $\ket{\bm{\sigma}^{(i)}}$ represents an array of spin configurations of $N$ spins, \emph{e.g.}, $\uparrow \uparrow \downarrow \cdots \downarrow$, and $\psi(\bm{\sigma}^{(i)})$ is the wavefunction value for the spin configuration $\ket{\bm{\sigma}^{(i)}}$. The goal is to use neural networks to parameterize the wavefunction and obtain the ground state wavefunction using the variational Monte Carlo described in~\cref{sec:QM_Overview}.

\subsubsection{Technical Challenges} 

Learning the ground states of quantum spin systems faces several key challenges, including incorporating symmetries of the wavefunction, learning ground state sign structures, and extending approaches to diverse lattice geometries.

\vspace{0.1cm}\noindent\textbf{Preserving Symmetries:} In spin systems, the learned ground state should satisfy certain symmetric structures. Quantum spin systems exhibit rich and intriguing symmetries that are not present in traditional deep learning tasks, such as image object detection. Different from images, lattices are periodic grids with additional symmetries, such as rotations and reflections, which can be classified into 17 wallpaper groups, namely, 17 different plane symmetry groups that make various planar patterns invariant to the corresponding transformations. While most powerful neural networks can learn these symmetries automatically from data according to the universal approximation theorem, this is often hard to achieve due to the enormous solution space and the difficulty of optimization. Incorporating symmetries of the ground state into the neural network structure can guarantee the symmetries of the learned ground state and improve the data efficiency and facilitate finding the optimal solution. 

\vspace{0.1cm}\noindent\textbf{Learning Sign Structures:} In quantum mechanics, the sign structure of a wavefunction, in general, refers to the phase of the complex probability amplitude associated with a quantum state. It is challenging to learn the accurate sign structure of the ground state. Sometimes the ground state of quantum spin systems exhibits severe sign problems, where small changes in the spin configuration can cause a change in the sign of the wavefunction, making it difficult for neural quantum states to converge. This phenomenon is even more severe in a frustrated regime and makes it challenging for neural networks to capture complex sign structures of the ground state.

\vspace{0.1cm}\noindent\textbf{Multiple Geometries:} Most existing methods only work for 1D chains or 2D square lattices. However, the lattice geometry of a magnetic material can be far richer than a simple square lattice and has significant effects on its ground state and thus its magnetic properties. The resulting magnetic frustration from this rich geometry provides a host for more exotic magnetic properties to emerge. Therefore, it is crucial to extend the neural network to handle various lattice geometries. 

\subsubsection{Existing Methods} 
Neural quantum states (NQS) have emerged as a powerful variational ansatz for approximating the ground states of quantum many-body systems. NQS can be classified into five different categories based on the type of neural networks, as shown in~\cref{fig:QM_overview}.~\citet{carleo2017solving} propose a pioneering work that uses restricted Boltzmann machine (RBM) to represent quantum states. Due to the success of using RBM as a variational ansatz~\citep{gao2017efficient, deng2017quantum, chen2018equivalence, choo2018symmetries, chen2023systematic}, researchers start exploring more expressive deep learning methods to represent quantum states, such as feed-forward neural networks~\citep{cai2018approximating, saito2018machine, saito2018method, saito2017solving}. Later on, convolutional neural networks (CNNs)~\citep{liang2018solving, choo2019two, szabo2020neural} are applied to 2D square lattices and are found to represent highly entangled systems effectively. However, CNN cannot be naturally used on non-grid lattices or even random graphs, which necessitated the exploration of graph neural networks (GNNs)~\citep{yang2020scalable, kochkov2021learning} for dealing with arbitrary geometric lattices. Moreover, autoregressive and recurrent neural networks (RNNs) are applied to represent quantum states, enabling direct sampling of spin configurations~\citep{sharir2020deep, hibat2020recurrent, luo2021gauge}.

\begin{table}[t]
	\centering
	\caption{Summary of different works on how to address several challenges in solving quantum many-body problems for spin systems, including incorporating symmetries of wavefunctions, learning sign structures, and processing multiple geometries. To consider symmetric ground state structures, solutions include averaging output over transformed inputs according to symmetries or using group convolution. For learning sign structures, solutions include using complex-valued networks to implicitly consider phase, separately modeling amplitude and phase, or incorporating the known Marshall sign rule as the reference sign structure in some special cases. For application on multiple geometries, solutions include processing random graphs and various lattice geometries.}
        \resizebox{1\textwidth}{!}{
	\begin{tabular}{lcc}
		\toprule[1pt]
		  Challenges & Solutions & Methods \\
		\midrule
            \multirow{3}{*}{Symmetry} & \multirow{2}{*}{Averaging} & \cite{nomura2005dirac} \cite{nomura2021helping} \cite{ferrari2019neural}\\
            & & \cite{choo2018symmetries} \cite{choo2019two} \cite{chen2023systematic}\\
            & Group Convolution & \cite{roth2021group}\\
            \cmidrule{1-3}
            \multirow{3}{*}{Sign Structure} & Complex-Valued & \cite{carleo2017solving} \cite{choo2019two} \cite{sharir2020deep}\\
            & Separate Modeling & \cite{cai2018approximating} \cite{szabo2020neural} \cite{kochkov2021learning} \cite{fu2022lattice}\\
            & Marshall Sign Rule & \cite{choo2019two}\\
            \cmidrule{1-3}
            \multirow{2}{*}{Multiple Geometries} & Random Graphs & \cite{yang2020scalable} \cite{kochkov2021learning}\\
            & Various Lattices & \cite{roth2021group} \cite{fu2022lattice}\\
		\bottomrule
	\end{tabular}
        }  
	\label{tab:NQS challenge}
\end{table}

In addition to the different neural network types that affect the expressiveness of neural quantum states, various methods also focus on addressing some of the challenges mentioned above, as shown in Table~\ref{tab:NQS challenge}. Incorporating the symmetries of ground states in neural networks can help reduce the hypothesis space. Effectively capturing sign structures of wavefunctions is crucial for neural quantum states to converge easily to optimal solutions. Moreover, the development of single neural quantum states that can function across multiple lattices could significantly enhance their practical usefulness and versatility.

\vspace{0.1cm}\noindent\textbf{Preserving Symmetries:} To capture the symmetry of ground states, most work~\citep{nomura2005dirac, nomura2021helping, ferrari2019neural, choo2018symmetries, choo2019two, chen2023systematic} use the symmetry-averaging technique, which involves transforming the input according to the symmetry group transformation and then taking the average of each output as the final predicted ground state value. Another approach to preserve symmetry is to use group equivariant convolution proposed in~\citep{cohen2016group}. In GCNN~\citep{roth2021group}, authors propose a general framework to use group equivariant convolution to consider the full wallpaper groups and demonstrate the effectiveness of GCNN on square and triangular lattices. GCNN can be mapped to symmetry-averaging models by masking some filters between hidden layers. It is also worthwhile to mention that many quantum many-body systems feature local gauge invariance. To preserve gauge symmetries,~\citet{luo2021gauge} proposes a gauge equivariant neural network quantum state for both abelian and non-abelian discrete gauge group.~\citet{luo2022gauge} designs gauge equivariant neural quantum state for abelian continuous gauge group.~\citet{chen2022simulating} develops Gauge-Fermion FlowNet that simultaneously fulfills fermionic symmetry and gauge symmetry.

\noindent\textbf{Learning Sign Structures: }In addition to capturing amplitudes of the ground state, sign structure is also crucial to be learned. Some works learn the amplitude and phases jointly by using a single neural network with complex-valued parameters~\citep{carleo2017solving, choo2019two, sharir2020deep}.~\citet{choo2019two} uses Marshall sign rule as a reference sign structure and incorporate it into the network design. The Marshall sign rule provides a simple sign structure that is known for bipartite graphs in some extremal limits, such as $J_1=0$ or $J_2=0$ for the $J_1-J_2$ Heisenberg model. However, for ground states in more complex frustrated regimes, there's no such simple prior sign structure to use.~\citet{cai2018approximating} modifies the feed-forward neural network into two branches to separately predict amplitude and sign of ground states, which are then multiplied together. They use the cosine function as the activation function for predicting the sign, which is suitable for capturing the oscillating features of the input spins.~\citet{kochkov2021learning} predicts log amplitude and phase of wave functions separately and shows that predicting phase directly enables effective generalization of the learned sign structure.~\citet{szabo2020neural} models amplitude and sign structure using two real-valued neural networks. Specifically, they compute the global phase by summing over predicted phasors for each local spin. Additionally, they adopt a two-stage optimization approach. First, they keep the amplitude of wave functions of all the spin configurations to be the same and only optimize the phase to minimize the system energy. This stage could provide a good initial sign structure since optimal sign structures depend weakly on amplitudes~\citep{szabo2020neural, marshall1955antiferromagnetism}. And then, the sign structure and amplitude are optimized simultaneously during the second stage. 

\vspace{0.1cm}\noindent\textbf{Multiple Geometries: } Most work mentioned above only use the square lattice as the test bed. A practical useful wavefunction ansatz should be applicable and work well across different lattice geometries. GNA~\citep{yang2020scalable} proposes universal wavefunction ansatz and conducts experiments on hard-core Boson systems over 2D Kagome lattices, triangular lattices, and randomly connected graphs.~\citet{kochkov2021learning} designs another GNN-based ansatz that uses sublattice encoding to denote the node’s location in a unit cell that respects the lattice symmetries. In addition to using GNN to achieve applicability on arbitrary lattices, LCN~\citep{fu2022lattice} proposes lattice convolution that uses virtual vertices to augment original lattices to transform them into square lattices, so that a regular CNN can be applied.

\subsubsection{Optimization Methods}
\label{sec:qm_disc_optim}
There are several ways to optimize the neural network quantum state. The straightforward approach is to calculate the system energy directly as the loss function and use gradient descent methods in deep learning, such as SGD and Adam, to update the network parameters~\citep{roth2021group, fu2022lattice}. The energy gradient is given as
\begin{align}\label{eq:quantum_spin_gradient}
   \Delta E_k = \langle E_{loc} O_k^*\rangle - \langle E_{loc} \rangle \langle O_k^* \rangle,
\end{align}
where $O_k = \frac{\partial log\psi(\bm{\sigma};\bm{\theta})}{\partial \theta_k}$ is the variational derivative with respect to the $k$-th network parameter, and $O_k^*$ is the complex conjugate of $O_k$. And $E_{loc} = \sum_{j} H_{ij} \frac{\psi(\bm{\sigma}^{(j)}; \bm{\theta})}{\psi(\bm{\sigma}^{(i)}; \bm{\theta})}$ is the local energy with respect to spin configuration $\bm{\sigma}^{(i)}$. $\langle \cdot \rangle$ denotes the expectation value over all the sampled spin configurations. 
To sample spin configurations from the desired probability distribution $p(\bm{\sigma}^{(i)}) = \frac{|\psi(\bm{\sigma}^{(i)})|^{2}}{\sum_{i}|\psi(\bm{\sigma}^{(i)})|^{2}}$ that is defined by the wavefunction $\psi$, we can use the Markov Chain Monte Carlo (MCMC) method described in~\cref{sec:QM_Overview}. For instance, if we consider a spin system governed by the Ising model, the proposed spin configuration in Markov Chain can be obtained by randomly flipping a spin in a lattice. So the proposal probability is symmetric, such that $g(\bm{\sigma}'|\bm{\sigma}) = g(\bm{\sigma}|\bm{\sigma}')$. Thus, the acceptance probability can be simplified in~\cref{eq: simplified_acceptence}. For other systems, we need to use a more general sampling method, and the Hasting correction is often used.
\begin{align}
    A(\bm{\sigma}', \bm{\sigma}) = \mathrm{min} \Bigl\{ 1, \frac{p(\bm{\sigma}')}{p(\bm{\sigma})} \Bigr\}.
    \label{eq: simplified_acceptence}
\end{align}

Another approach to optimize the neural network quantum state is to use stochastic reconfiguration (SR)~\citep{sorella2007weak} that represents an imaginary-time evolution process in the variational space. When a quantum state undergoes imaginary time evolution, it eventually converges to the ground state of the system. In stochastic reconfiguration, network parameters are updated as
\begin{align}\label{eq:quantum_sr}
    \bm{\theta} \leftarrow \bm{\theta} - \eta S^{-1} \bm{\Delta E}, 
\end{align}
where $\eta$ is the learning rate, $\bm{\Delta E}$ is the energy gradient, and $S_{ij} = \langle O_i^* O_j\rangle - \langle O_i^* \rangle \langle O_j \rangle$. The only difference from the gradient descent is the presence of a covariance matrix $S$. Stochastic reconfiguration is generally more robust and less sensitive to the learning rate. However, if we directly evaluate~\cref{eq:quantum_sr}, the limitation is that the size of matrix $S$ equals the number of neural network parameters, making it computationally expensive to compute its inverse for neural networks with large parameters. To reduce the complexity of SR, people often use iterative solvers, such as conjugated gradient (CG), to make the complexity of SR become linear in the number of parameters~\citep{neuscamman2012optimizing}. This is also routinely used in NetKet~\citep{vicentini2112netket}, a machine learning toolbox for quantum physics. \revisionOne{Recently, MinSR and related methods~\citep{chen2023efficient, rende2024simple} reformulate the $S$ matrix to have the size of the number of the sampled configurations, which is significantly smaller than the number of parameters for modern deep neural networks. Such approaches allow a scaling of SR linear with the number of variational parameters and are especially useful in the regime of small batch dimension.} Another alternative optimization method proposed by~\citet{kochkov2018variational} that can overcome the limitation of SR is imaginary time supervised wavefunction optimization (IT-SWO). IT-SWO interpolates between the energy gradient and stochastic reconfiguration methods~\citep{kochkov2021learning}. It optimizes the wavefunction ansatz to maximize the overlap between the current variational state $\psi(\bm{\sigma};\bm{w})$ and the imaginary-time evolved state $(I - \beta H)\psi(\bm{\sigma};\bm{r})$, where $\bm{w}$ and $\bm{r}$ represent the parameters of current state and the state at the end of last optimization iteration. At each optimization iteration, IT-SWO first updates the target state and keep it fixed during the current iteration, and then performs multiple inner steps with stochastic gradient descent to update the current state.  

\subsubsection{Datasets and Benchmarks}\label{sec:quantum_spin_data}
In contrast to traditional machine learning tasks, models used to determine the ground state of quantum spin systems cannot be trained on a pre-existing dataset. Instead, the model is trained for a specific quantum spin system, which is defined by the lattice and Hamiltonian. During each step of the training process, data are dynamically sampled from the wavefunction (neural network) of a quantum system. This approach is known as concurrent machine learning as described by~\citet{han2020integrating}. Typically, a variety of lattice systems are considered, such as square, honeycomb, triangular, and kagome lattices. The most commonly used Hamiltonian is $J_1\text{-}J_2$ quantum Heisenberg model, which is the prototypical model for studying the magnetic properties of quantum materials.~\citet{wu2023variational} also create variational benchmarks for quantum many-body problems. In terms of evaluation metrics, the energy of the systems usually serves as a measure of how closely the approximated ground state aligns with the true ground state. A lower energy indicates a more accurate approximation.

\subsubsection{Open Research Directions}
Neural network quantum states have shown promise in representing the ground state of quantum spin systems. but several challenges still need further exploration. 
First, designing neural wavefunctions with provable sufficient expressiveness remains an open problem, especially for quantum systems exhibiting highly frustrated regimes and strong correlations. Second, a comprehensive benchmark that can consistently assess different methods on different quantum systems is highly needed, and the work by~\citet{wu2023variational} is an endeavor in this direction.
Finally, in variational Monte Carlo, Markov Chain Monte Carlo (MCMC) is commonly used to sample spin configurations from the probability distribution determined by wavefunctions and then calculate the system energy. However, performing exact sampling with MCMC is difficult, and the samples may still exhibit correlations, leading to inaccurate energy estimations. Usually, to decrease autocorrelation between samples, N annealing MCMC steps are added between two samples, where N represents the size of the system. However, this makes the sampling process computationally expensive for larger systems. A potential solution to this challenge is proposed in~\citep{sharir2020deep}. They use an autoregressive model to represent quantum states, which bypass the MCMC sampling and can support more efficient and exact sampling.

\subsection{Learning Ground States for Many-Electron Systems}
\label{sec:QM_electron}
\noindent{\emph{Authors: Xuan Zhang, Nicholas Gao, Stephan Günnemann, Shuiwang Ji}}\newline

Another important application of neural wavefunctions is to model many-electron systems such as molecules.  
Studying many-electron systems is at the core of quantum chemistry, where properties of molecules are directly computed from first principles based on quantum physics. Specifically, accurately describing the ground states of molecules is of great interest because the ground state determines the most stable state of a molecule and is important to the understanding of its structural and chemical properties.
Compared to quantum spin systems, the spin of the electrons does not occur in the Hamiltonian and, thus, can be fixed a priori~\citep{foulkes2001quantum}. As a result, the wavefunction only acts on the spatial coordinates in $\R^3$ of each of the $N$ electrons. 
Additionally, since electrons can move freely in space, the input space of the neural wavefunction becomes a continuous space.
Nevertheless, the search space for suitable wavefunctions still grows exponentially with the number of electrons $N$.
Moreover, the fermionic nature of electrons significantly increases the difficulty of the problem~\citep{ceperley1991fermion}, due to which an additional antisymmetry constraint must be satisfied by the wavefunctions. For example, it has been shown that solving the sign problem, which can arise for fermions due to Pauli exclusion, is NP-hard~\citep{troyer2005computational} for certain related but different quantum Monte Carlo problems.

Although the wavefunction becomes continuous, in quantum chemistry it is common to approximate a wavefunction as a linear combination of a set of basis functions so that the wavefunction can be represented as coefficients of the basis functions.
When such a discrete (and antisymmetric) basis set is used, the same formalism in~\cref{sec:QM_spin} can be applied. These methods are called second-quantization methods and have been successfully applied to molecules~\citep{
choo2020fermionic, barrett2022autoregressive,  herzog2023solving}.
Alternatively we can work directly with continuous-space wavefunctions. These methods are called first quantization methods and have gained popularity recently because of their flexibility beyond the choice of the basis set as well as the good performance that they demonstrated.
More detailed comparison between the first and second quantization can be found in~\citet{hermann2022ab}.
In this section, we mainly discuss learning ground states of molecules with continuous-space neural wavefunctions to contrast with the methods in~\cref{sec:QM_spin}. However, it should be noted that the use of a discrete basis is the cornerstone of many electronic structure methods such as DFT, and many important concepts in continuous-space NQS have been proposed and routinely used in discrete-space, such as Slater determinants and the neural backflows~\citep{luo2019backflow}, which we will discuss in details later.

Other than molecules, similar methods have been applied to superfluid and homogeneous electron gas (HEG), which we will also review briefly for completeness.
Furthermore, to get a more complete description of many-electron systems, excited states~\citep{entwistle2023electronic, pfau2023natural, feldt2020excited} can also be studied using similar approaches as the ground states. However, the details are out of the scope of our discussion in this section.

\subsubsection{Problem Setup}\label{sec:many-electron-setup}
Molecules are composed of electrons and atomic nuclei. Within the Born-Oppenheimer approximation~\citep{born1927quantentheorie}, atomic nuclei are treated as fixed particles, hence quantum states are completely determined by electrons' spins and 3D coordinates. At ground states, spins of electrons can be determined by chemical rules, such as the Aufbau principle, the Hund's rule and the Pauli exclusion principle. Hence, we are able to define the wavefunction solely in terms of the electron coordinates. Formally, given $N^\uparrow$ electrons with spin-up, $N^\downarrow$ electrons with spin-down. The set of their 3D Cartesian coordinates is defined as $\bm{r} = [\bm{r}_1, \dots, \bm{r}_{N^\uparrow+N^\downarrow}]\in \mathbb{R}^{(N^\uparrow+N^\downarrow)\times 3}$, where the first $N^\uparrow$ electrons have spin-up and the last $N^\downarrow$ electrons have spin-down. A wavefunction $\psi: \mathbb{R}^{(N^\uparrow+N^\downarrow)\times3}\to \mathbb{R}$ maps the set of coordinates to a scalar value. In the continuous case, the Hamiltonian operator $\hat{H}:(\mathbb{R}^{(N^\uparrow+N^\downarrow)\times 3}\to \mathbb{R})\to(\mathbb{R}^{(N^\uparrow+N^\downarrow)\times 3}\to \mathbb{R})$ is a function that maps a wavefunction to another function, defining the energy of a molecule, and is defined as
\begin{equation}
    [\hat{H}\psi](\bm{r}) = -\sum_i \frac{1}{2}\nabla_i^2 \psi(\bm{r}) + V(\bm{r})\revisionOne{\psi(\bm{r})},
\end{equation}
where the first term represents the kinetic energy and the second term represents the Coulomb potential, which is defined as
\begin{equation}   
    V(\bm{r}) = \sum_{i<j} \frac{1}{\Vert\bm{r}_i-\bm{r}_j\Vert_2} - \sum_{i,I} \frac{z_I}{\Vert\bm{r}_i-\bm{c}_I\Vert_2} + \sum_{I<J}\frac{1}{\Vert\bm{c}_I-\bm{c}_J\Vert_2},
\end{equation}
where $\bm{c}_I$ denotes the coordinate of an atomic nucleus and $z_I$ denotes its atomic charge. The terms define Coulomb potential between electron-electron pairs, electron-atom pairs, and atom-atom pairs, respectively. Note that although in general a wavefunction is complex-valued, we can work with real-valued wavefunctions here since the Hamiltonian is Hermitian, and therefore its eigenvalues and eigenfunctions are real-valued. Additionally, it is noteworthy that the Hamiltonian does not depend on electron spins. Thus, one can fix the spins a priori. Given the Hamiltonian, the local energy $E_{loc}(\bm{r}) = \frac{[\hat{H}\psi] (\bm{r})}{\psi(\bm{r})}$ (as introduced in~\cref{sec:QM_Overview}) can be expressed as
\begin{equation}
    E_{loc}(\bm{r}) = -\frac{1}{2} \sum_i^{(N^\uparrow+N^\downarrow)\times 3} \left[{\partial_i^2 \log|\psi(\bm{r})|} + \left(\partial_i \log|\psi(\bm{r})|\right)^2 \right] + V(\bm{r}),
\end{equation}
where $i$ goes through all $(N^\uparrow+N^\downarrow)\times 3$ spatial coordinates.

A fundamental constraint for a many-electron system is that its wavefunction must be antisymmetric upon permutation of two electrons with the same spin, a concept originating from Pauli exclusion.
In quantum mechanics, exchanging two indistinguishable particles does not affect the probability density of particles. In our case, two electrons cannot be distinguished if they have the same spin. Hence, $\psi(\dots, \bm{r}_i, \dots, \bm{r}_j,\dots)^2 = \psi(\dots, \bm{r}_j, \dots, \bm{r}_i,\dots)^2$, for any $(i,j)$ with same spins. 
Further, indistinguishable particles are classified into bosons, such as photons, and fermions, such as electrons, according to their exchange symmetry~\citep{feynman1965feynman}, which refers to whether the wavefunction $\psi$ remains unchanged or changes sign upon exchanging the positions of two particles.
Electrons are fermions, so the wavefunction must be antisymmetric upon permutation of two electrons with the same spin.
, \emph{i.e.}, $\psi(\dots, \bm{r}_i, \dots, \bm{r}_j,\dots) = -\psi(\dots, \bm{r}_j, \dots, \bm{r}_i,\dots)$.
This antisymmetry property leads to the Fermi-Dirac statistics in particle distributions, and thus fundamentally changes the behavior of fermions. 
Consequently, the task of finding ground states can be formulated as a constrained optimization problem. In the context of variational Monte Carlo, the wavefunction $\psi$ is approximated by a parametrized class of functions $\psi_\theta$. In this case, learning ground states is equivalent to the following optimization problem:
\begin{align}
    &\psi_{\bm\theta}: \mathbb{R}^{(N^\uparrow+N^\downarrow)\times 3} \to \mathbb{R}\\
    \min_{\bm{\theta}} \quad & \mathbb{E}_{p_{\bm\theta}} [E_{loc}(\bm{r}; \bm\theta)], \quad p_{\bm\theta} \propto \psi_{\bm\theta}^2 \label{eq:many_elec_optim_obj}\\
    \textrm{s.t.} \quad & \psi_{\bm\theta}(\dots, \bm{r}_i, \dots, \bm{r}_j,\dots) = -\psi_{\bm\theta}(\dots, \bm{r}_j, \dots, \bm{r}_i,\dots), \text{ where} \label{eq:fermion_anti}\\
    & \text{pair } (i,j) \text{ satisfies } 1\leq i,j \leq N^\uparrow \text{ or }  N^\uparrow + 1 \leq i,j \leq N^\uparrow +N^\downarrow. \nonumber
\end{align}
In the above optimization objective, the energy expectation is calculated as an average over samples obtained using Monte Carlo sampling. By the variational principle mentioned previously (~\cref{eq: vmc}), the energy expectation of any wavefunction is guaranteed to be larger than the ground state energy, and the lower bound is attained when $\psi_\theta$ converges to the ground state wavefunction.

\ifshowname\textcolor{red}{(Nicholas)}\else\fi
The above formulation provides a framework for obtaining the ground-state energy for a single molecule. In this section, we additionally consider the setting for jointly optimizing for multiple geometries. For example, we are usually interested in studying the change in energy based on structural changes in a molecule. Joint optimization improves computational efficiency by eliminating the need to optimize again for every nucleus configuration.
Formally, we define the potential energy surface (PES) as a function $E:\mathbb{M}\rightarrow \R$, where 
$\mathbb{M}=\{M=\{\bm{c}_i,z_i\}_{i=1}^{\vert M\vert}, \mathbf{\bm{c}_i}\in \mathbb{R}^3, z_i\in\mathbb{Z}\}$
is the set of possible molecules, maps from the molecular structure (coordinates and charges of nuclei) to the energy.
Classically, to obtain a PES, one needs to repeat single structure calculation multiple times.
The advent of neural network-based solution makes it possible to model the ab-initio solutions for PES with a single model. 
Concretely, in this setting, one is interested in finding a wavefunction $\psi_{\bm\theta}: \R^{(N^\uparrow+N^\downarrow)\times 3} \times \mathbb{M} \rightarrow \R$, where the wavefunction is now also dependent on the molecular structure, in addition to the electron coordinates.
Following~\citet{gao2023generalizing}, we call such $\psi_{\bm\theta}$ the \emph{generalized wavefunction}.
Note that this formulation should not be confused with the Schrödinger equation without the Born-Oppenheimer approximation where the nuclei are treated as waves and are thus considered as part of the wavefunction. This is not the case here, \revisionOne{as} we still only model the electronic wavefunction but condition the wavefunction on the molecular structure. 
Finally, using the generalized wavefuncion, the potential energy surface can be derived as $E(M)=\int\psi_{\bm\theta}(\bm{r}, M)\hat{H}_M\psi_{\bm\theta}(\bm{r}, M) d\bm{r}$ where $\hat{H}_M$ refers to the Hamiltonian of molecule $M$.

\ifshowname\textcolor{red}{(Xuan)}\else\fi
Additionally, superfluids and homogeneous electron gas can also be modeled as fermions in continuous space. However, these problems do not involve nuclei and use a different potential energy in the Hamiltonian. Another major difference is that these problems are periodic in space. Nevertheless, the general approaches for the single-molecule setting can still be applied.


\subsubsection{Technical Challenges}\label{sec:many-electron-challenge}
There are several challenges related to finding many-electron ground states with QMC, including satisfying the fermion antisymmetry constraint, designing expressive neural networks for individual electrons (orbitals), achieving good optimization, and effectively learning generalized wavefunctions for multiple geometries to improve computational efficiency.

\vspace{0.1cm}\noindent\textbf{Fermion Antisymmetry: }
As introduced in~\cref{sec:many-electron-setup},  fermion antisymmetry is a hard constraint imposed by quantum physics, and a neural wavefunction for electrons must adhere to it strictly. Failing to encode the antisymmetry constraint will void the variational guarantees and result in unphysically lower energies. Although deep neural networks can approximate arbitrarily complex functions, imposing such hard constraints poses a unique challenge. 

\vspace{0.1cm}\noindent\textbf{Orbital modeling: }
Electrons interact with each other via the Coulomb potential and Pauli exclusion, which can result in highly non-linear landscapes in wavefunctions. Therefore, the networks must have a strong capacity to model the wavefunction of each electron (dubbed as orbitals) while accounting for the interactions with other electrons.
Additionally, quantum physics gives us some prior knowledge of the system, which may be hard to model directly with neural networks. Thus, incorporating physics knowledge in orbital modeling is important to obtain solutions that respect physics.

\vspace{0.1cm}\noindent\textbf{Optimization: }
Although in principle we can get arbitrary approximation accuracy with VMC, it is challenging to achieve effective optimization of neural wavefunctions towards ground states. 
This is in part due to high accuracy requirement of the problem. The chemical accuracy is defined as 1 kcal/mol (1.594 mE\textsubscript{h} or 0.043 eV)~\citep{pfau2020ab}, which is very small compared to the total energy. For example, the $N_2$ molecules, the error in energy estimation must be lower than 0.2\% to be useful for chemical applications~\citep{gerard2022goldstandard}. Consequently, effective optimization methods are crucial to obtain accurate and stable optimization.

\vspace{0.1cm}\noindent\textbf{Multiple Geometries: }
\ifshowname\textcolor{red}{(Nicholas)}\else\fi 
There are some unique challenges related to the multiple geometries setting.
Firstly, special considerations are necessary to make the learned wavefunctions adaptable to various molecular configurations, including different nucleus positions and variable numbers of nuclei and electrons (\emph{e.g.}, ionic systems) while respecting the fermion antisymmetry.
Secondly, as the PES $E$ is an observable metric, it is invariant to the Euclidean group $E(3)$, \emph{i.e.}, translations, rotations, and reflections of the molecule, as well as the permutation group $S_M$. 
However, as an abstract concept, the electronic wavefunction does not exhibit such symmetries. 
Thus, the challenge is to design generalized wavefunctions that result in invariant energies. 
It can be shown that to obtain such a behavior one needs to design symmetry-breaking covariant wavefunctions.
Thirdly, prior knowledge gives us additional constraints about limit behaviors. One such property is size-consistency, \emph{i.e.}, the energy of a duplicated non-interacting system is twice the energy of the single system. Implementing such behaviors into wavefunctions remains a challenge to reduce the function search space.
Lastly, while the wavefunction is directly linked to the energy, obtaining the energy from the wavefunction remains expensive as it requires numerical integration. Approximate inference methods promise to accelerate the process and enable high-resolution PES.

\subsubsection{Existing Methods}
Recently, VMC-based neural networks have shown strong ability in modeling ground states of many-electron systems. Classical methods such as DFT or CCSD(T) either result in unreliable results in strongly correlated settings, \emph{e.g.}, when bonds break, or scale unfavorably with the system size. VMC coupled with deep neural networks has shown to be able to outperform classical methods~\citep{pfau2020ab, gerard2022goldstandard}. 
While DFT is orders of magnitudes faster than deep VMC calculations, significantly higher accuracies can be obtained in VMC calculations thanks to the variational principle. Further, deep VMC offers clear path forward with advances in optimization and neural architecture while the exact form of the exchange correlation functional in DFT remains a mystery. Compared to accurate wave function theory like CCSD(T), deep VMC scales more favorably in theory ($\mathcal{O}(N^4)$ vs $\mathcal{O}(N^7)$). However, CCSD(T) typically runs faster on all reasonably accessible structures while often yielding lower relative energy errors. Nonetheless, deep VMC frequently succeeds in challenging multi-reference systems where CCSD(T) results are unreliable.
Moreover, although CCSD(T) can be applied to larger molecules, a smaller basis set must be picked. In the following, we briefly introduce how the challenges listed in Section~\ref{sec:many-electron-challenge} are resolved by existing methods. We first describe how to encode fermion antisymmetry in networks, in particular with Slater determinants. Next, we describe how networks are designed in existing methods. With these two components, we can already have a working neural wavefunction model. We then discuss how to effectively optimize the networks to reach ground states. Finally, we describe strategies to reuse and accelerate the computations via generalized wavefunctions. The challenges and existing methods are summarized in Table~\ref{tab:many_electron_challenge}.

\vspace{0.1cm}\noindent\textbf{Fermion Antisymmetry: }
To design wavefunctions that satisfy the fermion antisymmetry (Equation~(\ref{eq:fermion_anti})), a well-established method is the Slater determinant~\citep{slater1929theory}. The Slater determinant wavefunction is computed as the determinant of a matrix which is constructed by applying $N$ molecular orbital functions to each of the $N$ electrons so that each row of the matrix encodes one electron. The key motivation is that when two electrons are swapped, the two corresponding rows in the matrix are also swapped, so its determinant will change sign.
Formally, let $\bm{\phi}^\uparrow$ and $\bm{\phi}^\downarrow:\mathbb{R}^3\times \mathbb{R}^{(N^\uparrow+N^\downarrow)\times 3}\to \mathbb{R}^{N^\uparrow+N^\downarrow}$ be two single-orbital functions that map the 3D electron coordinates to a $(N^\uparrow+N^\downarrow)$-dimensional feature vector, where $\bm\phi^\uparrow$ is used to encode spin-up electrons and $\bm\phi^\downarrow$ is used to encode spin-down electrons.
$\bm\phi^\uparrow$ and $\bm\phi^\downarrow$ take an electron coordinate as well as all electron coordinates as input and produce a $(N^\uparrow+N^\downarrow)$-dimensional vector. The objective of single orbital functions is to generate an embedding for each electron and the information from all electrons is used to provide context information.
After encoding each electron with $\bm\phi^\uparrow$ or $\bm\phi^\downarrow$, a set of $N^\uparrow+N^\downarrow$ feature vectors is obtained, each containing $N^\uparrow+N^\downarrow$ elements. The features are stacked into a matrix where each row represents one electron. The Slater determinant wavefunction $\psi$ is then computed as the determinant of that matrix:
\begin{equation}
    \psi(\bm{r}) = \det\begin{bmatrix} \bm\phi^\uparrow(\bm{r}_1;\bm{r})^T\\ \vdots \\ \bm\phi^\uparrow(\bm{r}_{N^\uparrow};\bm{r})^T \\ \bm\phi^\downarrow(\bm{r}_{N^\uparrow+1};\bm{r})^T \\ \vdots \\ \bm\phi^\downarrow(\bm{r}_{N^\uparrow+N^\downarrow};\bm{r})^T \end{bmatrix}.
\end{equation}
Note that the orbital function for spin-up and spin-down electrons are different so that the antisymmetry is present only exchanged.

\begin{table}[t]
    \centering
    \caption{Summary of challenges and existing methods for learning many-electron ground states in continuous space formulation. For electrons, a special challenge arises from fermion antisymmetry imposed by quantum physics. Most existing wavefunction models solve it through Slater determinants but have different network designs to model orbital functions. Moreover, to make learning accurate and practical, it is crucial to achieve effective optimization. Finally, the diversity and flexibility of molecules require methods to handle multiple geometries to increase computational efficiency.}
    \small
    \begin{tabular}{@{}l c c c c@{}}\toprule
         Challenges & Fermion Antisymmetry & Orbital Modeling & Optimization & Multiple Geometries\\ \midrule
         Methods &
        \begin{tabular}[t]{@{}c@{}}
        \textbf{Main method:} \\
            Slater determinant \\  
        \textbf{Others:} \\  Pairwise construction  \\ \revisionOne{Hidden Fermions} \\ Explicit construction \\
        AGP \\
        
        \end{tabular} &
        \begin{tabular}[t]{@{}c@{}}
            PauliNet \\ FermiNet \\ FermiNet+SchNet \\ PsiFormer \\ Moon \\ WAP-net \\ MP-NQS
        \end{tabular} &
        \begin{tabular}[t]{@{}c@{}}
           \textbf{Framework:}   \\VMC \\ DMC \\ DiffVMC\\
           \textbf{Optimizer:} \\ KFAC \\ CG
        \end{tabular} &
        \begin{tabular}[t]{@{}c@{}} \vspace{-6pt} \\ DeepErwin \\ PESNet \\ PESNet++ \\
        PlaNet \\ Globe \\ TAO \\
        \end{tabular}\\\bottomrule
    \end{tabular}
    \label{tab:many_electron_challenge}
\end{table}

For example, when $N^\uparrow=2$ and $N^\downarrow=1$ (Li atom), fermion antisymmetry is ensured by Slater determinants when $\bm{r}_2$ and $\bm{r}_3$ are exchanged, as demonstrated below:
\begin{equation}
    \psi(\bm{r}_1, \bm{r}_2, \bm{r}_3) = \det \left[\begin{smallmatrix}\phi^\uparrow_1(\bm{r}_1)&\phi^\uparrow_2(\bm{r}_1)& \phi^\uparrow_3(\bm{r}_1)\\
    \phi^\uparrow_1(\bm{r}_2)&\phi^\uparrow_2(\bm{r}_2)& \phi^\uparrow_3(\bm{r}_2)\\
    \phi^\downarrow_1(\bm{r}_3)&\phi^\downarrow_3(\bm{r}_2)& \phi^\downarrow_3(\bm{r}_3)\\
    \end{smallmatrix}\right]
    =-
    \det \left[\begin{smallmatrix}\phi^\uparrow_1(\bm{r}_1)&\phi^\uparrow_2(\bm{r}_1)& \phi^\uparrow_3(\bm{r}_1) \\
    \phi^\uparrow_1(\bm{r}_3)&\phi^\uparrow_2(\bm{r}_3)& \phi^\uparrow_3(\bm{r}_3) \\
    \phi^\downarrow_1(\bm{r}_2)&\phi^\downarrow_2(\bm{r}_2)& \phi^\downarrow_3(\bm{r}_2) \\
    \end{smallmatrix}\right]
    =-\psi(\bm{r}_1, \bm{r}_3, \bm{r}_2).
\end{equation} 
To further increase expressiveness, multiple Slater determinants can be computed, each with a different set of orbital functions, and the final wavefunction is the linear combination of Slater determinants. When $k$ Slater determinants are used, letting $w_p\in \mathbb{R}$ be the weights, the final wavefunction is computed as:
\begin{equation}
    \psi(\bm{r}) = \sum_{p=1}^k w_p \det\begin{bmatrix} \vdots \\ \bm{\phi}^{\uparrow p}(\bm{r}_{i};\bm{r})^T \\ 
    \vdots \\ \bm{\phi}^{\downarrow p}(\bm{r}_{j};\bm{r})^T \\ \vdots \end{bmatrix}.
\end{equation}

Besides Slater determinants, there are other ways to achieve antisymmetry. DeepWF~\citep{han2019solving} and \citet{pang2022on} propose to enforce the antisymmetry to every electron pair. The wavefunction is defined in the form of $\prod_{i<j}(f(\bm{r}_i;\bm{r})-f(\bm{r}_j;\bm{r}))$ where $f$ outputs a scalar value. When a pair of ($\bm{r}_i$, $\bm{r}_j$) is swapped, the sign of the product will be changed. This strategy is shown to be a special case of the Slater determinant (it can be written as the determinant of a Vandermonde matrix~\citep{pang2022on}) but with less computational cost.
\revisionOne{The hidden fermions approach augments Slater determinants with virtual orbitals and virtual particles, and is first applied to discrete systems~\citep{robledo2022fermionic} and then to continuous systems (for modeling the wavefunction of atomic neuclei)~\citep{lovato2022hidden}.}
\citet{lin2023explicitly} generalizes the sum-of-product determinant computations by explicitly considering all possible permutations of electrons. The final wavefunction is the sum of results from all permutations, \emph{i.e.}, $\sum_\pi \text{sign}(\pi) g(\pi(\bm{r}))$, where $\pi$ iterates over all permutations $\text{sign}(\pi)$ gives the sign of each permutation, and $g$ is a function maps the permuted $\bm{r}$ to a scalar. This however leads to a factorial complexity. 
\ifshowname\textcolor{red}{(Nicholas)}\else\fi
Finally, antisymmetric geminal power (AGP) wavefunctions have shown great success in modeling superfluids with neural-network wavefunctions~\citep{lou2023neural}. There one uses pairwise orbital functions $\phi:\R^3\times\R^3\times\R^{(N^\uparrow \times N^\downarrow)\times 3}\rightarrow\R$, and constructs the wavefunction as $\psi(\bm{r})=\det\Phi$, where $\Phi_{i,j}=\phi(r_i, r_j, \bm{r}), i\in\{1, \hdots, N^\uparrow\}, j\in\{N^\uparrow+1, \hdots, N^\uparrow+N^\downarrow\}$.
\begin{figure}[t]
    \centering
    \includegraphics[width=\textwidth]{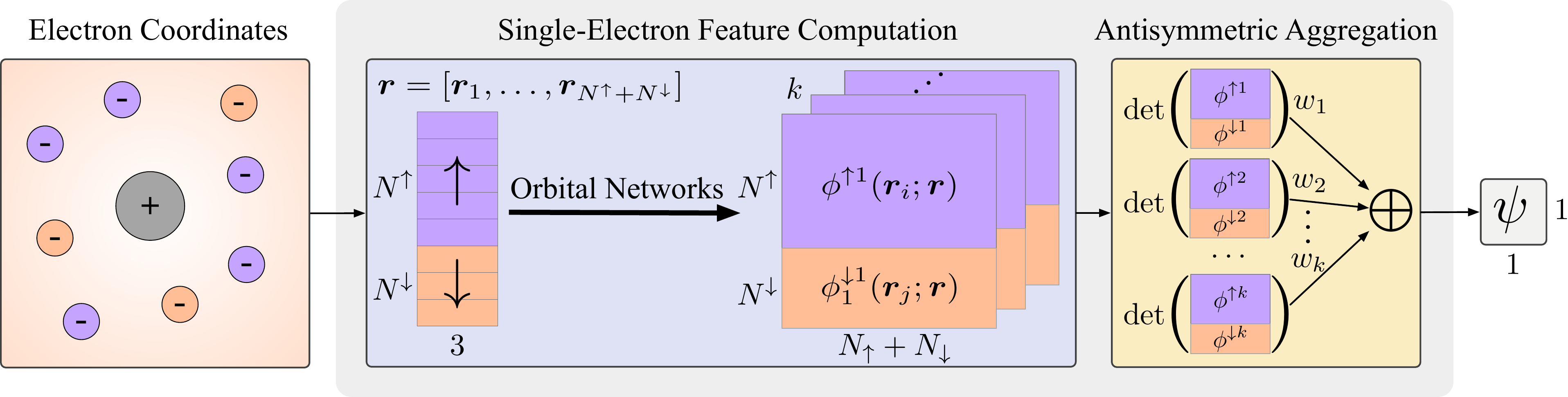}
    \caption{Pipeline of many-electron wavefunction computation with Slater determinants, illustrated for molecules. The input to the network is a set of 3D electron coordinates with $N^\uparrow$ electrons with spin-up and $N^\downarrow$ electrons with spin-down. The spin structure ($\uparrow$ or $\downarrow$) as well as the positions of atomic nuclei are fixed. A neural network is used to produce $k$ $(N^\uparrow+N^\downarrow)$-dimensional  features vectors for each electron, which are then concatenated into $k$ $(N^\uparrow+N^\downarrow)\times(N^\uparrow+N^\downarrow)$ matrices. Finally, the determinants of these matrices are computed and a linear combination of them gives the final wavefunction value.}
    \label{fig:many_electron}
\end{figure}

\vspace{0.1cm}\noindent\textbf{Orbital Modeling: } 
Although Slater determinants solves the exchange antisymmetry for many-electron systems, it does not provide any guarantee on the accuracy of the optimized wavefunction. To accurately model the ground-state wavefunction, the orbital functions $\bm\phi^\uparrow$ and $\bm\phi^\downarrow$ must stem from a flexible function family. 
Classically, orbital functions are modeled as single-particle orbitals from solutions for the single-particle Schrödinger's equation. FermiNet~\citep{pfau2020ab} and PauliNet~\citep{hermann2020deep} successfully use neural networks to model orbital functions while using the Slater determinant as antisymmetric aggregation, where all 3D electron coordinates are first encoded as a $(N^\uparrow+N^\downarrow)\times k$-dimensional feature vector with permutation-equivariant neural networks $\bm\phi^\uparrow_\theta$ and $\bm\phi^\downarrow_\theta$. The vectors are then concatenated into $k$ $(N^\uparrow+N^\downarrow)\times (N^\uparrow+N^\downarrow)$ matrices (\emph{e.g.}, $k= 16$). Finally, the determinants of these matrices are computed and the final wavefunction value is the linear combination of determinants. As a result, the parameters in the network are the parameters in the orbital networks and the combination weights of determinants. Due to their prominent performances, follow-up works mostly follow the same pipeline, which is shown in Figure~\ref{fig:many_electron}. 
To build a complete representation of the quantum state, one must consider all pairwise information between all particles, including electrons and nuclei. To this end, commonly used input features are relative vectors and distances between electron-electron and electron-nucleus pairs.
Additionally, a symmetric part $e^{J(\bm{r};\bm\theta)}$ can be multiplied to the final wavefunction, which is called the \emph{Jastrow factor}.

\vspace{0.1cm}
\noindent\textbf{Orbital Modeling --- Single-Electron Feature Computation:}
To capture the complex physics interactions, it is necessary that the orbital networks $\bm\phi^\uparrow_\theta$ and $\bm\phi^\downarrow_\theta$ consider all electron positions collectively. Hence, when encoding one electron, all other electrons are also encoded to provide context information. As a result, $\bm\phi^\uparrow_\theta$ and $\bm\phi^\downarrow_\theta$ must be able to gather information from other electrons, \emph{i.e.}, for the $i$-th electron, instead of computing $\bm\phi_\theta^\sigma(\bm{r_i})$, $\sigma\in\{\uparrow, \downarrow\}$, $\bm\phi_\theta^\sigma(\bm{r_i};\bm{r})$ is computed.
This idea, known as \emph{neural backflow}, is first proposed in~\citet{luo2019backflow} for electronic systems on the lattice in first quantization. It subsequently inspired later work and is adopted to study molecules~\citep{hermann2020deep, pfau2020ab}.
PauliNet~\citep{hermann2020deep} uses distance-based convolution to gather information from neighborhood electrons similar to SchNet~\citep{schutt2018schnet}, where convolution weights are computed based on the relative distances: $\bm{h}'_i= \sum_j \bm{w}(\Vert\bm{r}_j-\bm{r}_i\Vert_2;{\bm\theta})\odot \bm{f}(\bm{h}_j)$, where $\odot$ denotes element-wise product and $\bm{w}$ computes differently for different spins. A similar computation is also applied for nuclei.
FermiNet~\citep{pfau2020ab} uses mean field information about the electronic structure and pairwise distance features between each pair of electrons. Concretely, FermiNet maintains two computation branches that compute one-particle features and pairwise features, respectively. At each layer, both one-particle features $\bm{h}_i$ and pairwise features $\bm{h}_{ij}$ are averaged over electrons to get global representations for each spin, which are further concatenated to the one-particle features for the next layer $\bm{h}'_i = \bm{f}([\bm{h}_i, \sum_{j, \sigma(j)=\uparrow} \bm{h}_j, \sum_{j, \sigma(j)=\downarrow} \bm{h}_j, \sum_{j, \sigma(j)=\uparrow} \bm{h}_{ij}, \sum_{j, \sigma(j)=\downarrow} \bm{h}_{ij}])$, where $\sigma(i)$ denotes the spin of the $i$-th electron and $[\cdot]$ denotes concatenation.
FermiNet+SchNet~\citep{gerard2022goldstandard} replaces the simple interactions in FermiNet by integrating the convolutions from PauliNet.
By abandoning the mean field of FermiNet, PsiFormer~\cite{glehn2023a} achieves significantly lower energies by using an attention mechanism to capture all pairwise electron-electron interactions.
In contrast, Moon~\citep{gao2023generalizing} replaces the global mean field of FermiNet by local mean fields on the nuclei by using continuous convolutions resulting in similar accuracy to PsiFormer.
For superfluids and homogeneous electron gas, similar networks can be used but the input coordinates must be embedded with periodic functions to respect the periodicity of the problem~\citep{pescia2022neural, cassella2023discovering}.
MP-NQS~\citep{pescia2023message} models quantum states of HEG with message passing networks.

Fermion antisymmetry does not require $\bm\phi^\uparrow_{\bm\theta}$ and $\bm\phi^\downarrow_{\bm\theta}$ to be the same mappings. Hence, in the most general setting, they should be able to produce different computations.
While traditionally choosing $\bm\phi^\uparrow_\theta$ and $\bm\phi^\downarrow_\theta$ to be different functions lead to better variational energies, for the ground state of a singlet-state system, \emph{i.e.}, all electron spins are paired, we do have $\bm\phi^\uparrow_\theta=\bm\phi^\downarrow_\theta$. In the neural network setting, \citet{gao2023samplingfree} has shown that implementing this constraint for singlet-state systems improves energies and accelerates optimization in neural network-based wavefunctions. However, in practice, most of the parameters of these two networks can be shared. Concretely, at each layer of the network, each electron is encoded as a feature $\bm{h}_i\in \mathbb{R}^d$ where $d$ is the hidden dimension. The feature at the next layer is computed as $\bm{h}'_i = \bm{f}(\bm{h}_i, \bm{h}^\sigma, \{\bm{h}_j\}_{j\neq i})$ where $\bm{h}^\sigma, \sigma\in\{\uparrow, \downarrow\}$ is a feature that which represents global information for different spins, which distinguishes spin-up and spin-down electrons,  For example, $\bm{h}^\uparrow$ and $\bm{h}^\downarrow$ can be computed as the average of spin-up electron features and spin-down electron features, respectively. As a result, when two electrons with the same spins are swapped, $\bm{h}^\uparrow$ and $\bm{h}^\downarrow$ will not be influenced. Hence, their feature vectors will also be swapped. Otherwise, if two electrons with different spins are swapped, $\bm{h}^\uparrow$ and $\bm{h}^\downarrow$ will change and their feature vectors will become entirely different. 

\vspace{0.1cm}
\noindent\textbf{Orbital Modeling --- Incorporating Physics:}
Recent studies show that the incorporation of physics knowledge can have an important impact on performance.
Among those, wavefunction has decaying behavior at long distances, and envelope functions are used to ensure this fundamental behaviour. Essentially, the neural wavefunctions are multiplied with a mask function such that the wavefunction vanishes when electrons are far away from the nuclei. For example, FermiNet~\citep{pfau2020ab} uses a simple exponential envelope. In the VMC setting, this also ensures to have a finite MCMC integration. 
Moreover, due to the singularities in the potential energy when two particles overlap, the wavefunctions must have discontinuous derivatives at such configurations, known as electron-electron cusp and electron-nuclear cusp. However, modeling such non-smooth behaviors is challenging for neural networks. A common way to handle electron-electron cusps is to include an explicit divergent term in the wavefunction. For example, PauliNet~\citep{hermann2020deep} handles electron-electron cusps by multiplying the wavefunction by $\exp\left(\sum_{i<j}-\frac{a_{ij}} {1+\|\bm{r}_i-\bm{r}_j\|}\right)$, with $a_{ij}$ being spin-dependent constants, which is part of the Jastrow factor. FermiNet~\citep{pfau2020ab} proposes that the cusp conditions can be modeled by using distances as input features, \emph{e.g.}, $\|\bm{r}_i - \bm{r}_j\|$ since they are not differentiable when two particles overlap.
For superfluids and homogeneous electron gas, periodic envelope functions can be used to improve convergence~\citep{lou2023neural, cassella2023discovering}.

Moreover, another commonly used strategy to incorporate physics is to make use of classical solutions.
A commonly used strategy is to pretrain the orbital networks $\phi_{\bm\theta}$ to match the classical orbitals, such as the ones obtained with Hartree-Fock methods.
Although \citet{gerard2022goldstandard} shows that this is not always beneficial to pretrain with more accurate classical solutions.
In contrast, PauliNet~\citep{hermann2020deep} directly uses Hartree-Fock solution as part of the orbital functions $\phi_{\bm\theta}$ instead of pretraining.
For homogeneous electron gas, WAP-net~\citep{wilson2022wave} multiplies the orbital function with Hartree-Fock plan wave orbitals, computed with transformed coordinates.

\vspace{0.1cm}\noindent\textbf{Optimization: }
Same to lattice systems, the neural wavefunction $\psi_\theta$ can be optimized via variational Monte Carlo (VMC). 
As defined in~\cref{eq:many_elec_optim_obj}, the objective to minimize the energy expectation, $\mathcal{L}(\bm{r}; \bm{\theta}) = \mathbb{E}_{p_{\bm\theta}} [E_{loc}(\bm{r}; \bm\theta)]$,  $p_{\bm\theta} \propto \psi_{\bm\theta}^2$. Since the expectation cannot be integrated analytically, we need to estimate the gradient from samples. One common way in machine learning to compute gradient through expectation is to use the stochastic gradient where the overall gradient is computed as the average gradient from each sample. However, the stochastic gradient cannot be used in our case because updating the parameters will also change the underlying probability distribution $p_\theta$.
To account for the distribution shift, a closed form gradient can be computed using the Hermitian property of the Hamiltonian~\citep{ceperley1977monte}. For real-valued wavefunctions it is given by
\begin{equation}\label{eq:many_electron_gradient}
    \nabla_{\bm\theta} \mathcal{L}(\bm{r};\bm\theta) = 2\mathbb{E}_{p_{\bm\theta}}\left[\left(E_{loc}(\bm{r};\bm\theta) - \mathbb{E}_{p_{\bm\theta}}[E_{loc}(\bm{r};\bm\theta)] \right) \nabla_{\bm\theta}\log |\psi_{\bm\theta}(\bm{r};\bm\theta)|\right].
\end{equation}
Compared to the spin systems (~\cref{eq:quantum_spin_gradient}), we omit the complex conjugate since $\psi_{\bm\theta}$ is real-valued.
From here we can use sample means to evaluate the expectations since we no longer need to differentiate through expectations and thus the gradient is unbiased.
However, to estimate the expectations correctly, we need to generate samples following $p_{\bm\theta}$. 
To this end, MCMC with Metropolis-Hasting can be used to generate such samples. 
Given a batch of current samples $\bm{r}$ (which are randomly initialized), we randomly perturb each of them with a Gaussian noise to get $\tilde{\bm{r}}_i=\bm{r}_i + \delta\bm{r}_i$, where $\delta\bm{r}_i\sim \mathcal{N}\left(0, \sigma\right)$. 
We then decide whether to accept the perturbed samples with Metropolis-Hasting rejection.
Specifically, for each electron $i$, we compute the ratio $q = p_{\bm\theta}(\tilde{\bm{r}}_i) / p_{\bm\theta}(\bm{r}_i)$ and at the same time uniformly sample a random number $a$ from $[0, 1]$. We then compare the value of $q$ and $a$. If $q \geq a$, we keep the perturbed sample and let $\bm{r}_i = \bm{\tilde{r}_i}$. Otherwise, we reject the proposal and keep $\bm{r}_i$ unchanged. We can prove that the samples generated with this procedure will converge to $p_{\bm\theta}$. 
$\sigma$ controls how different the proposals are. Practically, we control $\sigma$ so that the acceptance ratio is around 0.5.

Another important component in optimization is the choice of optimizer. Second-order optimizers such as natural gradients are found to be critical to achieve accurate optimization. Compared to first-order methods which update the parameters directly follow the reverse direction of gradients, which corresponds to the steepest direction in Euclidean space, natural gradient preconditions the gradient with the inverse of the Fishier information matrix so that the updates in the steepest direction in terms of the distribution. Concretely, the parameters are updated as $\bm{\theta}\leftarrow \bm{\theta} - \eta F^{-1}\nabla_{\bm\theta} \mathcal{L}(\bm{r};\bm\theta)$. When dealing with unnormalized wavefunctions, the parameter update is equivalent to the stochastic reconfiguration with 
$F_{ij}\propto \mathbb{E}_{p_{\bm\theta}}\left[\left(\mathcal{O}_i -  \mathbb{E}_{p_{\bm\theta}}[\mathcal{O}_i]\right)\left(\mathcal{O}_j -  \mathbb{E}_{p_{\bm\theta}}[\mathcal{O}_j]\right)\right]$ \revisionOne{and} $\mathcal{O}_i = \frac{\partial \log |\psi_{\bm\theta}(\bm{r})|}{\partial \theta_i}$.
However, \revisionOne{as described in~\cref{sec:qm_disc_optim}, directly} computing $F^{-1}$ is infeasible for large models. \revisionOne{In addition to various acceleration methods introduced in~\cref{sec:qm_disc_optim}, } 
by making certain assumptions to the Fisher matrix, KFAC~\cite{martens2015optimizing} accelerates the computation by factorizing the Fishier matrix with Kronecker products. Alternatively, as commonly used for learning neural quantum states,the conjugated gradient (CG) method can be employed, which approximates the term  $F^{-1}\nabla_{\bm\theta} \mathcal{L}(\bm{r};\bm\theta)$~\citep{neuscamman2012optimizing, carleo2017solving, gao2021ab, vicentini2022netket}.

Besides VMC, there exist other methods to optimize neural wavefunctions. In diffusion Monte Calo (DMC), each sample is additionally assigned with a weight such that the weighted average of sampled energies can be closer to the true ground state energy. To achieve this, the weights are computed based on imaginary time evolution. In DMC the sampling is done with Langevin dynamics, where the samples are generated following the quantum drift (or the score in machine learning), defined as  $\nabla_{\bm{r}} \log \psi$. The process approximates the iterative application of the imaginary-time evolution operator $\psi \leftarrow e^{-\tau \hat{H}}\psi$, where $\tau$ is the evolution time. The score gives a 3D vector for each electron, which points towards the direction of higher probability density.~\citet{ren2022towards, wilson2021simulations} first train a FermiNet with VMC and then use DMC to further approach the ground states.
Moreover, DiffVMC~\citep{zhang2023score} combines VMC and DMC by parameterizing the score directly. Instead of updating the weights for samples, DiffVMC updates the parametrized score function directly through a specially designed loss function based on score matching~\citep{hyvarinen2005estimation}.


\ifshowname \textcolor{red}{Nicholas} \else\fi

\vspace{0.1cm}
\noindent\textbf{Generalized Wavefunctions for Multiple Geometries: }
Learning generalized Wavefunction that either covers the complete PES of a given molecule or across different compounds, faces a set of various challenges that do not apply to single structure calculations.

\vspace{0.1cm}\noindent\textbf{Multiple Geometries --- Generalized Orbitals:} In learning generalized wavefunctions, a key challenge is to adapt the molecular orbital function $\phi_i$ to the molecular structure. There have been various approaches on accomplishing this. The first work by \citet{scherbela2022solving} (DeepErwin) tackles the problem by sharing most of the parameters across different structures and only retraining specific weights for each structure. Concurrently, \citet{gao2021ab} proposes a two-layered network approach to adapting the orbital functions called Potential Energy Surface Network (PESNet). In PESNet, the orbital functions are parametrized by another neural network that only acts on the nuclei, similar to supervised surrogate models. This avoids the need for retraining completely and has to be optimized only once to model a whole PES of a molecule. However, while the weight-sharing approach can be transferred to different sets of atoms, PESNet has no such capabilities.

When learning generalized orbitals for different compounds, \emph{i.e.}, varying sets of nuclei, the problem of parameterizing molecular orbital functions $\{\phi_i\}_{i=1}^N$ is aggravated by the varying number of molecular orbitals $N$ which corresponds to the number of electrons. While many weights can still be shared, one still needs to optimize the wavefunction separately for each molecule\citep{scherbela2022solving}. Two concurrent works tackle this problem and  avoid the individual optimizations: Transferable Atomic Orbitals (TAOs) \citep{scherbela2023towards} and Graph-learned orbital embeddings (Globe) \citep{gao2023generalizing}.
In TAO, the molecular orbital functions are constructed as linear combinations of atomic orbital functions $\phi_i = \sum_{j}a_j\varphi_j$ similar to Hartree-Fock (HF) theory. In fact, TAO uses classical HF calculations to obtain the coefficients $a_j$. In contrast, Globe does not rely on classical HF calculations but builds on a two-layered network structure like PESNet and localizes orbitals as points in space. The parameters of the orbitals are then learned by spatial message passing similar to SchNet~\citep{schutt2018schnet}. In their respective evaluations, the authors find TAO to perform better in transferring the wavefunction to new molecules as the HF calculation provides a strong inductive bias~\citep{scherbela2023towards} while Globe shows strong capabilities in learning various compounds' ground states simultaneously~\citep{gao2023generalizing}.

\vspace{0.1cm}\noindent\textbf{Multiple Geometries --- Symmetries:} As the energy $E$ is observable, it is invariant with respect to the Euclidean group $E(3)$. Concretely, let $U_R$ be the unitary operator associated with the rotation matrix $R$, the rotated Schrödinger equation $U_R \hat{H}U_R^\dagger\psi=E\psi$ with $U_R\hat{H}U_R^\dagger=-\frac{1}{2}\sum_i \nabla_i^2 + \sum_{i<j}\frac{1}{\Vert\bm{r}_i-\bm{r}_j\Vert} - \sum_{i,I}\frac{z_I}{\Vert\bm{r}_i - \bm{c_I}R\Vert} + \sum_{I<J}\frac{1}{\Vert \bm{c}_I - \bm{c}_J\Vert}$ being the rotated Hamiltonian operator.
From this formulation, one can see that rotating an eigenfunction $\psi$ of $\hat{H}$ solves the rotated Schrödinger equation, \emph{i.e.},
\begin{align}
    U_R \hat{H}U_R^\dagger U_R\psi&=EU_R\psi,\\
    U_R\hat{H}\psi&=EU_R\psi,\\
    \hat{H}\psi&=E\psi.
\end{align}
Thus, to obtain invariant energies, one must have an equivariant wavefunction. A simple implementation would be invariant wavefunctions like PauliNet~\citep{hermann2020deep} but as wavefunctions do not have to be invariant, this severely restricts the function class and typically results in higher energies.
Instead, current approaches either rely on using equivariant coordinate frames that rotate with the nuclear structure~\citep{gao2021ab,gao2022sample,gao2023generalizing} or augmenting the training data by arbitrary rotations~\citep{scherbela2023towards}.
In addition to the Euclidean group, the PES is also invariant to the permutation group of the nuclei $S_M$. Integrating this symmetry is typically achieved by relying on summations over nuclei in the orbital functions rather than concatenations~\citep{gao2021ab}.

\vspace{0.1cm}\noindent\textbf{Multiple Geometries --- Size-Consistency:} Similar to symmetries which tell us the exact change of the wavefunction under certain actions, size consistency is prior information about the energy of the system. Specifically, size consistency refers to the change in energy depending on the size of the modeled system. In the limit case, where the system can be decomposed into two non-interacting subsystems, the energy of the whole system is simply the sum of the energies of the subsystems. Though, as it is only phrased in the limit case of non-interacting systems it cannot be phrased as a symmetry in a strict way. Nonetheless, restricting the functional form of neural wavefunctions by size consistency reduces the search space of potential function classes and results in better generalization~\citep{gao2023generalizing}. It can be shown that to implement size consistency, one needs to restrict orbital function $\phi$ to have decaying receptive fields such that particles do not interact with each other given sufficient distance~\citep{gao2023generalizing}. This is incompatible with the widely used FermiNet architecture which strongly relies on global averages. The Molecular orbital network (Moon)~\citep{gao2023generalizing} implements this by relying decaying spatial filters in a message-passing scheme similar to SchNet~\citep{schutt2018schnet}.

\vspace{0.1cm}\noindent\textbf{Multiple Geometries --- Energy Surfaces:} As the wavefunction is directly linked to the energy, one could for each structure solve the integral $\int\psi_\theta(\bm{r}, M)\hat{H}_M\psi_\theta(\bm{r}, M) d\bm{r}$ numerically to obtain the respective energy. However, due to the inherently expensive process of Monte Carlo integration, this proves costly if thousands to millions of states have to be evaluated. For single structure calculations, the final energy is often approximated as a running average over the energies observed during training, \emph{e.g.}, the last 20\% of observed training energies. However, this does not translate to the generalized setting where molecular structures are often sampled from some continuous distribution~\citep{gao2021ab}. Thus, the structure will not be observed multiple times. \citet{gao2022sample} tackles this issue by introducing Potential learning from ab-initio Networks (PlaNet). In PlaNet, one translates the idea of averaging training energies from the single-molecule setting to PES modeling by averaging the observed energy surfaces at each time step.
In practice, this means that at each time step one fits a function to the observed energies. These functions are then temporally averaged in a first-order Taylor approximation by averaging the parameters of the function.

\subsubsection{Datasets and Benchmarks}

Same as quantum spin systems (Section~\ref{sec:quantum_spin_data}), the training data is sampled according to the distribution defined by the neural wavefunction. As a result, there is no need to generate a dataset beforehand. Instead, the data is defined in terms of the atom coordinates of a geometry. On the other hand, due to the variational nature of the optimization process, accuracies are evaluated by the average energy as well as the standard deviation estimated from samples, and a lower energy represents a more accurate result. Commonly tested systems are small or heavy atoms such as \ce{N} or \ce{Fe}, small or large molecules such as \ce{N2} or \ce{CCl4}~\citep{glehn2023a}, some special atomic configurations such as \ce{H10}, compound structures such as the Benzene dimer~\citep{ren2022towards, glehn2023a}. as well as transition energies for molecular systems, defined as the difference between the ground state energies after and before a chemical process, such as the automerization of cyclobutadiene (\ce{C4H4})~\citep{spencer2020better} and the ionization of \ce{Fe}~\citep{glehn2023a}.

\subsubsection{Open Research Directions}
There are several remaining challenges for modeling many-electron systems with VMC. First, due to the fermion antisymmetry constraint, most existing methods use Slater determinants. However, optimizing through determinants could introduce extra difficulty. It is still to be seen whether we can effectively achieve fermion antisymmetry using methods other than Slater determinants. Second, currently most methods model wavefunctions explicitly. Modeling wavefunctions in real space is similar to modeling probability density in generative machine learning. Moving towards implicit modeling could be an interesting direction. Finally, 
one of the most pressing challenges lies in computational efficiency. As the computational complexity scales $O(N^4)$ with the number of electrons $N$, current calculations are limited to at most 80 electrons.
it is important to further scale the methods. This could be achieved by more efficient sampling, better optimization, and enabling more effective weight sharing across systems. For example, the deep learning library Jax~\citep{jax2018github} has shown good accelerations for energy evaluation by improving implementations. Scaling is important to extend QMC methods to larger molecular systems or materials.

\clearpage
\hypertarget{AI for Density Functional Theory}{\section{AI for Density Functional Theory}} \label{sec:dft}

\begin{figure}
    \centering
    \includegraphics[width=\textwidth]{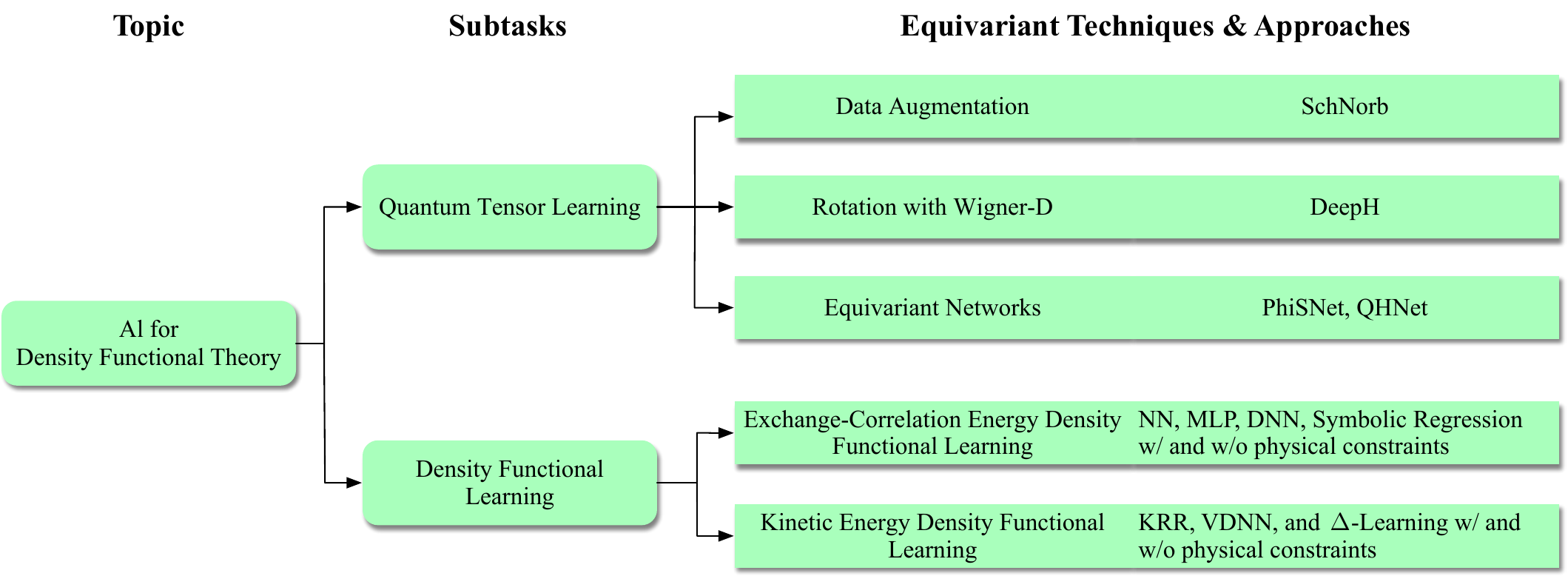}
    \caption{{\ifshowname\textcolor{red}{Figure: Zhao} \else\fi}An overview of tasks and methods in AI for density functional theory (DFT). In the quantum tensor learning subtask, invariant quantum tensor networks include SchNorb~\cite{schutt2019unifying} and DeepH~\cite{deeph}. SchNorb encourages the equivariance by training with data augmentation and DeepH guarantees the equivariance by rotation with Winger D-matrix. Meanwhile, equivariant quantum tensor networks, including PhiSNet~\cite{unke2021se} and QHNet~\cite{yu2023efficient}, intrinsically consider the equivariance of matrix by tensor product and tensor expansion. Another subtask is density functional learning, primarily focused on approximating exchange-correlation (XC) energy and kinetic energy. Several machine learning approaches have been employed for approximating exchange-correlation energy density functionals. These include neural network (NN)~\cite{Tozer1996MLXC,Dick2019MLCF,Dick2020NeuralXC,Ryabov2020NNXC,Lei2019invariantdescriptors,Gedeon2022MLXCDiscontinuity}, multiple layer perceptron (MLP)~\cite{nagai2020completing}, deep neural network (DNN)~\cite{kirkpatrick2021pushing,Pokharel2022Exact}, and symbolic regression~\cite{Ma2022Evolving}, implemented both with and without physical constraints. In terms of approximating kinetic energy density functionals, approaches such as kernel ridge regression (KRR)~\cite{Snyder2012MLXC,Snyder2015Denoising}, voxel deep neural network (VDNN)~\cite{ryczko2022toward}, and $\Delta$-Learning~\cite{Ramakrishnan2015DeltaLearning} have been used, again both with and without the inclusion of physical constraints. Alternatively, Ref.~\cite{Brockherde2017Bypassing} aims to learn Hohenbergy-Kohn map between external potential and electron density using KRR, thereby bypassing the Kohn-Sham equation.}
    \label{fig:dft_overview}
\end{figure}

In this section, we first introduce the basic knowledge of density functional theory (DFT) in Section~\ref{sec:dft:overview}. Then in Section~\ref{sec:dft:QTLearning}, we formulate the quantum tensor learning problem and describe a couple of state-of-the-art machine learning models. In Section~\ref{sec:dft:MLXCKE}, we review the recent progress of machine learning approaches for more accurate density functionals. We point out some promising future directions, \emph{e.g.} quantum tensors as physics-based features for deep learning and machine learning density functionals. We summarize the structure of this section in Figure \ref{fig:dft_overview}.


\subsection{Overview}
\label{sec:dft:overview}

\noindent{\emph{Authors: Xiaofeng Qian, Haiyang Yu, Alexandra Saxton, Zhao Xu, Xuan Zhang, Shuiwang Ji}}\vspace{0.3cm}\newline
\emph{Recommended Prerequisites:~\cref{sec:QM_Overview}}\newline

{\ifshowname\textcolor{red}{Revised by Xiaofeng}\else\fi}In principle, modeling the quantum states of physical systems, such as molecules or materials, requires the explicit form of their wavefunctions $|\psi \rangle$ by solving the many-body Schr\"odinger equation. However, as the number of electrons in the system increases, the Hilbert space of the system grows exponentially, resulting in high computational and memory consumption. \revisionOne{While lower accuracy approximations can be successfully employed in \textit{ab initio} calculations on hundreds of atoms, it becomes impractical to solve the Schr\"odinger equation exactly or with high accuracy \textit{ab initio} calculations, such as full configuration interaction methods or complete active space SCF methods,} for even small molecular systems with just tens of electrons. Therefore, for the time being, the methods in Section \ref{sec:qt} cannot be directly used to obtain the many-body wavefunctions by solving the Schr\"odinger equation (Equation~(\ref{sec:qt eq:sch})) for large and complex systems. 

\revisionOne{Note that this section will be highly related to symmetry and equivariant graph networks which has been introduced in Section~\ref{subsec:cont_feat} and Section~\ref{subsec:cont_equi}.}

\subsubsection{Density Functional Theory}

\begin{figure}
    \centering
    \includegraphics[width=\textwidth]{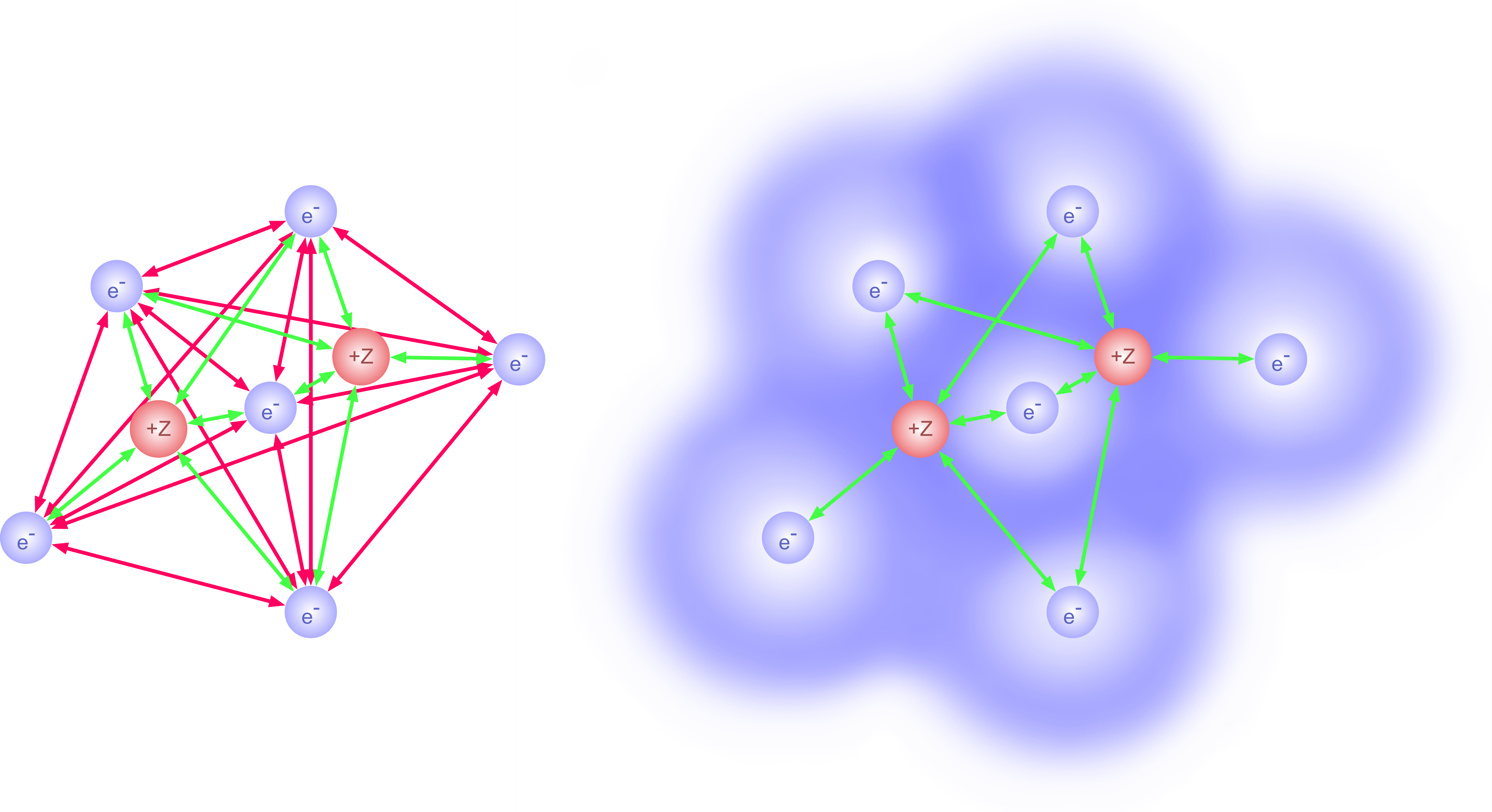}
    \caption{
    \ifshowname\textcolor{red}{Figure: Allie. Caption: Xuan, Xiaofeng}\else\fi
    An illustration contrasting the interacting many-body perspective (left) with the DFT perspective within the Kohn-Sham framework (right) for modeling electronic structure in an atomistic system (\emph{e.g.}, molecules). Red spheres represent nuclei and blue spheres represent electrons. Red and green edges represent interactions between electron-electron pairs and electron-nucleus pairs (\emph{e.g.}, via Coulomb potential), respectively. The blue shading on the right represents the electron density. \textbf{Left:} In the interacting many-body view, the wavefunction of the system is defined w.r.t. the coordinates of all particles. Hence the interactions between all electron-electron and electron-nucleus pairs are explicitly considered, leading to an exponential growth of the dimension for many-body wavefunctions with the increasing number of electrons. \textbf{Right:} In the DFT Kohn-Sham picture, the electron-electron interactions are replaced by the interaction between each electron and the average effect of all other electrons, modeled with an electron density. Since such interactions are equivalent for different electrons, modeling the electronic wavefunction of the system effectively reduces to modeling multiple single-electron wavefunctions. The lighter shading around each electron illustrates the exclusion of the electron itself in the electron density to avoid the self-interaction.}
    \label{fig:opendft:wavefunction_vs_density}
\end{figure}

{\ifshowname\textcolor{red}{Xiaofeng}\else\fi} In practice, first-principles density functional theory (DFT)~\cite{Hohenberg64,dft} and \textit{ab initio} quantum chemistry methods~\cite{szabo2012modern} are widely used to solve the Schr\"odinger equation with different approximations and provide near-form wavefunctions in polynomial time complexity. In quantum chemistry methods such as Hartree-Fock theory, the total wavefunction of a system is represented by the Slater determinant of noninteracting electrons which satisfies the exact antisymmetry upon exchanging two fermionic electrons (\emph{i.e.}, identical particles with spin 1/2). In the Hartree-Fock theory, the electron-electron Coulomb interaction and the exact exchange interaction are precisely taken into account, while the additional electron correlation energy beyond the exact exchange is not considered. Alternatively, the Hohenberg-Kohn DFT theorem~\cite{Hohenberg64} shows that (HK-1) the ground-state electron density $\rho({\bm r)}$ (a three-dimensional quantity) uniquely determines the external potential (such as electron-nuclei interaction) and thus the Hamiltonian, and (HK-2) the ground-state total energy is a functional of electron density minimized by the ground-state density $\rho_{\bf{gs}}$, \emph{i.e.}, $E[\rho] \geq E[\rho_{\bf{gs}}]$ for any trial/approximate density $\rho \neq \rho_{\bf{gs}}$. In practice, the many-body interacting system may be mapped onto many one-body noninteracting systems within the DFT Kohn-Sham (KS) framework~\cite{dft}, where individual electrons are subject to a mean field that depends on the total electron density, as illustrated in Figure~\ref{fig:opendft:wavefunction_vs_density}. The corresponding Kohn-Sham electronic total energy $E_{\bf{KS}}[\rho]$ is given by
\begin{equation}
      E_{\bf{KS}}[\rho] = E_{\bf kin}[\rho] + E_{\bf H}[\rho] + E_{\bf ext}[\rho] + E_{\bf XC}[\rho],
\label{eq: kohn-sham-energy-functional}
\end{equation}
where $E_{\bf kin}$ is the kinetic energy of noninteracting electrons,  $E_{\bf H}$ is the Hartree term originating from electron-electron Coulomb interaction, $E_{\bf ext}$ denotes the external potential energy \emph{e.g.} due to the interaction between electrons and nuclei, and $E_{\bf XC}$ denotes the exchange-correlation (XC) energy. Since the system is mapped onto a noninteracting one in the Kohn-Sham framework, the total electron density $\rho(\bm{r})$ can be obtained by summing over the contributions from individual non-interacting electrons in the orthonormal eigenstates $\psi_i$,
\begin{equation}
      \rho(\bm{r}) = \sum_{i} f_i \psi_i^*(\bm{r}) \psi_i({\bm r}),
\label{eq: kohn-sham-electron-density}
\end{equation}
where $f_i$ denotes the occupation number in state $\psi_i(\bm r)$. Since the total number of electrons, $N_e$, is fixed for a given system, thus $\int \rho(\bm{r}) d\bm{r} = N_e$. For orthonormal eigenstates,  $\langle \psi_i | \psi_j \rangle = \int \psi_i^*(\bm{r}) \psi_i({\bm r}) d \bm{r} = \delta_{ij}$, hence $\sum_i f_i = N_e$. Correspondingly, the kinetic energy $E_{\bf kin}$ can be calculated by directly evaluating the expectation value of the kinetic energy operator  $\hat{T} = -\frac{\hbar^2}{2 m_e} \nabla^2$ for all occupied electronic states,
\begin{equation}
      E_{\bf kin}[\rho] = \sum_{i} f_i \langle \psi_i | \hat{T} | \psi_i \rangle 
      = \sum_{i} f_i \int \psi_i^*({\bm r}) \left[ -\frac{\hbar^2}{2 m_e} \nabla^2 \right]  \psi_i({\bm r}) d \bm{r}, 
\label{eq: kohn-sham-kinetic-energy}
\end{equation}
where $\hbar$ stands for the reduced Planck's constant and $m_e$ stands for the electron mass. 
Furthermore,
\begin{equation}
      E_{\bf H}[\rho] = \frac{e^2}{8 \pi \varepsilon_0 } \iint
      \frac{ \rho(\bm{r}) \rho(\bm{r'}) }
      {|\bm{r} - \bm{r'}|} d \bm{r} d \bm{r'},
\label{eq: kohn-sham-hartree-energy}
\end{equation}
\begin{equation}
      E_{\bf ext}[\rho] = \int 
      V_{\bf{ext}}(\bm{r}) \rho(\bm{r}) d \bm{r},
\label{eq: kohn-sham-external-potential-energy}
\end{equation}
where $e$ stands for the elementary charge, $\varepsilon_0$ stands for vacuum permittivity, and $V_{\bf{ext}}$ denotes the external potential. For the cases of molecules and materials, the external potential simply comes from the electron-nuclei Coulomb interaction,
\begin{equation}
    V_{\bf{ext}}(\bm{r}) = - \frac{e^2}{4 \pi \varepsilon_0} \sum_I
    \frac{Z_I}{|\bm{r} - \bm{R}_I|},
\label{eq: kohn-sham-external-potential}
\end{equation}
where $\bm{R}_I$ and $Z_I$ denote the position and charge of the nuclei $I$, respectively. 
Similarly, the exchange-correlation (XC) energy $E_{\bf XC}[\rho]$ can also be readily evaluated for a given electronic density $\rho(\bm{r})$.

\subsubsection{The Kohn-Sham Equation}
{\ifshowname\textcolor{red}{Xiaofeng}\else\fi}
According to the second Hohenberg-Kohn theorem, one can minimize Equation~(\ref{eq: kohn-sham-energy-functional}) by varying the electron density using the variational principle, thereby achieving the ground-state density and total energy. In practice, the method of Lagrange multipliers is applied under the constraint of total electrons $N_e$, where the Lagrange function is constructed as follows:
\begin{equation}
      \mathcal{L}[{|\psi_i}\rangle] = E_{\bf{KS}} - \sum_i \epsilon_i \left( \langle \psi_i | \psi_i \rangle - 1 \right).
\label{eq: kohn-sham-Lagrange}
\end{equation}
Thus at the stationary points of Lagrange function $\mathcal{L}$, we have 
\begin{equation}
      \frac{\delta \mathcal{L}} {\delta \langle \psi_i |} 
      = H_{\bf{KS}} | \psi_i\rangle - \epsilon_i | \psi_i\rangle = 0.
\label{eq: kohn-sham-Lagrange-equation}
\end{equation}
We then obtain the DFT Kohn-Sham equation,
\begin{equation}
      H_{\bf{KS}} | \psi_i\rangle = \epsilon_i | \psi_i\rangle,
\label{eq: kohn-sham-equation}
\end{equation}
where $H_{\bf{KS}}$ is the Kohn-Sham Hamiltonian, and $\epsilon_i$ and $\psi_i$ are the corresponding Kohn-Sham eigen energies and eigen wavefunctions, respectively. More explicitly,
\begin{equation}
      H_{\bf{KS}} = -\frac{\hbar^2}{2 m_e}\nabla^2
      + V_{\bf{H}}(\bm{r})
      + V_{\bf{XC}}(\bm{r})
      + V_{\bf{ext}}(\bm{r}),
\label{eq: kohn-sham-hamiltonian}
\end{equation}
where $V_{\bf{H}}$ is the Hartree potential as 
\begin{equation}
V_{\bf{H}}(\bm{r}) = \frac{e^2}{4 \pi \varepsilon_0}\int \frac{\rho(\bm{r})} {|\bm{r} - \bm{r'}|} d \bm{r'}. 
\label{eq: kohn-sham-Hartree-potential}
\end{equation}
$V_{\bf{XC}}$ is the exchange-correlation potential as 
\begin{equation}
V_{\bf{XC}}(\bm{r}) \equiv \delta E_{\bf{XC}}[\rho]/\delta \rho(\bm{r})
\label{eq: kohn-sham-XC-potential}
\end{equation}
and $V_{\bf{ext}}$ is the external potential defined above in Equation~(\ref{eq: kohn-sham-external-potential-energy}) for molecules and materials. Alternatively, the three potentials can be considered as an effective potential or self-consistent field (SCF) for individual electron, with $V_{\bf{eff}}(\bm{r}) = V_{\bf{SCF}}(\bm{r}) = V_{\bf{H}}(\bm{r}) + V_{\bf{XC}}(\bm{r})
     + V_{\bf{ext}}(\bm{r})$, thus
\begin{equation}
       H_{\bf{KS}} = -\frac{\hbar^2}{2 m_e}\nabla^2 + V_{\bf{SCF}} (\bm{r}). 
\label{eq: kohn-sham-hamiltonian-SCF}
\end{equation}
More details of theoretical background, technical implementation, and applications of DFT can be found in many excellent books~\cite{ParrYang1995DFT,EngelDreizler2011DFT,DreizlerGross2012DFT,Fiolhais2003DFTPrimer,Kaxiras_Joannopoulos_2019,CohenLouie2016Fundamentals,martin2016Interacting,martin2020electronic,koch2015chemist,Sholl2009DFT,Yip2005Handbook,Giustino2014modeling} and review papers~\cite{payne1992rmp,Kohn1999RMP,Kummel2008ODDFT,Jones2015RMP}.
Nevertheless, with the exact exchange-correlation energy functional (though unknown by far), ground-state total energy as well as many other ground-state properties such as atomic forces and electric polarization can be derived exactly. Furthermore, although Kohn-Sham eigen wavefunctions $\psi_i(\bm{r})$ and eigen energies $\epsilon_i$ obtained from the DFT Kohn-Sham equation correspond to the non-interacting fictitious system, it turns out that electronic structure such as band structure, the density of states, \emph{etc.} are fairly well described for many materials and compounds in practice, except strongly correlated materials, \emph{etc.} It is therefore highly desirable to (1) develop machine learning approaches to predict full quantum mechanical Hamiltonian for arbitrary materials and molecules with arbitrary structures since it determines the underlying physical and chemical properties, and (2) develop more accurate density functionals beyond the state-of-the-art using machine learning methods under the known physical constraints. 

\revisionOne{There are a few major errors that arise in traditional DFT calculations with approximated XC energy functionals, such as delocalization error, self-interaction error, and strong correlation error~\cite{verma2020trends, bryenton2023delocalization, wrighton2023some}.  
In particular, self-interaction error is a type of delocalization error.
Specifically, the density of each electron is delocalized within the mean-field density and interacts with a field that already includes its own contribution (Equation~\ref{eq: kohn-sham-hartree-energy}), resulting in a spurious self-interaction. 
This error is a specific instance of a more general error arising from electron density delocalization and is especially prominent in predicting dissociation curves.
In contrast, the simple Hartree-Fock (HF) method has exact exchange energy which exactly cancels the excess energy from self-interaction error, thus the HF method is exact for single-electron systems. 
Hybrid DFT functionals incorporate a fraction of exact HF exchange energy which resolves part of self-interaction error, however the computational cost of the exact exchange energy is higher compared to typical GGA and LDA functional. 
In addition, for multi-electron systems, there arises correlation energy contributions from electron-electron interactions which can be categorized into static correlation and dynamical correlation. 
Typical DFT functionals miss the proper description of multi-determinant effects and dynamical electron fluctuations, resulting in the static and dynamical correlation errors. Advanced methods, such as dynamical mean-field theory, have been developed to improve the treatment of both static and dynamic correlation beyond DFT. However, they often come with a high computational cost and fall beyond the scope of this survey.}

 \subsubsection{Density Functional Theory In Practice}

{\ifshowname\textcolor{red}{Haiyang, revised by Xiaofeng}\else\fi}

Although DFT significantly simplifies the way of solving the many-body Schr\"odinger equation, the exact numerical solution of the Kohn-Sham equation in Equation~(\ref{eq: kohn-sham-equation}) in principle needs infinite spatial grids to represent the exact wavefunctions. To avoid modeling the infinite spatial space, a set of predefined basis functions $\left\{\phi_j \right\}$ is introduced to approximate the single electronic wavefunctions. Commonly used basis sets include plane-wave basis, real space grids, wavelets, and localized atomic basis sets such as Slater-Type Orbitals (STO)~\cite{Slater1930STO}, Gaussian-Type Orbitals (GTO)~\cite{Boys1950GTO}, and Numerical Atomic Orbitals (NAO)~\cite{Koepernik1999NAO, junquera2001numerical}. Here we focus on the localized atomic orbital basis functions $\left\{\phi_j \right\}$ with an analytical form which are often represented by the product of a radial function and spherical harmonics. Specifically, it is approximated using a linear combination of basis functions defined as
\begin{equation}
    \psi_i(\bm{r})=\sum_j c_{ij} \phi_j(\bm{r}),
\end{equation}
where $c_{ij}$ is the $j$-th coefficient of the electronic wavefunction $\psi_i$ associated with basis function $\phi_j$, forming wavefunction coefficient matrix $\bm{C}_{e}$. $\bm{C}_{e}$ can be obtained by solving the DFT Kohn-Sham equation in the matrix form as a generalized eigenvalue problem, 
\begin{equation}
    \bm{H} \bm{C}_{e} = \bm{\epsilon} \bm{S} \bm{C}_{e},
    \label{eq: kohn-sham}
\end{equation}
where $\bm{H}:= H_{\bf{DFT}} = H_{\bf{KS}}$ (Equation~(\ref{eq: kohn-sham-hamiltonian})) is Hamiltonian with $h_{i j} \equiv \braket{\phi_i| {\bm{H}}|\phi_j} = \int \phi_i^*(\bm{r}) {\bm{H}} \phi_j(\bm{r}) d\bm{r}$, 
incorporating the interactions of different particles,
$\bm{S} \in \mathbb{R}^{N_o \times N_o}$ is the overlap matrix with $\bm{S}_{ij} \equiv \braket{\phi_i|\phi_j} = \int \phi_i^*(\bm{r}) \phi_j(\bm{r}) d\bm{r}$, representing the integral of a pair of predefined orbital basis, and $\bm{\epsilon}$ is a diagonal matrix where each diagonal element  $\epsilon_{ii}$ represents the eigen energy for the corresponding eigen wavefunction $\psi_i$. Depending on the nature of the system, $\bm{C}_{e}$ and $\bm{H}$ may be $\in \mathbb{R}^{N_o \times N_o}$ or $\mathbb{C}^{N_o \times N_o}$, where $N_o$ is the number of orbitals, and each atom may have multiple associated orbitals.  

\begin{figure}[t]
    \centering
    \includegraphics[width=0.9\textwidth]{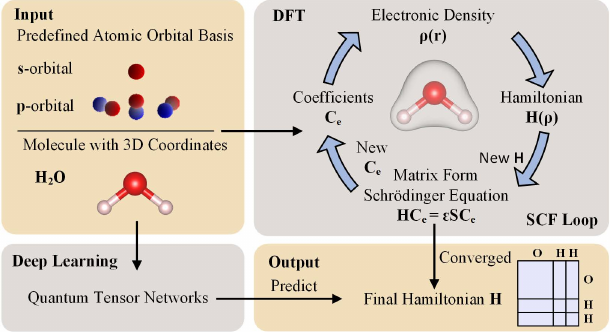}
    \caption{{\ifshowname\textcolor{red}{Haiyang, revised by Yaochen}\else\fi}The pipeline of the DFT calculations and deep learning methods to obtain the Hamiltonian matrix. The DFT calculation uses the predefined atomic orbital basis associated with the molecule and its coordinates to optimize Hamiltonian matrix $\bm{H}(\rho)$ iteratively within the SCF loop until it reaches convergence towards the total energy minimum/minima. In contrast, deep learning method uses the proposed quantum tensor networks to directly predict the final Hamiltonian matrix, taking atomic types and coordinates as inputs. This eliminates the iterative optimization, thereby accelerating the DFT calculations.}
    \label{fig:opendft:dft_scf}
\end{figure}

As discussed above, by mapping many-body interacting systems onto many one-body non-interacting systems using the Kohn-Sham approach~\cite{dft}, it is possible to compute electronic charge density using the single-particle electronic wavefunctions represented by wavefunction coefficients $\bm{C}_{e}$ and basis functions $\left\{\phi_j \right\}$. Consequently, the Hamiltonian matrix $H_{\bf{KS}}$ can be determined. To find $\bm{C}_{e}$ the solution of the Kohn-Sham equation (Equation~(\ref{eq: kohn-sham})), the SCF algorithm~\cite{payne1992rmp, cances2000convergence, kudin2002black} is commonly applied to improve the solutions $\bm{C}_{e}$ iteratively until the convergence is reached. When there are $N_T$ steps in total, the time complexity of the DFT algorithm is $\mathcal{O}(N_o^3 N_T)$, with each step being $\mathcal{O}(N_o^3)$. 
However, running iterative SCF calculations for large systems is computationally expensive.
To address this issue, deep learning models have been proposed to consider the interactions among the atoms and directly predict the Hamiltonian matrix. As shown in Figure~\ref{fig:opendft:dft_scf}, the final quantum tensors commonly obtained by self-consistent DFT calculations, such as Hamiltonian matrix, can be predicted by quantum tensor networks, which is discussed in detail in Section~\ref{sec:dft:QTLearning}. Meanwhile, another category of methods take use of optimization stratigies such as stochastic gradient descent~\cite{li2023d4ft} to replace the SCF loop accelerating the optimization stage.

{\ifshowname\textcolor{red}{Xiaofeng}\else\fi}According to \cite{Hohenberg64} and \cite{dft}, the ground-state total energy in Equation~(\ref{eq: kohn-sham-energy-functional}) is \textit{exact} for many-body system if we have the exact exchange-correlation energy functional $E_{\bf{XC}}[\rho]$. This can be more explicitly seen by re-writing the ground-state electronic total energy in Equation~(\ref{eq: kohn-sham-energy-functional}) using the Kohn-Sham eigen energies from the Kohn-Sham equation in Equations~(\ref{eq: kohn-sham-equation}) or (\ref{eq: kohn-sham}), 
\begin{equation}
 E_{\bf{KS}} = \sum_i f_i \epsilon_i + E_{\bf{XC}}[\rho] 
 - \frac{e^2}{8 \pi \varepsilon_0} \iint \frac{ \rho(\bm{r}) \rho(\bm{r'}) } {|\bm{r} - \bm{r'}|} d \bm{r} d \bm{r'} 
 - \int V_{\bf{ext}}(\bm{r}) \rho(\bm{r}) d \bm{r}.
\label{eq: kohn-sham-energy-functional-2}
\end{equation}
The ground-state density $\rho(\bm{r})$ uniquely determines the Hamiltonian $H_{\bf{KS}}$, thus determines Kohn-Sham eigen energies $\epsilon_i$. Subsequently, the second, third, and last terms of the above Kohn-Sham total energy are completely decided. While eigen energies $\epsilon_i$ of the first term and the Hartree energy of the third term are easy to evaluate, the key challenge of the Kohn-Sham DFT lies in the unknown exchange-correlation energy functional $E_{\bf{XC}}[\rho]$ and the corresponding exchange-correlation potential $V_{\bf{XC}}(\bm{r}) \equiv \delta E_{\bf{XC}}[\rho] / \delta \rho(\bm{r})$. Two main categories of XC energy functionals have been developed in the past, including the Local Density Approximation (LDA)~\cite{CA1980, VWN1980, PZ1981, PW1992} where the XC energy depends on the local electron density $\rho(\bm{r})$ only, and Generalized Gradient Approximation (GGA)~\cite{PBE1996, PW91, Becke1988, LYP1988} where the XC energy depends on both the local electron density $\rho(\bm{r})$ and its gradient $\nabla \rho(\bm{r})$. In addition, hybrid XC functionals~\cite{B3LYP1993, HSE03} have been proposed and widely used to (partially) include exact exchange, and meta-GGA functionals~\cite{TPSS2003,SCAN2015,SCAN2016,SCAN2020} have been proposed to include high-order gradients of electron density, kinetic energy density, \emph{etc.} These methods gradually climb Jacob's ladder~\cite{Perdew2001Jacobladder} with better accuracy at the price of higher computational cost. 
While all these approximations rely on the physical constraints and intuitions with great success in the fundamental materials and molecular research, the exact XC energy functional has not been achieved yet, presenting a unique opportunity for AI/ML approaches to tackle this challenge. We briefly summarize the recent progress in learning density functionals in Section~\ref{sec:dft:MLXCKE} and discuss potential future directions in this area in Section~\ref{sec:dft:future-directions}.

\subsection{Quantum Tensor Learning}
\label{sec:dft:QTLearning}
\noindent{\emph{Authors: Haiyang Yu, Zhao Xu, Limei Wang, Yaochen Xie, Xiaofeng Qian, Shuiwang Ji}}\newline

{\ifshowname\textcolor{red}{Haiyang}\else\fi}In DFT calculations, quantum tensors, such as Hamiltonian matrix $\bm{H}$ and wavefunction coefficients $\bm{C}_e$, can describe quantum states of physical systems and determine various critical physical properties, including total energy, charge density, electric polarization, \textit{etc}. To accelerate the DFT calculations, various deep learning models~\cite{schutt2019unifying, gastegger2020deep, unke2021se, deeph, gong2023general, li2023deep} have been proposed to directly predict the quantum tensors. The predicted quantum tensors are used to derive the physical properties at a reasonable level of accuracy, thereby accelerating the optimization process in the electronic structure calculations.

\subsubsection{Problem Setup}
In this section, we focus on the task of predicting the Hamiltonian matrix, which is the key quantum tensors in accelerating the DFT algorithm. We denote the input 3D molecule as $\bm{M} = (\bm{z}, C)$ consisting of atom types $\bm{z} = \left( z_1, \dots z_n \right) \in \mathbb{Z}^n$ and atom coordinates $C = \left[ \bm{c}_1, \dots, \bm{c}_n \right] \in \mathbb{R}^{3 \times n}$, where $n$ is the number of atoms. We aim to develop deep learning models to  predict target quantum tensors for input molecular geometries. Specifically, the Hamiltonian matrix $\bm{H} \in \mathbb{R}^{N_o \times N_o}$ is one of the prediction targets that can be used to derive various physical properties, where $N_o$ represents the number of electronic orbitals. 


\subsubsection{Technical Challenges}
\begin{figure}[t]
    \centering
    \includegraphics[width=\textwidth]{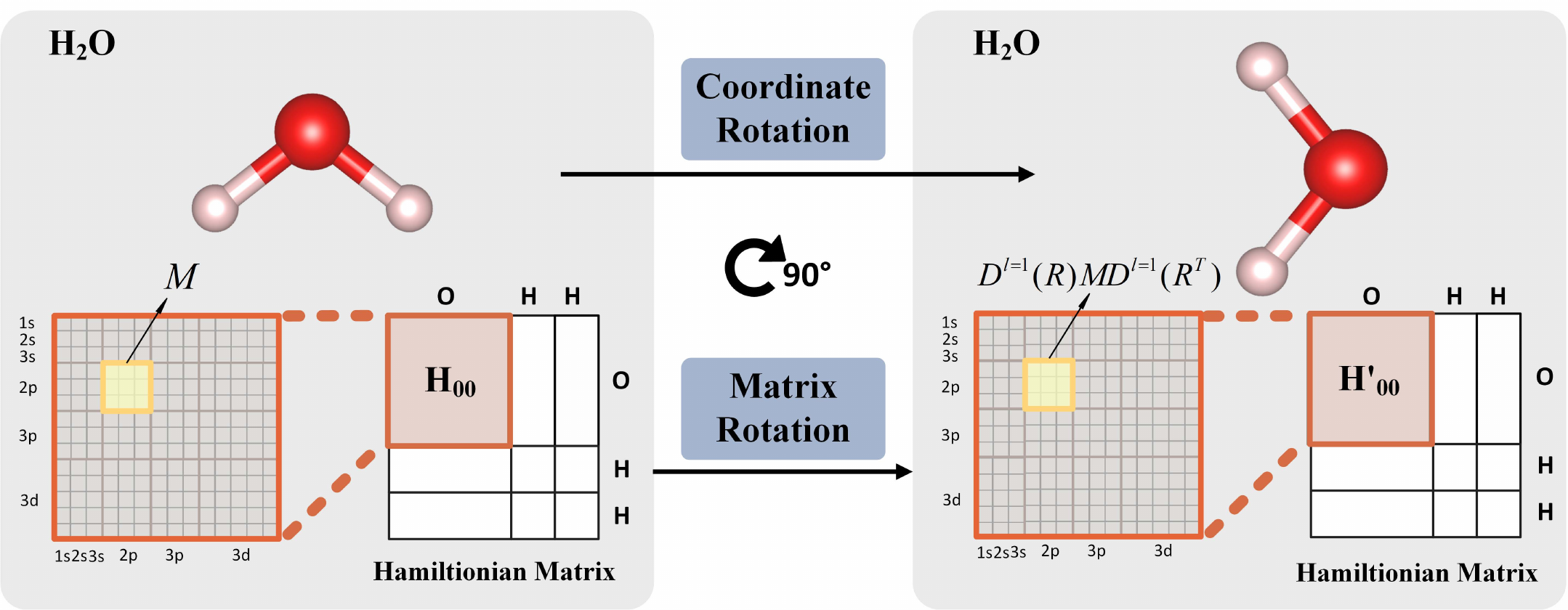}
    \caption{{\ifshowname\textcolor{red}{Haiyang, revised by Yaochen}\else\fi}The equivariance of Hamiltonian matrix $\bm{H}$.  When the coordinates of the molecule are rotated using a rotation matrix $R$, the corresponding Hamiltonian matrix is rotated accordingly. Specifically, the Hamiltonian block $B$ represents the orbital interaction between the oxygen $2p$ orbitals, and it is rotated to $D^{\ell=1}(R) B D^{\ell=1}(R^{T})$ using Wigner D-matrix $D^{\ell}(R)$.}
    \label{fig:DFT_equivariance}
\end{figure}

There are several challenges to tackle for quantum tensor learning. The first challenge is symmetry. Quantum tensor learning requires geometric deep learning models to guarantee the intrinsic permutation, translation, and rotation equivariance for quantum tensors. While geometric deep learning models usually maintain equivariant features, the predicted quantum tensors are generally composed of equivariant matrices. As shown in Figure \ref{fig:DFT_equivariance}, when the input molecule is rotated, the block $B$ divided by orbitals in the Hamiltonian matrix is rotated to $D^{\ell_1}(R) B D^{\ell_2}(R^{T})$, where $D^{\ell}(R)$ is the Wigner D-matrix for rotation $R$ with rotation order $\ell$. This raises the need to design equivariant architectures that build equivariant matrices from equivariant features. Another challenge is the flexibility of the model. Since the size of quantum tensors varies significantly with the chemical elements in the system, a flexible architecture is required to apply geometric deep learning models to diverse systems. Moreover, efficiency is also a challenge for equivariant networks. To maintain the equivariance, the operations in these models usually have a considerable computation cost compared to invariant networks.

\subsubsection{Existing Methods}


{\ifshowname\textcolor{red}{Haiyang}\else\fi}Currently, several graph neural networks are proposed to learn the interactions among the atoms and predict the quantum tensor like SchNorb~\cite{schutt2019unifying}, PhiSNet~\cite{unke2021se}, DeepH~\cite{deeph} and QHNet~\cite{yu2023efficient}. They are composed of three parts, including establishing node-wise interaction, building pairwise features, and constructing the quantum matrix. 
The node-wise interaction module encodes the atomic and geometric information between atoms, and builds the node-wise equivariant features using a message passing scheme~\cite{gilmer2017neural}. In addition, since the final Hamiltonian matrix is constructed with blocks representing the pairwise interaction of atoms, pairwise features are trained to learn such interactions.
Finally, the matrix building module expands the pairwise features to matrices and then constructs the final quantum tensors corresponding to the atomic orbitals of the input atoms.

{\ifshowname\textcolor{red}{Limei}\else\fi}

\vspace{0.1cm}\noindent \textbf{Node-Wise Interaction:} The node-wise interaction module is used to construct representations of atoms by aggregating information from neighbors following the Message Passing Neural Networks (MPNNs) framework~\cite{gilmer2017neural}. Specifically, the features $\bm{h}$ of each node $i$ in layer $t$ are updated based on 
\begin{equation}
\begin{aligned}
      \bm{m}_i^{t+1} &=\sum_{j\in\mathcal{N}(i)} M_t\left(\bm{h}_i^t, \bm{h}_j^t, \bm{h}_{ij}\right), \\
      \bm{h}_i^{t+1} &= U_t \left(\bm{h}_i^t, \bm{m}_i^{t+1}\right). \\
\end{aligned}
\end{equation}
Here $\mathcal{N}(i)$ is the neighboring node set of node $i$, $\bm{h}_{ij}$ is the edge feature between node $i$ and node $j$, $\bm{m}_{}^{t+1}$ is the hidden variable at node $i$, and $U_t$ and $M_t$ denote the update and message functions at layer $t$.
Note that the final prediction target, such as the Hamiltonian matrix, is an equivariant matrix, \emph{i.e.}, if the input molecule is rotated by a rotation matrix $R$, each block $B_{ij}$ of the Hamiltonian matrix $\bm{H}$ should be transformed to $D^{\ell_i}(R)B_{ij}D^{\ell_j}(R^T)$ accordingly, as shown in Figure~\ref{fig:DFT_equivariance}. Here $D^\ell(R)\in\mathbb{C}^{(2\ell+1)\times(2\ell+1)}$ is Wigner D-matrix of $R$. Therefore, it is crucial to ensure equivariance while constructing node features. One approach is to first construct invariant node features, followed by additional operations in the pairwise feature building and matrix construction steps to ensure or encourage equivariance. 
For example, SchNorb and DeepH construct invariant node features by aggregating the features and distances of neighboring nodes. Since the initial node features and distances are $SE(3)$-invariant, the constructed node features are also invariant.
Alternatively, another approach focuses on constructing equivariant node features directly. For example, PhiSNet and QHNet construct equivariant node features using spherical harmonics and tensor products, as introduced in Section~\ref{subsec:cont_equi}, in each message passing layer, which ensures their equivariance to continuous symmetry transformations. Specifically, the features of each node are obtained by aggregating the tensor product between the features of each neighboring node and an equivariant filter. The filter depends on the spherical harmonics of the direction vector. With such operations, the methods can ensure equivariance at each layer. It is worth noting that the computational cost of a tensor product is significantly larger than a linear layer due to the need to multiply node features with CG matrix for each path. QHNet is much more efficient than PhiSNet by reducing the number of tensor products in the network. 

\begin{figure}[t]
    \centering
    \includegraphics[width=0.99\textwidth]{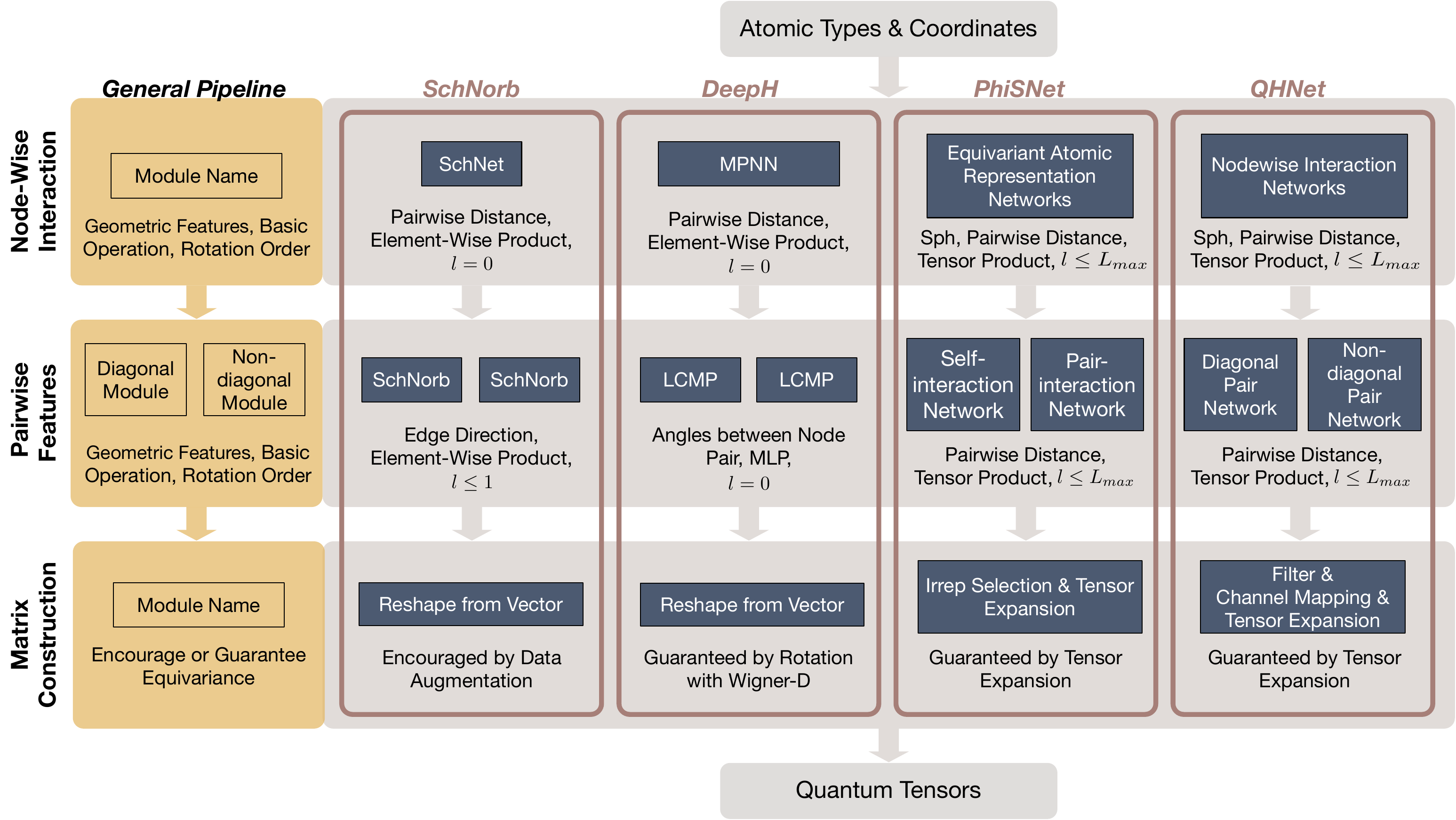}
    \caption{{\ifshowname\textcolor{red}{Zhao}\else\fi}An overview of quantum tensor networks and methods in AI for density functional theory. Quantum tensor networks take the atomic types and coordinates of a given molecule as input and output the predicted quantum tensor, such as the Hamiltonian matrix. Typically, quantum tensor networks consist of three sequential modules: the node-wise interaction module, pairwise feature building module, and matrix construction module. The node-wise interaction module updates the node features based on neighboring nodes within a cutoff distance. The pairwise feature module creates features to describe the relationship between atom pairs, with a diagonal module for pairs of a single atom and a non-diagonal module for pairs of two atoms. Pairwise features are then used to construct target matrices for node pairs, which are assembled to output the quantum tensor for the entire molecule. In node-wise interaction and pairwise feature modules, existing methods use different geometric features, basic operations, and rotation orders in their designs. Note that Sph denotes the spherical harmonics of edge direction. In the matrix construction module, the equivariance of the constructed matrix is either encouraged or guaranteed through various techniques. This figure provides module names and essential information about each module for existing methods, including SchNorb~\cite{schutt2019unifying}, DeepH~\cite{deeph}, PhiSNet~\cite{unke2021se}, and QHNet~\cite{yu2023efficient}.}
    \label{fig:opendft:main}
\end{figure}


{\ifshowname\textcolor{red}{Zhao}\else\fi}

\vspace{0.1cm}\noindent \textbf{Building Pairwise Features:} Since the quantum matrix encodes interactions between atom pairs, it is critical to construct pairwise feature vectors for atom pairs. A typical approach is to process diagonal pairwise features $f_{ii}$ and non-diagonal pairwise features $f_{ij}$ separately, as they correspond to pairs of single or two atoms. However, it's also possible to process diagonal and non-diagonal pairs in the same way. Furthermore, pairwise features can be either invariant or equivariant to rotation, depending on how edge orientation is used.
In DeepH, edge features are updated in its local coordinate message passing (LCMP) layer and then serve as pairwise features for atom pairs that are connected by these edges. Since self-loops are added in advance for diagonal atoms, LCMP can output both diagonal and non-diagonal features. Here, the obtained pairwise features are invariant to rotation ($\ell=0$) because edge orientations have been converted to local coordinates before inputting to the LCMP layer. SchNorb computes invariant scalar for each edge and uses it to scale edge orientation in global coordinates. Hence, scaled edge features are equivariant and of rotation order $\ell \leq 1$. Without self-loops, diagonal and non-diagonal pairwise features are then obtained from scaled edge features in different manners. Unlike the above two methods, QHNet and PhiSNet employ tensor products in their pairwise interaction modules, which leads to equivariant pairwise features of order $\ell \leq L_{max}$. Specifically, QHNet considers attentive scores of two atoms for non-diagonal pairs and uses tensor products for both diagonal and non-diagonal pairs. In contrast, PhiSNet uses the tensor product only for non-diagonal pairs and builds diagonal pairwise features like regular message passing.
Corresponding module names, rotation orders, and key elements used for building pairwise features in the above methods are summarized in Figure~\ref{fig:opendft:main}.

{\ifshowname\textcolor{red}{Zhao}\else\fi}

\vspace{0.1cm}\noindent \textbf{Matrix Construction and Equivariance:} After obtaining pairwise features, the final step is to build the quantum matrix, such as the Hamiltonian matrix. The molecular quantum matrix is composed of multiple pairwise blocks containing interactions between atoms. Here, the pairwise block is denoted by $B_{ij}$, where $o_i$ and $o_j$ are the numbers of orbitals of atom $i$ and $j$, respectively. Depending on the nature of physical properties and systems, $B_{ij} \in \mathbb{R}^{o_i\times o_j}$ or $\mathbb{C}^{o_i\times o_j}$. Since pairwise features are vectors, we must convert them into block matrices. SchNorb and DeepH reshape the flattened edge feature vector directly into a matrix, while QTNet and PhiSNet use tensor expansion to expand a single irrep vector into the matrix. Note that different atoms have varying numbers of orbitals. Thus, pairwise blocks $B_{ij}$ are in different shapes, but outputting blocks with various shapes is challenging. To build blocks with various shapes, SchNorb, DeepH, and QTNet firstly construct immediate blocks with full orbitals and then extract exact blocks according to atom types. In contrast, PhiSNet maintains a record to guide which channel of irrep should be selected to build the block for each pair. Finally,  blocks $B_{ij}$ are assembled to construct the whole quantum matrix for the molecular system. Among the above-summarized methods, QTNet and PhiSNet adopt tensor expansion to ensure rotation equivariance of the quantum matrix. DeepH applies the inverse of the Wigner D-matrix to convert the predicted local quantum matrix back to global coordinates, thereby ensuring its rotation equivariance. SchNorb cannot guarantee rotation equivariance, but it augments data via rotation to encourage the encoding of rotational symmetry. Figure~\ref{fig:opendft:main} summarizes how the above methods construct the matrix and maintain matrix equivariance.

\vspace{0.1cm}\noindent \textbf{Equivariant Networks on Quantum Tensor Predictions:}
Due to the intrinsic equivariance nature of Hamiltonian matrix and other quantum tensors, equivariant networks~\cite{unke2021se, yu2023efficient, gong2023general, zhong2023transferable, li2023deep} have become the mainstream methods to obtain these quantum tensors. Especially, tensor expansion stands out as a powerful technique for constructing equivariant matrix from equivariant features. 
As listed in Table~\ref{tab:opendft:equivariantQTNet}, these equivariant networks integrate tensor expansion with various techniques to construct Hamiltonian matrices that satisfy the intrinsic symmetries required for specific tasks. 
For the basic Hamiltonian matrix prediction, PhiSNet builds the entire matrix with the atom-orbital quadruple irrep selection. This selection assigns a channel index on the irrep used for tensor expansion to each quadruple $(\text{atom}_1, \text{atom}_2, \text{orbital}_1, \text{orbital}_2)$.
While QHNet follows the a fashion of TFN, it introduces learnable parameters in filter operation and maps the channels of output equivariant matrix block to the  full orbital matrix.
When considering the spin-orbital coupling, the spin equivariance follows rotation order $\ell = \frac{1}{2}$ and $\ell = -\frac{1}{2}$.
Addressing this non-integral equivariance challenge involves employing basis transformation techniques to revert the basis back to an integral basis. Similarly, the basis transformation technique can be used to resolve the time-reversal equivariance issue for Hamiltonian matrix with magnetic momentum.

\begin{table}[t]
    \centering
    \caption{Equivariant Quantum Tensor Networks. They develop different modules to apply tensor expansion to obtain equivariant matrix from equivariant features. Currently, batch training can be applied on QHNet with various molecules while other implementations focusing on single same system during training and testing. Meanwhile, the implementation of DeepH-E3, HamGNN and xDeepH can be applied on both molecule and material systems.}
    \resizebox{0.99\textwidth}{!}{
    \begin{tabular}{l|ccccc}
    \toprule
       Model    & Tasks & Parameters & Techniques in Matrix Construction \\
       \midrule
       PhiSNet~\cite{unke2021se} &  Hamiltonian                             & \xmark & Atom-orbital quadruple Irrep selection \\
       QHNet~\cite{yu2023efficient}    &  Hamiltonian                       & \cmark & Filter \& Channel mapping  \\
       DeepH-E3~\cite{gong2023general} &  Hamiltonian with spin             & \xmark & Basis transformation  \& Orbital pair irrep selection\\
       HamGNN~\cite{zhong2023transferable}   &  Hamiltonian with spin       & \xmark & Basis transformation  \& Orbital pair irrep selection  \\
       xDeepH~\cite{li2023deep}   &  Hamiltonian with spin and magnetic momentum   & \xmark & Basis transformation  \& Orbital pair irrep selection \\
    \bottomrule
    \end{tabular}
    }
    \label{tab:opendft:equivariantQTNet}
\end{table}

\subsubsection{Datasets and Benchmarks} 
For the quantum tensor learning task, MD17~\cite{schutt2019unifying, gastegger2020deep} provides Hamiltonian matrices for the molecular geometries in the trajectories, with each trajectory corresponding to a single molecule. MD17 consists of 4 molecules: water, ethanol, Malondialdehyde, and Uracil. Furthermore, mixed MD17~\cite{yu2023efficient} combine four molecular trajectories in the MD17 dataset together and provide a quantum tensor dataset with multiple molecules in the training and testing sets. 


However, commonly used MD17 dataset contains four distinct molecules. 
To enhance the generalization ability to the molecule space, Quantum Hamiltonian (QH9) dataset~\cite{yu2023qh9} is a dataset of precise Hamiltonian matrices of molecular geometries. 
The open-source software PySCF (Python‐based simulations of chemistry framework)~\cite{sun2018pyscf} is used to compute the Hamiltonian matrix. 
QH9 consists of two kinds of datasets: static and dynamic. 
\begin{itemize}
\item The QH-stable dataset consists of 130,831 stable molecular geometries, coming from a subset of the QM9 dataset~\cite{ramakrishnan2014quantum}. To explore the performance in both in-distribution and out-of-distribution (OOD) scenarios, two tasks are created on the dataset: (1) random split creates the QH-stable-iid; (2) split based on the number of constituting atoms would yield QH-stable-ood. 
\item The QH-dynamic dataset contains 2,998 molecular dynamics trajectories, where each trajectory has 100 geometries. Two split strategies on QH-dynamic yield two tasks: (1) QH-dynamic-300k-mol splits training/validation/test set based on different molecules; (2) QH-dynamic-300k-geo allows different geometries of the same molecule in training, validation, and test set. 
\end{itemize}
The statistics of these datasets are shown in Table~\ref{tab:dft_data}. 
Based on the curated QH9 dataset, \citet{yu2023qh9} also demonstrates that the state-of-the-art method QHNet~\cite{yu2023efficient} is capable of predicting Hamiltonian matrices for molecules of any kind and generalized well on unseen molecules. 
Specifically, QHNet trained on QH9 reached an MAE (Mean Absolute Error, between predicted Hamiltonian matrix and groundtruth) of $83.12 \times 10^{-6} E_h$ on the mixed MD17 dataset.

\begin{table}[t]
\centering
\caption{Statistics of datasets for quantum tensor learning, including MD17~\cite{chmiela2017machine}, mixed MD17~\cite{yu2023efficient}, QH9~\cite{yu2023qh9}. 
}
\resizebox{0.95\textwidth}{!}{
\begin{tabular}{l|ccccc}
\toprule
Datasets & \# geometries & \# molecules & \# training & \# validation & \# test \\\midrule
MD17-water & 4,900 & 1 & 500 & 500 & 3900\\
MD17-ethanol & 30,000 & 1 & 25,000 & 500 & 4,500 \\ 
MD17-Malondialdehyde & 26,978 & 1 & 25,000 & 500 & 1,478 \\ 
MD17-Uracil & 30,000 & 1 & 25,000 & 500 & 4,500  \\ 
mixed MD17 & 91,878 & 4 & 75,500 & 2,000 & 14,378 \\ 
QH-stable-iid & 130,831 & 130,831 & 104,664 & 13,083 & 13,084 \\ 
QH-stable-ood & 130,831 & 130,831 & 104,001 & 17,495 & 9,335 \\ 
QH9-dynamic-300k-geo & 299,800 & 2,998 & 239,840 & 29,980 & 29,980 \\ 
QH9-dynamic-300k-mol & 299,800 & 2,998 & 239,800 & 29,900 & 30,100 \\
\bottomrule
\end{tabular}}
\label{tab:dft_data}
\end{table}

\subsubsection{Open Research Directions}
{\ifshowname\textcolor{red}{Haiyang, revised by yaochen, and Xiaofeng}\else\fi}
Quantum tensors from DFT, such as Hamiltonian matrix $\bm{H}$ and density matrix $\bm{D}$, are not only useful for computing the electronic structures, but also can serve as physics-based features to predict accurate molecular properties such as total energy and electronic polarization. The reason why PhiSNet, SchNorb, DeepH, QHNet/QTNet, \emph{etc.} with small $\ell$ cutoff work so well for molecules and materials lies in the fact that most of these compounds are dominated by low-energy chemistry/physics where the eigen wavefunctions possess significant atomic-orbital like characteristics. For the same reason, one can directly construct atomic-like maximally localized Wannier functions (MLWFs)~\cite{Marzari1997mlwf,Souza2001mlwf,Marzari2012rmp} and quasi-atomic orbitals (QOs)~\cite{qian2008qo, qian2010qotransport} from molecular orbitals or eigen wavefunctions and obtain accurate Hamiltonians for the low-energy regime (\emph{e.g.}, from the lowest eigen energies to a few eVs above the Fermi level), which has enabled the discovery of novel materials and physics such as quantum spin Hall insulators~\cite{Qian2014QSHI,Marrazzo2018QSHI}, Berry curvature memory effect~\cite{Wang2019BerryCurvatureMemory, Xiao2020BerryCurvatureMemory}, and nonlinear photocurrent~\cite{Wang2020GeneralizedSC}. Accurate prediction of quantum tensors such as Hamiltonian will be highly crucial and valuable for accelerating the materials discovery in future, as recently demonstrated by DeepH~\cite{deeph, gong2023general, li2023deep}. Furthermore,
existing work OrbNet~\cite{qiao2020orbnet} employs the quantum tensors as node and edge features in the orbital graphs, resulting in a significant enhancement of molecular energy prediction performance.
Due to the significant time and computational cost associated with DFT calculations, OrbNet uses GNF-xTB~\cite{grimme2013simplified, grimme2016ultra}, a fast semi-empirical method, to approximate the quantum tensors.
The quality of these approximated quantum tensors directly influences the performance of deep learning models.
Therefore, it is crucial to address the challenge of obtaining accurate quantum tensors within a reasonable time to build the physical-based input features~\cite{Bai2022graph}.
Quantum tensor networks have the potential to address this challenge by accurately predicting quantum tensors to construct physics-based features for deep learning models, such as accurate deep learning force field for studying both electronic and structural phase transition in quantum materials~\cite{Li2021natrevmat}.

\subsection{Density Functional Learning}
\label{sec:dft:MLXCKE}

\noindent{\emph{Authors: Alex Strasser, Xiaofeng Qian}}\newline

{\ifshowname\textcolor{red}{Alex, Xiaofeng}\else\fi}

Machine learning has been applied to model density functionals in order to predict the exchange-correlation energy functional, kinetic energy functional~\cite{Snyder2012MLXC} for orbital-free DFT, the universal functional, corrections to density functional, and more. The approaches vary widely, with most being numerical, but some are some symbolic~\cite{Ma2022Evolving}. Some predictions start from scratch, some start from the previously established functionals~\cite{Zheng2004MLXC}, or from both~\cite{Ma2022Evolving}, and other approaches incorporate exact physical constraints into the functional form~\cite{Hollingsworth2018Exact,Pokharel2022Exact,nagai2022machine,dick2021highly,Gedeon2022MLXCDiscontinuity}. More details can be found in the recent perspectives and review articles~\cite{Burke2015PerspectiveMLXC,Manzhos2020MLDFT,Kalita2021Learning,perdew2021artificial,Pederson2022MLDFT,Fiedler2022MLDFT,Kulik2022Roadmap,Nagai2023development}.

\subsubsection{Machine Learning Exchange-Correlation Energy Functionals}
\label{sec:dft:MLXCKE:MLXC}

The exchange-correlation (XC) energy (the last term in~\cref{eq: kohn-sham-energy-functional}) is the most challenging part of the DFT-KS equation. Most widely used XC energy functionals are designed in analytical forms and fitted to various sets of known physical constraints, such as LDA and GGA mentioned above. Many of the calculations in applying ML to XC energy functionals implement a well-known quantum chemistry software called PySCF (The Python-based Simulations of Chemistry Framework)~\cite{sun2018pyscf}, \emph{e.g.},  \citet{nagai2020completing,nagai2022machine,kirkpatrick2021pushing,bystrom2022cider}. While great improvement has been demonstrated by incorporating more exact constraints~\cite{TPSS2003, SCAN2015, SCAN2016, SCAN2020}, they are still approximations of the unknown exact XC energy functional. Machine learning can be used to enhance the accuracy of XC functionals, which are the primary source of error in typical KS-DFT calculations~\cite{kim2013understanding,crisostomo2023seven}, so we summarize some of the progress in this area. 

The application of ML to XC functionals started with~\citet{Tozer1996MLXC}, which used a neural network to approximate the XC potential using the ZMP density inversion method~\cite{ZMP1994}, resulting in geometries comparable to LDA and substantially more accurate eigenvalues. Another early study~\cite{Zheng2004MLXC} improves the widely-used B3LYP functional, using a neural network to optimize the three parameters used to determine the relative contributions of exact exchange functional, local spin density exchange functional, Becke88 exchange functional, as well as the LYP and VMN correlation energy functionals.

Recently, thanks to the rapid development of deep learning methods and their surprising capability to capture nonlinear patterns, data-driven approaches have been used to estimate the precise exchange-correlation energy functional~\cite{Bogojeski2020MLXC,nagai2020completing,dick2021highly,kirkpatrick2021pushing,bystrom2022cider,trepte2022data,sparrow2022uniting,bystrom2023nonlocal,Dick2019MLCF,Dick2020NeuralXC,Ryabov2020NNXC,Lei2019invariantdescriptors,kasim2021learning}, which demonstrates exceptional accuracy on main-group chemistry and represents a state-of-the-art achievement in the field. Unlike approximation techniques, data-driven approaches can learn a theoretically unbiased (exact) estimator of the XC energy functional from real data because they do not impose any approximations on the functional form. 
Specifically, to learn the exact XC energy functional from data, \citet{nagai2020completing} builds a multiple layer perceptron (MLP) to learn the mapping from local density descriptor (human-curated feature) to local XC potential functional. However, this early attempt is an end-to-end paradigm, purely learning from data, and does not consider any physical constraints on the DFT system. It typically requires a large number of data points to reach desirable performance. 

Since XC functionals that include exact constraints tend to have more predictive power and are more generalizable~\cite{kaplan2023predictive}, identifying ways of incorporating these constraints into an ML-based XC functional is important. The two general approaches of imposing these exact constraints on an ML XC functional are 1) analytical, which guarantees adherence, and 2) data-driven, which primarily comes from training the ML model on data that obeys those constraints, such as data produced using the SCAN functional, and will not guarantee perfect adherence. Of the 17 known exact constraints for a semi-local XC functional, \citet{Pokharel2022Exact} argues that six of these constraints would be conducive to an analytical application in ML models through post-processing steps, input restrictions, or choosing separate exchange and correlation models. These constraints include (for the exchange energy) negativity, spin-scaling, uniform density scaling, a tight bound for two-electron densities, (for the correlation energy) non-positivity, and (for exchange and correlation together) the general Lieb-Oxford bound. 

Several works combine data-driven XC energy fitting with exact physical constraints, including fractional charge/fractional spin constraints~\cite{kirkpatrick2021pushing}, linear/nonlinear constraints~\cite{sparrow2022uniting}, physical asymptotic constraints~\cite{nagai2022machine}, and others~\cite{trepte2022data,brown2021mcml}. 
For example, one essential constraint of the DFT system is that electrons are treated as a continuous charge density rather than discrete particles. However, the continuous XC functionals cannot handle the derivative discontinuity of the XC energy at integer–electron numbers~\cite{Perdew1982derivativediscontinuities,Perdew1983derivativediscontinuities}. To meet this physical constraint and address DFT's problematic delocalization error~\cite{bryenton2023delocalization},~\citet{kirkpatrick2021pushing} defines a fictitious system to enable fractional charge and spin,  takes local features of electron density and trains a neural network to estimate the local energies, which are aggregated to obtain the XC energy. The resulting functional, DeepMind21 (DM21), demonstrates excellent performance on a bond-breaking dataset as well as across the QM9~\cite{ramakrishnan2014quantum} and GMTKN55~\cite{goerigk2017look} databases, superior to the three best hybrid functionals tested and reproducing multiple disassociation curves~\cite{kirkpatrick2021pushing}. \revisionOne{The generalizability of DM21 has been challenged by~\cite{gerasimov2022commentdm21}, such as whether the part of the bond breaking benchmark data was implicitly included in the training data leading to ``memorization'' vs ``understanding'', etc. \citet{kirkpatrick2022responsetocommentdm21} pointed out in response that DM21 differs from the infinite separation limit, etc. Recently, \citet{zhao2024dm21} benchmarked the DM21 functional on transition metal chemistry (TMC) and found that DM21 achieves comparable or sometimes better accuracy than hybrid B3LYP functional, however, DM21 faces challenges in achieving self-consistent field convergence for TMC molecules. Nevertheless, the performance of DM21mu supports the conclusion of imposing exact constraints to enhance generalization of machine learning functionals.}

\citet{Gedeon2022MLXCDiscontinuity} proposes an approach to train a $N_e$ neural network with a piece-wise linearity which reproduces the derivative discontinuity of the XC energy.
Similarly, \citet{sparrow2022uniting} designs a novel set of bell-shaped spline functions as the basis to embed the linear and nonlinear constraints as well as incorporate the implicit smoothness constraint as a regularization term in the learning objective. One group examined the effects of imposing a spin-scaling constraint and the general Lieb-Oxford bound when attempting to reproduce the SCAN functional in a deep neural network, showing improvements from the constraints but limited generalizability when attempting to move from data without chemical bonding to chemically bound systems~\cite{Pokharel2022Exact}.  
Furthermore, to satisfy physical asymptotic constraints, \citet{nagai2022machine} breaks down the XC energy functional into different terms (\emph{e.g.}, spin-up exchange, spin-down exchange energy, correlation energy), analytically imposes asymptotic constraints on different neural modules and aggregates all the NNs' output. In total, they analytically imposed 10 constraints -- 5 for the exchange part and 5 for the correlation part, and the physical constraints on the neural network enabled convergence in cases where the unconstrained NN did not converge. The application of the constraints was made easier by using the SCAN XC functional as a base, but they provide a way of imposing the same constraints for other base XC functionals. 
\citet{bystrom2022cider,bystrom2023nonlocal} approximate the exchange energy functional using a nonparametric estimation that measures the similarity between the current data and existing data points using a Gaussian process model~\cite{williams2006gaussian}, while imposing the uniform scaling constraint. The resulting significant accuracy improvement when testing on the Minnesota Database 2015B~\cite{yu2016mn15L} and also shows promising generalizability to solid-state systems. Finally, \citet{dick2021highly} impose a local Lieb-Oxford bound, finding this constraint to aid generalizability along with the uniform scaling, spin-scaling, and non-negativity. Trained on the SCAN results of 21 molecules, the neural network was tested on the diet-GMTKN55 dataset~\cite{gould2018diet} and was competitive with SCAN and hybrid functionals.

Another development addresses the issue of only using the converged energies and densities to train a model by allowing information about each iteration of the self-consistent KS solution to backpropagate through a deep neural network -- a Kohn-Sham regularizer~\cite{li2021kohn}. With this extra data while training on only two exact energies and densities in a 1D H$_2$ dissociation curve, the model was able to achieve chemical accuracy for the entire dissociation curve. The authors later extend this model to include spin-density for spin-polarized systems and test on weakly correlated systems, the domain of standard DFT calculations~\cite{kalita2022well}. They find that incorporating spin-density while training on energies and densities on atoms substantially reduces the error and improves convergence for equilibrium molecules, approaching chemical accuracy. As another way to go beyond converged energies in training, \citet{dick2021highly} assign an explicit function of iteration number in the loss function in order to penalize the slow convergence, leading to a smooth functional without convergence issues. The use of automatic differentiation enabled the extraction of more information contained in the electron density, thereby further expanding the training inputs. The same authors developed a metric for assessing XC functionals based on both energy and density errors since both are approximated in DFT calculations.


\subsubsection{Machine Learning Kinetic Energy Functionals}

{\ifshowname\textcolor{red}{Alex, Xiaofeng}\else\fi}

Rather than learning the XC functional, a different approach is to learn a density functional for the kinetic energy (KE), that is, KEDF. Compared to the KS approach, where the KE operator acts on the KS orbitals, data-driven machine learning KEDF instead allows one to neglect the KS orbitals entirely, resulting in the so-called orbital-free DFT (OF-DFT). While KS-DFT scales with $\mathcal{O}(N_e^3)$, OF-DFT scales quasi-linearly. One difficulty with this approach is that the gradient descent method used to find the energy minimum requires an accurate gradient, but the gradient of the KE functional is noisy and not well-behaved. The functional derivative of KE arises from varying Equation~(\ref{eq: kohn-sham-energy-functional}) with respect to electron density, but retaining a general form of KE rather than the quantum mechanical KE operator as in Equation~(\ref{eq: kohn-sham-kinetic-energy}). In OF-DFT, the focus is approximating the KE functional, such as the very first DFT method using the Thomas-Fermi model. The KE functional derivative is used explicitly in the Euler-Lagrange Equation~(\ref{eq: Euler-equation-KS}) in order to find the self-consistent density, given as 
\begin{equation}
\frac{\delta T_{\mathrm{s}}}{\delta \rho(\mathbf{r})} = \mu - V_{\mathrm{eff}}(\mathbf{r}),
\label{eq: Euler-equation-KS}
\end{equation}
where $T_s$ is the non-interacting KS kinetic energy, $V_{\mathrm{eff}}$ is the effective or KS potential, and $\mu$ is the chemical potential which ensures the constraint of total $N_e$ electrons. A substantial amount of work has been done to address the noisy functional derivative problem, such as the development of nonlinear gradient denoising~\cite{Snyder2015Denoising}. One reason this is relevant is that the kinetic energy is in the same order as the total energy and is significantly larger than the XC energy, so an inaccurate approximation of the KE has a much larger impact than an inaccuracy in an XC approximation.

In the first pioneering work to apply ML for KE density functional approximation~\cite{Snyder2012MLXC}, a kernel ridge regression was used to approximate a KE density functional, achieving chemical accuracy (mean absolute error below 1 kcal/mol) in a 1D analog to OF-DFT of noninteracting fermions in a 1D box. The same approach has been explored in more detail~\cite{li2016understanding} and also applied to bond breaking for various 1D diatomic molecules, achieving chemical accuracy with 20 training data points, much better than usual OF-DFT errors~\cite{Snyder2013OFDFT}. On ten atoms ranging from H to Ne and 19 molecules,~\citet{seino2018semi} uses density and its gradients up to the third order as explanatory variables in a neural network, demonstrating superiority over the majority of the other 27 semi-local KE density functionals in comparison.~\citet{golub2019kinetic} show that using up to a fourth-order term in a gradient expansion of the kinetic energy density as an input into a neural network allows OF-DFT to reproduce the KS kinetic energy density very closely, for both solid state and molecular systems. Rather than using the density gradient directly,~\citet{yao2016kinetic} show that the reduced density gradient, a dimensionless quantity, may be more informative as a neural network input, demonstrating strong predictive power of a convolutional neural network for seven alkanes with better performance than other GGAs at hydrocarbon bonding. One recent work~\cite{Ryczko2022MLDMC} uses slices of electron density in a voxel deep neural network (VDNN) to predict the kinetic energy of a graphene lattice within chemical accuracy, and they also showed that one can use Monte Carlo-based optimization instead of gradient-based optimization for a 1D model system. 

Just as in the XC case, KE functionals have exact constraints that must be applied to find physically motivated KE functionals, such as Pauli positivity, asymptotic limit, and coordinate scaling~\cite{levy1988exact, holas1995exact, aldossari2023constraint}. Similarly, there have been attempts to impose these exact constraints when developing ML KE functionals, the first of which applied the coordinate scaling condition in a 1D analog test~\cite{Hollingsworth2018Exact}. Imposing the constraint for an ML functional by kernel ridge regression showed substantially improved accuracy for a 1D Hooke's atom case, but no improvement in the case of bond stretching in a 1D H$_2$ study. The Pauli positivity condition requires that the Pauli potential over all space is non-negative. It is satisfied if the KEDF only includes the Thomas-Fermi and von Weizs{\"a}cker terms, but it is not met in the fourth-order expansion used by~\citet{golub2019kinetic}. Neither the scaling or asymptotic limit conditions are met by~\citet{yao2016kinetic}, although the authors point out that the physical constraints can be met in their approach via the training data fed to the convolutional neural network.

Finally, some studies aimed to learn a density functional other than XC or KE, such as the total energy functional, or learn corrections to a density functional rather than the functional itself~\cite{Bogojeski2020MLXC,Mezei2020Noncovalent}. One study developed a kernel ridge regression model to learn the difference ($\Delta$-learning~\cite{Ramakrishnan2015DeltaLearning}) in the energies from a low-level (\emph{e.g.} DFT) calculation and a high-level calculation using coupled-cluster with single, double, and perturbative triple excitations (CCSD(T))~\cite{Bogojeski2020MLXC}. This correction factor is a functional of the input DFT densities, and it allows for highly accurate predictions with the low computational cost of a standard KS-DFT calculation that scales with $\mathcal{O}(N_e^3)$ instead of $\mathcal{O}(N_e^7)$ for the CCSD(T) calculations. Another interesting approach uses machine learning to recommend the best already-established density functional approximation for a given system, outperforming $\Delta$-learning models as well as any of the other 48 tested approximations~\cite{duan2023transferable}. Perhaps the most unique development is to learn the map from potential to density directly, called a Hohenberg-Kohn (HK) map, which can be done at a smaller computational cost and avoid the problem of the functional derivative~\cite{Brockherde2017Bypassing}.



\subsubsection{Datasets and Benchmarks}

{\ifshowname\textcolor{red}{Alex, Xiaofeng}\else\fi}


Compilations of highly accurate chemical data calculated at higher theory levels than DFT, such as CCSD(T), serve as an indispensable resource for the development of ML density functionals. These datasets can be used to train and test the accuracy of a density functional and compare with the performance of other functionals, especially useful for ML-based density functionals that require large amounts of highly accurate data for their training. Great efforts have been made to develop high quality datasets. For example, ACCDB is a collection of chemistry databases~\cite{morgante2019accdb}, including five previously established databases (GMTKN, MGCDB84~\cite{mardirossian2017thirty}, Minnesota2015, DP284~\cite{hait2018dipole,hait2018polarizability}, and W4-17), two new reaction energy databases automatically generated~\cite{margraf2017automatic} (W4-17-RE from W4-17 and MN-RE from Minnesota 2015B), and a new database for transition metals, which can be used as a benchmark for the development of density functionals. The GMTKN database consists of GMTKN55~\cite{goerigk2017look} and MB08-165~\cite{korth2009mindless}. The Minnesota database includes Database 2015~\cite{haoyu2015nonseparable}, Database 2015A~\cite{yu2016mn15}, and Database 2015B~\cite{yu2016mn15L}. Collectively, the ACCDB contains 10,049 structures and 8,656 unique reference data points (44,931 if the reaction energies are included), providing a substantial amount of high-quality data for training and testing. Another useful dataset is the SOL62 database~\cite{trepte2022data,zhang2018performance}, consisting of the cohesive energies of 62 solids (40 non-metals and 22 metals). 

All these databases can be used as training and testing data for the development of machine-learned density functionals, as well as the evaluation of the functional. The accuracy of the prediction compared to those from higher theory levels around ten times more accurate enables a good benchmark for comparison, and several of these databases include the information of accuracy comparison for many other XC functionals. The datasets are split into various subsets to evaluate the functionals in different applications, such as atomization energies, barrier heights, bond energies, noncovalent interactions, and more.  

In summary, many different machine learning approaches have been developed which provide more and more accurate density functionals (such as XC or KE functionals) and improve the predictions, even approaching chemical accuracy, with much lower computational cost compared to the higher level calculations. In Section~\ref{sec:dft:future-directions} we discuss potential directions in this area for further exploration.

\subsubsection{Open Research Directions}
\label{sec:dft:future-directions}

{\ifshowname\textcolor{red}{Alex, Xiaofeng}\else\fi}

\vspace{0.1cm}\noindent \textbf{ML-Based XC Functional:} 
One exciting area of research is using symbolic regression to find the mathematical expression of density functionals that best fits the dataset, offering a mechanism of creating functionals from scratch or improving previously known functionals. \citet{Ma2022Evolving} is able to reconstruct a previously known functional from a starting point of a small library of mathematical instructions (\emph{e.g.}, multiplication, exponentiation, or building blocks of existing functionals) using an evolutionary search procedure. They used the same method to iterate the $\omega$B97M-V functional through a regularized evolutionary algorithm in order to get a new functional -- Google Accelerated Science 22 (GAS22), with improved error on the MGCDB84 dataset. This fundamentally novel approach can be applied to other density functionals, such as KE functional for OF-DFT or  universal functional, with a larger library of mathematical instructions for a wider search space which is more likely to find improved or novel functionals. The initial library was limited to only four (conveniently chosen) instructions in the first case, and for the second case the library included five arithmetic operations, six simple power operations, and the enhancement factors in the PBE, RPBE, B88, and B97 functionals. This library can be expanded to include more mathematical operations as well as the components of many previous functionals that were included in this study. Furthermore, the method can include density information, regularization, and exact constraints to restrict the functional form to be more physically motivated. 

Another symbolic approach to ML for density functionals is through the use of automated feature engineering (AFE) and Q-learning, a reinforcement learning method. A recent investigation has used AFE to produce a feature generation tree where features are combined by mathematical operations into a physically meaningful equation, the exploration of which is guided by a deep Q-network (DQN)~\cite{xiang2021physics}. The result demonstrated the improved classification and regression scores with less computation time for three materials databases compared to primary feature sets as well as another recent feature generation and selection technique, SISSO~\cite{ouyang2018sisso}. This novel method may be leveraged towards the discovery of an analytical XC functional by using AFE to produce equation-like features and applying the DQN to select the optimum features for further exploration. Thus, this AFE+DQN method offers another promising approach to the discovery of symbolic XC functionals, in addition to the evolutionary automated ML approach described above. 

\revisionOne{
Addressing the major errors of DFT, such as delocalization error, self-interaction error, and strong-correlation issues, can potentially be addressed with machine learning. 
One of the potential approaches is to train on data that meets exact constraints, even if the model outputs are not forced to meet those constraints, such as machine learning XC functionals DM21 and DM21mu
DM21 was trained on fractional charge and fractional spin systems, and DM21mu was trained on data that meets the uniform electron gas constraint~\cite{kirkpatrick2021pushing}. Both significantly outperform DM21m which was trained without the constrained data. Regarding the challenges in strongly correlated systems, DFT+U is a computational technique that adds a Hubbard U parameter to account for strong interactions, such as properly describing a metal-insulator transition. Machine learning can help address this strong correlation error, such as by using Bayesian optimization to predict the Hubbard U parameter in a way that is more accurate than the conventional linear response approach and has been successfully applied to transition metal oxides, europium chalcogenides, and narrow-gap semiconductors~\cite{yu2020DFTUBayes}. Finally, it is possible to impose exact constraints on the outputs of the ML model itself in neural network models or symbolic regressions. Previous examples were mentioned in Section~\ref{sec:dft:MLXCKE:MLXC}, but more exact constraints remain to be explored~\cite{Pokharel2022Exact}.}


\vspace{0.1cm}\noindent \textbf{ML-Based KE Functional:}  With the substantial scaling advantage of OF-DFT to KS-DFT, there is significant interest in developing accurate kinetic energy density functionals (KEDFs), and machine learning methods have demonstrated powerful towards that end. One area of future development in this area is to impose physical constraints on those KEDFs for ML models, whether that is through an analytical approach or a data-driven approach. It is still not entirely clear the full impact of imposing the exact constraints on the KEDFs on the accuracy of an ML model. In fact, because the KE functionals are much less well-explored compared to XC functionals, more work is needed to even establish these physical constraints by studying the behavior of the exact KEDF, \emph{e.g.} addressing six open questions regarding the exact KEDF raised in a recent review article~\cite{wrighton2023some}. 

While there has been much work addressing the noisy functional derivative of the KEDF, there are more opportunities for avoiding gradient-based optimization entirely, such as using Monte Carlo-based optimization and erasing the need for evaluating a functional derivative of the KE, a method that still needs to be extended to the three-dimensional case~\cite{ryczko2022toward}. There is also room for exploring more methods that do not make use of the functional derivative~\cite{Brockherde2017Bypassing}. Furthermore, exclusively learning on converged densities and energies limits the accuracy of ML models. While there have been steps forward to expand the kinds of training data used with a KS regularizer~\cite{li2021kohn,kalita2022well} or automatic differentiation~\cite{dick2021highly}, one further possibility is to find the converged external potentials to which the unconverged densities correspond to. Some insightful suggestions have been made along these lines by \cite{ryczko2022toward}. Another issue which is repeatedly raised is the non-uniform sampling and density distributions, which cause issues for ML models. 

{\ifshowname\textcolor{red}{Alex}\else\fi}Overall, there has been substantial progress in ML for density functionals, among which the improvement that arises from the imposition of physical constraints is particularly interesting. One area for future improvement is that many of these functionals are fitted to or tested only on atomic or molecular systems, and their generalizability to solid-state systems needs to be further tested~\cite{perdew2021artificial,Pokharel2022Exact,Pederson2022MLDFT}. Therefore, the use of datasets such as LC20 from~\cite{sun2011self} or SOL62~\cite{trepte2022data,zhang2018performance} for training and testing may advance ML density functionals for solid-state systems~\cite{bystrom2023nonlocal}. Similarly, most of the ML-based density functional predictors train their ML model with a small number of molecules, making their model hard to generalize in unseen molecules or solid-state systems that are different from the training data. For example, \cite{nagai2020completing} incorporates three small molecules in the training set, which showed promising results across first- and second-row molecules, but it hasn't been tested on transition metal or extended systems yet. Since different kinds of molecules vary greatly in their properties, \emph{e.g.}, organic and inorganic, small molecules and macromolecules, and open-shell and close-shell molecules, developing a ML-based density functional that generalizes well across different groups of molecules as well as solids is an important direction to explore in the future.  

{\ifshowname\textcolor{red}{Alex}\else\fi}
\vspace{0.1cm}\noindent \textbf{Accelerating KS-DFT Optimization:}
ML can be used to improve the convergence and/or reduce the computational complexity for KS-DFT. For example, a recent work ~\cite{li2023d4ft} employed stochastic gradient descent on the energetic quantities and embedded the orthonormality constraint on the wave functions as part of the objective function, which allows the prediction of the ground state energy and magnetic state more efficiently.

{\ifshowname\textcolor{red}{Xiaofeng}\else\fi}

\vspace{0.1cm}\noindent \textbf{Going Beyond Atomistic Scale:}
Last but not least, the AI/ML development for quantum mechanics presented in Section
~\ref{sec:qt} may allow efficient generation of more accurate datasets, which will in turn advance the development of the ML models towards the exact density functional. It is anticipated that quantum tensor learning and/or accurate ML density functionals will bring transformative impact to many other scientific fields such as organic and inorganic chemistry, condensed matter physics, materials design for electrical, mechanical, aerospace, nuclear, civil, and environmental engineering, as well as biological science and pharmaceutical research such as protein folding and drug design. In particular, it will enable the generation of more accurate datasets for the AI/ML development of molecules (Section~\ref{sec:mol}), proteins (Section~\ref{sec:prot}), materials (Section~\ref{sec:mat}), and molecular docking (Section~\ref{sec:dock}), \emph{etc.}

\clearpage
\hypertarget{AI for Molecular Simulation}{\section{AI for Small
Molecules}} \label{sec:mol}





\begin{figure}[htbp]
    \centering
\includegraphics[width=\textwidth]{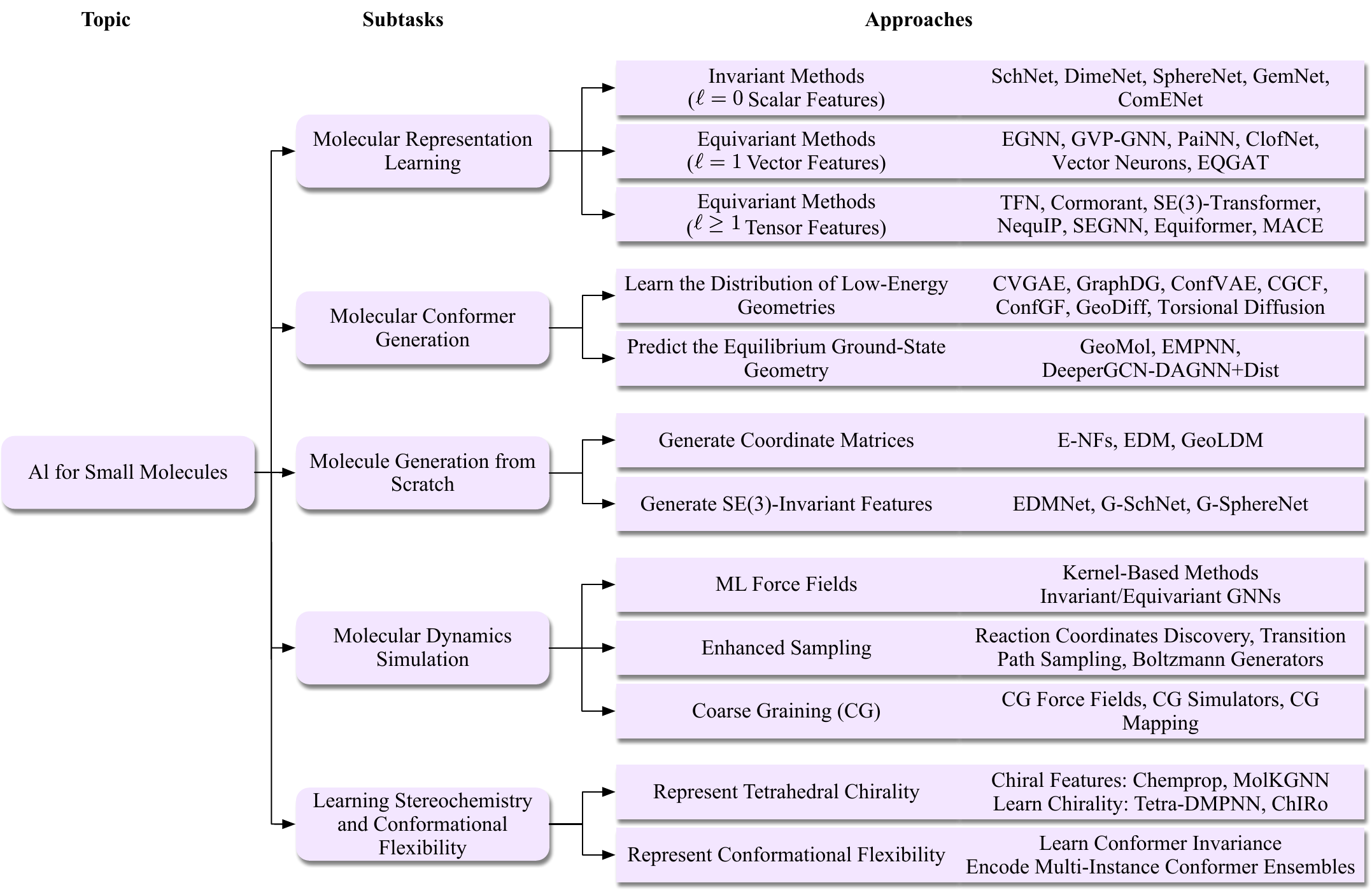}
    \caption{An overview of the tasks and methods in AI for small molecules. In this section, we consider five tasks, including molecular representation learning, molecular conformer generation, molecule generation from scratch, molecular dynamics simulation, and learning stereochemistry and conformational flexibility. In molecular representation learning, the $\ell=0$ case corresponds to invariant methods, including SchNet \cite{schutt2018schnet}, DimeNet~\cite{gasteigerdirectional}, SphereNet~\cite{liu2022spherical}, GemNet~\cite{gasteiger2021gemnet}, and ComENet~\cite{wang2022comenet}. The $\ell=1$ case corresponds to equivariant methods with order-$1$ vector features $\bm{v} \in \mathbb{R}^{d\times 3}$, including EGNN~\cite{satorras2021n}, GVP-GNN~\cite{jing2021learning}, PaiNN~\cite{schutt2021equivariant}, ClofNet~\cite{du2022se}, Vector Neurons~\cite{deng2021vector}, and EQGAT~\cite{le2022equivariant}. The $\ell\geq 1$ case corresponds to equivariant methods with order-$\ell$ features $\bm{h}^{\ell}\in\mathbb{R}^{d\times(2\ell+1)}$, including TFN~\cite{thomas2018tensor}, 3d-steerable CNNs~\cite{3d_steerableCNNs}, Cormorant~\cite{anderson2019cormorant}, $SE(3)$-Transformer~\cite{fuchs2020se}, NequIP~\cite{batzner20223}, SEGNN~\cite{brandstetter2022geometric}, Equiformer~\cite{liao2023equiformer}, and MACE~\cite{batatia2022mace}. In molecular conformer generation, one category of methods aims to learn the distribution of low-energy geometries, including CVGAE~\cite{mansimov2019molecular}, GraphDG~\cite{simm2019generative}, ConfVAE~\cite{xu2021end}, CGCF~\cite{xu2021learning}, ConfGf~\cite{Shi2021LearningGF}, GeoDiff~\cite{xu2022geodiff}, and Torsional Diffusion~\cite{jing2022torsional}. Another category of methods aims to predict only the equilibrium ground-state geometry, including GeoMol~\cite{ganea2021geomol}, EMPNN~\cite{xu20233d} and DeeperGCN-DAGNN+Dist~\cite{xu2021molecule3d}. In molecule generation from scratch, one category of method aims to directly generate coordinate matrices of 3D molecules, including E-NFs~\cite{satorras2021en}, EDM~\cite{hoogeboom2022equivariant}, and GeoLDM~\citep{xu2023geometric}. Another category of methods implicitly generates 3D atom positions from $SE(3)$-invariant features, including EDMNet~\cite{hoffmann2019generating}, G-SchNet~\cite{gebauer2019symmetry}, and G-SphereNet~\cite{luo2022an}. In molecular dynamics simulation, research directions including ML force fields~\cite{unke2021machine}, enhanced sampling~\cite{sidky2020machine}, and coarse-graining approaches~\cite{noid2023perspective} are briefly introduced. Learning stereochemistry has focused on encoding tetrahedral chirality by employing heuristic features (Chemprop \cite{yang2019analyzing}, MolKGNN \cite{liu2022interpretable}) or designing chiral message passing operations (Tetra-DMPNN \cite{pattanaik2020message}, ChIRo\cite{adams2021learning}). Representing conformational flexibility has involved learning conformer invariance \cite{adams2021learning} or explicitly encoding multi-instance conformer ensembles \cite{axelrod2020molecular,chuang2020attention}. }
\label{fig:mol}
\end{figure}

In chemistry, a small molecule refers to a relatively low molecular weight organic compound. It typically comprises a small number of atoms, usually less than 100, and has a defined chemical structure. Small molecules are contrasted with macromolecules, such as proteins, nucleic acids, and polymers, which are much larger in size and often have complex structures. The use of AI approaches in small molecule learning allows for the development of more accurate and efficient methods for molecular predictive and generative tasks. In this section, we consider several key tasks in AI for molecular learning, 
including molecular representation learning, molecular conformer generation, molecule generation from scratch, molecular dynamics simulation, and representation learning of stereoisomerism and conformational flexibility, as summarized in Figure~\ref{fig:mol}. \revisionOne{More tasks related to proteins, materials, and molecular interactions are introduced in Sections~\ref{sec:prot}, \ref{sec:mat}, and \ref{sec:dock}.}

\subsection{Overview}

\noindent{\emph{Authors: Meng Liu, Shuiwang Ji}}\newline


Since machine learning approaches modeling 2D molecular graphs have been widely explored and achieved promising results~\cite{gilmer2017neural,wu2018moleculenet,yang2019analyzing,hu2020open,wang2022advanced,molecule2022,edwards2022translation,velivckovic2023everything}, here, we focus on modeling 3D geometric molecules, a more challenging and practically meaningful perspective. The 3D geometry of a molecule plays a crucial role in many molecular predictive and generative tasks. To be specific, the 3D geometry of a molecule is a critical factor in determining many important properties, such as quantum properties~\cite{ramakrishnan2014quantum,schutt2018schnet, liu2022spherical,liu2021fast,liu2023symmetryinformed, unke2021spookynet}, \revisionOne{molecular spectra~\cite{zou2023deep}} and its binding affinity to a target protein, which is largely dependent on the complementary 3D shape of the molecule and the target protein~\cite{anderson2003process}. Therefore, modeling 3D molecular geometries using predictive and generative AI approaches has immense potential. Particularly, it can significantly enhance the accuracy of predicting molecular properties and generate new molecules with desired properties.

In molecular representation learning, given 3D molecular geometries, our objective is to learn informative representations for various downstream tasks, such as molecule-level predictions and atom-level predictions. These representations are expected to capture accurate structural and chemical features of molecules. This task is fundamental since it is the basis for many advanced topics, such as drug discovery, materials design, and chemical reactions. The next two tasks we considered in this section are generative tasks. Specifically, molecular conformer generation is a conditional generation task where we aim to generate low-energy geometries or equilibrium ground-state geometries given a 2D molecular graph. This can provide an alternative to computationally expensive methods like density functional theory for obtaining 3D molecular geometries, thereby having significant potential in accelerating molecular simulation applications. Further, in certain applications, the desired 2D molecular graph is unknown, and we are interested in generating desired 3D molecules from scratch. Thus, in the task of molecule generation from scratch, our goal is to model the distribution over the 3D molecular geometry space with generative approaches. This can be used as the first step to generate novel molecules with desired properties for drug discovery, material science, and other applications. In addition to generic molecular representation learning, it is particularly important to learn to simulate molecular dynamics, which allows us to capture the time-dependent behavior of molecular systems, providing invaluable insights into their physical properties and structural transformations. AI approaches are facilitating the field of molecular dynamics simulations mainly through improving the accuracy of force fields, enhancing sampling methods, and enabling effective coarse-graining. Lastly, we consider the inherent complexity of molecular structures for more effective representation learning. Specifically, we discuss the importance of molecular stereochemistry and conformational flexibility during molecular representation learning.

Since we are modeling molecules in 3D space, it is desired to take the underlying equivariance and invariance properties into consideration. Preserving the desired symmetry in 3D molecular learning tasks is crucial for obtaining accurate predictions and ensuring the physical constraints of the system. In addition, how to capture the 3D information accurately, such as distinguishing enantiomers of the same molecule, is another important consideration to achieve effective modeling.

\subsection{Molecular Representation Learning} \label{subsec:mol_representation_learning}

\noindent{\emph{Authors: Limei Wang, Youzhi Luo, Zhao Xu, Montgomery Bohde, Chaitanya K. Joshi, Haiyang Yu, Meng Liu, Simon V. Mathis, Alexandra Saxton, Yi Liu, Pietro Liò, Shuiwang Ji} \vspace{0.3cm}\newline
\emph{Recommended Prerequisites: Sections~\ref{subsec:cont_feat},~\ref{subsec:cont_equi}}\newline


In this section, we study the problem of molecular representation learning, which aims to learn informative representations of given input molecules. The learned representations can be used for various downstream tasks, such as molecule-level prediction and atom-level prediction. In addition, the representation learning models introduced in this section can be seen as backbones that enable more advanced applications, such as drug discovery and material design.

\subsubsection{Problem Setup}

\vspace{0.1cm}\noindent\textbf{Molecular Graphs and Point Clouds:} 
Molecules may be represented as 2D molecular graphs, which contain the graph topology (bonds between atoms) as well as node and edge features or as 3D molecular graphs, which additionally consider the 3D coordinates for each node. While the 2D representation suffices to describe the chemical identity of a molecule, the 3D configuration of the molecule (called \emph{conformer}) is relevant for determining many experimentally relevant properties of the molecule, such as its energy or electric dipole moment. Thus, we focus on methods for working with 3D molecular graphs in the remainder of this section.
Formally, we represent a 3D molecule as a point cloud with $n$ atoms as $\mathcal{M}=(\bm{z},C)$, where $\bm{z}=[z_1,...,z_n]\in\mathbb{Z}^n$ is the atom type vector and $C=[\bm{c}_1,...,\bm{c}_n]\in\mathbb{R}^{3\times n}$ is the atom coordinate matrix. To obtain a molecular graph from this point cloud, edges may then be added for example from the bonds (2D graph topology), from radial distance cut-offs or from the $k$ nearest neighbors. Because edge construction differs between methods (further discussed below), we refer to a molecule as its point cloud $\mathcal{M}=(\bm{z},C)$.

\vspace{0.1cm}\noindent\textbf{Task Formulations:} We aim to learn latent representations of 3D molecules which can be used for downstream prediction tasks and applications. Two types of downstream prediction tasks are of interest: \emph{molecule-level} predictions and \emph{atom-level} predictions. For molecule-level property prediction tasks, we aim to learn a function $f(\mathcal{M})$ to predict a property $y$ of any given molecule $\mathcal{M}$. Here, $y$ can be a real number (regression problems such as the energy of a conformer), an integer (classification problems such as toxicity), or a tensor (such as the electric dipole vector, or the tensor of inertia). If the target property $y$ is a scalar/tensorial quantity, it needs to be invariant/equivariant to changes in reference frame. For atom-level property prediction tasks, we aim to learn a function $f$ to predict the property $y_i$ of the $i$-th atom, such as per-atom forces for molecular simulation. Again, $y$ may be a scalar or tensorial target property.

\subsubsection{Technical Challenges}
Different from typical 2D graphs with topology only, the geometry of 3D structures poses unique challenges to 3D molecular modeling.

(1) The first challenge is that the learned representations correspond to physical geometric quantities and should follow the underlying symmetries for different applications \cite{bogatskiy2022symmetry}. To be specific, for tasks like energy prediction, the learned representations should be $SE(3)$-invariant. This means that if the input molecule is rotated or translated, the learned representations should remain unchanged. For tasks like per-atom force prediction, the representations should be $SO(3)$-equivariant. This is because if the input molecule is rotated, the prediction target (\emph{e.g.}, forces) should rotate accordingly.

(2) Another challenge is the theoretical expressive power of learned representations~\cite{joshi2023expressive}, which instantiates itself as practical limitations of models at distinguishing different 3D geometries of molecules, such as enantiomers and different conformers of the same molecule. 
Learning expressive molecular representations is crucial for applications like drug design and molecular simulations \cite{pozdnyakov2020incompleteness}. For example, the enantiomers of a chiral drug can interact very differently with other chiral molecules and proteins. Different conformers of the same molecule also have different potential energies and per-atom forces. 

(3) Thirdly, efficiency is an important factor to consider when designing a model for molecular representation learning. High efficiency enables fast training and inference, reduces computational resources, and enhances scalability to large-scale, real-world datasets.

\subsubsection{Overview of Existing Methods}

As indicated above, a 2D molecular graph contains the graph topology as well as the original node and edge features, base on which
a 3D molecular graph further considers
3D coordinates for each node.
Any geometric quantities, like distance, angle, and torsion angle, can be computed from the 3D coordinates.
More generally, as introduced in Section~\ref{sec:group}, each node has an order-$\ell$ $SE(3)$-equivariant node feature. 
From the perspective of tensor order, 
existing methods for 3D molecular representation learning can be categorized into invariant 3D graph neural networks (3D GNNs) with only $\ell=0$ scalar-type features~\cite{schutt2017quantum, smith2017ani, chmiela2017machine, zhang2018deep, zhang2018end, unke2019physnet, schutt2018schnet, ying2021transformers, zhou2023unimol, luo2023one, gasteigerdirectional, liu2022spherical, gasteiger2021gemnet, wang2022comenet}, equivariant 3D GNNs with $\ell=1$ vector-type features~\cite{schutt2021equivariant, jing2021learning, satorras2021n, du2022se, du2023new, tholke2022equivariant}, and equivariant 3D GNNs with higher order $\ell\geq1$ tensor features~\cite{thomas2018tensor, 3d_steerableCNNs, fuchs2020se, liao2023equiformer, batzner20223, batatia2022design, batatia2022mace}. Specifically, invariant methods directly take invariant geometric features such as distances and angles as input, and thus, all internal features remain unchanged regardless of transformations like rotation and translation of the input molecule. In contrast, internal features in equivariant methods should transform accordingly when the input molecule is rotated or translated. 

\begin{figure}[t]
    \centering
    \includegraphics[width=0.7\textwidth]{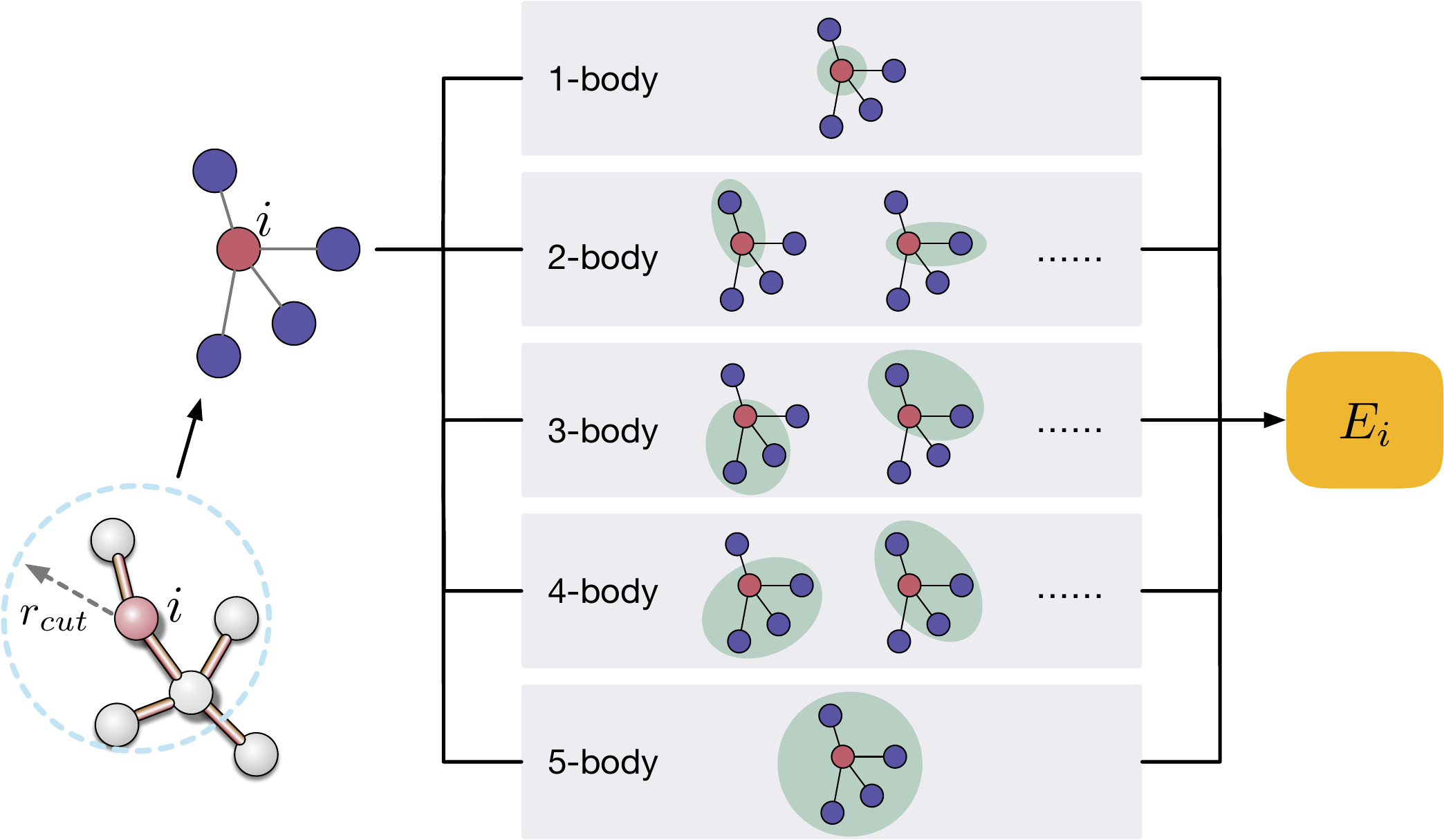}
    \caption{An illustration of body order expansion in molecular energy prediction. First, neighboring nodes and edges of the central atom $i$ are determined by a cutoff $r\textsubscript{cut}$. Then, $v$ body term considers all combinations of the central atom $i$ and $v-1$ of its 1-hop neighbors. Finally, the local energy of atom $i$ is computed as a linear combination of all body-ordered functions. Compared to body order expansion~\cite{brown2004quantum, braams2009permutationally}, the standard message passing~\cite{gilmer2017neural} only considers a body order of 2 as it solely involves the central atom and one neighboring atom in each message.}\label{fig:many_body}
\end{figure}

In addition to tensor order, existing 3D GNN layers can be further categorized from the perspective of body order. Body order originates from the decomposition of potential energy surface (PES) as a linear combination of body-ordered functions. Traditional approaches~\cite{brown2004quantum, braams2009permutationally} show that body order expansion as illustrated in Figure~\ref{fig:many_body} leads to high accuracy and fast convergence in approximating the PES of molecular and material systems. In these approaches, the total molecular energy $E=\sum_i E_i$ is the summation of the local energy of every atom in the molecule, and the local energy of atom $i$ is written in the form of body-ordered expansions as
\begin{equation}
	\begin{aligned}
    E_i = &f_1(z_i) + \sum_{j\in\mathcal{N}(i)} f_2\left(\sigma^i_j;z_i\right) + \sum_{j_1 <j_2, j_1, j_2\in\mathcal{N}(i)} f_3\left(\sigma^i_{j_1}, \sigma^i_{j_2};z_i\right) \\
    &+ \cdots + \sum_{j_1<...<j_v, j_1,...,j_v\in\mathcal{N}(i)} f_{v+1}\left(\sigma^i_{j_1}, ..., \sigma^i_{j_v};z_i\right)+ \cdots ,
    \end{aligned}
    \label{eq:many_body} 
\end{equation}
where $\mathcal{N}(i)$ is the set of all neighbor atoms of atom $i$; $\sigma^i_j=(z_j, \bm{r}_{ij})$ denotes the state of the neighbor atom $j$, including atom type $z_j$ and the position $\bm{r}_{ij}=\bm{c}_i-\bm{c}_j$ of atom $j$ relative to atom $i$; $f_1(z_i)$ is a constant energy term that is only related to atom type of atom $i$, and $f_{v+1}\left(\sigma^i_{j_1}, ..., \sigma^i_{j_v};z_i\right) (v>0)$ captures the many-body interaction among atom $i$ and its 1-hop neighboring atoms $j_1,...,j_v$. In the $(v+1)$-body term of Equation~(\ref{eq:many_body}), all combinations of any $v$ different neighboring atoms $j_1,...,j_v$ of atom $i$ are considered, and $f_v(\cdot)$ is invariant to permutations of $j_1,...,j_v$. Therefore, the standard message passing~\cite{gilmer2017neural} 
\begin{equation}
    \begin{aligned}
    \bm{m}_i &= \sum_{j\in\mathcal{N}(i)} M \left(\bm{h}_i, \bm{h}_j, \bm{h}_{ij}\right), \\
    \bm{h}_i^{\prime} &= U \left(\bm{h}_i, \bm{m}_i\right)
    \end{aligned}
    \label{eq:node_centered_message_passing}
\end{equation}
implements a 2-body term because each message involves the central atom and one neighbor atom. Here $\bm{h}_{ij}$ is the edge feature between node $i$ and node $j$, such as the edge length and edge type, and $U$ and $M$ are the update and message functions. Although standard message passing can further aggregate information from many nodes along edges through iterative layers, such aggregation is distinct from many-body interaction that is restricted within 1-hop of the central node. In this subsection, we discuss existing 3D molecular representation learning methods based on their tensor order as well as body order, as summarized in Figure~\ref{fig:tensor_order_body_order}. 
\begin{figure}[ht]
    \centering
    \includegraphics[width=\linewidth]{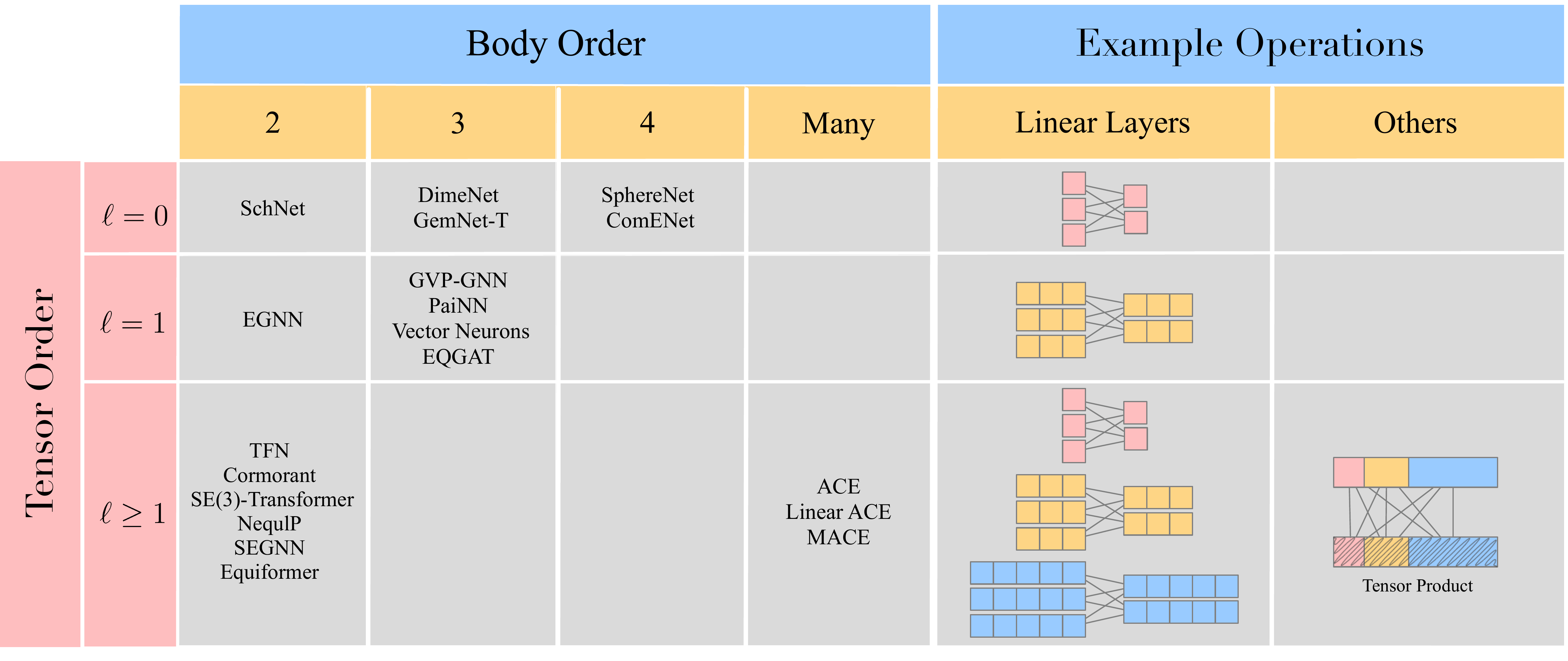}
    \caption{An overview of existing methods for molecule representation learning. We categorize existing methods based on the tensor order of features and the body order of GNN layers, which are two key design choices for building maximally powerful 3D GNNs, as discussed in~\citet{joshi2023expressive} and Section~\ref{subsec:mol_rep_open}. Invariant methods with $\ell=0$ scalar features are described in detail in Section~\ref{sec:l=0}, equivariant methods with $\ell=1$ vector features in Section~\ref{sec:l=1}, equivariant methods with $\ell \geq 1$ tensor features in Section~\ref{sec:l>1}, and higher body order methods in Section~\ref{subsec:higher_body}. In addition, different order features require specific operations to maintain $SE(3)$ equivariance. Here we list several example operations for methods with different tensor orders. Specifically, for the linear layers, each gray line between input and output features contains a learnable weight. The bias term can only be added to $\ell=0$ scalar features, as it would break the equivariance of $\ell \geq 1$ features. Additionally, tensor product, introduced in~\ref{subsec:cont_equi} and illustrated in Figure~\ref{fig:group}, is another crucial operation for equivariant methods with $\ell \geq 1$ tensor features as it can maintain $SE(3)$-equivariance of higher-order features. This figure is adapted from \citet{joshi2023expressive} with permission.}
    \label{fig:tensor_order_body_order}
\end{figure}

\begin{table}[t]
    \centering
    \caption{Summary of existing invariant 3D graph neural networks ($\ell=0$) for molecular representation learning, including SchNet~\cite{schutt2018schnet}, DimeNet~\cite{gasteigerdirectional}, GemNet~\cite{gasteiger2021gemnet}, SphereNet~\cite{liu2022spherical}, and ComENet~\cite{wang2022comenet}. Here $n$ and $k$ denote the number of nodes and the average degree in a molecule. The complexity depends on the calculation of the geometric features and the message passing schema.
    }
    \resizebox{\textwidth}{!}{
    \begin{tabular}{l|p{0.22\textwidth}ccc}
    \toprule
    Methods & \multicolumn{2}{c}{Invariant Geometric Features} & Body Order & Complexity \\
    \midrule
    SchNet & \multirow{5}{*}{\includegraphics[width=0.22\textwidth]{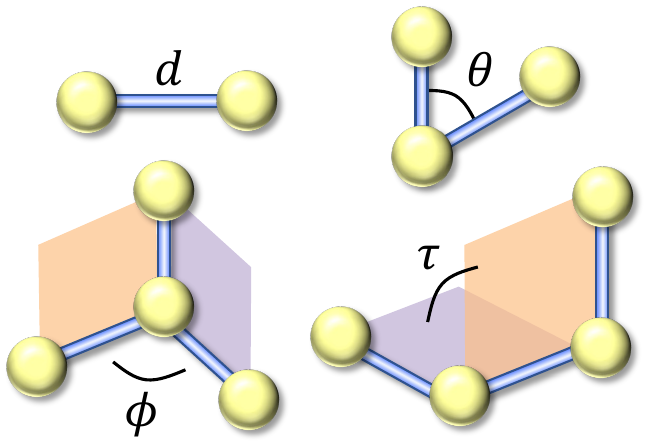}} & Pairwise distances $d$ & 2-body &$O(nk)$ \\
    DimeNet & & $d$ + Angles between edges $\theta$              & 3-body &$O(nk^2)$ \\
    GemNet & & $d, \theta$ + Angles between 4 nodes $\tau$       & 4-body &$O(nk^3)$ \\
    SphereNet & & $d, \theta$ + Angles between 4 nodes $\phi$    & 4-body &$O(nk^2)$ \\
    ComENet & & $d, \theta, \phi, \tau$                          & 4-body &$O(nk)$ \\
    \bottomrule
    \end{tabular}
    }
    \label{tab:mol_rep_invariant}
\end{table}

\subsubsection{Invariant Methods ($\ell=0$ Scalar Features)} \label{sec:l=0}

Invariant methods only maintain invariant node, edge, or graph features, which do not change if the input 3D molecule is rotated or translated. Invariant methods face a trade-off between improving their discriminative ability by considering many-body geometric features and maintaining their efficiency, as summarized in Table~\ref{tab:mol_rep_invariant}. Let $n$ and $k$ denote the number of nodes and the average degree in a molecule. Specifically, SchNet~\cite{schutt2018schnet} considers only pairwise distances as edge features $\bm{h}_{ij}$ in the node-centered message passing schema shown in Equation~(\ref{eq:node_centered_message_passing}), resulting in a complexity of $O(nk)$ and a body order of 2. DimeNet~\cite{gasteigerdirectional} further considers angles between each pair of edges with edge-centered message passing 
\begin{equation}
    \begin{aligned}
    \bm{m}_{ji} &= \sum_{k\in\mathcal{N}(j)\backslash\{i\}} M \left(\bm{h}_{ji}, \bm{h}_{kj}, \bm{h}_{kji}\right), \\
    \bm{h}_{ji}^{\prime} &= U \left(\bm{h}_{ji}, \bm{m}_{ji}\right),
    \end{aligned}
    \label{eq:edge_centered_message_passing}
\end{equation}
and the complexity is $O(nk^2)$. Here $\mathcal{N}(j)\backslash\{i\}$ is the set of neighboring nodes of node $j$ except for node $i$, $\bm{h}_{kji}$ is the feature of nodes $k, j$, and $i$, such as the angle $\theta_{kji}$, and $U$ and $M$ are the update and message functions.
GemNet~\cite{gasteiger2021gemnet} further considers two-hop dihedral angles, increasing body order to 4 and complexity to $O(nk^3)$. SphereNet~\cite{liu2022spherical} computes local 4-body angles between two planes. To reduce the complexity, SphereNet does not incorporate all possible angles, instead reducing the number of angles by selecting reference nodes to construct reference planes while retaining $O(nk^2)$ complexity. 
ComENet~\cite{wang2022comenet} defines complete geometric features that can distinguish all different 3D molecules that exist. Specifically, the distance and angles $d, \theta, \phi$ are 2-body, 3-body, and 4-body geometric features and can be used to identify local structures. Here a local structure means a central node and its 1-hop neighborhood. This is because $d_{ij}, \theta_{ij}, \phi_{ij}$ can determine the relative position of node $j$ in the local spherical coordinate system centered in $i$. In addition, The rotation angle $\tau$ further captures the remaining degree of freedom between local structures. Therefore, ComENet has the ability to generate a unique representation for each 3D molecule, able to distinguish all different 3D molecules in nature. Moreover, it follows the node-centered message passing schema in Equation~(\ref{eq:node_centered_message_passing}), and the complexity is only $O(nk)$ by selecting reference nodes within 1-hop neighborhood.

In addition to the methods that convert equivariant 3D information to invariant features like distances and angles~\cite{schutt2018schnet, gasteigerdirectional, liu2022spherical, gasteiger2021gemnet, wang2022comenet}, \citet{du2022se, du2023new} propose scalarization to obtain invariant features. Specifically, scalarization converts equivariant features into invariant features based on equivariant local frames. For example, given an equivariant frame $(\bm{e}_1, \bm{e}_2, \bm{e}_3)$, we can convert a 3D vector $\bm{r}_{ij} = \bm{c}_i - \bm{c}_j$ to $(\bm{r}_{ij} \cdot \bm{e}_1, \bm{r}_{ij} \cdot \bm{e}_2, \bm{r}_{ij} \cdot \bm{e}_3)$. Here $\bm{e}_1, \bm{e}_2, \bm{e}_3$ form an orthonormal basis. 
In addition to scalarization, ClofNet and LEFTNet~\cite{du2022se, du2023new} also use tensorization to convert invariant features to equivariant features. Therefore, these methods can maintain both invariant and equivariant internal features and require both invariant operations and equivariant operations (see Section~\ref{sec:l=1} and \ref{sec:l>1}) to update the internal features. 


\subsubsection{Equivariant Methods ($\ell=1$ Vector Features)} \label{sec:l=1}

The first category of equivariant 3D GNNs \cite{satorras2021n, du2022se,schutt2021equivariant, deng2021vector, jing2021learning, tholke2022equivariant} uses order 1 vectors as intermediate features and propagates messages via a restricted set of operations that guarantee $E(3)$ or $SE(3)$ equivariance, as summarized in Table~\ref{tab:mol_rep_l=1}. Let us denote a scalar feature by $\bm{s}\in \mathbb{R} ^ {d}$ and a vector by $\bm{v} \in \mathbb{R}^{d\times 3}$. As summarized in~\citet{schutt2021equivariant} and~\citet{deng2021vector}, operations on a vector $\bm{v}$ that can ensure equivariance include scaling of vectors $\bm{s} \odot \bm{v}$, summation of vectors $\bm{v}_1 + \bm{v}_2$, linear transformation of vectors $W\bm{v}$, scalar product $\lVert \bm{v} \rVert ^2, v_1 \cdot v_2$, and vector product $v_1 \times v_2$. Here $\odot$ denotes element-wise multiplication. Note that $v_1 \cdot v_2=\lVert v_1\rVert \lVert v_2\rVert \cos \theta$ and $v_1 \times v_2=\lVert v_1\rVert \lVert v_2\rVert \sin \theta \vec n$, therefore, using scalar product and vector product can implicitly incorporate angular and directional information.

Existing methods use these operations to update both scalar and vector features by propagating scalar as well as vector messages. For example, EGNN~\cite{satorras2021n} uses scaling of vectors and vector summation to ensure equivariance. To be specific, following the notation of Equation~(\ref{eq:node_centered_message_passing}), an EGNN layer updates node representation $\bm{h}_i$ and node coordinate $\bm{c}_i$ as
\begin{equation}
    \begin{aligned}
    \bm{m}_{ij} &= \phi_e \left(\bm{h}_i, \bm{h}_j, ||\bm{c}_i-\bm{c}_j||^2, \bm{h}_{ij}\right), \\
    \bm{c}'_i &= \bm{c}_i + C\sum_{j\neq i}(\bm{c}_i-\bm{c}_j)\phi_c(\bm{m}_{ij}), \\
    \bm{h}'_i &=\phi_h\left(\bm{h}_i, \sum_{j\neq i} \bm{m}_{ij}\right),
    \end{aligned}
    \label{eq:egnn}
\end{equation}
where $\phi_e$, $\phi_c$, and $\phi_h$ denote learnable functions and $C$ is a normalization factor. Different from EGNN which only considers a single vector for each edge, ClofNet~\cite{du2022se} employs complete frames that consist of three vectors for each edge. PaiNN~\cite{schutt2021equivariant} further includes linear transformation and scalar product in the network. GVP-GNN~\cite{jing2021learning} uses similar operations as PaiNN and was originally designed to learn representations for protein structures, but can also be adapted to molecules. Vector Neurons~\cite{deng2021vector} is originally designed for point cloud data and can be applied to molecules. It also employs linear transformation to achieve the linear operator for order 1 vectors. In addition to linear operators, Vector Neurons incorporate carefully designed non-linear, pooling, and normalization layers that are tailored for order 1 vectors while ensuring the desired equivariance. Notably, it uses a learnable direction, which is equivariant, to split the domain into two half-spaces, and then non-linear layers such as ReLU can be defined to map such two spaces differently. In addition to the aforementioned operations, EQGAT~\cite{le2022equivariant} uses cross product to update equivariant features during message passing. This enables interactions between type-1 vector features and allows for more comprehensive and expressive feature representations. Moreover, it uses attention mechanism to capture content and spatial information between nodes.

\begin{table}[t]
    \centering
    \caption{Comparisons of equivariant methods using $\ell=1$ vector features, including EGNN~\cite{satorras2021n}, ClofNet~\cite{du2022se}, PaiNN~\cite{schutt2021equivariant}, GVP-GNN~\cite{jing2021learning}, Vector Neurons~\cite{deng2021vector}, and EQGAT~\cite{le2022equivariant}. Here $\bm{s}\in \mathbb{R} ^ {d}$ denotes a scalar feature, and $\bm{v} \in \mathbb{R}^{d\times 3}$ denotes a vector feature. Existing methods use different operations to ensure equivariance.}
        \resizebox{\textwidth}{!}{
    \begin{tabular}{l|ccccc}
        \toprule[1pt]
        Methods & Scaling & Summation & Linear Transformation  & Scalar Product  & Vector Product  \\
            & $\bm{s} \odot \bm{v}$ & $\bm{v}_1 + \bm{v}_2$ & $W\bm{v}$ & $\lVert \bm{v} \rVert ^2, \bm{v}_1 \cdot \bm{v}_2$ & $\bm{v}_1 \times \bm{v}_2$ \\
        \midrule
        EGNN & \cmark & \cmark &  & \cmark &  \\
        ClofNet & \cmark & \cmark &  & \cmark &  \\
        PaiNN & \cmark & \cmark & \cmark & \cmark & \\
        GVP-GNN & \cmark & \cmark & \cmark & \cmark & \\
        Vector Neurons & \cmark & \cmark & \cmark & \cmark & \\
        EQGAT & \cmark & \cmark & \cmark & \cmark & \cmark \\
        \bottomrule
    \end{tabular}
     }
    \label{tab:mol_rep_l=1}
\end{table}

\subsubsection{Equivariant Methods ($\ell\geq1$ Tensor Features)} \label{sec:l>1}
\label{equivariant_greater_1_methods}


Another category of equivariant methods considers higher-order ($\ell\geq1$) features that have been discussed in Section~\ref{sec:group}. 
Most existing methods under this category use tensor products (TP) of higher-order spherical tensors to build equivariant representations and follow the general architecture in Figure~\ref{fig:equivariant_model_architechture} to update features, and differing in body order and technical details. 
For example, TFN~\cite{thomas2018tensor} and NequIP~\cite{batzner20223} follow the node-centered message passing scheme~\cite{gilmer2017neural} to update node features based on messages from neighboring nodes. Since each message contains the information of the central atom and one neighbor, these methods naturally have a body order of 2. $SE(3)$-Transformer~\cite{fuchs2020se} and Equiformer~\cite{liao2023equiformer} further enhance their model architectures with the attention mechanism. 
In addition, Cormorant~\cite{anderson2019cormorant} and SEGNN~\cite{brandstetter2022geometric} introduce different designs of non-linearity on higher-order features.

\begin{figure}[t]
    \centering
    \includegraphics[width=0.6\textwidth]{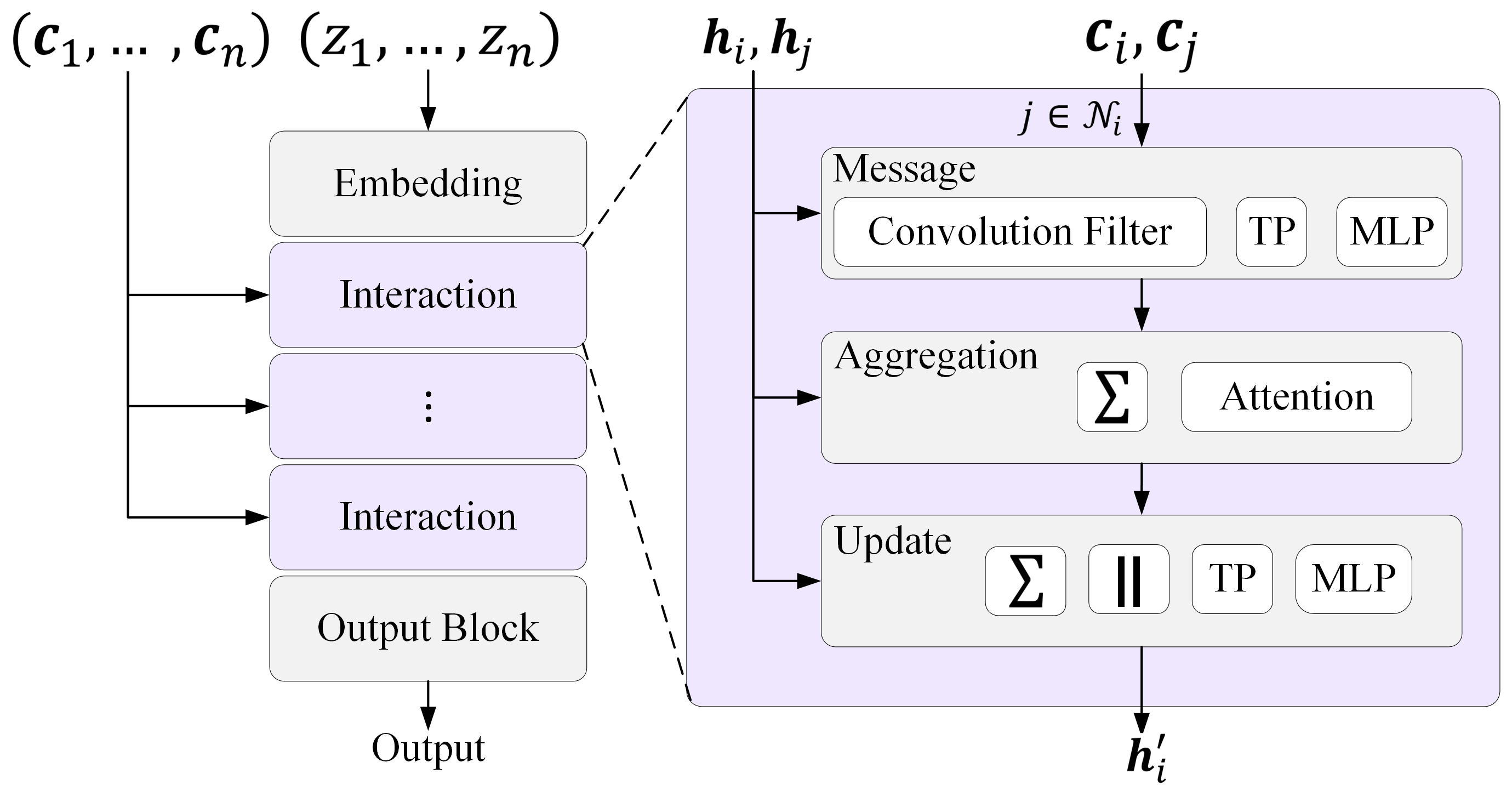}
    \caption{A general architecture of higher order equivariant models. Each model consists of several interaction blocks which perform pairwise message passing between atoms. Here, $\Sigma$ denotes summation, $\parallel$ denotes feature concatenation, TP refers to the Tensor Product of feature vectors, and MLP refers to multilayer perceptrons. The specific operations used in each Message, Aggregation, and Update blocks differ between models, but existing methods such as TFN~\cite{thomas2018tensor}, NequIP~\cite{batzner20223}, $SE(3)$-Transformer~\cite{fuchs2020se}, Equiformer~\cite{liao2023equiformer}, Cormorant~\cite{anderson2019cormorant}, and SEGNN~\cite{brandstetter2022geometric}  all fall under this framework.}\label{fig:equivariant_model_architechture}
\end{figure}

\noindent

Generally,
higher order equivariant models build multiple feature vectors or "channels" for each rotation order $\ell \leq \ell_{max}$. Each channel of order $\ell$ has a length of $2\ell + 1$. As such, the features of node $i$ can be indexed by $\bm{h}_{icm}^{\ell}$, where $\ell$ is the rotation order, $c$ is the channel index, and $m$ is the representation index $\left(-\ell \le m \le \ell \right)$.
A general architecture of higher order equivariant models is given in Figure~\ref{fig:equivariant_model_architechture}, and we describe each component below.

\vspace{0.1cm}\noindent\textbf{Nonlinear Functions:} In order to preserve equivariance, the nonlinear functions used in these models are restricted to those which act as scalar transforms in the representation index $m$. The nonlinear functions used by various models are shown in Table~\ref{tab:equivariant_nonlinear}.
Notably, Cormorant uses the tensor product as the only nonlinear operation, and $SE(3)$-Transformer uses attention instead of nonlinear activations found in other models.
\begin{table}[h]
    \centering
    \caption{The nonlinear functions used in higher order equivariant models. $\eta:\mathbb{R}\to \mathbb{R}$ is a nonlinear function such as SiLU or tanh, $\lVert \bm{h}_{c}^{\ell} \rVert = \sqrt{\sum_{m} {|\bm{h}_{cm}^{\ell}|}^2 }$, and $b_c^{\ell}$ is a learnable bias.}
    \begin{tabular}{c|c}
        \toprule[1pt]
        Methods & Nonlinear Functions, $g\left(\bm{h}_c^{\ell}\right)$ \\
        \midrule
        TFN & $\eta \left( \parallel \bm{h}_c^{\ell} \parallel + \text{ }b_c^{\ell}\right)\bm{h}_c^{\ell}$\\
        NequIP &   $\eta \left( \parallel \bm{h}_c^{\ell} \parallel \right)\bm{h}_c^{\ell}$\\
        SEGNN, Equiformer & $\eta \left(\bm{h}_c^{0} \right)\bm{h}_c^{\ell}$\\
        Cormorant, $SE(3)$-Transformer & $\bm{h}_c^{\ell}$\\
        \bottomrule
    \end{tabular}\par
    \label{tab:equivariant_nonlinear}
\end{table}

\vspace{0.1cm}\noindent \textbf{Linear Layers:}
The linear layers used in these models take the form as
\begin{equation}
W(\bm{h}^{\ell})=\sum_{c\prime}W_{cc\prime}^{\ell}\bm{h}_{ic\prime m}^{\ell}.
\end{equation}
The weights are constant across the $m$ dimension, which is required to maintain equivariance. Optionally, biases can be added for $l = 0$ features.
\smallskip

\vspace{0.1cm}\noindent\textbf{Convolution Filters:}
These models generally build convolution filters as the product of a learnable radial function and spherical harmonics. The specific filters used by several models are shown in Table~\ref{tab:equivariant_filter}.

\begin{table}[h]
    \centering
    \caption{The convolution filters used in higher order equivariant models. Here, $d_{ij}$ is the distance between nodes.}
    \begin{tabular}{c|c}
        \toprule[1pt]
        Methods & Convolution Filter, $F\left(\cdot\right)^{\ell}_{cm}$ \\
        \midrule
        TFN, NequIP, Equiformer, $SE(3)$-Transformer & $R_c^{\ell}(d_{ij}) Y_m^{\ell}\left( \frac{\bm{c}_{i} - \bm{c}_{j}}{d_{ij}}\right)$\\
        SEGNN & $Y_m^{\ell}\left( \frac{\bm{c}_{i} - \bm{c}_{j}}{d_{ij}}\right)$\\
        Cormorant & $R_c^{\ell}(d_{ij}, \bm{h}_{icm}^{\ell}, \bm{h}_{jcm}^{\ell}) Y_m^{\ell}\left( \frac{\bm{c}_{i} - \bm{c}_{j}}{d_{ij}}\right)$\\
        \bottomrule
    \end{tabular}\par
    \label{tab:equivariant_filter}
\end{table}

\vspace{0.1cm}\noindent\textbf{Message:}
Pairwise messages are then built using tensor products. All methods begin by taking the tensor product of the convolution filer and node features, however, some methods further augment these messages. The specific equations to compute messages in each model are shown in Table~\ref{tab:equivariant_message}.
In general, the tensor product of type $\ell_1$ and $\ell_2$ feature vectors produces outputs at all rotation orders $| \ell_1 - \ell_2 | \leq \ell_3 \leq \ell_1 + \ell_2$. Section ~\ref{subsec:cont_equi} describes the tensor product operations in more detail.

\begin{table}[h]
    \centering
    \caption{The equations for message computing in higher order equivariant models. $\phi\left(\cdot\right) = g\left(W\left(\cdot\right)\right)$. $\parallel$ denotes concatenation of features. The $\ell$, $c$, and $m$ message indices are omitted for brevity.}
    \resizebox{0.9\textwidth}{!}{
    \begin{tabular}{c|c}
        \toprule[1pt]
        Methods & Message $\bm{m}_{ij}$ \\
        \midrule
        TFN, NequIP, Cormorant, $SE(3)$-Transformer & $\mbox{TP}\left(F\left(\cdot\right), \bm{h}_j\right)$\\
        SEGNN & $\phi_2\left(\mbox{TP}\left(F\left(\cdot\right), \phi_1\left(\mbox{TP}\left(F\left(\cdot\right), \bm{h}_i \parallel \bm{h}_j \parallel d_{ij}\right)\right)\right)\right)$\\
        Equiformer & $W\left(\mbox{TP}\left(F\left(\cdot)\right), \phi\left(\mbox{TP}\left(F\left( \cdot \right), W\left(\bm{h}_i\right) + W\left(\bm{h}_j\right)\right)\right)\right)\right)$\\
        \bottomrule
    \end{tabular}}\par
    \label{tab:equivariant_message}
\end{table}

\vspace{0.1cm}\noindent\textbf{Aggregation:}
For each atom, messages are then aggregated over neighboring atoms. For all models, sum aggregation is used, however, $SE(3)$-Transformer and Equiformer first weigh the incoming messages using attention. The aggregation functions used by each model are shown in Table~\ref{tab:equivariant_aggre}. Note $SE(3)$-Transformer uses dot-product attention. Equiformer uses more powerful MLP attention, however, in order to preserve equivariance, only $\ell = 0$ features are used to compute attention scores.

\begin{table}[h]
    \centering
    \caption{The aggregation functions used in higher order equivariant models. $\alpha_{ij}$ are the attention scores.}
    \begin{tabular}{c|c}
        \toprule[1pt]
        Methods & Aggregated Message, $\bm{m}_i$ \\
        \midrule
        TFN, NequIP, Cormorant, SEGNN& $\sum_{j\in \mathcal{N}_i} \bm{m}_{ij}$\\
        $SE(3)$-Transformer, Equiformer & $\sum_{j\in \mathcal{N}_i} \alpha_{ij}\bm{m}_{ij}$\\
        \bottomrule
    \end{tabular}\par
    \label{tab:equivariant_aggre}
\end{table}

\vspace{0.1cm}\noindent\textbf{Update:}
Finally, the aggregated message is used to update the features for each node. The specific update functions used in each model are shown in Table~\ref{tab:equivariant_update}. In TFN and $SE(3)$-Transformer, no residual connection is used between layers. However, later works have shown this connection is crucial in order to retain chemical information such as atom type.

\begin{table}[h]
    \centering
    \caption{The update functions used in higher order equivariant models.}
    \begin{tabular}{c|c}
        \toprule[1pt]
        Methods & Updated Features, $\bm{h}_i^{\prime}$ \\
        \midrule
        TFN, $SE(3)$-Transformer& $\phi\left(\bm{m}_i\right)$\\
        NequIP& $\bm{h}_i + \phi\left(\bm{m}_i\right)W\left(\bm{m}_i\right)$\\
        Cormorant & $W\left(\bm{m}_i \parallel \bm{h}_i \parallel \mbox{TP}\left(\bm{h}_i, \bm{h}_i\right)\right)$\\
        SEGNN & $\bm{h}_i + W\left(\mbox{TP}\left(\phi \left(\mbox{TP}\left(\bm{h}_i \parallel \bm{m}_i, Y\left(\bm{c}_{i}\right)\right)\right), Y\left(\bm{c}_{i}\right)\right)\right)$\\
        \bottomrule
        \end{tabular}\par
        \label{tab:equivariant_update}
\end{table}

\subsubsection{Higher Body Order Methods} \label{subsec:higher_body}



Previously introduced equivariant graph neural networks only capture 2-body interactions in one layer as each message passed to the central atom only involves one neighbor atom. While body-ordered expansions in Equation~(\ref{eq:many_body}) capture many-body interactions, the computational cost of $(v+1)$-body term is at the order of the total number of $v$-atom combinations. To efficiently calculate many-body interactions, atomic cluster expansion (ACE)~\cite{drautz2019atomic,dusson2022atomic} is proposed to make the computational cost of many-body terms linear to the number of neighbor atoms.

In Equation~(\ref{eq:many_body}), the $(v+1)$-body term is a summation over neighbor atoms $j_1,...,j_v$ satisfying $j_1<...<j_v$, which sets an order constraint on neighbor atoms. The first step of ACE is to remove this order constraint by using the permutation-invariance property of $f_v(\cdot)$, thus simplifying Equation~(\ref{eq:many_body}) as
\begin{equation}
	\begin{aligned}
	E_i = &f_1(z_i) + \frac{1}{1!}\sum_{j\in\mathcal{N}(i)} f_2\left(\sigma^i_j;z_i\right) + \frac{1}{2!}\sum_{j_1 \ne j_2, j_1, j_2\in\mathcal{N}(i)} f_3\left(\sigma^i_{j_1}, \sigma^i_{j_2};z_i\right) \\ 
	&+ \cdots + \frac{1}{v!}\sum_{j_1\ne...\ne j_v, j_1,...,j_v\in\mathcal{N}(i)} f_{v+1}\left(\sigma^i_{j_1}, ..., \sigma^i_{j_v};z_i\right)+ \cdots.
	\end{aligned}
	\label{eq:many_body_permute} 
\end{equation}
Equation~(\ref{eq:many_body_permute}) does not have any constraints on neighbor atom orders, but the summation condition $j_1\ne...\ne j_v$ in the $(v+1)$-body term requires that any two neighbor atoms have to be different. To remove this constraint, ACE simplifies Equation~(\ref{eq:many_body_permute}) using spurious terms. Specifically, $(v+1)$-body spurious terms have the same mathematical form as $f_{v+1}\left(\sigma^i_{j_1}, ..., \sigma^i_{j_v};z_i\right)$ but contain repeated atoms among $j_1,...,j_v$. Let $s_{v+1}^k$ be the sum of spurious terms that take the form of $f_{v+1}\left(\sigma^i_{j_1}, ..., \sigma^i_{j_v};z_i\right)$ but only have $k$ $(0<k<v)$ different atoms among $j_1,...,j_v$, \emph{e.g.}, $s_{v+1}^1=\sum_{j\in\mathcal{N}(i)} f_{v+1}\left(\sigma^i_{j}, ..., \sigma^i_{j};z_i\right)$, we can obtain that
\begin{equation}
	 \sum_{j_1\ne...\ne j_v, j_1,...,j_v\in\mathcal{N}(i)} f_{v+1}\left(\sigma^i_{j_1}, ..., \sigma^i_{j_v};z_i\right)= \sum_{j_1,...,j_v\in\mathcal{N}(i)} f_{v+1}\left(\sigma^i_{j_1}, ..., \sigma^i_{j_v};z_i\right)-\sum_{k=1}^{v-1}s_{v+1}^k,\quad v\ge 2.
	 \label{eqn:spurious}
\end{equation}
Replacing all many-body terms by Equation~(\ref{eqn:spurious}), Equation~(\ref{eq:many_body_permute}) can be simplified as
\begin{equation}
	\begin{aligned}
	E_i=&f_1(z_i)+\frac{1}{1!}\sum_{j\in\mathcal{N}(i)}f_2(\sigma_j^i;z_i)+ \cdots +\frac{1}{v!}\left[\sum_{j_1,...,j_v\in\mathcal{N}(i)} f_{v+1}\left(\sigma^i_{j_1}, ..., \sigma^i_{j_v};z_i\right)-\sum_{k=1}^{v-1}s_{v+1}^k\right]+ \cdots \\
	=&f_1(z_i)+\left[\sum_{j\in\mathcal{N}(i)}\frac{f_2(\sigma_j^i;z_i)}{1!}-\sum_{v'>1}\frac{s_{v'+1}^1}{v'!}\right]+ \cdots +\left[\sum_{j_1,...,j_v\in\mathcal{N}(i)} \frac{f_{v+1}\left(\sigma^i_{j_1}, ..., \sigma^i_{j_v};z_i\right)}{v!}-\sum_{v'>v}\frac{s_{v'+1}^v}{v'!}\right] \\
 &+ \cdots \\
	=&g_1(z_i)+\sum_{j\in\mathcal{N}(i)} g_2\left(\sigma^i_j;z_i\right)+ \cdots + \sum_{j_1,...,j_v\in\mathcal{N}(i)} g_{v+1}\left(\sigma^i_{j_1}, ..., \sigma^i_{j_v};z_i\right)+ \cdots
	\end{aligned}
	\label{eq:many_body_no_res}
\end{equation}
In Equation~(\ref{eq:many_body_no_res}), spurious terms are rearranged so that all spurious terms on $v$ different atoms are subtracted from $(v+1)$-body term, and this subtraction result can be rewritten as the summation of a function (defined as $g_{v+1}(\cdot)$ here) over $j_1,...,j_v\in\mathcal{N}(i)$. Note that different from Equation~(\ref{eq:many_body}) or~(\ref{eq:many_body_permute}), the $v$ neighbor atoms $j_1,...,j_v$ are mutually independent of each other. There is no order restriction and any two neighbor atoms can be the same atom. 

The $(v+1)$-body term in Equation~(\ref{eq:many_body_no_res}) is the sum of $|\mathcal{N}(i)|^v$ interaction functions $g_{v+1}(\cdots)$, which is exponential with respect to the number of neighbor atoms for high body order terms. To reduce the computational complexity, ACE simplifies every body-ordered term in Equation~(\ref{eq:many_body_no_res}) to the product of atomic basis functions by density trick. Specifically, ACE uses a set of $L$ orthogonal basis functions $\phi_1,...,\phi_L$, in which all products of any $v$ basis functions form a new orthogonal basis function group. All interaction functions $g_{v+1}\left(\sigma^i_{j_1}, ..., \sigma^i_{j_v};z_i\right) (v>0)$ is expanded to a linear combination of the products of $v$ basis functions as
\begin{align}
    E_i = &g_1(z_i)+\sum_{j\in\mathcal{N}(i)}\sum_{\ell=1}^L c_{i,\ell}^{(1)} \phi_\ell\left(\sigma_j^i\right) 
     + \sum_{j_1,j_2\in\mathcal{N}(i)} \sum_{\ell_1,\ell_2=1}^L c_{i,\ell_1,\ell_2}^{(2)} \phi_{\ell_1}\left(\sigma_{j_1}^i\right) \phi_{\ell_2}\left(\sigma_{j_2}^i\right) \nonumber \\
     & + \cdots +\sum_{j_1,...,j_v\in\mathcal{N}(i)} \sum_{\ell_1,...,\ell_v=1}^L c_{i,\ell_1,...,\ell_v}^{(v)} \phi_{\ell_1}\left(\sigma_{j_1}^i\right) \cdots \phi_{\ell_v}\left(\sigma_{j_v}^i\right) + \cdots,
     \label{eq:many_body_basis}
\end{align}
where $c_{i,\ell_1,...,\ell_v}^{(v)}$ is the coefficient. Using the fact
\begin{equation}
	\begin{aligned}
	&\sum_{j_1,...,j_v\in\mathcal{N}(i)} \sum_{\ell_1,...,\ell_v=1}^L c_{i,\ell_1,...,\ell_v}^{(v)} \phi_{\ell_1}\left(\sigma_{j_1}^i\right) \cdots \phi_{\ell_v}\left(\sigma_{j_v}^i\right) \\
	= &\sum_{\ell_1,...,\ell_v=1}^L c_{i,\ell_1,...,\ell_v}^{(v)} \sum_{j_1,...,j_v\in\mathcal{N}(i)} \phi_{\ell_1}\left(\sigma_{j_1}^i\right) \cdots \phi_{\ell_v}\left(\sigma_{j_v}^i\right)\\
	= &\sum_{\ell_1,...,\ell_v=1}^L c_{i,\ell_1,...,\ell_v}^{(v)}\left[\sum_{j\in\mathcal{N}(i)}\phi_{\ell_1}\left(\sigma_{j}^i\right)\right]\cdots\left[\sum_{j\in\mathcal{N}(i)}\phi_{\ell_v}\left(\sigma_{j}^i\right)\right],
	\end{aligned}
\end{equation}
and defining the atomic basis function as $A_{i,\ell}=\sum_{j\in\mathcal{N}(i)}\phi_\ell(\sigma_j^i)$, Equation~(\ref{eq:many_body_basis}) can be simplified to
\begin{equation}
E_i = g_1(z_i) + \sum_{\ell=1}^L c_{i,\ell}^{(1)} A_{i,\ell}+\sum_{\ell_1,\ell_2=1}^{L} c_{i,\ell_1,\ell_2}^{(2)} A_{i,\ell_1}A_{i,\ell_2}  + \cdots + \sum_{\ell_1,...,\ell_v=1}^{L} c_{i,\ell_1,...,\ell_v}^{(v)} A_{i,\ell_1}\cdots A_{i,\ell_v}+\cdots .
\label{eq:many_body_density_trick}
\end{equation}
In this way, a linear growth with the number of neighbors in computational complexity can be maintained. Let $(v+1)$-body product basis vector $\bm{A}_i^{(v)}$ and coefficient vector $\bm{c}_i^{(v)}$ collect the coefficients $c_{i,\ell_1,...,\ell_v}^{(v)}$ and atomic basis products $A_{i,\ell_1}\cdots A_{i,\ell_v}$ over all possible $\ell_1,...,\ell_v$, respectively, we can write $E_i$ as $E_i=g_1(z_i)+\sum_{v>0}\bm{c}_i^{(v)T}\bm{A}_{i}^{(v)}$. The used basis functions $\phi_1,...,\phi_L$ are product of Bessel functions and spherical harmonics functions, which makes $\bm{A}_{i}^{(v)}$ not $SE(3)$-invariant. Hence, $\bm{A}_{i}^{(v)}$ is always symmetrized to the basis vector $\bm{B}_{i}^{(v)}$ through multiplying with Clebsch-Gordan coefficients, and the final equation for $E_i$ in ACE becomes
\begin{equation}
	E_i = g_1(z_i)+\sum_{v>0}\bm{c}_i^{(v)T}\bm{B}_{i}^{(v)}.
	\label{eq:many_body_ace}
\end{equation}
Based on Equation~(\ref{eq:many_body_ace}), many machine learning methods~\cite{batatia2022mace, batatia2022design, musaelian2023learning, kondor2018n, li2022group, bigi2023wigner,kovacs2023evaluation,musaelian2023scaling,batatia2023general, xu2024equivariant} are developed to capture high body order interactions.

Linear ACE and MACE are developed based on 
the theory of ``density trick''. Linear ACE sequentially builds particle basis, atomic basis, product basis, symmetrized basis, and finally uses the linear combination of symmetrized basis to construct high body-order features efficiently.
The Linear ACE model~\cite{kovacs2021linear} consists of only one layer while MACE ~\cite{batatia2022mace, batatia2022design} leverages tensor product and further extends to multiple ACE layers to enlarge the receptive field so that semi-local information is also incorporated through message passing. In contrast to using ACE to obtain aggregated messages for nodes, Allegro~\cite{musaelian2023learning} focuses operations on edges. Specifically, Allegro performs tensor products between edges around the central node and increases the order of body interaction through a stack of layers. The many-body embeddings produced by Allegro are analogous to ACE's symmetrized basis, although not strictly equivalent. 
PACE~\cite{xu2024equivariant} introduces the edge-boost operation which applies TP on the spherical harmonics and then take the aggregated boosted feature to perform the many-body interaction, aiming to approximate higher-degree $S_n \times SE(3) $ equivariant polynomial functions~\cite{dymuniversality}.  
In addition to the above methods, Wigner kernels~\cite{bigi2023wigner} develops body-ordered kernels calculated in a radial-element space with a cheaper cost that is linear to the maximum body order. N-body networks~\cite{kondor2018n, li2022group} is a hierarchical neural network that aims to learn atomic energies based on the decomposition of the many-body system. Among existing many-body methods, Linear ACE, MACE, PACE, and Allegro follow the general architecture summarized in Figure~\ref{fig:equivariant_model_architechture}.

\vspace{0.1cm}\noindent\textbf{Convolution Filters:} Convolution filters used in higher body order methods are summarized in Table~\ref{tab:higherbody_filter}. Similar to methods such as NequIP in Section~\ref{equivariant_greater_1_methods}, MACE builds convolution filters as the product of a learnable radial function and spherical harmonics.
PACE introduces the edge booster to take the TP of spherical harmonics with number of boosted times to obtain a higher degree polynomials, and concatenate all of them as the convolution.
Linear ACE and Allegro do not have convolution filters.

\begin{table}[h]
    \centering
    \caption{The convolution filters used in higher body order methods.}
    \begin{tabular}{c|c}
        \toprule[1pt]
        Methods & Convolution Filter, $F\left(\cdot\right)$ \\
        \midrule
        Linear ACE, Allegro & -\\
        MACE, PACE& $R_c^{\ell}\left(d_{ij}\right)Y_m^{\ell}\left(\frac{\bm{c}_{i} - \bm{c}_{j}}{d_{ij}}\right)$\\
        \bottomrule
    \end{tabular}\par
    \label{tab:higherbody_filter}
\end{table}

\vspace{0.1cm}\noindent\textbf{Message:} As shown in Table~\ref{tab:higherbody_message}, linear ACE builds messages as the product of a radial basis and spherical harmonics, where the radial basis $R_{c,z_iz_j}^{\ell}$ is coupled with atomic types of node $i$ and $j$. MACE and PACE build messages as the tensor product of the convolution filter and a linear transformation of node features. The linear transformations used by MACE are the same as in Section~\ref{equivariant_greater_1_methods}. 

Allegro builds messages by passing invariant features from the previous layer $\bm{x}_{ij}^{t-1}$ to an MLP, then multiplying the output with spherical harmonics.


\begin{table}[h]
    \centering
    \caption{The equations for message computing in higher body order methods.}
    \begin{tabular}{c|c}
        \toprule[1pt]
        Methods & Message, $\bm{m}_{ij}$\\
        \midrule
        Linear ACE & $R_{c,z_iz_j}\left(d_{ij}\right)Y_m^{\ell}\left(\frac{\bm{c}_{i} - \bm{c}_{j}}{d_{ij}}\right)$ \\
        MACE, PACE& $\mbox{TP}\left(F\left(\cdot\right), W\left(\bm{h}_j\right)\right)$ \\
        Allegro & $\mbox{MLP}\left(\bm{x}_{ij}^{t-1}\right)Y_m^{\ell}\left(\frac{\bm{c}_{i} - \bm{c}_{j}}{d_{ij}}\right)$ \\
        \bottomrule
    \end{tabular}\par
    \label{tab:higherbody_message}
\end{table}

\vspace{0.1cm}\noindent\textbf{Aggregation:} The aggregation functions used in higher body order methods are summarized in Table~\ref{tab:higherbody_aggre}. In linear ACE, MACE and PACE, the aggregation is implemented following atomic cluster expansion (ACE). According to ACE, message $m_{ij}$ is the 2-body particle basis for atoms $i$ and $j$. First, the two-body particle basis functions are summed over neighboring atoms to obtain the atomic basis for the central atom, $i$. Then, products of $v$ atomic basis functions create a $v+1$-body product basis. In order to preserve rotational equivariance, the product basis is multiplied with the generalized Clebsch-Gordan coefficients to form the symmetrized basis. Finally, the aggregated many-body message for each atom is constructed by a linear combination of symmetrized basis features with different body orders. Here, $\bm{\ell m}$ denotes $(\ell_1 m_1, \dots, \ell_v m_v)$ and an additional index $\eta_v$ is used to enumerate all paths of rotation orders $(\ell_1, \dots, \ell_v)$ which result in the desired output rotation order. In Allegro, messages from neighboring edges are summed around the central atom $i$ to obtain the aggregated message $m_i$ for atom $i$.

\begin{table}[h]
    \centering
    \caption{The aggregation functions used in higher body order methods.}
    \begin{tabular}{c|c}
        \toprule[1pt]
        Methods & Aggregation, $\bm{m}_{i}$\\
        \midrule
        Linear ACE & $\sum_{v}\sum_{\eta_v}W_{\eta_v}\sum_{\bm{\ell m}}\mathscr{C}_{\eta_v, \bm{\ell m}}^{\ell_o m_o}\prod_{\xi = 1}^{v}\sum_{j \in \mathcal{N}_i}\bm{m}_{ij}$ \\
        MACE & $\sum_{v}\sum_{\eta_v}W_{\eta_v}\sum_{\bm{\ell m}}\mathscr{C}_{\eta_v, \bm{\ell m}}^{\ell_o m_o}\prod_{\xi = 1}^{v} W \sum_{j \in \mathcal{N}_i}\bm{m}_{ij}$ \\
        PACE & $\sum_{v}\sum_{\eta_v}W_{\eta_v}\sum_{\bm{\ell m}}\mathscr{C}_{\eta_v, \bm{\ell m}}^{\ell_o m_o}\prod_{\xi = 1}^{v}W_{\xi} \sum_{j \in \mathcal{N}_i}\bm{m}_{ij}$ \\
        Allegro & $\sum_{j \in \mathcal{N}_i}\bm{m}_{ij}$ \\
        \bottomrule
    \end{tabular}\par
    \label{tab:higherbody_aggre}
\end{table}

\vspace{0.1cm}\noindent\textbf{Update:}  The update functions used in higher body order methods are summarized in Table~\ref{tab:higherbody_update}. 
Linear ACE does not need to update node features because it only has a single layer. MACE and PACE update the features for each node using the aggregated message and the node feature from the previous layer. In Allegro, the aggregated message $m_{i}$ is used to update two features of edge $e_{ij}$. The first is an equivariant feature $v_{ij}$ obtained by the tensor product of the aggregated message $m_{i}$ and feature $v_{ij}$ from the previous layer. Then, the invariant part of $v_{ij}$ is extracted to update the invariant feature $x_{ij}$ for edge $e_{ij}$. 

\begin{table}[h]
    \centering
    \caption{The update functions used in higher body order methods.}
    \begin{tabular}{c|c}
        \toprule[1pt]
        Methods & Update\\
        \midrule
        Linear ACE & - \\
        MACE, PACE & $\bm{h}_i^t = W\left(\bm{h}_i^{t-1}\right) + W\left(\bm{m}_i\right)$ \\
        \multirow{2}{*}{Allegro} & $\bm{v}_{ij}^t = \mbox{TP}\left(\bm{m}_i, \bm{v}_{ij}^{t-1}\right)$ \\
        & $\bm{x}_{ij}^t = \phi\left(\bm{x}_{ij}^{t-1} \| \bm{v}_{ij}^{t, \ell m = 00}\right)$ \\
        \bottomrule
    \end{tabular}\par
    \label{tab:higherbody_update}
\end{table}

\vspace{0.1cm}\noindent\textbf{Output:} The output modules of many-body methods differ from other equivariant methods. Higher body order methods first compute local energies for each atom, then the total energy of the molecule is the sum of local energies over all atoms. In Linear ACE, the invariant part of the aggregated message is extracted as the local energy for each atom. In MACE, the local energy of an atom is the sum of a fixed term and a learnable term. The fixed term is determined by atomic type and corresponds to the isolated energy of that atom, which is precomputed using DFT. The learnable term is derived from the invariant part of node features obtained in each layer. In Allegro, the invariant edge feature from the final layer is used to obtain pairwise energies. Next, pairwise energies around the central atom are scaled with a scaling factor that depends on both atom types, $\theta_{z_iz_j}$. These are then summed to obtain the local energy for the central atom. The output functions used in higher body order methods are summarized in Table~\ref{tab:higherbody_output}.

\begin{table}[h]
    \centering
    \caption{The output functions used in higher body order methods.}
    \begin{tabular}{c|c}
        \toprule[1pt]
        Methods & Output\\
        \midrule
        Linear ACE & $\sum_{i}\bm{m}_i^{\ell m = 00}$ \\
        MACE, PACE & $\sum_{i}\left[E_{\mbox{iso}, i} + \sum_{t=1}^{T-1}W\bm{h}_i^{t, \ell m=00} + \mbox{MLP}\left(\bm{h}_i^{T, \ell m = 00}\right)\right]$ \\
        Allegro & $\sum_{i}\sum_{j \in \mathcal{N}_i}\theta_{z_iz_j}\mbox{MLP}\left(\bm{x}_{ij}^T\right)$ \\
        \bottomrule
    \end{tabular}\par
    \label{tab:higherbody_output}
\end{table}

\subsubsection{Model Outputs} \label{sec:mol_rep_output}

Both invariant and equivariant methods should be able to deal with the symmetries for different tasks and applications. 
Invariant methods can produce $SE(3)$-invariant features directly, and some equivariant features may also be achieved based on the final invariant features. For example, in order to predict per-atom forces that are $SO(3)$-equivariant, invariant methods first predict the energy $E$ and then use the gradient of the energy w.r.t. atom positions $\bm{f}_i=-\frac{\partial E}{\partial\bm{c}_i}$ to compute the force of each atom, which can ensure energy conservation. Here $\bm{c}_i$ is the coordinate of node $i$. Equivariant methods can also use the predicted energy to compute forces or predict forces directly. For other equivariant prediction targets like the Hamiltonian matrix discussed in Section~\ref{sec:dft}, additional operations are necessary for invariant models to ensure equivariance, making equivariant methods more straightforward and suitable for such tasks.



\subsubsection{Datasets and Benchmarks} 


\begin{table}[t]
	\centering
	\caption{Statistics of QM9~\cite{ramakrishnan2014quantum}, MD17~\cite{chmiela2017machine}, rMD17~\cite{christensen2020role}, MD17@CCSD(T)~\cite{chmiela2018towards}, ISO17~\cite{schutt2018schnet}, and Molecule3D~\cite{xu2021molecule3d} datasets. We summarize the prediction tasks and the number of 3D molecule samples (\# Samples), maximum number of atoms in one molecule (Maximum \# atoms), and average number of atoms in one molecule (Average \# atoms).}
	\resizebox{0.95\textwidth}{!}{
        \begin{tabular}{l|ccccc}
        \toprule
        Datasets & Prediction Tasks &\# Samples & Maximum \# atoms & Average \# atoms \\\midrule
            QM9 &Predict energetic, electronic, and thermodynamic properties &130,831 &29 &18.0 \\
            MD17 &Predict energy and force &- &- &- \\
            rMD17 &Predict energy and force &- &- &- \\
            MD17@CCSD(T) &Predict energy and force &- &- &- \\
            ISO17 &Predict energy and force &645,000 &19 &19 \\
            Molecule3D &Predict 3D geometry and energetic and electronic properties &3,899,647 &137 &29.1 \\
        \bottomrule
	\end{tabular}}
	\label{tab:mol_rep_data}
\end{table}
Molecular representation learning methods are evaluated on various tasks, such as \revisionOne{quantum chemistry property prediction, energy prediction, and per-atom force prediction}. Table~\ref{tab:mol_rep_data} summarizes commonly used datasets, including QM9~\cite{ramakrishnan2014quantum}, MD17~\cite{chmiela2017machine}, rMD17~\cite{christensen2020role}, MD17@CCSD(T)~\cite{chmiela2018towards}, ISO17~\cite{schutt2018schnet}, and Molecule3D~\cite{xu2021molecule3d}. Typically, the mean absolute error and mean square error between the predicted and ground-truth values are used as evaluation metrics.

Specifically, QM9 dataset~\cite{ramakrishnan2014quantum} collects more than 130k small organic molecules with up to nine heavy atoms (CONF) from GDB-17 database~\cite{ruddigkeit2012enumeration}. For each molecule, the dataset provides its 3D geometry for the stable state (minimal in energy), along with corresponding harmonic frequencies, dipole moments, polarizabilities, energies, enthalpies, and free energies of atomization. All properties were calculated at the B3LYP/6-31G(2df,p) level of quantum chemistry. Typically, a separate model is trained for each property.

MD17 dataset~\cite{chmiela2017machine} includes molecular dynamic simulations of 8 small organic molecules, namely, aspirin, benzene, ethanol, malonaldehyde, naphthalene, salicylic acid, toluene, and uracil. For each molecule, the dataset provides hundreds of thousands of conformations and corresponding energies and forces. 
Revised MD17 (rMD17) dataset~\cite{christensen2020role} is a recomputed version of MD17 to reduce numerical noise. For each molecule in the original MD17, 100,000 structures are taken, and the energies and forces are recalculated at the PBE/def2-SVP level of theory using very tight SCF convergence and very dense DFT integration grid. Therefore, the dataset is practically free from numerical noise.
MD17@CCSD(T)~\cite{chmiela2018towards} is calculated based on the more accurate and expensive CCSD or CCSD(T) method and contains fewer molecules.
Typically, for MD17, rMD17, ad MD17@CCSD(T), a separate model is trained for each molecule, with the task of predicting the energy and force for each conformation.
To ensure energy conservation, most methods compute the per-atom force from the predicted energy, as discussed in Section~\ref{sec:mol_rep_output}, and the commonly used loss function is a combination of the energy loss and force loss
\begin{equation}
    \begin{aligned}
    \mathcal{L}= \lambda_{E} \mathcal{L}_{E}(\hat{E}, E) + \lambda_{f} \mathcal{L}_{f}(-\frac{\partial \hat{E}}{\partial C}, \bm{f}).
    \end{aligned}
    \label{eq:e_f_loss}
\end{equation}
Here, $\hat{E}$ is the predicted energy, $-\frac{\partial \hat{E}}{\partial C}$ is the computed force, $C$ is the atom coordinate matrix, $E$ and $\bm{f}$ are the ground-truth energy and force, $\mathcal{L}_{E}$ and $\mathcal{L}_{f}$ are energy and force loss functions, such as mean absolute error and mean square error, $\lambda_{E}$ and $\lambda_{f}$ are the weights for energy and force losses,  which are often set to 1 and 1000, respectively.

ISO17~\cite{schutt2018schnet} differs from MD17 in that it includes both chemical and conformational changes. The dataset contains molecular dynamics trajectories of 129 isomers with the same composition of $\text{C}_7\text{O}_2\text{H}_{10}$. Each trajectory consists of 5,000 conformations, resulting in a total of 645,000 samples. Unlike MD17 where a separate model is usually trained for each molecule, for ISO17, a typical setting is that a single model is trained across all 129 different molecules.

\revisionOne{Molecule3D~\cite{xu2021molecule3d} is a large-scale dataset with around 4 million molecules curated from PubChemQC~\cite{nakata2017pubchemqc}. For each molecule, the dataset provides its precise ground-state 3D geometry derived from DFT at the B3LYP/6-31G* level, as well as molecular properties such as energies of the highest occupied molecular orbital (HOMO) and the lowest unoccupied molecular orbital (LUMO), the HOMO-LUMO gap, and total energy. Although Molecule3D is primarily designed for predicting 3D geometries from 2D molecular graphs (Section~\ref{subsec:mol_conformer_generation}), in this subsection, we can directly take the ground-truth 3D geometries as input and test models' performance on property prediction tasks. It is worth noting that PCQM4Mv2 dataset~\cite{hu2020open,hu2021ogb} is curated from PubChemQC as well, but only provides 3D geometries for the training data. Therefore, it is often used for tasks such as predicting properties from 2D molecules~\cite{ying2021transformers}, pre-training~\cite{zaidi2023pretraining}, and 2D-3D joint-training~\cite{luo2023one}.}

In addition to the datasets mentioned above, some of the molecular representation learning methods discussed in this section are also evaluated on larger molecules such as proteins, materials, DNA, and RNA. Example datasets include MD22~\cite{chmiela2023accurate}, Atom3D~\cite{townshend2020atom3d}, and OC20~\cite{chanussot2021open}. \revisionOne{Note that the datasets we discussed above mainly include properties directly related to 3D molecular structures, where different molecular conformers can lead to varying properties. However, there are also critical molecular properties, such as ADMET properties, that typically rely on 2D molecular representations. Enhancing such property prediction by incorporating 3D information remains a crucial area of research. In addition, more datasets for molecular interactions, particularly the interactions between small molecules and proteins or materials, are introduced in Section~\ref{sec:dock}.}

\subsubsection{Open Research Directions} \label{subsec:mol_rep_open}

\vspace{0.1cm}\noindent\textbf{Learning from Both 2D and 3D Information:} 
Despite recent advances in molecular representation learning, several challenges require further exploration. One direction is the joint training from both 2D and 3D information of molecules~\cite{stark20223d, luo2023one}. While 3D information is crucial for accurately modeling the physical properties of molecules, it can be computationally expensive to calculate and hard to obtain experimentally. On the other hand, 2D information, such as the molecular graph, is computationally efficient to generate, but may not capture all the necessary information for accurate predictions. Therefore, exploring methods that can be jointly trained on both 2D and 3D information or transfer knowledge between 2D and 3D representations could lead to improved performance as well as efficiency in tasks such as property prediction and drug discovery. In addition, pre-training~\cite{zhou2023unimol} can further improve the generalization of models.

\vspace{0.1cm}\noindent\textbf{Expressivity and Computational Efficiency:} 
On the theoretical front, a challenge is developing provably expressive 3D GNNs that capture geometric interactions among atoms in a \emph{complete} or universal manner \cite{pozdnyakov2020incompleteness}, as elaborated in Section~\ref{subsec:universality_of_equiv_arch}.
Towards this goal, \citet{joshi2023expressive} provides a theoretical upper bound on the expressive power of geometric GNNs in terms of discriminating non-isomorphic geometric graphs, and shows that equivariant layers which propagate geometric information are more expressive than invariant ones, in general.
They identify key design choices for building maximally powerful equivariant GNNs: (1) depth, (2) tensor order, and (3) scalarization body order. As highlighted in this section, body order controls how well a network manages to capture the local geometry in a neighbourhood of a node, while higher tensor order enables a network to have higher angular resolution when representing geometric information. Finally, network depth controls the receptive field of an architecture, and in many current equivariant architectures increased depth implicitly also leads to increased body order.

While increasing all three properties theoretically improves the expressive power of geometric GNNs, several practical challenges hinder provably expressive models. Computing higher-order tensors \cite{passaro2023reducing} and many-body interactions \cite{batatia2022mace} scales the compute cost drastically, often limiting practically used networks to tensor order $l\leq2$ and body order $\nu \leq 4$. Further, there is early evidence of geometric oversquashing with increasing depth \cite{alon2021on}.
Future research may focus on improving the efficiency of many-body and higher-order equivariant GNNs to scale to larger biomolecules as well as larger datasets. 

\vspace{0.1cm}\noindent\textbf{Invariant Versus Equivariant Message Passing:} 
Invariant GNNs are significantly more scalable than equivariant GNNs and can be as powerful when working with fully connected graphs \citep{joshi2020transformers} and pre-computing non-local features \citep{gasteiger2021gemnet, wang2022comenet}. 
Similarly, some invariant GNNs build canonical reference frames to convert equivariant quantities into scalar features \citep{du2022se, duval2023faenet}, allowing non-linearities on all intermediate representations in the network.
Investigating the trade-offs between invariant and equivariant message passing is another fruitful avenue of research on molecular representation learning.
%

\subsection{Molecular Conformer Generation} \label{subsec:mol_conformer_generation}

\noindent{\emph{Authors: Zhao Xu, Yuchao Lin, Minkai Xu, Stefano Ermon, Shuiwang Ji}\vspace{0.3cm}\newline\emph{Recommended Prerequisites: Section~\ref{subsec:mol_representation_learning}}}\newline



As discussed in Section~\ref{subsec:mol_representation_learning}, the role of 3D molecular geometries in molecular representation learning is integral as they significantly enhance the accuracy of property prediction compared to the use of 2D graphs solely. This enhancement of 3D information is attributed to the fact that the physical configuration of a molecule largely influences its numerous properties. For example, isomers with identical atomic compositions can have vastly different melting points due to variations in their molecular structures. In immunology, the shape of antibodys' binding site, specifically the complementarity determining regions (CDRs), precisely determines the antigen they can recognize and bind to, which is critical for immune response. Hence, spatial information of molecules is highly desirable when working on real-world applications such as molecular property prediction, molecular dynamics, and molecule-protein docking. However, the acquisition of accurate 3D geometries through Density Functional Theory (DFT) is significantly challenging due to its high computational cost, thus limiting the widespread application of 3D molecular geometries. Consequently, the employment of machine learning models for the reconstruction of 3D molecular geometries emerges as a promising alternative, offering the potential to mitigate computational cost and make 3D geometries more accessible.

\subsubsection{Problem Setup}
Let the total number of atoms in the molecule be $n$. A 2D molecule is represented as $\mathcal{G}=(\bm{z}, E)$, where $\bm{z}=[z_1,...,z_n]\in\mathbb{Z}^n$ denotes atom type vector, and each $e_{ij}\in\mathbb{Z}$ in $E$ denotes the edge type between nodes $i$ and $j$. For a given 2D molecule $\mathcal{G}$, the corresponding 3D molecule further needs 3D geometries $C=[\bm{c}_1,...,\bm{c}_n]\in\mathbb{R}^{3\times n}$ where $\bm{c}_i$ denotes the 3D coordinate of the $i$-th atom. One form of  $C$ is associated with a potential energy, sampled from the potential energy surface corresponding to the Boltzmann distribution, which dictates that states of lower potential energy are more probable in a given environment. Geometries that correspond to lower energy or high probability states are generally more stable and thus, are more likely to be corroborated by experimental observations. The geometry that minimizes the potential energy or maximizes the distribution, known as the equilibrium ground-state geometry, is the most stable and critical one. The problem of molecular geometry reconstruction can be bifurcated into two distinct tasks. The first task, referred to as 3D geometry generation, involves training a generative model, denoted as $f_G$, with the aim of understanding the distribution $p(C|\mathcal{G})$ of low-energy geometries given the conditional 2D molecular graph $\mathcal{G}$. On the other hand, the second task, known as 3D geometry prediction, seeks to train a predictive model $f_P$ that is capable of directly estimating the equilibrium ground-state geometry $C_{eq}$ based on its corresponding 2D graph $\mathcal{G}$.

\subsubsection{Technical Challenges}
The reconstruction of 3D molecular geometries from 2D molecular graphs poses three major challenges. The first challenge is to ensure that the obtained conformers are geometrically valid in 3D space. For instance, it is possible for symmetric graph nodes to have identical embeddings due to the permutation invariance inherent to GNNs, leading to invalid geometries. Therefore, it is essential to distinguish these symmetric atoms and enforce their reconstructed coordinates are distinct because atoms should not overlap in 3D space. Besides, existing works~\citep{simm2019generative, xu2021molecule3d} consider Distance Geometry (DG) first and then reconstruct atom coordinates based on the distance matrix. In such cases, ensuring the 3D geometric validity of atom coordinates becomes particularly challenging due to the potential for the derived distance matrix to fail in constituting a valid Euclidean Distance Matrix (EDM). In addition to maintaining the 3D geometric validity of reconstructed conformers, the second challenge is to meet the chemical validity imposed on conformer fragments. For instance, aromatic rings or $\pi$ bonds restrict all their atoms on a planar surface, while many macrocycles and small rings are non-planar~\cite{wang2020improving}. It is desirable to have reconstructed geometries obeying such quantum rules and being chemically valid. An additional challenge in reconstructing 3D molecular geometry arises from the inherent symmetry of the geometry density function. Given initial systems with zero centers of mass (CoM)~\citep{xu2022geodiff, kohler2020equivariant}, the generative geometry distribution of conformers is often modeled as an invariant distribution in order to draw asymptotically unbiased samples with respect to the ground truth distribution~\citep{kohler2020equivariant}. Specifically, we must ensure the reconstructed conformer is subject to $SE(3)$-invariant position distributions. Let rotation matrix $R\in SO(3)\subset \mathbb{R}^{3\times 3}$ and translation vector $\bm{t}\in \mathbb{R}^{3}$ and $\bm{1}\in \mathbb{R}^n$, the $SE(3)$-invariant position distributions require $p(RC+\bm{t}\bm{1}^T|G)=p(C|G)$. In other words, rotated or translated conformers are regarded as identical because geometry reconstruction is independent of rotation and translation.

\begin{table}[t]
	\centering
	\caption{Summary of 3D outputs, model architecture, and distribution symmetry of several representative 3D molecular conformation generation methods. Among the various methods, CVGAE~\cite{mansimov2019molecular}, GraphDG~\cite{simm2019generative}, ConfVAE~\cite{xu2021end}, DMCG~\cite{zhu2022direct} use conditional variational autoencoder to generate molecular conformers, where CVGAE and DMCG directly generate 3D coordinates, and GraphDG and ConfVAE generate interatomic distances of molecular conformers. In the spirit of ConfVAE, CGCF~\cite{xu2021learning}  generates interatomic distances by taking advantage of the flow generative model. Moreover, ConfGf~\cite{Shi2021LearningGF} and GeoDiff~\cite{xu2022geodiff} employ zero-centering $E(3)$-equivariant models to directly generate 3D coordinates, and achieve an $E(3)$-invariant generative distribution. Conversely, Torsional Diffusion~\cite{jing2022torsional} applies an $SE(3)$-invariant diffusion model to generate torsions exclusively, preserving local structures such as bond length and angle, which are generated by RDKit. Additionally, GeoMol~\cite{ganea2021geomol}, DeeperGCN-DAGNN+Distance~\cite{xu2021molecule3d}, and EMPNN~\citep{xu20233d} implement predictive strategies for the generation of molecular conformers. }
	\begin{tabular}{l|ccc}
		\toprule[1pt]
		Methods & 3D Outputs & Architecture & Distribution Symmetry \\
		\midrule
        CVGAE & Coordinates & VAE & -\\
        DMCG & Coordinates & VAE & $SE(3)$-Invariant\\
        GraphDG & Distances & VAE & $E(3)$-Invariant\\
        ConfVAE & Distances & VAE & $E(3)$-Invariant\\
        CGCF & Distances & Flow & $E(3)$-Invariant\\
		ConfGF & Coordinates & Score Matching & $E(3)$-Invariant\\
		GeoDiff & Coordinates & Diffusion & $E(3)$-Invariant \\
		Torsional diffusion & Torsions & Diffusion & $SE(3)$-Invariant \\
  GeoMol & Coordinates & Predictive Model &  $SE(3)$-Invariant \\
		EMPNN & Coordinates & Predictive Model &  -\\
            DeeperGCN-DAGNN+Dist & Distances & Predictive Model &  $E(3)$-Invariant\\
		\bottomrule
	\end{tabular}
	\label{tab:mol_conf_gen}
\end{table}

\subsubsection{Existing Methods}

\vspace{0.1cm}\noindent\textbf{Generation-Based Methods:} While there are a plethora of molecular generative models available, this section focuses exclusively on those that represent recent and significant contributions to the field. Many earlier generative models, such as CVGAE~\cite{mansimov2019molecular}, GraphDG~\cite{simm2019generative}, ConfVAE~\cite{xu2021end} and CGCF~\cite{xu2021learning}, DMCG~\cite{zhu2022direct}, are developed based on variational autoencoders (VAEs) or flow model as their fundamental theory. On the other hand, current state-of-the-art generative models mostly rely on score matching and probabilistic denoising diffusion models with $E(3)$-equivariant/invariant modules. For instance, ConfGF~\cite{Shi2021LearningGF} develops a 3D generative model that uses score matching and obtains scores by computing chain rule derivatives from positions to distances. To generate the position of each atom in a molecule, ConfGF applies an $E(3)$-equivariant model to update atom positions during sampling. GeoDiff~\cite{xu2022geodiff} further extends ConfGF's capabilities by incorporating a zero-centering $E(3)$-equivariant diffusion probabilistic model. Although both methods can directly generate 3D coordinates of molecules, they do not consider chemical constraints, such as aromatic rings in which all atoms are at the same plane. As a result, they may produce chemically invalid conformers. In contrast, torsional diffusion~\cite{jing2022torsional, corso2023particle} employs an $SE(3)$-invariant diffusion model only to adjust conformers' torsions while retaining all local structures like bond length and angle generated by RDKit. By doing so, it takes advantage of the chemical knowledge introduced by RDKit. However, this approach heavily depends on RDKit-generated conformers and cannot refine local ring structures, such as macrocycles or small non-planar rings. Based on torsional diffusion, DiffDock~\cite{corso2022diffdock} advances protein-ligand docking as a conformer generation process conditioned on proteins. We describe the details of DiffDock in Section~\ref{sec:dock}.

\vspace{0.1cm}\noindent\textbf{Prediction-Based Methods:} Alternatively, other existing works formulate the task as a predictive task, which focuses on predicting the equilibrium ground-state conformer. One such example is GeoMol~\cite{ganea2021geomol}, which employs message passing neural networks (MPNNs) with geometric constraints to predict local structures to generate diverse conformers. To ensure geometrical validity, GeoMol applies a matching loss that effectively distinguishes symmetric atoms by searching for the best matching substructures to ground truths among all possible permutations of symmetric nodes. Another notable approach is DeeperGCN-DAGNN+Distance, as proposed in~\cite{xu2021molecule3d}. This method aims to predict the full distance matrix and then directly use the distance matrix in downstream tasks because pairwise distances implicitly provide 3D information. On the contrary, EMPNN~\cite{xu20233d} uses node indices to break node symmetries and explicitly outputs geometrically valid 3D coordinates of the ground-state conformer. 

To provide a comprehensive overview of the various generative and predictive methods in the field, we summarize representative approaches in Table~\ref{tab:mol_conf_gen}.

\subsubsection{Datasets and Benchmarks}

Geometric Ensemble Of Molecules (GEOM)~\cite{axelrod2022geom} is a dataset of high-quality molecular conformers, where the 3D molecular structures are first initialized by RDKit~\cite{rdkit} and then optimized by ORCA~\cite{neese2012orca} and CREST~\cite{grimme2019exploration} programs. It contains two subsets, GEOM-QM9 and GEOM-Drugs, which are the commonly used benchmark datasets for evaluating the performance of different molecular conformation generation methods. 
GEOM-QM9 dataset contains 133,258 small organic molecules from the original QM9 dataset~\cite{ramakrishnan2014quantum}. All molecules in this dataset have up to 9 heavy atoms and 29 total atoms including hydrogen atoms, with very smaller molecular mass and few rotatable bonds. The atomic types of heavy atoms are limited to carbon, nitrogen, oxygen, and fluorine. 
By contrast, GEOM-Drugs contains larger drug-like molecules, with a mean of 44.4 atoms (24.9 heavy atoms) and a maximum of 181 atoms (91 heavy atoms). Most importantly, these large molecules contain significant flexibility, \emph{e.g.}, with an average of 6.5 and up to a maximum of 53 rotatable bonds, which is challenging for learning the molecular conformation generation models.

\subsubsection{Open Research Directions}


While significant progress has been made in ML-based molecular conformation generation models, there remain several promising directions for future research. Firstly, the current benchmark datasets are all simulated in the vacuum environment, while in reality, molecular conformations are different in surrounding solvent environments. Future research could focus on incorporating solvent effects into the generative models, allowing them to generate conformations that reflect the realistic behavior of molecules in given solvent environments. Second, learning the conformation generative models often requires large amounts of training data. However, in many cases, limited data is available for specific classes of molecules or compounds. Future research could also explore transfer learning and few-shot learning techniques to leverage knowledge learned from a broader set of molecules and apply it to generate conformations for less-studied or novel compounds. This could significantly reduce the data requirements and improve the generalization capabilities of the models. Thirdly, existing methods mainly focus on the generation of low-energy conformers due to their stability. However, exploring molecular conformers in high-energy transition states (TS) is equally significant, as they are pivotal to the progress of chemical reactions~\cite{choi2023prediction, duan2023accurate}. Hence, future research could also concentrate on generating the TS structures for reactants and products, facilitating an enhanced understanding of the kinetics and mechanisms of chemical reactions. Addressing these directions will significantly advance the field and contribute to the development of more effective, efficient, and practically relevant molecular conformation generation methods.

\subsection{Molecule Generation from Scratch}
\label{subsec:mol_generation}




\noindent{\emph{Authors: Youzhi Luo, Minkai Xu, Stefano Ermon, Shuiwang Ji}\vspace{0.3cm}\newline\emph{Recommended Prerequisites: Section~\ref{subsec:mol_representation_learning}}}\newline

In Section~\ref{subsec:mol_representation_learning}, we study the problem of property predictions from given molecules. However, some other real-world problems, such as designing novel molecules for drugs, require us to model the reverse process, \emph{i.e.}, to obtain target molecules with given properties. Exhaustively searching target molecules in chemical space is impossible because the number of candidate molecules can be very large, \emph{e.g.}, there are around $10^{33}$ drug-like molecules~\cite{polishchuk2013estimation} in estimation. Recently, the significant progress in deep generative learning has motivated many researchers to generate novel molecules with advanced deep generative models, including variational auto-encoders (VAEs)~\cite{kingma2014auto}, generative adversarial networks (GANs)~\cite{goodfellow2014gen}, flow models~\cite{danilo2015variational} and diffusion models~\cite{ho2020denoising}. Some early studies~\cite{jin2018junction,you2018goal,shi2020graphaf} generate molecules in the form of 2D molecular graphs. However, these methods do not generate 3D coordinates of atoms in molecules, so they cannot distinguish molecules with the same 2D graphs but different 3D geometries, such as spatial isomers. Actually, many molecular properties, such as quantum properties or biological activities, are determined by 3D geometries of molecules. Hence, in this section, we focus on the 3D molecule generation problem. 

\subsubsection{Problem Setup} 

We represent a 3D molecule with $n$ atoms as $\mathcal{M}=(\bm{z},C)$, where $\bm{z}=[z_1,...,z_n]\in\mathbb{Z}^n$ is the atom type vector and $C=[\bm{c}_1,...,\bm{c}_n]\in\mathbb{R}^{3\times n}$ is the atom coordinate matrix. Here, for the $i$-th atom, $z_i$ is its atomic number and $\bm{c}_i$ is its 3D Cartesian coordinate. Our target is to learn a probability distribution $p$ over the 3D molecule space with generative models and sample novel 3D molecules from $p$. Note that different from the molecular conformer generation or prediction problem discussed in Section~\ref{subsec:mol_conformer_generation}, we do not generate 3D molecules from any conditional inputs like 2D molecular graphs, but instead generate them from scratch.

\subsubsection{Technical Challenges}

The central challenge in generating 3D molecules from scratch lies in achieving $SE(3)$-invariance when generating 3D atom positions. In other words, the generative model should assign the same probability to $\mathcal{M}$ and $\mathcal{M}^{\prime}$ if $\mathcal{M}^{\prime}$ can be obtained by rotating or translating $\mathcal{M}$ in 3D space. Generally, there are two strategies to generate 3D molecular structures. First, generative models may directly use the coordinate matrix as the generation targets or outputs. But the challenge is that the probabilistic modeling for coordinate matrices should be carefully designed to ensure invariance to $SE(3)$ transformations. Second, instead of directly generating coordinates, generative models may take some $SE(3)$-invariant 3D features, such as distances or angles, as generation targets. This strategy removes the necessity of explicitly considering $SE(3)$-invariance in generative models, but requires the generated 3D features to have valid values and complete 3D structure information so that 3D atom coordinates can be reconstructed from them.

\subsubsection{Existing Methods}

Three representative methods of directly generating coordinate matrices of 3D molecules are E-NFs~\cite{satorras2021en}, EDM~\cite{hoogeboom2022equivariant}, and GeoLDM~\citep{xu2023geometric}. They adopt multiple strategies to achieve $E(3)$-invariance, where $E(3)$ is a superset of $SE(3)$, including translation, rotation, and reflection. Specifically, to remove the freedom of translation, any 3D molecules are always zero-centered by reducing the centroid, \emph{i.e.}, the averaged 3D coordinates over all atoms, from each column of the coordinate matrix before being passed into generative models. In other words, the probability density captured by these approaches is non-zero only on coordinate matrices with zero centroids. In addition, the probability density of zero-centered coordinates is calculated by their corresponding latent variables, which are subject to CoM-free Gaussian distribution~\cite{kohler2020equivariant}. Mathematically, CoM-free Gaussian distribution ensures that the probability density is invariant to rotation and reflection. Flow and diffusion models are used to map between zero-centered coordinate matrices and latent prior variables in E-NFs and EDM, respectively. GeoLDM further proposes to first encode the zero-centered atoms into a zero-centered latent space, where each atom is represented with latent invariant features and latent equivariant coordinates. Then instead of the original coordinate matrices, GeoLDM learns to map between the latent variable and prior Gaussian distribution via latent diffusion models~\cite{stablediffusion}.

\begin{table}[t]
	\centering
	\caption{Summary of 3D outputs, model architecture, generation pipeline, distribution symmetry of several representative 3D molecule generation methods. Among these methods, E-NFs~\cite{satorras2021en}, EDM~\cite{hoogeboom2022equivariant}, and GeoLDM~\citep{xu2023geometric} directly generate 3D coordinates of atoms in molecules. They achieve $E(3)$-invariant by zero-centering coordinates and using CoM-free Gaussian distribution. On the other hand, EDMNet~\cite{hoffmann2019generating}, G-SchNet~\cite{gebauer2019symmetry}, and G-SphereNet~\cite{luo2022an} implicitly generate 3D positions of atoms by distances, angles, and torsion angles that are invariant to rotations and translations.}
	\resizebox{0.95\textwidth}{!}{
	\begin{tabular}{l|cccc}
		\toprule[1pt]
		Methods & 3D Outputs & Architecture & Pipeline & Distribution Symmetry\\
		\midrule
		E-NFs & Coordinates & Flow & One-shot & $E(3)$-Invariant \\
		EDM & Coordinates & Diffusion & One-shot & $E(3)$-Invariant \\
		EDMNet & Distances & GAN & One-shot & $E(3)$-Invariant \\
		G-SchNet & Distances & Autoregressive model & Autoregressive & $E(3)$-Invariant \\
		G-SphereNet & Distances + Angles + Torsion angles & Flow & Autoregressive & $SE(3)$-Invariant \\
        GeoLDM & Coordinates & Latent Diffusion & One-shot & $E(3)$-Invariant \\
		\bottomrule
	\end{tabular}
    }
	\label{tab:mol_gen}
\end{table}

In contrast to E-NFs and EDM, some methods implicitly generate 3D atom positions from $SE(3)$-invariant features. To represent complete structural information of a 3D molecule, one alternative to the coordinate matrix is Euclidean distance matrix that contains distances between every pairwise atoms in the molecule. EDMNet~\cite{hoffmann2019generating} is the first work studying generating 3D molecular structures in the form of Euclidean distance matrices. In EDMNet, various of novel loss functions are used to train a GAN model to generate valid Euclidean distance matrices so that 3D Cartesian coordinates can be successfully reconstructed. Different from one-shot methods like EDMNet, other methods adopt an autoregressive procedure to generate 3D molecules through step-by-step placing atoms in 3D space. Two representatives of autoregressive methods are G-SchNet~\cite{gebauer2019symmetry} and G-SphereNet~\cite{luo2022an}. In both methods, a complete 3D molecule is generated by multiple steps, and only one new atom is generated and placed to the local region of a reference atom at each generation step. Specifically, G-SchNet places the new atom to one of the candidate grid positions of the reference atom through sampling from distance distributions predicted by an autoregressive generative model. On the other hand, G-SphereNet generates distances, line angles, and torsion angles by autoregressive flow models to determine the relative position of the new atom to the reference atom. Because of the use of torsion angles, G-SphereNet captures $SE(3)$-invariant distributions. We summarize the key information of discussed 3D molecule generation methods in Table~\ref{tab:mol_gen}. \revisionOne{Note that among these discussed methods, only EDM can take molecular properties as conditional input and perform property-oriented generation, while other methods can only use implicit strategies to generate molecules with desirable properties, such as optimizing latent representations. Some other methods may consider more complicated conditional inputs, such as protein pockets~\cite{ragoza2022generating,liu2022generating}, which is introduced in Section~\ref{sec:dock}.}

\subsubsection{Datasets and Benchmarks}


Two benchmark datasets, QM9~\cite{ramakrishnan2014quantum} and GEOM-Drugs~\cite{axelrod2022geom}, are commonly used to evaluate the performance of different 3D molecule generation methods. QM9 dataset collects more than 130k small organic molecules from GDB-17~\cite{ruddigkeit2012enumeration} database. All molecules in QM9 have up to 9 heavy atoms (29 atoms including hydrogen atoms), and the element type of any heavy atom is always one of carbon, nitrogen, oxygen, and fluorine. The 3D atom coordinates of molecules in QM9 are calculated at the B3LYP/6-31G (2df, p) level of quantum chemistry by Gaussian software~\cite{2009Gaussian}. In addition to QM9, GEOM-Drugs is another dataset used to evaluate the performance of 3D molecule generation methods in generating larger and more complicated drug molecules. It collects 430k drug-like molecules with up to 181 atoms. The 3D atom coordinates of molecules in GEOM-Drugs are first initialized by RDKit~\cite{rdkit}, then optimized by ORCA~\cite{neese2012orca} and CREST~\cite{grimme2019exploration} software. In both datasets, the 3D coordinates of atoms in molecules are calculated by DFT. 

\subsubsection{Open Research Directions}

Despite that a lot of 3D molecule generation methods have been proposed in recent years, there exist several challenges hampering them to generate practically useful 3D molecules. First, most existing methods consider $E(3)$ symmetries, not $SE(3)$ symmetries, so they are also invariant to reflection. This invariance should be avoided in many biological and chemical applications where generative models are expected to discriminate 3D molecules with different chiralities. Additionally, it is crucial for the generated 3D molecules to meet chemical constraints in 3D positions of some local structures so that they are chemically valid and synthesizable. For instance, all atoms in a benzene ring are restricted to be in the same plane. However, it still remains challenging and under-explored to design generative models that satisfy all chemical constraints.

\subsection{Molecular Dynamics Simulation}

\noindent{\emph{Authors: Xiang Fu, Tommi Jaakkola}\vspace{0.3cm}\newline\emph{Recommended Prerequisites: Section~\ref{subsec:mol_representation_learning}}}\newline

Since its development in the 1950s, molecular dynamics (MD) simulation has evolved into a well-established and valuable technique for gaining atomistic insights into a wide range of physical and biological systems~\citep{alder1959studies, rahman1964correlations, frenkel2001understanding, schlick2010molecular, tuckerman2010statistical}. Through MD simulations, researchers can effectively characterize the potential energy surface (PES) that underlies the system and calculate macro-level observables based on the resulting MD trajectories. These observables play a crucial role in determining important material properties, such as the diffusivity of battery materials~\citep{webb2015systematic}, and provide valuable insights into physical mechanisms, such as protein folding kinetics~\citep{lane2011markov, lindorff2011fast}. However, the practical applicability of MD simulations is limited due to their high computational cost. This cost arises from two main factors: Firstly, in many applications that demand high accuracy, the energy and forces must be determined using quantum chemistry methods, which involve approximately solving the computationally expensive Schr\"{o}dinger equation (Section~\ref{sec:dft}). Secondly, when studying large and intricate systems like polymers and proteins, extensive simulations spanning nanoseconds to milliseconds are often necessary to investigate specific physical processes, while the time step size required for numerical stability is often at the femtosecond level. Conducting such simulations, even with less accurate classical force fields, incurs substantial computational expenses. 
In recent years, machine learning (ML) approaches have shown promise to accelerate MD simulations substantially. This section provides a brief overview of some forefronts of ML methods applied to MD simulations, encompassing ML force fields, ML augmented sampling methods, and ML-based coarse-graining methods. While we categorize this subsection under AI for small molecules, it is important to note that MD simulation is a versatile computational technique applicable to a wide range of molecules, including small organic molecules, biological macromolecules, and materials.

\begin{figure}[t]
    \centering
    \includegraphics[width=\textwidth]{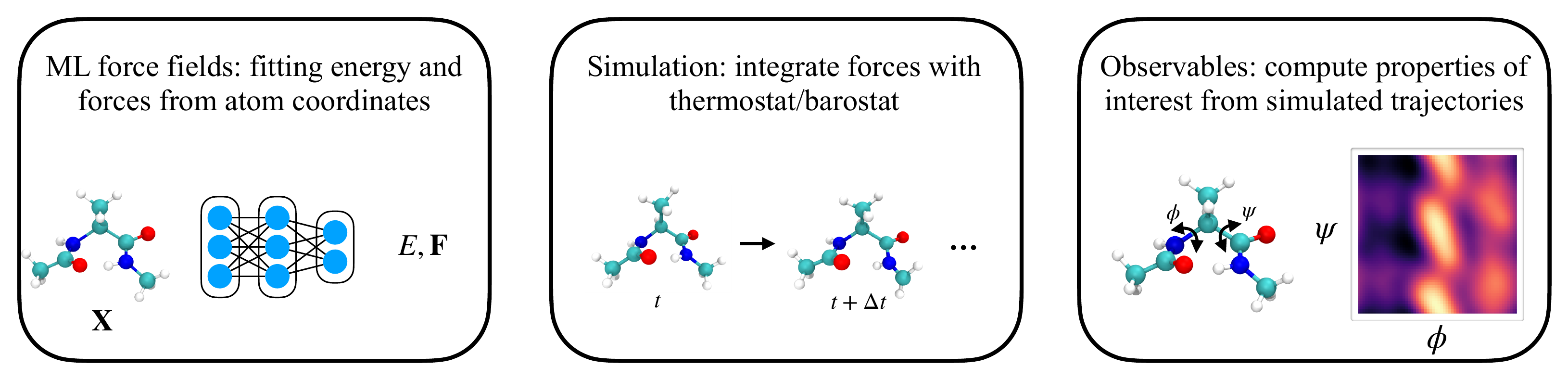}
    \caption{Simulating molecular dynamics with machine learning. To replace expensive quantum mechanical calculations, an ML force field is learned to predict energy and forces from atomic coordinates. With a learned force field, we can simulate MD by pairing it with an appropriate thermostat/barostat. From the simulated trajectories, properties of interest can be computed.}\label{fig:mlff_md}
\end{figure}

\subsubsection{Problem Setup}

Simulating molecular dynamics involves integrating Newton's equation of motion: $\frac{d^2 \bm x}{dt^2} = \bm m^{-1}\bm F(\bm x)$. The forces necessary for this integration are obtained by differentiating a potential energy function: $\bm F (\bm x) = -\frac{\partial E(\bm x)}{\partial \bm x}$. Here, $\bm x$ represents the state configuration, $\bm m$ represents the mass of the atoms, and $\bm F$ and $E$ represent the force and potential energy function, respectively. To replicate desired thermodynamic conditions, such as constant temperature or pressure, an appropriate thermostat or barostat is selected to augment the equation of motion with additional variables. The choice of these conditions depends on the specific system and task at hand. Through the simulation, a time series of positions $\{\bm x_t \in \mathbb{R}^{N \times 3}\}_{t = 0}^T$ (and velocities) is generated, where $t$ denotes the temporal order index and $T$ represents the total number of simulation steps, $N$ is the number of atoms in the molecule. From the time series, observables $O({\bm x_t})$ such as radial distribution functions (RDFs), virial stress tensors, mean-squared displacement (MSD), and free energy surfaces with respect to key reaction coordinates, can be computed. These observables play a crucial role in studying the structural and dynamical properties of various physical and biological systems. \cref{fig:mlff_md} summarizes the pipeline for using ML force fields (FFs) to simulate MD trajectories.

Obtaining the forces and energy for a given state requires classical or quantum mechanical calculations. While quantum mechanical calculation offers higher accuracy, it is computationally expensive. To accelerate MD simulation, one strategy is to fit a machine learning (ML) model that predicts $\bm F(\bm x)$ and $E (\bm x)$ from the atomic coordinates. These models, known as machine learning force fields, are trained to approximate atom-wise forces and energies using a training dataset: $\{\bm x_i, \bm F_i, E_i\}_{i=1}^{N_{\mathrm{data}}}$, where $ \bm x_i \in \R^{N \times 3}, \bm F_i \in \R^{N \times 3}, E_i \in \R$, $N_{\mathrm{data}}$ is the number of data points. The learned force field can then be used to simulate molecular dynamics by replacing the computationally expensive quantum mechanical calculations for obtaining energy and forces.

In addition to ML force fields, it is important to note that the primary goal of MD simulation is to extract macroscopic observables that characterize system properties. Due to the chaotic nature of molecular dynamics, it is neither practical nor necessary to recover the trajectories given the initial states exactly. Therefore, many approaches focus on augmenting existing force fields to achieve more efficient sampling or coarse-graining that aims at reducing the system's complexity. The design of sampling and coarse-graining methods is often influenced by the specific system/observable of interest.

\subsubsection{Technical Challenges}
First of all, the potential energy surfaces (PES) of molecular systems are often highly nonsmooth. Complex atomic interactions require expressive descriptors of the atomic environment. Ideally, the physical symmetry of energy ($E(3)$-invariant) and forces ($E(3)$-equivariant) should be respected. Expressive model architecture is a key technical problem in designing accurate ML force fields. The complex PES also poses technical challenges in effectively sampling diverse conformations, which motivates research in enhanced sampling methods.
Secondly, while the simulation time step is usually at the femtosecond level, the observable of interest can often be at a much longer time scale. Therefore, practically useful MD trajectories require simulations of millions to billions of steps to sample the dynamics. This practical need poses challenges to the efficiency, stability, and accuracy of the learned force fields. It is very hard to predict the performance of the learned force field in a simulation setting without actually running expensive simulations. Recent works have demonstrated that a lower force/energy prediction error does not imply a more stable and accurate simulation or observable calculation~\cite{fu2023forces}. The scale discrepancy between full-atom MD simulations and practical observables of interest also motivates research in coarse-graining approaches. 
In particular, the sampling of rare atomistic events is an important but difficult problem due to the combination of the two challenges stated above: the complex potential energy surface may have high energy barriers between different local minima, making rare events such as transitions between different metastable states hard to sample. Consequently, these transitions happen at a much longer time scale than the time scale a learned force field operates on. To summarize, the technical challenges of learning MD simulation root in the inherent complexity of the potential energy surface of atomistic systems and the computational complexity in calculating energy and forces for large spatiotemporal scales.

\subsubsection{Existing Methods}

ML force fields~\cite{behler2007generalized, khorshidi2016amp, smith2017ani, artrith2017efficient, chmiela2017machine, chmiela2018towards, zhang2018deep, zhang2018end, thomas2018tensor, jia2020pushing, gasteigerdirectional, schoenholz2020jax, noe2020machine, doerr2021torchmd, kovacs2021linear, satorras2021en, unke2021spookynet, park2021accurate, tholke2022equivariant, gasteiger2021gemnet, friederich2021machine, liu2022spherical, li2022graph, batzner20223, takamoto2022towards, musaelian2023learning} have attained incredible accuracy and data/compute efficiency that makes them promising for replacing quantum mechanical calculations in many applications. Different model architectures have been explored, including kernel-based methods, feed-forward neural networks, and message passing neural networks. These models are designed to respect the physical symmetry principle, including the $E(3)$-invariance of energy and $E(3)$-equivariance of forces. Much of the molecular representation learning research has been motivated by MD applications. They have been covered with more details in \cref{subsec:mol_representation_learning}.

In an effort to enhance the sampling process, machine learning (ML) methodologies have been employed to uncover crucial reaction coordinates~\cite{sidky2020machine, mehdi2023enhanced} (also known as collective variables). These serve as prerequisites for implementing specific advanced sampling techniques, such as Meta Dynamics~\cite{laio2002escaping, barducci2008well}. Identifying reaction coordinates can also play a significant role in elucidating the Molecular Dynamics (MD) process, notably in fitting a Markov state model for studying the transitions occurring between metastable states in protein molecules~\cite{mardt2018vampnets}. Moreover, ML techniques are being harnessed for learning coarse-grained force fields~\cite{husic2020coarse, wang2019machine}, coarse-grained and latent-space simulators~\cite{fu2022simulate, vlachas2021accelerated, sidky2020molecular}, and coarse-graining mapping~\cite{wang2019coarse, wang2022generative, kohler2023flow}. Simulation in the coarse-grained space is usually much more efficient but involves trade-offs over accuracy.
Learning coarse-grained mapping encompasses the discovery of a coarse-graining scheme capable of preserving the essential information of the molecular state, as well as facilitating coarse-graining back mapping (predicting the distribution of fine-grained states corresponding to a coarse-grained state).

It's important to acknowledge that the research areas mentioned above possess extensive histories in their respective fields and continue to be vigorously developed. The materials and references explored here provide just a very preliminary overview. For a more in-depth understanding, we direct interested readers towards more exhaustive surveys on these topics~\cite{sidky2020machine, unke2021machine, noid2023perspective}.

\subsubsection{Datasets and Benchmarks}
Small molecules~\cite{chmiela2017machine, chmiela2023accurate, eastman2023spice} have been a popular testbed for ML force field development. The widely used MD17 dataset~\cite{chmiela2017machine} contains MD data for eight small molecules generated from path-integral molecular dynamics simulations, with updated versions MD17@CCSD(T)~\cite{chmiela2018towards} and rMD17~\cite{christensen2020role} that use higher levels of theory and are more accurate. 
Other systems of interest include bulk water~\cite{zhang2018deep}, various crystalline solid materials (\emph{e.g.}, Li-ion electrolytes~\cite{batzner20223}), and amorphous materials (\emph{e.g.}, polymer~\cite{fu2022simulate}). With these datasets, force and energy prediction error over a test dataset is a common benchmarking strategy in existing work. Some papers~\cite{stocker2022robust, zhang2018deep, batzner20223} also study the stability of simulation and certain observables such as the distribution of interatomic distances, radial distribution function, diffusion coefficient, and so on. In particular, a recent benchmark study~\cite{fu2023forces} compared a series of existing ML force fields over a wide range of systems and tasks and found a misalignment between force/energy prediction performance and simulation performance, which shows the inefficacy of using force and energy prediction as the sole evaluation protocol. 

For biomolecules, one focus of existing studies is to recover their free energy surface (FES) with respect to key reaction coordinates. Alanine dipeptide~\cite{noe2020machine} is a standard benchmarking molecule due to its well-understood reaction coordinates and FES: there are two main conformational degrees of freedom: dihedral angle $\phi$ of $\mathrm{C-N-C_{\alpha}-C}$ and dihedral angle $\psi$ of $\mathrm{N-C_{\alpha}-C-N}$, with six FES minima over these two reaction coordinates. It has been studied in many papers focusing on sampling the Boltzmann distribution~\cite{fu2023forces}, transition path sampling~\cite{holdijk2022path}, and coarse-grained MD studies~\cite{wang2019machine, vlachas2021accelerated, greener2021differentiable}. More complex biomolecules, such as the small protein Chignolin have also been studied in existing works~\cite{husic2020coarse, wang2022generative}.
For materials, past works have studied Li-ion battery electrolytes, such as LiPS~\cite{batzner20223} and solid polymer electrolytes~\cite{fu2022simulate}, while looking at the radial distribution function and Li-ion diffusivity as the key observables. Finally, we note that MD simulation is a broad area with diverse applications and datasets available. We are only covering some of the most popular ones that were studied with ML methods. 

\subsubsection{Open Research Directions}

The precision and efficacy of current machine learning (ML) force fields present ample opportunities for further refinement. Predominantly, existing methodologies are built upon kernel-based or message passing schemes across a graph formed with a predetermined radius cutoff. A promising avenue to enhance ML force fields' proficiency involves accurately and efficiently capturing long-range interactions. Given that the primary motivation behind ML force fields is to expedite Molecular Dynamics (MD) simulations, designing neural architectures that prioritize computational efficiency and parallelizability without compromising accuracy is critical. For instance, existing works have explored strictly local ML potentials~\cite{musaelian2023learning}. In practice, simulation instability is a recurrent issue when applying ML force fields. Active learning strategies~\cite{vandermause2020fly, ang2021active} that focus on gathering new data from states where the learned model underperforms can help address these instability concerns and decrease the number of ground truth calculations required for training a dependable ML force field.
As a crucial avenue for extending MD simulations to broader spatial and temporal scales, the implementation of coarse-graining methods in both space (by converting atoms into coarse-grained beads) and time (by employing larger time steps, such as through learning time-integrated dynamics) are of great importance. Future research should strive to further comprehend and characterize the information retained and forfeited in a coarse-graining (CG) scheme. It's also imperative to explore ways to create more effective CG schemes that retain the maximum amount of information within a specified computational budget.
Lastly, the endeavor to learn to sample rare events constitutes another vibrant area of research. In this domain, various methods such as ML-based collective variable discovery~\cite{sidky2020machine}, transition path sampling~\cite{holdijk2022path}, and deep generative models for modeling the Boltzmann distribution~\cite{noe2019boltzmann} represent promising directions to advance this research theme.

\subsection{Learning Stereoisomerism and Conformational Flexibility}

\noindent{\emph{Authors: Keir Adams, Connor W. Coley}\vspace{0.3cm}\newline\emph{Recommended Prerequisites: Section~\ref{subsec:mol_representation_learning}}}\newline

One potential advantage of 3D graph neural networks (GNNs) over their 2D counterparts is their capacity to natively model the structural differences between stereoisomers, molecules that share the same 2D molecular graph but have differing spatial arrangements of atoms in 3D. Stereoisomerism is commonly induced by tetrahedral chiral centers (\emph{e.g.}, carbon atoms with four non-equivalent bonded neighbors); double bonds with different E/Z (cis/trans) configurations; and chiral axes of atropisomers, allenes, or other helical molecules (Figure  \ref{fig:stereochemistry}) \citep{eliel1994stereochemistry}. 
Notably, a molecular graph may have many different stereoisomers; a molecule with $N$ tetrahedral chiral centers can have up to $2^N$ stereoisomers, without even considering E/Z isomerism or axial chirality. Two stereoisomers can be classified as either diastereomers, or enantiomers. Enantiomers are mirror-image chiral molecules that cannot be superimposed via thermodynamically-permissible conformational changes (\emph{e.g.}, rotations about chemical bonds). Diastereomers generally have distinct chemical properties altogether, while enantiomers exhibit identical physicochemical properties in many situations unless interacting with other chiral molecules (such as proteins), in which case they may exhibit wildly different properties \citep{mcconathy2003stereochemistry, chhabra2013review}.
Hence, the ability of graph neural networks to learn the subtle influence of stereoisomerism is crucial for practical applications across domains ranging from medicinal chemistry to chemical catalysis. 
Stereoisomerism has been overlooked as an aspect of molecular identity because the majority of benchmarks used for molecular property prediction do not require careful treatment given their high aleatoric uncertainty and underrepresentation of stereoisomers in the datasets.

\begin{figure}[t]
    \centering
    \includegraphics[width=0.8\textwidth]{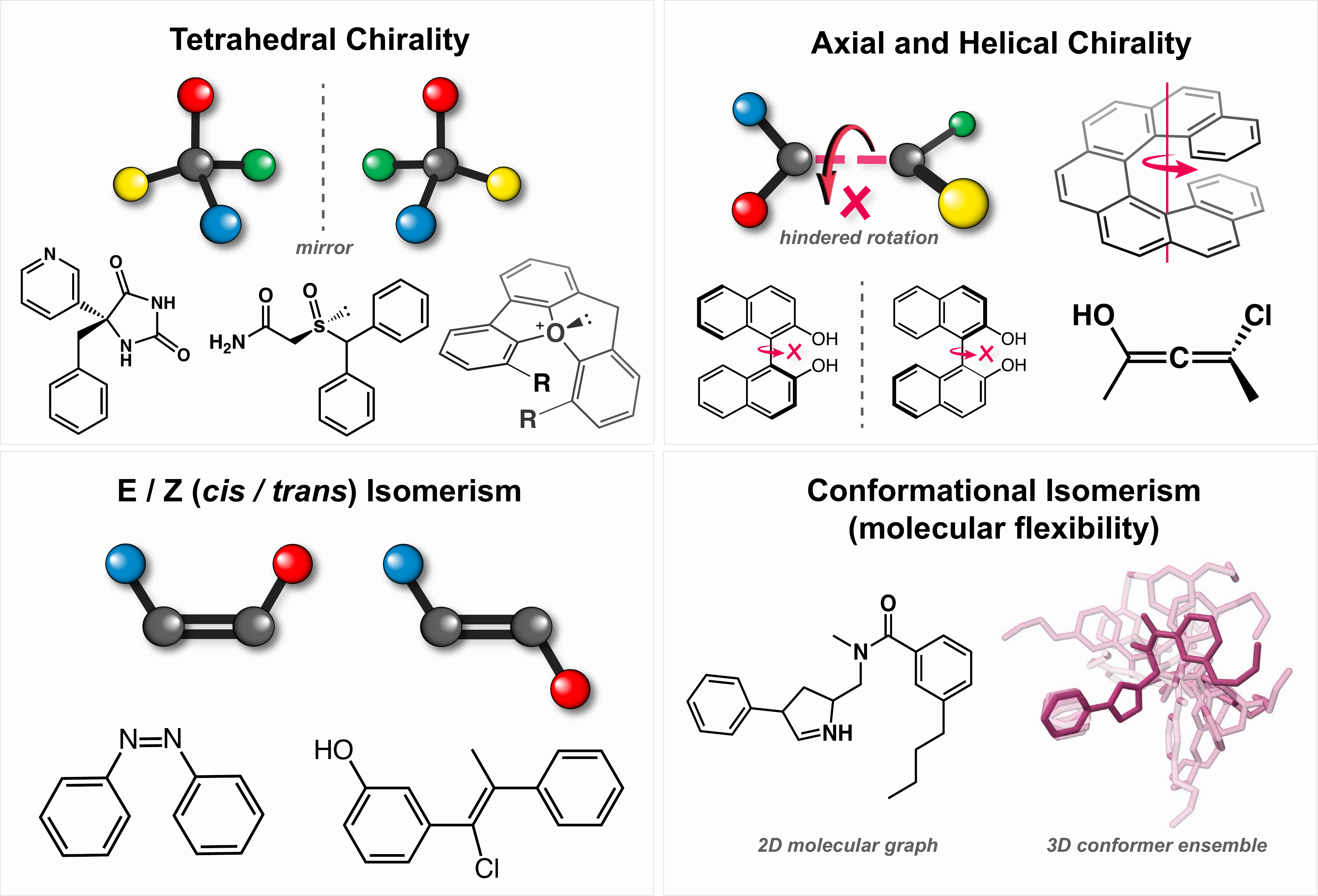}
    \caption{Stereochemistry is an important yet often overlooked aspect of molecular identity that describes the differing orientations or arrangements of atoms in 3D space for molecules which share a common 2D graph topology. Stereoisomerism can be caused by a variety of local structural characteristics; shown here are common forms of stereochemistry that are particularly relevant for medicinal chemistry, chemical catalysis, and organic chemistry. \textit{Tetrehedral chirality} (or point chirality) describes the differing orientations of four non-equivalent chemical groups around a stereogenic center. Tetrahedral chiral centers are inverted upon reflection, and hence induce enantiomerism. Tetrahedral chirality is often associated with carbon atoms, but can arise elsewhere, such as in sulfinyl or oxonium compounds. \textit{Axial chirality} is caused by the (non-planar) orientations of four chemical substituents around a chiral axis. Axial chirality is found in allenes due to the propeller-like arrangement of adjoining double ($\pi$) bonds, and commonly in atropisomers, where bulky substituents restrict rotation about a single ($\sigma$) bond. Helical chirality is also a form of axial chirality, although its structural origin is different. Like tetrahedral chirality, axial chirality leads to enantiomerism. \textit{E/Z isomerism} is caused by the differing \textit{cis} or \textit{trans} configurations of (planar) double bonds. Unlike tetrahedral or axial chirality, E/Z isomerism does not produce pairs of enantiomers. These forms of stereochemistry yield stereoisomers that typically cannot be interconverted on practical timescales without undergoing chemical reactions. Any given stereoisomer can also have a distribution of structurally-distinct but rapidly-interconverting conformational isomers, or \textit{conformers}, owing to the molecule's 3D flexibility. Observable molecular properties are often related to thermodynamic averages over the entire conformer ensemble.}
    \label{fig:stereochemistry}
\end{figure}

Conformational isomerism is yet another form of stereochemistry which describes how a single molecule can adopt many different low-lying structures on the potential energy surface (PES), collectively called the conformer ensemble 
\citep{wolf2007dynamic, eliel1994stereochemistry}. Section \ref{subsec:mol_representation_learning} described representation learning on static and previously-known 3D molecular structures, such as DFT-optimized ground-state molecular geometries from the QM9 dataset \citep{ramakrishnan2014quantum}. In reality, molecules are not static structures, but are instead constantly interconverting between different conformations through intramolecular motions such as chemical bond rotations and smaller vibrational perturbations. 
The energetic penalty for these motions is environment-dependent (\emph{e.g.}, solvent-dependent), and the rate of interconversion between conformers is highly temperature-dependent. For instance, at room temperature, cyclohexane undergoes a chair flip (10 kcal/mol energetic barrier) with a characteristic period of microseconds \citep{hendrickson1961molecular}, whereas a bulky biaryl system like 1,1'-Binaphthyl interconverts between its (R)- and (S)- atropisomers (23 kcal/mol energetic barrier) on a time scale of hours \citep{meca2003racemization}.
Colloquially, two conformers would be considered to belong to distinct ``stereoisomers'' if they do not thermally interconvert on a practical timescale (\emph{e.g.}, cannot be isolated at room temperature); two conformers that are mutually accessible would be described as corresponding to the same ``stereoisomer''.

Many experimentally observable chemical properties depend on the full distribution of thermo-dynamically accessible conformers. On the other hand, some may depend on a particular (higher-energy) geometry that is not known \textit{a priori}, such as the active binding pose of a ligand. The PES can also be substantially altered by intermolecular interactions (\emph{e.g.}, with solvent molecules), making it challenging to identify \textit{a prior} which molecular structure(s) significantly contribute to observable properties without performing prohibitively expensive simulations. Although 3D GNNs have primarily been developed to encode individual 3D structures (Section~\ref{subsec:mol_representation_learning}), recent works have attempted to represent conformational flexibility by explicitly encoding conformer ensembles \citep{axelrod2020molecular, chuang2020attention}. This may be impactful for predicting distribution-dependent molecular properties such as Boltzmann-averaged ligand-protein binding affinities \citep{miller1997ligand, gilson2007calculation}; chemical reaction rates and selectivities \citep{hansen2016prediction, guan2018aaron}; and entropic contributions to free energies \citep{mezei1986free, chen2004calculation}.

\subsubsection{Problem Setup}

For a given 2D molecular graph $\mathcal{G} = (\bm{z}, E)$, where $\bm{z}$ is the vector of atom types (\emph{e.g.}, atomic numbers) and $E$ denotes the graph adjacency matrix, we can formally describe its thermodynamically-accessible conformer ensemble as a set $\mathcal{C}_\mathcal{G} = \{C_i\}_{i=1}^{|\mathcal{C}_\mathcal{G}|}$ of structurally-distinct 3D molecular geometries $C_i \in \mathbb{R}^{3 \times n}$, each annotated with a (free) energy. 
Although the conformer ensemble is actually a continuous distribution, it is common to describe it with a discrete set of conformers by imposing a sub-Angstrom minimum root mean square distance (RMSD) threshold between any $C_i$ and $C_j$.
$\mathcal{C}_\mathcal{G}$ can be divided into $S$ disjoint subsets corresponding to the distribution of conformers available to each stereoisomer $\mathcal{C}_\mathcal{G}^s = \{C^s_k\}_{k = 1}^{|\mathcal{C}_\mathcal{G}^s|}$ of the molecular graph so that $\mathcal{C}_\mathcal{G} = \mathcal{C}_\mathcal{G}^1 \cup \mathcal{C}_\mathcal{G}^2 \cup ... \cup \mathcal{C}_\mathcal{G}^S$. 
The decision of which conformers belong to disjoint subsets can be somewhat subjective, but is typically based on their ability to interconvert on whatever timescale is relevant to the application at hand. 
Each conformer in a distribution can be assigned a statistical (Boltzmann) weight $p_{C^s_i} = \exp(\frac{- e_i} {k_B T}) / \sum_j \exp(\frac{- e_j}{k_B T})$ corresponding to its expected presence under experimental conditions, where $e_i$ is the (free) energy of conformer $C^s_i$, $k_B$ is the Boltzmann-constant, and $T$ is the temperature. Some example stereochemical representation learning tasks include classifying a given conformer as one of many stereoisomers, contrasting the learned representations of conformers belonging to different stereoisomers, or training a supervised model to predict the properties of non-interconvertible stereoisomers from sampled conformers. This final supervised learning task aims to learn $\hat{f}(C^s_k \in \mathcal{C}_\mathcal{G}^s; \bm{\theta}) \approx f(\mathcal{C}_\mathcal{G}^s)$, where $\hat{f}$ is a neural network with weights $\bm{\theta}$. Tasks related to learning on conformer ensembles include predicting Boltzmann-averaged properties $\langle y\rangle_{k_B} = \sum_i p_{C_i} f({C_i})$ from a small subset of the full conformer ensemble $\hat{f}(\{C^s_k\}_{k=1}^{K \ll |\mathcal{C}_\mathcal{G}^s|}; \bm{\theta}) \approx \langle y\rangle_{k_B}$, where $f({C_i})$ is a per-conformer property, or identifying a property-active conformer amongst a set of (non-active) decoy conformers.

\subsubsection{Technical Challenges}

Learning molecular stereochemistry and conformational flexibility presents multifaceted modeling challenges. Because stereoisomers have the same molecular graph, 2D GNNs are inherently limited in their ability to distinguish stereoisomers with different chemical properties. Often, practitioners augment a molecular graph with simple atom (node) or bond (edge) features that store stereochemical information such as the handedness of chiral centers or the configuration of double bonds \citep{yang2019analyzing}. However, commonly used features (such as R/S chiral atom tags) are global properties that do not act in accordance with local graph convolutions \citep{pattanaik2020message}, have restricted representation learning power \citep{adams2021learning}, and do not account for all forms of molecular chirality. 3D GNNs can also be limited in their ability to express certain stereochemistries according to their symmetry properties. For instance, many mathematically simple 3D GNNs with $E(3)$-invariant features cannot distinguish the mirror-image structures of enantiomers. As a result, more complex networks with either 4-body interactions or equivariant features are often needed to robustly express chirality from 3D molecular structures \citep{liu2022spherical, gasteiger2021gemnet, thomas2018tensor}. Further, adequately predicting properties of stereoisomers $f(\mathcal{C}^s_{\mathcal{G}})$ from a single 3D structure $C_k^s$ requires the neural network to learn an invariance over 3D conformations to avoid confusing which conformers belong to which stereoisomer \citep{adams2021learning}. Meanwhile, simultaneously encoding multiple conformers (at least) linearly scales the computational cost of training/inference while also making network optimization significantly more challenging \citep{axelrod2020molecular}.

Modeling conformer ensembles and molecular flexibility raises additional challenges associated with the cost of obtaining high-quality conformer ensembles, especially at inference time.  Namely, if a prediction model is trained with conformers obtained from expensive quantum chemical or molecular dynamics simulations, then the same simulations are likely required at inference time in order to avoid a domain shift reducing model accuracy. On the other hand, it may be difficult to accurately predict structure-sensitive properties of high-quality conformers when only encoding cheap conformers that are not faithful representations of the ground-truth conformers. For instance, it has been observed that encoding conformers optimized with molecular mechanics force fields to predict ground-state quantum properties of DFT-optimized molecules can lead to substantial loss in model accuracy compared to the case where ground-truth conformers are used directly \citep{stark20223d, pinheiro2022impact}. Similarly, it may be challenging to accurately predict properties of \textit{unknown} property-active conformers when only modeling non-active conformers that are randomly sampled from the ensemble (\emph{e.g.}, predicting protein-ligand binding affinity without knowing the relevant ligand poses \emph{a priori}). Although using a 2D GNN may side-step the challenges of obtaining quality conformers, 2D GNNs often cannot adequately learn functions that are highly sensitive to molecular geometry.

There are also challenges associated with collecting high-quality datasets for benchmarking and model development. When developing models to predict distribution-dependent properties obtained from simulated conformers, it is crucial that exhaustive conformer simulations are performed at a sufficiently high level of theory in order to avoid missing important (low-energy or property-active) conformers or assigning undue statistical weight to unrealistic geometries. These conformer searches should ideally be performed in a setting that reflects physical conditions, such as considering the influence of solvent molecules on the PES. Additionally, developing new models for stereochemical representation learning is often impeded by a lack of high-quality datasets that simultaneously 1) include properties of multiple stereoisomers for each molecular graph, 2) include properties that are sensitive to molecular stereochemistry, and 3) include properties with high signal-to-noise ratios. 

\subsubsection{Existing Methods}

\vspace{0.1cm}\textbf{Representing Molecular Stereochemistry}: 
Existing works have chiefly focused on encoding tetrahedral chirality, and occasionally E/Z (or cis/trans) isomerism, through special tokens in molecular SMILES strings or through atom (node) and bond (edge) attributes in the 2D molecular graph \citep{yang2019analyzing}.
SMILES strings can natively store the handedness of tetrahedral chiral centers via `@' and `@@' tokens that indicate whether the ordering of neighboring atoms (as provided in the string) is clockwise (CW) or counterclockwise (CCW). `$\backslash$' and `$\slash$' tokens similarly store the configuration of double bonds. Hence, sequence encoders like transformers or recurrent neural networks can in principle express these forms of stereochemistry. Any 2D graph neural network that uses node and edge attributes may similarly include binary one-hot features that store the local configurations around tetrahedral centers or double bonds, based on a specified ordering of atoms. A related strategy creates graph convolution kernels that use the sign of the tetrahedral volume under a specific atom ordering \citep{liu2022interpretable}. However, using atom or bond orderings that are sensitive to arbitrary bookkeeping breaks the permutation invariance of graph neural networks.
Instead of relying on local atom orderings, the absolute R/S handedness of tetrahedral chiral centers and E/Z configuration of double bonds can instead be encoded via heuristic Cahn-Ingold-Prelog (CIP) rules that specify the global priority ranking of neighboring atoms \citep{cahn1966specification}. However, small edits to a molecular graph (\emph{e.g.}, replacing a carbon atom with a silicon atom) can flip the parity of these global atom/bond tags without substantially changing the 3D molecular geometry (Figure \ref{fig:chiral_tags}), potentially making the learned stereochemical representations non-smooth.

\begin{figure}[t]
    \centering
    \includegraphics[width=0.4\textwidth]{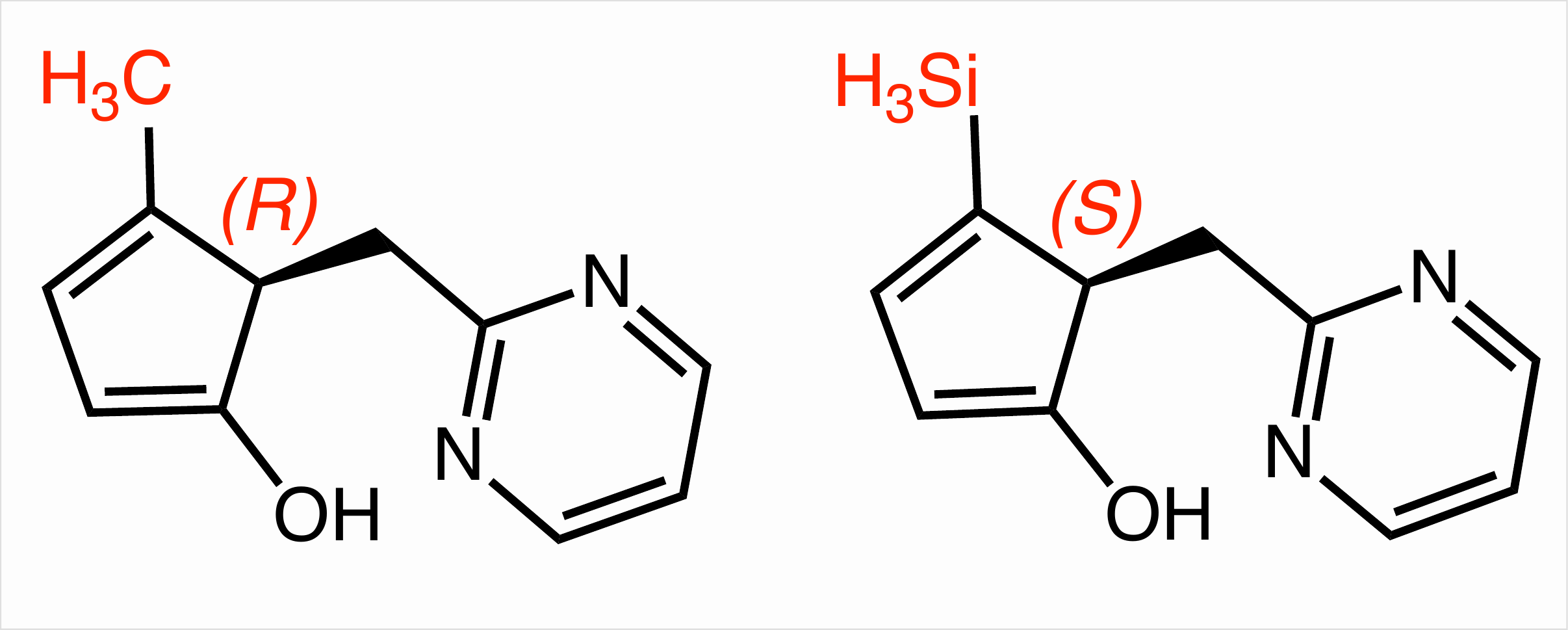}
    \caption{Global (R/S) chiral atom tags do not conserve local 3D geometric features. Small changes to the molecular graph, such as swapping a carbon atom for a silicon atom, can flip the handedness of a tetrahedral chiral centers (according to CIP rules) without affecting the 3D molecular geometry or chemical reactivity.}
    \label{fig:chiral_tags}
\end{figure}

\citet{pattanaik2020message} and \citet{adams2021learning} introduce alternative methods of encoding tetrahedral chirality in graph neural networks without using heuristic rules or breaking permutation invariance. In each message passing layer of the GNN, \citet{pattanaik2020message} alters the sum-pooling operation (Equation~(\ref{eq:node_centered_message_passing})) to instead enumerate and encode all 12 permutations of the four neighboring atoms around each chiral center, thereby learning 2D atom representations that are sensitive to tetrahedral chirality. \citet{adams2021learning} introduces ChIRo, which learns 3D representations of tetrahedral chirality by specially encoding the dihedral/torsion angles of each internal chemical bond. Although a conformer is required as input to the model, ChIRo is natively invariant to conformational changes caused by bond rotations, and using a cheap conformation generated by RDKit is shown to be sufficient. These approaches are empirically demonstrated to be superior to solely using local/global chiral atom tags when predicting chiral-dependent properties, at the expense of being computationally less efficient.

$SE(3)$-invariant 3D GNNs, such as SphereNet \citep{liu2022spherical}, can also learn representations that are sensitive to tetrahedral chirality. In this case, it is important that multiple conformers per molecule are used as training-time data augmentation in order to force the networks to learn an approximate conformer invariance \citep{adams2021learning}. However, in addition to being computationally demanding and requiring many conformers to train, 3D GNNs currently underperform in chiral property prediction tasks on small molecules compared to simply using 2D GNNs with chiral atom tags.

\vspace{0.1cm}\noindent \textbf{Representation Learning on Conformer Ensembles}:
The most common strategy for learning on conformer ensembles is to formulate the task as a multiple-instance learning (MIL) problem \citep{dietterich1997solving, maron1997framework, ilse2018attention}. In this setting, multiple conformer instances of a given molecule are individually encoded into a set of fixed-length embeddings and then pooled or otherwise aggregated to obtain a single embedding of the entire conformer ensemble \citep{axelrod2020molecular, chuang2020attention}. Formally, the ensemble embedding can be represented as
\begin{equation}
    \begin{aligned}
    \bm{h}_k &= \hat{f}(C_k; \bm{\theta}), \\
    \bm{h}_{\mathcal{C}_\mathcal{G}} &= \sum_{k=1}^{K} a_k \bm{h}_k.
    \end{aligned}
\label{eq:conformer_ensemble_encoding}
\end{equation}
Here, $\hat{f}(C_k; \bm{\theta})$ is any machine learning model that learns an embedding $\bm{h}_k$ of a conformer $C_k$, such as a 3D GNN. For instance, \citet{axelrod2020molecular} employs SchNet \citep{schutt2018schnet}, augmented with additional chemical features, as an underlying 3D GNN. Many other works employing MIL for molecular machine learning tasks have used non-neural models or hand-crafted 3D feature vectors to encode $\bm{h}_k$ \citep{zahrt2019prediction, zankov2021qsar, weinreich2021machine}.
$a_k$ can be set to a constant in order to weight each encoded conformer equally in the ensemble-level representation $\bm{h}_{\mathcal{C}_\mathcal{G}}$, as in sum- or mean-pooling. Alternatively, $a_k$ can be learned attention coefficients that assign relative importance to each encoded conformer, which may be used to identify key conformer instances in the ensemble without needing to predict instance-level labels \citep{chuang2020attention}. Another approach uses max pooling to aggregate single-instance conformer representations \citep{liu2021fast}.
Because each conformer instance may be in a random reference frame, typically only $l=0$ invariant representations are aggregated ($\bm{h}_k= \bm{h}_k^{l=0} $).

To avoid the cost of sampling or encoding multiple conformers at inference time, other works have attempted to implicitly model conformational flexibility by encoding a single ``effective" structure obtained by averaging multiple conformers in structure-space \citep{weinreich2022ab}, by learning conformer invariance via conformer-based data augmentation during training \citep{adams2021learning}, or by considering multiple conformations during the collection of training labels \citep{suriana2023flexvdw}. On the other hand, not explicitly encoding individual conformers from the ensemble may preclude the model from identifying key conformer instances, or otherwise reduce the model's sensitivity to important 3D structures.

\subsubsection{Datasets and Benchmarks}

\vspace{0.1cm}\textbf{Representing Molecular Stereochemistry}:
Enantiomers share common physio-chemical properties such as dipole moments or HOMO-LUMO gaps, meaning that popular benchmarks developed for 3D representation learning, like QM9 \citep{ramakrishnan2014quantum} or PubChemQC \cite{nakata2017pubchemqc}, cannot be used to evaluate the ability of models to learn chiral-sensitive representations even if the molecules in these datasets have well-defined chirality. On the other hand, biological measurements such as protein-ligand binding affinity or toxicity measurements can be influenced by chirality as well as other forms of stereochemistry. However, experimental datasets such as those contained in MoleculeNet \citep{wu2018moleculenet} typically do not contain measurements for multiple stereoisomers of the same molecule, preventing straightforward evaluations of whether models learn the effects of stereochemistry. Further, the subtle effects of chirality may be obfuscated by experimental noise. As a result, existing works benchmark models on simulated datasets that have been specially curated to display acute sensitivity to molecular stereochemistry, particularly tetrahedral chirality.

\citet{pattanaik2020message} filters the D4 dopamine
receptor protein-ligand docking screen performed by \citet{lyu2019ultra} to curate a dataset of 287,468 drug-like molecules with a Bemis-Murcko 1,3-dicyclohexylpropane scaffold that have at least one tetrahedral chiral center. Each molecule is also constrained to have a pair of enantiomers or diastereomers present in the dataset. They further subdivide this dataset into a subset containing enantiomer pairs with differences in docking scores above a threshold, and another subset containing enantiomer pairs with only one chiral center.  \citet{adams2021learning} also uses receptor-ligand docking to evaluate chirality-aware models. To control for stochasticity in docking simulations, they dock enantiomers from PubChem3D \citep{bolton2011pubchem3d} with low molecule weight and few rotatable bonds to a small docking box (PDB-ID: 4JBV), and only retain 34,560 pairs of enantiomers with differences in docking scores above a statistically significant threshold. In both works, models are evaluated based on their capability to correctly rank-order stereoisomer pairs by their predicted docking scores.

Beyond benchmarking on simulated datasets, \citet{adams2021learning} and \citet{mamede2020machine} curate datasets containing experimentally-measured optical activity (``L" \textit{versus} ``D" classifications) for one-chiral center enantiomers, sourced from the Reaxys database \citep{lawson2014making}. ``L" \textit{versus} ``D" classification is especially interesting as a benchmarking task because optical activity is difficult to simulate without expensive \textit{ab initio} calculations, L/D labels have no correlation to R/S chiral tags, and the optical rotation for one enantiomer can be directly inferred if its value is known for the other enantiomer. Moreover, predicting optical activity is practically useful for assigning the absolute configurations of chiral molecules. We envision that the collection and public dissemination of similar datasets containing chirality-sensitive properties of interest will be instrumental in furthering the field of chiral molecular representation learning.

\vspace{0.1cm}\noindent\textbf{Representation Learning on Conformer Ensembles}:
Few datasets have been developed to benchmark deep learning models on predicting the properties of conformer ensembles due to the resources required to obtain high-quality conformer ensembles and their associated properties for a large library of molecular compounds. \citet{axelrod2020molecular} has benchmarked a handful of 2D, 3D, and 3D-ensemble models on their ability to classify bioactive hits from a library of 278,758 drug-like molecules from the GEOM-DRUGS dataset \citep{axelrod2022geom}, each annotated with experimental inhibition data against the SARS-CoV 3CL protease. Each molecule contains numerous conformers generated with the CREST program \citep{pracht2020automated}, which are used to build models that encode ensembles each containing up to 200 conformers. On this task, models that explicitly encode multiple conformers in a multi-instance ensemble do not outperform baseline models that only encode a single conformation. The MIL models also require orders of magnitude more resources to train. It is important to note that this experimental dataset is very unbalanced, containing just 426 bioactive hits, which may complicate model optimization. \citet{chuang2020attention} introduces a small synthetic dataset containing 1157 biaryl ligands with at most one rotatable bond, each containing an average of 13.8 conformers generated with OMEGA \citep{hawkins2010conformer}. They evaluate the ability of a multi-instance attention model to identify whether an encoded ensemble contains a key conformer instance with a specific bidentate coordination geometry. On this toy task, however, simple random forest baselines using ECFP4 molecular fingerprints outperformed the MIL models.

Recent works have created new datasets containing large conformer ensembles that could potentially be used for conformer ensemble learning. In particular, \citet{grambow2023cremp} introduces the CREMP dataset, consisting conformer ensembles for 36,198 macrocyclic peptides generated with CREST. \citet{siebenmorgen2023misato} introduces MISATO, a dataset containing 10-ns molecular dynamics traces for 16,972 protein-ligand complexes in explict water solvent.

\subsubsection{Open Research Directions}
Designing the next generation of geometric models to better represent molecular stereochemistry chiefly requires the development of new benchmarking datasets that contain multiple stereoisomers per molecule, labeled with properties that are sensitive to stereochemistry. In real-world applications, however, it may be impractical or infeasible to collect data on multiple stereoisomers due to experimental or computational budgets. Hence, another promising direction is to design models or training strategies that can learn stereochemistry-sensitive representations without needing to be exposed to multiple stereoisomers for each molecular graph during training time. Additionally, while current methods have focused solely on encoding tetrahedral chirality, future work could explore how to represent other forms of medicinally-relevant stereochemistry, such as atropisomerism.

Based on the limited number of existing studies, 3D conformer ensemble models are currently unable to outperform traditional 3D models that only encode a single conformer, despite the rich structural information contained in a conformer ensemble. Future work could investigate if this is due to the use of out-of-date 3D model architectures in existing studies, the inadequacy of the MIL framework to capture conformer flexibility, or the inherent difficulties of optimizing models that simultaneously encode structurally diverse ensembles, among other possible factors. The computational burden of both generating and encoding multiple conformers during training and inference also presents a practical barrier to the widespread adoption of MIL models for encoding conformer ensembles. New strategies to efficiently account for conformational flexibility should be explored. Finally, because simulating high-quality conformer ensembles in physically realistic environments is often impractical for large virtual screens at inference time, future work could investigate the transferability of models that instead encode readily-available conformer ensembles obtained from inexpensive algorithms or generative models.

\clearpage
\hypertarget{AI for Protein Science}{\section{AI for Protein
Science}} \label{sec:prot}

{\ifshowname\textcolor{red}{Keqiang}\else\fi}Proteins consist of a chain of amino acids at the primary level. They can fold into 3D structures to perform many biological functions. Recent breakthroughs in graph neural networks, diffusion models, and 3D geometric modeling enable the use of machine learning to boost the discovery of novel proteins. In this section, we focus on three AI for protein science topics, including protein folding (protein structure prediction) in~\cref{sec:protein_folding}, protein representation learning in~\cref{sec_prot_pred}, protein backbone structure generation in \cref{sec_prot_gen}. \revisionOne{In this section, we only discuss individual protein molecules. For protein and small molecule interactions, such as binding prediction and structure-based drug design, can be found in~\cref{sec:openmi_binding} and~\cref{sec:sbdd}. Another area of research not covered in this survey is protein inverse folding, which aims to predict protein's amino acid sequences based on given protein structures. This is critical in drug discovery and synthetic biology by allowing researchers to design proteins with therapeutic properties. More details can be found in the recent benchmark~\citep{gao2024proteininvbench}.}

\subsection{Overview}

\noindent{\emph{Authors: Keqiang Yan, Limei Wang, Cong Fu, Tianfan Fu, Yi Liu, Jimeng Sun, Shuiwang Ji}}\newline

{\ifshowname\textcolor{red}{Keqiang Yan}\else\fi}We first 
give formal definitions of different levels of protein structures, and then introduce protein folding, protein representation learning, and protein backbone generation tasks. After that, we introduce geometric constraints that need to be considered for the latter two tasks. An overview of this section is shown in Figure~\ref{fig:prot_overview}.

We first describe the four-level structure of a protein.  
(1) Amino acid is the basic building block of protein. 
Proteins consist of chains of amino acids, with each amino acid containing nitrogen ($N$), alpha-carbon ($C_{\alpha}$), carbon ($C$), and oxygen ($O$) atoms, as well as atoms in the side chain (known as R-group). The side chain determines the amino acid category. 
The amino acid chain is also known as the protein's primary structure; 
(2) based on the primary structure, secondary structures are locally folded structures that form based on interactions within the protein backbone; (3) tertiary structure is the three-dimensional structure of a single polypeptide chain (polypeptide chain refers to a string of amino acids connected together by peptide bonds); (4) quaternary structure describes the association between multiple polypeptide chains. They characterize the structure of a protein at different levels of complexity. In this work, we mainly focus on primary and tertiary structures.

\vspace{0.1cm}\noindent\textbf{Notations of Protein Structures:\ifshowname
\textcolor{red}{Limei, Keqiang}\else\fi}
Formally, a full-atom level protein structure can be represented as 
\begin{equation}
\label{eqn:protein_structure}
    \mathcal{P}_{\text{full}} = (\bm{z},\mathcal{C}).
\end{equation}
Here $\bm{z}=[z_1,...,z_n]\in\mathbb{Z}^n$ is the amino acid type vector, where each $z_i$ denotes the type of the $i$-th amino acid and $n$ denotes the number of amino acids in the protein. There are 20 commonly occurring amino acids that are used to build proteins in living organisms. The amino acid chain represents the primary structure of the protein and is folded into a 3D structure, where $\mathcal{C} = [\mathcal{C}_1, ..., \mathcal{C}_n]\in \mathbb{R}^{n \times k \times 3}$ denotes the coordinate matrix of the protein. Note that distinct from other sections, we use $\mathcal{C}$ to denote the coordinate matrix in this section in order to avoid notation conflict with carbon atom $C$. Each $\mathcal{C}_i$ includes the coordinates of all atoms in the amino acid $i$, including $N, C_{\alpha}, C, O$, and side chain atoms. $k$ is the maximum number of atoms in each amino acid. If we only consider the $C_{\alpha}$ atom in each amino acid, a protein structure can be represented as 
\begin{equation}
    \mathcal{P}_{\text{base}} = (\bm{z},\mathcal{C}^{C_{\alpha}}),
\end{equation}
where $\mathcal{C}^{C_{\alpha}}=[\bm{c}_1^{C_{\alpha}},...,\bm{c}_n^{C_{\alpha}}]\in\mathbb{R}^{3\times n}$ denotes the coordinate matrix of $C_{\alpha}$ atoms. Similarly, a protein backbone structure can be represented as
\begin{equation}
    \mathcal{P}_{\text{bb}} = (\bm{z},\mathcal{C}^{C_{\alpha}}, \mathcal{C}^{N}, \mathcal{C}^{C}),
\end{equation}
where $\mathcal{C}^{C_{\alpha}}=[\bm{c}_1^{C_{\alpha}},...,\bm{c}_n^{C_{\alpha}}]\in\mathbb{R}^{3\times n}$, $\mathcal{C}^{N}=[\bm{c}_1^{N},...,\bm{c}_n^{N}]\in\mathbb{R}^{3\times n}$, and $\mathcal{C}^{C}=[\bm{c}_1^{C},...,\bm{c}_n^{C}]\in\mathbb{R}^{3\times n}$ denote coordinate matrices of $C_{\alpha}, N, C$ atoms. In the following parts, $\mathcal{P}$ is used to denote a general representation of protein structures.

\begin{figure}[t]
    \centering
\includegraphics[width=\textwidth]{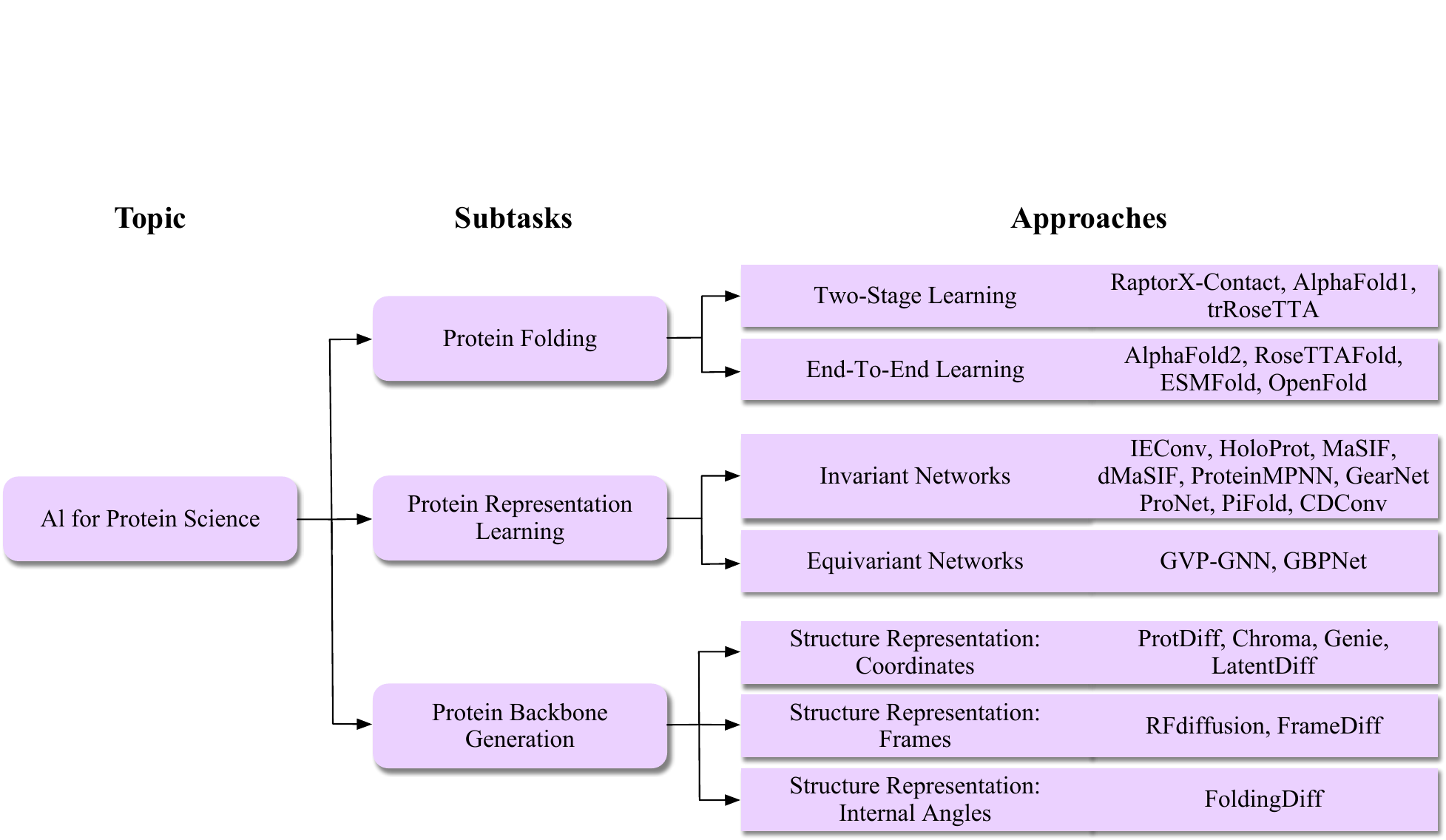}
    \caption{An overview of the tasks and methods in AI for protein science. 
    In this section, we focus on three subtasks, including protein folding, protein representation learning, and protein backbone generation. 
    The methods for protein folding can be categorized into two classes, before and after AlphaFold2~\citep{jumper2021highly}: (1) 
two-stage learning: RaptorX-Contact~\citep{wang2017accurate}, AlphaFold1~\citep{senior2020improved}, trRoseTTA~\citep{du2021trrosetta}, (2) end-to-end learning: AlphaFold2~\citep{jumper2021highly}, RoseTTAFold~\citep{baek2021accurate}, ESMFold~\citep{lin2023evolutionary}, OpenFold~\citep{ahdritz2022openfold}. 
    The methods for protein representation learning are grouped into the invariant networks, including IEConv~\cite{hermosilla2021intrinsicextrinsic}, HoloProt~\cite{somnath2021multiscale}, MaSIF~\cite{gainza2020deciphering, gainza2023novo}, dMaSIF~\cite{sverrisson2021fast}, ProteinMPNN~\cite{dauparas2022robust}, GearNet~\cite{zhang2023protein}, ProNet~\cite{wang2023learning}, PiFold~\cite{gao2023pifold}, and CDConv~\cite{fan2023continuousdiscrete}, and equivariant networks, including GVP-GNN~\cite{jing2021learning} and GBPNet~\cite{aykent2022gbpnet}. For protein backbone generation, the methods are grouped in terms of structure representations they use. Specifically, ProtDiff~\citep{wu2022protein}, Chroma~\citep{ingraham2022illuminating}, LatentDiff~\citep{fu2023latent}, and Genie~\citep{lin2023generating} use 3D Euclidean coordinates as the structure representation for protein backbone structure, while RFdiffusion~\citep{watson2022broadly} and FrameDiff~\citep{yim2023se} use frame representations. Besides, FoldingDiff~\citep{wu2022protein} uses internal angles to represent protein backbone structures.}
\label{fig:prot_overview}
\end{figure}

\vspace{0.1cm}\noindent\textbf{Protein Folding:} \ifshowname
\textcolor{red}{Tianfan Fu}\else\fi 
\ The three-dimensional (3D) geometric structure of proteins plays a crucial role in determining their function. The specific arrangement and spatial organization of atoms within a protein molecule are essential for its interactions with other molecules, such as substrates, cofactors, ligands, and other proteins. Traditional X-ray crystallography is indeed considered an expensive and resource-intensive method for determining protein structures~\cite{ilari2008protein}. Machine learning methods were proposed to automatically predict the protein structure based on the amino acid sequence. 
Protein folding, also known as protein structure prediction, aims to predict protein 3D structure (coordinates of all the atoms in both backbone and side chain, denoted $\mathcal{C}$ in~\cref{eqn:protein_structure}) based on the amino acid sequence $\bm{z}$.

\vspace{0.1cm}\noindent\textbf{Protein Representation Learning:\ifshowname
\textcolor{red}{Limei, Keqiang, Cong}\else\fi} Protein representation learning aims to learn informative representations for protein structures. The learned representations can be used for a wide range of predication tasks, including enzyme reaction classification~\cite{webb1992enzyme, hermosilla2021intrinsicextrinsic, hermosilla2022contrastive, zhang2023protein, fan2023continuousdiscrete}, protein inverse folding~\cite{ingraham2019generative, jing2021learning, hsu2022learning, dauparas2022robust, gao2023pifold}, and protein-ligand binding affinity prediction~\cite{wang2004pdbbind, liu2015pdb, ozturk2018deepdta, karimi2019deepaffinity, somnath2021multiscale, wang2023learning}, as shown in Figure~\ref{fig:protein_property_prediction}. Protein representation learning can significantly speed up the processes of protein screening and new protein discovery.

\vspace{0.1cm}\noindent\textbf{Protein Backbone Structure Generation:} \ifshowname
\textcolor{red}{Cong Fu}\else\fi As described above, a protein backbone consists of a chain of amino acid backbones, each of which contains the nitrogen ($N$), alpha-carbon ($C_{\alpha}$), and carbon ($C$) atoms.
These backbone atoms determine the secondary structure and overall shape of a protein, significantly affecting the protein functions. Hence, generating protein backbones is of great importance in \emph{de novo} protein design. Specifically, the protein backbone generation task is to learn a generative model $p_\theta$ that can model the density distribution of real protein backbones $p_{\mathcal{P}_{\text{bb}}}$ and then we can sample a novel protein backbone $\hat{\mathcal{P}}_{\text{bb}}$ that satisfies $p_\theta(\hat{\mathcal{P}}_{\text{bb}}) \approx p_{\mathcal{P}_{\text{bb}}}(\hat{\mathcal{P}}_{\text{bb}})$. In practice, instead of jointly modeling the density of protein backbone atom positions and amino acid types, most studies formulate this problem as a conditional generation task, where atom positions are first sampled from the learned generative model, and then amino acid types are predicted from generated structures using a trained inverse folding model. 

\ifshowname
\textcolor{red}{Keqiang Yan}\else\fi The goal for the protein backbone structure generation task is to create a model distribution that can easily generate samples to imitate the real data distribution. However, several challenges must be addressed to achieve this goal, including establishing a bijective mapping between data distribution and prior distributions such as Gaussians, ensuring distribution $E(3)$/$SE(3)$-invariance, employing $E(3)$/$SE(3)$-equivariant message passing, and efficient modeling of protein structures.

\ifshowname
\textcolor{red}{Keqiang Yan}\else\fi In this survey, for protein representation learning and protein backbone generation tasks, we mainly focus on structure-based instead of sequence-based predictive and generative methods for the following reasons. Firstly, protein representation learning is a complex task that requires the consideration of both protein structure and sequence. A significant proportion of protein functionalities are influenced by the structure and cannot be deduced directly from the sequence alone. And changes in the structure can result in different properties for the same protein sequence.
Secondly, in protein generation, a key objective is to generate new protein structures that meet specific structure constraints, such as containing specific sub-structures, possessing particular secondary structures, and binding to particular molecules and antigens. These geometric constraints can be incorporated into protein structure generation methods as conditions but cannot be directly addressed from the protein sequence generation perspective.
Particularly, protein generation using deep learning approaches is largely under-explored, and there is not much work published on this topic. Recent research trend shows diffusion models have great capacity and achieve the best performance. Thus, regarding deep learning approaches, we focus on diffusion models in this survey.

\subsection{Protein Folding}
\label{sec:protein_folding}

\noindent{\emph{Authors: Tianfan Fu, Alexandra Saxton, Shuiwang Ji, Jimeng Sun}}\newline

Different from small molecules that consist of a few atoms discussed in Section~\ref{sec:mol}, proteins are macromolecules composed of a large number of atoms (mostly 1,000 to 10,000), posing greater challenges in estimating their native structure. 
In this section, we first formulate the protein folding problem, then identify the major challenges and discuss the existing methods and datasets. Finally, we point out a couple of potential directions for future work.


\begin{figure}[t]
\centering
\includegraphics[width=\textwidth]{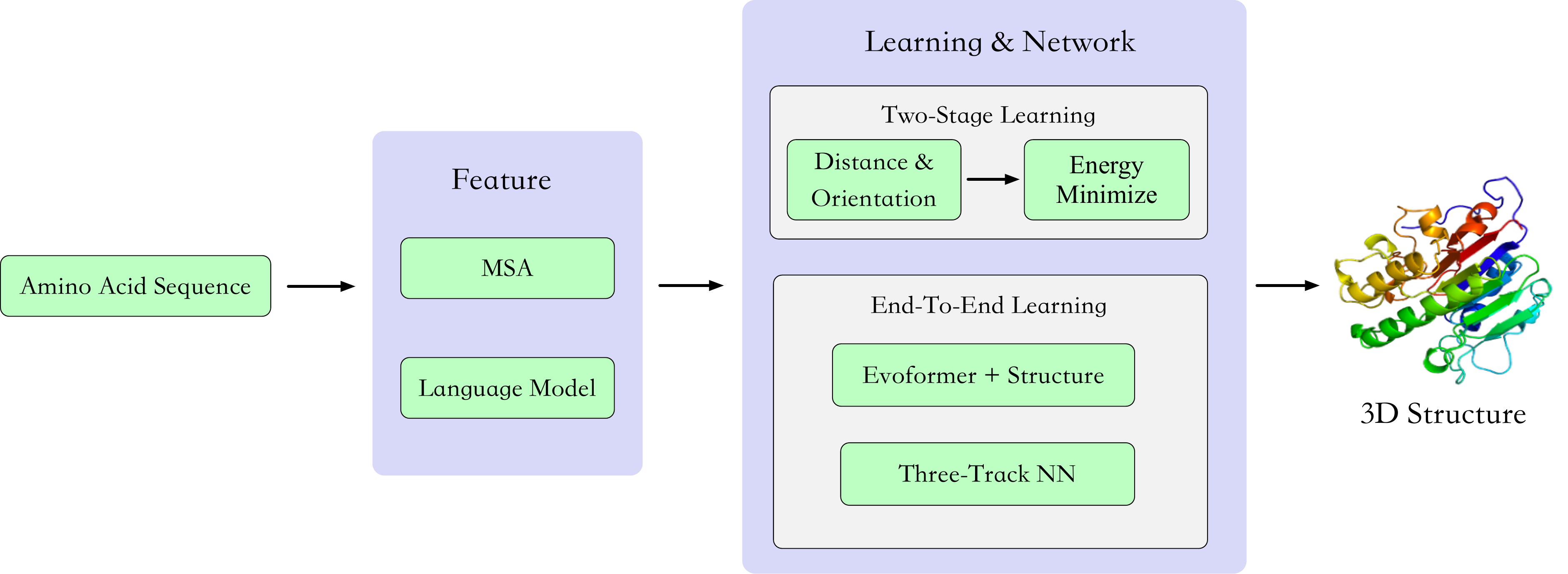}
\caption{Summarization of the protein folding algorithms. Existing methods, including RaptorX-Contact~\citep{wang2017accurate}, 
AlphaFold1~\citep{senior2020improved}, 
trRoseTTA~\citep{du2021trrosetta}, 
AlphaFold2~\citep{jumper2021highly}, RoseTTAFold~\citep{baek2021accurate}, 
AlphaFold-Multimer~\citep{evans2021protein}, 
ESMFold~\citep{lin2023evolutionary}, and
 OpenFold~\citep{ahdritz2022openfold}, follow this pipeline and their respective modules are summarized in Table~\ref{tab:protein_folding_methods}.}
\label{fig:folding}
\end{figure}

\subsubsection{Problem Setup}
Protein folding, also known as protein structure prediction, aims to predict protein 3D structure (including coordinates of all the atoms in both backbone and side chain, denoted $\mathcal{C}$ in~\cref{eqn:protein_structure}) based on the amino acid sequence $\bm{z}$. 

\subsubsection{Technical Challenges}


\vspace{0.1cm}\noindent\textbf{Generator with $E(3)$/$SE(3)$-Equivariance:} The neural network is designed to maintain consistency when applied to protein structures undergoing $SE(3)$ transformations. \revisionOne{More details about $E(3)$/$SE(3)$-equivariance can be found in~\cref{sec:group}.}

\vspace{0.1cm}\noindent\textbf{Physical Constraints:} Protein folding is governed by fundamental physical principles that dictate the spatial arrangement of atoms within the protein structure. One of these principles is the concept of bond lengths, which refers to the average distances between atoms participating in chemical bonds. In protein molecules, the distances between bonded atoms are relatively fixed, meaning they have characteristic and well-defined values. These fixed bond lengths are determined by the types of atoms involved and the specific chemical bonds formed between them. Another principle is the distance between arbitrarily paired atoms can not be too short to avoid a clash. It is necessary to incorporate these physical constraints into end-to-end models. 

\vspace{0.1cm}\noindent\textbf{Computational Efficiency:} Proteins can adopt an astronomical number of possible structures due to the flexibility of their backbone and side chains. Exploring this vast structural space to identify the most energetically favorable folded state is computationally demanding.

\subsubsection{Existing Methods}

Existing work on protein folding can be classified into two categories, known as two-stage prediction and end-to-end prediction. ~\cref{tab:protein_folding_methods} and~\cref{fig:folding} summarize the major difference between existing approaches. Before the development of geometric deep learning, to circumvent the straightforward generation of 3D coordinates, most of the earlier methods leveraged a two-stage learning process: the first stage is to predict the pairwise distance and orientation (\emph{e.g.}, torsion angles), while the second stage is to design a differentiable potential function as the optimization surrogate. The pairwise distance and orientation are invariant under $SE(3)$ transformation. 
Prominent approaches include RaptorX-Contact~\citep{wang2017accurate}, AlphaFold1~\citep{senior2020improved}, trRosetta~\citep{du2021trrosetta}, \emph{etc}, and are essentially non-end-to-end approaches. 
Then, the emergence of geometric deep learning, especially $E(3)$/$SE(3)$ neural networks, enables the buildup of an end-to-end system for protein structure prediction. Specifically, in 2020, AlphaFold2~\citep{jumper2021highly} has demonstrated remarkable accuracy in predicting the 3D structures of proteins in the 14-th CASP (Critical Assessment of Structure Prediction) competition (a biennial community-wide competition in the field of protein structure prediction). 
It concatenates Evoformer (a variant of the transformer) and the $SE(3)$-equivariant structure module. The structure module aims to transform the representation into a 3D structure. It first generates the coordinates of the backbone sequentially: For each residue, instead of the global coordinate, it produces the relative position to the previous residue, which is parameterized by a rotation matrix (three learnable parameters) and transition vector (three learnable parameters). 
It was followed by a couple of works, including RoseTTAFold~\citep{baek2021accurate}, 
AlphaFold-Multimer~\citep{evans2021protein}, ESMFold~\citep{lin2023evolutionary}, 
UniFold~\citep{li2022uni}, 
OpenFold~\citep{ahdritz2022openfold}, \emph{etc.} 

Specifically, drawing inspiration from AlphaFold2,  RoseTTAFold~\citep{lin2023evolutionary} introduced an innovative three-track neural network, enabling the joint modeling of 1D protein sequence, 2D distance map, and 3D coordinate information. By adopting this approach, impressive precision was achieved in predicting protein folding structures, on par with the performance of AlphaFold2. 
However, the original implementation of protein structure prediction is prohibitively time-consuming and resource-intensive. 
One major computational bottleneck lies in multiple sequence alignment (MSA). 
MSA is used for almost all the protein folding methods (before and after AlphaFold2) and plays a critical role in the final performance. 
The primary motivation for performing MSA is to identify and understand the functional and structural constraints on biological molecules. By aligning sequences from different species or within a single organism, researchers can identify conserved regions that are critical for maintaining the function of the molecule. MSA exhaustively searches over large-scale protein structure databases to identify similar amino acid sequences to reveal insight of biological evolutionary relationships and enhance the input feature. However, the MSA procedure is typically time-consuming and resource-demanding due to its brute-force essence. To alleviate this issue, ESMFold~\citep{lin2023evolutionary} pretrains a large language model on amino acid sequences and uses it to replace MSA with a powerful neural representation, which is shown to accelerate the whole process significantly. The other neural architectures of ESMFold follow AlphaFold2. 
OpenFold~\citep{ahdritz2022openfold} develops a memory-efficient version of AlphaFold2, and curates OpenProteinSet, one of the largest public MSA databases (five million protein structures). 
In addition, OpenFold releases the code to benefit the whole community. 
AlphaFold-Multimer~\citep{evans2021protein} enhances the prediction performance of AlphaFold in the context of multi-chain protein complex structure via incorporating more multi-chain proteins in training data. 

\begin{table}[t]
\centering
\caption{Summary of existing protein folding approaches, including 
RaptorX-Contact~\citep{wang2017accurate}, 
AlphaFold1~\citep{senior2020improved}, 
trRoseTTA~\citep{du2021trrosetta}, 
AlphaFold2~\citep{jumper2021highly}, RoseTTAFold~\citep{baek2021accurate}, 
AlphaFold-Multimer~\citep{evans2021protein}, 
UniFold~\citep{li2022uni}, 
ESMFold~\citep{lin2023evolutionary}, 
and OpenFold~\citep{ahdritz2022openfold}. 
Two-stage learning typically consists of (1) prediction of pairwise distance and orientation and (2) energy minimization. 
}
\resizebox{\textwidth}{!}{
\begin{tabular}{l|ccccc}
\toprule[1pt]
Methods  & Feature & Learning  & Network & Symmetry \\
\midrule 
RaptorX-Contact  & MSA & Two-Stage & Residual CNN & $SE(3)$-Invariant \\ 
AlphaFold1  &  MSA & Two-Stage & Residual CNN & $SE(3)$-Invariant \\
trRoseTTA  & MSA & Two-Stage & Residual CNN & $SE(3)$-Invariant \\ 
AlphaFold2  & MSA & End-To-End & Evoformer + Structure & $SE(3)$-Equivariant \\ 
RoseTTAFold  & MSA & End-To-End &  Three-Track NN & $SE(3)$-Equivariant \\ 
UniFold  & MSA & End-To-End & Evoformer + Structure & $SE(3)$-Equivariant \\ 
ESMFold  & Language model & End-To-End & Evoformer + Structure & $SE(3)$-Equivariant \\ 
OpenFold  & MSA & End-To-End & Evoformer + Structure & $SE(3)$-Equivariant \\ 
AlphaFold-Multimer & MSA & End-To-End & Evoformer + Structure & $SE(3)$-Equivariant \\ 
\bottomrule[1pt]
\end{tabular}
}
\label{tab:protein_folding_methods}
\end{table}

\subsubsection{Datasets and Benchmarks}

Protein Data Bank (PDB)~\cite{berman2000protein} is the most well-known public protein structure database, where the protein structure is determined by X-ray crystallography~\cite{ilari2008protein}. 
The PDB collects, validates, and disseminates experimentally determined atomic coordinates and related information, such as experimental methods, resolution, and bibliographic references. 
The PDB houses over 180,000 protein structures, and the number of structures in the PDB is constantly growing as researchers determine and deposit new structures. 
For AlphaFold2, the dataset comes from two sources. Specifically, 75\% of the training samples are from a self-distillation dataset from Uniclust30~\cite{mirdita2017uniclust}; 25\% are from protein data bank (PDB)~\cite{berman2000protein}. 
It removes some repetitive samples, and the final dataset contains around 475K protein structures.

The Critical Assessment of Protein Structure Prediction (CASP) is a biennial competition that aims to evaluate state-of-the-art methods in protein structure prediction. CASP provides a platform for researchers and computational methods to assess their ability to predict the three-dimensional structure of proteins accurately. 
During CASP, participants are given a set of protein sequences for which the experimental structures have been determined but are kept confidential. The participants use their computational methods to predict the corresponding protein structures without any prior knowledge of the experimental structures. These predictions are then evaluated and compared to the experimental structures to assess the accuracy and quality of the predictions. 
AlphaFold1 dominates CASP13, which was held in 2018. 
After two years, in CASP14 held in 2020, AlphaFold2 achieved nearly 90 Global Distance Test (GDT) scores, roughly equivalent to X-ray crystallography's accuracy. It became the first computational method to predict protein structures with near experimental accuracy and is called the gold standard of protein folding. Many follow-up works focus on reproducing AlphaFold2 and matching its performance, including RoseTTAFold~\citep{baek2021accurate}, OpenFold~\citep{ahdritz2022openfold}, ESMFold~\citep{lin2023evolutionary}, \emph{etc.} 
For example, thanks to the use of the large language model instead of MSA, ESMFold achieves up to 60$\times$ speedup while maintaining accuracy~\citep{lin2023evolutionary}. 
More recently, \citep{deepmind2023performance} expanded the application range of AlphaFold2 to joint structure prediction of complexes including proteins, nucleic acids, small molecules, ions, and modified residues, and demonstrated significant improvement over existing approaches, including traditional methods like Vina~\citep{trott2010autodock}, and state-of-the-art deep learning methods like DiffDock~\cite{corso2022diffdock}).

\subsubsection{Open Research Directions}

A couple of challenges remain unsolved and hinder the practical use of protein folding. 
First, most current methods can accurately predict the structures of proteins with single chains and would degrade significantly for multi-chain proteins. 
The key reason is a multi-chain protein is more complex than a single-chain one with limited available data. 
Second, most of the current methods rely heavily on MSA and degrade significantly when dealing with proteins that are different from the training set and MSA database. Thus, enhancing the generalization ability to dissimilar proteins is a critical problem.  
Third, in the real world, proteins do not exist in isolation but often interact with other proteins, RNA, DNA and small molecules. Therefore, predicting protein structures in different contexts (\emph{e.g.}, RNA-protein complex, DNA-protein complex, drug-protein complex) is also an important challenge, where the available data is also limited. So, future research aims to predict the structure of a protein in these complicated scenarios. \revisionOne{Towards this direction, AlphaFold3~\citep{abramson2024accurate} has achieved remarkable results in predicting structures and interactions of almost all life molecules, including proteins, DNA, RNA, ligands, etc. Predicting interactions among biomolecules, such as protein-protein interaction, can help us better understand complex cellular functions, often with high specificity and regulation. This understanding of biomolecule interactions can further facilitate therapeutic development.}

\subsection{Protein Representation Learning}
\label{sec_prot_pred}

\noindent{\emph{Authors: Limei Wang, Yi Liu, Cong Fu, Michael Bronstein, Shuiwang Ji}}\vspace{0.3cm}\newline
\emph{Recommended Prerequisites: Sections~\ref{subsec:cont_feat},~\ref{subsec:cont_equi},~\ref{subsec:mol_representation_learning}}\newline

\begin{figure}[t]
    \centering
    \includegraphics[width=\textwidth]{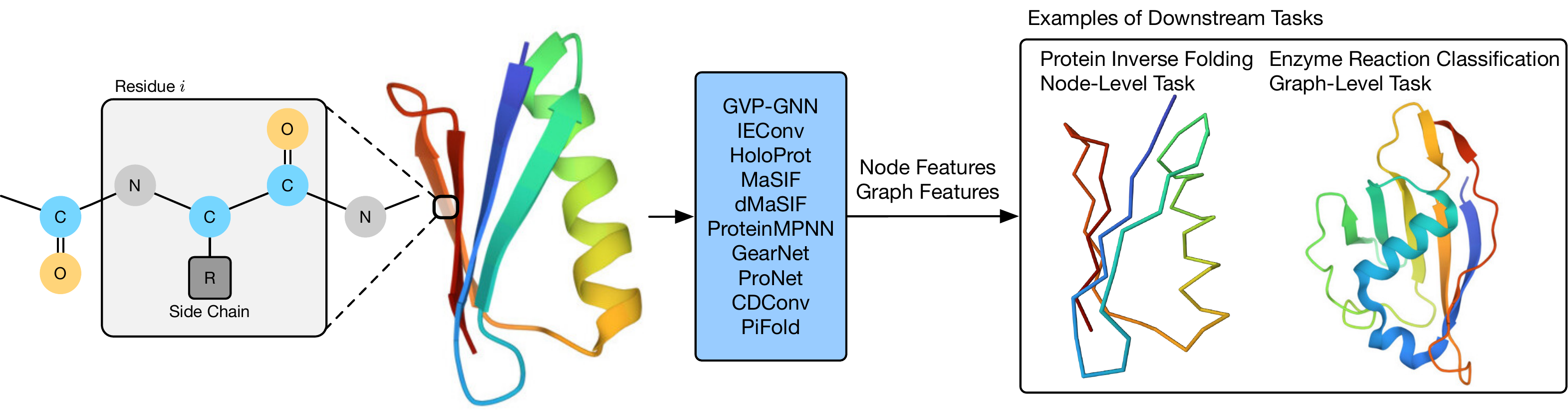}
    \caption{Illustrations of protein structures, representation learning methods, and examples of downstream tasks. Proteins consist of one or more amino acid chains. When two or more amino acids bond to form a peptide, water molecules are removed, and the remaining part of each amino acid is called an amino acid residue. The left part of the figure shows the detailed structure of a residue. In the middle, we list existing protein representation learning methods, including GVP-GNN~\cite{jing2021learning}, IEConv~\cite{hermosilla2021intrinsicextrinsic}, HoloProt~\cite{somnath2021multiscale}, MaSIF~\cite{gainza2020deciphering, gainza2023novo}, dMaSIF~\cite{sverrisson2021fast}, ProteinMPNN~\cite{dauparas2022robust}, GearNet~\cite{zhang2023protein}, ProNet~\cite{wang2023learning}, CDConv~\cite{fan2023continuousdiscrete}, and PiFold~\cite{gao2023pifold}. Different downstream tasks are shown in the right, including node-level tasks such as protein inverse folding~\cite{ingraham2019generative} and graph-level tasks such as enzyme reaction prediction~\cite{hermosilla2021intrinsicextrinsic}.}
\label{fig:protein_property_prediction}
\end{figure}



Different from small molecules discussed in Section~\ref{sec:mol}, proteins are macromolecules with a large number of atoms, making protein representation learning more challenging. In this section, we highlight the challenges associated with protein representation learning. We also summarize existing methods designed specifically for protein structural learning.

\subsubsection{Problem Setup}
The objective of protein representation learning is to learn a suitable representation that can encode important information about the given protein sequence and structure. Well-learned protein representations can be used to facilitate many downstream tasks, such as protein property prediction tasks that we focus on in this survey. Specifically, protein property prediction tasks can be classified into two categories, known as protein-level tasks and node-level tasks. For protein-level tasks like enzyme reaction classification, we aim to learn a function $f$ to predict the property $y$ of any given protein $\mathcal{P}$, and $y$ can be a real number (regression problem) or categorical number (classification problem). For node-level tasks like inverse protein folding, we aim to learn a function $f$ to predict the property $y_i$ of the $i$-th amino acid. 
\subsubsection{Technical Challenges}
Complex protein structures pose several significant challenges that need to be addressed for protein representation learning as follows. 

\vspace{0.1cm}\noindent\textbf{Computational Efficiency:} A significant challenge in protein representation learning lies in the size of proteins. Proteins can contain hundreds or even thousands of amino acids, which makes them much larger than small molecules. As a result, computational efficiency is a critical bottleneck. Effective methods are expected to address this challenge by efficiently handling the large size of proteins without compromising the prediction accuracy.

\vspace{0.1cm}\noindent\textbf{Multi-Level Structures:} Proteins are complex molecules made up of amino acids, and each amino acid comprises several atoms, as shown in Figure~\ref{fig:protein_property_prediction}. Therefore, methods should capture the amino acid level information and probably further the details at the atomic level to generate more accurate predictions. This requires a multi-level representation of proteins, which is challenging for the design of machine learning models.

\vspace{0.1cm}\noindent\textbf{Preserving Symmetries:} Methods should follow the desired symmetry. Specifically, the output of the model should be $SE(3)$-invariant, which means that the predicted results should not change with respect to the rotation and translation transformation of the protein structure. This is because the function and properties of a protein do not depend on its orientation or position in 3D space, but rather on its chemical composition and spatial arrangement of atoms.

\vspace{0.1cm}\noindent\textbf{Expressive Power:} Accurately distinguishing different protein structures poses another significant challenge in protein representation learning. For protein structures that cannot be matched via $SE(3)$ transformation, effective methods should be able to distinguish them. This requires the methods to have great expressive power.

\subsubsection{Existing Methods}
\begin{table}[t]
    \centering
    \caption{Summary of existing protein learning methods, including GearNet~\cite{zhang2023protein}, CDConv~\cite{fan2023continuousdiscrete}, GVP-GNN~\cite{jing2021learning}, ProteinMPNN~\cite{dauparas2022robust}, PiFold~\cite{gao2023pifold}, IEConv~\cite{hermosilla2021intrinsicextrinsic}, ProNet~\cite{wang2023learning}, HoloProt~\cite{somnath2021multiscale}, MaSIF~\cite{gainza2020deciphering, gainza2023novo}, and dMaSIF~\cite{sverrisson2021fast}. The complexity of a method is typically influenced by the number of nodes in the corresponding graph. Different methods can incorporate different levels of protein structures and have varying expressive power, which affects their ability to distinguish different protein structures.
    }
    \begin{tabular}{l|cccc}
    \toprule[1pt]
    Methods & Node (Complexity) & Level of Structures & Network & Symmetry\\
    \midrule
   GearNet & Amino acid & $C_{\alpha}$ & $\ell=0$ & $E(3)$-Invariant \\
   CDConv & Amino acid + Pooling & $C_{\alpha}$ & $\ell=0$ & $E(3)$-Invariant \\
   GVP-GNN  & Amino acid & Backbone & $\ell\le 1$ & $E(3)$-Equivariant\\
   ProteinMPNN & Amino acid & Backbone & $\ell=0$ & $E(3)$-Invariant\\
   PiFold & Amino acid & Backbone & $\ell=0$ & $SE(3)$-Invariant\\
   IEConv  & Atom + Pooling & All-Atom & $\ell=0$ & $E(3)$-Invariant \\
   ProNet  & Amino acid & All-Atom & $\ell=0$ & $SE(3)$-Invariant \\
   HoloProt  & Amino acid + Surface & $C_{\alpha}$ + Surface & $\ell=0$ & $SE(3)$-Invariant \\
   MaSIF, dMaSIF  & Surface & Surface & $\ell=0$ & $SE(3)$-Invariant \\
    \bottomrule[1pt]
    \end{tabular}
    \label{tab:protein_pro_pred_methods_comparsion}
\end{table}
In Section~\ref{subsec:mol_representation_learning}, we discussed recent studies on representation learning for small molecules with 3D structures, where both invariant and equivariant methods were proposed to learn accurate representations. However, proteins are macromolecules with a large number of atoms and inherent multi-level structures, presenting significant challenges, as detailed in the previous section. Therefore, it is not practical to directly apply methods designed for small molecules to proteins. In this section, we summarize existing methods that are specifically designed to process protein
structures, focusing on strategies for dealing with a large number of atoms  and capturing inherent multi-level structures, as well as how to build more powerful and symmetry-aware representation learning methods, as summarized in Table~\ref{tab:protein_pro_pred_methods_comparsion}, to tackle the above challenges.

Existing methods use different strategies to deal with the large number of atoms in proteins. For example, 
IEConv~\cite{hermosilla2021intrinsicextrinsic} treats each atom as a node in a protein graph and employs several hierarchical pooling layers to reduce the number of nodes. In addition, the pooling operations enable multi-scale protein analysis and help the model learn different levels of protein representations.  
In contrast, methods including GearNet~\cite{zhang2023protein}, ProNet~\cite{wang2023learning}, GVP-GNN~\cite{jing2021learning}, ProteinMPNN~\cite{dauparas2022robust}, PiFold~\cite{gao2023pifold}, and CDConv~\cite{fan2023continuousdiscrete} treat each amino acid as a node in the graph and consider the information of atoms in each amino acid as special node and edge features. Since each amino acid contains many atoms, the graph size in these methods is significantly smaller than that of IEConv~\cite{hermosilla2021intrinsicextrinsic}, resulting in more efficient methods.


Furthermore, existing methods capture different levels of protein structure. For example, GearNet~\cite{zhang2023protein} considers only the $C_{\alpha}$ atom in each amino acid, and this structural encoder is trained by leveraging multiview contrastive learning and different self-prediction tasks. 
GVP-GNN~\cite{jing2021learning} takes the unit vectors in the directions of as $\bm{c}_i^{N} - \bm{c}_i^{C_{\alpha}}, \bm{c}_i^{C} - \bm{c}_i^{C_{\alpha}}, \bm{c}_j^{C_{\alpha}} - \bm{c}_i^{C_{\alpha}}$ as inputs, leading to a complete description of the protein backbone structure. IEConv~\cite{hermosilla2021intrinsicextrinsic} treats each atom as a node and considers both the Euclidean distance between atoms and the shortest path with covalent or hydrogen bonds. Thus, it can capture the full-atom structure of a protein. Similarly, ProNet~\cite{wang2023learning} can also capture the full-atom structure. The difference is that ProNet treats each amino acid, rather than each atom, as a node. It uses Euler angles between two backbone triangles as edge features and dihedral angles in the side chain as additional node features. This strategy effectively captures both the backbone and side-chain structures, leading to an expressive-powerful and efficient description of the protein all-atom structure.
In addition to atomic coordinates, protein surfaces play a crucial role in understanding molecular interactions and protein functions. HoloProt~\cite{somnath2021multiscale} goes beyond considering only $C_{\alpha}$ atoms and incorporates surface structures to capture coarser details of the protein. Similarly, MaSIF~\cite{gainza2020deciphering, gainza2023novo} and dMaSIF~\cite{sverrisson2021fast} specifically recognize the importance of protein surfaces in their respective approaches, highlighting their significant role in the analysis and understanding of protein interactions.

To build powerful and symmetry-aware representation learning methods, existing methods learn different order representations for each node. As shown in Table~\ref{tab:protein_pro_pred_methods_comparsion}, most existing methods consider only scalar features, and the feature order is 0. For GVP-GNN~\cite{jing2021learning} and GBPNet~\cite{aykent2022gbpnet}, the feature order is 1, as it considers directional vectors as node and edge features. The directional features are used to update the learned features for each node. However, currently, there are no higher-order methods designed for protein representation learning, mainly due to the large scale of protein structures. It would be interesting to explore the power of high-order (and many-body) methods discussed in Section~\ref{sec:group} and Section~\ref{subsec:mol_representation_learning} for protein representation learning.

\subsubsection{Datasets and Benchmarks}
\begin{table}[t]
	\centering
	\caption{Some statistic information of CATH 4.2 curated by~\citet{ingraham2019generative}, Fold dataset~\cite{hou2018deepsf, hermosilla2021intrinsicextrinsic}, Enzyme Reaction dataset~\cite{hermosilla2021intrinsicextrinsic}, Enzyme Commission (EC) dataset~\cite{gligorijevic2021structure}, and Gene Ontology dataset~\cite{gligorijevic2021structure}. We summarize the prediction tasks and the number of protein samples (\# Samples), maximum number of amino acids in one protein (Maximum \# amino acids), and average number of amino acids in one protein (Average \# amino acids).}
	\resizebox{0.95\textwidth}{!}{
        \begin{tabular}{l|ccccc}
        \toprule
        Datasets & Prediction Tasks &\# Samples & Maximum \# amino acids & Average \# amino acids \\\midrule
            CATH 4.2 &Protein inverse folding, predict amino acid sequence &19,752 &500 &233 \\
            Fold &Protein fold classification &16,292 &1,419 &168 \\
            Enzyme Reaction &Enzyme reaction classification &37,428 &3,725 &299 \\
            Gene Ontology &Gene Ontology (GO) term prediction &36,635 &997 &258 \\
            Enzyme Commission &Enzyme Commission (EC) number prediction &19,198 &998 &299 \\
        \bottomrule
	\end{tabular}}
	\label{tab:protein_rep_data}
\end{table}

Representation learning methods for proteins with 3D structures are evaluated on various tasks, such as amino acid type prediction and protein function prediction. Table~\ref{tab:protein_rep_data} summarizes commonly used datasets, including CATH 4.2 curated by~\citet{ingraham2019generative}, Fold dataset~\cite{hou2018deepsf, hermosilla2021intrinsicextrinsic}, Enzyme Reaction dataset~\cite{hermosilla2021intrinsicextrinsic}, Enzyme Commission (EC) dataset~\cite{gligorijevic2021structure}, and Gene Ontology dataset~\cite{gligorijevic2021structure}.

CATH 4.2 dataset curated by~\citet{ingraham2019generative} is used for the task of inverse folding, also called computational protein design (CPD) or fixed backbone design, which aims to infer an amino acid sequence that can fold into a given structure. The dataset is collected based on the CATH hierarchical classification of protein structure~\cite{orengo1997cath} and is split into 18,024 structures for training, 608 for validation, and 1,120 for testing. The evaluation metrics include perplexity and recovery. Perplexity measures the ability of the model to give a high likelihood to held-out sequences, and recovery evaluates predicted sequences versus the native sequences of templates. In addition to CATH 4.2, some other datasets are also used to test the performance of models on the inverse folding task, including CATH 4.3~\cite{hsu2022learning}, TS 50~\cite{li2014direct, jing2021learning}, and TS 500~\cite{qi2020densecpd, gao2023pifold}.

Fold dataset~\cite{hou2018deepsf, hermosilla2021intrinsicextrinsic} is a collection of 16,712 proteins with 3D structures curated from the SCOPe 1.75 database~\cite{murzin1995scop}, and each of the proteins is labeled with one of 1,195 fold classes. The fold classes indicate the secondary structure compositions, orientations, and connection orders of proteins. To evaluate the generalization ability of models, three test sets are used, namely Fold, Superfamily, and Family. Specifically, the Fold test set consists of proteins whose superfamily are unseen during training, the Superfamily test set consists of proteins whose family are unseen during training, and the Family test set consists of proteins whose family are present during training. Among these three test sets, Fold is the most challenging one as it differs most from the training data set. The dataset is divided into 12,312 proteins for training, 736 for validation, 718 for Fold, 1,254 for Superfamily, and 1,272 for Family. Accuracy is the evaluation metric for the fold classification task.

Enzyme Reaction dataset~\cite{hermosilla2021intrinsicextrinsic} is a collection of enzymes, which are proteins that act as biological catalysts and can be classified with enzyme commission (EC) numbers~\cite{webb1992enzyme} based on the reactions they catalyze. In total, this dataset contains 37,428 proteins with 3D structures from 384 classes, and the EC annotations are downloaded from the SIFTS database~\cite{dana2019sifts}. The dataset is divided into 29,215 proteins for training, 2,562 for validation, and 5,651 for testing, with each EC number represented in all three splits. Accuracy is used as the evaluation metric for this task.

Enzyme Commission (EC) dataset~\cite{gligorijevic2021structure} is also a collection of enzymes. However, unlike the enzyme reaction dataset that forms a protein-level classification task, this dataset forms 538 binary classification tasks based on the three-level and four-level 538 EC numbers~\cite{webb1992enzyme}. Additionally, the enzymes collected in this dataset are different from those in the Enzyme Reaction dataset.  In total, this dataset contains 19,198 proteins, with 15,550 for training, 1,729 for validation, and 1,919 for testing. This multi-label classification task is evaluated using two metrics, namely protein-centric maximum F-score ($\text{F}_{\text{max}}$) and pair-centric area under precision-recall curve ($\text{AUPR}_{\text{pair}}$). For more detailed information on these metrics, please refer to relevant papers~\cite{gligorijevic2021structure, wang2022lm, zhang2023protein, fan2023continuousdiscrete}.

Gene Ontology dataset~\cite{gligorijevic2021structure} is used for the prediction of protein functions based on Gene Ontology (GO) terms~\cite{ashburner2000gene} and forms multiple binary classification tasks. Specifically, GO classifies proteins into hierarchically related functional classes organized into three different ontologies, namely biological process (BP) with 1,943 classes, molecular function (MF) with 489 classes, and cellular component (CC) with 320 classes. The dataset is divided into 29,898 proteins for training, 3,322 for validation, and 3,415 for testing. The evaluation metrics are the same as those used for Enzyme Commission (EC) dataset~\cite{gligorijevic2021structure}.

In addition to the datasets mentioned above, Atom3D~\cite{townshend2020atom3d} is also commonly used to test the performance of protein representation learning methods. Specifically, Atom3D is a unified collection of datasets concerning the 3D structures of biomolecules, including proteins, small molecules, and nucleic acids. It includes several datasets for protein-related tasks, such as Protein Interface Prediction (PIP), Residue Identity (RES), Mutation Stability Prediction (MSP), Ligand Binding Affinity (LBA), Ligand Efficacy Prediction (LEP), Protein Structure Ranking (PSR). The LBA task is described in detail in Section~\ref{sec:dock}.

\subsubsection{Open Research Directions}
Despite the recent advances in protein representation learning, several challenges remain unresolved, and certain directions remain underexplored. For example, while current methods can effectively capture the full-atom structure of proteins, it is still uncertain whether these methods can accurately capture the important local substructures, such as $\alpha$-helix and $\beta$-sheet, in the secondary structure, and their spatial arrangement in the tertiary structure. Additionally, incorporating accessible surface area and protein domains into protein representation learning methods is important for a more comprehensive understanding of protein structure and function, potentially enhancing performance as well. For example, MaSIF~\cite{gainza2020deciphering, gainza2023novo} focuses on solvent-excluded protein surfaces represented as meshes. On the other hand, dMaSIF~\cite{sverrisson2021fast} employs oriented point clouds to model protein surfaces. Meanwhile, although both $\ell=0$ and $\ell=1$ equivariant methods have been proposed for protein representation learning, there is currently a lack of higher-order methods, which have been extensively investigated for small molecules.

In addition, protein dynamics is a significant and actively evolving field that focuses on studying the motions, conformational changes, and interactions of proteins over time, which are critical for understanding the function and behavior of proteins. One direction to explore protein dynamics is by leveraging temporal GNNs in conjunction with the protein representation learning methods we introduced. By incorporating temporal information and accounting for the dynamic nature of protein structures and interactions, temporal GNNs offer a promising avenue for unraveling the intricacies of protein dynamics. Another important aspect is protein circuit design, which holds great potential for advancing our understanding of protein function and behaviors.

\revisionOne{Moreover, predicting the mutation effects for proteins is also a crucial task related to addressing challenges in genetic disease, climate, agriculture, etc. It often requires learning a "fitness landscape" which maps protein sequences or structures to their resulting properties. To facilitate this research, ProteinGym~\citep{notin2024proteingym} was proposed as a large scale benchmark specifically developed for protein fitness prediction and design, consisting of large scale deep mutational scanning assays, curated clinical dataset with mutation effects annotated by experts, and a robust evaluation framework.}

\subsection{Protein Backbone Structure Generation}
\label{sec_prot_gen}

\noindent{\emph{Authors: Keqiang Yan, Cong Fu, Yi Liu, Shuiwang Ji}}\vspace{0.3cm}\newline
\emph{Recommended Prerequisites: Sections~\ref{subsec:mol_conformer_generation},~\ref{subsec:mol_generation}}\newline

{\ifshowname\textcolor{red}{Keqiang Yan}\else\fi}
 In this section, we first describe the aforementioned challenges for protein backbone structure generation in detail, then discuss how previous methods address these challenges from two perspectives, including protein structure representations and diffusion processes. The pipeline of protein generation with diffusion model is illustrated in~\cref{fig:protein_generation}. As mentioned above, we focus on diffusion models for protein backbone structure generation in this survey. Aside from this line of works, there are also works that are based on flow matching~\citep{bose2023se3stochastic, yim2023fast}.

\subsubsection{Problem Setup}
The protein backbone generation task is to learn a generative model $p_\theta$ that can model the density distribution of real protein backbones $p_{\mathcal{P}_{\text{bb}}}$ and then we can sample a novel protein backbone $\hat{\mathcal{P}}_{\text{bb}}$ that satisfies $p_\theta(\hat{\mathcal{P}}_{\text{bb}}) \approx p_{\mathcal{P}_{\text{bb}}}(\hat{\mathcal{P}}_{\text{bb}})$.

\subsubsection{Technical Challenges}
{\ifshowname\textcolor{red}{Keqiang Yan}\else\fi}
\vspace{0.1cm}\noindent\textbf{Sophisticated Data Distribution:} Generation tasks need the sampling of new data points from the data distribution. However, the distribution of real protein structures is unknown, sparse, and intractable to sample from. Thus, establishing a bijective mapping between the data distribution and prior distributions, such as Gaussians, is crucial. 

{\ifshowname\textcolor{red}{Keqiang Yan}\else\fi}

\vspace{0.1cm}\noindent\textbf{Distribution $E(3)$/$SE(3)$-Invariance:} Distribution $SE(3)$-invariance, which arises from the nature of real protein structure distribution, needs to be satisfied. Specifically, for a protein backbone structure $\mathcal{P}_{\text{bb}}$ sampled from real data distribution $p_{\mathcal{P}_{\text{bb}}}$, if we apply 3D rotation transformations $R \in \mathbb{R}^{3\times3}, |R| = 1$ for $SE(3)$ and translation transformations $b \in \mathbb{R}^{3}$, the geometric structure of the given protein remains unchanged. Hence, we have $p_{\mathcal{P}_{\text{bb}}}(\mathcal{P}_{\text{bb}}) = p_{\mathcal{P}_{\text{bb}}}(R\mathcal{P}_{\text{bb}} + b)$ for real data distribution $p_{\mathcal{P}_{\text{bb}}}$. Some early works also consider distribution $E(3)$-invariance in which $|R| = \pm 1$, but this imposes incorrect inductive bias since natural proteins are chiral molecules and sensitive to reflection. However, for completeness of review, we still include works that consider distribution $E(3)$-invariance. 

{\ifshowname\textcolor{red}{Cong Fu}\else\fi}

\vspace{0.1cm}\noindent\textbf{$E(3)$/$SE(3)$-Equivariant Networks:} In order to achieve distribution $E(3)$/$SE(3)$-invariance, the neural networks should satisfy the equivariance property. As discussed above, two identical protein structures up to a group transformation should satisfy $p_{\mathcal{P}_{\text{bb}}}(\mathcal{P}_{\text{bb}}) = p_{\mathcal{P}_{\text{bb}}}(R\mathcal{P}_{\text{bb}} + b)$. 
For a protein in 3D space, first, we can let the coordinates have zero centroids by subtracting the mean values of the coordinates. In this way, the translational invariance for the distribution is naturally satisfied.
Therefore, we only need to consider rotational invariance, that is, $p_{\mathcal{P}_{\text{bb}}}(\mathcal{P}_{\text{bb}}) = p_{\mathcal{P}_{\text{bb}}}(R\mathcal{P}_{\text{bb}})$ 
should be satisfied. Through the total probability rule, we have 
$p(\mathcal{P}_{\text{bb}}) = \int p(\mathcal{P}_{\text{bb}}|\bm{Z})p(\bm{Z}) d\bm{Z}$, where $\bm{Z}$ denotes the latent variables. Similarly, we also have 
$p(R\mathcal{P}_{\text{bb}}) = \int p(R\mathcal{P}_{\text{bb}}|R\bm{Z})p(R\bm{Z}) dR\bm{Z}$.

If we sample latent variables $\bm{Z}$ from zero-centered Gaussian distribution, then $p(R\bm{Z}) = p(\bm{Z})$ can be easily satisfied. And in order to achieve $p(R\mathcal{P}_{\text{bb}}|R\bm{Z}) = p(\mathcal{P}_{\text{bb}}|\bm{Z})$, we should make the networks mapping from $\bm{Z}$ to $\mathcal{P}_{\text{bb}}$ to be equivariant to $R$. Thus, $p(R\mathcal{P}_{\text{bb}}) = p(\mathcal{P}_{\text{bb}})$ holds and distribution $E(3)$/$SE(3)$-invariance is satisfied.

{\ifshowname\textcolor{red}{Keqiang Yan}\else\fi}

\vspace{0.1cm}\noindent\textbf{Computational Efficiency:} Efficient modeling of protein structures is also crucial. In addition to the aforementioned challenges, it is worth noting that the exploration space of protein structures is incredibly vast, while known protein structures are limited. Therefore, elegant protein representations must be established not only to ease the modeling difficulty but also to minimize geometric bias in machine learning models.  

\begin{figure}[t]
    \centering
    \includegraphics[width=\textwidth]{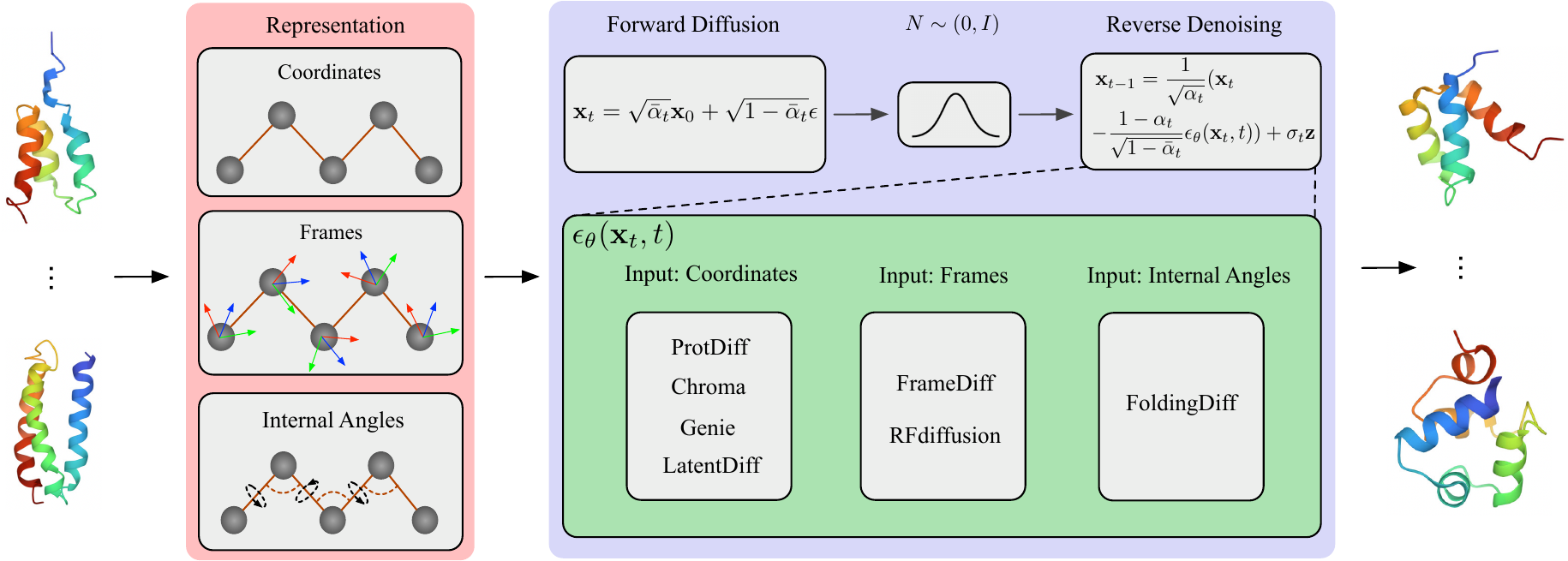}
    \caption{Pipeline of protein generation with diffusion models. The process begins by diffusing protein backbone structures to Gaussian noise, which is then denoised using a learned denoising network to generate novel protein structures. Protein backbones can be represented in various ways, such as coordinates, frames, or internal angles. Different methods have been proposed for each representation. For coordinates representation, methods include ProtDiff~\citep{trippe2022diffusion}, Chroma~\citep{ingraham2022illuminating}, LatentDiff~\citep{fu2023latent} (coordinates are in the latent space), and Genie~\citep{lin2023generating}. For frames representation, methods include FrameDiff~\citep{yim2023se} and RFdiffusion~\citep{watson2022broadly}. For internal angles representation, the method is FoldingDiff~\citep{wu2022protein}.}
    \label{fig:protein_generation}
\end{figure}

\subsubsection{Existing Methods}
\noindent\textbf{Protein Structure Representations:{\ifshowname\textcolor{red}{Keqiang}\else\fi}} To address the efficiency issue of protein structure modeling, recent approaches either generate protein backbone structures alone~\cite{wu2022protein, trippe2022diffusion, fu2023latent, lin2023generating, yim2023se}, or first generate protein backbone structures and then predict the full atom-level protein structures~\cite{ingraham2022illuminating, watson2022broadly} due to high complexity and vast exploration space of complete protein structures. 
When modeling protein backbone structures, different 3D representations are used, resulting in different modeling costs and diffusion processes. 
Specifically, ProtDiff~\cite{trippe2022diffusion},  Chroma~\cite{ingraham2022illuminating}, LatentDiff~\citep{fu2023latent}, and Genie~\cite{lin2023generating} represent protein backbone structures by alpha carbon positions $\mathcal{P}_{\text{bb}}$, without considering positions of $N, C, O$ atoms. LatentDiff further encodes the protein backbone structure into the compact latent space to reduce modeling complexity. Both FoldingDiff~\cite{wu2022protein} and RFdiffusion~\cite{watson2022broadly} represent protein backbone structures based on positions of $C_\alpha, N, C$ atoms. However, FoldingDiff~\cite{wu2022protein} uses relative bond and torsion angles along amino acid chains, and RFdiffusion~\cite{watson2022broadly} uses the 3D positions of $C_\alpha$ and $3\times3$ rotation matrices representing the rigid-body orientation of each residue in a global reference frame. Recently,  FrameDiff~\cite{yim2023se} further considers positions of $O$ atoms and represents protein backbone structures using 3D positions of $C_\alpha$, $3\times3$ rotation matrices, and rotation torsion angles of $O$. The representations, corresponding levels of structural granularity, and modeling spaces of previous works are summarized in Table~\ref{tab:pro_rep_tab1}. 


\begin{table}[t]
    \centering
    \caption{{\ifshowname\textcolor{red}{Keqiang 90\%, cong 10\%}\else\fi}Summary of protein 3D representations used by previous works as well as the corresponding levels of protein structural granularity and modeling space. $N$denotes the number of amino acids. $f$ denotes the downsampling factor in LatentDiff. Among them, ProtDiff~\citep{wu2022protein}, Chroma~\citep{ingraham2022illuminating}, and Genie~\citep{lin2023generating} only consider positions of alpha carbons, LatentDiff~\citep{fu2023latent} consider the node positions in the latent space, while FoldingDiff~\citep{wu2022protein} and RFdiffusion~\citep{watson2022broadly} further consider positions of carbon and nitrogen atoms in protein backbone structures. FrameDiff~\citep{yim2023se} considers all atoms in protein backbone structures.
    }
    \small
    \begin{tabular}{l|ccc}
    \toprule[1pt]
    Methods & Protein 3D Representations & Structural Granularity & Modeling Space \\
    \midrule
   ProtDiff & Coordinates& $C_{\alpha}$ & $\mathbb{R}^{N\times3}$ \\
   FoldingDiff  & Bond and torsion angles & $C_{\alpha}, C, N$ & $[0, 2\pi )^{6N}$ \\
   Chroma  & Coordinates & $C_{\alpha}$ & $\mathbb{R}^{N\times3}$\\
   RFdiffusion  & Coordinates + Frame rotation angles &$C_{\alpha}, C, N$& $\mathbb{R}^{N\times3} SO(3)^{N}$\\
   LatentDiff & Coordinates (latent space) & $C_{\alpha}$ & $\mathbb{R}^{\frac{N}{f} \times3}$\\
   Genie  & Coordinates & $C_{\alpha}$ & $\mathbb{R}^{N\times3}$ \\
   FrameDiff  & Coordinates + Frame rotation angles & $C_{\alpha}, C, N, O$ & $\mathbb{R}^{N\times3} SO(3)^{N} [0, 2\pi )^{N}$  \\
    \bottomrule[1pt]
    \end{tabular}
    \label{tab:pro_rep_tab1}
\end{table}




\vspace{0.1cm}\noindent\textbf{Diffusion Models:}{\ifshowname\textcolor{red}{Keqiang Yan}\else\fi} Given different protein structure representations, corresponding diffusion processes need to be established to address challenges in protein backbone generation, including establishing the bijective mapping between data distribution and prior distribution to enable sampling, ensuring distribution $E(3)$/$SE(3)$-invariance, and $E(3)$/$SE(3)$-equivariant property of neural networks. 

{\ifshowname\textcolor{red}{Keqiang Yan}\else\fi}3D Euclidean coordinates of alpha carbons are used in ProtDiff~\cite{trippe2022diffusion}, LatentDiff~\citep{fu2023latent}, and Genie~\cite{lin2023generating}, with relatively simple diffusion process. Specifically, ProtDiff, LatentDiff, and Genie use the zero-mean distribution to get rid of influences of translations in 3D space and achieve distribution translation invariance. Moreover, LatentDiff encodes protein structures into the latent space to reduce the modeling complexity. Beyond this, to address influences of rotation transformations, ProtDiff uses the $E(3)$-equivariant network EGNN and achieves distribution $O(3)$-invariance, while LatentDiff and Genie achieve distribution $SO(3)$-invariance which is sensitive to reflections by using SE(3)GNNs~\citep{schneuing2022structure} and $SE(3)$-equivariant IPA layers, respectively. Corresponding forward and reverse diffusion processes for 3D coordinates similarly used in the image domain are established to enable the sampling of protein structures. One limitation of this structure representation is that only the positions of alpha carbons in the backbone structure are considered, and an additional generation step is needed to generate the positions of $N$, $C$, and $O$ atoms. 

{\ifshowname\textcolor{red}{Cong Fu}\else\fi}As mentioned above, diffusion models need to transform original data into Gaussian noise and learn a denoising network to mimic and generate realistic data from Gaussian noise. Most methods transform Euclidean coordinates by adding isotropic Gaussian noise, which appears as an uncorrelated diffusion process and could break some common structure constraints of proteins. As a result, the models need to have extra designs to learn these  correlations from data. To avoid this, Chroma~\cite{ingraham2022illuminating} introduces a correlated diffusion process that transforms proteins into random collapsed polymers and uses designed covariance models to encode the chain and radius of gyration constraints. Additionally, Chroma designs a random graph neural network that can capture long-range information with sub-quadratic scaling, and the network predicts pairwise inter-residue geometries and then optimize 3D protein structures in a $SE(3)$-equivariant manner. 

{\ifshowname\textcolor{red}{Keqiang Yan}\else\fi}Bond and torsion angles along the backbone structure are used by FoldingDiff~\cite{wu2022protein}, with the assumption that the lengths of chemical bonds along the backbone chain follow practical constraints. Due to the $E(3)$-invariant nature of bond and torsion angles, a simple sequence model is used to achieve distribution $E(3)$-invariant. To enable the sampling process from Gaussian noise, FoldingDiff applies the forward and reverse diffusion processes similar to coordinates to bond angles and torsion angles, regardless of the fact that bond angles and torsion angles belong to compact Riemannian Manifolds instead of Euclidean space. 

{\ifshowname\textcolor{red}{Keqiang Yan 70\%, Cong Fu 30\%}\else\fi}Frame representations consisting of 3D positions of alpha carbons and relative rotation angles of amino acid planes are used by RFdiffusion~\cite{watson2022broadly} and FrameDiff~\cite{yim2023se}. Specifically, to achieve distribution $SE(3)$-invariance, FrameDiff uses the zero-mean distribution to get rid of influences of translations in 3D space and achieve distribution translation invariance for 3D positions of alpha carbons. Beyond this, to address influences of rotation transformations, FrameDiff achieves distribution $SO(3)$-invariance which is sensitive to reflections by using $SE(3)$-equivariant IPA layers. And RFdiffusion modifies RoseTTAFold, a powerful protein structure prediction method, as the denoising network and achieves $SE(3)$-equivariance property. To enable the sampling process from Gaussian noise, different from previous works~\cite{wu2022protein, lin2023generating, wu2022protein} using diffusion processes established in Euclidean space, FrameDiff proposes solid $SO(3)$ diffusion forward and reverse processes for the rotation matrices of amino acid planes which belong to compact Riemannian Manifolds. Also, for the RFdiffusion, noise should be added to the rotation matrix, so Brownian motion on the manifold of $SO(3)$ is adopted. And the diffusion processes of alpha carbon positions in FrameDiff and RFdiffusion are similar to ProtDiff and Chroma. Moreover, RFdiffusion uses a self-conditioning mechanism that uses the denoising network output as the template input to the subsequent denoising step, which is similar to the recycling in AlphaFold2~\citep{jumper2021highly}. The diffusion space and achieved distribution symmetries of previous works are summarized in~\cref{tab:protein_rep_gen_methods_comparsion}.

\begin{table}[t]
    \centering
    \caption{{\ifshowname\textcolor{red}{Keqiang}\else\fi}Demonstration of diffusion processes for different protein representations and achieved distribution symmetries of previous works. Among them, ProtDiff~\citep{wu2022protein} and FoldingDiff~\citep{wu2022protein}  achieve distribution $E(3)$-invariance and treat chiral protein structures as the same, while Chroma~\citep{ingraham2022illuminating}, RFdiffusion~\citep{watson2022broadly}, LatentDiff~\citep{fu2023latent}, Genie~\citep{lin2023generating}, and FrameDiff~\citep{yim2023se} achieve distribution $SE(3)$-invariance and have better generation performances.
    }
    \begin{tabular}{l|ccc}
    \toprule[1pt]
    Methods & Network & Diffusion Space & Distribution Symmetry \\
    \midrule
   ProtDiff & $E(3)$-Equivariant & Euclidean &  $E(3)$-Invariance  \\
   FoldingDiff  & $E(3)$-Invariant & Angle &   $E(3)$-Invariance\\
   Chroma  & $E(3)$-Equivariant & Euclidean &  $SE(3)$-Invariance\\
   RFdiffusion  & $SE(3)$-Equivariant & Euclidean + $SO(3)$ &  $E(3)$-Invariance\\
   LatentDiff & $SE(3)$-Equivariant & Euclidean &  $SE(3)$-Invariance\\
   Genie  & $SE(3)$-Equivariant & Euclidean &  $SE(3)$-Invariance \\
   FrameDiff  & $SE(3)$-Equivariant & Euclidean + $SO(3)$ + $SO(2)$ &   $SE(3)$-Invariance \\
    \bottomrule[1pt]
    \end{tabular}
    \label{tab:protein_rep_gen_methods_comparsion}
\end{table}






\subsubsection{Datasets and Benchmarks}{\ifshowname\textcolor{red}{Keqiang Yan 50\%, Cong Fu 50\%}\else\fi}
For now, there are no standard benchmark datasets for the task of protein backbone structure generation. Early protein backbone generation methods are usually evaluated on selected protein structures from PDB~\citep{berman2000protein} or other protein structure libraries. Specifically, ProtDiff uses 4269 single-chain protein structures with the number of amino acids in the range $[40, 128]$, while FoldingDiff uses the CATH~\citep{ingraham2019generative} dataset with 24316 structures for training, 3039 structures for validation, and 3040 structures for testing. Chroma~\cite{ingraham2022illuminating} queries non-membrane X-ray protein structures with a resolution of 2.6 \AA \ or better, and an additional set of 1725 non-redundant antibody structures was added, resulting in 28819 structures in total. In RFdiffusion~\cite{watson2022broadly}, RoseTTAFold (RF) is pre-trained on a mixture of several data sources, including monomer/homo-oligomer and hetero-oligomer structures in the PDB, AlphaFold2 data having pLDDT $>$ 0.758, and negative protein-protein interaction examples generated by random pairing. Then, RFdiffusion is trained on monomer structures in the PDB used for RF training. LatentDiff curates about 100k training data from Protein Data Bank (PDB)~\citep{berman2000protein} and Swiss-Prot data in AlphaFold Protein Structure
Database (AlphaFold DB)~\citep{jumper2021highly, varadi2022alphafold}. Genie uses 8766 protein domains, with 3,942 domains having at most 128 residues from the Structural Classification of Proteins-extended (SCOPe) dataset, and FrameDiff uses 20312 protein backbones from PDB~\citep{berman2000protein} for training. When evaluating model performance, the widely-used metrics include the scTM score (higher is better) and novelty of generated protein backbone structures compared with the training structures.

\subsubsection{Open Research Directions{\ifshowname\textcolor{red}{Keqiang}\else\fi}} 
Recent protein generative models mainly focus on the backbone level of protein structures but can not generate full atom-level protein structures in a one-step manner. Additionally, recent methods are mainly designed for random protein structure generation or conditional generation given protein substructures. Another potential direction beyond this will be generating protein structures satisfying desirable properties. 


\clearpage
\hypertarget{AI for Materials Science}{\section{AI for Materials
Science}} \label{sec:mat}

In this section, we discuss the applications of AI techniques in materials science. We first give an overview introduction about crystalline materials and elaborate formal definitions of physical symmetries of crystalline materials in Section~\ref{sec:mat_overview}. Next, we discuss two common and fundamental tasks, material representation learning problem and the material generation problem in Section~\ref{sec:mat_pred} and~\ref{sec:mat_gen}, respectively.
\ifshowname\textcolor{red}{Yi Liu}\else\fi Furthermore, we include three advanced topics,
including ordered crystalline materials characterization in \cref{sec:mat_char}, disordered crystalline materials characterization in \cref{sec:local_disorder}, and phonon calculations in \cref{sec:phonon}.

\subsection{Overview}
\label{sec:mat_overview}

\noindent{\emph{Authors: Youzhi Luo, Yuchao Lin, Yi Liu, Shuiwang Ji}}\newline





In addition to small molecules and proteins, AI methods have been used for modeling crystalline materials, which are another family of large chemical compounds formed by periodic repetitions of atoms in 3D space. Crystalline materials are the foundation of many real-world industrial applications, such as semiconductor electronics, solar cells, and batteries~\cite{butler2018machine}. Due to the dramatic demand of the industry, materials science has emerged to study a variety of fundamental research, such as predicting material properties (\emph{e.g.}, formation energy) and designing novel materials with target properties. For a long time, the research progress in these problems was relatively slow due to heavy reliance on either expensive lab experiments or time-consuming materials simulations. Recently, inspired by the success of AI methods, especially machine learning models on molecules, many studies have tried to apply these models to crystalline materials related problems~\cite{choudhary2023large,du2023m2hub}. Nonetheless, different from molecules, the arrangement of atoms in crystalline materials has a complicated periodic arrangement of repeating unit cells and atoms. Hence, crystalline materials have very different physical symmetries from molecules, and developing effective AI models for them requires explicitly capturing these symmetries in models.


In this and the following sections, we describe the structure of a crystalline material by lattice vectors and one of its unit cells, \emph{i.e.}, the smallest repeatable structures. Specifically, let the number of atoms in any unit cell be $n$, then a crystalline material $\mathcal{M}$ is represented as $\mathcal{M}=(\bm{z}, C, L)$. Here, $\bm{z}\in\mathbb{Z}^n$ is the atom type vector where the $i$-th element $z_i$ of $\bm{z}$ denotes the atom type (atomic number) of the $i$-th atom in the unit cell. $C=[\bm{c}_1,...,\bm{c}_n]\in\mathbb{R}^{3\times n}$ is the coordinate matrix where $\bm{c}_i$ denotes the 3D coordinate of the $i$-th atom in the unit cell. $L=[\bm{\ell}_1,\bm{\ell}_2,\bm{\ell}_3]\in\mathbb{R}^{3\times 3}$ is the lattice matrix, and the three lattice vectors $\bm{\ell}_1,\bm{\ell}_2,\bm{\ell}_3$ describe the three periodicity vectors along which the atoms periodically repeat themselves. 
In physics, there exist several well-defined symmetry transformations for crystalline materials, including the following permutation, $E(3)$, and periodic transformations.
\begin{itemize}
	\item \textbf{Permutation transformations} produce a new material by permuting the atom orders in $\mathcal{M}=(\bm{z},C,L)$, \emph{i.e.}, exchanging the elements in $\bm{z}$ and column vectors in $C$ with the same order.
	\item \textbf{E(3) transformations}, or rigid transformations, change the 3D coordinates of $\mathcal{M}=(\bm{z},C,L)$ by translation, rotation, or reflection in 3D space. Specifically, $C$ and $L$ are replaced by $RC+\bm{t}\bm{1}^T$ and $RL$, respectively, where $\bm{t}\in\mathbb{R}^3$ is an arbitrary translation vector, $R\in\mathbb{R}^{3\times 3}$ is an orthogonal matrix satisfying $R^TR=I$, and $\bm{1}$ is an $n$-dimensional vector whose elements are all $1$s.
	\item \textbf{Periodic transformations} map the 3D coordinates of $\mathcal{M}=(\bm{z},C,L)$ to periodically equivalent coordinates, which can be formally described as replacing $C$ by a new coordinate matrix $C'=C+LK$, where $K\in\mathbb{Z}^{3\times n}$ is an arbitrary integer matrix.
\end{itemize}

\begin{figure}[H]
    \centering
    \includegraphics[width=\textwidth]{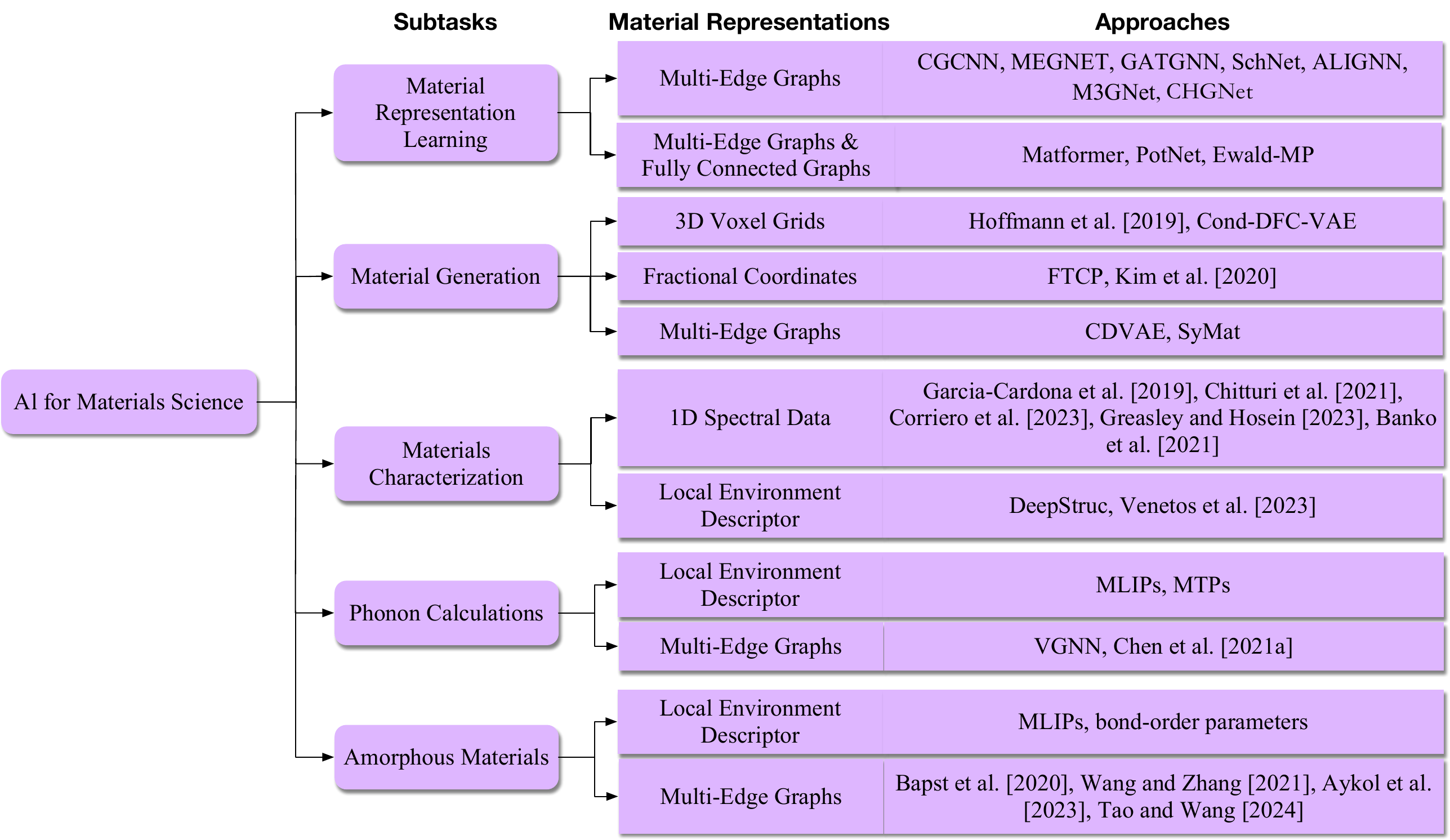}
    \caption{
    An overview of the tasks and methods in AI for materials science. In this section, we consider five tasks, including material representation learning, material generation, materials characterization, phonon calculations, and amorphous materials. In material representation learning, we discuss two categories of methods using different graph representations of materials. The first category of methods, including CGCNN~\cite{xie2018crystal}, MEGNET~\cite{chen2019graph}, GATGNN~\cite{louis2020graph}, SchNet~\cite{schutt2018schnet}, ALIGNN~\cite{choudhary2021atomistic}, M3GNet~\cite{chen2022universal}, 
    and CHGNet~\cite{deng2023chgnet} only use multi-edge graphs. The second category of methods, including Matformer~\cite{yan2022periodic}, PotNet~\cite{lin2023efficient}, and Ewald-MP~\cite{kosmala2023ewald}, use multi-edge graphs and fully connected graphs to capture periodic information. In material generation, we discuss three categories of methods using three different representations of 3D material structures as generation targets. The first category of methods, including~\citet{hoffmann2019generating} and Cond-DFC-VAE~\cite{court20203}, generate 3D voxel grids. The second category of methods, including FTCP~\cite{ren2022invertible} and~\citet{kim2020generative}, generate fractional coordinates. The third category of methods, including CDVAE~\cite{xie2022crystal} and SyMat~\cite{luo2023symmetryaware}, generate materials in the form of multi-edge graphs. In materials characterization, we discuss two categories of methods using different representations of materials. The first category of methods, including~\citet{structure_neutronscattering},~\citet{chitturiAutomatedPredictionLattice2021},~\citet{corrieroCrystalMELANewCrystallographic2023},~\citet{greasleyExploringSupervisedMachine2023} and~\citet{bankoDeepLearningVisualization2021}, uses 1D spectral data to represent materials and then to predict material structures. The second category of methods, including DeepStruc~\cite{kjaer2023deepstruc} and~\citet{venetos2023machine}, takes advantage of local environment descriptors of materials to model disordered materials. 
    \emph{Note that both Section~\ref{sec:mat_char} and Section~\ref{sec:local_disorder} describe materials characterization while Section~\ref{sec:local_disorder} specifically addresses scenarios for disordered materials. For the sake of clarity, these two sections have been integrated into a single task, materials characterization, in this figure.}
    In phonon calculations, we discuss two categories of methods using different representations of materials. The first category of methods, including MLIPs~\cite{MORTAZAVI2020100685} and MTPs~\cite{Zuo2020}, applies local environment descriptors of materials to represent material structure. The second category of methods, including VGNN~\cite{okabe2023virtual} and \citet{https://doi.org/10.1002/advs.202004214}, uses multi-edge graphs to capture periodic information.
    \revisionOne{For amorphous materials, we discuss two categories of material representations. The first uses local environment descriptors to represent material structures in the form of MLIPs~\cite{Deringer2017MLIPamorph, Deringer2018MLIPSi} or bond-order parameters~\cite{boattini2020autonomously}. The second uses multi-edge graphs to represent information of amorphous materials, including ~\citet{bapst2020unveiling}, ~\citet{wang2021inverse}, ~\citet{aykol2023predicting}, and ~\citet{tao2024artificial}.}
    }
\label{fig:openmat_overall}
\end{figure}


\ifshowname\textcolor{red}{Yi Liu}\else\fi In other words, for an arbitrary material $\mathcal{M}=(\bm{z},C,L)$, if $\mathcal{M}'$ is obtained by applying one of the above transformations on $\mathcal{M}$, we should consider $\mathcal{M}$ and $\mathcal{M}'$ as different representations of the same material. Ideally, the learned property prediction function $f$ and material distribution $p$ should be symmetry-aware, \emph{i.e.}, satisfying $f(\mathcal{M})=f(\mathcal{M}')$ and $p(\mathcal{M})=p(\mathcal{M}')$. In the following subsections, we review and discuss existing AI methods for material representation learning and material generation, as well as emerging topics including ordered/disordered materials characterization and phonon calculation, and compare them mainly from the perspective of capturing symmetries. See an overview of our covered methods in Figure~\ref{fig:openmat_overall}.

\subsection{Material Representation Learning}
\label{sec:mat_pred}

\noindent{\emph{Authors: Keqiang Yan, Yuchao Lin, Youzhi Luo, Yi Liu, Shuiwang Ji}}\newline
\newline\emph{Recommended Prerequisites: Section~\ref{sec:mat_overview}}\newline

\ifshowname\textcolor{red}{Youzhi}\else\fi
We first discuss material representation learning for property prediction in this section. The major challenge of developing material representation learning models lies in capturing symmetries in crystalline materials, particularly the invariance to periodic transformations. To overcome this challenge, existing studies have proposed numerous crystal graph representation construction methods and crystal graph neural network models. We elaborate on them in the next subsections.  In this work, we mainly focus on graph representation learning for materials using geometric information. Aside from this line of works, there are also works that are coordinate-free~\citep{goodall2020predicting,goodall2022rapid,wang2021compositionally,zhang2022composition}.

\subsubsection{Problem Setup}
\ifshowname\textcolor{red}{Yi Liu}\else\fi
Material representation learning requires learning a function $f$ to predict the property $y$ of any given material $\mathcal{M}$, and $y$ can be a real number (regression problem) or categorical number (classification problem).

\subsubsection{Technical Challenges}

\ifshowname\textcolor{red}{Keqiang}\else\fi Crystalline material representation learning aims to predict physical and chemical properties of crystalline materials based on their lattice structures. As already illustrated in the above section, different from small molecules or proteins, crystalline materials consist of a smallest unit cell structure and corresponding periodic repeating patterns in 3D space. Thus, unique geometric symmetries and model designs need to be established for crystalline materials. Specifically, when rotation transformations are applied to $C$ and $L$ together, or when translation transformations are applied to $C$ alone, the crystal structure remains unchanged, which is described as Unit Cell $E(3)$ invariant property by \citet{yan2022periodic}. Beyond this, when periodic transformations are applied, the crystal structure remains the same and the corresponding graph representation should be the same, which is periodic invariant described in Matformer~\cite{yan2022periodic}. Additionally, periodic patterns that indicate the orientations that a unit cell repeats in 3D space are also crucial for crystal structural modeling. For the crystal neural network design, due to the fact that crystal structures can have more than two hundred atoms in a unit cell, it remains challenging to consider higher body order interactions between atoms. However, higher-order interactions are arguably indispensable to achieve complete geometric representations for crystal structures. Additionally, powerful and efficient networks need to be designed for crystal structures as there could be more than two hundred atoms in the unit cell.



\begin{figure}[t]
    \centering
    \includegraphics[width=\textwidth]{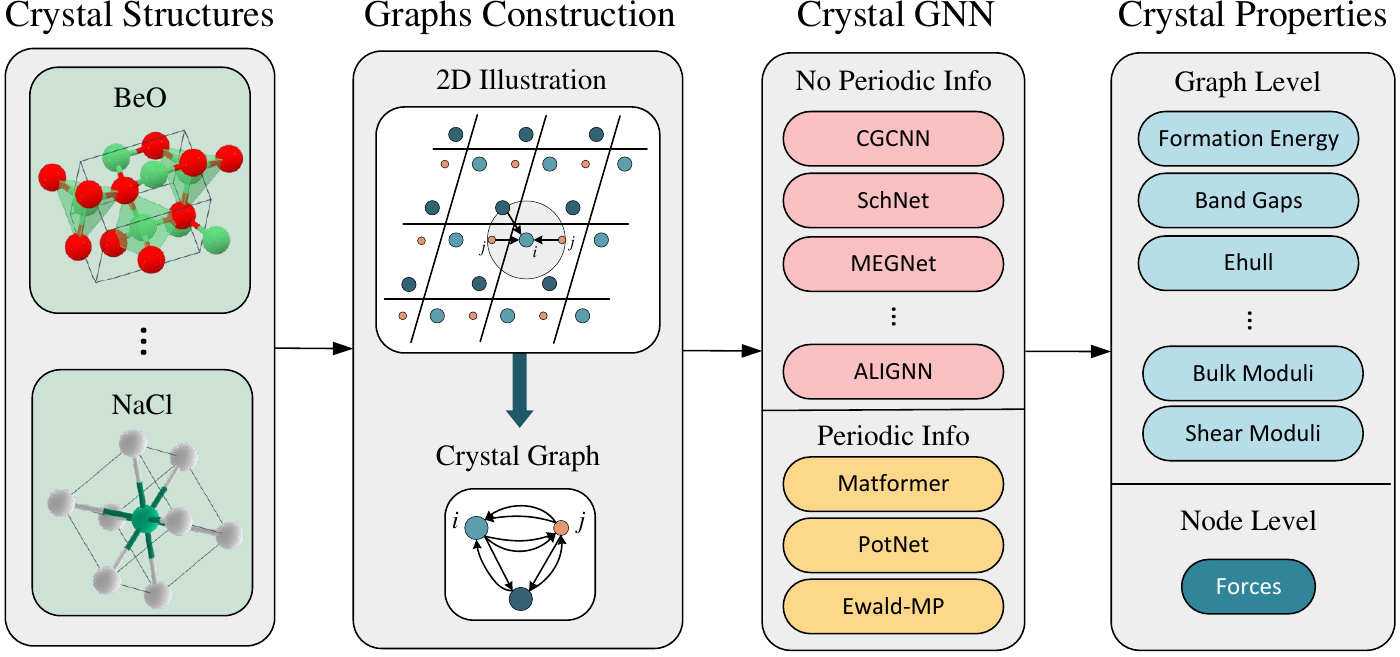}
    \caption{\ifshowname\textcolor{red}{Yuchao Lin, Keqiang Yan 40\%}\else\fi 
 Pipeline of material representation learning. A crystalline material is transformed into a crystal graph representation, subsequently serving as the input for a crystal graph message passing neural network. Then the models are trained to accurately predict the desired properties of the crystal. Notably, a message passing process can be distinguished based on its incorporation of periodic information, such as lattice lengths or infinite potential summations. Networks without periodic information include models such as CGCNN~\cite{xie2018crystal}, SchNet~\cite{schutt2018schnet}, MEGNet~\cite{chen2019graph}, and ALIGNN~\cite{choudhary2021atomistic}, among others. Conversely, networks that incorporate periodic information include Matformer~\cite{yan2022periodic}, PotNet~\cite{lin2023efficient}, and Ewald-MP~\cite{kosmala2023ewald}. Additionally, the predicted output can be classified into graph-level properties, including formation energies and band gaps, or node-level properties, which include forces.}
    \label{fig:Mat_procedure}
\end{figure}

\subsubsection{Existing Methods}

\begin{table}[t]
    \centering
    \caption{\ifshowname\textcolor{red}{Keqiang Yan}\else\fi Summary of crystal graph representations used by previous works. It can be seen that all previous works satisfy periodic invariance and achieve $E(3)$ invariance. Among these methods, Matformer~\citep{yan2022periodic}, PotNet~\citep{lin2023efficient}, and Ewald-MP~\citep{kosmala2023ewald} encode periodic patterns explicitly. CGCNN~\citep{xie2018crystal}, MEGNET~\citep{chen2019graph}, GATGNN~\citep{louis2020graph}, SchNet~\citep{schutt2018schnet}, Matformer~\citep{yan2022periodic}, and PotNet~\citep{lin2023efficient} only uses two body bond information, while ALIGNN~\citep{choudhary2021atomistic}, M3GNet~\citep{chen2022universal}, and CHGNet~\citep{deng2023chgnet} further uses three body angle information.
    }
    \begin{tabular}{l|ccccc}
    \toprule[1pt]
    Methods & Periodic Invariant & Symmetry & Periodic Pattern & Body Order & Complete \\
    \midrule
    CGCNN &\cmark& $E(3)$ invariant &\xmark&2&\xmark \\
    MEGNET &\cmark& $E(3)$ invariant &\xmark&2&\xmark \\
    GATGNN &\cmark& $E(3)$ invariant &\xmark&2&\xmark \\
    SchNet &\cmark& $E(3)$ invariant &\xmark&2&\xmark \\
    Matformer &\cmark& $E(3)$ invariant &\cmark&2&\xmark \\
    PotNet &\cmark& $E(3)$ invariant &\cmark&2&\xmark \\
    Ewald-MP & \cmark & $E(3)$ invariant & \cmark & 2 & \xmark \\
    ALIGNN &\cmark& $E(3)$ invariant &\xmark&3&\xmark \\
    M3GNet &\cmark& $E(3)$ invariant &\xmark&3&\xmark \\
    CHGNet &\cmark& $E(3)$ invariant &\xmark&3&\xmark \\
    \bottomrule[1pt]
    \end{tabular}
    \label{tab:crystal_rep}
\end{table}

\ifshowname\textcolor{red}{Keqiang Yan}\else\fi 

As shown in Figure~\ref{fig:Mat_procedure}, a typical procedure of material representation learning contains constructing crystal graph representations and employing crystal graph neural networks to predict properties. We discuss several representative crystal graph representation methods and crystal graph neural network models below. 

\vspace{0.1cm}\noindent\textbf{Crystal Graph Representations:} The periodic nature of crystals poses unique challenges for crystal representation learning as mentioned above. Specifically, to tackle challenges (1) and (2), 
CGCNN~\cite{xie2018crystal} proposes to represent the infinite crystal structure by multi-edge crystal graph built upon the unit cell structure, which has been verified to be periodic invariant by~\citet{yan2022periodic}. Concretely, multi-edge crystal graph maps a given atom and all its duplicates in 3D space as a single node, and models the interactions between node pairs by recording pairwise Euclidean distances as multiple edges. Due to the effectiveness and simplicity of the multi-edge crystal graph, it has been widely used by follow-up works, including MEGNET~\citep{chen2019graph}, GATGNN~\citep{louis2020graph}, and ALIGNN~\citep{choudhary2021atomistic}. Additionally, the use of pairwise Euclidean distances enables multi-edge crystal graph invariant to $E(3)$ transformations including rotations, translations, and reflections in 3D space. Thus, multi-edge crystal graphs are periodic invariant and $E(3)$ invariant, only considering pairwise Euclidean distances with a body order of 2. Beyond this, to enable higher body order interactions, ALIGNN proposes to further include angles between bonds without breaking periodic invariance, and M3GNet and CHGNet includes three-body interactions in a similar way. However, these methods only consider local interactions of given atoms and cannot capture periodic patterns, which are the key differences between crystal structures and molecule structures. To explicitly capture the periodic patterns, Matformer~\cite{yan2022periodic} proposes to encode three periodic vectors by using six self-connecting edges, whose combination can fully capture the lengths of periodic vectors and angles between them. Additionally, recent two works take advantage of Ewald summations to capture periodic information. PotNet~\cite{lin2023efficient} proposes to consider the infinite interactions between any two nodes in a crystal structure by using infinite summations of multiple types of potentials, and thus periodic patterns are captured by those summations. In a similar vein, Ewald-MP~\cite{kosmala2023ewald} tackles the problem of long-range interactions in periodic structures by applying the decomposition from Ewald summation. By decomposing the aggregation of message passing into a short-range signal, modeled in real space, and a long-range signal, modeled in Fourier space, periodic structure information is captured explicitly. A summary of crystal graph representations of previous works is shown in Table ~\ref{tab:crystal_rep}.




\ifshowname\textcolor{red}{Yuchao Lin}\else\fi

\vspace{0.1cm}\noindent\textbf{Crystal Graph Neural Networks:} 
The effectiveness of modern graph neural networks in predicting material properties relies on the symmetry and high-order information they incorporate as well as the used message passing fashions. Existing networks such as CGCNN~\citep{xie2018crystal}, MEGNET~\citep{chen2019graph}, GATGNN~\citep{louis2020graph}, and SchNet~\citep{schutt2018schnet} employ radius crystal graphs and consider only two-body distances as edge features during message passing, similar to the methods used in molecular representation learning. Concretely, SchNet and MEGNET use common graph convolution networks, whereas CGCNN employs the sigmoid gate operation to the concatenation of node and edge features, and GATGNN employs the attention mechanism to weight node features during message passing. Moreover, as mentioned above, Matformer~\citep{yan2022periodic}, PotNet~\citep{lin2023efficient}, and Ewald-MP~\citep{kosmala2023ewald} incorporate additional features during message passing to address the limitation of lacking periodic information of the previous methods. Specifically, Matformer explicitly encodes lattice structure information into self-connecting edges. In addition, PotNet considers infinite interatomic potential summations by summing up long-range and short-range interactions and encoding them into the edge features of fully-connected graphs. In contrast, Ewald-MP decouples the long-range interactions from the short-range message passing and combines the short and long-range node embeddings after each layer. To reduce complexity compared to fully-connected graphs, both message passings are built upon cutoff graphs, a distance cutoff in real space and a frequency cutoff in Fourier space. While most of the above methods are based on interatomic (two-body) information, incorporating three-body information can also enhance material representations, as demonstrated by ALIGNN~\citep{choudhary2021atomistic}, M3GNet~\citep{chen2022universal}, and CHGNet~\citep{deng2023chgnet}. These methods convert the crystal graph into a line graph, where the original edges become node features, and the angles between edges become edge features. They subsequently apply graph neural networks, similar to CGCNN, to learn material representations.

\begin{table}[t]
	\centering
	\caption{\ifshowname
\textcolor{red}{Keqiang Yan}\else\fi Dataset statistics of the Materials Project-2018.6.1(MP)~\citep{chen2019graph}, JARVIS~\citep{choudhary2021atomistic}, and MatBench~\citep{matbench}. The number of crystals in the largest scale tasks of corresponding datasets, number of regression tasks, and number of classification tasks are summarized.}
		\begin{tabular}{l|ccc}
			\toprule
			Datasets & Largest scale task &  \# regression tasks & \# classification tasks \\
			\midrule
			MP & 69,239 & 4 & 2 \\
			JARVIS & 55,722 & 29 & 10 \\
            MatBench & 132,752 & 10 & 3 \\
			\bottomrule
		\end{tabular}
	\label{tab:mat_dataset_stat}
\end{table}

\ifshowname\textcolor{red}{Keqiang Yan}\else\fi

\subsubsection{Datasets and Benchmarks} 
There are three widely-used crystal property prediction benchmarks as shown in Table \ref{tab:mat_dataset_stat}, including the Materials Project-2018.6.1~\citep{chen2019graph}, JARVIS~\citep{choudhary2021atomistic}, and MatBench~\citep{matbench}. The Materials Project-2018.6.1 has four widely-used regression tasks for the properties of formation energy, band gap, bulk moduli, and shear moduli. JARVIS includes 29 crystal property regression tasks and 10 crystal property classification tasks. The widely used regression tasks in JARVIS are formation energy, bandgap (OPT), bandgap (MBJ), Ehull, and total energy. MatBench consists of 10 regression tasks and 3 classification tasks. Most of the crystal properties are calculated by using Density Function Theory (DFT) based methods. 

\subsubsection{Open Research Directions} First, current deep learning based crystal property prediction methods are mainly designed for regression and classification tasks. A possible future direction would be exploring the higher rotation order crystal properties including atomic forces, dielectric tensors, \emph{etc.} Second, current works for crystal property prediction tasks are not geometrically complete, and geometric completeness for infinite crystal structures is a challenging yet important topic.

\subsection{Material Generation}
\label{sec:mat_gen}

\noindent{\emph{Authors: Youzhi Luo, Shuiwang Ji}}\newline
\newline\emph{Recommended Prerequisites: Section~\ref{sec:mat_overview}}\newline

In this section, we focus on the problem of generating crystalline materials. In this problem, a key challenge is achieving invariance to all symmetry transformations in probabilistic modeling frameworks of generative models. We elaborate the details of several existing crystalline material generation methods and their captured symmetries in the following subsections.

\subsubsection{Problem Setup}
Material generation learns a probabilistic distribution $p$ over the material space so that novel materials can be generated by sampling from $p$.

\subsubsection{Technical Challenges}

For the crystalline material generation problem, we aim to learn a probability distribution $p$ over the material space with generative models, and sample novel crystalline materials from $p$. The key challenge of this problem is incorporating all symmetries of crystalline materials into generative models. In other words, if two crystalline materials $M$ and $M'$ can be mutually transferred to each other by symmetry transformations, they should be assigned to the same probability by generative models. Particularly, the symmetry transformations of 3D material structures include not only $E(3)$ transformations that commonly exist in other chemical compounds, but also periodic transformations that are unique to crystalline materials. 
Hence, we cannot simply apply the existing 3D molecule generation methods (Section~\ref{subsec:mol_generation}) to the crystalline material generation problem because they do not consider invariance to periodic transformations. It poses significant challenges to incorporate the invariance to periodic transformations into existing $E(3)$-invariant generative models in 3D molecule generation. Also, periodic transformations do not preserve the distances between every pair of two atoms so they are not Euclidean transformations. Hence, we cannot generate 3D material structures using 3D features like distances, angles, or torsion angles as generation targets.

\subsubsection{Existing Methods} 

Generally, two strategies can be used to ensure invariance to periodic transformations. First, we can implicitly determine the atom positions in materials by generating 3D features or representations (\emph{e.g.}, 3D voxel grids) that are internally invariant to periodic transformations. Two representative methods using this strategy are Cond-DFC-VAE~\cite{court20203} and the method proposed in~\citet{hoffmann2019data}. They both convert crystalline materials to 3D voxel grids, smooth 3D voxel grids to 3D density maps, and use 3D density maps as the generation targets. Specifically, a 3D voxel grid is obtained from the 3D crystalline material by extracting the information of all atoms in a 3D cube. For any grid point in the 3D voxel grid, its voxel value is non-zero if there is an atom in its corresponding position, and non-zero voxel values contain the information of atom types. Because 3D voxel grids are usually very sparse and not suitable to serve as the direct generation targets, they are smoothed to 3D density maps where most values are non-zero. Crystalline materials are generated by first generating 3D density maps with a VAE model~\cite{kingma2014auto}, then segmented to 3D voxel grids by a U-Net model~\cite{ronneberger2015u}. 3D voxel grids or 3D density maps are invariant to periodic transformations, but not invariant to $E(3)$ transformations, so Cond-DFC-VAE and the method in~\citet{hoffmann2019data} both fail to capture $E(3)$ symmetries.

In addition, the other strategy is to directly generate the lattice matrix $L$ and coordinate matrix $C$, but tailored probabilistic modeling is needed for generative models so that for any integer matrix $K\in\mathbb{Z}^{3\times n}$, $p(C)=p(C+LK)$ always holds. Two early methods, FTCP~\cite{ren2022invertible} and the method in~\citet{kim2020generative}, propose to generate the lattice parameters and fractional coordinate matrices. Specifically, lattice parameters are $\ell_2$-norms of three lattice vectors and angles between every two lattice vectors, and for a crystalline material $M=(\bm{z},C,L)$, its fractional coordinate matrix is defined as $F=L^{-1}C$. It can be easily demonstrated that lattice parameters and fractional coordinate matrices are invariant to rotation and reflection transformations. However, fractional coordinate matrices assume an order among atoms so they are not permutation-invariant, and they are not invariant to translation and periodic transformations. Instead of directly generating fractional coordinate matrices, two recent methods, CDVAE~\cite{xie2022crystal} and SyMat~\cite{luo2023symmetryaware}, propose to generate crystalline materials in the form of 3D graphs. They use VAE models to generate the aforementioned lattice parameters, initialize atom coordinates randomly, and iteratively refine atom coordinates by score matching models~\cite{song2019generative}. Particularly, in both methods, the atom coordinates refinement is done by $E(3)$-equivariant graph neural network models on the multi-edge graph~\cite{xie2018crystal}, a 3D graph representation of the crystalline material. Since multi-edge graphs do not assume the order of atoms and are invariant to periodic transformations, their probabilistic modeling of atom coordinates refinement ensures invariance to permutation and periodic transformations. Despite these similarities, CDVAE and SyMat apply score matching to different targets in the coordinate refinement process. CDVAE directly applies score matching to atom coordinates, which fails to achieve translation-invariant. Differently, SyMat applies score matching to pairwise distances between atoms so as to achieve invariance to all $E(3)$ transformations. We summarize the key information and the captured symmetries of all crystalline material generation methods discussed in this section in Table~\ref{tab:mat_gen}.

\begin{table}[t]
	\centering
	\caption{Summary of 3D outputs, model architecture, and the captured symmetries in several representative crystalline material generation methods. Among these methods,~\citet{hoffmann2019generating} and Cond-DFC-VAE~\cite{court20203} generate 3D voxel grids, which are invariant to permutations and periodic transformations, but not invariant to rotations, reflections, and translations. FTCP~\cite{ren2022invertible} and~\citet{kim2020generative}, generate fractional coordinates and only achieves invariance to rotations and reflections. CDVAE~\cite{xie2022crystal} and SyMat~\cite{luo2023symmetryaware} generate materials in the form of multi-edge graphs. They both achieve invariance to permutations, rotations, reflections, and periodic transformations. However, CDVAE fails to achieve invariance to translations due to directly applying score matching to coordinates, while SyMat achieves it because it applies score matching to pairwise distances.}
	\resizebox{0.95\textwidth}{!}{
	\begin{tabular}{l|cccccc}
		\toprule[1pt]
		Methods & 3D Outputs & Architecture & \makecell[c]{Permutation\\invariant} & \makecell[c]{Rotation \& Reflection\\invariant} & \makecell[c]{Translation\\invariant} &
		 \makecell[c]{Periodic\\invariant} \\
		\midrule
		\citet{hoffmann2019data} & 3D voxel grids & VAE \& U-Net & \cmark & \xmark & \xmark & \cmark \\
		Cond-DFC-VAE & 3D voxel grids & VAE \& U-Net & \cmark & \xmark & \xmark & \cmark \\
		FTCP & Fractional coordinates & VAE & \xmark & \cmark &\xmark & \xmark  \\
		\citet{kim2020generative} & Fractional coordinates & GAN & \xmark & \cmark & \xmark & \xmark \\
		CDVAE & Multi-edge graphs & VAE \& Score matching & \cmark & \cmark & \xmark & \cmark \\
		SyMat & Multi-edge graphs & VAE \& Score matching & \cmark & \cmark & \cmark & \cmark \\
		\bottomrule
	\end{tabular}
}
\label{tab:mat_gen}
\end{table}

\subsubsection{Datasets and Benchmarks}

\begin{table}[t]
	\centering
	\caption{Some statistic information of Perov-5, Carbon-24, and MP-20 datasets~\cite{xie2022crystal}. We summarize the number of 3D molecule samples (\# Samples), maximum number of atoms in one molecule (Maximum \# atoms), and average number of atoms in one molecule (Average \# atoms).}
		\begin{tabular}{l|ccc}
			\toprule
			Datasets & \# Samples & Maximum \# atoms & Average \# atoms \\
			\midrule
			Perov-5 & 18,928 & 5 & 5.0 \\
			Carbon-24 & 10,153 & 24 & 9.2 \\
                MP-20 & 45,231 & 20 & 10.4 \\
			\bottomrule
		\end{tabular}
	\label{tab:mat_gen_data}
\end{table}

For a long time, there are no standard benchmark datasets for the material generation task. Early material generation methods are usually evaluated on manually selected materials from the Materials Project database~\cite{jain2013commentary} or other data libraries. Until recently,~\citet{xie2022crystal} curate three benchmark datasets Perov-5, Carbon-24, and MP-20 for the evaluation of different material generation methods. Perov-5 dataset collects 18,928 perovskite materials from an open material database for water splitting~\cite{castelli2012computational,castelli2012new}. All materials in Perov-5 have 5 atoms in a unit cell. Carbon-24 dataset collects 10,153 materials whose 3D structures are optimized by AIRSS~\cite{pickard2006high,pickard2011ab} at 10 GPa. All materials in Carbon-24 only contain carbon atoms and have up to 24 atoms in a unit cell. MP-20 dataset is composed of 45,231 materials whose energies above the hull and formation energies are smaller than 0.08 eV/atom and 2 eV/atom, respectively. All materials in MP-20 are obtained from Materials Project database and have up to 20 atoms in a unit cell. See Table~\ref{tab:mat_gen_data} for some statistical information of these three datasets.

\subsubsection{Open Research Directions}

Though several crystalline material generation methods have been proposed recently, some challenges remain unsolved and prevent them from practical use. First, recent methods including CDVAE and SyMat are based on score-matching models. They achieve better performance than earlier methods, but take much higher computational costs in refining atom coordinates for thousands of iterations by score-matching models. An efficient generative model that can generate good material samples with a reasonable time cost is desirable in real-world applications, but designing such a model remains challenging. Second, it is important to ensure that the crystalline materials generated by models are practically synthesizable. However, to our knowledge, no standard evaluation metric has been used to measure the synthesizability of crystalline materials generated by generative models in the literature. Introducing such synthesizability metrics can be useful in filtering out materials that cannot be practically synthesized, and it is also interesting yet challenging to design novel material generation methods that can optimize the synthesizability metrics of their generated materials.

\subsection{Materials Characterization}
\label{sec:mat_char}

\noindent{\emph{Authors: Elyssa F. Hofgard, Aria Mansouri Tehrani, Yuchao Lin, Shuiwang Ji, Tess Smidt}}\newline
\newline\emph{Recommended Prerequisites: Section~\ref{sec:mat_overview}}\newline

\ifshowname\textcolor{red}{Aria}\else\fi

In the last two sections, we described the ML methods for predicting materials' properties based on their crystal structures as well as for generating new crystal structures. However, perhaps a more fundamental challenge is the accurate and efficient experimental determination of crystal structures. Beyond long-range ordering, a spectrum of local disorders from short-range order in amorphous materials to correlated disorders in Prussian Blue analogs can manifest and influence materials' properties \cite{simonov2020designing, kholina2022metastable}. Therefore, to determine the exact crystal structure of a material, typically, a combination of instruments and characterization techniques such as X-ray and neutron scattering and spectroscopies are used.

The measurements are usually done in laboratories or at large-scale facilities (\emph{e.g.}, synchrotron beamlines) and require time-consuming and careful modeling of the data. To give more perspective, modern X-ray detectors can generate as many as 1,000,000 images per day, and data post-processing, analysis, and interpretation of the experiments can take over a year \cite{doucet2020machine, chen2021machine,wang2017x}. The ability of machine learning methods to process big data, and find patterns in complex data, and the computer vision algorithms for the autonomous detection of images can play a significant role in accelerating the existing workflows at beamline user facilities by providing immediate feedback during the experiments \cite{wang2017x, doucet2020machine, sullivan2019braggnet, yanxon2023artifact, wang2017x, bankoDeepLearningVisualization2021, ozer2022towards, venderley2022harnessing}. 

Here, we discuss some of the challenges and opportunities of integrating machine learning methods with characterization techniques. We specifically review the existing works on using scattering and spectroscopy techniques to predict the average and local crystal structures (and their inverse problem). We note that, due to the huge variation and complexity of the characterization methods, we only scratch the surface of possibilities.

\subsubsection{Problem Setup}
\ifshowname\textcolor{blue}{Yuchao} \& \textcolor{red}{Aria}\else\fi

As discussed in the overview, the potential applications of ML in materials characterization methods are huge due to the inherent diversity of materials characterization methods. Therefore, we focus our problem setup into two main categories, as illustrated in~\cref{fig:struct_char}:


\begin{itemize}
\item Crystal structure prediction: This category involves using the output of the experimental characterization techniques, such as the one-dimensional (1D) spectra of X-ray diffraction, to predict three-dimensional (3D) crystal structures, which can be described by atomic positions along with three lattice vectors or other crystal structure parameters such as crystal systems, Bravais lattice types, unit cell lengths, and cell angles.
\item Its Inverse Problem: The inverse of the aforementioned procedure forms the second category, whereby predicting the output of the characterization methods using the crystal structure. For example, using the crystal structure of materials, such as atomic positions and lattice vectors, to reconstruct one-dimensional (1D) and two-dimensional (2D) diffraction spectra.
\end{itemize}

\revisionOne{
\begin{figure}[t]
    \centering
    \includegraphics[width=0.9\textwidth]{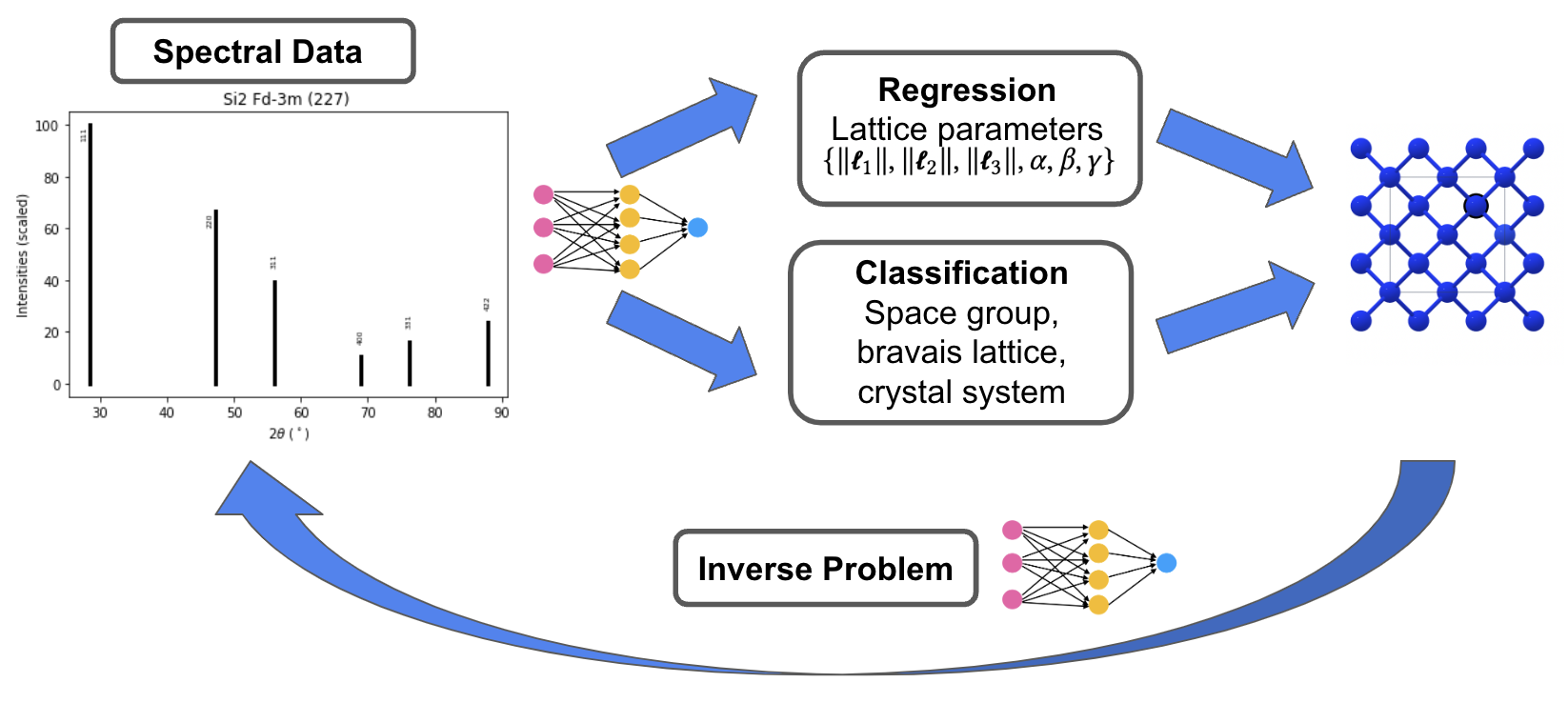}
    \caption{\ifshowname\textcolor{red}{Elyssa\%}\else\fi 
 Potential applications of ML in materials characterization methods. Given spectral data, ML can be used for crystal structure prediction. This could entail a regression task (predicting the lattice parameters or atomic positions) or a classification task (predicting space group, bravais lattice, or crystal system). In the inverse task, ML is used to reconstruct diffraction spectra given the crystal structure.}
    \label{fig:struct_char}
\end{figure}
}

\subsubsection{Technical Challenges}
\ifshowname\textcolor{red}{Elyssa}\else\fi

The process of reconstructing a 3D crystal structure or relevant parameters from a scattering pattern is not trivial. The scattering signal can be described by a complex wave function, where the amplitude represents the magnitude of the scattering and the phase represents the position and arrangement of the scattering centers. However, in practice, only the intensities (square of the amplitude) are measured or recorded. Without the phase information, a diffraction pattern may not map uniquely to the crystal structure and vice versa. This is known as the phase problem \cite{siviaNovelExperimentalProcedure1991}. 
The problem is further complicated as scattering data is obtained on a per-material basis. The experimental data can vary depending on the material, quality of the sample, and what scientific questions experimentalists are asking. Thus, there could be many modalities of data per experiment. It is thus crucial to develop ML models that generalize to experimental as well as simulated data.

As earlier stated in this section, crystals are periodic and highly symmetric structures. For example, for the inverse problem of reconstructing scattering data from crystal structures, equivariance should be preserved (\emph{e.g.}, applying an element of a crystal's symmetry group to the input will lead to the same scattering output). Thus, these problems require the design of symmetry preserving and equivariant ML methods, which is challenging.

\subsubsection{Existing Methods}
\ifshowname\textcolor{red}{Elyssa}\else\fi

\vspace{0.1cm}\noindent\textbf{Spectral Data as Input:} First, we describe existing work that provides spectral data as input to an ML algorithm. Most existing work uses this to either classify the crystal symmetry (crystal class, Bravais lattice, or space group) or to find the crystal lattice parameters in a regression problem. The structure of a crystalline material can be described by three lattice vectors and a unit cell. The unit cell can be specified according to six lattice parameters which are the lengths of the cell edges ($\Vert \bm{\ell}_1\Vert,\Vert \bm{\ell}_2\Vert,\Vert \bm{\ell}_3\Vert$) and the angles between them ($\alpha,\beta,\gamma$). In 3D space, there are seven crystal classes and 14 Bravais lattices. 

Many previous studies have used CNNs for lattice type classification or lattice parameter regression. \citet{structure_neutronscattering} develop a CNN classifier to predict Perovskite crystal systems and lattice parameters $\{\Vert \bm{\ell}_1\Vert,\Vert \bm{\ell}_2\Vert,\Vert \bm{\ell}_3\Vert,\alpha,\beta,\gamma\}$ from neutron scattering data. They then train a random forest regression model for each of the crystallographic symmetries studied to predict the lattice parameters. Both models yielded lower accuracy for lower symmetry crystal systems. They concluded that more sophisticated models were necessary to handle experimental data. \citet{chitturiAutomatedPredictionLattice2021} assume that the crystal systems are already known and use 1D CNNs to predict lattice parameters for each crystal system. The 1D CNNs were able to predict unit cell lengths but unable to predict unit cell angles of monoclinic or triclinic systems (lower symmetry classes). \citet{corrieroCrystalMELANewCrystallographic2023} also use CNNs and random forests to predict the crystal system and space group.

While CNNs may be a popular choice, \citet{greasleyExploringSupervisedMachine2023} demonstrate that more conventional supervised learning algorithms such as Support Vector Machine (SVM) and Complement Naive Bayes (CNB) classifiers performed equally as well to a neural network for multi-phase identification with experimental and simulated XRD spectra. Other groups have also employed variational autoencoder (VAE) architectures. The latent space of a VAE is a ''compressed'' representation of the training samples, so this can be an effective strategy for learning meaningful, continuous representations of materials properties from scattering data. For example, \citet{bankoDeepLearningVisualization2021} find that the latent space provides direct visual evidence of the clustering properties of the encoder model and distribution of the main reflection axes in the XRD patterns. 

However, it is crucial to note that these methods all take 1D spectra as input. A CNN for example assumes translational invariance in the input data, a feature that may not be present in these spectra. Thus, the symmetries and assumptions of ML models should be considered before applying them to powder spectra. Additionally, it would be worthwhile to develop methods that utilize equivariance in a more meaningful way.

\vspace{0.1cm}\noindent\textbf{Spectral Data as Output:} The inverse problem, such as using structures or other physical properties to predict diffraction patterns, has not been studied as extensively using ML techniques. \citet{chengDirectPredictionInelastic2023} develop an ML-based framework that can predict both one-dimensional and two-dimensional inelastic neutron scattering (INS) spectra from the structure (atomic coordinates and elemental species). This study extends work done in \citet{https://doi.org/10.1002/advs.202004214} using equivariant neural networks to predict the phonon density of states. They first employ an autoencoder to represent the 2D spectrum (a function of momentum and energy transfer) in latent space, as otherwise there would be $\sim$300 $\times$ 300 map in momentum/energy space to predict. They then use a Euclidean neural network for feature prediction in the latent space and reconstructed the 2D spectrum. Note this approach could also be applied to 1D spectra, perhaps without the autoencoder step. Equivariant neural networks thus represent a promising avenue for the inverse problem of reconstructing spectral data from crystal structures. 

\subsubsection{Datasets and Benchmarks}
\ifshowname\textcolor{red}{Elyssa}\else\fi

Using synthetic data for training X-ray/neutron scattering ML models is inevitable due to the lack of available experimental data. However, experimental data are often quite different than simulated data. Simulated datasets with structural information include the Materials Project \cite{jain2013commentary}, the Inorganic Crystal Structure Database (ICSD) \cite{dRecentDevelopmentsInorganic2019}, and the Cambridge Structural Database (CSD) \cite{groomCambridgeStructuralDatabase2016}. One approach to developing more generalizable models is to perform data augmentation on simulated datasets to address possible experimental factors such as peak shift, broadening, texture, and noisy background. Another approach is to train on simulated data but test on experimental data (an avenue that, as expected, doesn't tend to produce satisfactory results). To our knowledge, there does not exist an experimental dataset for benchmarking performance of ML algorithms for X-ray/neutron scattering due to the wide variety of experimental setups and issues investigated. For developing robust ML methods, such benchmarking datasets should be created.

\subsubsection{Open Research Directions}
\ifshowname\textcolor{red}{Elyssa}\else\fi

In the future, it would be useful to build models that employ the principles of symmetry and Fourier transforms to effectively represent scattering data in either real space, reciprocal space, or both. A constant theme in scattering experiments is the acquisition of data in reciprocal space to inform something traditionally represented in real space. Neural networks that operate in the frequency domain have been developed. \citet{li2021fourier} develop a neural network architecture defined in Fourier space and \citet{yiEdgeVaryingFourierGraph2022} extend this to perform graph convolutions in the Fourier domain. This approach could potentially be synthesized with equivariant neural networks and applied to this problem. Due to the phase problem, an equivariant neural network that understands Fourier space and can exchange information with real space could be quite powerful. 

Another future direction could be exploring different ways of representing materials through equivariant operations and how these relate to powder spectra. This could lend itself to a different equivariant ML approach. In general, current models perform worse with lower symmetry structures (classifying and predicting lattice parameters) as well as experimental data, so this should be addressed in future work. 

\subsection{Local Structure and Disordered Materials Characterization} \label{sec:local_disorder}

\noindent{\emph{Authors: Aria Mansouri Tehrani, Tuong Phung, Yuchao Lin, Shuiwang Ji, Tess Smidt}}\newline
\newline\emph{Recommended Prerequisites: Sections~\ref{sec:mat_overview},~\ref{sec:mat_char}}\newline

\ifshowname\textcolor{red}{Aria}\else\fi

The crystallographic methods we have discussed in the last section are useful for creating structural models for the average atomic positions. However, they neglect that crystalline materials can possess disorders (random or correlated). Disorder has been shown to exist and significantly influence the property of crystalline materials \cite{simonov2020designing, kholina2022metastable, venetos2023machine, cheetham2016defects}. Additionally, short-range order is even more vital in materials with limited or without long-range orders, such as nanostructures and amorphous materials. \cite{li2020predicting, martin2002designing} One approach to probe the local structure uses atomic pair distribution function (PDF) analysis, which is the Fourier transform of total scattering from X-ray, neutron, or electron scattering of powder or single crystal samples \cite{young2011applications, kjaer2023deepstruc, liu2019using}. Alternatively, spectroscopies analysis such as nuclear magnetic resonance (NMR) can provide further insight into the local structures \cite{venetos2023machine}.

\subsubsection{Problem Setup}
\ifshowname\textcolor{blue}{Yuchao}\else\fi

In this task, models are taking advantage of the results of characterization techniques that probe the local structures of materials, \emph{e.g.}, the pair distribution function (PDF) and nuclear magnetic resonance (NMR), to predict or generate the structures of materials, including three-dimensional (3D) atomic positions within the laboratory coordinate system \cite{kjaer2023deepstruc}, or chemical properties corresponding to the local atomic frame of reference, such as magnitude, anisotropy and orientation of NMR chemical shift tensor \cite{venetos2023machine}, which describe how a nucleus is influenced by the electronic environment surrounding it.

\subsubsection{Technical Challenges}
\ifshowname\textcolor{red}{Aria}\else\fi

Solving the short-range order in materials is a notoriously challenging task experimentally and computationally. For example, incorporating even small disorders such as site-sharing or point defects in DFT calculations requires expensive supercell calculations of many different configurations. These calculations are almost impossible for amorphous materials, therefore hindering the possibility of creating large training data for ML. Consequently, ML models cannot rely on standard computational data. Unfortunately, experimental characterizations of short-range order are also not trivial. Some methods to gain insights into local structures are PDF analysis of powder diffraction, 3D-$\Delta$PDF modeling of single crystal diffraction (using the Yell computer program), or merging scattering and spectroscopic techniques \cite{simonov2014yell, venetos2023machine}. The total scattering techniques require large quantities of phase pure powder samples suitable for neutron or high-quality single crystal samples, high-intensity neutron or X-ray sources at large facilities, domain and beamline experts to perform the experiments, and arduous refinements of the diffuse scattering. On the ML side, it is therefore critical to construct representations that can effectively capture the local atomic structure, develop models that are data efficient, and can predict tensorial properties.

\subsubsection{Existing Methods}

\emph{Ab initio} solving of crystal structures from atomic pair distribution function is extremely challenging and so far has only been done for highly symmetric nanostructures \cite{juhas2006ab}. Recent work has developed a deep generative model called DeepStruc, that can solve a simple monometallic nanoparticle structure directly from a PDF using a conditional variational autoencoder \cite{kjaer2023deepstruc}. PDF, also known as $G(r)$, which represents the histogram of real-space interatomic distances, is defined as 

$$\begin{aligned}
G(r) = 2/\pi \int_{Q_{\min}}^{Q_{\max}}Q[S(Q)-1]\sin(Qr)dQ,
\end{aligned}$$

where $Q$ is the scattering vector, and $S(Q)$ is the total scattering structure function that depends on the measured X-ray scattering intensities and the atomic form factor. And the structures of monometallic nanoparticles are represented as graphs, $\mathcal{G} = (X, A)$, where $X \in \mathbb{R}^{N\times 3}$. Here, $X$ is the node feature matrix, and the interatomic connections are described by the adjacency matrix $A \in \mathbb{R}^{N\times N}$. On this basis, they utilize conditional deep generative models to synthesize data conditioned on PDF by solving the unassigned distance geometry problem (uDFO) and in essence capturing the relationship between the atomic structures and PDF. Finally, they show that by using experimental data, their model can successfully predict the crystal structures of some simple monometallic nanoparticles \cite{juhas2006ab}.

Beyond scattering techniques, nuclear magnetic resonance (NMR) is a powerful spectroscopy tool that is often used in conjunction with powder X-ray diffraction to elucidate the local environment of materials. The NMR chemical shift tensor encodes both the average electronic environment of an atom represented by chemical shift as well as anisotropies that contain additional structural data. These anisotropies are evident by the line shape in an NMR measurement and can be used to infer the local chemical bondings \cite{venetos2023machine}. Exploiting the advances in equivariant geometric deep learning methods to directly predict tensorial properties while preserving the input symmetries, researchers have recently developed a model to predict Si chemical shift tensors in silicates \cite{venetos2023machine}. In this paper, the rotational equivariance is implemented using the MatTEN package, which utilizes the tensor field network and e3nn while, for comparison, symmetry-invariant models have also been constructed. The result shows that equivariant models outperform symmetry invariant ones by 53\,\%, highlighting the application of equivariant geometric deep learning models in predicting symmetry-dependent tensorial properties in characterization measurements. 

\subsubsection{Datasets and Benchmarks}
\ifshowname\textcolor{blue}{Yuchao}\else\fi

Since there is a lack of widely recognized benchmarks within this specialized field, datasets are often simulated and self-generated to develop ML models. An example is the PDF dataset~\cite{kjaer2023deepstruc}, where a total of 3,742 structures of monometallic nanoparticles were generated through the atomic simulation environment (ASE) alongside their corresponding PDF. Particularly, seven types of monometallic nanoparticle structures, such as simple cubic (sc), body-centered cubic (bcc), and face-centered cubic (fcc), among others, are included. These are constructed across a size range spanning from 5 to 200 atoms. Additionally, a subset of \emph{ab initio} NMR chemical shift tensors of relaxed structures computed by~\citet{sun2020enabling} is used in~\citet{venetos2023machine}, which comprises 421 unique silicate structures, with 1,387 unique silicon sites and different numbers of bridging oxygen atoms. These examples underline the opportunity of creating more comprehensive datasets in the future, which would help facilitate this line of research.

\subsubsection{Open Research Directions}

\ifshowname\textcolor{red}{Tuong}\else\fi

Since the local atomic environments in a material play a significant role in determining its overall properties, understanding the complex interplay between local environment geometries and material properties is crucial to designing next-generation materials with desired properties. We would like a local environment descriptor that is invariant to translation, rotation, and permutation of atoms of the same species, as these symmetry operations do not change physical properties. Additionally, recent work has introduced a metric called local prediction rigidity (LPR) to assess to what extent the global quantities can be rigorously be assigned to the local, atom-centered contributions \cite{chong2023robustness}.

We can consider spherical harmonics (detailed in \cref{sec:spherical_harmonics}) as a natural starting point for coming up with such a descriptor. Spherical harmonics are a very nice set of basis functions for signals on the sphere and are well-suited for describing local atomic environments. This is because atoms don't like to be too close to each other, so they naturally spread across a sphere. Consequently, higher degrees of spherical harmonics are not needed in order to capture a local environment due to the angular spacing of atoms. They also transform as the irreducible representations of $SO(3)$, making them invariant under rotations. Considering this, spectra, which are quantities that can be computed from spherical harmonic coefficients, seem like a natural choice to characterize geometry. 

Given a local environment, one can express it as a sum of radial Dirac delta functions as 

$$\begin{aligned}
\sum_{i=1}^{N} v_i \delta(\bm{r}_i).
\end{aligned}$$

From this function, a spherical harmonic signal $\bm{x}$ is obtained by expanding this function into its spherical harmonic coefficients as
$$\begin{aligned}
\bm{x} = a_{\ell,m} = \int \left(\sum_{i=1}^{N} v_i \delta(\bm{r}_i - \bm{x})\right) Y_{m}^\ell(\bm{x}) d\bm{x} = \sum_{i=1}^{N} v_i Y_{m}^\ell(\bm{r}_i).
\end{aligned}$$
Taking this spherical harmonic signal $x$, one can calculate the spectra of order $d$ by computing repeated symmetric tensor tensor products ($x^{\otimes (d+1)}$) and extracting the scalar and pseudoscalar coefficients ($\bm{x}^{\otimes (d+1)} \longrightarrow (0, e) \oplus (0, o)$). The first, second, and third order spectra are more commonly known as the power spectrum, bispectrum, and trispectrum, respectively. The bispectrum in particular is effective in characterizing local environments, being more expressive than the power spectrum but less computationally expensive than the trispectrum. 

Spectra also have other nice properties including smoothness (small perturbations to the original geometry lead to small perturbations in the resulting spectra), invertibility (the original geometry can be decoded up to a global rotation), and being fixed-length (for a given value of $l$). They can also be clustered (\emph{e.g.}, using k-means clustering), enabling the identification of geometric trends within a given class of materials or across different material classes.

Beyond further developing appropriate, expressive local representations, an essential avenue moving forward is to utilize the power of ML to accelerate the time-consuming characterization processes of disordered materials. For example, computer vision can be useful for the rapid detection of Bragg peak shapes, making corrections, and identifying artifacts, among others, while in some cases, the reliability of equivariant neural networks to preserve symmetry can be exploited to incorporate domain knowledge.

\subsection{Phonon Calculations} \label{sec:phonon}

\noindent{\emph{Authors: Adriana Ladera, Tess Smidt}}\newline
\newline\emph{Recommended Prerequisites: Section~\ref{sec:mat_overview}}\newline

A phonon is the quantization of energy of vibrations in a lattice, analogous to photons being the quantization of the electromagnetic wave \cite{Kittel2004}. Calculations of these phonons are crucial to understanding the thermal and dynamical properties of materials. First principles phonon calculations have become especially available due to advances in efficient density functional theory (DFT) codes and high-performance computing,  but experimental and computational challenges in efficient phonon calculations still remain. In this section, we detail current obstacles (force prediction accuracy, limited resources, periodicity of crystallographic materials complicating training data and the learning of current models), existing methods (equivariant neural networks for direct prediction of phonon density of states,  moment tensor potentials trained on \emph{ab initio} molecular dynamics trajectories), and promising future directions for the integration of ML in phonon studies.

\subsubsection{Problem Setup}

In phonon calculations, crystal structures, including atomic positions, atomic species, and lattice vectors of the unit cell, are given to calculate phonon properties, which comprise computations such as the phonon dispersion relation (PDR) (commonly known as the phonon band structure), and the phonon density of states (PDOS). PDR relates the phonon momentum, normally along a high symmetry path in the Brillouin zone, and the angular frequency of the phonon in each branch. It is significant for studying phonon-related properties, such as phonon-phonon interactions and electron-phonon coupling, with applications in thermal and electronic transport. PDOS summarizes phonon dispersions by integrating over the wave vector and summing over each branch. Formally, PDOS is defined as
\begin{equation}
    g(\omega)=\frac{1}{N}\sum_{q_j}\delta(\omega-\omega_{q_j}),
\end{equation}
where $N$ is the number of unit cells in the crystal, $q$ is the wave vector $j$ is the band index, and $\omega_{q_j}$ is the phonon frequency \citet{TOGO20151}. PDOS is important in understanding several material properties, such as superconductivity, electrical transport, and vibrational properties. When predicting PDOS, PDR, and other phononic calculations, graph neural network models often have no prior knowledge of interatomic forces and other characteristics are used, other than atomic positions, masses, and atomic species, whereas in ML interatomic potentials (MLIPs), the models are trained on ab initio molecular dynamics trajectories.

\subsubsection{Technical Challenges}

For studying thermal properties and PDR, DFT simulations offer accurate approximations, but the computational cost quickly increases in the case of nanoporous and low-symmetry materials. Lower k-point grids and smaller supercells and plane-wave
cutoff energy are often to bypass computational constraints, but unsurprisingly resulting in the poor accuracy of the yielded PDR. Additionally, computational conditions can still produce nonphysical frequencies in phonon dispersion diagrams \cite{MORTAZAVI2020100685}. Obtaining PDOS via both experimental and computational methods poses a challenge, as ab initio calculations for complex materials demand high computational costs and inelastic scattering often requires limited resources \cite{https://doi.org/10.1002/advs.202004214} such as high flux neutron sources or synchrotron X-rays \cite{10.1063/5.0055593}. 

Several recent advances in machine learning for material science suggest a new paradigm for materials studies. However, the 3D nature of atomic systems and the periodicity and symmetry of crystallographic materials further complicate any potential learning of PDOS for a regular neural network, as this requires expensive data augmentation to learn different rotations and translations of the 3D coordinate systems. These problems therefore highlight the need for a more efficient  strategy for obtaining PDOS. Turning towards ML methods, GNNs represent atoms and their bonds as graph nodes and edges, respectively, offering a natural representation of atomic systems.  Symmetry-augmented GNNs \cite{geiger2022e3nn} additionally, hold an advantage due to the innate symmetry of crystal materials. Challenges in representation and neural network design choices, still persist, however, such as the difference between properties in real space and reciprocal space, and the fixed length of output properties (\emph{i.e.}, scalar outputs) \cite{PhysRevB.80.184302}. This is problematic for materials properties with varying degrees of dimensions, such as the number of phononic bands \cite{RevModPhys.73.515}.

\subsubsection{Existing Methods}

In this section, we categorize approaches for efficient and accurate prediction and analysis of phononic properties of materials in terms of using local environment descriptors and geometric graph neural networks.

As a prime example of local environment descriptors, \citet{MORTAZAVI2020100685} utilize ML interatomic potentials (MLIPs) to train on computationally efficient \emph{ab initio} molecular dynamics trajectories, providing an alternative and efficient method to DFT simulations. In MLIPs, the potential energy surface is described as a function of the local environment descriptors which are invariant to rotations, translations, and inversions of homonuclear atoms \cite{10.1063/1.4966192}. Under the umbrella of MLIPs are moment tensor potentials (MTPs) \cite{Zuo2020}, which can approximate any interatomic interactions \cite{MORTAZAVI2020100685}, therefore able to evaluate phononic properties comparable to density functional perturbation theory methods without being computationally expensive. In addition, \citet{LADYGIN2020109333} use MTPs to reproduce phonon properties with high accuracy compared to DFT-obtained data. Using active learning of MTPs as an advantage, training is conducted on molecular dynamics runs of Al, Mo, Ti, and U.  The active learning approach automatically
fits the MTP only on configurations in which there is significant extrapolation in data, which greatly reduces the number of DFT calculations
required for training the MTP. The error between MTP and DFT results for PDOS of Al and U and phonon dispersion diagrams of Mo and Ti is far smaller than the error between DFT and experimental data. Similar errors are produced when comparing MTP and DFT for vibrational free energy and entropy results. 

The geometric graph neural network learns material properties from 3D atomic positions, atomic species, and interatomic distances based on graph neural networks. For instance, \citet{okabe2023virtual} propose to augment GNNs with their Virtual Node Graph Neural Network (VGNN), which manages output properties with variable or even arbitrary dimensions. \citet{okabe2023virtual} present three versions of VGNN, each implemented with the symmetry-aware graph  Euclidean convolutional 
neural networks \cite{geiger2022e3nn}.
With the atomic positions and masses represented as a periodic graph, this input is passed through a series of
convolution layers, which compute the tensor product of the input features, separated by nonlinear layers, which introduce the complexity to the model. Vector virtual nodes (VVN) is the simplest model with $m$-atom crystal structure input and $3\times m$ $\Gamma$-phonon energies. The matrix virtual nodes (MVN) are slightly more accurate and computationally expensive for complex materials. Lastly, the momentum-dependent matrix virtual nodes (k-MVN) is the most complex and, given random $k$-points in the Brillouin zone, is able to predict the entire phonon band structure. This is done with virtual dynamical matrices 14 (matrices analogous to the phonon dynamical matrices) in the crystal graphs. Training data of the materials then enables the matrix elements to be learned from optimizing the neural network. Each of the listed models has no prior knowledge of interatomic forces and only takes the crystal structure as input. This methodology provides a computationally feasible strategy for obtaining full phonon band structures from the crystal structures of complex materials. \citet{https://doi.org/10.1002/advs.202004214} capture the main features of PDOS using a Euclidean neural network in 3 dimensions ($E(3)$NN), as implemented by the $E(3)$NN open-source repository \cite{geiger2022e3nn}. The $E(3)$NN is equivariant to 3D rotations, translations, and inversion, and therefore is able to preserve all geometric inputs as well as their crystallographic symmetries. The dataset is a phonon database of 1,521 crystallographic semiconductor compounds, containing PDOS data-based density functional perturbation theory \cite{Petretto2018}. $E(3)$NN successfully reproduces key features in experimental data and predicts a broad number of high phononic specific heat capacity materials directly from atomic structure without being computationally expensive.

\subsubsection{Datasets and Benchmarks}

To the best of our knowledge, there does not exist a standard database for phonon calculations that are explicitly for ML methods. From the representative works, however, such datasets include a $\Gamma$-phonon database with approximately 146,000 materials from the Materials Project \cite{okabe2023virtual} and a database of the PDOS data based on DPFT from 1,521 semiconductor compounds \cite{Petretto2018}. However, the onset of ML applications in phonon calculations demands a standard database for material phonon data from which ML methods could be easily applied.

\subsubsection{Open Research Directions}

While the main focus of lattice dynamics studies in the work of \citet{LADYGIN2020109333} are single-component systems, there is promising work in multi-component systems as well. Specifically, MTP performs well in covalently and ionically bonded systems \cite{Grabowski2019} as well as metallic systems \cite{NOVIKOV201974}. Additionally, the Euclidean neural network workflow of \citet{https://doi.org/10.1002/advs.202004214} highlights a framework that could aid in high-throughput screening in the search for promising thermal materials candidates, while drawing connections between the structural symmetry of materials and their phononic properties.

\subsection{\revisionOne{Amorphous Materials}} \label{sec:amorphous}

\noindent{\emph{Authors: Alex Strasser, Keqiang Yan, Xiaofeng Qian}}\newline
\newline\emph{Recommended Prerequisites: Section~\ref{sec:mat_overview}}\newline

\revisionOne{Unlike crystalline materials, which hold both short-range and long-range order, amorphous materials lack long-range order due to the absence of translational invariance, thus requiring different computational approaches to study their properties. 
Interestingly, amorphous materials often hold short-range order in their hidden inherent structures, for examples, in liquids, glasses, and metallic glasses~\cite{Stillinger1982hiddernstructure,debenedetti2001supercooled,sheng2006atomic,tanaka2019revealing}. 
Amorphous materials have many applications in batteries, phase-change materials, photocatalysis, nonlinear optics, solar cells, and more. However, due to their non-equilibrium structure and lack of long-range order, structure determination, characterization, and synthesis all become more pressing challenges~\cite{liu2024amorphous}.}

\subsubsection{Technical Approaches and Challenges}

\revisionOne{Molecular dynamics (MD) simulations, based on \emph{ab initio} theory or classical force fields, are commonly used to model amorphous materials by tracking molecular motion over time. This is achieved by calculating the atomic forces and stresses and iteratively integrating Newton's equations of motion. In MD simulations of amorphous systems, the process usually begins with a liquid-state configuration, which is then rapidly quenched to an amorphous state, called a melt-quench simulation. Monte Carlo (MC) simulations provide an alternative approach to study amorphous materials and, combined with MD simulations to sample inherent structures. To address the challenges of structure determination and characterization, computational techniques are often combined with experiment. For example, reverse Monte Carlo (RMC) method can help extract local atomic structure by using experimental metrics such as the radial distribution function. Since these structures may not be unique, refinement techniques, such as force-enhanced atomic refinement~\cite{madanchi2024future}, are often employed for improvement.}

\revisionOne{Computational methods such as special quasirandom structures (SQS) have been developed to construct a small set of configurations that statistically approximate the true random distribution of atoms or defects in 
disordered or alloys~\cite{Zunger1990SQS,Wei1990SQS}. Effective Hamiltonian can be fit via a cluster expansion method, which can then be used for subsequent simulations of amorphous materials using computational packages such as the Alloy Theoretic Automated Toolkit (ATAT)~\cite{vandeWalle2002ATAT} and the Clusters Approach to Statistical Mechanics (CASM) code ~\cite{Puchala2023CASM}.}

\subsubsection{Existing Methods}

\revisionOne{Recently, machine learning approaches have been applied to study amorphous systems~\cite{reiser2022graph}. Here we highlight a few examples, though this list is by no means exhaustive. }

\revisionOne{As mentioned above, MD is a prominent tool for modeling amorphous materials, where machine learning force fields  become critically important for the study of amorphous materials.} 

\revisionOne{
In the last decade, MLIPs such as Gaussian approximation potential have been developed for amorphous carbon~\cite{Deringer2017MLIPamorph} or amorphous silicon~\cite{Deringer2018MLIPSi}.
Very recently, graph neural networks (GNNs) have been applied to investigate amorphous materials~\cite{bapst2020unveiling, wang2021inverse, aykol2023predicting, tao2024artificial}. \citet{bapst2020unveiling} showed that GNN can efficiently and accurately predict the long-time dynamics and propensity map and extract a remarkable structural length scale for dynamics in glassy systems. \citet{wang2021inverse} established an inverse design approach by integrating GNN with swap Monte Carlo method to efficiently sample the glass landscape and predict the associated plastic resistance, which reveals a geometrically ultrastable but energetically metastable state. }

\revisionOne{\citet{boattini2020autonomously} utilized unsupervised machine learning approach to uncover structural heterogeneities in supercooled liquids without relying on dynamical information, which enables the identification of hidden local structures in supercooled liquids. They encoded the local environment of structural snapshots in bond-order parameters and then applied a neural network autoencoder combined with Gaussian mixture models. The resulting structural order parameter obtained from machine learning enables the detection of the largest structural variation in supercooled liquids, indicating the great potential of ML approach for identifying nontrivial structural fingerprints for amorphous materials.}

\revisionOne{In addition, generative models have been recently used to generate a wide range of structures up to an arbitrary size, which was first done in 2D~\cite{kilgour2020generating} and then extended to 3D disordered systems~\cite{madanchi2024simulations3D} and compositionally complex bulk metallic glasses~\cite{zhou2023generative}. This is possible by modeling a context ``kernel'' of a correlation length that represents the longest static structural correlations, and then extrapolating from the structure contained within the correlation length to arbitrary size.}

\subsubsection{Open Directions}

\revisionOne{While the above examples demonstrated the potentials of AI/ML approaches, there are many opportunities to explore further. For example, it will be highly desirable to develop more advanced inverse design approaches to enable generation of novel amorphous materials with desired physical properties. Recently, several diffusion models and large language models (LLMs) have shown great promise for crystalline materials generation~\cite{xie2022CDVAE,jiao2023DiffCSP,antunes2024crystallm,yan2024mat2seq,zeni2025mattergen}. However, extending these methods to amorphous materials with large supercells remains largely unexplored. Furthermore, developing more accurate GNN models for large amorphous materials requires more efficient sampling techniques and overcoming computational bottlenecks when generating ground-truth data from DFT calculations. Finally, how to make GNN models both computation and memory efficient is another challenge, which is crucial for studying large-scale long-time dynamics across multiple time and length scales and understanding short-range order and potentially intermediate-range order in different amorphous materials.}

\clearpage
\hypertarget{AI for Molecular Interactions}{\section{AI for Molecular
Interactions}} \label{sec:dock}

As described in Sections~\ref{sec:mol},~\ref{sec:prot}, and~\ref{sec:mat}, AI has revolutionized the field of molecular learning, protein science, and material science. While AI for individual molecules has been extensively studied, the physical and biological functions of molecules are often driven by their interactions with other molecules. In this section, we further introduce AI for molecular interactions, where we particularly consider the interactions between small molecules and proteins or materials.

\subsection{Overview}

\noindent{\emph{Authors: Meng Liu, Shuiwang Ji}}\newline


This research area focuses on using AI to gain insights into the molecular mechanisms that govern interactions between small molecules and other substances, which has great potential in advancing our understanding of molecular interactions and providing practical solutions for a wide range of challenges in life science and material science.

For both molecule-protein and molecule-material interactions, as illustrated in Figure~\ref{fig:openmi_categorization}, we categorize existing tasks into predictive tasks and generative tasks. Note that our categorization is based on the nature of the tasks, rather than the methods employed to perform them. To be specific, the predictive task related to the interaction of small molecules and proteins is binding (or docking) prediction. Binding refers to the process by which a small molecule, namely as a ligand, binds to a target protein based on their native shape complementary and chemical interactions~\cite{fischer1894einfluss}, also known as the ``lock-key'' model. The target protein is usually associated with human disease. The drug molecule binds to the target protein to inhibit or activate it to treat human diseases. This task includes binding pose prediction and binding affinity prediction. On the other hand, the generative task for such interaction is to generate molecules that can bind to given target proteins, known as structure-based drug design~\cite{anderson2003process}. Both protein-ligand binding prediction and structure-based drug design are fundamental and challenging problems in drug discovery. 
In terms of the interaction of small molecules and materials, to our knowledge, only predictive tasks have been investigated in existing works. \revisionOne{Specifically, we are interested in predicting total system energy (S2E) and per-atom force (S2F) from a molecule-material pair structure. These values can be used to compute specific properties such as adsorption energy and transition state energy.} In addition, given the initial structure of a molecule-material pair, it is highly desired to predict the relaxed final structure (IS2RS) and the energy at its relaxed state (IS2RE). These tasks are critical for many problems in material science, such as electrocatalyst design for renewable energy storage~\cite{zitnick2020introduction}. The generative tasks for molecule-material interactions remain unexplored, and we discuss the potential opportunities in Section~\ref{subsubsec:mol_mat_future}. An overview of the methods covered in this section is shown in Figure~\ref{fig:mi_overview}.


Ensuring the preservation of desirable symmetry in 3D space is a crucial aspect of molecular interaction tasks. This unique symmetry is distinct from the consideration of a single molecule, as it encompasses multiple instances involved in the interaction. In particular, it is essential to take into account the symmetry properties of each molecule within the context of the entire interaction system. 

\begin{figure}[t]
    \centering
    \includegraphics[width=\textwidth]{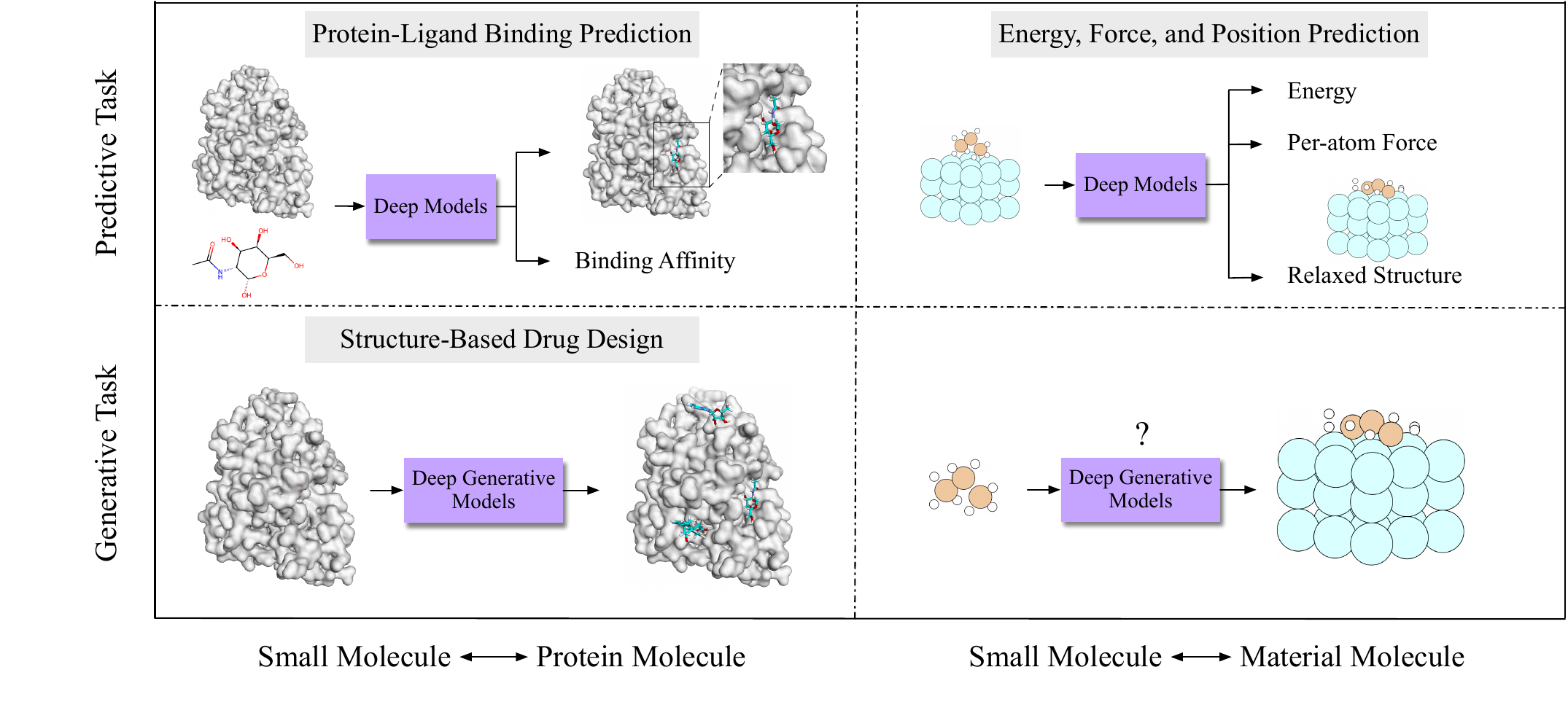}
    \caption{An illustration of our covered tasks in molecular interactions. The bidirectional arrows represent interactions. For both molecule-protein interaction and molecule-material interaction, we categorize existing tasks into predictive tasks and generative tasks. In protein-ligand binding prediction, we aim to predict the binding pose of the ligand and the strength of the binding, namely binding affinity. In structure-based drug design, it is desired to generate 3D ligand molecules that can bind to the given target proteins. In the predictive tasks of molecule-material pairs, we are interested in predicting the \revisionOne{energy} and per-atom force for given molecule-material pair structures. In addition, it is of interest to predict the relaxed final structure with minimum energy given the initial structure as input. The generative tasks for molecule-material interactions are unexplored and we discuss the possible directions in Section~\ref{subsubsec:mol_mat_future}}. 
    \label{fig:openmi_categorization}
\end{figure}

\begin{figure}[t]
    \centering
\includegraphics[width=\textwidth]{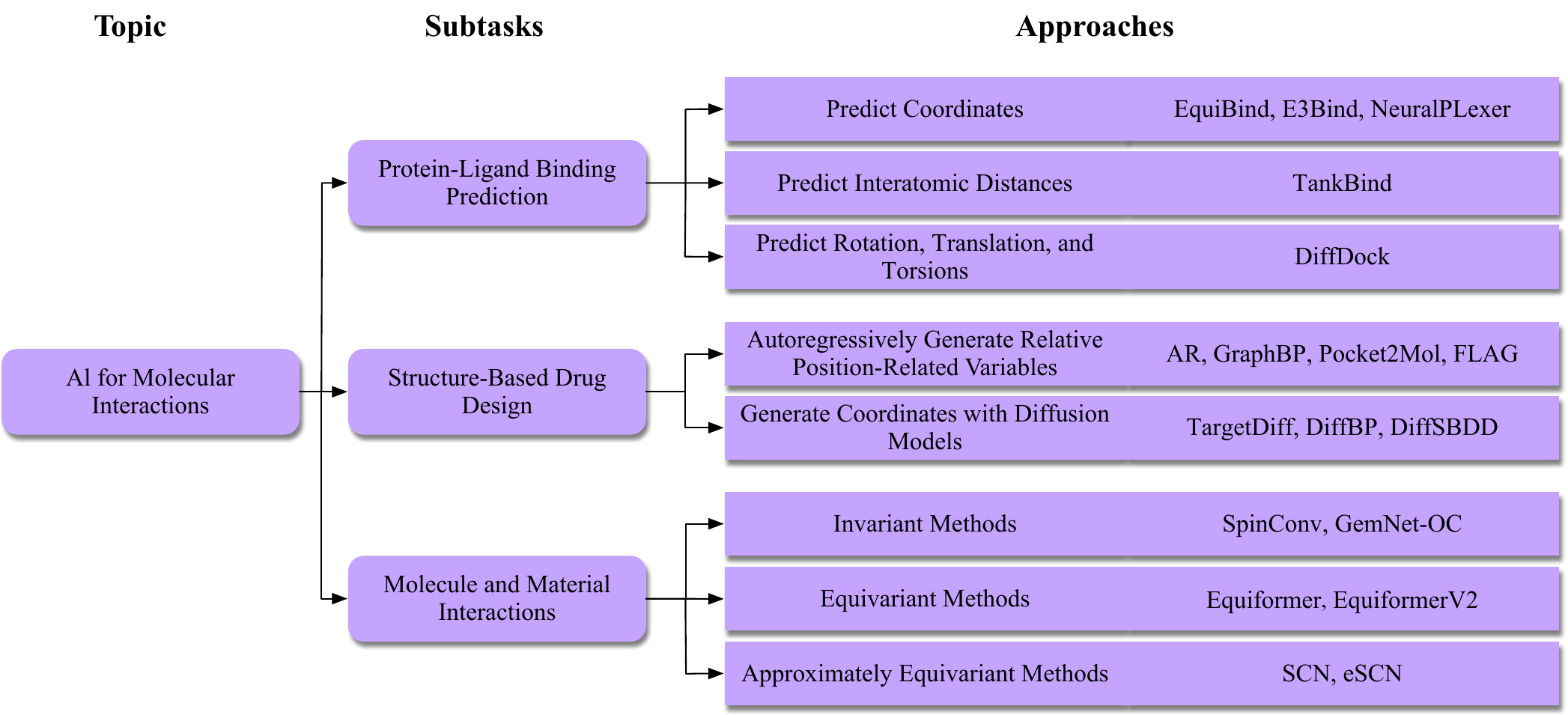}
\caption{An overview of the tasks and methods in AI for molecular interactions. This section considers three tasks, including protein-ligand binding prediction, structure-based drug design, and energy, force, and position prediction for molecule-material pairs. In protein-ligand binding prediction, one category of methods, including EquiBind~\cite{stark2022equibind}, E3Bind~\cite{zhang2023ebind}, and NeuralPLexer~\cite{qiao2023statespecific}, aims to directly predict the 3D coordinates of ligands. In comparison, TankBind~\cite{lu2022tankbind} predicts the interatomic
distances between protein segments and ligands. Besides, DiffDock~\cite{corso2022diffdock} generates the rotation, translation, and torsion angles of seed conformers given by RDKit. In structure-based drug design, one category of methods, including AR~\cite{luo20213d}, GraphBP~\cite{liu2022generating}, Pocket2Mol~\cite{peng2022pocket2mol}, and FLAG~\cite{zhang2023molecule}, aims to generate ligand atoms/fragments autoregressively by modeling their relative position-related variables. Another category of methods, including TargetDiff~\cite{guan2023targetdiff}, DiffBP~\cite{lin2022diffbp}, and DiffSBDD~\cite{schneuing2022structure}, considers generating 3D coordinates of all ligand atoms directly via diffusion models. 
In the prediction tasks for molecule-material pairs, the invariant methods are SpinConv~\cite{shuaibi2021rotation} and GemNet-OC~\cite{gasteiger2022gemnetoc}, and the equivariant methods include Equiformer~\cite{liao2023equiformer} and EquiformerV2~\citep{liao2023equiformerv2}. Besides, SCN~\cite{zitnick2022spherical} and eSCN~\cite{passaro2023reducing} are approximately equivariant methods.} 

\label{fig:mi_overview}
\end{figure}







\subsection{Protein-Ligand Binding Prediction}
\label{sec:openmi_binding}

\noindent{\emph{Authors: 
Hannes St{\"a}rk, Yuchao Lin, Shuiwang Ji, Regina Barzilay, Tommi Jaakkola}\vspace{0.3cm}\newline\emph{Recommended Prerequisites: Sections~\ref{subsec:mol_representation_learning}, \ref{sec_prot_pred}}\newline

In this section, we study the protein-ligand binding problem. We aim to make inferences about how a small molecule, potentially a drug, interacts with a protein. For this purpose, we discuss molecular docking and binding affinity prediction, both of which are important for fields such as drug discovery or molecular biology.

\subsubsection{Problem Setup}

In docking, we are given a protein structure (its amino acid identities and atom coordinates) and the molecular graph of a small molecule (ligand). The goal is to predict the atom positions with which the ligand most likely binds to the protein. This task can be divided into the scenario where the docking location (pocket) is approximately known and the blind docking scenario without any prior knowledge. Performing well at either means having a high fraction of approximately correct predictions. Meanwhile, with binding strength prediction, we refer to predicting a ranking or scalar binding affinity that indicates the strength with which a ligand binds to a protein; that is, roughly the fraction of times the ligand and protein can be observed in a bound vs. unbound state. The inputs for this could, \emph{e.g.}, be the bound protein-ligand structure or a protein structure and the ligand's molecular graph. 

In both docking and binding strength prediction, it is important to consider the discussed nuances of the problem setup. Additionally, in both tasks, it is relevant whether one has access to the protein's structure when bound to the ligand (holo-structure), the unbound protein structure (apo-structure), the structure of the protein bound to another ligand, or only a computationally generated structure from, \emph{e.g.}, AlphaFold2. In docking evaluations, knowledge of the bound structure is often assumed, which is unrealistic for application purposes, and comparisons in the other scenarios are desirable. 

To present docking and binding strength prediction in a formal context, the structure of a ligand can be expressed as $\mathcal{M}=(A, E, C)$, where $A=[\bm{a}_1, \cdots, \bm{a}_n]$ refers to the atomic properties, for instance, atomic types, of all $n$ atoms contained in the molecule. The edge features, which could include bond types and bond lengths, are denoted by $E = [\bm{e}_1, \cdots, \bm{e}_l]$ for all $l$ chemical bonds in the molecule. Meanwhile, the 3D coordinate matrix is expressed as $C=[\bm{c}_1,\cdots, \bm{c}_n] \in \mathbb{R}^{3 \times n}$. Similarly, the structure of a protein or a known pocket can be represented as $\mathcal{P}=(B, S)$. Here, $B=[\bm{b}_1, \cdots, \bm{b}_m]$ represents the node features, including either amino acid types or atomic types, of all the amino acids or atoms within the designated protein or pocket, depending on the levels of granularity as detailed in Section~\ref{sec:prot}.
Additionally, the 3D coordinates corresponding to either the alpha-carbons of amino acids or the atoms within the structure are symbolized by $S=[\bm{s}_1,\cdots, \bm{s}_m] \in \mathbb{R}^{3\times m}$. 
The primary goal in docking is to sample from the distribution of docking poses based on the protein/pocket structure, which can be represented by $p\textsubscript{pose}(C|B, S, A, E)$. Methods can either directly model the distribution $p\textsubscript{pose}(C|B, S, A, E)$ and then sample, or predict $k$ geometries of ligand conformers such that $f\textsubscript{pose}(B,S,A,E)\mapsto [C_1, \cdots, C_k]$. On the other hand, the objective of binding strength prediction is to estimate the binding affinity, to provide a ranking, or to make a binary binding vs. non-binding prediction. We unify these targets as $q$ where $q\in \mathbb{R}$ or $q\in \mathbb{Z}^+$ or $q\in [0, 1]$, and the task is represented as $f\textsubscript{strength}(A, E, C, B, S)\mapsto q$. Note that for binding strength prediction, the geometric information of ligands or the protein/pocket may not be given.

\subsubsection{Technical Challenges}
In our description of molecular docking, we strive to predict the most likely binding pose of a ligand. This is only the global mode of the Boltzmann distribution that describes the probability of each possible ligand pose conditioned on the protein. Ideally, one would want a generative model that replicates this high-dimensional distribution with sparse support, which is especially challenging considering that there usually only is data for one of its modes. Producing the lowest energy (\emph{i.e.}, highest probability) pose is already a difficult problem, given the large space of plausible ligand configurations.

A challenge for docking compared to protein structure prediction is that docking methods cannot rely on vast amounts of sequence data for evolutionary information to constrain the set of plausible structures which partially explains the success of protein structure prediction before geometric deep learning had large impacts on molecular docking. Data concerns also pose another technical challenge; the amount of easily accessible training data with reasonable quality is $\sim$20k samples. While there are more complexes in the PDB, it requires expert knowledge and is hard to clean this data from, \emph{e.g.}, complexes that are only spurious interactions with very low affinities. 

Similar data concerns impede progress in binding strength prediction with few ($\sim$20k) data points for 3D structures and noisy measurements (also sequence-based data) preventing a successful affinity predictor for general protein-ligand combinations. However, an already currently useful strategy is training protein-specific predictors that only take ligands as input (under the condition that sufficient binding affinity data is available for the specific protein or can be gathered in an active learning setup). A potential direction towards general protein-ligand affinity predictions could be using geometric deep learning to aid or approximate statistical mechanics methods for calculating binding free energies. This, again, is hampered by the difficulty of modeling the Boltzmann distribution and partitions thereof.

Lastly, we discuss the symmetries involved in the two tasks. Docking is an $SE(3)$-equivariant task; rotating or translating the input protein structure should result in a corresponding rotation and translation of the generated ligand poses. Technically, for a roto-translation $g \in SE(3)$ and its corresponding group action $\rhd$, this task requires $p\textsubscript{pose}(g\rhd C |B, g\rhd S, A, E) = p\textsubscript{pose}(C|B, S, A, E)$ and $f\textsubscript{pose}(B, g \rhd S, A, E)=g\rhd f\textsubscript{pose}(B, S, A, E)$. In contrast, binding strength prediction is an $SE(3)$-invariant task as rotation and translation to the system do not affect the prediction. Formally, for a roto-translation $g \in SE(3)$, it is desired that $f\textsubscript{strength}(A, E, g\rhd C, B, g\rhd S)= f\textsubscript{strength}(A, E, C, B, S)$.

\begin{table}[t]
	\centering
	\caption{Summary of deep learning methods for blind protein-ligand docking w.r.t. symmetry. EquiBind~\cite{stark2022equibind} and E3Bind~\cite{zhang2023ebind} apply $E(3)$-equivariant networks to predict ligand coordinates. TankBind~\cite{lu2022tankbind} estimates interatomic distances between protein pocket candidates and the ligand together with an affinity for each candidate. DiffDock~\cite{corso2022diffdock} employs an $SE(3)$-equivariant diffusion model over rotations, translations, and torsions to sample candidates before ranking them. NeuralPLexer~\cite{qiao2023statespecific} predicts a contact map and applies it to an $E(3)$-equivariant diffusion model to generate ligand coordinates.}
	\begin{tabular}{l|ccc}
		\toprule[1pt]
		Methods & Outputs & Architecture & Network Symmetry \\
		\midrule
		EquiBind & Coordinates & Regression-Based & $E(3)$-Equivariant \\
            E3Bind & Coordinates & Regression-Based & $E(3)$-Equivariant \\
            TankBind & Interatomic Distances & Regression-Based & $E(3)$-Invariant\\
            DiffDock & Rotation/Translation/Torsions & Generative & $SE(3)$-Equivariant \\
            NeuralPLexer & Coordinates & Generative & $E(3)$-Equivariant \\
		\bottomrule
	\end{tabular}
	\label{tab:binding_prediction}
\end{table}

\subsubsection
{Existing Methods} 
We categorize the protein-ligand docking models into three distinct types: traditional search-based docking, regression-based docking, and generative docking. In order to highlight the recent progress of deep learning made in the field of blind protein-ligand docking, we provide an outline of blind protein-ligand docking methods, specifically with regard to their treatment of symmetry in Table~\ref{tab:binding_prediction}.

\vspace{0.1cm}\noindent\textbf{Traditional Search-Based Docking:} Traditional approaches employ an $E(3)$-invariant scoring function that assigns likelihoods to ligand poses together with an optimization algorithm to find the scoring function's global minimum, \emph{i.e.}, the most likely pose \cite{trott2010autodock, koes2013smina, halgren2004glide}. The most common scoring functions consist of physics-inspired terms of invariant quantities, such as interatomic distances, and use very few learned parameters. More recently, there have been deep learning parameterizations that employ, \emph{e.g.}, 3D CNNs \cite{mcnutt2021gnina}. Importantly, most of these methods are developed for docking to known pockets (with exceptions \citep{Hassan2017QVinaW}) and struggle with the larger search space in blind docking, leading to long inference times. Furthermore, their scoring functions are sensitive to deviations of the input protein from the bound structure \citep{corso2022diffdock, karelina2023AlphaFoldDocking}, which limits their ability for docking to computationally generated or unbound protein structures.

\vspace{0.1cm}\noindent\textbf{Regression-Based Docking:} More recent deep learning methods significantly speed up blind docking by directly predicting ligand binding poses with $E(3)$-equivariant/invariant GNN instead of parameterizing a scoring function for a search algorithm. Of the regression-based approaches, EquiBind~\citep{stark2022equibind} produces its prediction by finding key points in the protein that characterize the binding pocket and superimposing the ligand with them. Meanwhile, Tankbind~\citep{lu2022tankbind} splits the protein into pocket candidates and predicts protein-ligand distances and an affinity score for each of them. E3Bind~\citep{zhang2023ebind} uses the same pocket candidates but produces pair representations for them which are iteratively decoded by updating initial ligand coordinates. A disadvantage of these regression-based methods is that they are forced to predict a single pose even though multiple configurations are plausible and could have a significant likelihood under the Boltzmann distribution. This often leads to unphysical predictions with steric clashes and self-intersections \citep{corso2022diffdock}.

\vspace{0.1cm}\noindent\textbf{Generative Docking:} To resolve the mismatch between the docking task and regression-based solutions, the first proposed generative model is DiffDock~\citep{corso2022diffdock}. Its diffusion model is parameterized by Tensor Field Networks~\citep{thomas2018tensor} and predicts updates to noisy ligand translations, rotations, and torsion angles before ranking generated samples with a confidence model. Meanwhile, NeuralPLexer~\citep{qiao2023statespecific} predicts a contact map conditioned on which an $E(3)$-equivariant diffusion model generates ligand coordinates and refolds the protein structure from, \emph{e.g.}, an unbound structure to the bound structure. The mentioned generative models are also able to dock to unbound or computationally generated protein structures with a reasonable degree of accuracy.

\subsubsection{Datasets and Benchmarks}
An important dataset of 3D structures of small molecules bound to proteins is PDBBind~\citep{liu2017forging} which curates complexes from the Protein Data Bank (PDB)~\citep{berman2003announcing} if they have binding affinities available and meet additional quality criteria. It consists of ~20k complexes with ~4k unique proteins. A common dataset split \citep{stark2022equibind} is based on time with complexes older than 2019 in the training data and newer ones as test data. Less stringent criteria than PDBBind for selecting complexes are applied by BindingMOAD \citep{Wagle2023BindingMoad}, which extracts ~40k protein-ligand structures from PDB. APObind \citep{aggarwal2021apobind} provides unbound protein structures for each of its protein-ligand complexes. Helpful for approaches for peptides, Propedia~\citep{Martins2021propedia} extracts protein-peptide complexes from PDB. While the amount of structure data is limited to these magnitudes, there is considerably more protein-ligand binding affinity data without structures in ChEMBL~\citep{mendez2019chembl} with ~20 million activity measurements.

To evaluate docking predictions, it is common to estimate the fraction of correct predictions, which is defined as a generated ligand pose whose RMSDT to a ground truth structure is below a specified threshold. Additionally, the number of steric clashes in the generated structure is a relevant metric. To gauge the performance of binding strength prediction, the metrics depend on the task, such as accuracy for binary classification (binding vs. not binding), ranking correlation for correctly ranking a set of ligands' binding strengths, or MAE for a binding affinity prediction. For evaluation that resembles the real-world docking problem, it is desirable for the field to evaluate docking to unbound or computationally generated protein structures (apo-structures) instead of presuming the holo-structures as input.

\subsubsection{Open Research Directions}
While the advances in molecular docking are impressive, the task is still far from solved with, \emph{e.g.}, 22\% accuracy in DiffDock when docking to structures from ESMFold. Possible improvements could stem from better generative models for biophysical structures, more meaningful feature embeddings, or more expressive 3D architectures. Nevertheless, the structure prediction capabilities promise a path toward integrating them with downstream binding strength predictors. Unlocking such approaches or similar methods to jointly leverage the available structure data and the larger amounts of sequence-based binding affinity data is promising, and initial successes exist \citep{moon2023versatile}. Additional help for accessing these larger amounts of data would be creative, problem-specific approaches for dealing with the noise in affinity measurements. 

Furthermore, there is great value in extending molecular docking to additionally model the conformational change of the protein during binding. 
We think this is more meaningful and realistic as the binding usually changes the conformation of proteins in practice.
Lastly, we wish to draw attention to the potential of generative models for statistical mechanics approaches for calculating or comparing protein-ligand interaction strengths/probabilities instead of relying on regressing on experimental affinity measurements.

\subsection{Structure-Based Drug Design}
\label{sec:sbdd}

\noindent{\emph{Authors: Meng Liu, Tianfan Fu, Michael Bronstein, Jimeng Sun, Shuiwang Ji}\vspace{0.3cm}\newline\emph{Recommended Prerequisites: Sections~\ref{subsec:mol_representation_learning}, \ref{sec_prot_pred}}\newline

In this section, we consider structure-based drug design (SBDD), a generative task for protein-ligand interaction. In this task, we aim at generating 3D molecules, known as ligands, that can bind tightly to a specific protein (a.k.a. target protein), which can be formulated as a conditional generation problem.

\subsubsection{Problem Setup}

Formally, following Section~\ref{sec:openmi_binding}, we let $\mathcal{M}=(A, E, C)$ denote the ligand molecule and $\mathcal{P}=(B, S)$ denote a protein binding site. Overall, the goal of this task is to learn the conditional distribution $p(\mathcal{M}|\mathcal{P})$ from observed protein-ligand pairs.

\subsubsection{Technical Challenges}

The unique symmetry challenge arises due to the fact that this generative task involves multiple molecules interacting with each other, rather than single molecules. This leads to a more complex symmetry challenge than modeling individual molecules. Particularly, the symmetries of individual molecules must be considered in the context of their relative positions and orientations with respect to each other. Specifically, if we rotate or translate the protein binding site, the generated molecules yielded by the generative models should be rotated or translated accordingly. Mathematically, the learned conditional distribution should satisfy $p(\mathcal{M}|\mathcal{P})=p(g\rhd \mathcal{M}|g\rhd \mathcal{P})$, where $g \in SE(3)$ and $\rhd$ represents its corresponding group action. To achieve this, the molecule generated by the model should be equivariant to the $SE(3)$ transformation of the protein.

In addition, this task faces the challenge of an extremely vast search space. To be specific, the chemical space containing all possible molecules is estimated to exceed $10^{60}$. In addition, the 3D molecules also have an additional conformation space. However, only a minuscule fraction of this space is relevant to drug discovery, as the molecules need to meet specific criteria to be considered ``drug-like''. Thus, how to effectively and efficiently model and explore such space while considering the interactions with target proteins is a fundamental consideration in this task.

\begin{table}[t]
	\centering
	\caption{Summary of existing methods for structure-based drug design in terms of adopted generative approaches, employed networks, level of modeled structures, and 3D output variables. Among these methods, AR~\cite{luo20213d}, GraphBP~\cite{liu2022generating}, and Pocket2Mol~\cite{peng2022pocket2mol} generate atoms autoregressively by modeling relative position-related variables, which can be used to determine the position of the new atom. Further, instead of generating atoms, FLAG~\cite{zhang2023molecule} considers generating fragments autoregressively. In comparison, TargetDiff~\cite{guan2023targetdiff}, DiffBP~\cite{lin2022diffbp}, and DiffSBDD~\cite{schneuing2022structure} use diffusion models to directly generate 3D coordinates of all atoms in a one-shot schema.}
        \resizebox{\columnwidth}{!}{
	\begin{tabular}{l|cccccc}
		\toprule[1pt]
		Methods & Generative Approach & Network & Level of Structures  & 3D Outputs \\
		\midrule
		AR & Autoregressive models & $\ell=0$  & Atom & Distribution of atom occurrence \\
            GraphBP & Autoregressive flow & $\ell=0$ & Atom & Relative distances, angles, and torsions \\
            Pocket2Mol & Autoregressive models & $\ell=1$ & Atom, bond & Relative coordinates \\
            FLAG & Autoregressive models & $\ell=0$ & Fragment & Relative rotation angles \\
            TargetDiff & Diffusion models & $\ell=1$ & Atom & Coordinates \\
            DiffBP & Diffusion models & $\ell=1$ & Atom & Coordinates\\
            DiffSBDD & Diffusion models &$\ell=1$  & Atom  & Coordinates\\
		\bottomrule
	\end{tabular}
 }
	\label{tab:openmi_sbdd}
\end{table}

\subsubsection{Existing Methods} 

Early studies either generate molecular SMILES strings conditional on the 3D information of target proteins~\cite{skalic2019target,xu2021novo}, or use estimated docking scores as the reward function to guide the molecule generative model~\cite{li2021structure,fu2022reinforced}. They do not explicitly model the crucial interactions between the ligand molecule and the target protein in 3D space. ~\citet{ragoza2022generating} converts protein-ligand complex structures to atomic density grids and then uses generative approaches for 3D image data to tackle the task. A limitation is that the aforementioned equivariance property is not preserved, since 3D CNNs~\cite{Ji:TPAMI2012} is not an $SE(3)$-equivariant operation for 3D grid data.

Recently, with the development of geometric deep learning and generative modeling, protein-ligand complexes are naturally modeled as 3D geometries and their intricate interactions and symmetry constraints can be effectively encoded. Generally, we can categorize these recent methods into two categories, as summarized in Table~\ref{tab:openmi_sbdd}. The first type of method generates atoms autoregressively based on the current context, which includes the binding site and previously generated atoms. To preserve the aforementioned desired $SE(3)$-equivariance property, at each autoregressive step, these methods consider modeling relative position-related variables of the new atom \emph{w.r.t.} the current context, instead of generating its 3D coordinates directly. To be specific, AR~\cite{luo20213d} uses an invariant 3D GNN (with feature order $\ell = 0$) to model the distributions of atom occurrence in 3D positions by taking the relative distances between the query position and the current context as input. Using such invariant distances \emph{w.r.t.} the context and invariant 3D GNN together ensures the modeled distribution is equivariant to the rotation and translation of the context. In comparison, GraphBP~\cite{liu2022generating} and Pocket2Mol~\cite{peng2022pocket2mol} model the relative position of the new atom \emph{w.r.t.} the selected focal atom at each step. Specifically, GraphBP first constructs a local spherical coordinate system (SCS) at the focal atom, which is equivariant to the context's rotation and translation. It then generates the invariant distance, angle, and torsion \emph{w.r.t.} the reference SCS through an invariant 3D GNN. Pocket2Mol uses an equivariant neural network (with feature order $\ell=1$) as the encoder and the obtained equivariant features of the focal atom can be used to generate the relative position of the new atom equivariantly. Pocket2Mol also explicitly generates bonds. Instead of using atoms as building blocks, FLAG~\cite{zhang2023molecule} considers generating 3D molecules fragment-by-fragment. Such fragment vocabulary can be obtained from chemical priors and can help generate valid and realistic molecules. At each step, FLAG assembles the new fragment to the current context and then predicts the rotation angle of the new fragment \emph{w.r.t.} the selected focal fragment. The $SE(3)$-equivariance can be preserved by FLAG similarly to GraphBP. Among the above methods, AR, Pocket2Mol, and FLAG are trained via a mask-fill schema, in which atoms or fragments are randomly masked and the model is trained to recover them. In contrast, GraphBP is trained by maximizing the log-likelihood of the trajectory of atom placement steps, thanks to the exact likelihood computation of flow models.

Another line of methods, such as TargetDiff~\cite{guan2023targetdiff}, DiffBP~\cite{lin2022diffbp}, and DiffSBDD~\cite{schneuing2022structure}, considers generating 3D coordinates of all atoms directly. Compared to the above autoregressive sampling methods, such a one-shot generation fashion does not require an order among atoms and can consider global interactions of the entire ligand molecule. Following the framework of EDM~\cite{hoogeboom2022equivariant}, these methods apply diffusion models~\cite{ho2020denoising} in continuous and discrete space to model atom coordinates and atom types, respectively. The denoising step is modeled by an equivariant GNN (with feature order $\ell=1$)~\cite{satorras2021en}. To circumvent the difficulty of maintaining translation equivariance in the diffusion process, they shift the Center of Mass (CoM) of the system to diffuse and denoise the coordinates in the linear subspace only. Moreover, with a loose notation, since the latent variables follow a rotationally invariant Gaussian distribution $p(\bm{r}_T)$ and the transition distribution $p(\bm{r}_{t-1}|\bm{r}_t,\mathcal{P})$ is equivariant, the aforementioned equivariance property can be achieved~\cite{kohler2020equivariant}. Without employing diffusion models, VD-Gen~\cite{lu20233d} introduces a learnable refinement technique known as virtual dynamics. This method iteratively repositions randomly initialized particles within the pocket, aligning them with ground-truth molecular atoms.

}

In addition to deep generative models, reinforcement learning (RL) has also been used for structure-based drug design~\citep{li2021structure,fu2022reinforced}, which formulates the drug molecule generation process as a Markov decision process (MDP). Unlike generative models that explicitly build the continuous data distribution, reinforcement learning selects the action from discrete space that would receive the maximal reward (\emph{e.g.}, docking score in structure-based drug design). Specifically, DeepLigBuilder~\citep{li2021structure} builds a policy network to select the appropriate actions (whether/where to add atom, which atom to add) to grow the molecule in the target pocket; Reinforced genetic algorithm (RGA)~\citep{fu2022reinforced} leverages a policy network that selects the discrete action space (mutation, crossover position) of the genetic algorithm (GA) intelligently, which suppresses the random-walk behavior in genetic algorithm. 

It is worth mentioning that earlier molecular optimization methods based on SMILES, SELFIES, or molecular graphs can also achieve competitive performance in structure-based drug design if we incorporate the docking score as optimization goal~\citep{huang2021therapeutics,gao2022sample}. These methods are SMILES variational autoencoder (SMILES-VAE)~\citep{gomez2018automatic},  junction tree variational autoencoder (JTVAE)~\citep{jin2018junction}, graph convolutional policy network (GCPN)~\citep{you2018goal}, molecular graph-level genetic algorithm (Graph-GA)~\citep{jensen2019graph}, graph autoregressive flow (GraphAF)~\citep{shi2020graphaf}, Multi-constraint molecule sampling (MIMOSA)~\citep{fu2021mimosa}, \emph{etc.}  


\subsubsection{Datasets and Benchmarks}

First, we briefly introduce several structure-based drug design datasets, including CrossDocked2020, PDBBing, DUD-E, and scPDB. 
(1) CrossDocked2020~\citep{francoeur2020three} is a widely used benchmark dataset for evaluating the performance of various methods on structure-based drug design. CrossDocked2020 contains an extensive collection of 22,584,102 docked protein-ligand complexes. These complexes are generated through cross-docking, where ligands associated with a specific pocket are docked into each receptor assigned to that pocket by Pocketome~\citep{kufareva2012pocketome}, using the smina docking software~\citep{koes2013smina}. There are a total of 2,922 pockets and 13,839 ligands covered in CrossDocked2020. Given the variability in the quality of these complexes, it is common in existing studies to include a filtering step. This step involves removing complexes with root-mean-squared deviation (RMSD) of the binding pose that exceeds a certain threshold. This aims to encourage the model to generate ligand molecules with higher binding affinity.
(2) PDBBind is an extensive repository derived from the Protein Data Bank (PDB)~\citep{berman2003announcing}, containing experimentally determined binding affinity data for protein-ligand complexes~\citep{wang2004pdbbind}. It comprises 19,445 protein-ligand pairs. (3) Directory of useful decoys, enhanced (DUD-E) provides a directory of useful decoys for protein-ligand docking~\citep{mysinger2012directory}. It consists of 22,886 protein-ligand complexes and their affinities against 102 distinct protein targets. (4) scPDB is a refined version of the Protein Data Bank (PDB) specifically tailored for structure-based drug design, enabling the identification of optimal binding sites for protein-ligand docking~\citep{meslamani2011sc}. It contains 16,034 protein-ligand pairs over 4,782 proteins and 6,326 ligands.

To assess the performance of different generative methods, several categories of metrics are commonly used. The first category involves measuring the quality of the generated molecules, including their chemical validity, novelty, and diversity. Additionally, comparing the distributions of specific variables, such as bond length~\citep{ragoza2022generating,liu2022generating,peng2022pocket2mol}, bond angles~\citep{ragoza2022generating}, and the occurrence of different motifs~\citep{peng2022pocket2mol}, between the generated molecules and a reference set can provide further insights into the quality of the generated molecules. The second category aims to estimate the binding affinities between the generated molecules and the target proteins by using the Vina energy or deep learning-based scoring functions~\citep{ragoza2022generating}. Comparing the binding affinities of the generated molecules to those of reference molecules helps assess their effectiveness in binding to the target. The last category includes measuring other important properties, such as drug-likeness QED (Quantitative Estimate of Drug-likeness)~\citep{bickerton2012quantifying} and SA (synthesizability accessibility)~\citep{ertl2009estimation}.
\citep{huang2021therapeutics} incorporates an SBDD benchmark that compares five machine learning approaches under the same number of docking oracle calls (5K).

\subsubsection{Open Research Directions}

Despite the progress of deep learning approaches in structure-based drug design, how to effectively and efficiently model the vast chemical space to generate valid and synthesizable molecules is still a predominant challenge. Incorporating essential chemical priors, such as motif fragments and scaffolds, could be a direction to tackle this challenge. For example, to enable fragment-based drug design, DiffLinker \cite{igashov2022equivariant} uses an $E(3)$-equivariant 3D-conditional diffusion model similar to DiffSBDD to link disconnected molecular fragments (pharmacophores) into a single molecule, while it can take the surrounding protein pocket into consideration as conditional information. In addition, a recent work~\cite{adams2022equivariant} proposes a shape-based 3D molecule generation approach, which could be another promising direction to narrow down the modeling space. 

Considering that a molecule must satisfy many properties, such as solubility and permeability, to become drug-like~\cite{bickerton2012quantifying}, another remaining challenge is to simultaneously optimize multiple drug-like properties of generated drug candidates, while retaining its binding affinity for the target protein. To our knowledge, existing works do not explicitly optimize such properties during generative modeling.

\subsection{Molecule and Material Interactions}

\noindent{\emph{Authors: Zhao Xu, Limei Wang, Meng Liu, Montgomery Bohde, Yuchao Lin, Shuiwang Ji}\vspace{0.3cm}\newline\emph{Recommended Prerequisites: Section~\ref{subsec:mol_representation_learning}}\newline



This subsection describes research problems related to interactions between molecules and materials, \revisionOne{including predicting the energy and per-atom force of molecule-material pair structures (S2E and S2F) and predicting the
relaxed final structure and the energy at its relaxed state from a given initial structure (IS2RS and IS2RE).} In addition, we discuss the unexplored generative tasks for molecule-material interactions in Section~\ref{subsubsec:mol_mat_future}.


\subsubsection{Problem Setup} 
\label{sec:mol_mat_problem}


Let the total number of atoms in a molecule-material pair be $n$, and the paired structure $\mathcal{S}$ is represented as $\mathcal{S}=(\bm{z}, C)$. Here, $\bm{z}\in\mathbb{Z}^n$ is the atom type vector indicating the atom type (atomic number) of all $n$ atoms in the structure. $C=[\bm{c}_1,...,\bm{c}_n]\in\mathbb{R}^{3\times n}$ is the coordinate matrix where $\bm{c}_i$ denotes the 3D coordinate of the $i$-th atom in the structure. The first problem that has garnered the attention of the research community is predicting the energy of molecule-material pair structures (S2E). This task is to learn a function $f_E$ to predict the property $e\in\mathbb{R}$ for any given pair structure $\mathcal{S}$, where $e$ is a real number. The second problem of interest is predicting per-atom force given a structure's atomic types and positions as input (S2F). Here, the goal is to learn a function $f_F$ to predict force matrix $F\in\mathbb{R}^{3 \times n}$ for any given structure $\mathcal{S}$. Per-atom forces drive structure relaxation until the pair structure reaches its relaxed state with an energy minimum. The third problem aims to learn a function $f_{RE}$ to predict the structure's energy $e_{rel}\in\mathbb{R}$ at its relaxed state, given its initial structure $\mathcal{S}_{init}$ as input (IS2RE). Typically, the initial structure $\mathcal{S}_{init}$ is heuristically determined. Similar to IS2RE, the last problem IS2RS aims to learn a function $f_{RS}$ to predict the relaxed final structure given its initial structure as input. In this problem, the target $C_{rel}\in\mathbb{R}^{3 \times n}$ represents atomic positions at the relaxed state of the given structure. 
\revisionOne{In addition to total energy, adsorption energy is a critical property for understanding interactions between a molecule or adsorbate and a catalyst surface. Adsorption energies of reaction intermediates can act as powerful descriptors, often correlating with experimental outcomes such as catalytic activity or selectivity. The adsorption energy ($ e_{ads}$) is calculated as the energy of the adsorbate-surface system ($e_{sys}$) minus the energy of the clean surface ($e_{slab}$) and the energy of the adsorbate in the gas phase or reference state ($e_{gas}$) 
\begin{equation*}
    e_{ads} = e_{sys} - e_{slab} - e_{gas}.
\end{equation*}
Unlike the task of predicting the energy from a given initial structure, which we discussed above, calculating adsorption energy involves finding the global minimum energy across all possible adsorbate placements and configurations. Note that in some papers, to simplify, they also refer to the $e_{ads}$ based on local minimum energy for a given initial structure as adsorption energy, instead of considering global minimum energy. Therefore, depending on the dataset, the energy terminology we use in the following may refer to total energy for a molecule-material pair or adsorption energy (local or global minimum energy).}
The Datasets released in the Open Catalyst Project~\cite{chanussot2021open} provide absorbate-catalyst pair structures and serves as a testbed for the problems related to molecule-material interactions described above.

\subsubsection{Technical Challenges}

For different problems defined above, there exist distinct challenges. First, for the energy prediction problem (S2E), the model prediction has to be rotationally invariant because energy is a structure-level property that is invariant to the rotation of molecule-material pairs. In contrast to energy, the force prediction problem (S2F) aims to predict per-atom force vectors. Hence, force prediction has to be equivariant to the rotation of the structure. Similarly, the relaxed energy prediction problem (IS2RE) and the relaxed structure prediction problem (IS2RS) have the same challenge of maintaining invariance and equivariance, respectively. In addition to symmetry, IS2RE and IS2RS problems have another challenge that the initial structure only provides a rough hint about the relaxed structure. Therefore, the model must consider structure relaxation, including molecule and material atoms, to obtain accurate predictions. 

\subsubsection{Existing Methods}

\begin{table}[t]
	\centering
	\caption{Comparison of existing methods for energy, force, and position prediction of molecule-material pairs. Different methods focus on different tasks and use different pipelines to solve the IS2RE and IS2RS tasks. Note that the methods introduced in Section~\ref{subsec:mol_representation_learning} can also be applied to the tasks presented in this section. However, for the sake of brevity, this table only includes several state-of-the-art methods, including ForceNet~\cite{hu2021forcenet}, GNS+NoisyNode~\cite{godwin2022simple}, Uni-Mol+~\cite{lu2023highly}, Equiformer~\cite{liao2023equiformer}, EquiformerV2~\cite{liao2023equiformerv2}, SpinConv~\cite{shuaibi2021rotation}, GemNet-OC~\cite{gasteiger2022gemnetoc}, SCN~\cite{zitnick2022spherical}, and eSCN~\cite{passaro2023reducing}.}
        \resizebox{\columnwidth}{!}{
	\begin{tabular}{l|cccccc}
		\toprule[1pt]
		Methods & Task & Pipeline for IS2RE and IS2RS tasks & Network & Symmetry \\
		\midrule
            ForceNet & S2F/IS2RS & Relax & - & - \\
            GNS+NoisyNode & IS2RE/IS2RS & Direct & - & - \\
            Uni-Mol+ & IS2RE & Relax & $\ell=0$ & Invariant \\
            Equiformer & IS2RE & Direct & $\ell=1$ & $SE(3)$/$E(3)$-Equivariant \\
            EquiformerV2 & S2EF/IS2RE/IS2RS & Relax & $\ell>1$ & $SE(3)$/$E(3)$-Equivariant \\
		SpinConv & S2EF/IS2RE/IS2RS & Direct/Relax & $\ell=0$ & Invariant \\
            GemNet-OC & S2EF/IS2RE/IS2RS & Relax & $\ell=0$ & Invariant \\
            SCN & S2EF/IS2RE/IS2RS & Direct/Relax & $\ell>1$ & Approximately equivariant \\
            eSCN & S2EF/IS2RE/IS2RS & Relax & $\ell>1$ & Approximately equivariant \\
		\bottomrule
	\end{tabular}
 }
	\label{tab:openmi_oc}
\end{table}
Table~\ref{tab:openmi_oc} provides a summary of existing methods, including their symmetry type, network order, tasks, and the pipeline used in the original paper. As discussed in Section~\ref{subsec:mol_representation_learning}, both invariant and equivariant methods can predict the $SE(3)$-invariant energy $e$. Once the energy is predicted, the $SO(3)$-equivariant force vectors $F$ acting on atoms can be obtained using the formula $F_i=-\frac{\partial e}{\partial \bm{c}_i}$. \revisionOne{This can ensure energy conservation but requires additional computational steps. Therefore, methods like SpinConv~\cite{shuaibi2021rotation} and GemNet-OC~\cite{gasteiger2022gemnetoc} opt for predicting force directly to speed up the model. Practically, direct force prediction may lead to better performance when we have sufficiently large training datasets.}
As discussed in Section~\ref{subsec:mol_representation_learning}, one main challenge for invariant methods is efficiency. Existing methods suffer from high computational cost~\cite{gasteigerdirectional, gasteiger2021gemnet} when incorporating more geometric information, like angles and torsion angles, into the network. On the other hand, one main challenge for equivariant methods is that, explicitly imposing physical constraints into the model architecture limits the capacity of the network~\cite{schutt2021equivariant}. Consequently, recent studies try to design GNN models that are both efficient and expressive without explicit physical constraints. For example, ForceNet~\cite{hu2021forcenet} directly uses atom coordinates in a scalable manner, but implicitly imposes physical constraints by using data augmentation.

To address the challenge of IS2RE and IS2RS tasks, where only initial structures are given, methods that consider structure relaxation are needed. There are primarily two solutions. The first is to use a well-trained S2F model to iteratively update the structure from the initial one. Atom positions are updated step-by-step based on the predicted forces of the current structure until the predicted forces approach zero. While this method can accurately simulate structure relaxation, it requires numerous steps to achieve the final output. Recently, both SCN~\cite{zitnick2022spherical} and eSCN~\cite{passaro2023reducing} use this indirect approach and note that they are approximately equivariant but with high order $\ell>1$. Incorporating both eSCN and a recent direct method Equiformer~\cite{liao2023equiformer}, EquiformerV2~\cite{liao2023equiformerv2} achieves state-of-the-art performance among these indirect methods. It applies eSCN convolution layers to Equiformer network structure with several additional techniques, including separable $\mathbb{S}^2$ activation and layer normalization for vectors of $\ell=0$ and those of $\ell>0$. In contrast to indirect methods, the direct approach aims to model the relations between initial and relaxed structures, with the output typically being the difference $C_{rel} - C_{init}$~\cite{godwin2022simple}. Hence, direct methods are faster in training and prediction but less accurate than indirect methods. Equiformer~\cite{liao2023equiformer}, an attention-incorporated equivariant architecture with $\ell=1$, achieves state-of-the-art performance among direct methods. Currently, the NoisyNode technique proposed in ~\cite{godwin2022simple} is widely used in many other direct approaches. However, it's worth noting that both GNS+NoisyNode~\cite{godwin2022simple} and ForceNet~\cite{hu2021forcenet} are based on the GNS framework, which is neither invariant nor equivariant. These methods take atom coordinates as input and use rotation data augmentation to improve performance. The symmetry of Uni-Mol+~\cite{lu2023highly} is somewhat nuanced as its main architecture is invariant, while it uses EGNN~\cite{satorras2021n} to update atom coordinates and sustain equivariance.


\subsubsection{Datasets and Benchmarks}
Open Catalyst 2020 (OC20) dataset~\cite{chanussot2021open} is a valuable resource for testing machine learning models on the problems related to molecule-material interactions described in this subsection. This dataset provides absorbate (molecule)-catalyst (material) pair structures and includes three tasks, namely Structure to Energy and
Forces (S2EF),  Initial Structure to Relaxed Energy (IS2RE), and Initial Structure to Relaxed Structure (IS2RS), which correspond to the tasks introduced in Section~\ref{sec:mol_mat_problem}. 

IS2RE and IS2RS tasks use the same dataset which contains the input initial structures and the output relaxed structures and energies. The dataset is originally split into training, validation, and test sets. The training set contains 460,328 structures. For both validation and testing sets, there for four subsets, namely in-domain (ID), out-of-domain adsorbate (OOD Ads), out-of-domain catalyst (OOD Cat), and out-of-domain adsorbate and catalyst (OOD Both). The ID set consists of structures from the same distribution as training. The OOD Ads set consists of structures with unseen adsorbates, the OOD Cat set consists of structures with unseen element compositions for catalysts, and the OOD Both set consists of structures with unseen catalysts and adsorbates. 
The four validation sets contain 24,943, 24,961, 24,963, and 24,987 structures, respectively. The testing sets contain 24,948, 24,930, 24,965, and 24,985 structures, respectively. Note that the labels for the test sets are not publicly available. Therefore, researchers need to submit their results to the Open Catalyst Project (OCP) leaderboard to evaluate their models on the test sets.

The dataset for S2EF contains more structures compared to the other two tasks. This is because the dataset for S2EF contains not only the initial and relaxed structures but also the intermediate structures in the relaxation trajectory, \emph{etc}. The dataset comprises 133,934,018 structures for training, while for the ID, OOD Ads, OOD Cat, and OOD Both validation sets, it contains 999,866, 999,838, 999,809, and 999,944 structures, respectively. Similarly, the ID and OOD testing sets include 999,736, 999,859, 999,826, and 999,973 structures, respectively.

\revisionOne{Importantly, the adsorption energy in OC20 does not necessarily correspond to the global minimum energy since it only considers one initial configuration for the molecule-material pair. Therefore, researchers further propose OC20-Dense \cite{lan2022adsorbml}, which contains a dense sampling of initial configurations. Specifically, OC20-Dense includes two splits, validation set for development and test set for evaluation. Each split consists of approximately 1,000 unique adsorbate-material combinations from the four subsets, ID, OOD Ads, OOD Cat, and OOD Both, in the validation set and test set of the OC20 dataset. For each combination, the authors perform a dense sampling of initial configurations and calculate relaxations using DFT to find the global minimum energy.}

In addition to OC20~\cite{chanussot2021open} and OC20-Dense \cite{lan2022adsorbml}, OC22~\cite{tran2023open} is curated recently and focuses on oxide electrocatalysts. Similarly, it contains three tasks, namely Structure to Total Energy and Forces (S2EF-Total), Initial Structure to Total Relaxed Energy (IS2RE-Total), and Initial Structure to Relaxed Structure (IS2RS). One main difference between OC20 and OC22 is that OC22 employs total energy targets rather than adsorption energy targets. This makes the models more general and enables the calculation of more properties, but it is more challenging.

\subsubsection{Open Research Directions}
\label{subsubsec:mol_mat_future}


\noindent
Though significant progress has been made over the last several years, there are still remaining challenges in modeling molecule-material interactions. Firstly, the relaxed energy/structure in the IS2RE/IS2RS challenge does not necessarily correspond to the global minimum adsorption energy for a molecule-material pair, and is sensitive to changes in the initial structure. Recently, \cite{lan2022adsorbml} used a brute force strategy to calculate relaxed energies across many initial configurations. In order to efficiently predict global minima, it will be necessary to use more advanced strategies for sampling initial configurations or build models that do not depend on specified initial configurations. Secondly, although many materials in molecule-material interactions are periodic, to our knowledge, there do not exist any models which explicitly model such periodicity when modeling a molecule-material pair. \revisionOne{Instead, models only consider material atoms within a predefined cutoff radius to the small molecule, which may contain only one or a few unit cells of the material. Furthermore, top performing models all use direct force predictions instead of gradient-based force calculations, which require additional computational steps. However, such direct force predictions may not obey energy conservation laws. Developing a method that ensures energy conservation without significant additional cost is another challenge and an important direction for future research.} Finally, models are currently incapable of incorporating all physical properties of the system. Many models have seen improved performance by including molecular dynamics information, however, current works cannot model properties such as magnetic or charge effects which can significantly impact the relaxed energy/structure.


\vspace{0.1cm}\noindent\textbf{Generation of Molecule-Material Pairs:} 
One promising direction for generative tasks in this field is to generate periodic materials conditioned on given molecules. This task involves training a generative model to produce new periodic material structures with specific properties, such as appropriate adsorption energies, when a specific molecule is present. To accomplish this, the generative model would need to be trained on a dataset of periodic material structures with different adsorbate molecules and their corresponding adsorption energies. Such a generative task has many potential applications, including designing new materials for electrocatalysts.

Although recent studies have explored unconditional material generation, as described in Section~\ref{sec:mat_gen}, developing a generative model that accurately captures the complex interactions between absorbate molecules and periodic materials is challenging. It requires a deep understanding of the underlying physics and chemistry. Moreover, similar to the structure-based drug design task introduced in Section~\ref{sec:sbdd}, the vast data space that needs to be modeled makes this task even more challenging. Additionally, this task presents a unique challenge in that the 3D geometry of the molecule is not static and will be influenced by the generated material molecule. Therefore, when generating periodic materials conditioned on given molecules, the generative model needs to account for the interplay between the molecule and the material, including the induced changes in the molecular geometry.



\clearpage
\newcommand{\jacob}[1]{\textcolor{blue}{#1}}
\newcommand{\jjacob}[1]{\textcolor{blue}{#1}}
\newcommand{\jjjacob}[1]{\textcolor{blue}{#1}}

\renewcommand{\jacob}[1]{#1}
\renewcommand{\jjacob}[1]{#1}
\renewcommand{\jjjacob}[1]{#1}

\hypertarget{AI for Partial Differential Equations}{\section{AI for Partial Differential Equations}}\label{sec:pde}

In this section, we detail advances in the field of AI for solving Partial Differential Equations (PDEs). We overview the general formulation of PDE modeling and motivate machine learning methods in this context in~\cref{sec:pde_overview}. We discuss forward problems in~\cref{sec:pde_forward} and inverse tasks in~\cref{sec:pde_inverse}.

\begin{figure}[htbp]
    \centering
    \includegraphics[width=0.99\textwidth]{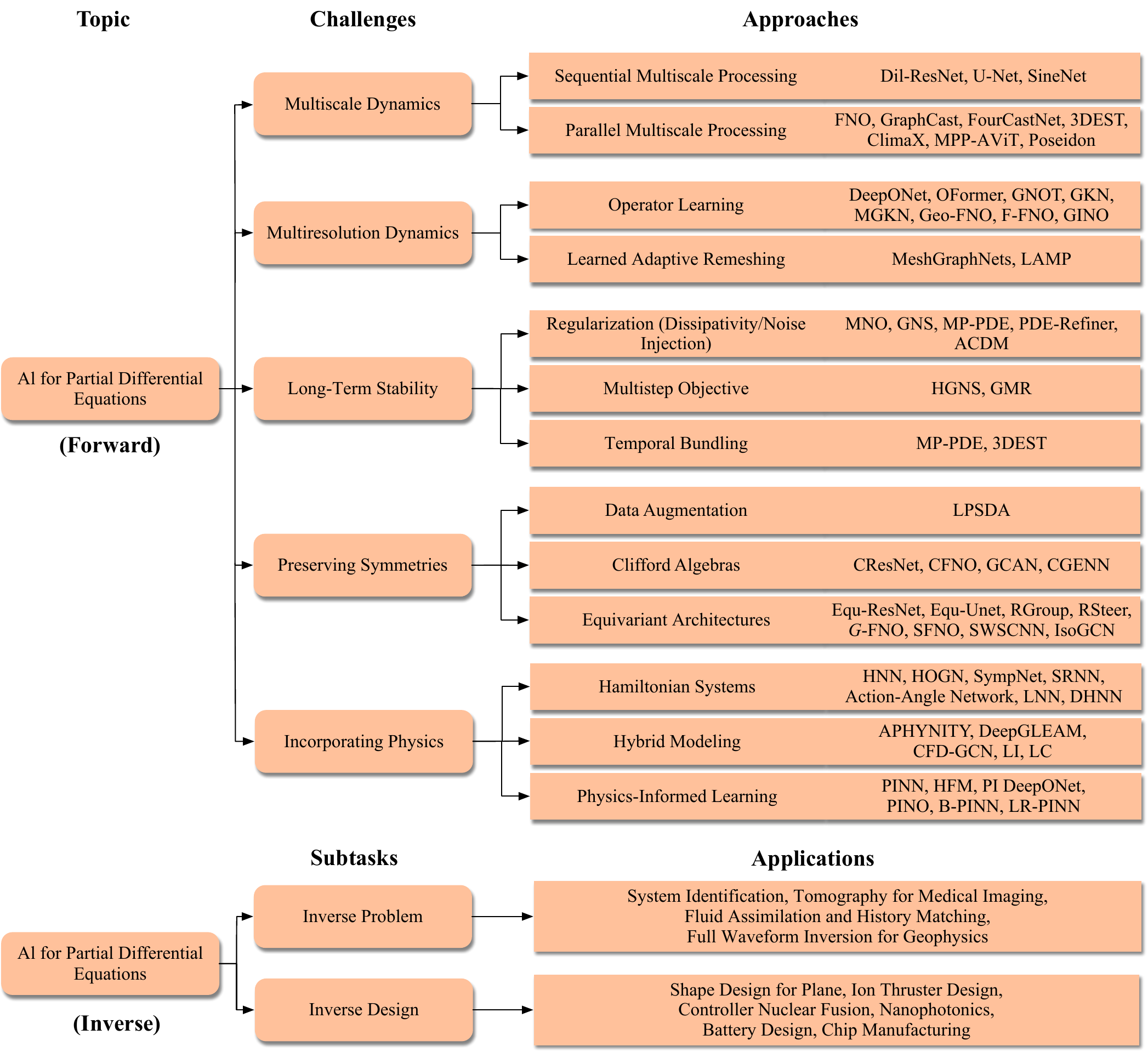}\vspace{-0.3cm}
    \caption{Overview of AI for forward modeling and inverse modeling of partial differential equations (PDEs). In~\cref{sec:pde_forward}, we consider the forward modeling task, that is, mapping from the initial timesteps of the numerical solution of a PDE to later timesteps. We identify and detail four fundamental challenges that have defined the development of neural PDE solvers. Multi-scale dynamics arise in systems where physics evolve on a continuum from local to global scales~\revisionOne{\cite{stachenfeld2021learned, gupta2022towards, zhang2024sinenet, li2021fourier,lam2022graphcast, pathak2022fourcastnet, bi2022pangu,nguyen2023climax, mccabe2023multiple, herde2024poseidon}}, while multi-resolution dynamics occur in systems with fast-evolving, isolated regions that require greater resources for stable simulation~\cite{pfaff2021learning, wu2022learning,lu2019deeponet,li2022transformer,hao2023gnot,li2020neural,li2020multipole,li2022fourier,li2023geometry}. Solvers that use explicit schemes
    encounter error in their inputs introduced by previous predictions and thus must consider methods for maintaining rollout stability~\cite{li2021learning,sanchez2020learning,brandstetter2022message,lippe2023pde,kohl2023turbulent,wu2022subsurface,han2021predicting}. Equivariant architectures and training techniques enforce symmetries of the system, enabling improved generalization and sample complexity~\cite{wang2021incorporating, wang2022approximately,wang2023relaxed,brandstetter2022lie,brandstetter2023clifford,ruhe2023clifford,ruhe2023geometric, helwig2023group,bonev2023spherical,esteves2023scaling,horie2021isometric}. Lastly, incorporating physics into architectures enables predictions to maintain physical consistency and reduces the difficulty of the learning task 
    ~\revisionOne{\cite{greydanus2019, sanchez2019hamiltonian, jin2020sympnets, chen2020Symplectic, daigavane2022learning, cranmer2020lagrangian, sosanya2022dissipative, yin2021augmenting, wu2021deepgleam, belbute2020combining, kochkov2021machine, tompson2017accelerating, raissi2019physics, raissi2020hidden, wang2021learning, li2021physics, yang2021b, cho2024hypernetwork}}. In~\cref{sec:pde_inverse}, we consider the inverse modeling task, including inverse problems and inverse design. Specifically, the task considered by inverse problems is to infer the unknown parameters of the system given observed dynamics, while the task for inverse design is to optimize the system based on a predefined objective. These two subtasks of inverse modeling have various applications across the science and engineering fields.}\label{fig:pde_challenges}
\end{figure}

\subsection{Overview}\label{sec:pde_overview}

\noindent{\emph{Authors: Jacob Helwig, Ameya Daigavane, Tess Smidt, Shuiwang Ji}}\newline

A PDE mathematically describes the behavior of a system \jacob{through} an unknown multivariate function \jacob{$u$ by prescribing constraints relating $u$ and its partial derivatives}. PDEs are frequently applied in a variety of disciplines to model the space-time evolution of physical processes, such as airflow around an airfoil with the Navier-Stokes equations~\revisionOne{\cite{pfaff2021learning,li2022fourier,bonnet2022airfrans}}, global weather patterns with the shallow water equations~\cite{gupta2022towards}, or optical design with Maxwell's equations~\cite{brandstetter2023clifford}. Additional real-world applications of PDE modeling include weather forecasting~\citep{pathak2022fourcastnet}, carbon dioxide storage~\cite{wen2022u,wen2023real}, seismic wave propagation~\cite{yang2021seismic,sun2022accelerating,yang2023rapid,sun2023next}, material sciences, and volcanic activities~\cite{rahman2022generative}. In many real-world applications, it may be intractable to obtain the functional form of the PDE solution, and therefore, the PDE must be solved numerically. Due to the widespread application of PDE modeling, over a century of work has been dedicated to developing classical numerical PDE solvers~\cite{brandstetter2022message}.


\revisionOne{The development of classical solvers has largely been defined by choice of spatial discretization scheme. In Eulerian schemes, the continuous spatial domain is discretized into finitely many mesh points on which c}lassical solvers \revisionOne{employ} numerical approximations of derivatives~\cite{quarteroni2008numerical,Bartels2016}, such as  forward difference approximations of the form
\begin{align*}
    \partial_x u(x, t) &\approx \frac{u(x + \revisionOne{\Delta_X}, t) - u(x, t)}{\revisionOne{\Delta_X}},
\end{align*}
\revisionOne{where $x$ and $x+\Delta_X$ are points on the computational mesh.}
These approximations improve in accuracy as the \revisionOne{mesh spacing} \revisionOne{$\Delta_X$} decreases to $0$. \revisionOne{Lagrangian schemes instead discretize the particles of the modeled material~\cite{lucy1977numerical,gingold1977smoothed}, with common applications including fluid dynamics~\cite{sanchez2020learning,toshev2024neural,toshev2024lagrangebench} or the behavior of a deformable solid such as a metal or fabric under an external force~\cite{pfaff2021learning}. While Lagrangian models enjoy a number of advantages over their Eulerian counterparts~\cite{price2011smoothed}, common classical Lagrangian models such as Smoothed Particle Hydrodynamics (SPH) encounter new challenges such as particles unphysically clumping together due to negative pressure or stress~\cite{price2012smoothed}.}

 \revisionOne{To numerically advance the solution in time from $u_t$ to $u_{t+\Delta_T}$, where $\Delta_T>0$, classical solvers employ either \textit{implicit} or \textit{explicit} time integration schemes, where we use $u_t$ to denote $u$ evaluated at time point $t$, giving a function of space. Explicit schemes obtain $u_{t+\Delta_T}$ directly from $u_{t}$, with the forward Euler method given by
\begin{equation}\label{eq:fwd_eul}
   u_{t+\Delta_T}\approx u_t + \Delta_T(\partial_tu)_{t}
\end{equation}
being one of the simplest examples. Note that dependence on the spatial derivatives of $u$ in the approximation given by~\cref{eq:fwd_eul} arises through their relationship to $\partial_t u$, later formalized with~\cref{eq:pde}. Alternatively, implicit methods utilize the form of~\cref{eq:pde} to derive an operator $\mathcal{A}$ such that
\begin{equation}\label{eq:implicit}
    \mathcal{A}\left(u_t, u_{t+\Delta_T}\right)=\mathbf{0},
\end{equation}
giving a set of (possibly non-linear) equations which can be solved to obtain the unknown $u_{t+\Delta_T}$ given $u_t$.}
\revisionOne{In general, e}xplicit schemes \revisionOne{tend to be} simpler to implement than implicit schemes, yet usually require smaller step sizes \revisionOne{$\Delta t$} to achieve similar accuracy or even to converge for stiff problems which may exhibit sharp discontinuities~\cite{cfl1928}. \revisionOne{However, while implicit methods can remain stable for larger time step sizes, numerically solving the system of equations given by~\cref{eq:implicit} often requires an iterative numerical solver which may take many iterations to converge within the desired tolerance. In summary, explicit schemes tend to require many inexpensive steps, whereas implicit schemes often enable fewer steps at a greater cost per step.}

While classical approaches such as \revisionOne{SPH~\cite{price2012smoothed},} the Finite Element Method and the Finite Difference Method \citep{quarteroni2008numerical,Bartels2016} have been proven to be effective, they require high computational effort. Furthermore, they often need to be carefully tailored on a task-by-task basis to ensure numerical stability. Large systems that are prevalent in industry applications of PDE modeling can require extensive computational resources and hundreds or even thousands of CPU hours~\cite{lam2022graphcast}.


To address these shortcomings, deep learning models have emerged as a general framework to produce solutions orders of magnitude faster than their numerical counterparts. This efficiency is primarily achieved via the ability of neural networks to take \revisionOne{substantially} larger time steps~\cite{kochkov2021machine,toshev2024lagrangebench}, learn on more coarse spatial discretizations compared to classical solvers~\cite{pfaff2021learning, stachenfeld2021learned}, and use explicit forward methods instead of implicit methods~\cite{tang2020deep,wu2022subsurface}. Unlike time-consuming iterative methods for implicit schemes, neural solvers learn a direct mapping from past states to future states, with fewer restrictions on the resolution of the data~\cite{kochkov2021machine}. Additionally, neural solvers can easily be optimized and evaluated using parallelized GPU operations, while the design of GPU-compatible classical solvers may offer limited benefit due to their iterative nature, and furthermore requires intimate familiarity with complex numerical methods. \revisionOne{These efficiency boosts can become particularly pronounced when uncertainty quantification is required, such as weather forecasting, as advances in generative modeling can be leveraged to obtain a Monte Carlo estimate of variability in the prediction with far less effort than relying on an ensemble of classical solvers~\cite{kohl2023turbulent,lippe2023pde,price2023gencast}.}
Most importantly, neural solvers have the ability to adapt to the task at hand and can be trained to generalize across initial conditions~\cite{li2021fourier, gupta2022towards}\revisionOne{,} PDE parameters~\cite{brandstetter2022message, tran2021factorized}\revisionOne{, and geometries~\cite{li2022fourier,li2023geometry,bonnet2022airfrans,hao2023gnot}}. Further, unlike classical solvers, neural solvers can learn dynamics directly from observed data, an ability that is especially useful when the underlying equations are unknown~\cite{lienen2022learning}.

Motivated by their prominence in \revisionOne{real-world} applications of PDE modeling and their challenging nature, many works focus on designing neural solvers for the class of time-evolving PDEs. Formally, a time-evolving PDE is a system of equations relating the derivatives of an unknown function $u: U\to\mathbb R^m$ of space and time~\cite{olver2014introduction}, where $U=\mathbb X\times\mathbb T$ consists of the spatial domain $\mathbb X$ and the temporal domain $\mathbb T$. Given $U$, we consider time-evolving PDEs given by a set of equations~\cite{brunton2023machine, evans2022partial}
\begin{align}\label{eq:pde}
\nonumber\partial_tu + \mathcal D \left(x, t, u,\partial_{x}u, \partial_{xx}u, \ldots\right)&=\mathbf{0} & (x,t)&\in U, 
\\
u(x,0) &= u_0(x) & x&\in \mathbb X,
\\
\nonumber
\mathcal B u(x,t) &=\mathbf{0}& (x,t)&\in\partial \mathbb X\times \mathbb T,
\end{align}
where $\mathcal D$ is a differential operator relating the partial derivatives of the solution $u$ on the space-time domain $U$, $\mathcal B$ is a differential operator relating derivatives on the boundary of the spatial domain $\partial\mathbb X$, and $u_0$ is the initial condition describing $u$ at time $t=0$. To solve this PDE, we must identify a function $u(x,t;\gamma)$ satisfying the constraints in~\cref{eq:pde}, either in analytical or numerical form, where $\gamma=(u_0, \mathcal B,\gamma_P)$ denotes the PDE configuration describing the initial condition $u_0$, boundary condition $\mathcal B$, and PDE parameters $\gamma_P$. There are several learning frameworks that exist for approximating this solution. Approaches for the forward problem, discussed in~\cref{sec:pde_forward}, utilize a forecasting model as a learned solver to map past numerical solutions to future solutions. Alternatively, inverse problems and inverse design, detailed in~\cref{sec:pde_inverse}, consider the reverse direction, where the task is instead to map from observed solution data to the \jacob{PDE configuration $\gamma$} or to optimize the design of a system based on some criterion. 

\subsection{Forward Modeling}\label{sec:pde_forward}

\noindent{\emph{Authors: Jacob Helwig, Ameya Daigavane, Rui Wang, Kamyar Azizzadenesheli, Anima Anandkumar, Rose Yu, Tess Smidt, Shuiwang Ji}}\newline

In this section, we overview the progress of machine learning models developed for forward PDE problems. We formalize the forward task for neural PDE solvers in~\cref{sec:pde_prob} before outlining the primary challenges that have shaped their development in~\cref{sec:pde_chall}. In~\cref{sec:msDyn,sec:mrDyn,sec:roSty,sec:phySym,sec:inPhy}, we discuss models and techniques that have emerged in response to these challenges, as well as datasets and benchmarks for these models in~\cref{sec:pdebench}, before closing with a discussion of remaining challenges and future directions in~\cref{sec:pde_future}. 

\subsubsection{Problem Setup}\label{sec:pde_prob}

In forward problems, models are tasked with predicting future states of the system given initial conditions or historical observations as inputs~\cite{kovachki2021neural,li2021physics,li2021learning,brandstetter2022message, gupta2022towards, li2021fourier, sanchez2020learning, stachenfeld2021learned,wen2023real,yang2023rapid}. Models are trained on a set of numerical solutions $\left\{u^{(j)}\right\}_{j=1}^n$, where  $u^{(j)}(x,t)\coloneqq u(x,t;\gamma^{(j)})$ and the PDE configurations $\gamma^{(j)}$ vary depending on the setting. For example, the $\gamma^{(j)}$ may correspond to varying initial conditions~\cite{li2021fourier,rahman2022u,gupta2022towards} or PDE parameters~\cite{li2020multipole,yang2021seismic,brandstetter2022message, tran2021factorized}. 

Numerical PDE solutions discretize the solution domain $U$ into a finite set of collocation points on which the solver will approximate the value of the solution function. For a PDE solution $u$, denote $u_t$ as the $t$-th time step in a uniform discretization of the temporal domain $\mathbb T$ with step size $\Delta_T$, that is, $u_t(x)\coloneqq u(x,t\Delta_T)$, where $x$ is any point in the discretization of the spatial domain $\mathbb X$. Additionally, let the PDE solutions at consecutive time points be denoted as {$u_{k:(k+K)}\coloneqq \left\{u_k,u_{k+1},\ldots,u_{k+K}\right\}$}.
The dynamics forecasting task is then defined as:
\begin{equation}
\phi_\theta(u_{0:(k-1)}^{(j)}) = \widehat{u_{k:T}^{(j)}},
\end{equation}
where $\phi_\theta$ is optimized based on a suitably chosen loss $\mathcal L$ as 
\begin{equation}\label{eq:pde_loss}
    \phi_\theta = \underset{\phi_\theta:\theta\in\Theta}{\arg\min}\:\mathbb E_{u^{(j)}}\left[\mathcal L\left(\phi_\theta\left(u_{0:(k-1)}^{(j)}\right),u_{k:T}^{(j)}\right)\right].
\end{equation}

In many cases, the form of the PDE gives rise to a solution set closed to the action of a symmetry group $G$. That is, if $u$ is a function that satisfies~\cref{eq:pde}, then for all group elements $g\in G$, $L_gu$ also satisfies~\cref{eq:pde}, where for a function $f$, $L_gf(x)\coloneqq f(g^{-1}x)$ denotes $f$ transformed by $g$. In such a case, it is desirable to constrain the model search space such that these symmetries are automatically respected, that is $\phi_\theta\left(L_gu_0\right) = L_g\phi_\theta\left(u_0\right)$. Such constraints have been shown to improve generalization and sample complexity for learned solvers via explicit encoding in equivariant architectures such as equivariant CNNs~\cite{wang2021incorporating, helwig2023group} and equivariant GCNs~\cite{horie2021isometric}, and through data augmentation~\cite{brandstetter2022lie}. We discuss this further in~\cref{sec:phySym}.

\subsubsection{Technical Challenges}\label{sec:pde_chall}

We next identify five key challenges encountered by neural solvers in the forward modeling setting, each of which have given rise to a variety of solutions that have shaped machine learning research in the field of PDE modeling.

\vspace{0.1cm}\noindent\textbf{Multi-Scale Dynamics} (\cref{sec:msDyn}): As physics evolve on multiple spatial scales, capturing the interactions within and between dynamics on each scale is vital in producing high-quality numerical solutions to PDEs. However, it is challenging to do this effectively, particularly at global scales, without excessively trading off computational efficiency or sacrificing performance at local scales.   
    
\vspace{0.1cm}\noindent\textbf{Multi-Resolution Dynamics} (\cref{sec:mrDyn}): Many systems possess isolated regions of fast-evolving dynamics that require higher-resolution discretizations relative to others to maintain solver stability. Therefore, the ability to model irregular geometries balances a trade-off between simulation accuracy and computational effort by enabling resources to be allocated dynamically in space and time.
    
\vspace{0.1cm}\noindent\textbf{Long-Term Stability} (\cref{sec:roSty}): 
    Evolving a system for many time steps can lead to an accumulation of error that causes the predicted rollout to diverge from the ground truth. As this error accumulation does not naturally occur until test time, it can be difficult to condition models to be stable during inference.   
    
\vspace{0.1cm}\noindent\textbf{Preserving Symmetries} (\cref{sec:phySym}): Many PDEs have intrinsic symmetries which are often used to find reduced-order models and improve solution efficiency. For machine learning, symmetry may be used as an inductive bias to reduce the difficulty of the learning task and narrow the size of the model search space. Furthermore, Noether's theorem~\cite{halder2018noether, noether1971invariant} establishes a connection between symmetries and conservation laws, which implies that models that uphold symmetries can produce physically consistent predictions.

\vspace{0.1cm}\noindent\textbf{Incorporating Physics} (\cref{sec:inPhy}): 
    Since machine learning models are fundamentally statistical, they are prone to make scientifically implausible predictions when trained solely on data without explicit constraints. Thus, leveraging known physical principles to guide deep learning models is crucial for learning the correct underlying dynamics instead of simply fitting the observed data, which may contain spurious, non-physical trends. This can be accomplished by imposing constraints on the loss function and the design of the architecture, or by appropriately augmenting traditional physics-based models with neural nets.

In the sections that follow, we discuss the above challenges in greater detail and advances made by previous works to address them, outlined in~\cref{fig:pde_challenges}.

\subsubsection{Existing Methods: Multi-Scale Dynamics}\label{sec:msDyn}

    Dynamics evolve and interact at multiple scales in many physical systems. For example, turbulent flows exhibit a hierarchy of localized regions of turbulent motion of different sizes, known as \textit{eddies}, in which energy from one scale is dispersed to eddies at the next smallest scale~\cite{pope2000turbulent}. While the behavior of particles is often most strongly associated with particles in the immediate neighborhood,  architectures composed of layers that only consider local information such as ResNets~\cite{he2016deep} rely on stacks of many layers to propagate long-range signals, and therefore demonstrate inferior performance in evolving dynamics
    ~\cite{li2021fourier, gupta2022towards, ruhe2023geometric}. Thus, a primary factor for faithful and efficient simulation via machine learning is the incorporation of multi-scale processing mechanisms~\cite{li2020multipole,gupta2022towards,rahman2022u,wen2023real} that balance the complexity tradeoff while maintaining sufficient local information flow. 

    
   \citet{stachenfeld2021learned} implement this mechanism in their Dil-ResNet using blocks of convolution layers with sequentially increasing dilation rates to process information beginning from local up to global scales followed by sequentially decreasing dilation rates to process from global to local. Following a similar philosophy, \citet{gupta2022towards} study several variants of the U-Net architecture~\cite{ronneberger2015u}, which uses downsampling and upsampling in place of dilation to traverse between local and global scales. In both cases, the mechanism processes local and global information sequentially while managing complexity by increasing the receptive field with a fixed kernel size, which we visualize in~\cref{fig:msDyn}. \revisionOne{\citet{zhang2024sinenet} examine the internal representation learned by U-Nets for time-evolving PDEs and show that the feature maps undergo latent evolution such that as the feature map is propagated through the architecture, it gradually evolves from the input field to the predicted field~\cite{chen2019neural}. Therefore, feature maps from the downsampling path are outdated with respect to feature maps in the upsampling path, implying that skip connections between these two paths force the model to aggregate temporally misaligned features. As these skip connections are vital for restoring high-resolution features to the upsampled feature maps, simply removing them is not an option, and therefore, \citet{zhang2024sinenet} propose to mitigate the amount of misalignment with their proposed SineNet architecture. SineNet partitions latent evolution across multiple lightweight U-Nets by composing them into an architecture resembling a sinusoid. In doing so, the degree to which the downsampled feature maps are outdated with respect to the upsampled feature map in each U-Net is reduced. Under a fixed parameter budget, SineNet is shown to substantially improve performance compared to a single U-Net.}

    \begin{figure}[t]
        \centering
        \includegraphics[width=\textwidth]{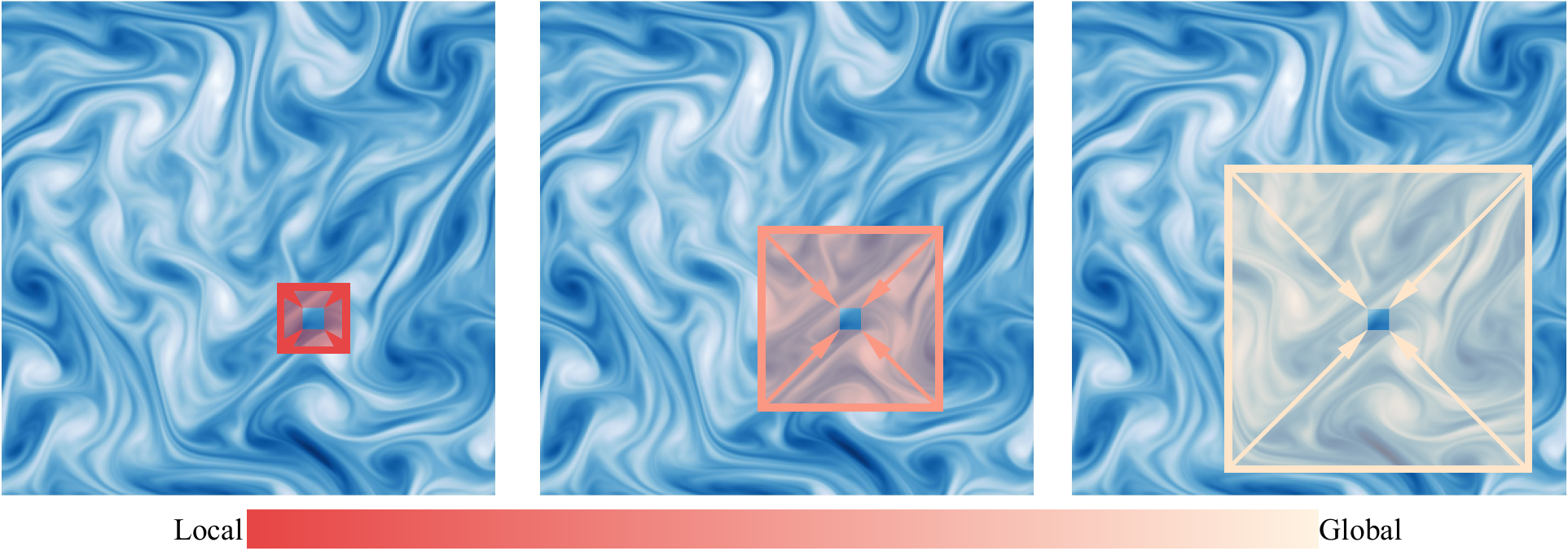}
        \caption{Multi-scale dynamics. Many systems exhibit dynamics consisting of interacting components with sizes ranging from local to global scales. A primary example is turbulent flows, which possess a hierarchy of eddies that decay down to the smallest scale, referred to as the Kolmogorov scale~\cite{pope2000turbulent}. Constructing machine learning models with multi-scale processing mechanisms is therefore key for high-fidelity simulation. These mechanisms aggregate information on each scale to update the latent representation at each mesh point. Here, we visualize a mechanism that performs aggregation and updates on each scale sequentially, as considered by \citet{stachenfeld2021learned} and \citet{gupta2022towards}, however, mechanisms such as those proposed by \citet{li2021fourier} and \citet{lam2022graphcast} act in parallel.}\label{fig:msDyn}
    \end{figure}


   In contrast to sequential processing mechanisms, the Fourier Neural Operator (FNO) proposed by~\citet{li2021fourier} processes multi-scale information in parallel~\cite{gupta2022towards}. \revisionOne{Operator learning tasks arise when the ground truth mapping $\mathcal K:\mathcal X\to\mathcal Y$ to be learned is between function spaces $\mathcal X$ and $\mathcal Y$~\cite{lu2019deeponet}, where the function spaces being considered are often Banach spaces~\cite{kovachki2021neural,seidman2022nomad}. Given a set of $n$ function pairs $\left(u^{(j)}, \mathcal K\left(u^{(j)}\right)\right)$ with $u^{(j)}\in\mathcal X$, operator learning frameworks first discretize the function pairs into point-wise evaluations on a computational grid, as described in~\cref{sec:pde_prob}, and subsequently leverage the discretized training data to learn a parametric map $\phi_\theta$ approximating the ground truth operator $\mathcal K$. If $\mathcal K$ is the forward operator advancing the PDE solution in time, then the loss given in~\cref{eq:pde_loss} can be understood as an operator learning objective. However, if the architecture of $\phi_\theta$ depends on the grid spacing of the discretization, as is the case for many convolutional neural networks, then $\phi_\theta$ will not be able to generalize well to discretizations with spacing differing from that of the training data~\cite{kovachki2021neural}, although adaptations to the convolutional framework have emerged to rectify this limitation~\cite{raonic2024convolutional}.} 
   
   \revisionOne{Instead, the family of \textit{neural operators} construct $\phi_\theta$ such that it can} generalize beyond the resolution of the discretization of the training data\revisionOne{.}
   \revisionOne{Neural operators} are de facto models for scientific computing and physics phenomena dealing with PDEs and furthermore have been shown to possess universal approximation abilities for continuous operators between Banach spaces~\cite{kovachki2021neural}. Among neural operator architectures, FNO \jacob{performs convolutions in the frequency domain, where convolution is realized via point-wise multiplication. FNO parameterizes convolution kernels in the frequency domain, that is, it directly learns the Fourier transform of kernels. Because the lower frequency modes of the transform are theoretically invariant to changes in spatial resolution, this parameterization enables FNO to generalize beyond the resolution of the training data}. \jacob{Additionally, }when dealing with regular grids and domains,  \jacob{the projection into the frequency domain} is often carried out using the Fast Fourier transform, making FNO among the most computationally efficient neural operator models. FNO has been successfully applied to many large-scale applications, including weather forecasting~\cite{pathak2022fourcastnet} and climate mitigation acts~\cite{wen2023real}. \jacob{This has\revisionOne{,} in large part\revisionOne{,} been enabled by the global Fourier convolutions in FNOs which efficiently process information on multiple scales.} In \jacob{the frequency domain}, low frequency modes represent information on a global scale, with higher frequency modes containing local information, and thus, multi-scale processing in the frequency domain occurs in parallel through the point-wise multiplication. To manage complexity, frequency modes above a fixed cutoff are set to zero, reducing the number of operations and parameters.
    
    \jjjacob{FNO has inspired a series of follow-up works, including the Factorized FNO (F-FNO)~\cite{tran2021factorized}. F-FNO introduced several modifications to the FNO architecture and training procedures to enable stability in deeper architectures, including processing of frequencies along each spatial dimension separately, effectively factorizing the transform to reduce the number of parameters per layer. \citet{poli2022transform} developed a more efficient FNO-type architecture based on the principle of only applying one transform per forward pass of the model, as opposed to the expensive forward and inverse transforms required per layer of the FNO architecture. \citet{poli2022transform} additionally replace the Discrete Fourier Transform as used by FNOs with the Discrete Cosine Transform for its energy compaction properties and real-valued output, whereas \citet{gupta2021multiwaveletbased} construct neural operators using the multiwavelet transform (MWT). \citet{brandstetter2022geometric} use a variant of the Fourier transform, the Clifford Fourier Transform, in their CFNO to encode geometric relationships between the scalar and vector fields describing the PDE solution. Similarly, \citet{helwig2023group} utilize the geometric principal of symmetry in their $G$-FNO to perform rotation and reflection equivariant convolutions in the frequency domain, which we discuss further in~\cref{sec:phySym}. Several works have proposed U-Net and FNO hybrids, such as U-FNO~\cite{wen2022u}, UNO~\cite{rahman2022u}, and U-F2Net~\cite{gupta2022towards}.}
    
    \citet{guibas2021adaptive} extend FNOs into the vision transformer framework~\cite{dosovitskiy2021an} \revisionOne{with the Adaptive Fourier Neural Operator (AFNO)}, which was later extended to forecasting global weather by \citet{pathak2022fourcastnet} with their FourCastNet architecture. To efficiently leverage attention as a multi-scale processing mechanism, \revisionOne{AFNO} uses the Fourier transform as an inexpensive token mixer. \citet{bi2022pangu} also build on vision transformers for weather forecasting, instead using patch embedding \jacob{to reduce dimensionality along the spatial dimensions} in their 3D Earth Specific Transformer (3DEST) to manage the quadratic complexity incurred by attention. \citet{lam2022graphcast} propose GraphCast in the same setting, a GNN operating on a graph representing the state of the global weather\revisionOne{. The edge set for GraphCast contains} seven different lengths of edges for efficiently passing messages long-range, ranging from a few edges spanning long distances to hundreds of thousands of localized short edges. As weather phenomena range from localized blizzards to heatwaves spanning multiple continents~\cite{gupta2022towards}, and the training data spans nearly a half-century~\cite{hersbach2020era5}, efficient multi-scale processing is particularly important for the task considered by these works. In large part due to effective choice of this processing mechanism, each of these models outperform the numerical weather prediction model currently used for delivering real-world forecasts on various tasks while operating at a fraction of the cost~\cite{bi2022pangu, lam2022graphcast, pathak2022fourcastnet}.

    \jjjacob{Similar to~\citet{pathak2022fourcastnet} and~\citet{bi2022pangu}, \citet{nguyen2023climax} employ vision transformers for weather and climate modeling. However, instead of focusing on one specific task where the spatial domain, input variables, and target variables are fixed, \citet{nguyen2023climax} leverage multiple climate and weather datasets spanning a variety of tasks in pre-training ClimaX, a climate and weather foundation model. Since the variables to be modeled vary from dataset to dataset, \citet{nguyen2023climax} propose a flexible encoding scheme which first tokenizes each  variable independently before mixing tokens using cross attention with a learned query. ClimaX is also trained to predict a variety of lead times, that is, the duration of time between the input state and the target state. \citet{nguyen2023climax} demonstrate the ability of ClimaX to be fine-tuned for tasks diverse from pre-training tasks, including forecasting on regional and global spatial scales and predictions with lead times ranging from a few hours to more than a month. Furthermore, \citet{nguyen2023climax} show that pre-training ClimaX improves the fine-tuned performance on downstream tasks compared to directly training a randomly initialized version of their architecture for that task.}


    \revisionOne{Several works have extended physics foundation models beyond weather and climate by pre-training on diverse governing equations for few-shot learning of downstream dynamics. \citet{subramanian2024towards} utilize the FNO architecture~\cite{li2021fourier}, while \citet{hao2024dpot} scalably integrate ViT attention by building on the AFNO~\cite{guibas2021adaptive}. \citet{mccabe2023multiple} also use a ViT backbone, instead utilizing axial attention~\cite{huang2019ccnet,ho2019axial} along the spatial and temporal dimensions for efficiency in designing their Multiple Physics Pretrained Axial ViT (MPP-AViT) architecture. Alternatively, \citet{herde2024poseidon} manage complexity with downsampling and upsampling in composing shifted window attention layers~\cite{liu2021swin,liu2022swin} to construct a U-shaped, multi-scale, hierarchical foundation model referred to as \textit{Poseidon}. To maximize the number of pre-training examples, Poseidon adopts a similar training strategy to ClimaX~\cite{nguyen2023climax} in training to predict at all possible lead times. Each of these works demonstrates the advantages of fine-tuning versus training from scratch in terms of sample complexity for equations and tasks distinct from those observed during pre-training, including inverse problems~\cite{mccabe2023multiple}, time-independent PDEs~\cite{herde2024poseidon}, out-of-distribution PDE parameters~\cite{subramanian2024towards}, and number of spatial dimensions~\cite{hao2024dpot}. Scalability of the physics foundation model paradigm is furthermore demonstrated by experiments showing improving results with model size~\cite{subramanian2024towards,hao2024dpot,mccabe2023multiple,herde2024poseidon} as well as amount and diversity of pre-training data~\cite{herde2024poseidon}.}

\subsubsection{Existing Methods: Multi-Resolution Dynamics}\label{sec:mrDyn}

    \begin{figure}[t]
        \centering
        \includegraphics[width=\textwidth]{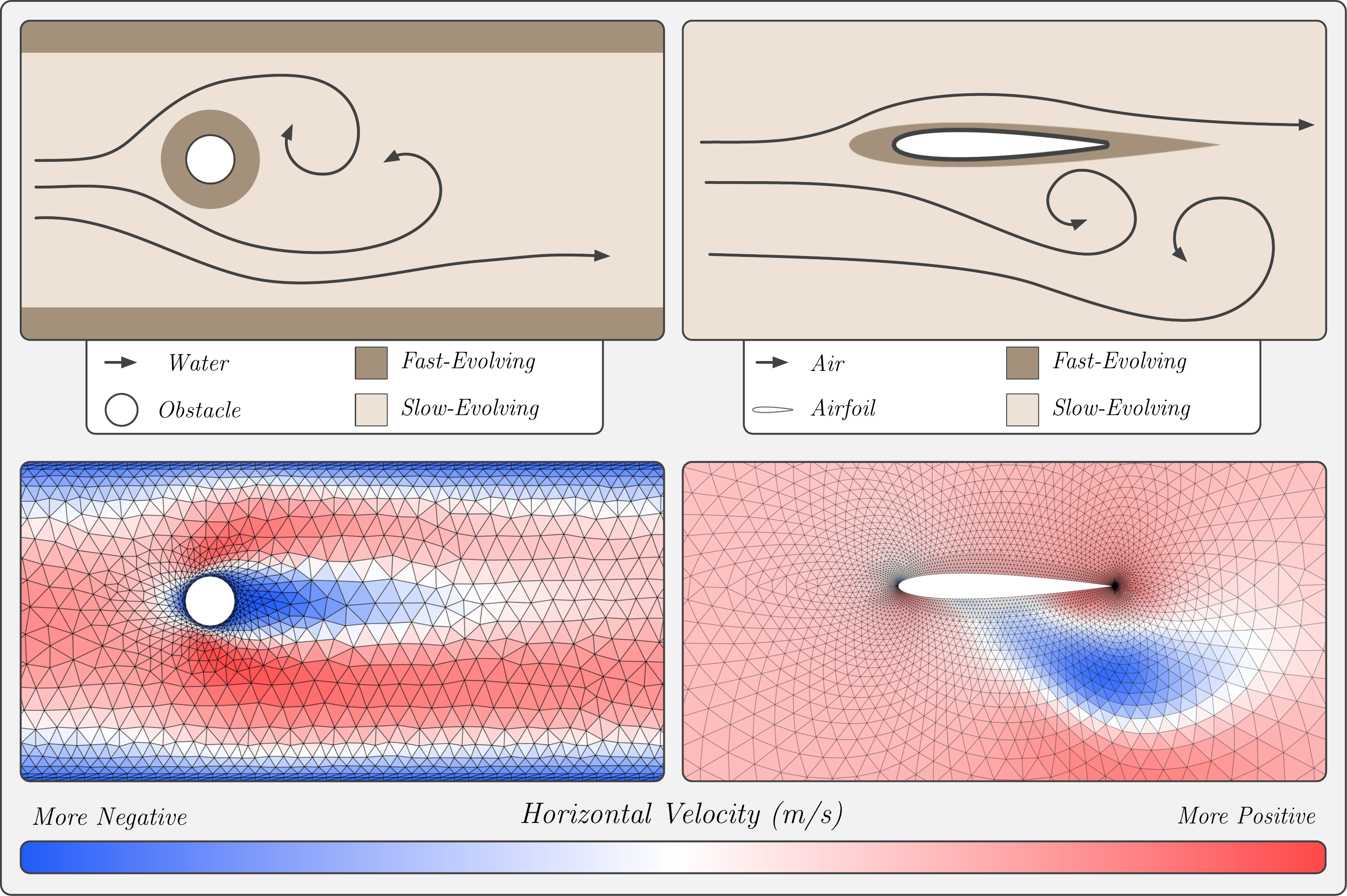}
        \caption{Multi-resolution dynamics, data from~\citet{pfaff2021learning}. Systems with localized regions of fast-evolving dynamics, such as fluid flow around a cylinder (left) or air flow around an airfoil (right), require high-resolution discretizations in these regions for dynamics to be stably resolved. Irregular discretizations of the domain can manage this cost by allocating high resolution in regions of high gradient and coarse resolution elsewhere. However, since irregularly discretized functions cannot be modeled by architectures such as CNNs, there has been a call for GNNs for simulating dynamics~\cite{pfaff2021learning}. Furthermore, the location in space of these high gradient regions can shift as the system evolves, thus requiring the discretization to dynamically adapt. While traditional re-meshing algorithms can be expensive, \citet{pfaff2021learning} and \citet{wu2022learning} propose learned alternatives for adaptive mesh refinement that reduce this cost.}\label{fig:mrDyn}
    \end{figure}

    The ability to model non-uniform discretizations of the PDE domain is important for balancing the tradeoff between computational cost and solution accuracy. Classical numerical solvers rely on the assumption that the solution is sufficiently smooth between collocation points to maintain stability~\cite{kochkov2021machine}. However, phenomena such as shockwaves and solid objects impeding flows, as we visualize in~\cref{fig:mrDyn}, introduce local regions of steep gradient that require expensive high-resolution discretizations to maintain this smoothness~\cite{berger1984adaptive}. Uniform discretizations wastefully allocate the same high resolution required in these isolated regions even in areas where the dynamics are slower-evolving~\cite{wu2022learning}. To address this limitation, non-uniform meshes allocate fine resolution in high-gradient regions and coarse resolution elsewhere. Furthermore, the geometry of the mesh can adapt as high-gradient regions shift in space, as is commonly the case for time-evolving PDEs~\cite{berger1984adaptive}. 

    While machine learning methods have been shown to allow for coarser discretizations than numerical methods due to their ability to learn a direct mapping~\cite{kochkov2021machine, stachenfeld2021learned}, neural networks still benefit from a certain level of continuity in the solution space. In addition to the inefficiency for dynamics with isolated regions of high gradient, surrogate models built on the CNN architecture
    cannot directly model non-rectangular domains, for example, fluid flow around a cylinder. These limitations have given rise to a number of approaches for modeling dynamics on non-uniform meshes and learned mesh adaptation.
    
    \jacob{Due to their ability to learn with unstructured data, GNNs~\cite{kipf2017semisupervised,gilmer2017neural} have been a primary choice for modeling dynamics on irregular meshes.} \citet{pfaff2021learning} take \jacob{this approach in their MeshGraphNets framework, where they train a message-passing GNN to model a time-evolving PDE autoregressively. The input graph for MeshGraphNets is constructed with the nodes representing the PDE solution evaluated on a non-uniform mesh at the current time step, and the target nodes as the solution at a future time step. The MeshGraphNets framework} additionally involves a second GNN that predicts the \textit{sizing tensor} for each node. The predicted sizing tensor is then used by an adaptive mesh refinement algorithm to update the mesh geometry as dynamics evolve. However, this re-meshing algorithm is expensive and furthermore may not produce the optimal geometry in terms of the tradeoff between solution accuracy and cost of the mesh~\cite{wu2022learning}. Thus, \citet{wu2022learning} propose Learning controllable Adaptive simulation for Multi-resolution Physics (LAMP), a faster, data-driven approach to re-meshing using reinforcement learning. They jointly optimize the dynamics GNN and a mesh refinement policy, where the policy is selected by simultaneously minimizing the error of the dynamics GNN and the number of nodes in the mesh. \jacob{Beyond standard message passing, \citet{janny2023eagle} utilize global attention in their dynamics GNN to model turbulent flows on a mesh, managing complexity by pooling nodes prior to attention computations.}

    \jacob{DeepONets are a general operator learning framework beyond GNNs developed based on the Universal Approximation Theorem for Operators~\cite{chen1995universal} which have demonstrated success on a variety of operator regression tasks including learning the solution operator for PDEs~\cite{lu2019deeponet}. The input to DeepONet is a function $v$ discretized onto an arbitrary geometry and a query point $x$. DeepONets then aim to map $v$ to a given target function $u(x)$ evaluated at point $x$. In the context \revisionOne{of} a PDE, $v$ may be the boundary conditions, initial conditions, or forcing term, and $u$ is the PDE solution. The primary components in the DeepONet framework are a branch network $\mathbf h$, which encodes $v$ to $\mathbf h(v)\in\mathbb R^p$, and a trunk network $\mathbf k$ to encode $x$ to $\mathbf k(x)\in\mathbb R^p$. While the trunk network is often chosen to be a MLP, the architecture of the branch network can be freely chosen dependent on the discretization of the input function. For example, if $v$ is discretized onto a regular grid, $\mathbf h$ can be a CNN, while a GNN branch network can be used for irregular discretizations~\cite{lu2022comprehensive}. DeepONet then takes the dot product of the output vectors from the two networks to approximate the target function evaluated at the query point as $u(x)\approx\mathbf h(v)\cdot\mathbf k(x)$. Intuitively, while the branch network $\mathbf h$ learns basis coefficients conditioned on $v$, the trunk network learns basis functions evaluated at point $x$, and can even be replaced with a fixed basis determined by the training data, such as the Proper Orthogonal Decomposition~\cite{bhattacharya2021model} as in \citet{lu2022comprehensive}.}
    
    \jacob{The DeepONet framework is flexible, with the only constraint being that the discretization of the input function be identical for all training pairs, however, several works have extended DeepONet to relieve this constraint~\cite{kovachki2021neural}. This mesh-independence is a defining factor for neural operators, which, as discussed in~\cref{sec:msDyn}, aim to learn PDE solution operators while maintaining the ability to generalize beyond the discretization of the training data~\cite{kovachki2021neural}. The attention mechanism has been shown to be a special case of a neural operator layer~\cite{kovachki2021neural}, and thus, there have been several works developing transformer-based frameworks for modeling dynamics on irregular meshes. For $n$ tokens with a $d$-dimensional embedding, \citet{cao2021choose} removes softmax and views the $n\times d$ query, key, and value matrices as $d$ learned basis functions evaluated at $n$ mesh points. Attention calculations then facilitate an integration-based interpretation, bearing \revisionOne{resemblances} to a Fourier-type kernel integral transform or a Petrov-Galerkin-type projection.} 
    
    \jacob{Under this interpretation, \citet{cao2021choose} propose Fourier-type attention and Galerkin-type attention, the latter of which has linear complexity and is proven to possess quasi-optimal approximation capacity. However, since both Fourier and Galerkin-type attention are based on self-attention, they cannot handle the setting where the discretization of the input function $v$ differs from that of the target function $u$, \textit{i.e.}, the query points $x$. OFormer~\cite{li2022transformer} therefore leverages cross-attention to adapt Galerkin-type attention to this case, with keys and values as the embedded discretization of $v$ and the queries as the $x$ embeddings. This conditions the embeddings for the query points $x$ on the input function, a formulation shown to have connections to DeepONet. \citet{li2022transformer} \revisionOne{empirically} demonstrate that this spatial encoding is sufficiently expressive such that given the embeddings, the PDE is reduced to an ODE in latent space. Specifically, following the embedding of $x$, temporal evolution of the system no longer requires spatial updates and can be accomplished in latent space simply through recurrent application of a point-wise MLP to each of the embeddings.} 
    
    \jacob{\citet{hao2023gnot} build on OFormer by extending to the case where there are multiple input functions that are discretized on different grids in constructing their General Neural Operator Transformer (GNOT). GNOT furthermore replaces the point-wise MLP applied in Transformer encoder layers with a Geometric Gating Mechanism. This mechanism consists of a mixture of MLPs, where the mixture weights for a given query point $x$ are determined by a gating MLP, effectively permitting the point-wise update to vary with $x$.} 
    
    \jacob{GNNs have also played a fundamental role in the development of neural operators. \citet{li2020neural} demonstrate a connection between message passing in GNNs and the Green's function formulation of the PDE solution in their Graph Kernel Network (GKN). Under this formulation, the solution of a linear PDE can be written as an integral involving a kernel function. Then, the Monte Carlo approximation to this integral produces a sum that closely resembles message passing given an assumption of locality on the integral, that is, by restricting neighbors of each mesh point to be the mesh points within a fixed radius. Because the Monte Carlo approximation includes normalization by the neighborhood size, learned message passing in GKN can generalize well to graphs with an arbitrary number of neighbors per node, an ability which becomes relevant when performing inference on a mesh with a resolution differing from that observed during training. \citet{li2020neural} extend the Green's function formulation to non-linear PDEs through the inclusion of non-linearities, and improve the efficiency of GKN through a Nystr\"om approximation of the kernel which reduces the Monte Carlo sum over the full neighborhood to a sum over randomly sampled sub-neighborhoods.} 
    
    \jacob{While the locality assumption imposed on integrals by GKN ensures computational efficiency, it does not allow long-range interactions between distant mesh points and thus cannot capture global properties of the solution operator~\cite{li2020multipole}. \citet{li2020multipole} therefore propose the Multipole Graph Kernel Network (MGKN) for efficiently modeling long-range interactions. MGKN is centered on a V-cycle algorithm inspired by multi-grid methods~\cite{han2021predicting} and the classical Fast Multipole Method, a method originally developed to approximate pairwise interactions for $n$-body simulation in $\mathcal O(n)$ time using a hierarchical decomposition of space to model increasingly distant particle interactions~\cite{cipra2000best,greengard1987fast}. The V-cycle algorithm operates similar to a graph U-Net~\cite{gao2019graph}, processing a hierarchy of graphs representing interactions at increasingly distant scales along a downsampling and upsampling path. MGKN composes multiple V-cycles wherein latent graphs are processed using message passing, with graphs processed at each scale of the downsampling path obtained as a subgraph of the graph processed at the previous scale.  \citet{li2020multipole} demonstrate improved performance and efficiency of MGKN relative to GKN. Additionally, since message passing on a given graph in MGKN is done identically to GKN, MGKN retains the ability to generalize to discretizations differing from those observed in training.}

    \jacob{\citet{li2022fourier} instead look to model long-range interactions on irregular geometries using global Fourier convolutions similar to the Fourier Neural Operator (FNO) discussed in~\cref{sec:msDyn}, a challenge since the Fast Fourier Transform is restricted to regular grids. The Geometry-aware FNO (Geo-FNO) therefore uses a geometric Fourier transform in its first and final convolution layers. The geometric transform in the first layer maps from a function on the irregularly-meshed spatial domain to a function in the frequency domain corresponding to a uniform grid using a learned mapping from mesh coordinates to uniform grid coordinates. This enables use of the computationally-efficient Fast Fourier Transform for the remaining convolutions. In the final layer, the inverse geometric transform is applied to map back to the irregularly-meshed spatial domain. This architecture was later optimized for depth alongside the conventional FNO by \citet{tran2021factorized} with their Factorized FNO (F-FNO). }

    \jacob{Despite its efficiency in capturing long-range interactions on irregular geometries, use of the geometric Fourier transform by Geo-FNO prevents it from generalizing well to discretizations differing from those observed during training~\cite{li2023geometry}. \citet{li2023geometry} thus integrate the GKN and FNO architectures in the Geometry-Informed Neural Operator (GINO), which leverages a GKN encoder and decoder with an FNO processor in latent space. Specifically, a GKN module is used to obtain a latent representation on the input irregular mesh. Because the radius graph-based message passing utilized by GKN allows for arbitrary mesh points to be queried, this irregularly-meshed representation can be queried on a regular grid. The FNO processor in GKN can then process long-range interactions in this representation efficiently using the fast Fourier transform. Finally, a second GKN is applied to the output regularly-meshed representation to map back to the input geometry following an analogous approach to the encoder GKN. This framework allows GINO to generalize beyond the resolution of the training data while also enabling modeling of long-range dependencies on irregular meshes. \citet{li2023geometry} demonstrate the ability of GINO to accurately model turbulent dynamics on irregular geometries in 3 spatial dimensions with a large number of mesh points. }


\subsubsection{Existing Methods: Long-Term Stability}\label{sec:roSty}

    Time-evolving PDEs are numerically solved by discretizing the temporal domain into time steps on which the solver produces solutions. \revisionOne{As discussed in~\cref{sec:pde_overview}, t}his solution can be obtained using an \revisionOne{\textit{explicit}} scheme, wherein the solution at a given time point is directly calculated using the preceding solution as $u_{t+1} = F\left(u_t\right)$ for some function $F$, or \revisionOne{\textit{implicit}} schemes, which entail solving a system of (possibly non-linear) equations involving $u_t$ and $u_{t+1}$~\cite{olver2014introduction}. Although explicit schemes appear to require less computational effort than implicit, classical solvers that advance time using an explicit scheme can exhibit \revisionOne{\textit{conditional stability}}, meaning that the discretization in time must be chosen sufficiently fine to prevent the solver from diverging ~\cite{cfl1928,olver2014introduction}. Such PDEs for which explicit methods require significantly finer time discretizations compared to the smoothness of the actual solution are termed \emph{stiff}. As a result, implicit methods have been traditionally preferred for solving stiff PDEs. Nonetheless, many neural surrogates utilize explicit schemes for convenience, and have been shown to outperform classical solvers on computationally inexpensive coarse discretizations with large time steps~\cite{kochkov2021machine, stachenfeld2021learned}. However, explicit schemes inevitably introduce error to the inputs of the model, and thus, increasing the robustness of neural solvers to noisy inputs is key for enabling stable predictions over many time steps. 

    The task often considered in this setting is to predict the rollout up to time step $T$ conditioned on the first $k$ solutions, that is, the mapping {$(u_0, u_1,\ldots,u_{k-1})\mapsto(u_k,u_{k+1},\ldots,u_T)$}. In what follows, we take $k=1$ for simplicity in notation such that the mapping to be learned is from the time $0$ solution to the remaining $T-1$ steps, but this is not necessary in general. For such a task, explicit schemes train $\phi_\theta$ for a one-step prediction of $u_{t+1}$ conditioned on the ground-truth solution at $u_t$, and at test time predict the full rollout by applying the trained network autoregressively $T$ times~\cite{li2021learning,brandstetter2022message, sanchez2020learning, stachenfeld2021learned,lippe2023pde,kohl2023turbulent}. This one-step training strategy has been shown to be more effective than training recurrently to predict the full rollout $u_1,u_2,\ldots,u_T$~\cite{tran2021factorized}. However, it is not representative of the task at test time, since for $t>1$, the input to the model will not be the ground truth $u_t$ as in training, but rather $u_t+\varepsilon_t$, where $\varepsilon_t$ is the error accumulated up to time $t$ through autoregressive prediction of $u_t$. \jjacob{Because $\varepsilon_t$ is often monotonically increasing with $t$, rolling out chaotic dynamics such as turbulent flows for many time steps can be prohibitively difficult for machine learning methods to do accurately in terms of mean squared error. However, stable, long-time simulation of dynamics that exhibit accurate behavior elsewhere can hold value, for example, in terms of the Fourier spectrum~\cite{li2021learning,lippe2023pde,kohl2023turbulent}, principal components~\cite{li2021learning}, Pearson correlation~\cite{lippe2023pde}, rate of change~\cite{kohl2023turbulent}, and other summary statistics which characterize the behavior of the system over long time horizons.}
    
    \jjacob{\citet{li2021learning} study chaotic dynamics with the physical property of \textit{dissipativity}, which ensures that regardless of their initial conditions, given sufficient evolution, the dynamics will eventually arrive and remain in a particular set of states referred to as the \revisionOne{\textit{absorbing set}}, allowing reproducible statistics to be computed even for trajectories with diverse starting points. \citet{li2021learning} therefore look to induce dissipativity in training their Markov Neural Operator (MNO), enabling accurate computation of statistics from predicted rollouts. MNO is an autoregressive instantiation of FNO with soft and hard dissipativity constraints. During training, the soft constraint is applied in the form of a dissipativity-inducing loss, whereas the hard constraint is an unlearned post-processing step which forces the predicted trajectory back into a pre-determined stable region should MNO predict a transition outside of this region. MNO is also trained using a Sobolev loss, which is suggested to more reliably capture high-frequency details in the dynamics.}

    \jjacob{Other works have also taken regularization-based approaches to enhancing long-term stability, including} the Lyapunov regularizer ~\cite{Rong2022Lyapunov} and
    adversarial noise injection~\cite{sanchez2020learning, brandstetter2022message}. Noise injection approaches intentionally corrupt inputs during training with an approximation to \jjacob{the prediction error} $\varepsilon_t$. \citet{sanchez2020learning} apply this strategy in training their Graph Network-based Simulator (GNS) by assuming that $\varepsilon_t$ follows a 0-mean Gaussian distribution with variance chosen as a hyperparameter. Although this approach is convenient since the noise distribution can be  easily sampled, the normality assumption may not be valid, and furthermore, the variance hyperparameter \jjacob{controlling the noise level} must be carefully tuned. \citet{brandstetter2022message} instead obtain the noise to train their Message Passing PDE solver (MP-PDE) directly from the model to reduce the distributional shift between the training noise and the test noise. They accomplish this by letting $\varepsilon_t=\phi_\theta\left(u_{t-1}\right)-u_t$ such that the input to the model is $u_t+\varepsilon_t=\phi_\theta\left(u_{t-1}\right)$. Thus, the noise added to the inputs during training is directly sourced from the model, as it will be during testing. 
    
    \jjacob{\citet{lippe2023pde} maintain the normality assumption in $\varepsilon_t$ in their PDE-Refiner framework and propose to choose several noise levels based on the amplitudes in the Fourier spectra of the PDE solution. This sequential noise injection strategy is motivated by analysis showing that standard MSE-loss with 1-step training objectives neglects frequency components with smaller amplitudes, and that errors on these components accumulate over many time steps, permeating into higher-amplitude components where errors are more noticeable. To improve the 1-step prediction of smaller amplitudes, \citet{lippe2023pde} apply their neural solver multiple times to advance the rollout by 1 time step, with each forward pass following the first one serving to iteratively refine the solver's initial prediction on increasingly smaller amplitudes. Since smaller noise levels correspond to smaller amplitudes, \citet{lippe2023pde} achieve this by injecting noise at increasingly smaller levels to the model's predictions at each refinement step during both training and inference, where, similar to denoising diffusion models~\cite{ho2020denoising}, the prediction target for refinement steps is the injected noise which is subsequently subtracted away from the predicted solution during inference. On several time-evolving fluid dynamics tasks, PDE-Refiner demonstrates not only substantial increases in rollout stability using only 3 refinement steps, but also improved sample complexity and the ability to detect instability in predictions.} 

    \jjacob{\citet{kohl2023turbulent} also employ diffusion models in the setting of time-evolving fluid dynamics for enhanced rollout stability. However, instead of a refinement-based approach, \citet{kohl2023turbulent} frame 1-step prediction as a conditional generation task, where their Autoregressive Conditional Diffusion Model (ACDM) is tasked with denoising both the state at the next time step as well as the previously predicted states with noise added. By learning to denoise the conditioning states, \citet{kohl2023turbulent} aim to increase the robustness of ACDM to errors $\varepsilon_t$ introduced in prediction of the previous states. \citet{kohl2023turbulent} show that ACDM can preserve the correct statistics for the flow over longer rollouts in a variety of fluid simulation settings. Additionally, the ability of ACDM to produce diverse yet physically-consistent samples from the posterior distribution of PDE solutions is demonstrated, an ability relevant for uncertainty quantification.}

    \jjacob{While~\citet{tran2021factorized} found training for recurrent prediction of full rollouts to be suboptimal compared to 1-step prediction, several works have trained models to recurrently predict only a few steps ahead~\cite{wu2022subsurface,lam2022graphcast}.} \citet{wu2022subsurface} \jjacob{implement this strategy in} optimizing their Hybrid Graph Network Simulator (HGNS) using a multi-step objective. The total loss is a weighted sum of the loss at each time step, with the one-step loss weighted heaviest so that the optimization initially targets short-term predictions before fine-tuning for longer-term predictions. This method is more stable than full recurrent prediction since it only predicts several steps ahead, and not the full rollout. 
    
    \jacob{Several works have successfully adopted fully-recurrent prediction in latent space through the use of autoencoders~\cite{han2021predicting,wu2022accel}.  \citet{han2021predicting} specifically consider rolling out irregularly-meshed dynamics for many time steps with their Graph Mesh Reducer Transformer (GMR-Transformer). In this framework, a GNN encoder and decoder are trained in an autoencoding fashion to map the system at a given timestep to and from a latent vector, respectively. This reduced latent representation is memory-efficient and therefore allows \citet{han2021predicting} to recurrently train a Transformer to predict the latent vectors for each state of the rollout, which can subsequently be upsampled from the latent space using the GNN decoder. GMR-Transformer demonstrates the ability to accurately predict rollouts of fluid dynamics in irregularly-meshed domains for hundreds of time steps.}
    
    While the previously discussed architectures predict only one step ahead, \citet{brandstetter2022message} make the observation that since each forward propagation introduces some error, reducing the number of calls to the model required to predict a rollout could reduce the total accumulated error. Instead of predicting only one time step ahead as $\phi_\theta\left(u_t\right)=\hat u_{t+1}$, \citet{brandstetter2022message} train their model to predict $l$ steps ahead with one forward propagation as {$\phi_\theta\left(u_t\right)=\left(\hat u_{t+1}, \hat u_{t+2},\ldots,\hat u_{t+l}\right)$}. For example, to predict $10$ timesteps with $l=2$, only 5 forward propagations are required instead of 10. \jacob{Following a similar philosophy, \citet{bi2022pangu} introduce hierarchical temporal aggregation for their learned weather forecasting model, wherein several models are trained with different time step sizes. At inference time, the rollout is divided between the models such that the minimal number of forward passes are required to advance the system forward to the target lead time.}

\subsubsection{Existing Methods: Preserving Symmetries}\label{sec:phySym}
    Dynamic systems are governed by the laws of physics, with symmetries of systems related to these laws through Noether's theorem~\cite{noether1971invariant, wang2021incorporating}. The symmetry group of a PDE characterizes the transformations under which solutions remain solutions, \emph{e.g.}, for a PDE with rotation symmetry, rotating the solution function produces a function that is also a solution. Symmetries such as rotation invariance are understood intuitively as the lack of canonical reference frame that allows, for example, a 2-dimensional flow rotated by $90^\circ$ to remain equally physically plausible. Other symmetries such as translation invariance arise in PDEs with infinite domains or periodic boundaries~\cite{holmes2012turbulence}. Priors that enforce symmetries can improve generalization and sample complexity by reducing the size of the \revisionOne{model search} space~\cite{raissi2019physics, wang2021incorporating,brandstetter2022lie}. \jjjacob{Furthermore, PDEs with spherical domains commonly arising in global weather forecasting applications~\cite{esteves2023scaling,bonev2023spherical} have led to the application of tailored architectures to ensure that symmetries are preserved.}

    As a method of instilling learned equivariance, \citet{brandstetter2022lie} propose \jjacob{Lie Point Symmetry Data Augmentation (LPSDA) to improve sample complexity and the generalization ability of neural solvers by leveraging symmetries of the PDE. Similarly, \citet{akhound2023lie} introduce a symmetry loss in training the Physics-Informed DeepONet proposed by \citet{wang2021learning}, allowing the network to learn a family of PDE solutions related by a symmetry transformation by learning only one member. }

    Equivariant CNNs, which are composed of convolutional layers that automatically encode the desired symmetry~\cite{ cohen2016group, Cohen2017-STEER, 3d_steerableCNNs, Weiler2019_E2CNN, worrall2019deep, weiler2023EquivariantAndCoordinateIndependentCNNs}, \jjacob{present an alternative path to achieving equivariance.} \citet{wang2021incorporating} consider a variety of symmetries in constructing their Equ-ResNet and Equ-Unet for dynamics forecasting, including \textit{exact} scale and rotation symmetries. However, dynamics often only exhibit approximate symmetries due to, for example, external forces~\cite{wang2022approximately}. \citet{wang2022approximately} thus relax equivariance constraints \jjjacob{in 2 spatial dimensions} in constructing their RGroup and RSteer CNNs for approximately equivariant group and steerable convolutions, respectively~\cite{cohen2016group,Cohen2017-STEER}. \jjjacob{\citet{wang2022approximately} learn their convolution kernels $\psi(h)$ as a linear combination of equivariant kernels $\psi_l(h)$, where the weight $w_l(h)$ corresponding to the $l$-th kernel in the linear combination is learned and is itself a function on the group as opposed to a scalar value. In doing so, exact equivariance is recovered when the $w_l$ are identical for all $l$. \citet{wang2023relaxed} demonstrated that the learned $w_l$ adapt as expected when the symmetry of the mapping to be learned is equal to, less than, or completely absent relative to the symmetry encoded by the network. \citet{wang2023relaxed} furthermore extended approximately rotation-equivariant group convolutions to 3 spatial dimensions with the R-Equiv architecture following a similar approach as in 2 spatial dimensions. However, the number of parameters increases in each kernel since the number of possible $90^\circ$ rotations increases from 2 dimensions to 3. Therefore, to improve parameter efficiency, \citet{wang2023relaxed} apply a rank-1 tensor decomposition to their relaxed kernels, as is done in separable group convolutions~\cite{knigge2022exploiting}. In experiments on turbulent flows with a range of symmetry levels, \citet{wang2023relaxed} highlight the advantages of their approach and furthermore offer interpretability in the weights $w_l$ as to how dynamics break symmetries.} \jjacob{Beyond equivariant convolutions, \citet{holderrieth2021equivariant} equivariantly model stochastic fields by extending steerability constraints to Gaussian Processes and Conditional Neural Processes. }
    
    Instead of steerable or group convolutions, \citet{ruhe2023clifford} \jjacob{encode symmetries using Clifford algebras. Multivectors, the elements of Clifford algebras, have scalar components, vector components, and higher-order components representing plane and volume segments, with multivector multiplication defined with the \textit{geometric product}~\cite{brandstetter2023clifford}. \citet{brandstetter2023clifford} note that the standard practice of stacking vector and scalar fields comprising PDE solutions along the channel dimension in neural solvers does not model the geometric relationships between fields well. Instead, \citet{brandstetter2023clifford} represent these fields as multivectors, resulting in multivector feature maps and kernels that are convolved with Clifford CNN layers and Clifford FNO layers operating via the geometric product in the CResNet and CFNO architectures, respectively.} 
    
    \jjacob{As dynamics tasks often involve a target which is a geometric transformation of the input, \citet{ruhe2023geometric} build on \citet{brandstetter2023clifford} by learning compositions of transformations with their Geometric Clifford Algebra Network (GCAN). The construction of the GCAN is based on the result that transformations of an arbitrary multivector $v$ by $g\in E(n)$ (\textit{i.e.}, rotations, reflections, and translations) can be achieved through geometric products of $v$ with other multivectors chosen dependent on $g$. Group action layers in the GCAN therefore apply learned transformations to the input multivector through geometric products with learned multivectors, where the range of possible transformations is determined by the choice of the basis for the algebra, ranging from $E(n)$ to $SO(n)$. Through appropriate selection of multivector representation for various data types, \citet{ruhe2023geometric} demonstrate the flexibility of the GCAN, allowing for simulation of rigid body transformations with GCA-MLP and GCA-GNN, as well as simulation of fluid dynamics with GCA-CNN.}
    
    \jjacob{\citet{ruhe2023clifford} extend this work by using Clifford algebras in the derivation of their $O(n)$-equivariant Clifford Group Equivariant Neural Network (CGENN). Unlike previous equivariant architectures, CGENNs achieve equivariance through symmetry properties of several multivector operations. Specifically, \citet{ruhe2023clifford} prove the $O(n)$-equivariance of the geometric product, multivector grade projections, wherein all the multivector components excluding the $k$-th order part are set to 0, and polynomial functions of multivectors. Equivariant linear layers in the CGENN architecture then learn coefficients used to linearly combine grade projections of multivectors comprising the channels of neural representations. Additionally, Geometric Product Layers consider pair-wise interactions between channels by learning to linearly combine compositions of various grade projections with geometric products applied to channels. \citet{ruhe2023clifford} demonstrate performance gains with the CGENN architecture on a variety of diverse tasks wherein symmetries play a role, including $n$-body simulation in 3 spatial dimensions. } 
    
    While the previously discussed \jjacob{equivariant} CNNs perform convolution in physical space, \citet{helwig2023group} extend group equivariant convolutions~\cite{cohen2016group} to a frequency domain parameterization with the $G$-FNO architecture. \jjacob{$G$-equivariant convolutions convolve kernels and feature maps that are functions on the group $G$, whereas the discrete Fourier transform $\mathcal F$ is only defined for functions on the grid $\mathbb Z^d$, thereby complicating the use of the Convolution Theorem which enables Fourier-space convolutions in FNOs. However, the groups considered by \citet{helwig2023group} are the semi-direct product of the plane $\mathbb Z^2$ with a subgroup $S$, where $S$ is either $90^\circ$ rotations for $G=p4$ or roto-reflections for $G=p4m$. Using this decomposition, group convolutions can be expressed as a sum of planar convolutions with a kernel $\psi$ transformed by an element of $S$. Applying the Convolution Theorem to these planar convolutions gives point-wise multiplication with the Fourier transform of the kernel transformed by an element $s$ of $S$, $\mathcal FL_s\psi$. Finally, \citet{helwig2023group} apply symmetries of the Fourier transform which allow orthogonal transformations, including elements of $S$, to commute with the Fourier transform, giving $L_s\mathcal F\psi$ and enabling equivariant convolutions parameterized in the frequency domain.} In addition to the benefits brought in terms of multi-scale processing as discussed in~\cref{sec:msDyn}, this allows for superior generalization to discretizations with different resolution relative to physically parameterized alternatives~\cite{li2021fourier}.

    \jjjacob{Similar to FNOs, spherical CNNs~\cite{cohen2018spherical,kondor2018clebsch,esteves2020spin} learn the \textit{generalized} Fourier transform of convolution kernels to address challenges associated with convolving spherical signals in a rotation-equivariant manner. Spherical data arises in the context of PDEs for global climate and weather forecasting tasks, where the fields to be modeled, such as wind velocity or air pressure, are defined on the globe. However, application of conventional CNNs to such spherical data requires that it be projected into the plane, resulting in distorations~\cite{cohen2018spherical}. The vision transformer-based weather forecasting methods discussed in~\cref{sec:msDyn} encounter the same distortion due to their reliance on splitting the input spherical fields into sequences of equal-sized, square patches~\cite{dosovitskiy2021an}. Furthermore, it is difficult to correctly model boundary conditions after projecting from the sphere to the plane, as boundary points which appear spatially distant following the projection may be immediately adjacent in reality.} 
    
    \jjjacob{To address these challenges, \citet{bonev2023spherical} introduce the spherical FNO (SFNO), a spherical CNN for modeling global weather, and demonstrate the strengths of their architecture in stably generating year-long forecasts more than 1,000 time steps in length. \citet{esteves2023scaling} similarly employ spin-weighted spherical CNNs (SWSCNNs)~\cite{esteves2020spin} for global forecasting, and introduce several enhancements including an optimized calculation of the generalized Fourier transform, spectral batch norm, spectral pooling, and spectral residual connections which enable training of spherical CNNs at scale. Additionally, SWSCNNs equivariantly model vector fields on the sphere~\cite{esteves2020spin}, a challenge beyond scalar fields since the rotation of individual vectors must be considered in addition to rotation of the field. This ability could enhance modeling of physical quantities such as the vector\revisionOne{-valued} velocity field for global winds.}

    \jjacob{\citet{horie2021isometric} further consider modeling dynamics on irregular geometries with their IsoGCN architecture, an $E(n)$-equivariant GNN which simplifies Tensor Field Networks~\cite{thomas2018tensor} for improved space-time complexity by not relying on spherical harmonics and by building on the linear message passing scheme used in Graph Convolutional Networks~\cite{kipf2017semisupervised}. IsoGCN leverages the IsoAM, an $E(n)$-equivariant adjacency matrix representation containing spatial information. In addition to equivariance, \citet{horie2021isometric} demonstrate physical motivation in their construction of the IsoAM by proving that convolution, contraction, and tensor product operators applied to IsoAM and the tensor field of node features can yield various differential operators applied to the tensor field, including the gradient, divergence, and Jacobian operators. As a result of the improved computational efficiency in IsoGCN, \citet{horie2021isometric} demonstrate the ability to simulate the evolution of a heat field on CAD objects in three spatial dimensions on a mesh with more than 1 million collocation points. \revisionOne{\citet{toshev2023learning} similarly consider three spatial dimensions, but instead utilize the $E(3)$-equivariant} SEGNN \revisionOne{architecture}~\cite{brandstetter2022geometric} to simulate \revisionOne{a} flow of fluid particles \revisionOne{under a Lagrangian scheme}.}
    
\subsubsection{Existing Methods: Incorporating Physics}\label{sec:inPhy}

While deep neural networks are universal function approximators, practitioners often have insight into the behavior of physical systems. By carefully designing architectures to automatically respect physical laws that these systems obey, the learning task is simplified and the ability of the network to generalize over similar systems is improved. This is because these rules are difficult to learn from data directly, particularly in the small data regime. Further, encoding physical laws often increases the interpretability of network outputs, as they can be directly related to concepts practitioners are familiar with, which is in stark contrast to the usual treatment of neural networks as a black box modeling tools.

Hamiltonian Neural Networks~(HNNs)~\citep{greydanus2019} incorporate physics knowledge in the form of Hamiltonian mechanics for faithful modeling of Hamiltonian systems. In general, these systems are described by position $q(t)$ and canonical momenta $p(t)$, which evolve in time according to Hamilton's equations as
\begin{align}
    \label{eqn:hamiltons}
    \dot{q} = \revisionOne{\partial_p H}, \
    \dot{p} = -\revisionOne{\partial_q H},
\end{align}
where $\dot{q}$ and $\dot{p}$ are the derivatives of $q$ and $p$ with respect to time. Hamiltonian systems are everywhere -- the motion of planets under the influence of gravity, particles impacted by electromagnetic forces, and blocks attached to springs all follow Hamiltonian mechanics. A key property of Hamiltonian systems is that as the system \revisionOne{state denoted by $u(t)=\left(q(t),p(t)\right)$} evolves over time, the Hamiltonian $\revisionOne{H(u(t))}$ is \emph{conserved}, that is, it remains constant. Loosely, the Hamiltonian $H$ captures the amount of energy in the system.
HNNs~\citep{greydanus2019} propose learning this Hamiltonian $H$ directly from dynamics data. \revisionOne{Instead of directly supervising their model $\phi_\theta$ such that $\phi_\theta(u(t))\approx H(u(t))$, \citet{greydanus2019} apply supervision on the gradients of the network to satisfy~\cref{eqn:hamiltons} as
\begin{equation}\label{eq:hnn_loss}
    \phi_\theta = \underset{\phi_\theta:\theta\in\Theta}{\arg\min}\:\mathbb E_{q^{(j)}, p^{(j)},t}\left[\left\lVert\begin{matrix}\partial_p\phi_\theta\left(u^{(j)}(t)\right)-\dot{q}^{(j)}(t) \\ \partial_q\phi_\theta\left(u^{(j)}(t)\right)+\dot{p}^{(j)}(t)\end{matrix}\right\rVert_2\right],
\end{equation}
where, similar to the approach taken by Physics Informed Neural Networks discussed later, the partial derivatives of $\phi_\theta$ in~\cref{eq:hnn_loss} are computed exactly using automatic differentiation.}
The time evolution of the Hamiltonian system is then computed by numerically integrating~\cref{eqn:hamiltons} \revisionOne{with $\phi_\theta$ in place of $H$} over time using an explicit Runge-Kutta method of order $4$.
The HNN model accurately learns the time evolution of simple Hamiltonian systems such as an oscillating pendulum without dissipating energy, making predictions that maintain consistency with Hamilton's equations. In contrast, a standard fully-connected neural network trained on the same data is unable to learn trajectories that conserve $H$, resulting in physically implausible predictions.


\revisionOne{\citet{sanchez2019hamiltonian} extend HNNs to the Neural ODE framework~\cite{chen2019neural} with their Hamiltonian ODE graph network (HOGN) by backpropagating through the numerical integrator such that the optimization of $\phi_\theta$ takes the form of
\begin{equation}\label{eq:hnnode_loss}
    \phi_\theta = \underset{\phi_\theta:\theta\in\Theta}{\arg\min}\:\mathbb E_{q^{(j)}, p^{(j)},t}\left[\mathcal L\left(\operatorname{Integrator}\left(\Delta_T, u^{(j)}(t), \left(
        \partial_p\phi_\theta , -\partial_q\phi_\theta
    \right)\right), u^{(j)}(t+\Delta_T)\right)\right],
\end{equation}
where the numerical integrator approximates the integration given by 
\begin{equation}\label{eq:num_int}
     u(t+\Delta_T) = u(t) + \int_{t}^{t+\Delta_T}\dot{u}(\tau)d\tau\approx\operatorname{Integrator}\left(\Delta_T, u(t), \dot{u}\right).
\end{equation}
Unlike~\cref{eq:hnn_loss}, \cref{eq:hnnode_loss} does not impose \textit{explicit} supervision on the temporal derivatives of $u(t)$, and instead learns these quantities \textit{implicitly} by backpropagating through the numerical integrator. As this implicit approach instead only requires the system state $u$ instead of the temporal derivatives $\dot{u}$ for model training, a number of works have extended it beyond Hamiltonian systems. These works commonly perform numerical integration using the forward Euler method given in~\cref{eq:fwd_eul}, as it is one of the most straightforward implementations of~\cref{eq:num_int}. \citet{sanchez2020learning,pfaff2021learning,toshev2024neural,toshev2024lagrangebench} train models to predict per-particle acceleration in Lagrangian simulations which are integrated twice to update particle positions. For Eulerian simulations, many works predict the residual $d_t$ between $u_{t}$ and $u_{t+1}$~\cite{pfaff2021learning,stachenfeld2021learned,lippe2023pde,price2023gencast}, where the update $u_{t+1}=u_t+d_t$ can be interpreted as integration with the forward Euler method.}

In theory, the current state $u(t)$ of a Hamiltonian system completely determines its state $u(t')$ at all future times $t' > t$. 
With this motivation, SympNets~\cite{jin2020sympnets} directly learn the mapping from the current system state $u(t)$ to the future system state $u(t')$ using symplectic normalizing flows to avoid integrating over time. This makes SympNets more efficient over longer rollouts, as they avoid accumulation of numerical errors present during numerical integration. Symplectic Recurrent Neural Networks~\cite{chen2020Symplectic} further improve upon Hamiltonian Neural Networks by using symplectic integrators such as the leapfrog method, which are a better fit for Hamiltonian systems than explicit Runge-Kutta schemes because they explicitly match the form of~\cref{eqn:hamiltons}. Thus, the symplectic integrator will conserve the learned Hamiltonian $H$ when integrating over time up to numerical precision. Further, \citet{chen2020Symplectic} propose training over longer rollouts produced by sampling the model recurrently instead of single-step predictions. This helps avoid the distributional shift problem inherent when recursively sampling from the model's predictions at each time step, as discussed in~\cref{sec:roSty}. Finally, to account for noise in the system observables, \citet{chen2020Symplectic} update the initial state $u_0 = (q_0, p_0)$ via gradient descent.
These modifications improve the accuracy of the HNN when modeling noisy, real-world systems.

Instead of general Hamiltonian systems, Action-Angle Networks~\cite{daigavane2022learning} leverage properties of the special class of \emph{integrable} Hamiltonian systems by learning a symplectic transformation of position $q$ and momenta $p$ to slow-varying action variables and fast-varying angle variables.
For integrable systems, the dynamics in the action-angle space are effectively linear, which makes it both easier to learn and more efficient to numerically integrate compared to HNNs and Neural ODEs~\cite{chen2019neural}.

As opposed to Hamiltonian systems, Lagrangian Neural Networks (LNN)~\cite{cranmer2020lagrangian}  model Lagrangian systems where the form of the canonical momenta $p$ is not necessarily known. Instead, Lagrangian mechanics provide the necessary insight to relate the time evolution of the position $q$ as
\begin{align}
    \label{eqn:lagrange}
    \ddot q = \left(\frac{\partial^2 L}{\partial \dot{q}^2}\right)^{-1}\left(\frac{\partial L}{\partial q} - \dot{q}\frac{\partial^2 L}{\partial \dot{q}\partial q}  \right),
\end{align}
where $\dot{q}$ and $\ddot{q}$ are the first and second derivatives of the position $q$ with respect to time, and $L(q, \dot{q})$ is the Lagrangian of the system. Analogous to HNNs with the Hamiltonian $H$, LNNs model the Lagrangian $L$ via a neural network. Then, by numerically integrating~\cref{eqn:lagrange}, the time evolution of $q(t)$ can be obtained.

Finally, \citet{sosanya2022dissipative}
augment HNNs to additionally predict a Rayleigh dissipation function $D$ together with the Hamiltonian $H$. This allows the network to capture external forces such as friction which dissipate energy. Such forces cannot be captured in the original HNN framework because the HNN learns energy-conserving dynamics. The Dissipative HNN shows improved performance on predicting the time evolution of damped spring-block systems and the velocity fields of ocean surface currents.

While Hamiltonian mechanics can describe a large number of physical systems, in many real world scenarios, the system may not be sufficiently well understood or only partially observed. If the underlying PDEs describing the system are only partially known, physical knowledge can be leveraged in a hybrid setup. In this context, deep neural networks can be applied in conjunction with PDE-based methods to learn the residual between assumed governing equations and observed data. 
A representative example is the APHYNITY framework proposed by \citet{yin2021augmenting}, which operates on the premise that dynamics can be decomposed into physical (known) and augmented (residual) components as
\begin{equation}
    \partial_tu + (\mathcal D + \phi_\theta)\left(x, t, u,\partial_xu, \partial_{xx}u, \ldots\right) = 0,
\end{equation}
where $\phi_\theta$ represents the data-driven component that complements the known operator $\mathcal D$. When learning the parameters of $\phi_\theta$, numerical integration is used to generate predictions at various steps based on $\mathcal D + \phi_\theta$ given an initial state. More importantly, it efficiently augments physical models with deep data-driven networks in such a way that the data-driven model only models what cannot be captured by the physical model. To achieve this, other than prediction loss, an additional L2 norm term $\|\phi_\theta\|_2$ is imposed on $\phi_\theta$. This avoids the situation that all or most of the dynamics could be captured by neural nets and the physics-based models contribute little to learning.

Similarly, DeepGLEAM~\cite{wu2021deepgleam} is a method used for predicting COVID-19 mortality by directly combining the mechanistic epidemic simulation model GLEAM with neural nets. GLEAM~\cite{balcan2009multiscale} is a PDE-based model that characterizes complex epidemic dynamics based on meta-population age-structured compartmental models. DeepGLEAM employs a DCRNN~\cite{li2017diffusion} to learn the errors made by GLEAM, resulting in enhanced performance for one-week ahead COVID-19 death counts predictions. 

Beyond applications to observed dynamics data, hybrid methods can be used to substitute the computationally intensive components of classical solvers or learn corrections for classical solvers applied on inexpensive but error-inducing coarse discretizations. \citet{belbute2020combining} introduce a novel approach termed CFD-GCN that combines graph convolutional neural networks with a Computational Fluid Dynamics (CFD) simulator. This hybrid method aims to generate accurate predictions of high-resolution fluid flow. It runs a fast CFD simulator on a coarse triangular mesh to generate a lower-fidelity simulation, which is subsequently enhanced by upsampling it to finer meshes using the K-nearest neighbor interpolation technique. The fine-grained simulation is then processed by a graph convolutional neural network, which further refines the predictions for specific physical properties. Similarly, \citet{kochkov2021machine} utilize CNNs to perform learned interpolation and learned correction on coarse velocity components produced by classic numerical solvers, leading to significant speedup in simulating high-resolution fluid velocity fields. Moreover,  \citet{tompson2017accelerating} replace the numerical solver for solving Poisson’s equations, which is the most computationally expensive step in the procedure of traditional Eulerian fluid simulation, with \revisionOne{a convolutional network}. This approach results in significant speedup and demonstrates physically consistent predictions with strong generalization abilities. 

As opposed to approximating the numerical PDE solution as in the previously discussed works, Physics-Informed Neural Networks (PINNs) aim to approximate the analytical solution by parameterizing the neural network $\phi_\theta$ as the PDE solution~\cite{raissi2019physics}. Using backpropagation, the spatial and temporal derivatives 
in~\cref{eq:pde} can be exactly evaluated and used as a regularizing agent in an effort to ensure that the constraints prescribed by~\cref{eq:pde} are approximately satisfied. Thus, for the operator $\mathcal T$ defined as 
\begin{align}
\mathcal T\phi_\theta(x,t)&=\partial_t\phi_\theta + \mathcal D \left(x, t, \phi_\theta,\partial_x\phi_\theta, \partial_{xx}\phi_\theta, \ldots\right) & (x,t)&\in U,    
\end{align}
 the network $\phi_\theta$ can be optimized over the parameter space $\Theta$ as
\begin{equation}\label{eqn:pinn_loss}
   \phi_\theta = \underset{\phi_\theta:\theta\in\Theta}{\arg\min}\:\lambda_{\mathcal T}\mathbb E_{x,t\in U}\left[\lVert\mathcal T\phi_\theta(x,t)\rVert\right] + \lambda_{\mathcal B}\mathbb E_{x,t\in \partial \mathbb X\times \mathbb T}\left[\lVert\mathcal B\phi_\theta(x,t)\rVert\right] + \revisionOne{\lambda_0}\mathbb E_{x\in\mathbb X}\left[\lVert\mathcal \phi_\theta(x,0)-u_0(x)\rVert\right].    
\end{equation}
\revisionOne{$\lambda_{\mathcal T}$, $\lambda_{B}$, and $\lambda_0$} are coefficients for balancing different loss terms, which require careful tuning. 
PINNs have found real-world applications in biomedical analyses of blood flow~\cite{raissi2020hidden} with the Hidden Fluid Mechanics framework, and have been coupled with data-driven neural solvers such as the Physics-Informed DeepONet~\cite{wang2021learning} and Physics-Informed Neural Operator~\cite{li2021physics} to improve sample complexity and even allow for fully self-supervised training. \citet{yang2021b} further propose B-PINNs, which extends the concept of PINNs into the Bayesian framework. Under this approach, a PINN is used as the prior for solving partial differential equations (PDEs), while the Hamiltonian Monte Carlo method is employed to draw samples from the resulting posterior distribution. Compared with PINNs, B-PINNs not only provide uncertainty quantification but also obtain more accurate predictions on noisy data due to their ability to avoid overfitting.

\revisionOne{PINNs draw several key contrasts to the operator-learning approach detailed in~\cref{sec:msDyn,sec:mrDyn}. The operator learning paradigm trains $\phi_\theta$ to inductively map functions in the input space to the target function space. While this enables generalization over PDE configurations $\gamma$, such as initial conditions or PDE parameters, training requires discretization of the functions. In contrast, if $\phi_\theta$ is a PINN, training can be mesh-free, as $\phi_\theta$ takes the form of a function approximating the analytical solution to the PDE. However, this ties the trained model to one particular realization of $\gamma$, as the solution of a PDE with configuration $\gamma$ is not likely to also be the solution with configuration $\gamma'$, thereby requiring re-training of $\phi_\theta$ for each new instance of the PDE. To mitigate re-training cost, \citet{cho2024hypernetwork} consider a meta-learning framework which includes a hypernetwork for parameterizing their low-rank PINN (LR-PINN) dependent on the PDE parameters $\gamma_P$. Specifically, LR-PINN is trained on a variety of realizations of a given PDE with varying PDE parameters $\gamma_P^{(j)}$. For the PDE parameterized by $\gamma_P^{(j)}$, the weight matrices $\mathbf{W}^{(j)}$ that form the linear layers of LR-PINN are decomposed using the singular value decomposition as $\mathbf{W}^{(j)}=\mathbf{U}\operatorname{diag}(\mathbf{\Sigma}(\gamma_P^{(j)}))\mathbf{V}$, where the $\mathbf{U}$ and $\mathbf{V}$ matrices are shared across all examples, while the singular values $\mathbf{\Sigma}(\gamma_P^{(j)})$ are output from a hypernetwork whose input is $\gamma_P^{(j)}$. Following training, given a new instance of the PDE parameterized by $\gamma_P^{\star}$, only the singular values of the weight matrices are tuned starting from the hypernetwork-initialized $\mathbf{\Sigma}(\gamma_P^{\star})$. This greatly accelerates fine-tuning, as all remaining weights are frozen, and furthermore, the singular values from the hypernetwork are likely to be near-optimal.}

\subsubsection{Datasets and Benchmarks}\label{sec:pdebench}
\begin{table}[t]
	\centering
	\caption{Selected PDE datasets for forward modeling. We highlight challenging datasets that have arisen from neural PDE solver benchmarks~\revisionOne{\cite{takamoto2022pdebench, bonnet2022airfrans, gupta2022towards, toshev2024lagrangebench}} and works introducing methodologies~\revisionOne{\cite{sanchez2020learning,tran2021factorized, pfaff2021learning,li2022fourier}}. These datasets model a variety of fields across 1,2 and 3 spatial dimensions, and include challenging tasks such as fast moving and turbulent dynamics, large time step prediction, conditional prediction, and irregular geometries.}
        \resizebox{\columnwidth}{!}{
		\begin{tabular}{m{5cm}>{\centering\arraybackslash}m{3cm}>{\centering\arraybackslash}m{3.5cm}>{\centering\arraybackslash}m{2cm}>{\centering\arraybackslash}m{6cm}}
			\toprule
			Dataset & Source & Fields Modeled & Spatial Dimensions & Task Details \\
                \midrule 
                Compressible Navier-Stokes & \citet{takamoto2022pdebench} & Density, Pressure, Velocity & 1,2,3 & Initial Mach number=1 and viscosity=$1\times10^{-8}$ yield fast-moving and turbulent dynamics\revisionOne{.} \\
                Shallow Water & \citet{gupta2022towards} & Density, Velocity, Vorticity & 2 & Global weather on a rectangular domain with 48 hour time step\revisionOne{.} \\
                Incompressible Navier-Stokes & \citet{gupta2022towards} & Pressure, Velocity, Vorticity & 2 & Includes conditional task: advance the state of the system conditioned on variable timestep size and forcing term\revisionOne{.} \\
                TorusVis and TorusVisForce & \citet{tran2021factorized} & Vorticity & 2 & Variable viscosity coefficient and forcing term. Includes time-varying forcing term\revisionOne{.} \\
                CylinderFlow and AirFoil & \citet{pfaff2021learning} & Momentum, Pressure, Density & 2 & \revisionOne{Flow about obstacle on irregular mesh.} \\
                DeformingPlate and FlagDynamic & \citet{pfaff2021learning} & Position, von-Mises Stress & 3 & \revisionOne{Lagrangian simulation of structural mechanics.} \\
                \revisionOne{Water, Sand, Goop, MultiMaterial, WaterRamps, SandRamps, FluidShake, and Continuous} & \revisionOne{\citet{sanchez2020learning}} & \revisionOne{Position} & \revisionOne{2,3} & \revisionOne{Lagrangian simulation of various materials interacting with ramps, external forcing, and variable friction.} \\
                \revisionOne{Decaying Taylor-Green Vortex, Reverse Poiseuille flow, Lid-driven cavity, and Dam break} & \revisionOne{\citet{toshev2024lagrangebench}} & \revisionOne{Position} & \revisionOne{2,3} & \revisionOne{Lagrangian simulation of fluid dynamics interacting with turbulence, external forcing, moving boundaries, and dam breakage.} \\
                \revisionOne{Elasticity} & \revisionOne{\citet{li2022fourier}} & \revisionOne{Stress} & \revisionOne{2} & \revisionOne{Incompressible Rivlin-Saunders material deformed about irregularly-shaped void.} \\
                \revisionOne{Plasticity} & \revisionOne{\citet{li2022fourier}} & \revisionOne{Deformation} & \revisionOne{2} & \revisionOne{Deformation due to impact with irregularly-shaped die.} \\
                \revisionOne{Pipe} & \revisionOne{\citet{li2022fourier}} & \revisionOne{Horizontal Velocity} & \revisionOne{2} & \revisionOne{Incompressible flow through a curved pipe on irregular mesh.} \\
                \revisionOne{Airfoil} & \revisionOne{\citet{li2022fourier}} & \revisionOne{Mach number} & \revisionOne{2} & \revisionOne{Transonic steady-state flow about an airfoil on irregular mesh.} \\
                \revisionOne{AirFRANS} & \revisionOne{\citet{bonnet2022airfrans}} & \revisionOne{Velocity, Pressure, Kinematic Viscosity} & \revisionOne{2} & \revisionOne{Reynolds-averaged steady-state flow about an airfoil on irregular mesh with variable angle-of-attack and Reynolds number.} \\
			\bottomrule
		\end{tabular}
        }
	\label{tab:pde_data}
\end{table}


The rise of neural PDE solvers has elicited many datasets for forward PDE modeling, several of which we highlight here and summarize in~\cref{tab:pde_data}. \citet{takamoto2022pdebench} \revisionOne{release} PDEBench, which contains numerical solution data for 8 different PDEs with varying spatial dimensions. Beyond forward problems, \citet{takamoto2022pdebench} also consider inverse problems, a task we discuss in~\cref{sec:pde_inverse}. Perhaps the most challenging PDEBench dataset is the compressible \revisionOne{Navier-Stokes} equations for modeling the density, pressure, and velocity fields of a compressible fluid, which~\citet{takamoto2022pdebench} include in one, two and three spatial dimensions. Fluids with velocities approaching \revisionOne{(subsonic)} or exceeding \revisionOne{(supersonic)} the speed of sound must be considered compressible, that is, as having a density \jacob{which varies due to pressure}~\cite{vreugdenhil1994numerical,anderson2011ebook}. Thus, \citet{takamoto2022pdebench} \revisionOne{release} versions of this data with the initial Mach number, quantifying the ratio between the velocity of the fluid to the speed of sound in the fluid~\cite{anderson2011ebook}, as high as 1. Further, the low viscosities considered by~\citet{takamoto2022pdebench} produce highly turbulent dynamics which must be resolved at small scales for stable simulation~\cite{kochkov2021machine}. 

\citet{gupta2022towards} consider a particularly difficult realization of the shallow water equations generated using the global atmospheric model developed by~\citet{klower2milankl}. This PDE is derived by depth-integrating the Navier-Stokes equations, and, despite the name, can model fluids beyond water~\cite{vreugdenhil1994numerical}. This dataset of over $5{,}000$ trajectories models global pressure, wind velocity, and wind vorticity fields. \citet{gupta2022towards} consider the task of advancing the system by 48 hour intervals, a coarse mapping that is especially challenging to learn. \citet{gupta2022towards} also \revisionOne{release} data for modeling the incompressible Navier-Stokes equations generated by the $\Phi_\textrm{Flow}$ solver~\cite{Holl2020Learning}, and consider an interesting conditional task in which the learned solver makes predictions to future timesteps conditioned on varying timestep sizes and forcing terms. Similarly, \citet{tran2021factorized} \revisionOne{provide} data for modeling the vorticity field of a fluid with the incompressible Navier-Stokes equations while generalizing over the viscosity coefficients and forcing terms. 

\revisionOne{The previous datasets emphasize Eulerian systems that are spatially discretized onto square, uniformly-spaced meshes, and therefore are largely dominated by convolutional models. However, as discussed in~\cref{sec:mrDyn}, architectures that can model dynamics on irregular geometries can balance the tradeoff between solution accuracy and computational cost, and furthermore enable modeling of a more general class of problems. \citet{pfaff2021learning} demonstrate this flexibility in their dataset modeling flow around a cylinder governed by the incompressible Navier-Stokes equations, as well as a flow around an airfoil governed by the compressible Navier-Stokes equations, both of which are visualized in~\cref{fig:mrDyn}. \citet{pfaff2021learning} extend beyond Eulerian fluid dynamics problems with the release of structural mechanics problems modeling a flag blowing in the wind and a deformable metal plate. Importantly, these structural mechanics settings are modeled in the Lagrangian paradigm which, as discussed in~\cref{sec:pde_overview}, differs from Eulerian schemes by discretizing particles instead of space.}

\revisionOne{Multiple works have released data with an exclusive focus on Lagrangian simulations. \citet{sanchez2020learning} consider several datasets comprising Lagrangian simulations of a variety of materials in diverse settings. The materials modeled are water, a viscous material referred to as \textit{goop}, sand, and mixtures of these materials, while the considered settings include two and three spatial dimensions, as well as ramp obstacles, external forcing in the form of a shaking container, and variable friction. The LagrangeBench datasets from \citet{toshev2024lagrangebench} introduce a diverse array of Lagrangian fluid simulation datasets. While the decaying Taylor-Green vortex datasets simulate the onset of turbulence, \citet{toshev2024lagrangebench} introduce a spatially-dependent external force with the reverse Poiseuille flow datasets. Additionally, the lid-driven cavity datasets impose both static and dynamic boundaries, whereas the dam-break dataset models dynamics of a fluid suddenly released from a container. Besides the dam-break setting, all LagrangeBench datasets are available in two and three spatial dimensions.}

\revisionOne{Similar to~\citet{pfaff2021learning}, \citet{li2022fourier} also consider irregular geometries arising in the context of both structural mechanics and fluid dynamics simulations. In the structural mechanics setting, \citet{li2022fourier} release data for modeling stress of an incompressible Rivlin-Saunders material deformed about a void with a randomly-sampled, non-square geometry. \citet{li2022fourier} additionally release data for modeling the time-dependent deformation of a plastic material following impact with an irregularly-shaped die given the geometry of the die. \citet{li2022fourier} go on to study fluid dynamics problems modeling the velocity of incompressible Navier-Stokes flow though a curved pipe constructed from third-degree polynomials. Finally, \citet{li2022fourier} release data for modeling the Mach number of a transonic flow over an airfoil governed by the Euler equation, where \textit{transonic} refers to the property of the flow as having regions moving at both subsonic and supersonic speeds. \citet{bonnet2022airfrans} model a flow around an airfoil instead governed by the incompressible Reynolds-Averaged Navier-Stokes equations. To maintain the assumption of incompressibility, \citet{bonnet2022airfrans} reduce the Mach number to less than 0.3, increasing difficulty by varying the angle and Mach number of the flow around the airfoil. Unlike the time-dependent fluid dynamics datasets from~\citet{pfaff2021learning}, both~\citet{li2022fourier} and \citet{bonnet2022airfrans} consider steady-state fluids modeling in which the solution is static as a function of time.}

We note that while these works have curated an impressive selection of challenging PDEs foundational to the study of \revisionOne{neural} solvers, solvers performing well on these tasks may not immediately generalize well to real-world applications of PDE modeling. Dynamics encountered in \revisionOne{real-world} settings can involve an interplay with complex external forces and occur in large, irregularly-shaped domains. In such settings, the previously discussed challenges of multi-scale processing (\cref{sec:msDyn}), multi-resolution modeling (\cref{sec:mrDyn}), and rollout stability (\cref{sec:roSty}) increase both in difficulty and importance. Future benchmarks should design tasks to explicitly probe these areas in more realistic scenarios. 

    \subsubsection{Open Research Directions}\label{sec:pde_future}

    We close this section on forward modeling with a discussion of challenges faced by neural solvers that have largely been unaddressed by current works.


    A primary limitation of learned solvers is the requirement of an adequate number of training data generated by costly numerical solvers~\cite{raissi2019physics, brandstetter2022lie}, which is particularly problematic at the industry scale. Thus, improving the ability to generalize and sample complexity of learned solvers is necessary to justify and reduce this cost for their adaptation to real-world settings. Toward this goal, a richer subfield in the learned solver literature targeting techniques for out-of-distribution (OOD) dynamics should develop. In contrast to many current works that train solvers to generalize over initial conditions or PDE parameters from the same distribution as the training set, the goal of this work should be to enable learned solvers to accurately infer dynamics beyond those observed during training. Models excelling in this OOD setting will permit the span of the PDE solutions in the training set to be a subspace of the span of those in the test set, thereby improving sample complexity. OOD settings have been examined in the literature, such as~\citet{kochkov2021machine}, who study the performance of their hybrid classical\revisionOne{-}neural solver under OOD domain size, external forcing, and PDE parameters. \citet{stachenfeld2021learned} similarly study OOD domain size, as well as OOD initial conditions and rollout length. However, there is a gap in the literature around works developing principled approaches in the regime of OOD dynamics. Initial works in this area have treated the parameters of the differential equation as environments in the context of meta-learning and trained a model that can be transductively fine-tuned at test time to adapt to unseen, OOD environments~\cite{wang2022metalearning, mouli2023metaphysica, kirchmeyer2022generalizing}. \revisionOne{Similarly, as discussed in~\cref{sec:msDyn}, fine-tuning physics foundation models has been shown to enable data-efficient learning of downstream dynamics modeling tasks~\cite{subramanian2024towards,hao2024dpot,mccabe2023multiple,herde2024poseidon}. During the pre-training stage, the model can learn the elements of simulation that are shared across diverse physics. In contrast to models trained from scratch, this learned structure has been shown to offer an initialization enabling data-efficient learning of dynamics that are OOD from the pre-training dataset.}

    A second factor limiting applicability is that much of the literature focuses on problems with only one or two spatial dimensions despite the prevalence of three-dimensional problems in \revisionOne{real-world} applications of PDE modeling. While many architectures discussed in this section admit an immediate extension to three spatial dimensions, in practice, three-dimensional modeling presents obstacles in the form of limited memory that must be carefully handled~\cite{wu2022subsurface, lam2022graphcast, bi2022pangu}. Beyond memory requirements, the optimization of a three-dimensional model is naturally more challenging than its two-dimensional counterpart due to the increased size of the search space. Furthermore, in three dimensions, challenging dynamics not present in lower dimensions may be introduced, such as the prominent case of turbulent flows, where the transition to three dimensions induces chaos to a degree that is unseen in two-dimensional flows due to the energy cascade discussed in~\cref{sec:msDyn}~\cite{lienen2023generative}. Thus, future works should look to design neural solvers that are both scalable to three spatial dimensions and that have sufficient inductive biases for the optimization to effectively navigate the search space while maintaining sufficient expressiveness to faithfully model more challenging dynamics.      

    Additionally, as neural networks struggle to model non-smooth functions, modeling systems with sudden changes, such as the trajectory of a ball bouncing off a wall, remains a challenge. Such problems represent an extreme version of \emph{stiffness}, because the timescale of such drastic interactions is orders of magnitude smaller than the usual time step size for advancing the system. \citet{chen2020Symplectic} propose a method to handle one-time interactions of such a kind
    by augmenting the update equation in their integrator with a \emph{rebound} module. \citet{kim2021stiff} propose efficient methods for computing the gradients and appropriate normalization of Neural ODEs~\cite{chen2019neural} to model stiff systems.
    However, there is still a great need to identify more general solutions to model stiff systems.

\subsection{Inverse Problem and Inverse Design}\label{sec:pde_inverse}

\noindent{\emph{Authors: Tailin Wu, Xuan Zhang, Cong Fu, Rui Wang, Jacob Helwig, Rose Yu, Shuiwang Ji, Jure Leskovec}}\newline

    \begin{figure}[t]
        \centering
        \includegraphics[width=\textwidth]{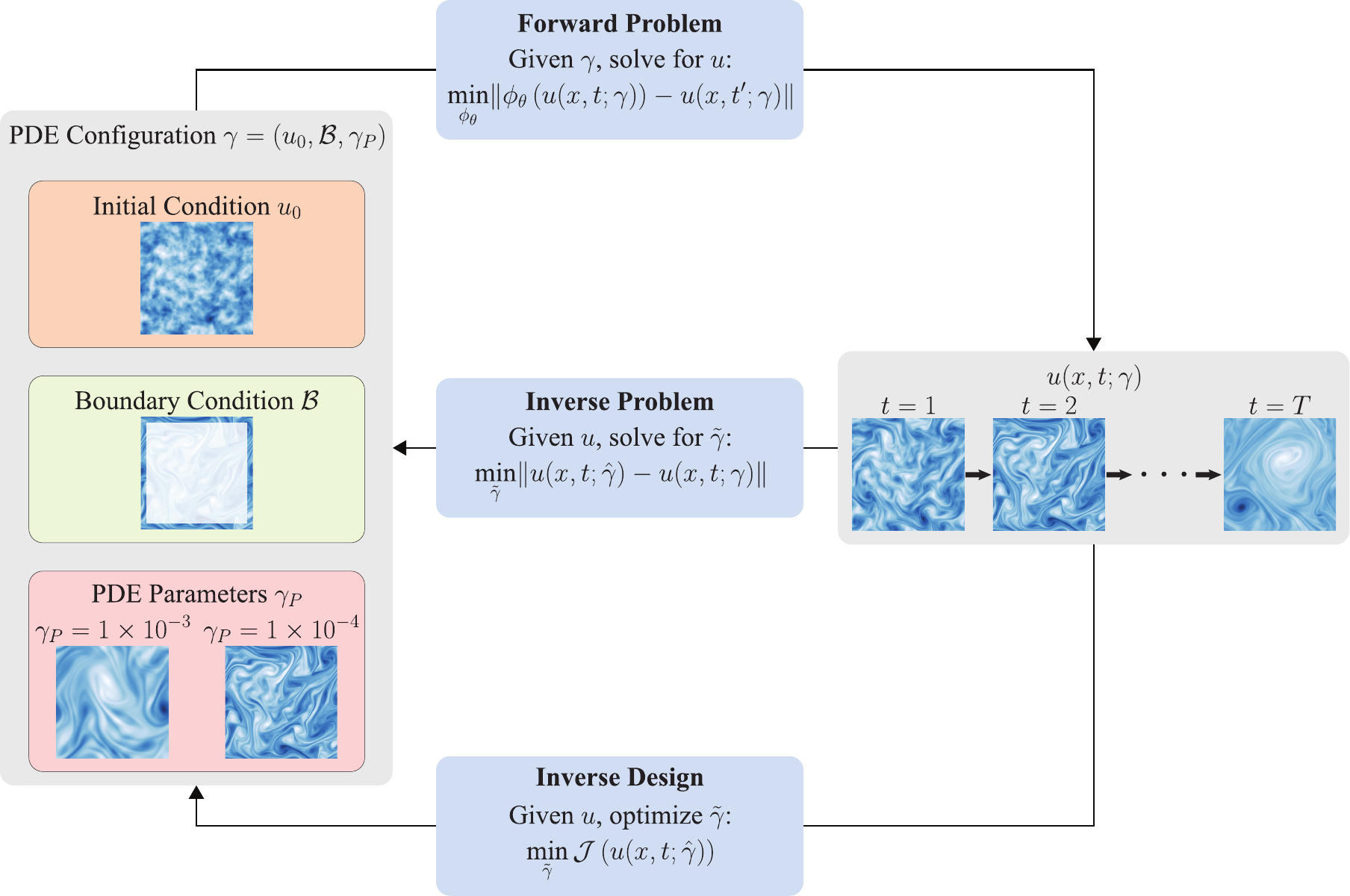}
        \caption{Illustration and comparison of forward problems, inverse problems, and inverse design. The solution to a PDE $u(x,t;\gamma)$ sampled on grid points $(x,t)$ discretizing space-time is induced by the PDE configuration $\gamma=(u_0, \mathcal B, \gamma_P)$ describing the initial conditions $u_0$, boundary conditions $\mathcal B$, and PDE parameters $\gamma_P$. In forward problems, the task is to learn the mapping from the solution induced by a particular choice of $\gamma$ at earlier time steps $t$ to the solution at later time steps $t'>t$ using the forecasting model $\phi_\theta$. Conversely, inverse problems consider the task of identifying a subset of the PDE configuration $\tilde\gamma \subset\gamma$, such as the initial conditions $u_0$, that generated the observed rollout data. The data is assumed to have originated from a forward model $u(x,t;\gamma)$, and the estimated configuration $\hat\gamma$ is optimized by minimizing the discrepancy between $u(x,t;\hat\gamma)$ and the observed data $u(x,t;\gamma)$, where $\hat\gamma$ denotes the union of the estimated components of the configuration $\tilde \gamma$ with the known components. Lastly, inverse design involves identifying $\tilde\gamma\subset\gamma$ such that the resulting rollout $u(x,t;\hat\gamma)$ optimizes some criterion $\mathcal J$, such as identifying the shape of an airplane wing that minimizes drag.}\label{fig:invFwd}
    \end{figure}

    In~\cref{sec:pde_forward}, we have delved into the advances and challenges of neural PDE solvers for simulating the \emph{forward} evolution of PDEs. The reverse direction is equally exciting, including (1) \emph{inverse problems}, where the task is to infer the unknown parameters or state of the system given (partial) observations of the dynamics, and (2) the emerging direction of AI-assisted \emph{inverse design}, where the task is to optimize the system (parameters or components such as initial or boundary conditions) based on a predefined objective. Both tasks are universal across science and engineering. In~\cref{fig:invFwd}, we conceptualize forward problems, inverse problems, and inverse design.

\subsubsection{Problem Setup}\label{sec:inverse_setup}

Let $u(x,t;\gamma)$ be a forward model that describes a physical process and is induced by the PDE configuration $\gamma=(u_0,\mathcal B,\gamma_P)$ describing the initial conditions $u_0$, boundary conditions $\mathcal B$, and PDE parameters $\gamma_P$. Furthermore, let $\tilde\gamma\subset\gamma$ be the properties to be recovered (in inverse problems) or the design parameters to be optimized (in inverse design). Finally, let $\mathcal{J}$ be an objective function which evaluates the quality of recovery or design. The inverse problem and inverse design can then be formulated as an optimization problem~\citep{lu2021physics} as
\begin{align}\label{eq:pde_inv_obj_no_constr}
    \tilde\gamma=\underset{\tilde\gamma}{\arg\min}\: \mathbb E_{x,t}\left[\mathcal{J}\left( u (x,t;\hat\gamma) \right)\right],
\end{align}
where $\hat\gamma$ represents the union of the components of the PDE configuration to be estimated with the remaining components that are assumed to be known. \revisionOne{$\hat\gamma$ can be finite-dimensional vectors or infinite-dimensional functions (\emph{e.g.}, an initial condition or boundary shape defined by a function).\footnote{\revisionOne{In this case, a neural network may be used to represent $\hat\gamma$ \citep{lu2021physics}. Alternatively, neural operator-based methods \citep{molinaro2023neural} can also be used to infer $\hat\gamma$.}}} For example, when modeling a dynamic system, $u$ typically defines what the rollout would be when a certain initial condition and boundary condition is given and $\mathcal{J}$ measures the difference between the simulated rollout induced by $\hat\gamma$ and the observed or targeted rollout. In the above formulation, $u$ is fixed and can be modeled with a classical PDE solver. To accelerate and improve the optimization, $u$ can also be a learned model and can be made to be differentiable. In this case, additional constraints on $u$ may be required to ensure physical consistency, and thus, the joint optimization of $\tilde\gamma$ and $u$ is constrained as:
\begin{align}\label{eq:pde_inv_obj_constr}
\tilde\gamma,u=\underset{\tilde\gamma,u\::\:\mathcal{C}(u, \tilde\gamma) \geq 0}{\arg\min}\: \mathbb E_{x,t}\left[\mathcal{J}\left( u (x,t;\hat\gamma) \right)\right]
\end{align}
where $\mathcal{C}$ can be the constraints stemming from the PDE or other constraints from multi-objective optimization~\citep{lu2021physics}.

\vspace{0.1cm}\noindent\textbf{Inverse Problems versus Inverse Design:} \ifshowname
\textcolor{red}{Cong}\else\fi  Despite the similarity suggested by their name, inverse problems and inverse design have different meanings in the context of PDEs. An inverse problem refers to the setting where some or all of the initial conditions, boundary conditions, or coefficients of the PDE are unknown, where the objective is then to determine or recover these unknowns from the observed data. An inverse problem typically assumes that the observed data is physically plausible and represents the solution to the PDE. For instance, in fluid mechanics, the observed data might be the vorticity field, and only the initial condition is unknown. Then, the inverse problem would be to determine the initial condition $u_0$ that would produce such a vorticity field. Alternatively, inverse design refers more specifically to a design or optimization methodology in which a predefined objective is given, and the goal is to optimize the system configuration based on the objective. For instance, given a surrogate model $u$ that can simulate forward fluid dynamics, the objective might be to design a surface that can guide the fluid flowing to the desired location. For inverse design, an exact solution may not necessarily exist, however, we may still want to optimize the proposed solution to satisfy the objective as much as possible. In some sense, inverse design can also be thought of as a specific type of inverse problem, where the goal is not just to determine unknown parameters or coefficients, but to design a system that behaves in a certain way.

\vspace{0.1cm}\noindent\textbf{Applications of Inverse Problems:}
    Here we describe several examples of inverse problems that hold potential for AI to create new opportunities, which we outline in~\cref{fig:pde_challenges}.

    \begin{itemize}
    \item \textbf{Fluid dynamics grounding:} \ifshowname
\textcolor{red}{Cong}\else\fi Learning a surrogate model of fluid dynamics typically requires the use of an expensive classical solver to obtain training data. An alternative approach is to consider an inverse problem, where the task is to infer the underlying dynamics solely based on a multi-view video of a 3D dynamical fluid scene~\citep{guan2022neurofluid}.
    \item \textbf{System identification:} \ifshowname
\textcolor{red}{Cong}\else\fi Traditionally, estimating the physical properties of an object requires conducting many physical experiments and the use of specifically designed algorithms. A promising inverse problem here is to infer the physical properties directly from visual observations~\citep{li2023pac}.
    \item \textbf{Full waveform inversion for geophysics:} \ifshowname\textcolor{red}{Xuan}\else\fi
    In geophysics, underground properties such as density or wave speed can be inferred from the measurement of seismic waves on the ground surface, a problem termed as \emph{full waveform inversion}~\citep{lin2023physics}. These underground properties are important for applications such as energy exploration or earthquake early warning, which are otherwise difficult to measure due to the large scale of the problem.
    \item \textbf{Fluid assimilation and history matching:} \ifshowname\textcolor{red}{Xuan}\else\fi Fluid assimilation aims to recover the entire fluid field from sparse observations in the spatio-temporal domain~\citep{zhao2022learning}. Fluid assimilation can be applied to model underground flow. The geological model is adjusted such that the predictions match the historical observations, a task termed as \emph{history matching}~\citep{tang2021deep}.
    \item \textbf{Tomography for medical imaging:} \ifshowname\textcolor{red}{Xuan}\else\fi Tomography aims to recover internal structures of an object using only surface measurements. For example, in medical imaging, electrical impedance tomography (EIT)~\citep{guo2023transformer} can infer the status of internal organs by measuring the voltage distribution on skin when an electrical current is injected, which avoids intrusive measurements or radiation exposure. 
\end{itemize}

\vspace{0.1cm}\noindent\textbf{Applications of Inverse Design:}
Here we identify a few applications where AI-assisted inverse design can play a significant role and where vast opportunities lie.
\begin{itemize}
  \item \textbf{Shape design for planes:} In aerodynamics, an important challenge is designing the shape of planes to minimize drag~\cite{athanasopoulos2009parametric}. This involves simulating the air fluid dynamics and its interaction with the boundary shape of the plane.
  \item \textbf{Ion thruster design:} In aerospace engineering, the design of efficient thrusters is highly important. For example, the Hall effect thruster (HET) is one of the most attractive electric propulsion (EP) technologies, since it has high specific impulse and high thrust density. One key question is how to design the shape and material arrangement of the thruster, given its complicated plasma dynamics~\cite{hara2019overview}. 
  \item \textbf{Controlled nuclear fusion:} Solving controlled nuclear fusion can pave the way for unlimited clean and cheap energy. In magnetic confinement with Tokamak, one of the two main approaches to controlled nuclear fusion, a key challenge is to optimize the external magnetic field and wall design in order to shape the plasma into configurations with good stability, confinement and energy exhaust~\cite{ambrosino2009design,degrave2022magnetic}.
 \item \textbf{Chip manufacturing:} Many processes in chip manufacturing involve inverse design. One important application is plasma deposition. Specifically, the problem is how to design the shape of the dielectric cell so that the deposition of the plasma onto a substrate is as smooth as possible~\cite{hara2023effects}. 
  
  \item \textbf{Shape design for underwater robots:} In underwater robots, an important problem is to design the shape of the robots to achieve multiple objectives, including minimizing drag, improving energy efficiency, improving dirigibility, and improving certain acoustics properties~\cite{saghafi2020optimal}.
  \item \textbf{Addressing climate change:}  Inverse design can play a significant role in many approaches to address climate change, including improving materials for buildings, optimizing carbon capture, solar geoengineering, and design of carbon credits and policy~\cite{rolnick2022tackling}.
  \item \textbf{Nanophotonics:} \ifshowname\textcolor{red}{Xuan}\else\fi Nanophotonics focuses on designing structures with a scale close to the wavelength of electromagnetic waves. Developing principled methods for designing micro-scale structures, nano-scale structures, or topological patterns to interact with light has important implications in applications such as laser generation, data storage, chip design, and solar cell design~\citep{molesky2018inverse}.
  \item \textbf{Battery design:} \ifshowname\textcolor{red}{Xuan}\else\fi  Deep learning-enabled inverse design has vast potential in battery design. For example, it can be used for the inverse design of battery interphases, which is important for developing high-performance rechargeable batteries~\citep{bhowmik2019perspective}. Besides the battery itself, hyperparameter-searching techniques in machine learning can be used to accelerate the experimental exploration of high-cycle-life charging protocols of lithium-ion batteries ~\citep{attia2020closed}, which is critical for electric cars.
\end{itemize}

\subsubsection{Technical Challenges}
\label{sec:challenges_inverse}

\vspace{0.1cm}\noindent \textbf{Common Challenges:} As inverse problems and inverse design involve the forward modeling task to evaluate $u(x,t;\gamma)$ in~\cref{eq:pde_inv_obj_constr,eq:pde_inv_obj_no_constr}, challenges encountered in forward problems discussed in~\cref{sec:pde_chall} are typically also present here. \revisionOne{An important challenge is \textbf{speed}, because inverse problems and inverse design require forward modeling as an essential component (which is already a computationally intensive process) and necessitate additional optimization with respect to the properties to be recovered or designed. Therefore, improving the speed for solving inverse problems and inverse design is a common challenge.}  Another common challenge in both inverse problems and inverse design is \textbf{adversarial modes}. This can occur when the high-dimensional parameters are inferred or designed with a deep learning-based surrogate model. Parameters with noisy, adversarial modes can occur that are not physically plausible, but achieve an excellent loss~\cite{zhao2022learning,wu2024compositional}. In the following, we illustrate several further unique challenges.

\vspace{0.1cm}\noindent \textbf{Challenges for Inverse Problems} 
\begin{itemize}

\item \textbf{Objective mismatch:} \ifshowname\textcolor{red}{(xuan)}\else\fi When the forward model and the inverse problem objective are jointly optimized, the forward model might sacrifice the physical constraints given by the underlying PDE in exchange for increased optimality in the inverse problem objective, resulting in physically inconsistent solutions. 

\item  \textbf{Ill-posedness:} In many applications, a complete measurement is often not available, which makes the inverse problem ill-posed and the solution non-unique. For example, when modeling a fluid, it is not feasible to track the movement of every fluid element. Thus, sparse measurement must be used. Another example is in tomography, where the problem is fundamentally ill-posed, as we try to infer inner structure solely from measurements on the boundary. \ifshowname\textcolor{red}{(xuan)}\else\fi

\item \textbf{Indirect observation:} \ifshowname
\textcolor{red}{Cong}\else\fi In some scenarios, it is hard or expensive to conduct direct measurement of physical states of an object or solution field. Instead, we may only be able to afford to video the object moving and interacting with the environment. Inferring the unknown parameters only from visual observations then poses a significant challenge.

\item \textbf{Incorporating Physics:} \ifshowname\textcolor{red}{(Rui)}\else\fi
Just as it is challenging to incorporate physics principles into the forward problem, it is equally important to ensure that inverse models adhere to the desired physical laws. It is crucial to extract relevant physical knowledge from well-established theories and incorporate it into the design of inverse models, while still maintaining sample and training efficiency and accuracy without any compromises.

\end{itemize}

\vspace{0.1cm}\noindent \textbf{Challenges for Inverse Design} \ifshowname\textcolor{red}{(Tailin)}\else\fi

\begin{itemize}
\item \textbf{Complex design space:} A fundamental challenge in inverse design, especially for real-world applications, is that the design space is hierarchical, heterogeneous, and consisting of many components that may be combined in many different ways. Take rocket design as an example. On a high level, a rocket consists of an airframe, a propulsion system, and a payload, each of which may consist of hundreds of parts. Thus, it presents a significant challenge to \emph{represent} the complex design space and to \emph{optimize} with respect to the chosen representation. 

\item  \textbf{Multiple (contradicting) objectives:} Real-world engineering design problems typically have multiple objectives that may contradict one another. For example, to design a phone battery, we simultaneously want the battery to have a long lifespan and be lightweight. These two objectives contradict each other, and thus, we must find a balanced trade-off. 

\item  \textbf{Temporally changing importance for multiple objectives:} In different scenarios, the importance of objectives may differ from one another. For example, as a rocket launches and transitions from ground to space, it will encounter drastically different environments, resulting in varying importance for its objectives of air resistance, fuel efficiency, and structure durability.

\end{itemize}

\subsubsection{Existing Methods}
\label{sec:existing_methods_inverse_design}

\vspace{0.1cm}\noindent\textbf{Inverse Problem:}
    \ifshowname
\textcolor{red}{Cong}\else\fi Recently, neural radiance field (NeRF)~\citep{guan2022neurofluid} has been applied to fluid dynamics grounding and system identification. NeuroFluid infers the underlying fluid dynamics from sequential visual observations by jointly training a particle transition model and a particle-driven neural renderer. PAC-NeRF~\citep{li2023pac} designs a hybrid Eulerian-Lagrangian representation of the neural radiance field combined with a differentiable simulator for estimating both physical properties and geometries of dynamic objects from sequential visual observations.
    
    \ifshowname\textcolor{red}{Xuan}\else\fi \citet{zhao2022learning} tackles the problem of fluid assimilation from sparse fluid rollout observations, as well as the full waveform inversion problem. The objective is to find an initial condition such that the simulated rollout is close to the observed rollout on sparse measurement locations. To enable an adaptive spatial resolution, a mesh-based data representation is used in conjunction with a learned GNN model~\citep{pfaff2021learning} as a forward model to forecast dynamics from the initial condition. To tackle the ill-posedness, \citet{zhao2022learning} propose to learn a latent vector for the entire fluid field, then infer the quantities to be recovered at each mesh point from the concatenation of the latent vector and the mesh coordinate.
    
    \ifshowname\textcolor{red}{(Rui/Tailin)}\else\fi

    Another important class of methods to address inverse problems is Physics-Informed Neural Networks (PINNs)~\cite{raissi2019physics}. PINNs are a class of methods addressing forward and inverse problems simultaneously. They parameterize the solution function as a neural network optimized (using backpropagation) with an objective that consists of both a data loss, which penalizes the discrepancy of the neural network solution with observed data, and a physics-informed loss, which penalizes the violation of the provided PDE. During training, unknown parameters of the PDE or the system can also be learned. Furthermore, \citet{lu2021physics} develop a new PINN method with hard constraints (hPINN) to solve PDE-constrained inverse design while avoiding commonly-seen optimization issues in PINNs~\cite{krishnapriyan2021characterizing}. The hPINN method utilizes two different techniques to enforce hard constraints on PINNs. One is the penalty method that gradually increases coefficients of the loss terms of boundary conditions and PDEs throughout training. The second technique involves the augmented Lagrangian method, which employs carefully selected multipliers in each iteration to enforce the constraints effectively.

    \revisionOne{A closely related class of method is Neural Radiance Field (NeRF) \cite{mildenhall2021nerf}. This approach inputs sample points along a ray into the neural network, which in turn outputs the color and density associated with that point. This inherently differentiable representation can address inverse problems in graphics and vision, such as reconstructing geometry from a set of images or disentangling scene lighting and material properties from these images. \citet{wang2021neus} proposed a geometric reconstruction method based on NeRF. This approach uses multiview images of a given scene as input, similar to the original NeRF. It outputs a geometry represented by a Signed Distance Function (SDF). While the original NeRF's volume rendering and density representation face limitations in reflecting accurate geometry, this method establishes a link between SDF and NeRF-based volume rendering. The original NeRF's density is replaced with SDF, and modifications to the rendering equation ensure the accumulated SDF density weights are both unbiased and occlusion-aware.
    \citet{zhang2021nerfactor} offers another technique which takes multiview images as input to separately output the geometry, albedo, material properties, and lighting characteristics of a scene. It predicts material parameters using a pre-trained BRDF decoder and employs MLPs to represent lighting via a low-resolution image. The geometry used involves the original NeRF density and the normal processed through an MLP. Such methods have been applied in various inverse problems in science, for example in cryo-electron microscopy \cite{Zhong2020Reconstructing}, computed tomography \cite{corona2022mednerf}, and mechanics \cite{mowlavi2023topology}.}

    \ifshowname\textcolor{red}{(Rui)}\else\fi 
    In situations where it becomes essential to identify the accurate governing equations for practical problem-solving, various attempts have been made to derive the precise mathematical formulation based on observed data. The conventional approach~\cite{schaeffer2017learning, brunton2016discovering, kaiser2018sparse} often involves selecting from a wide dictionary of potential candidate functions and finding the combination of a subset that minimizes the discrepancies between the model predictions and the observed data. Recently, many studies have also employed neural networks to augment the dictionary of candidate functions or to capture more intricate relationships between these functions. For instance, \citet{rudy2017data} utilize neural networks as supplementary candidate functions in addition to predefined basis functions to model more complex dynamics. \citet{martius2016extrapolation, sahoo2018learning} introduce EQL which utilizes neural nets to identify complex governing equations from observed data. Rather than relying on conventional activation functions, they employ predefined basis functions, including identity and trigonometric functions. Additionally, they integrate custom division units into the framework to capture division relationships within the potential governing equations. However, generalizability and over-reliance on high-quality measurement data remain critical concerns in this research area.

\vspace{0.1cm}\noindent\textbf{Inverse Design:} \ifshowname\textcolor{red}{Tailin}\else\fi
    Past methods to address inverse design are mostly based on domain-specific classical solvers, which are extremely computationally expensive. \revisionOne{In scenarios where the PDE is known, the adjoint method can be used to optimize the parameter $\hat\gamma$ by constructing a Lagrangian with the objective function and adjoint variables, deriving and solving the adjoint equations, and combining these solutions to compute the objective function's gradients for parameter updates \citep{edmunds1972optimal,protas2008adjoint,sirignano2022online}. Although the adjoint method can compute gradients efficiently, it requires the specific form of the PDE to be known and is highly sensitive to the initial conditions.} Recently, with the success of neural PDE solvers, AI-assisted inverse design has also emerged, but largely remains unexplored. One notable work is by \citet{allen2022inverse}, which uses backpropagation through time (BPTT) over the entire differentiable physical simulation to design the boundary for particle-based simulations. However, this method is still computationally expensive, as it must compute the gradient three times for hundreds of steps of simulation in the input space. \citet{wu2022learning} introduce BPTT in the latent space for inverse design, which improves both the runtime and accuracy compared to inverse design in the input space. For Stokes flow, \citet{tao2020functional} develop a method to simulate and optimize Stokes systems governed by design specifications with different types of boundary conditions.  \citet{li2022fluidic} further introduce an anisotropic constitutive model for topology optimization that can generate new topological features that differ drastically from the initial shapes and enable flexible modeling of both free-slip and no-slip boundary conditions. \revisionOne{A notable recent work is by \citet{wu2024compositional}, which introduces a compositional generative inverse design method CinDM. CinDM learns a diffusion model for generating the joint variable of state trajectory and system boundary. During inference, CinDM achieves compositional inverse design by averaging multiple diffusion models, each conditioned on subsets of the design variables and as a whole conditioned by the design objective. Experiments show that CinDM is able to design initial states and boundary shapes that are more complex than those in the training data. For instance, it discovers formation flying—a technique involving the strategic arrangement of multiple airfoils to reduce drag—despite being trained solely on the dynamics of a single airfoil interacting with airflow.}
    
    The above initial works of AI-assisted inverse design are limited to relatively simple and idealized scenarios. Therefore, there is a massive gap between the tasks considered by these works and those in 
real-world engineering in terms of the following aspects: (1) \emph{Complexity of the physics:} The physics in real-world systems may be multi-resolution or even multi-scale, making efficient and accurate simulation difficult. 
(2) \emph{Complexity of the design:} Real-world systems consist of many parts, requiring the system to be designed in a more hierarchical and structured way. (3) \emph{Generality and diversity:} The tasks tested above are restricted to a specific domain, and are not diverse enough to test the methods' generality across multiple disciplines.  
 These challenges provide great opportunities to develop novel neural representations and methods for proposing improved designs. A related work is by \citet{degrave2022magnetic}, which for the first time, employs deep reinforcement learning (RL) for shaping fusion plasma, and demonstrates that deep RL is able to control such complex systems. This work further demonstrates the feasibility of such a method on complex physical systems and serves as inspiration for the community to work on more challenging problems that have the potential to offer long-term beneficial impacts on humanity.

\subsubsection{Datasets and Benchmarks}

\ifshowname
\textcolor{red}{Cong, Xuan}\else\fi For the NeRF-related inverse problem, datasets are multi-view images of a dynamic scene or object generated by simulation engines, such as MLS-MPM~\citep{hu2018moving, hu2019difftaichi} and DFSPH~\citep{bender2015divergence}.
In the context of fluid assimilation, rollout data can be simulated using a classical solver such as finite element method solver~\citep{logg2012automated} used by~\citet{zhao2022learning}.  Lastly, \citet{deng2022openfwi} put forth an extensive benchmarking suite of 12 full waveform inversion datasets.

\ifshowname\textcolor{red}{Tailin}\else\fi In the domain of inverse design, as introduced in~\cref{sec:existing_methods_inverse_design}, different works have tested their methods using their respective domain-specific datasets, such as the datasets considered by~\citet{allen2022inverse} for designing  shape for airfoil and surfaces for particle-based fluid flows, \revisionOne{the dataset introduced by \citet{wu2024compositional} for compositional inverse design of multiple airfoils,} and the dataset employed by~\citet{wu2022learning} for designing boundaries to control smoke in fluid flow. However, there has not been a standard benchmark to evaluate different inverse design methods systematically. Furthermore, compared to real-world engineering tasks, the current datasets are significantly lacking in terms of the complexity of the physics and the difficulty of the design. This presents an excellent opportunity for the community to introduce more diverse and more complex benchmarks in terms of physics and the design task.

\subsubsection{Open Research Directions}
\ifshowname
\textcolor{red}{Cong}\else\fi For the inverse problem, there are several possible future directions to explore: (1) \emph{Uncertainty quantification}: Many inverse problems are ill-posed, and this instability can lead to high uncertainty in the solution. Uncertainty quantification is therefore crucial in these cases, as it can help describe uncertainties associated with the solution. (2) \emph{Improved training techniques}: Complex or ill-posed inverse problems present difficulty in training deep neural networks, motivating future research to develop novel training strategies and regularization techniques. 

\ifshowname\textcolor{red}{Tailin}\else\fi For inverse design, the challenges (\cref{sec:challenges_inverse}) and limitations of current works (\cref{sec:existing_methods_inverse_design}) also point toward exciting future directions. We identify several exciting opportunities. (1) \emph{Developing novel representations}: The hierarchical, heterogeneous, and complex design space presents ample opportunity to design suitable representations that balance faithfulness and efficiency. (2) \emph{Developing new optimization methods}: The design space is typically hybrid, consisting of discrete variables, such as the number for each part, and continuous variables, such as the shape for each part and how parts are composed. This complex space presents an exciting opportunity for the development of novel optimization methods. (3) \emph{Developing more general methods across domains:} The diversity of real-world tasks also calls for more general methods to tackle multiple domains.



















\clearpage
\section{Related Technical Areas of AI}\label{sec:other}
In addition to the challenges specific to individual science areas, there are several technical challenges that are shared across multiple domains in the field of AI for science. In particular, we identify the following four common technical challenges: out-of-distribution generalization, interpretability, \revisionOne{foundation models powered by self-supervised learning}, and uncertainty quantification. These challenges have long been recognized in the field of AI and machine learning, but they take on increased significance in the context of AI for science due to the unique characteristics of the data and tasks involved. In this section, we discuss the current limitations, existing approaches, and potential research opportunities related to these four challenges.

\subsection{Interpretability}

\noindent{\emph{Authors: Hongyi Ling, Yaochen Xie, Ada Fang, Marinka Zitnik, Shuiwang Ji}}\newline

Interpretability, despite its ubiquity in the machine learning field, lacks a unified mathematical definition. Its meaning can differ based on the context. It sometimes refers to a model's inherent ability to offer humanly
understandable interpretations of its predictions, a characteristic commonly observed in models such as decision trees. On the other hand, interpretability can also refer to an in-depth understanding of intricate models. For example, an interpretation highlights how distinct input graph patterns, \emph{e.g.}, a substructure, can lead to a certain GNN behavior, such as maximizing a target prediction. Within the scope of this work, we narrow our focus to instance-level interpretations which provide input-dependent explanations for each input graph. From this perspective, an interpretation sheds light on significant patterns or components of an input graph crucial for its prediction. Notably, different components of the input graphs may contribute to the model's predictions to varying extents. Thus, an effective interpretation method precisely identifies those components and patterns that significantly impact the predictions, enabling a comprehensive understanding of the underlying factors driving the predictions of models. 

Geometric deep learning (GDL) models have demonstrated significant potential in solving various problems in quantum, molecular, material, and protein science. However, to assess the scientific plausibility of GDL model outcomes, it is essential to achieve interpretability of results. Unfortunately, most GDL models lack interpretability and are often treated as black boxes, which hampers their reliability and limits their applicability in scientific domains. Here we explore the importance of interpretability with the incorporation of explainable artificial intelligence (XAI) with models. XAI aims to track the contributions of specific components of the input instance to the final predictions and identify the parts that carry information indicative of the prediction label. By understanding how model outputs are determined, the trustworthiness of their predictions increases. Additionally, XAI can test if model predictions are faithful to physical laws, which in turn will help improve the quality of existing GDL models. Precise interpretation techniques of model weights and features provide domain experts with deeper insights into the underlying mechanisms learned by these models, allowing the acquired knowledge from the model to guide future research directions. Interpretability of models can be particularly valuable for design of new compounds through identification of important substructures in molecules, materials, and proteins for particular properties. 

\subsubsection{Existing XAI Methods.}
While many XAI methods have been developed to study graph neural networks \cite{ying2019gnnexplainer, yuan2021explainability, gui2022flowx, baldassarre2019explainability, huang2022graphlime, schnake2021higher, xie2022task}, they mainly focus on 2D graphs. According to \citet{yuan2020explainability}, existing approaches can be mainly categorized into four classes, namely, gradients/feature-based methods, perturbation-based methods, decomposition methods, and surrogate methods. Gradients/feature-based methods, which rely on either feature values or gradients to evaluate feature importance, have been particularly popular because of their simplicity and the intuition they provide about feature importance. Perturbation-based methods analyze the change in prediction when input features are perturbed to generate importance scores. Decomposition methods decompose prediction scores and back-propagate these scores layer by layer until the input space to compute importance scores. These approaches provide more insights into each layer of graph neural networks. Surrogate-based methods sample some similar data to a given input example and fit a simple and interpretable model like a decision tree. The explanations from the surrogate model are used to explain the original predictions. These techniques are valuable for interpreting the behavior of complex models. For a deeper understanding of graph XAI, we recommend referring to the recent surveys \cite{yuan2020explainability}. 

Despite the progress made in XAI for 2D graph neural networks, XAI for GDL models or 3D graphs remains an underexplored field. Existing GDL methods \cite{wang2022visnet, tubiana2022scannet} aim to interpret their architecture through systematic analysis and visualization of the learned representations. These representations, categorized into distinct clusters, are aligned with specific physical or chemical properties. However, the prediction mechanisms of these models and the contribution of input graph components to predictions remain unknown.
There are unique challenges and opportunities in this domain due to the higher dimensionality of the geometric data and the complexity of the models. 
Although gradient/feature-based and perturbation-based methods are useful, they are insufficient to provide a complete explanation for the importance of geometric features. On the other hand, decomposition methods and surrogate-based methods cannot be easily applied to GDL models.  
Recently, \citet{miao2022interpretable} propose a new perturbation-based method specifically designed for 3D points. This approach uses a learnable interpreter model to introduce random noise to each 3D point. The interpreter model is trained together with the GDL model used to predict labels. The amount of the learned random noise is then used to generate importance scores for each input point. However, this work only focuses on interpreting the GDL models with invariant predictions and doesn't consider the invariance and equivariance of the explanations \cite{crabbe2023evaluating}. Thus, there is a need for more XAI techniques specifically for equivariant GDL models. 

\subsubsection{Potential Application Scenarios}

The contributions of interpretability with XAI to research science can be broadly categorized into the following four perspectives, with several potential applicable scenarios for each perspective.

\vspace{0.1cm}\noindent\textbf{Improving Trustworthiness of GDL Models:}
XAI techniques aim to provide insight into model behaviour and predictions, such as identifying important features and substructures of inputs. Interpretability of model predictions allows researchers to better understand underlying model mechanisms and in turn promotes trustworthiness of models. In molecular property prediction, XAI could validate faithfulness to physical rules for outputs of GDL models, such as the role of the chemical structure and functional groups of molecules in determining molecular properties. 
Similarly, in protein fold classification, XAI techniques can help identify the most important amino acid residues or secondary structure elements for predicting a specific fold, thereby verifying whether GDL models capture secondary structural features and assisting scientists in using model outputs to make research decisions. In material property prediction, XAI could be used to validate if the model is focusing on the correct elements and structure in the material for prediction. For learning ground state of quantum spin systems, using XAI to probe how perturbations of spin configuration and electron positions change the energy of the system will assist in validating if the learned energy is physically consistent. Application of XAI to GDL models through identifying substructures, feature importance, and effects of perturbations can be a valuable method for verifying if models are exhibiting scientifically consistent behaviour to promote trustworthiness of model predictions. 

\vspace{0.1cm}\noindent\textbf{Enabling Further Scientific Knowledge Discovery:}
XAI may reveal patterns and insights from model predictions that can help researchers discover new hypotheses and research questions, potentially leading to discoveries of new scientific knowledge. For example, when performing molecule energy prediction, XAI can provide valuable insights into the importance of substructures and perturbations of features of different conformers of the same molecule and their corresponding energy levels, assisting future research on the generation of molecular conformers. Furthermore, for protein science it could identify key secondary structure, or amino acid residues responsible for a given predicted property. This could guide further investigation of the identified substructures of the proteins. In the scenario of studying complex systems such as quantum mechanisms and PDEs, XAI can be used to understand the behavior of the systems and identify the most important variables and factors that contribute to a system's behavior of interest. By using XAI to gain insights of how features and weights affect the model's internal representations and decision-making processes, scientists can gain insight into the underlying physical principles and test new hypotheses about unknown and under-explored systems.

\vspace{0.1cm}\noindent\textbf{Diagnosing and Improving Existing Models:}
XAI enables researchers to improve the existing models by examining and ensuring that GDL models satisfy physical rules. The presence of scientifically erroneous model explanations also helps expose potential biases or errors in the model, which in turn improves model quality. For example, in modeling global weather patterns using shallow water equations, it is important to ensure that the GDL models used to solve these equations satisfy physical laws such as the conservation of mass, momentum, and energy. In molecular ML researchers can also use XAI to probe feature importance and validate if it aligns with chemically expected features such as atomic number, bond angle, \emph{etc}. XAI techniques can help researchers identify whether the predictions of GDL models adhere to these physical laws, and identify which physical constraints the model needs to better satisfy. This could be applicable in finding many-electron ground states, by checking if outputs satisfy fermion antisymmetry constraints. XAI of GDL models is important for validation of results by domain experts and is helpful for revealing limitations in predictions for improvement of existing models.


\vspace{0.1cm}\noindent\textbf{Facilitating Design of Drugs and Materials:}
XAI can identify the critical substructures or functional groups that contribute a desired property in drug discovery and material design.
For example, in molecular interactions between small molecules and proteins, XAI techniques can identify critical parts of the protein and ligand that contribute to predicting binding affinity. By obtaining information about specific groups of amino acids that determine affinity and the location of binding sites, researchers can design drugs that are more selective and only interact with the desired protein, reducing the risk of off-target effects. In material science, identification of elements and packing orientations that lead to a molecular property prediction can help guide discovery of new materials with particular properties. For protein science, XAI may identify particular substructure of a protein linked to predicted properties that researchers could use to design \textit{de novo} proteins with similar properties. Generative models for drugs and materials could also benefit from XAI to better understand proposed designs. For example, in generating a drug for a given protein, using XAI to highlight the importance of particular substructures in the protein and the generated drug for binding can be helpful for researchers to understand generated compounds and direct future design. XAI can guide the design of new compounds through identification of patterns and features that are important for a desirable or undesirable property.


\clearpage

\subsection{Out-of-Distribution Generalization}

\noindent{\emph{Authors: Xiner Li, Shurui Gui, Shuiwang Ji}}\newline


The out-of-distribution (OOD)~\cite{gulrajani2020search, arjovsky2019invariant} problem focuses on the common learning scenario where test distribution shifts from training distribution, which substantially degrades model performances in scientific discovery tasks, as shown in Figure~\ref{fig:ood_airs}. The mismatching of the distribution is commonly referred to as distribution shifts, including several concepts of covariate shift~\cite{shimodaira2000improving}, concept shift~\cite{widmer1996learning}, and prior shift~\cite{quinonero2008dataset}. This problem occurs in diverse application scenarios~\cite{miller2020effect, sanchez2020learning, myers2014information, gui2020featureflow} and is tied to various fields, such as transfer learning~\cite{weiss2016survey, torrey2010transfer, zhuang2020comprehensive}, domain adaptation~\cite{wang2018deep}, domain generalization~\cite{wang2022generalizing}, causality~\cite{pearl2009causality, peters2017elements}, and invariant learning~\cite{arjovsky2019invariant, ahuja2021invariance}. 

Currently, OOD generalization methods and studies in AI for science can significantly improve overall task performances as well as generalization abilities across various domains.
While numerous studies exist on general OOD generalization~\cite{arjovsky2019invariant, peters2016causal, tzeng2017adversarial, lu2021invariant, rosenfeld2020risks, ahuja2021invariance, sun2016deep, ganin2016domain} and non-Euclidean OOD generalization~\cite{wu2022dir, chen2022ciga, zhu2021shift, bevilacqua2021size, li2023graph, gui2023joint}, the realm of OOD methodologies for scientific applications remains largely uncharted. 
In this section, we aim to summarize the research on OOD approaches within the field of AI for science and emphasize the importance of further exploration.

\subsubsection{Background and Settings}
The distribution shift problem is studied under various settings including transfer learning~\cite{weiss2016survey, torrey2010transfer, zhuang2020comprehensive}, domain adaptation~\cite{wang2018deep}, domain generalization~\cite{wang2022generalizing}, causality~\cite{pearl2009causality, peters2017elements}, and invariant learning~\cite{arjovsky2019invariant, ahuja2021invariance}. 

In domain adaptation scenarios, we target transferring knowledge from one (source) domain to another (target) domain with distribution shifts between domains. Specifically, we can access both source domain samples with labels and target domain samples. According to the accessibility of labels in target domains, domain adaptation is typically categorized into semi-supervised and unsupervised settings. Unsupervised domain adaptation~\cite{pan2010domain, patel2015visual, wilson2020survey} is the most popular setting because it does not require any labeled samples in the target domains. The basic and most common idea is aligning the distributions between the source and target domains, mitigating the distribution shifts. This goal can be often done by discrepancy minimization~\cite{long2015learning, sun2016deep, kang2019contrastive} and adversarial training~\cite{ganin2015unsupervised, tsai2018learning, ajakan2014domain, ganin2016domain, tzeng2015simultaneous, tzeng2017adversarial}. However, domain adaptation necessitates the pre-collected target domain samples, shrinking its application scope, \emph{e.g.}, privacy-sensitive applications. 

Without the requirement of pre-collected target samples, domain generalization~\cite{wang2022generalizing, li2017deeper, muandet2013domain, deshmukh2019generalization} instead delves into the prediction for unseen domains, providing more practical solutions. Despite the prosperity of these areas, domain adaptation and generalization methods are still in need of robustly theoretical and intuitive analysis. Meanwhile, as the development of causality~\cite{pearl2009causality, peters2017elements}, one common sense is that generalization is logically implausible without interventions and inductive biases. Therefore, environment partitions~\cite{ganin2016domain, zhang2022nico++} are generally used as the indicator to imply the interventions that distributions come from. 

Causality~\cite{peters2016causal, pearl2009causality, peters2017elements} and invariant learning~\cite{arjovsky2019invariant, rosenfeld2020risks, ahuja2021invariance} can serve as the theoretical foundations for the out-of-distribution analysis, formulating various distribution shifts as graphical models or structural causal models (SCMs). Stemming from the independent causal mechanism assumption, the discovered causal correlations in SCMs are stable and ultimately endowed with physical laws. Therefore, learning causal mechanisms empowers deep models with generalization ability, leading to causality-based out-of-distribution analyses.

\citet{peters2016causal} firstly introduces the concept of invariant predictions and proposes the learning strategy of optimal predictors invariant across all interventions. Motivated by the invariant learning principle, \citet{arjovsky2019invariant} formulates the interventions as environment partitions and proposes the invariant predictor learning strategy as an optimization process, namely, invariant risk minimization (IRM). IRM considers one of the most popular data generation assumptions, later known as the partially informative invariant feature (PIIF) assumption. Subsequently, numerous invariant learning works~\cite{rosenfeld2020risks, ahuja2021invariance, chen2022does, lu2021invariant}, endowed with causality, propose to solve distribution shifts formulated by various assumptions including fully informative invariant feature (FIIF) and anti-causal assumptions~\cite{rosenfeld2020risks, ahuja2021invariance, chen2022does}, which makes these assumptions the popular basis of causally theoretical analyses for OOD problems.


\begin{figure}[t]
    \centering
    \includegraphics[width=\textwidth]{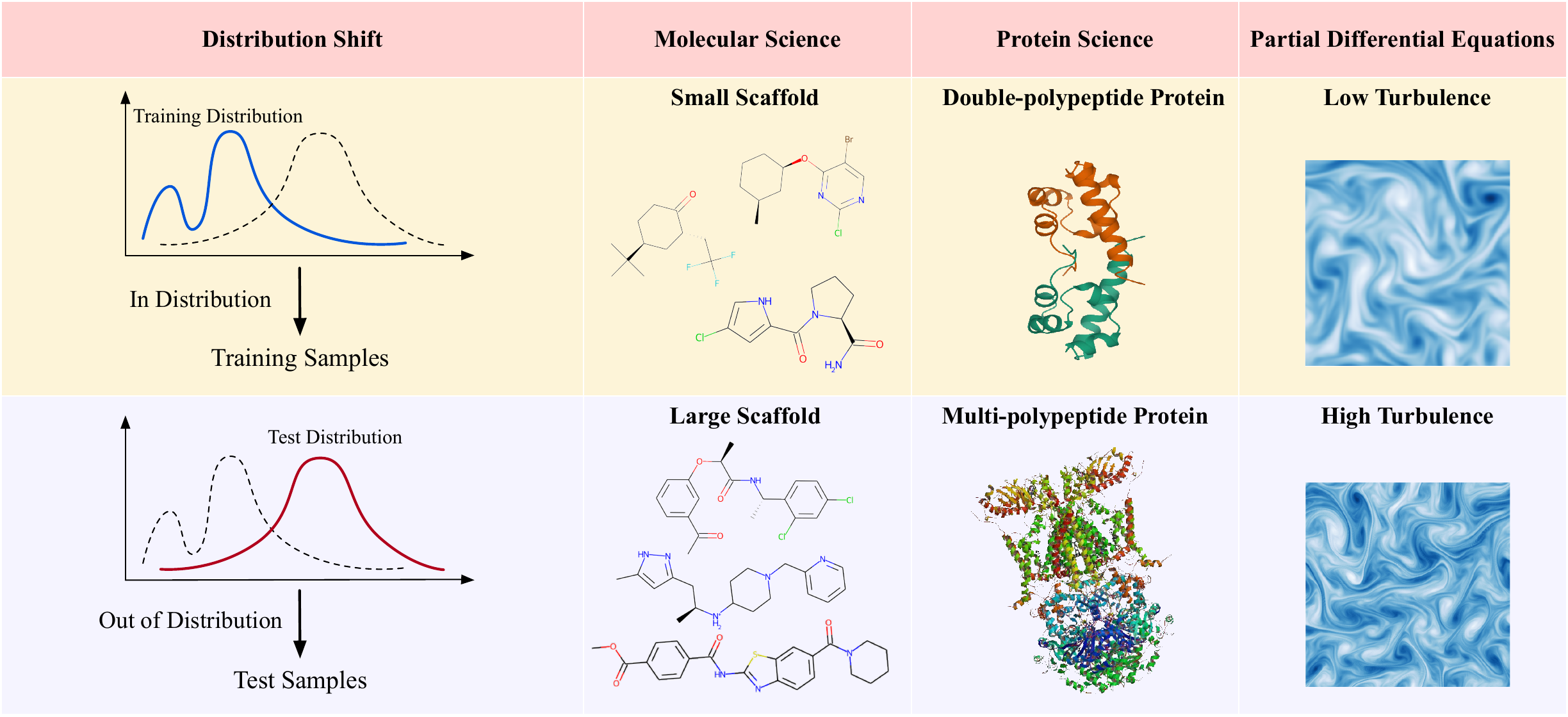}
    \caption{Illustrations for OOD in the field of AI for science. The out-of-distribution (OOD) problem is universal among scientific tasks, where training and test samples are from different distributions. In molecular science, different molecule sizes and scaffolds are major sources of distribution shifts. In protein science, the complexity of 3D protein structures, along with the vast array of potential variations in composition and folding, renders the generalization to unseen distributions a formidable challenge. In PDEs, generalizing from higher viscosity to lower viscosity in time-evolving modeling is a difficult task since lower viscosity leads to more turbulent flows, giving rise to more chaotic dynamics and challenges in simulation.}\label{fig:ood_airs}
\end{figure}
 
\subsubsection{OOD in AI for Quantum Mechanics} 
In the domain of quantum mechanics, the OOD problem frequently emerges when determining the wavefunction of quantum systems~\cite{yang2020scalable,kochkov2021learning,roth2021group,fu2022lattice}.
For example, as the sizes of quantum systems increase, the space needed to model the wavefunction grows exponentially, and the interactions between spins or particles become more intricate. Additionally, different geometries of systems will also change the underlying physical interactions dramatically. It is challenging to apply a wavefunction ansatz to a larger lattice or a different molecule.
Some works address OOD issues by better encoding intrinsic interaction modes that can be shared across system sizes and geometries.
~\citet{botu2015adaptive} compare the fingerprints of new structures with those in the training dataset and mandate a fresh QM calculation if it is out of the predictable domain when one or more components of the structure's fingerprints lies outside the training range.
QM-GNN~\cite{guan2021regio} implements supplemental QM descriptors to facilitate the prediction of out-of-domain unseen examples.
~\citet{caro2022out} initiate a study of out-of-distribution generalization in Quantum Machine Learning (QML). They prove out-of-distribution generalization for the task of learning an unknown unitary, a fundamental primitive for a range of QML algorithms, with a broad class of training and test distributions, showing that one can learn the action of a unitary on entangled states having trained only product states.

\subsubsection{OOD in AI for Density Functional Theory} 
Within the realm of DFT, OOD situations commonly arise in the context of quantum tensor learning.
For the task of quantum tensor prediction, current models are trained on systems with only tens of atoms due to computational complexity~\cite{schutt2019unifying, unke2021se, deeph}. However, in practice, systems can contain hundreds even thousands of atoms. The size shift of quantum systems engenders prediction difficulties without a trivial solution.
Several existing works put forward the problem of severe performance drop outside the defined applicability domain~\cite{pereira2017machine,li2016pure}, while few works offer feasible solutions to address this issue. A general and realistic direction for future studies is to perform training using data of different sizes under the invariant risk minimization framework~\cite{peters2016causal,arjovsky2019invariant} using size as the environment.

\subsubsection{OOD in AI for Molecular Science}
The OOD challenge in molecular science arises from the vast and intricate chemical space that AI models must navigate, with many potential challenges stemming from data limitations, model architectures, and evaluation metrics~\cite{gomez2018automatic, chen2018rise, feinberg2018potentialnet}. One primary challenge is the limited coverage of chemical space by training data, which can lead to biased predictions and model performance degradation on unseen molecules. This issue arises due to the immense size and complexity of the chemical space, with a virtually infinite number of potential compounds~\cite{polishchuk2013estimation}. For example, general GNNs are not capable of generalizing to large molecules when they are trained with small molecules. Motivated by pioneer causality-related invariant learning works~\cite{arjovsky2019invariant, peters2016causal}, \citet{bevilacqua2021size} propose to address the size shifts by introducing size-invariant graph representations. As many graph OOD learning methods~\cite{wu2022dir, chen2022ciga, gui2023joint} using subgraph-based graph modeling emerge recently, \citet{yang2022learning} introduces a molecule-specific invariant learning method for drug discovery. In science fields, \citet{sharifi2021out} formulate the drug response prediction in cancers as an OOD problem and propose Velodrome under a semi-supervised setting.
Besides OOD learning strategies, molecule OOD generation is also an emerging realistic direction. Current drug discovery experiments are expensive, and in-distribution generation will not provide innovative molecule structures. Therefore, OOD molecule generation is crucial for drug discovery. Recently, as the emergence of energy-based methods~\cite{elflein2023out} and molecule generations~\cite{liu2021graphebm}, \citet{lee2022mog} combine both energy-based generation and OOD detection to generate molecule out of the known molecular distribution. In addition, a score-based method molecular out-of-distribution diffusion (MOOD)~\cite{lee2022exploring} is proposed to generate novel and chemically meaningful molecule by utilizing gradients to guide the generation process to high property score regions.


\subsubsection{OOD in AI for Protein Science}
OOD in AI for protein science is a critical research topic due to the immense diversity of protein structures and functions, as well as the continually evolving knowledge of protein sequence-structure-function relationships~\cite{koehl2002sequence, petrey2005protein}. The ability to generalize AI models for predicting protein structures, protein-protein interactions, or even protein-drug interactions beyond the training data distribution would accelerate progress in areas such as drug discovery, precision medicine~\cite{ashley2016towards}, and protein engineering~\cite{goldenzweig2016automated}. One primary challenge in this context is the limited availability of high-quality experimental data, since the lack of domains is critical for OOD generalization. Therefore, a potential solution is incorporating domain knowledge and physical principles into AI models~\cite{jumper2021highly}. These approaches can help AI models learn more transferable and robust features that generalize better to novel protein sequences or complexes. ProGen~\cite{madani2020progen} is an unsupervised protein sequence generation method by using language models which include non-trivial OOD performance evaluations. \citet{gruver2021effective} find that ensemble models are more robust on OOD protein design than other methods. \citet{kucera2022conditional} propose an innovative protein sequence generation method with OOD generation evaluations. Finally, one possible direction is uncertainty estimation or OOD detection in protein science~\cite{hamid2018reliable, hamid2019self}, which is underexplored.


\subsubsection{OOD in AI for Material Science}
For material science, the OOD problem often arises due to the vast diversity of materials and their unique properties. Towards unseen OOD materials and compositions, the complexity of their structures, interactions, and properties present a significant challenge for AI-driven material discovery and optimization. 
Additionally, incorporating domain knowledge and physical principles into AI models can aid in learning more transferable and robust features that generalize better to novel materials and structures~\cite{murdock2020domain}.
\citet{kailkhura2019reliable} use an ensemble of simple models and propose a transfer learning technique exploiting correlations among different material properties to reliably predict material properties from underrepresented and distributionally skewed data.
\citet{sutton2020identifying} use subgroup discovery to determine domains of applicability of models within a materials class.
Another aspect of material science is material design and discovery~\cite{ghiringhelli2015big, xue2016accelerated, guo2019electrical}, which is influenced by the intricate nature of materials' structures and properties. Lastly, exploring OOD detection in material science~\cite{musil2018machine}, which remains largely unexplored, can be a promising future research direction.

\subsubsection{OOD in AI for Chemical Interactions}
 The OOD challenge is a critical issue in chemical interactions, particularly in the study of molecular interactions~\cite{cai2022binding, cai2022deepreal}, where models might struggle to generalize and provide accurate predictions when applied to new and unseen bindings. For instance, accurately predicting the docking efficacy of a drug candidate on a target protein that is significantly different from those in the training data is crucial for designing effective treatments. Recently, \citet{zhang2023learning} propose to consider protein-molecule interaction through subpocket-level similarities for drug generations, improving the model generalization ability. For drug-drug interactions (DDIs), \citet{tang2023dsil} devise a substructure interaction module, DSIL-DDI, to learn domain-invariant representations for DDI tasks, improving generalization ability and interpretability. To probe dark gene families, \citet{cai2023end} propose an innovative OOD meta-learning algorithm PortalCG to generalize from distinct gene families to dark gene family. Because of the challenge of the scarcity of receptor activity data, \citet{cai2022deepreal} propose a self-supervised method DeepREAL to mitigate distribution shifts. To assess the OOD generalization ability of previous drug-target interaction works, \citet{torrisi2022improving} provide a generalization ability evaluation by including systematic test sample separations.

\subsubsection{OOD in AI for Partial Differential Equations}
In the field of neural PDE solvers, deriving training data from classical solvers can be prohibitively expensive. Therefore, a practically useful neural PDE solver should be able to generalize to different systems, including those with different initial conditions, boundary conditions, and PDE parameters. 
MAgNet~\cite{boussif2022magnet} enables zero-shot generalization to unseen meshes, solving PDEs at a different resolution from that seen during training.
\citet{brandstetter2022message} add noise during training to encourage stability and address the distribution shift problem.
NCLaw~\cite{ma2023learning} embeds a network architecture that strictly guarantees standard constitutive priors (including rotation equivariance and undeformed state equilibrium) inside a differentiable simulation and optimize based on the difference between the simulation and the motion observation.
NCLaw can generalize to new geometries, initial/boundary conditions, temporal ranges, and even multi-physics systems after training on a single motion trajectory, achieving performance gains by orders-of-magnitude over previous neural network approaches on these typical OOD tasks.
Other works~\cite{kochkov2021machine, stachenfeld2021learned} study various OOD generalization abilities of learned models, including generalizing to conditions, rollout durations, and environment sizes outside the training distribution.
Future works can incorporate prior physical knowledge into deep learning surrogate models to obey the underlying physical laws and capture invariant information, thereby improving generalization ability across different systems.

\subsubsection{Datasets and Benchmarks}
To facilitate the development of OOD in scientific tasks, there have been prior benchmark works addressing the OOD problem in the scope of scientific tasks, providing schemes and evaluations for OOD learning on various real-world datasets.
OGB~\cite{hu2020open} focuses on graph datasets, identifies and splits different distributions respecting multiple domains. Wilds~\cite{koh2021wilds, sagawa2021extending} studies shifts on data collections from the wild covering multiple domains and data modalities. GOOD~\cite{gui2022good} considers the completeness of distribution shifts and benchmarks diverse graph tasks with numerous datasets and methods. 
DrugOOD~\cite{ji2022drugood} and CardioTox~\cite{han2021reliable} focus on molecular graph OOD problems, and are curated based on a large-scale bioassay databases ChEMBL~\cite{mendez2019chembl}, NCATS, and FDA~\cite{siramshetty2020critical}. ImDrug~\cite{li2022imdrug} evaluates several drug discovery tasks for imbalanced learning.
Further OOD studies for AI can benefit scientific tasks on the basis of these works. 

\subsubsection{Open Research Directions}
OOD scenarios are universal for AI in scientific fields, causing substantial deterioration in task performances; therefore, it is crucial to safeguard AI models in scientific domains from faltering in such situations to prevent adverse real-world consequences. 
We seek to underscore the significance of continued investigation and research into OOD strategies in the context of AI applications for scientific disciplines.
For further studies,  we point out that one promising direction is to identify and exploit causal factors~\cite{peters2016causal} in the training data that can constrain the behavior of optimized models on unseen test data. The model can generalize to OOD if the nature of the target distribution shift is known a priori, for example, enabling generalization to OOD orientations with models built in $SE(3)$ equivariance.

\clearpage
\subsection{Foundation and Large Language Models}

\noindent{\emph{Authors: Yaochen Xie, Carl Edwards, Qian Huang, Jacob Helwig, Jure Leskovec, Heng Ji, Shuiwang Ji}}\newline

Supervised learning of deep models usually requires a large amount of labeled data. However, in the case of scientific discovery, obtaining labeled data can be especially challenging due to factors such as the need for expert domain knowledge, high computational or experimental costs, or physical limitations. For example, computing the energy of molecules using DFT methods can take hours to days per molecule, depending on its size. Additionally, experimentally obtaining positively labeled data for drug discovery is costly and time-consuming, making deep models less applicable for rapid drug discovery at the early stage of global pandemics such as COVID-19. This difficulty has led to an emerging research area focusing on self-supervised learning (SSL). SSL techniques enable deep models to leverage unlabeled data and learn realistic data priors, such as physical rules and symmetries, without relying on extensive labeled datasets. Based on SSL, foundation models push this idea of leveraging data with no task labels to an extreme by aiming to pretrain a single model over these data that is easy to adapt for all tasks~\cite{Bommasani2021FoundationModels}. It essentially allows knowledge to be transferred as pre-trained representations from a general, usually self-supervised, task to a wide range of specific tasks of interest with limited labeled data. 
Specifically, large language models (LLMs) are the most versatile and powerful foundation models so far thanks to the label-free and rich supervision contained in the text data. LLMs enable even more flexible knowledge capturing and transfer due to their strong knowledge acquisition and reasoning abilities in scientific domains, including Physics, Computer Science, Chemistry, Biology, Medical Science \cite{Boiko2023EmergentAS, openai2023gpt4, nori2023capabilities, gupta2022matscibert}, \emph{etc}. One of the most exciting applications of LLMs in the sciences is generative modeling. While hallucination is a common problem for many LLM use-cases, it becomes a strength for discovering new drugs \cite{liu2021ai}, materials \cite{xie2023large}, and research ideas \cite{wang2023learning}.
So far, SSL-powered foundation and large language models are among the most promising directions to address the challenges of label acquisition and enable AI applications to a broader range of scientific problems. In the following subsections, we discuss the current challenges, focuses, and progress of SSL techniques, single-modal foundation models, and LLMs in the domain of scientific discovery.

\subsubsection{Self-Supervised Learning\ifshowname
\textcolor{red}{\ Yaochen}\else\fi}
SSL aims to construct informative learning tasks by deriving labels from the data itself, based on the associations within it. According to~\citet{xie2022self-survey}, SSL methods can be broadly categorized into contrastive and predictive approaches, depending on whether paired data are required in the learning process. Specifically, contrastive approaches involve multiple data modalities or augmentations to obtain positive data pairs to be discriminated from randomly sampled negative pairs, whereas predictive approaches auto-generate easy-to-compute and informative labels from certain subsets of dimensions of the data as the learning targets. SSL has shown its effectiveness and necessity in various fields~\cite{chen2020simple, devlin2019bert} in the paradigms of representation learning, pre-training, and auxiliary learning~\cite{xie2022self-survey}. 

\vspace{0.1cm}\noindent\textbf{SSL of Molecule and Protein Representations:} \revisionOne{In the context of AI for science, a majority of existing SSL work has focused on learning representation for molecules from their 2D graph formulations. In particular, general graph SSL work~\cite{you2020graph, xie2022self} has considered molecule representation learning as an important use case. In contrast, other studies have developed SSL methods specifically for molecular graphs, which allows for the integration of domain knowledge such as functional groups (motifs) co-occurrence~\cite{hu2019strategies, rong2020grover, li2021pairwise}, atom-bond associations~\cite{rong2020grover}, and reaction context~\cite{molecule2022}. These approaches have proven to be effective in leveraging the topology of molecular graphs as indicated by the chemical bonds but may miss certain geometry information of higher significance for certain tasks such as quantum properties predictions. To further use the 3D geometry information of molecules, \citet{liu2022pretraining} and \citet{stark20223d} propose to construct SSL tasks for molecules based on trans-modal associations. Technically, these approaches learn to maximize the mutual information between representations of 2D and 3D modalities of a molecule so that the representations are informative for multiple downstream tasks. Moreover, the Noisy Nodes technique has been proposed as a predictive SSL method in both pre-training~\cite{zaidi2023pretraining} and auxiliary learning~\cite{godwin2022simple} paradigms for 3D molecules. Specifically, Noisy Nodes approaches provide self-supervision by corrupting the atom coordinates and training GNNs to estimate the injected noise, which is in line with the idea of denoising autoencoders~\cite{vincent2008extracting, xie2020noise2same, batson2019noise2self}. The simple strategy is shown to be effective for 3D molecules and has been used in various following works~\cite {luo2023one, masters2022gps++}.} In addition to small molecules, there are also efforts on developing SSL approaches for proteins. Specifically, \citet{yu2023enzyme} use contrastive learning to train a model which can compare protein sequences against functional annotations, such as enzyme commission numbers, for functional understanding.

\vspace{0.1cm}\noindent\textbf{SSL of Neural PDE Solvers\ifshowname
\textcolor{red}{\ Jacob}\else\fi: } SSL has also been used in training neural PDE solvers, where the cost of training data generated by expensive numerical methods is a primary limitation of supervised solvers. To reduce this cost, \citet{raissi2019physics} propose the physics-informed neural network (PINN), which directly parameterizes the network as the PDE solution. The network is optimized with a self-supervised physics-informed loss derived using constraints on the solution specified by the PDE and was empirically validated by solving the Schr\"odinger equation in one spatial dimension. In a more challenging SSL setting, \citet{raissi2020hidden} infer the velocity and pressure field of a fluid flow using PINNs trained with constraints specified by the Navier-Stokes equations coupled with snapshots of the concentration of a scalar field such as dye advected by the flow. This application is particularly relevant in settings where pressure and velocity measurements are needed but only snapshots are accessible, such as biomedical analyses of blood flow to detect coronary stenoses~\cite{raissi2020hidden}. Additionally, neural solvers conceived in the supervised setting, such as DeepONet~\cite{lu2019deeponet}, have been extended to the SSL setting through the incorporation of a physics-informed loss~\cite{wang2021learning}. Unlike vanilla PINNs~\cite{raissi2019physics}, which are locked to one particular instance of the PDE, the physics-informed DeepONet proposed by~\citet{wang2021learning} can generalize over a family of PDEs, \emph{e.g.}, over initial conditions, and even demonstrated successful performance in experiments on the OOD regime. Furthermore, \citet{wang2021learning} report the physics-informed DeepONet outperforms its supervised counterpart.

\subsubsection{Single-Modal Foundation Models\ifshowname
\textcolor{red}{ Yaochen}\else\fi} \label{sec:single_modal_foundation}
The success of SSL techniques has given rise to the development of foundation models in vision~\cite{stablediffusion, kirillov2023segany,clipevent2022,TextVideo2022}, language~\cite{radford2018improving, devlin2019bert}, and medical~\cite{moor2023foundation} domains. Typically, foundation models are large-scale models pre-trained under self-supervision or generalizable supervision, allowing a wide range of downstream tasks to be performed in few-shot, zero-shot manners, with easy finetuning, or to be built upon learned embeddings. Similar to SSL techniques, they enable knowledge distillation and transfer from a large amount of unlabeled data to specific tasks with limited or even zero data. In this section, we focus on discussing foundation models that do not heavily rely on the natural language modality. Specifically, we explore the development of foundation models in the fields of protein and molecule analysis, where their versatility and potential impact are particularly evident, whereas in~\cref{sec:msDyn}, we have discussed a foundation model for forecasting weather and climate developed by~\citet{nguyen2023climax}.

\vspace{0.1cm}\noindent\textbf{Protein Discovery and Modeling:} Foundation models have shown great potential in AI for Science to address various challenges related to protein discovery and analysis. AlphaFold~\cite{jumper2021highly} and RoseTTAFold~\cite{baek2021accurate} are two foundation models that have made significant progress in predicting the geometry of protein folding. The trained models are then extended to perform more downstream tasks, including protein generation and protein-protein interaction (PPI). Specifically, RFdiffusion~\cite{watson2022broadly} fine-tunes RoseTTAFold to enable protein structure generation with a diffusion model. Instead of predicting structure from sequence, RFdiffusion performs unconditional generation from random noise, which can be further extended to conditional generation given certain functional motif or binding target. Similarly, Chroma~\cite{ingraham2022illuminating} is developed as a foundation protein diffusion model to enable protein generations conditioned on desired properties, including substructures and symmetry, which facilitates multiple downstream applications such as therapeutic development.
In addition, AlphaFold Multimer~\cite{evans2021protein} and ~\citet{humphreys2021computed} extend AlphaFold2~\cite{jumper2021highly} and RoseTTAFold~\cite{baek2021accurate}, respectively, to perform prediction tasks of PPI, or protein complexes without further fine-tuning. In addition to modeling the protein structure and geometry, the language model has also shown to be effective in multiple tasks related to protein design~\cite{madani2023large, melnyk2022reprogramming, zheng2023structure, hie2023efficient} even when only the sequential form of proteins is involved.

\vspace{0.1cm}\noindent\textbf{Molecule Analysis and Generation: }For molecule-related tasks, while various self-supervised learning (SSL) techniques have been proposed, there is currently no dominant non-language-based foundation model in this field. However, two promising threads of research work have emerged, focusing on different modalities of molecules: the molecular graph, in terms of either the 2D structure or the 3D geometry, and the sequential representation in terms of SMILES~\cite{weininger1988smiles, weininger1989smiles}. In the case of \emph{2D molecular graphs}, researchers have extended the success of graph-based SSL studies~\cite{wang2022molecular, molecule2022}. For example, \citet{fifty2023harnessing} formulate molecules as graphs and pre-trains GNN models with a great amount of simulated data to predict the binding energies for interactions between molecules and protein targets in simulation. Compared to typical pre-training approaches, \citet{fifty2023harnessing} demonstrate the potential of molecule foundation models in a wider range of downstream tasks, including few-shot docking and property predictions. 
Recent work also demonstrates the multi-tasking capability of foundation models built upon \emph{3D molecular graphs} when encoded appropriately. Specifically, \citet{flam2023language} formulate 3D molecule-related tasks as the auto-regressive generation on the sequentialized 3D coordinates of atoms. This framework enables the use of language model architectures on multiple tasks, including molecule generation, material generation, and protein binding site prediction.
On the other hand, existing work such as ChemGPT~\cite{chemgpt}, ChemBERTa~\cite{chithrananda2020chemberta, ahmad2022chemberta}, MolBert~\cite{fabian2020molecular}, \citet{schwaller2021mapping}, MegaMolBart \cite{megamolbartv2}, and \citet{tysinger2023can} 
focus on \emph{string representations} of molecules and adapt pre-training techniques from language models to molecule representation learning from large collections of such strings. 
Language models are also shown to be capable of molecule generation by producing SMILES strings. 
However, since SMILES strings were not designed specifically for generative modeling, many generated SMILES strings are chemically invalid. New string representations~\cite{grisoni2023chemical} for the generation purpose have been proposed, such as DeepSMILES~\cite{krenn2020self}, which avoids ring and parenthesis closing issues, and SELFIES~\cite{krenn2020self}, which proposes a formal grammar approach to ensure validity. SELFIES has been extended to incorporate groups~\cite{cheng2023group} to better capture meaningful molecular motifs. These string-based studies are among the first attempts to explore the power of large language models and have shown great potential in various tasks. In spite of the use of language models, these studies focus on the single modality of molecules and do not involve guidance and knowledge from natural language. 

\subsubsection{Natural Language-Guided Scientific Discovery\ifshowname
\textcolor{red}{ Carl, Yaochen, Qian}\else\fi}






Applying language models to the scientific domain is becoming increasingly popular due to its potential impact for accelerating scientific discovery \cite{hope2022computational}. A natural question is to ask why we want to integrate language into the scientific discovery process. Beyond the conspicuous and important task of extracting information from literature, there are a number of other compelling reasons. First, language enables scientists without computational expertise to leverage advances in AI. Second, language can enable high-level control over complex properties when designing novel artifacts (\emph{e.g.}, drug design going from low-level ``logP'' to high-level ``antimalarial''). Third, language can serve as a ``bridge'' between modalities (\emph{e.g.}, cellular pathways and drugs) when data is scarce. Beyond these three reasons, language has been developed as \emph{the} method by and for humans to abstractly reason about the world. In much the same way that science often relies on natural phenomenon (\emph{e.g.}, penicillin) for innovation, we can rely on linguistic phenomenon for abstraction and connection. 

Traditionally, natural language processing (NLP) has been developed with a focus on core tasks including translation and sentiment analysis. In the scientific domain, NLP tasks have focused on extracting information from the literature, such as named entity recognition \cite{li2016biocreative}, entity linking \cite{lai2021bert}, relation extraction \cite{wei2016assessing, lai2021joint}, and event extraction \cite{zhang2021fine}. NLP models have advanced rapidly in recent years and hence resulted in strong foundational models which can be easily applied to most NLP tasks \cite{devlin2019bert, raffel2020exploring, brown2020language}. Further, these systems have, to some extent, commonsense \cite{bian2023chatgpt} and reasoning \cite{NEURIPS2022_9d560961, yao2023tree, huang2022towards} abilities which may further advance AI research for science. However, the variety and complexity of scientific text still pose challenges to these systems. Thus, considerable effort has gone into constructing domain-specific language model variants \cite{beltagy2019scibert, liu2020self, michalopoulos2020umlsbert, gu2021domain,  meng-etal-2021-mixture, yasunaga2022linkbert, gupta2022matscibert, luo2022biogpt, Taylor2022GalacticaAL} to harness the valuable information contained therein. 
Building on these base models has powered a wide range of applications, ranging from large-scale information retrieval systems \cite{googlescholar, fricke2018semantic} to knowledge graph construction \cite{wang2020covid,zhang2021fine}, and from analogical search engines for scientific creativity \cite{kang2022augmenting} to scientific paper generation \cite{wang2018paper, wang2019paperrobot}. Although these applications are diverse, their common theme is the attempt to make sense of an overabundance of scientific information. Recently, work has further improved these scientific language models by introducing external knowledge from human-constructed databases into existing models \cite{lu2021parameter, lai2023keblm}, applying distillation for data augmentation \cite{wang2023understanding}, and augmenting models via retrieval \cite{naik2021literature, zamani2022retrieval}. 

Current advances in LLMs for science generally focus on addressing two key challenges. First, science discovery problems usually involve complicated data modalities such as the geometric status of a particle system. It is hence crucial to develop effective approaches to encode and integrate scientific modalities with the language modality. Second, due to various task formulations and limited data and model availability, the adaptation of general-purposed LLMs to scientific domains is non-trivial in terms of the learning task formulations and paradigms.
In this section, we discuss the frameworks and techniques of LLMs for science instantiated by existing works from the above two perspectives.





\vspace{0.1cm}\noindent\textbf{Multimodal Science with Language: }
To address the challenge of leveraging scientific data modalities, work has begun to investigate aligning natural language with modalities in the scientific domain. While some cases explore different variations of the original modality, such as molecule string representations \cite{guo2021multilingual}, significant interest has begun to grow in the integration of these naturally existing modalities and natural language for enabling control of the scientific discovery process. This is in large inspired by the success of models such as CLIP \cite{radford2021learning} (contrastive learning) and DALL-E \cite{ramesh2021zero} (joint sequence modeling) in the last two years. The high-level goal of integrating language with other modalities is to enable high-level function control (\emph{e.g.}, taste) rather than low-level property-specific control (\emph{e.g.}, solubility). The overall proposition is that similarly capable models in the scientific domain would vastly accelerate many aspects of the discovery process by enabling scientists to work with function rather than form in mind. Additionally, language is compositional by nature \cite{szabo2004compositionality, partee1984compositionality,LMSwitch2023}, and therefore holds promise for composing these high-level properties \cite{liu2022multi}. Such compositionality is shown in scientific tasks evaluated by \citet{edwards2021text2mol, edwards2022translation,su2022molecular,liu2022multi}.

\vspace{0.1cm}\noindent\textbf{Multimodal Science with Language --- Determining Modalities: }
In order to determine the appropriate problem formulation and model design for a given application, it's first necessary to determine the relevant input and output modalities. For example, if one wants to extract reactions from the literature, a text-to-text model \cite{vaucher2020automated} should be sufficient. However, to develop a contextual understanding of the reactions we are extracting, we might additionally incorporate figures (vision) and molecular structures. In the case of drug molecule generation and editing with high-level instructions, incorporating language as an input would be appropriate \cite{edwards2022translation, liu2023chatgpt, fang2023mol}. In the case of retrieving relevant literature about a drug, we may choose a molecular graph as input which is used to retrieve from a corpus of papers. Generally, given sufficient data for training, adding modalities to language will likely be helpful simply by grounding the model's understanding into the real world. However, obtaining multimodal data can be challenging in practice. As a rule of thumb, one can ask themselves the following three questions for moving from single-modal solutions to multimodality: 1) is multimodality a core part of my task, such as molecule captioning? 2) Should I add language to Section \ref{sec:single_modal_foundation} tasks? In other words, do I need the level of control and abstraction offered by natural language; or is there complementary information available as text? and 3) Will I meaningfully benefit from anything beyond language, or is all the information I need expressed as text?

\begin{figure}
    \centering
    \includegraphics[width=0.99\textwidth]{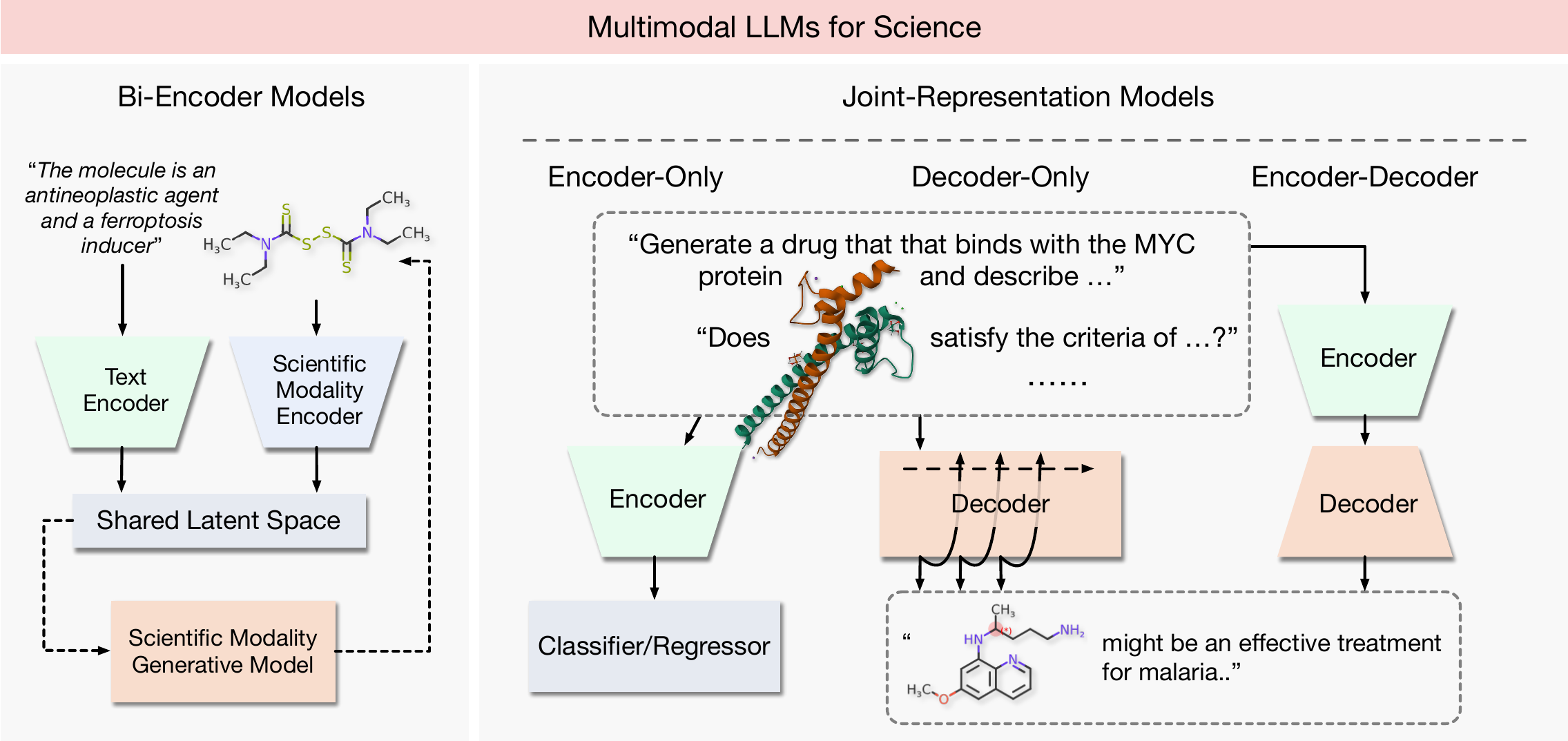}
    \caption{High-level architectures for multimodal scientific NLP. Molecule-language multimodality is used as a source for examples. The dotted lines in the bi-encoder diagram indicate the possible extension to a generative framework. Example inputs and outputs are shown with ``code-switched'' modalities (\emph{i.e.}, they are integrated into a single sequence). Encoders are used to generate an output representation which can be used for retrieval, classification, and regression, among other tasks. Decoder models are generally used for generative modeling applications. In some cases, non-LLM components can be used (such as for the decoder in an encoder-decoder model). We note that the extension of general-purpose LLMs via tools is another approach not shown in this figure. }
    \label{fig:llm-multimodal}
\end{figure}

\vspace{0.1cm}\noindent\textbf{Multimodal Science with Language --- Integrating Modalities: }
Now that one has decided to pursue a multimodal approach to their application, it is crucial to develop a framework to integrate the modalities. There are two common approaches as shown in~\cref{fig:llm-multimodal}, namely, Bi-Encoder Models and Joint Representation Models. An analogy can be drawn here to the distinction between cross-encoders and bi-encoders in information retrieval \cite{reddy2023inference}; bi-encoders allow fast comparisons and can be less data-intensive to train, but cross-encoders allow more fine-grained interaction between modalities. We now discuss the two approaches in detail. \emph{Bi-Encoder Models} consist of an encoder branch for text and a branch for the other modality such as molecules and proteins. They have the advantage of not requiring direct, early integration of the two modalities, allowing existing single-modal models to be integrated. Representative examples include Text2Mol \cite{edwards2021text2mol}, which proposes a new task of retrieving molecules from natural language queries, and CLAMP \cite{seidl2023enhancing}, which learns to compare molecules and textual descriptions of assays for drug activity prediction. BioTranslator \cite{xu2023multilingual} takes this to the extreme by learning a latent representation between text, drugs, proteins, phenotypes, cellular pathways, and gene expressions. Generally, these bi-encoder models are effective for cross-modal retrieval \cite{edwards2021text2mol, su2022molecular, liu2022multi, zhao2023adversarial}, but they may also be integrated into molecule \cite{su2022molecular, liu2022multi} and protein \cite{liu2023text} generation frameworks. 
\emph{Joint-Encoder Models}, on the other hand, seek to model interactions between multiple modalities inside the same network branch. These can be categorized by whether they incorporate a decoder or not. Encoder-only models can be used for prediction, regression, and (potentially slow) retrieval, but are unable to perform generative modeling. An example is KV-PLM \cite{zeng2022deep}, which trains an encoder-only language model on literature data with molecule names replaced by SMILES strings. A second category is encoder-decoder \cite{edwards2022translation, christofidellis2023unifying} or decoder-only models \cite{liu2023molxpt}. These can be used for cross-modal generative tasks, such as the ``translation'' between molecules and language proposed by \citet{edwards2022translation}, where molecules are generated to match a given textual description and vice versa. Interest has also arisen in using language to edit existing molecules for drug lead optimization \cite{liu2022multi, liu2023chatgpt}. Other work considers reaction sequences \cite{vaucher2020automated, vaucher2021inferring} or proteins \cite{gane2022protnlm}.

\vspace{0.1cm}\noindent\textbf{Adapting LLMs to Science Domains: }
Existing LLMs have predominantly been studied and developed for general purposes. These LLMs can be leveraged and adapted to specific science domains or even particular tasks, capitalizing on their inherent knowledge and priors. When performing such an adaptation, it becomes crucial to meticulously design the formulation of the scientific task as a sequential generative task and incorporate essential domain-specific context and knowledge into the LLM, either during or prior to task inference. Moreover, given limited domain-specific data, one has to trade-off between the overwhelming irrelevant knowledge from general domain and the reasoning capability during the adaptation.

\vspace{0.1cm}\noindent\textbf{Adapting LLMs to Science Domains --- Learning Task Formulations: }
LLMs are sequence-based models, so it is non-trivial to construct input and output sequences for different task formulations such as prediction, retrieval, and generation. As a general trend, however, most current language models in scientific domains adapt pretraining procedures from core NLP such as BERT \cite{devlin2019bert} (\emph{e.g.}, KV-PLM \cite{zeng2022deep}) or GPT \cite{radford2018improving} (\emph{e.g.}, \cite{liu2023molxpt}). As such, future work may find benefit in newer language model training objectives \cite{tay2023ul2}. Some work, however, has attempted to use additional signals from known properties \cite{ahmad2022chemberta}. Additional work is often needed for designing multimodal learning formulations. Contrastive learning paradigms are widely applied in the case of bi-encoder models due to their multi-branch design.
Learning tasks in joint-representation multimodal models are often designed to alleviate challenges with data scarcity. Strategies include multi-lingual \cite{edwards2022translation} and multi-task \cite{christofidellis2023unifying} learning. In addition to training models for multimodal tasks, existing single-modal models can be integrated with multimodal extensions to avoid formulating a new learning objective. For example, bi-encoder models can be combined with flows \cite{su2022molecular} or sequence generation models \cite{liu2022multi, liu2023chatgpt}. Further, classifier guidance can be used with an existing generative model \cite{ingraham2022illuminating}.

\begin{figure}
    \centering
    \includegraphics[width=0.99\textwidth]{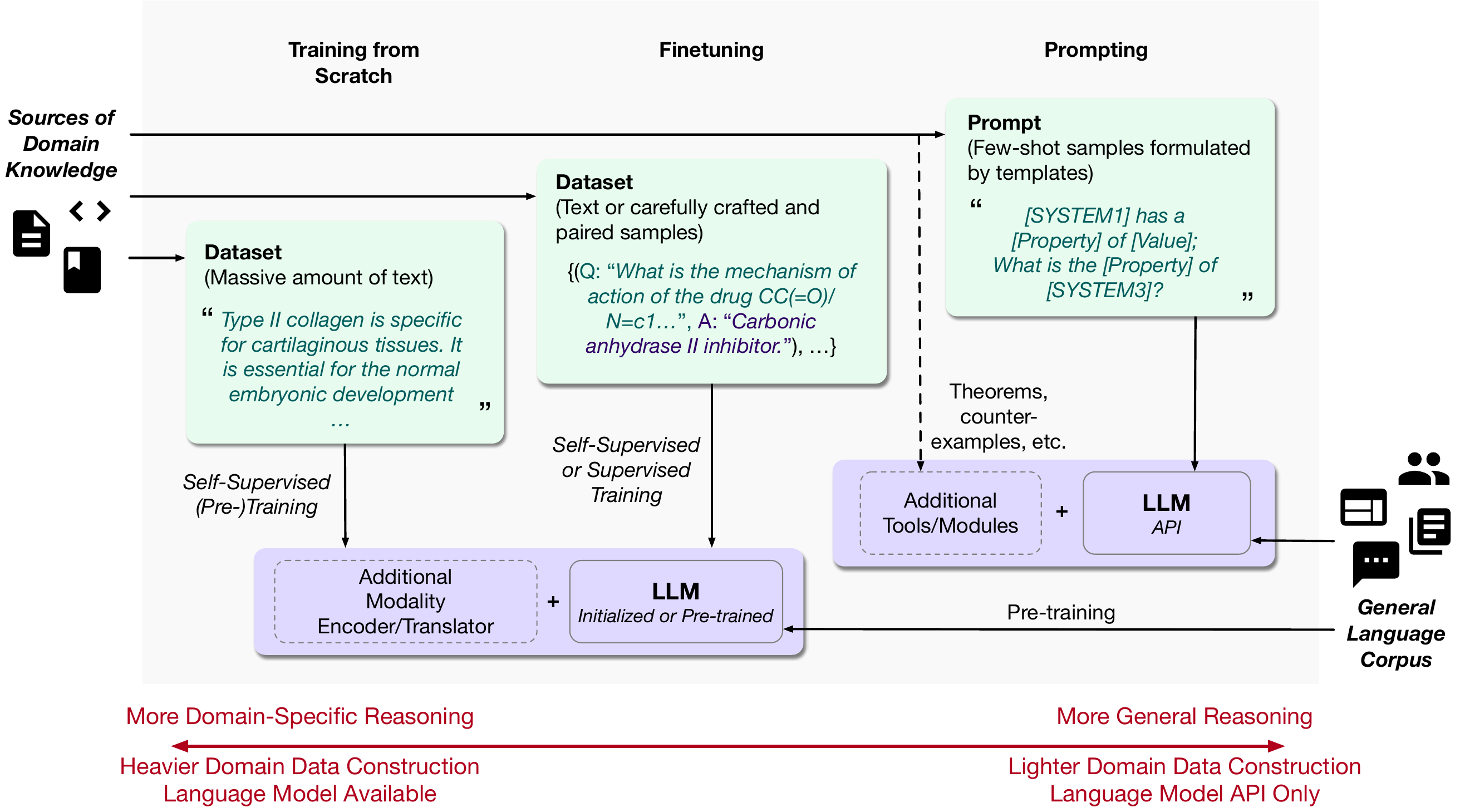}
    \caption{Three paradigms of adapting LLMs to science domains. One can construct datasets consisting of massive amounts of text from science domains and train LLMs from scratch in a self-supervised manner. The trained model can be used directly or further fine-tuned for specific tasks. Alternatively, one can fine-tune a pre-trained general-purposed LLM with less amount of text data, in a self-supervised manner, or paired samples, in a supervised manner, from science domains. In cases of proprietary LLMs with API access, one can adapt the model by prompting with carefully designed templates, where domain knowledge are provided as few-shot samples in the prompt or as explicit knowledge with additional tools or modules.
    Dataset examples are from Galactica~\cite{Taylor2022GalacticaAL} and ChEMBL~\cite{liang2023drugchat}, respectively.}
    \label{fig:llm-paradigms}
\end{figure}

\vspace{0.1cm}\noindent\textbf{Adapting LLMs to Science Domains - Learning Paradigms: }
Existing efforts have explored approaches to adapting LLMs for various specific applications. Taking into consideration varying levels of data availability and model accessibility, these adaptations can be achieved through several paradigms~\cite{liu2023pre, wang2023interactive}. In the science domain, there are three typical adaptation paradigms, including training domain-specific LLMs from scratch, fine-tuning general-purposed LLMs, and few/zero-shot learning with prompting. Their learning frameworks, dataset construction, and model access are compared in Figure~\ref{fig:llm-paradigms} and discussed in the following. 
LLM architectures themselves usually contain certain prior about reasoning. Given an effective LLM architecture, it is natural to \emph{train domain-specific LLMs from scratch} with highly customized data. Such an approach enables the highest flexibility with additional modules and blocks and helps learn better domain-specific knowledge given a fixed capacity of model. Once trained, those models can be further used in specific downstream tasks. However, to achieve a desired performance, a significant effort is to be made to construct the training dataset consisting of a large amount of text. For example, Galactica~\cite{Taylor2022GalacticaAL} constructs a large scientific corpus with data collected from papers, code, knowledge base, \emph{etc}. The LLM is then trained in a self-supervised manner~\cite{devlin2019bert, radford2018improving} on the collected domain-specific data, being able to tokenize math equations, SMILES, protein, and DNA strings. 
These approaches are capable of performing multiple downstream tasks such as drug discovery, repurposing, and interaction prediction. 
Taking advantage of the general reasoning capability of pre-trained language models, recent works also explore the \emph{fine-tuning of pre-trained LLMs} with domain-specific datasets. 
BioMedLM \cite{biomedlm} and med-PALM \cite{Singhal2023TowardsEM} are finetuned on biomedical domains from general LLMs GPT-2~\cite{radford2019language} and PaLM~\cite{chowdhery2022palm}, respectively. They have shown promising performance on the medical question–answering tasks. 
Fine-tuning can also be performed with less amount of but paired data in the supervised fashion.
For example, DrugChat~\cite{liang2023drugchat} constructs an instruction-tuning dataset consisting of more than 143k manually crafted question-answer pairs and covering more than 10k drug compounds. The LLM is then trained together with a GNN module on the constructed dataset. 

Due to the recent success and popularity achieved by the most advanced LLMs such as GPT-4~\cite{openai2023gpt4}, work has begun to adapt these general instruction-tuned models for the most challenging scientific discovery. 
As the advanced LLMs are mostly proprietary with API availability, their science domain adaptation is usually achieved by the paradigms of \emph{few-shot or zero-shot learning}, also known as in-context learning, through prompting. In particular, domain knowledge can be provided as a context in the prompt in the form of theories, facts, or examples. This paradigm has demonstrated its effectiveness in subjects like Social Science~\cite{Zhong2023GoalDD} and astronomy~\cite{galaxies11030063}. In the molecule domain, 
work has explored the chemical knowledge contained in these models in terms of language \cite{castro2023large} and code generation \cite{hocky2022natural, white2023assessment}. 
\citet{guo2023indeed} benchmark advanced LLMs on multiple tasks in the chemistry domain and demonstrate their competitive performance compared to task-specific machine learning approaches.
Recent work, such as CancerGPT \cite{li2023cancergpt} and SynerGPT \cite{synergpt}, also explores the applications of language models for drug synergy prediction. 
SynerGPT proposes novel LLM training strategies for in-context learning to explore the higher-level ``interactome'' between drug molecules in a cell. They extend their model to inverse drug design and context optimization for standardized assays. The proposed training strategies may enable a new type of foundation model based on the drug interactome. 
Further, one particular promising route is augmenting LLMs with external tools such that even complex tasks become textual \cite{Schick2023ToolformerLM, yao2023react}, with scientific examples like using the Web APIs of the National Center for Biotechnology Information (NCBI) for answering genomics questions \cite{Jin2023GeneGPTTL}. 
Existing LLMs are pretrained only from unstructured texts and fail to capture some domain knowledge. Recent solutions for domain knowledge-empowered LLMs include developing lightweight adapter framework to select and integrate structured domain knowledge to augment LLMs~\cite{lai2023keblm} and data augmentation for knowledge distillation from LLMs in the general domain to chemical domain~\cite{wang2023understanding}. External domain tools can also be integrated into language model prompts to allow these ``agents'' to access external domain knowledge \cite{bran2023chemcrow, Boiko2023EmergentAS, liu2023chatgpt}. Specifically, ChatDrug~\cite{liu2023chatgpt} enables LLM-powered drug editing in a few-shot manner by equipping LLMs with a retrieval and domain feedback module. Work is also done in the few-shot setting for both regression and classification \cite{jablonka2023gpt}, as well as Bayesian optimization \cite{ramos2023bayesian}. 

Pushing this paradigm to the extreme, there are also emergent efforts on developing LLM-based agents for scientific discovery by connecting LLMs with tools for conducting experiments, such as in Chemistry \cite{Boiko2023EmergentAS} and Machine Learning \cite{Zhang2023MLCopilotUT}. 
However, different science domains often rely on data in very different forms of modalities in practice, making it challenging for LLMs to be directly useful in many applications.

\subsubsection{Open Research Directions}
There are still remaining challenges towards scientific discoveries with foundation and large language models. We identify and discuss the following three challenges and opportunities.

\vspace{0.1cm}\noindent\textbf{Data Acquisition for Foundation Models: }
Large-scale data acquisition presents a significant challenge when developing SSL and foundation models for scientific applications, mainly due to the specialized nature of scientific data compared to general internet data with no easy way around it. There are existing efforts such as collecting domain-specific text from the internet, image-caption pairs from PubMedCentral's OpenAccess subset \cite{Lin2023PMCCLIPCL}, and scientific figures with captions from arXiv \cite{Hsu2021SciCapGC}. However, most of these efforts focus on the image and text modalities and largely overlap with web data. More work is needed on curating more realistic scientific data with diverse modalities, such as sensory and tabular data, to support building more customized foundation models for science. Data scarcity is a key challenge for language-based scientific multimodality, such as models trained on molecule-text pairs. 
Existing work has attempted to alleviate this challenge by applying multilingual pretraining strategies \cite{edwards2022translation} or by using entity linking to extract large quantities of noisy molecule-sentence pairs from the literature \cite{zeng2022deep, su2022molecular}. However, improved extraction of less noisy and more complete data from the literature will greatly benefit these tasks. \cite{yang2023bliam} investigates the use of language models for extracting additional drug synergy training tuples from literature.

\vspace{0.1cm}\noindent\textbf{Addressing Algorithmic Challenges for SSL and Foundation Models: }
Aside from data, the main technical challenges for SSL and foundation models for science typically include incorporating diverse modalities in the architecture, designing customized pretraining techniques for these modalities, and addressing domain distribution shifts. Recent methods have mainly focused on combining text and image modalities \cite{Liu2023PrismerAV, Koh2023GroundingLM, Alayrac2022FlamingoAV,niu2023ct} and more well-studied scientific units like molecules and proteins, with limited recent SSL/foundation models works on other modalities like graph \cite{Huang2023PRODIGYEI}, RNA expression \cite{Rosen2023TowardsUC} and benchmark on even more rare modalities like bacterial genomics and particle physics \cite{Tamkin2021DABSAD}. Finally, the dynamic data change in the realistic scientific discovery process also forces the pretrained models to face domain distribution shifts, as exemplified in the Wilds benchmark \cite{koh2021wilds}. More foundational work on designing robust SSL algorithms for diverse modalities is needed for applying AI for science in practice. For LLMs, since the knowledge from general domain is often overwhelming, developing better fusion models beyond perceivers can be a promising future direction.

\vspace{0.1cm}\noindent\textbf{Extending the Success to Broaden AI for Science Topics: }
Self-supervised learning (SSL) and foundation models have demonstrated promising performance in domains such as small molecules, proteins, and continuum mechanics. However, their methodologies and applications in other areas have received less attention. For example, SSL has been relatively less explored in the context of quantum systems. Learning tasks in quantum systems often revolve around modeling wavefunctions, and the neural network architectures used tend to be specific to lattice structures, making knowledge transfer between systems or tasks challenging.
Nonetheless, SSL holds significant potential in these fields as unlabeled data distributions can contain valuable information about the underlying symmetry and physical rules. SSL can play a crucial role in learning these rules as a prior, thereby facilitating the discovery of fundamental principles across various systems. 
Furthermore, adapting foundation models presents another promising avenue for the discovery of emerging and less-explored domains with limited data. Particularly, as demonstrated by~\citet{Taylor2022GalacticaAL, xu2023multilingual}, the text-based nature of LLMs enables them to capture and transfer knowledge among different systems more effectively and flexibly, bridging the gap between different domains and data modalities.


\clearpage

\subsection{Uncertainty Quantification}

\noindent{\emph{Authors: Yucheng Wang, Xiaoning Qian}}\newline






The capability of profiling and predicting properties of complex systems involved in previously discussed AI for Science tasks enables optimal and robust decision making for scientific discovery as well as automated generative capabilities. While developing deep forward prediction and generative models for inverse design under different conditions may have made significant advancements, reliable uncertainty quantification~(UQ) in these physics constrained prediction and generative models, such as neural ODEs~\cite{chen2019neural} as well as DeepONet~\cite{raissi2020hidden}, 
is critical to guarantee robust decision making under data and model uncertainty, however still requires investigation via collaboration of applied mathematics, computational science, and AI/ML researchers. Different UQ strategies have been developed in these research communities, from classical Bayesian model sensitivity analysis focusing on subsets of model parameters to the recent ensemble-based UQ in deep Bayesian learning. When integrating these UQ strategies into forward predictive and inverse generative models, scalability and efficiency are the utmost important factors to enable time-sensitive prediction and decision making in practice. Efficient and reliable approximate Bayesian computation and variational inference methods are to be developed to achieve desired performances of both predictive and computational criteria. 

\subsubsection{Uncertainty Quantification: Introduction and Background}
\emph{Uncertainty quantification}~(UQ) has been studied in various disciplines of applied mathematics, computational and information sciences, including scientific computation, statistic modeling, and more recently, machine learning. Traditional UQ aims at either quantitatively assessing prediction uncertainty or calibrating parameters of traditional physics-principled mechanistic models and data-driving machine learning models to address challenges of modeling complex systems due to enormous system complexity and data uncertainty~\cite{kennedy2001bayesian, psaros2023uncertainty}. 
When modeling complex systems, the uncertainties of a computational model can be from multiple sources. First, dynamics of real-world complex systems are typically modulated by many potential internal and external factors. Abstract computational modeling often can not cover all these factors, due to either missing information or computation limitations. Some factors affecting the system outcomes may be unknown or ignored for model construction. Second, even if all of the influencing factors are included, due to lack of knowledge, especially for data-driven black-box machine learning models, the selected model itself can be mis-specified with potential inductive bias. Third, the systems dynamics itself to be modeled can be intrinsically stochastic and non-stationary. Fourth, significant data uncertainty has to be taken care of as the observed data themselves are inevitably noisy and even corrupted due to the inherent sensor noise or the random perturbations from uncontrollable environmental factors. Finally, due to the limited precision of the modern digital computer hardware, the numerical results from different models may still contain errors. All these above uncertain sources contribute to the uncertainty of the final system output or model prediction.


\vspace{0.1cm}\noindent\textbf{Aleatoric and Epistemic Uncertainty:} 
Two types of uncertainties that have been identified and extensively investigated in computational modeling are the \emph{aleatoric uncertainty} and \emph{epistemic uncertainty}~\cite{kendall2017uncertainties, hullermeier2021aleatoric}. Aleatoric uncertainty, also known as~(a.k.a.) \emph{stochastic uncertainty} or \emph{data uncertainty}, refers to the uncertainty due to the intrinsic randomness of the physical process under investigation. For example, in a quantum spinning system, even if the quantum state of the system is known, the measurements with respect to the computational basis are typically random. In materials science experiments, since the noise of the sensor measurements can hardly be removed completely, the experimental results under the same condition may differ with some degree. In molecular property prediction, the predicted molecular properties can have significant uncertainty if only the 2D structure information is provided due to the incomplete representation  considering the actual 3D molecular geometry~\cite{hirschfeld2020uncertainty}. These uncertainties are irreducible even if more knowledge of the complex system or supplementary data become available. Epistemic uncertainty, a.k.a. \emph{systematic uncertainty} or \emph{model uncertainty}, represents the uncertainty due to the lack of knowledge of its physical process dynamics when modeling a complex system.  
Epistemic uncertainty can be reduced or removed as more and more knowledge or data becomes available, for which many Bayesian learning~\cite{cohn1996active,lampinen2001bayesian,titterington2004bayesian,Qian-PNAS,Qian2016,gal2017deep,goan2020bayesian,boluki2020learnable}, UQ~\cite{Yoon2013MOCU,lakshminarayanan2017simple,huang2017snapshot, sensoy2018evidential,Randy2019,amini2020deep,ywang2022agsys}, experimental design methods~\cite{kushner1964new, mockus2012bayesian, mariet2020deep,Zhao2020,lei2021bayesian,griffiths2023applications} 
have been developed as effective and efficient solution strategies. 

\vspace{0.1cm}\noindent\textbf{Importance of Uncertainty Quantification:}
The uncertainty quantification problem is of great importance in various disciplines for complex system modeling and scientific discovery. Knowing the uncertainty associated with a certain prediction will help us develop more reliable models and making better decisions, especially for some safety-critical applications~\cite{mcallister2017concrete}. As some modern machine learning models such as deep neural networks have great approximation capacity and expressiveness, the aleatoric uncertainty needs to be taken great care to avoid over-fitting. 
Moreover, online machine learning strategies, such as \emph{Bayesian active learning}~\cite{cohn1996active, gal2017deep,Zhao2021SMOCU,Zhao2021WMOCU,Zhao2021GPC} and \emph{Bayesian optimization}~\cite{kushner1964new, mockus2012bayesian}, can be combined with the inverse uncertainty quantification to facilitate new material and compound discovery~\cite{Solomou2018,Qian-PRM,Anjana2019,lei2021bayesian}. 

\subsubsection{Uncertainty Quantification in Computational Science}

\vspace{0.1cm}\noindent\textbf{Forward and Inverse Uncertainty Quantification:} In computational science, the quantification of uncertainty is typically categorised into \emph{forward uncertainty propagation} and \emph{inverse uncertainty quantification}. 
The objective of \emph{forward uncertainty propagation}, a.k.a. \emph{sensitivity analysis}~\cite{RAZAVI2021104954, rochman2014efficient, peherstorfer2018survey}
, is to measure how much the randomness of a certain input will result in the uncertainty of system output. By modeling input factors or model parameters as random variables with corresponding probability distributions, the randomness or uncertainty of the system output can be captured by forward uncertainty propagation. In many cases when the computational models are too complex such that the output random variable do not have the closed-form probability distribution, the forward uncertainty is often estimated by \emph{Monte Carlo}~(MC) sampling. Other forward UQ methods to alleviate the high computational cost of MC sampling include \emph{Taylor approximation}~\cite{fornasini2008uncertainty} and other \emph{surrogate modeling} strategies~\cite{box2007response}, which have been extended to UQ in deep learning as discussed in the next section. 

On the other hand, the \emph{inverse uncertainty quantification}, a.k.a. \emph{model calibration}~\cite{malinverno2004expanded, nagel2016unified, nagel2019bayesian}, aims at measuring how uncertain we are about the corresponding parameters of the system model or input factors that modulate the underlying physical process of the system, and then further reducing the relevant uncertainties. 

One powerful method to solve the inverse UQ problem is Bayesian modeling. Compared to the frequentist approaches modeling parameters as deterministic variables and derive point estimates that best fit the selected model with the observed data, 
Bayesian approaches consider model parameters as random variables and solve the Bayesian inverse problem to update the corresponding probabilities to derive predictive posterior belief of a certain outcome following the Bayes' theorem. As a simple illustration, assume that we want to quantify the uncertainty of the system input $X$, and let $Y$ denote the corresponding system output, which can be noisy. Bayes' theorem states
\begin{equation}
    p(X|Y) = \frac{p(Y|X) p(X)}{p(Y)}, 
\end{equation}
where $P(X)$ is the \emph{prior distribution} representing our prior belief of $X$ without any observation $Y$, $P(Y|X)$ is the \emph{likelihood}, the probability distribution of system output being $Y$ given $X$ based on the adopted model assumptions. The denominator $P(Y)$ is often called the \emph{evidence}, which is the marginal distribution $P(Y) = \int P(Y|X) P(X) d X$ over the randomness of $Y$. The \emph{posterior distribution} $p(X|Y)$ captures our updated belief of $X$ after observing the system output $Y$. The same idea can be applied to quantify the uncertainty of system parameters. We can quantify the inverse uncertainty by Bayesian inference based on Bayes's theorem to derive the probability distributions of the corresponding system input or model parameters.  

\vspace{0.1cm}\noindent\textbf{Other Notions of Uncertainty:}
Although the Bayesian uncertainty has long been the primary notion of uncertainty in various applications for its simplicity and soundness in both applied mathematics and probability theory, there are also many other notions of uncertainty other than Bayesian uncertainty. Those includes other methods with probabilistic predictions~\cite{nix1994estimating}, making  \emph{interval predictions}~\cite{koenker2005quantile, angelopoulos2021gentle}, assigning each prediction with a \emph{confidence score}~\cite{jumper2021highly}, as well as \emph{distance-based uncertainty}~\cite{sheridan2004similarity, liu2018molecular, hirschfeld2020uncertainty}. More recently, a variant of Bayesian uncertainty quantification methods called uncertainty quantification of the 4th kind~(UQ4K)~\cite{bajgiran2022uncertainty} has been proposed to alleviate ``brittleness of Bayesian inference'', which is a phenomenon that Bayesian inference could be sensitive to the choice of prior~\cite{owhadi2015brittleness}. In UQ4K, the authors have developed UQ in the game theory framework. Via a min-max game on the risk between the estimation of model parameters and the prior distribution, the authors promote a \emph{hypothesis testing notion of uncertainty}, which gets rid of the choice of prior and does not suffer from the ``Bayesian brittleness''. While those UQ methods are less explored compared to the Bayesian UQ approaches, they can be useful for certain applications with corresponding advantages over Bayesian UQ, for example, lower computational cost, better scalability, and solution properties with theoretical guarantee.

\subsubsection{Uncertainty Quantification in Deep Learning}
In machine learning, most of the existing UQ methods are based on Bayesian statistics and probability theory. 
One specific example is \emph{Bayesian linear regression}~\cite{box2011bayesian}, which is the corresponding Bayesian adaptation of \emph{linear regression}. Bayesian linear regression similarly considers a linear parametrized model from the input $\bm{x}$ to the output $y$ with observation noise $\epsilon$, but with model parameters $\bm{w}$ treated as a random vector with Bayesian inference to update the corresponding posterior rather than solving for deterministic point estimates. With training data $\{\bm{x}, y\}_{n=1}^N$, the posterior update rule gives:
\begin{equation}
p(\bm{w}|\{\bm{x}, y\}_{n=1}^N) = \frac{p(\{y\}_{n=1}^N|\{\bm{x}\}_{n=1}^N, \bm{w})p(\bm{w})}{p(\{y\}_{n=1}^N|\{\bm{x}\}_{n=1}^N)}.    
\end{equation}
The uncertainty is well-preserved in the posterior distribution of $\bm{w}$, and can be quantified further for the posterior predictive distribution on $\hat{y}$ given a new test point $\bm{\hat{x}}$. With $\bm{x}$ projected into the kernel space, we can further extend to another popular Bayesian learning method: \emph{Gaussian process regression}~(GPR)~\cite{williams2006gaussian}, 
as the Bayesian adaptation of \emph{kernel ridge regression}. Although many machine learning models have been implicitly derived in the traditional frequentist fashion, most of them can be adapted into the Bayesian counterparts with the UQ capability. 

There are also other non-Bayesian UQ methods in machine learning, including \emph{conformal prediction}~\cite{angelopoulos2021gentle} and \emph{quantile regression}~\cite{koenker2005quantile}, with the objective to predict an interval of outcome predictions instead of either deriving point estimates or updating distributions. 

\vspace{0.1cm}\noindent\textbf{UQ with Bayesian Neural Networks:} 
\emph{Bayesian neural networks}~(BNNs)~\cite{lampinen2001bayesian, titterington2004bayesian, goan2020bayesian} have been proposed as Bayesian counterparts of frequentist training of artificial neural networks (ANNs), by modelling the network parameters or activation as random variables. We here use a regression problem to illustrate the main idea, and assume that the model parameters $\bm{\theta}$ are modeled as random variables.

Given a data point $\bm{{x}}$ and $f_{\bm\theta}$, which is a prespecified neural network architecture $f$ with the model parameters $\bm{\theta}$, the \emph{aleatoric uncertainty} is typically modeled as the $\bm{{y}}$ being the output of neural network $f_{\bm{\theta}}(\bm{x})$ with a random noise $\bm{\epsilon}$ added: $\bm{{y}} = f_{\bm{\theta}}(\bm{{x}}) + \bm{\epsilon}$, which implicitly specified a distribution $p(\bm{y}|\bm{\theta}, \bm{x}, f)$ and further $p(\{ \bm{y}\}_{n=1}^{N}|\bm{\theta}, \{\bm{x}\}_{n=1}^N, f)$ under some independence assumption. The noise $\bm{\epsilon}$ is typically modeled as a zero-mean Gaussian random variable with the noise being a hyperparameter, or data-dependent through a neural network, which is also termed as \emph{mean variance estimation}~(MVE)~\cite{nix1994estimating}.

On the other hand, the \emph{epistemic uncertainty} is the uncertainty of model parameters $\bm{\theta}$ given the limited training data $\{\bm{x}, \bm{y}\}_{n=1}^{N}$. According to the Bayes' theorem, the posterior distribution of the model parameters $\bm{\theta}$ can be derived as:
\begin{equation}
\begin{split}
    p(\bm{\theta}|\{\bm{x}, \bm{y}\}_{n=1}^{N}, f) = \frac{p(\{ \bm{y}\}_{n=1}^{N}|\bm{\theta}, \{\bm{x}\}_{n=1}^N, f)p(\bm{\theta}| f)}{p(\{ \bm{y}\}_{n=1}^{N}|\{\bm{x}\}_{n=1}^N, f)} = \frac{p(\{ \bm{y}\}_{n=1}^{N}|\bm{\theta}, \{\bm{x}\}_{n=1}^N, f)p(\bm{\theta}| f)}{\int p(\{ \bm{y}\}_{n=1}^{N}|\bm{\theta}, \{\bm{x}\}_{n=1}^N, f)p(\bm{\theta}| f) d\bm{\theta}}.
\end{split}    
\end{equation}

With the posterior distribution of model parameters $p(\bm{\theta}|\{\bm{x}, \bm{y}\}_{n=1}^{N}, f)$, we can quantify the uncertainty of the model parameters $\bm{\theta}$. Given a new test data point $\bm{\hat{x}}$, the model prediction is the marginal distribution of $\bm{\hat{y}}$: 
\begin{equation}
p(\bm{\hat{y}}|\{\bm{x}, \bm{y}\}_{n=1}^{N}, \bm{\hat{x}}, f) = \int p(\bm{\hat{y}}|\bm{\theta}, \bm{\hat{x}}, f)p(\bm{\theta}|\{\bm{x}, \bm{y}\}_{n=1}^{N}, f) d\bm{\theta}.    
\end{equation}

The expectation $\mathbb{E}[\bm{\hat{y}}|\{\bm{x}, \bm{y}\}_{n=1}^{N}, \bm{\hat{x}}, f]$ can be used as the point prediction of $\bm{\hat{y}}$.  The total uncertainty of the forward prediction is the uncertainty of the marginal distribution $p(\bm{\hat{y}}|\{\bm{x}, \bm{y}\}_{n=1}^{N}, \bm{\hat{x}}, f)$. Different metrics can be used to quantify this uncertainty, such as the variance or the (differentiable) entropy. However, the denominator $\int p(\{ \bm{y}\}_{n=1}^{N}|\bm{\theta}, \{\bm{x}\}_{n=1}^N, f)p(\bm{\theta}| f) d\bm{\theta}$ is often intractable, which makes the computation of $p(\bm{\theta}|\{\bm{x}, \bm{y}\}_{n=1}^{N}, f)$ difficult. Various inference methods based on either \emph{Markov chain Monte Carlo}~(MCMC) sampling with corresponding gradient-based variants~\cite{welling2011bayesian} or \emph{variational inference}~\cite{blei2017variational} have been developed to address the challenge. We refer the readers to~\citet{jospin2022hands} for a comprehensive review of different approximate inference methods. 

\vspace{0.1cm}\noindent\textbf{Scalable Approximate Inference for Deep Neural Networks:} 
Bayesian inference becomes more challenging with millions of model parameters in modern deep neural network~(DNN) architectures. Many recent research efforts aim at scaling up the approximate inference algorithms, especially for efficient Bayesian inference with DNN architectures. Some commonly adopted approximation UQ methods include \emph{Bayes-by-backprop}~(BBB)~\cite{blundell2015weight}, \emph{Monte Carlo dropout}~(MC dropout)~\cite{gal2016dropout, gal2017concrete}, and \emph{deep ensemble}~(ensemble)~\cite{lakshminarayanan2017simple, huang2017snapshot}. BBB~\cite{blundell2015weight} is an optimization trick which can be coupled seamlessly with the backpropagation algorithm and the autodifferentiation training when the neural network parameters are modeled as random variables with reparameterizable variational distributions. 
MC dropout~\cite{gal2016dropout, gal2017concrete} is a simple and efficient way to provide uncertainty estimation on any model trained with dropout~\cite{srivastava2014dropout}, a regularization technique shutting down selected neurons randomly. By performing dropout at both training and testing time, MC dropout has been proven to be able to provide an approximation inference on random neural network weights. 
Deep ensemble~\cite{lakshminarayanan2017simple, huang2017snapshot} is another heuristic strategy to achieve effective predictive uncertainty by combining the models in different local optima from random initializations or multiple ``snapshots'' during training. Deep ensemble and its variants have been shown to have an interpretation in Bayesian perspective~\cite{pearce2020uncertainty}. Some other heuristic approaches to further scale up BNN inference have been developed by combining BNN with frequentist neural network training. For example, \emph{Bayesian last layer}~(BLL)~\cite{brosse2020last} has shown to be surprisingly effective for its simplicity by modelling only the last layer network parameters as random variables, which significantly reduces the number of uncertain parameters by fixing all the other parameters as point estimates except those in the final layer, under the premise that earlier layers are performing feature extraction and later layers are performing the final prediction task. These approximate inference methods can either be used alone or combined together to achieve better uncertainty estimation with improved scalability and computational efficiency. 

\vspace{0.1cm}\noindent\textbf{Evidential Deep Learning:} \emph{Evidential deep learning}~(EDL)~\cite{sensoy2018evidential, amini2020deep} is a recent emerging UQ strategy for deep learning based on \emph{Theory of Evidence}, a generalized Bayesian formulation. EDL explicitly considers the uncertainty due to the lack of evidence, which is the amount of support from data for certain prediction. The network output of a test data without the support of evidence is encouraged to be a predefined prior other than a confident prediction. To model the epistemic uncertainty, instead of considering the model parameters $\bm{\theta}$ to be random, EDL alternatively introduces a random vector $\bm{\pi}$, with the predictive distribution now becoming
\begin{equation}
   p(\bm{\hat{y}}|\bm{\theta}, \bm{\hat{x}}, f) = \int p(\bm{\hat{y}}|\bm{\pi}) p(\bm{\pi}|\bm{\theta}, \bm{\hat{x}}, f) d\bm{\pi}, 
\end{equation}
where $p(\bm{\hat{y}}|\bm{\pi})$ is typically a categorical distribution for classification and Gaussian for regression, similar as most of the existing DNN models. The prior distribution of random vector $\bm{\pi}$ is typically a conjugate prior for $p(\bm{\hat{y}}|\bm{\pi})$ and the neural network models the likelihood of $\bm{\pi}: p(\bm{\hat{x}}|\bm{\pi}, \bm{\theta}, f)$. The training procedure is similar as the usual neural network training except for an extra penalty term that increases as the posterior of $\bm{\pi}$ becomes far away from the prior distribution. Some of EDL models~\cite{sensoy2018evidential} can also be interpreted as a special case of BNN by performing amortized variational inference on the last layer activation. The EDL framework has been shown to be especially useful for out-of-distribution data detection in various classification~\cite{sensoy2018evidential, stadler2021graph} and regression tasks~\cite{amini2020deep}. We give a schematic illustration of EDL along with some other UQ methods for DNN including BNN and MVE in Figure~\ref{fig:uqmethods}.

\begin{figure}[h]
    \centering
    {\includegraphics[width = \textwidth]{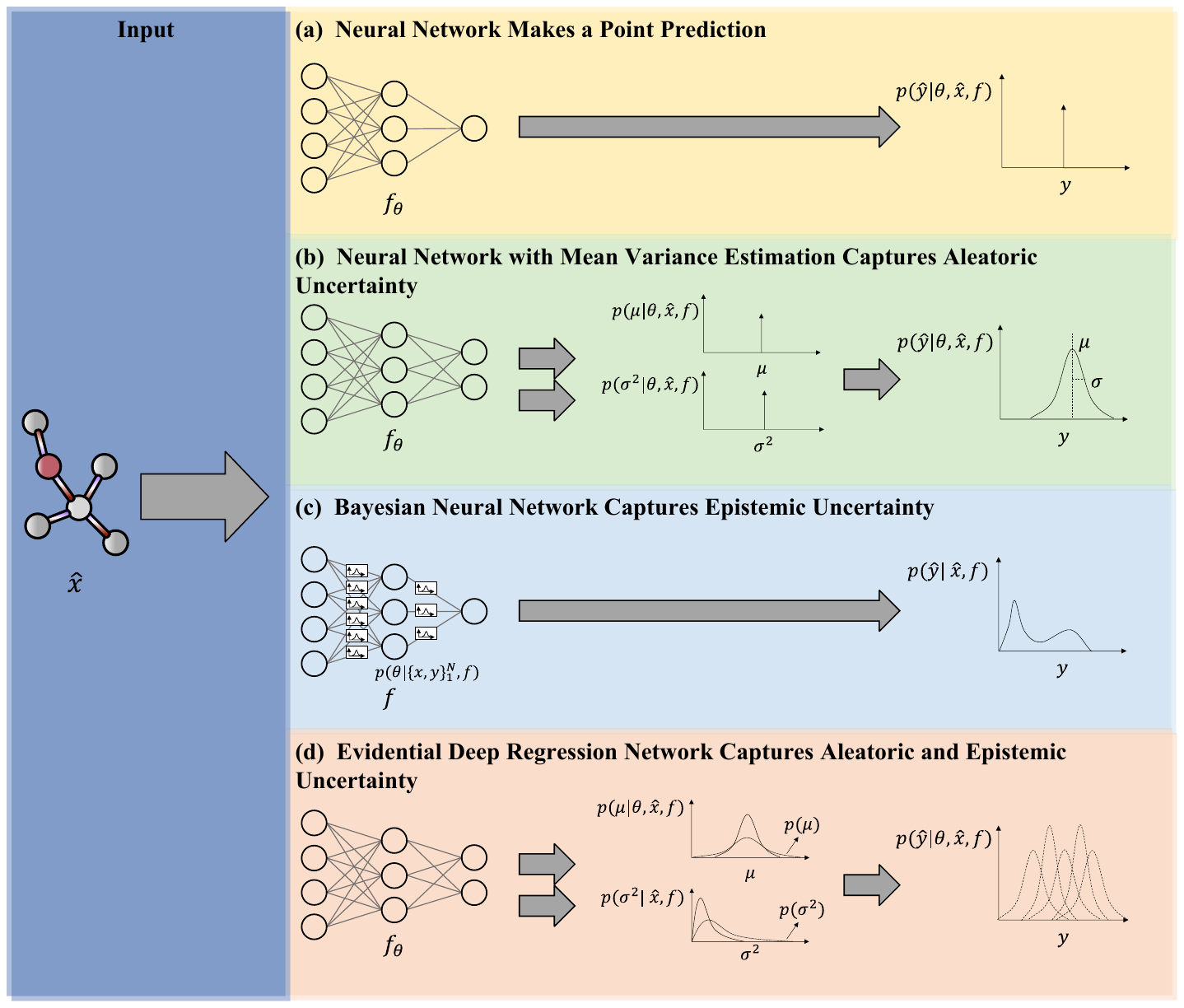}}
    \caption{Schematic illustration of different uncertainty quantification methods on a molecular energy prediction task, with $\hat{y}$ denoting the predicted energy of a given molecule. Note that (c) Bayesian Neural Network can be further combined with Mean Variance Estimation to capture both aleatoric and epistemic uncertainty.}
    \label{fig:uqmethods}
\end{figure}

\vspace{0.1cm}\noindent\textbf{Uncertainty Quantification for Graph Learning:}
When modeling dependency across different system components, graph neural networks (GNNs) have been developed with successes in diverse applications, including materials science, molecular biology, and quantum mechanics as detailed in previous sections. 
Existing UQ methods developed specifically for graph learning includes Graph DropConnect~(GDC)~\cite{hasanzadeh2020bayesian}, Bayesian graph convolutional neural network~(BGCNN)~\cite{zhang2019bayesian, pal2020non}, variational inference for graph convolution networks~(VGCN)~\cite{sigvae2019,vgrnn2019,elinas2020variational}, graph posterior network~(GPN)~\cite{stadler2021graph}, gaussian process with graph convolutional kernel~(GPGC)~\cite{fang2021gaussian}, and conformalized GNN~(CF-GNN)~\cite{huang2023uncertainty}. 
Being a generalization of MC Dropout for GCNN with the dropout rates as learnable parameters, GDC~\cite{hasanzadeh2020bayesian} provides UQ capability by modeling the neural network weights as Bernoulli random variables. BGCNN~\cite{zhang2019bayesian, pal2020non} further considers the topology uncertainty in graph structure by modeling the observed graph as a noisy observation of the true node relationship, and posterior inference on true node relationship is performed using the mixed membership stochastic block model~(MMSBM). VGCN~\cite{elinas2020variational} similarly models the structure uncertainty by performing variational inference on graph adjacency matrix and show this treatment can improve the model robustness to adversarial perturbation on graph structure. GPN~\cite{stadler2021graph} is a UQ method for node classification task based on EDL, which is shown to be effective in OOD node detection. GPGC~\cite{fang2021gaussian} is a graph Gaussian process model~\cite{venkitaraman2020gaussian} with the kernel function defined over a deep GCNN, whose parameters are learned through the variational inducing point~\cite{titsias2009variational}. CF-GNN~\cite{huang2023uncertainty} extends conformal prediction for the classification and regression tasks under an independence assumption to the graph-based models with theoretical analysis of validity conditions. The authors have shown satisfactory UQ performances with predicted intervals by CF-GNN covering the ground truth on various datasets.

Although all these previous efforts consider different uncertainty in graph learning,
few of them actually quantify and analyze the estimated uncertainty. The appropriate treatment of uncertainty in graph learning is still an under-explored area despite recent success of GNNs on various graph learning tasks in material and protein property prediction.

\subsubsection{Uncertainty Quantification in AI for Science}

\begin{table}[t]
	\centering
	\caption{Existing research and benchmark studies on UQ for PDE surrogate solutions, DFT-related tasks, molecular property prediction, and compound-protein binding prediction. }
	\begin{tabular}{lcm{6cm}}
		\toprule[1pt]
		Application & Method &  Adopted UQ Approach(es) \\\hline
        \multirow{17}{*}{\shortstack{Molecular Property }}& \citet{ryu2019bayesian} & MVE, MC Dropout \\
        & \citet{zhang2019bayesian2} & BNN\\
        & \citet{scalia2020evaluating} & MC Dropout, Ensemble\\
        & \citet{hirschfeld2020uncertainty} & MVE, Ensemble, GP, MC Dropout, RF~\cite{ho1995random}\\
         & \citet{tran2020methods} & MVE, BNN, MC Dropout, Ensemble, GP  \\    
        & \citet{hie2020leveraging} & Ensemble, GP, BNN\\
        & \citet{soleimany2021evidential} & EDL \\
        & \citet{yang2023explainable} & MVE, Ensemble \\
        & \citet{greenman2023benchmarking} & MVE, MC Dropout, Ensemble, GP, EDL\\
        & \citet{griffiths2022gauche} & GP \\
        & \citet{griffiths2023applications} & GP \\
        & \citet{li2023muben} & MVE, Ensemble, MC Dropout, BNN, Focal Loss~\cite{lin2017focal, mukhoti2020calibrating}, SWAG~\cite{maddox2019simple}\\
        & \citet{wollschlager2023uncertainty} & GP \\\hline
        Binding Affinity & \citet{hirschfeld2020uncertainty} & GP \\\hline
        \multirow{2}{*}{DFT}& \citet{fowler2019managing} & MVE, Ensemble\\
        & \citet{mahmoud2020learning}  & GP \\\hline
      PDE & \citet{psaros2023uncertainty} & GP, MCMC, Ensemble, MC Dropout, BNN, FP\\    
		\midrule
		\bottomrule 
	\end{tabular}\label{tab:uqresearch}
\end{table} 

Compared to other machine learning applications in computer vision and natural language processing, the problem of training under data scarcity is even more severe for scientific AI as the experiments typically being expensive and time-consuming to be conducted to collect meaningful data. 
Similar as any existing data-driven black-box model, the DNN-based scientific AI models can make erroneous but overconfident prediction for unseen input data~\cite{hein2019relu}, and are particularly vulnerable to adversarial attacks~\cite{goodfellow2014explaining}. Therefore, there has been a growing interest in equipping those scientific AI models with the UQ capability. Many different benchmark studies and research efforts have been made to test the idea of using Gaussian process~(GP)~\cite{hie2020leveraging, mahmoud2020learning, hirschfeld2020uncertainty, tran2020methods, griffiths2022gauche, griffiths2023applications, wollschlager2023uncertainty, li2023muben, psaros2023uncertainty, greenman2023benchmarking}, deep ensemble~\cite{fowler2019managing, scalia2020evaluating, hirschfeld2020uncertainty, tran2020methods, yang2023explainable, mariet2020deep, hie2020leveraging, greenman2023benchmarking, li2023muben}, MC dropout~\cite{ryu2019bayesian, zhang2019bayesian2, scalia2020evaluating, hirschfeld2020uncertainty, tran2020methods, greenman2023benchmarking, li2023muben}, Bayesian neural network~\cite{zhang2019bayesian2, tran2020methods, hie2020leveraging, li2023muben} and EDL~\cite{soleimany2021evidential, greenman2023benchmarking} to quantify the uncertainty of molecular property prediction~\cite{ryu2019bayesian, zhang2019bayesian2, scalia2020evaluating, hirschfeld2020uncertainty, tran2020methods, hie2020leveraging, soleimany2021evidential, griffiths2022gauche, yang2023explainable, greenman2023benchmarking, griffiths2023applications, li2023muben, wollschlager2023uncertainty}, compound-protein binding prediction~\cite{hirschfeld2020uncertainty}, ground-state density prediction tasks~\cite{fowler2019managing, mahmoud2020learning}, as well as PDE surrogate prediction tasks with physics-informed neural network~(PINN)~\cite{psaros2023uncertainty}, along with seismic inversion~\cite{smith2020eikonet,smith2022hyposvi}. We summarize existing UQ research and benchmark studies applied to the above disciplines with the adopted UQ approaches in Table~\ref{tab:uqresearch}. 

In particular, for chemical and protein molecular property prediction tasks, \citet{ryu2019bayesian} have shown that MVE and MC Dropout based UQ methods can be used to assess the data quality. In \citet{zhang2019bayesian2}, Stein variational gradient descent~(SVGD)~\cite{liu2016stein} inference algorithm has been implemented for BNN training to model the epistemic uncertainty and show that such an implementation can be used to mitigate the potential dataset bias and integrated into the active learning cycle to further improve the data efficiency. 
\citet{soleimany2021evidential} further test the idea of modeling the epistemic uncertainty via EDL, with the quantified uncertainty correlated with prediction error. \citet{yang2023explainable} also have tested MVE and Deep Ensemble based UQ on a molecular property prediction task, demonstrating the effectiveness in data noise identification and OOD data detection. \citet{griffiths2022gauche} and \citet{griffiths2023applications} use GP to model different types of uncertainties in molecular property prediction as well as molecular discovery tasks, which have been further extended to an open-source package to facilitate real-world scientific applications.

Among the DFT-related quantum mechanics computation tasks, \citet{fowler2019managing} have estimated different types of uncertainty using MVE and deep ensemble in a task to predict ground state electron density and show that the quantified uncertainty is informative to help detect inaccurate predictions. \citet{mahmoud2020learning} use a sparse Gaussian process to quantify the uncertainty in predicting the electronic density of states using the quasiparticle energy levels, and show that the predicted uncertainty can identify the problematic test structures. \citet{tran2020methods} provide a benchmark study of UQ on a task to predict the adsorption energies of materials given atomic structures, with the best-performing model being a GP added at the end of a convolutional neural network. \citet{wollschlager2023uncertainty} propose six desiderata for UQ in molecular force field and further introduce localized neural kernel~(LNK), a GNN-based deep kernel for GP-based uncertainty quantification, which is the first method to fulfill all of the six desiderata.  

For deep models as surrogates for solving PDE problems, \citet{psaros2023uncertainty} have conducted a comprehensive study by applying different UQ methods on PINN and DeepONet with various evaluation metrics on forward PDE problems, mixed PDE with known and unknown noise, and operator learning problems.
The authors also propose to combine generative adversarial network~(GAN) and GP as a functional prior~(FP) to harness historical data and reduce the computational cost. While the relative performance of different methods differs in different tasks, the quantified uncertainty is shown to be indicative of prediction error and informative for detecting OOD data.

There are also several benchmark studies aiming at comparing different UQ methods in terms of the informativeness to error, faithfulness of data fitting, and calibration in molecular property prediction tasks. \citet{scalia2020evaluating} have benchmarked different UQ methods for chemical molecular property prediction tasks with the results in favor of ensemble-based UQ. In \citet{hirschfeld2020uncertainty}, different UQ methods for small organic molecules property prediction have been tested showing that random forest~(RF)~\cite{ho1995random} and GP predictors on GNN-based features can provide the best UQ performance. \citet{greenman2023benchmarking} have provided another benchmark on UQ for protein property prediction with the results suggesting that no UQ method can perform  consistently better than other competitive methods. \citet{li2023muben} also benchmark UQ methods for molecular property prediction with various training schemes, network architectures as well as post-hoc calibration approaches, whose results suggest that different UQ methods may surpass others on different tasks with different experiment setups.

To summarize, by incorporating different UQ methods, existing scientific AI models for various tasks can get reasonable uncertainty without harming the predictive performance. Moreover, the quantified uncertainty can be useful for OOD data detection~\cite{fowler2019managing, mahmoud2020learning, yang2023explainable}, data noise identification~\cite{yang2023explainable}, and has the potential to be incorporated into active learning~\cite{zhang2019bayesian2, soleimany2021evidential, hie2020leveraging} and Bayesian experimental design~\cite{hie2020leveraging} cycles for data-efficient model training and new molecular discovery. 

\subsubsection{Open Research Directions}

\vspace{0.1cm}\noindent\textbf{Evaluation of Uncertainty Quantification (UQ):} 
One major difference between scientific AI modeling and other machine learning tasks is that often AI/ML models are considered as surrogates for more computationally expensive mechanistic models based on physics principles. Although existing research and benchmark studies have designed various UQ evaluation metrics by considering different aspects of uncertainty for AI/ML surrogates, comprehensive UQ evaluation is still challenging due to prohibitive computational cost to simulate all the underlying stochastic scenarios. It may be critical to construct new benchmark datasets based on corresponding high-fidelity computational methods and develop UQ evaluation metrics to help standardize UQ for scientific AI with guaranteed performance. 

\vspace{0.1cm}\noindent\textbf{Development of UQ Methods with Domain Knowledge:} As has been pointed out in many existing benchmark studies ~\cite{scalia2020evaluating, hirschfeld2020uncertainty, psaros2023uncertainty}, even though some UQ methods may perform relatively well on a specific task, there is no existing UQ method that can consistently outperform other methods with different setups on different evaluation metrics. As there is not a general rule of thumb for applying UQ methods, many existing deep models for science with UQ capability are developed by empirically testing different existing UQ methods. There is a lack of UQ methods with the properties of physical or biological process explicitly considered. The development of new UQ methods for scientific AI models with the integrated domain knowledge is a research direction to be considered in the future. 

\vspace{0.1cm}\noindent\textbf{Scalable UQ Approaches for Large Models: } Most of the existing UQ methods for Deep Neural Networks are either too simplistic and restrictive to achieve satisfactory UQ performance, \emph{e.g.}, MC dropout, or computationally too expensive to be deployed in practice, \emph{e.g.}, deep ensemble and BNN. As the large models with billions of parameters start dominating more and more tasks in natural  language processing and computer vision, there has been increasing interest in applying those large models on scientific discovery. Therefore, it is a promising research direction to develop new UQ methods, which is scalable for large models and can better trade-off between computational complexity and quality of the quantified uncertainty. Various types of approximate heuristics, stochastic gradient sampling variants, as well as variational inference techniques have been developed~\cite{graves2011practical, welling2011bayesian, li2016preconditioned, blundell2015weight,shi2017kernel, gal2016dropout, gal2017concrete,boluki2020learnable,dadaneh2020pairwise,fan2020bayesian}. New approximation strategies, including recent subspace-based methods~\cite{Izmailov19,NEURIPS2019_1113d7a7,RANK-1-BNN,ChenGhattas20,BayPOD}, may help further scale up UQ for large models. 
\clearpage

\section{Learning, Education, and Beyond}

The advancement of AI holds immense promise for accelerating scientific discovery, driving innovations, and solving complex problems across various domains. However, to fully harness the potential of AI for scientific research, new challenges are faced on education, workforce development, and public engagements. In this section, we first collect existing resources for fundamentals of each AI and Science field. Next, we identify three main paradigm shifts, including discipline boundaries, communities, and educational resources. Finally, we point out recent progress and call for future actions to construct our new knowledge and community system to support the ever-growing AI for Science field.

\subsection{Existing Resources for Fundamental AI and Science}

\noindent{\emph{Authors: Yuanqi Du, Yaochen Xie, Xiner Li, Shurui Gui, Tianfan Fu, Jimeng Sun, Xiaofeng Qian, Shuiwang Ji}}\newline

In the ever-evolving landscape of AI and scientific fields, traditional learning materials such as books and courses have long served as the fundamental for knowledge acquisition. Conferences, particularly within AI and each scientific community, have been the traditional medium for fostering collaboration and sharing groundbreaking research. In this section, we lay out existing resources of different types covering fundamentals of each individual field of AI and Science in Table~\ref{tab:airs-edu-fundamental} and Table~\ref{tab:airs-edu-book}. Specifically, the most representative resource types are books, courses and libraries developed for computational purposes.

\subsection{Paradigm Shifts in AI for Science}

\noindent{\emph{Authors: Yuanqi Du, Yaochen Xie, Xiner Li, Shurui Gui, Tianfan Fu, Jimeng Sun, Xiaofeng Qian, Shuiwang Ji}}\newline

Despite the accumulating resources for individual AI and Science fields, the emerging field of AI for Science continues to grapple with substantial paradigm shifts. Educational resource types, community levels, and knowledge collection methods specific to AI for Science remain fragmented, necessitating a need for consolidation and further development. We identify three main paradigm shifts in this section.

\vspace{0.1cm}\noindent\textbf{Transcending Discipline Boundaries:} The knowledge system in AI for Science should transcend disciplinary boundaries, promoting interdisciplinary collaborations to address multifaceted challenges and opportunities (Figure~\ref{fig:edu1}). Breaking traditional silos between scientific disciplines fosters a comprehensive understanding of complex problems and encourages innovative solutions. The integration of diverse perspectives from fields like physics, biology, chemistry, and artificial intelligence enhances knowledge breadth and facilitates the cross-pollination of ideas. This symbiotic relationship between AI and Science not only allows for shared problem-solving approaches but also enables principled advancements in AI research and development. By leveraging scientific principles, methodologies, and interdisciplinary techniques, breakthroughs in scientific discovery can be achieved to tackle pressing challenges in the AI for Science domain. Existing examples have already demonstrated the power of integrating AI and Science, such as interpreting neural networks with physical laws~\cite{sorscher2022beyond,di2023graph}, designing generative models with dynamical system, control and optimal transport~\cite{song2020score,xupoisson,liu2022flow,berner2022optimal}, solving grand challenges like protein structure prediction~\cite{jumper2021highly}, and more. This collaborative approach will pave the way for breakthroughs in scientific discovery and enable us to tackle the most pressing challenges in the realm of AI for Science.

\vspace{0.1cm}\noindent\textbf{Fostering a Diverse and Agile Community:} The AI for Science community celebrates diversity and flexibility, extending beyond the boundaries of AI and Science to include students, researchers, and practitioners with a shared interest in advancing the field. While traditional events like NeurIPS and ACS meetings have showcased the forefront of AI and Science research, there is a growing recognition of the need to expand beyond these established platforms. For example, there has been a significant increase in AI-related articles in chemical journals, with a 133\% rise in ACS Omega from 2020 to 2021~\cite{imberti2022diving}. Initiatives such as AI for Science workshops and symposiums have emerged, aiming to foster dialogue and collaboration between the AI and Science communities (Table~\ref{tab:airs-edu-ai4science}). These events range from local gatherings to global conferences and facilitate collaborations between industry and research institutions (Figure~\ref{fig:edu2}). By embracing diversity, the community nurtures creativity, critical thinking, and unconventional problem-solving, harnessing a wealth of expertise and insights from individuals with diverse backgrounds. This collaborative environment propels the progress of AI for Science.

\begin{figure}[t]
    \centering
    {\includegraphics[width = \textwidth]{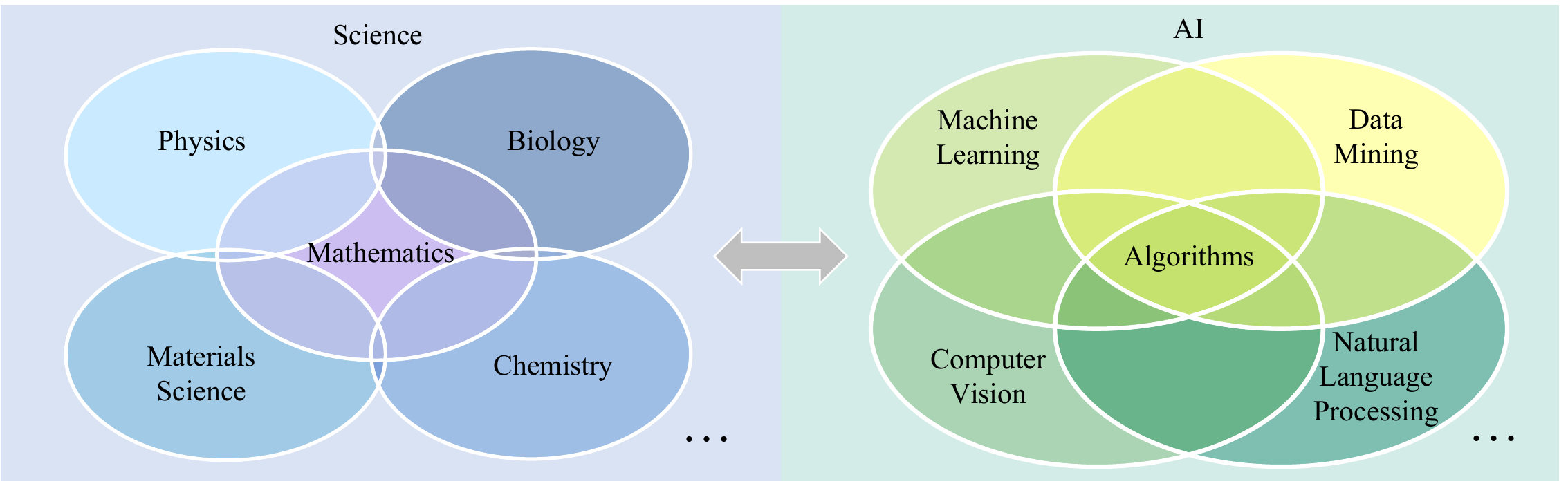}}
    \caption{Traditional scientific fields have recognized the power of interdisciplinary collaborations, leading to the emergence of new fields. Similarly, the intersection of AI and Science promises to forge new frontiers, as these two domains merge their strengths and synergize to tackle challenges in both AI and Science. Note that there are far more scientific fields and cross-domain overlappings that cannot be illustrated in this figure.}
    \label{fig:edu1}
\end{figure}

\vspace{0.1cm}\noindent\textbf{Enriching Educational Landscape:} The educational resources in AI for Science are expanding beyond what we have seen before. As AI continues to revolutionize the scientific landscape, the demand for high-quality educational resources in AI for Science is growing rapidly. To meet this demand, numerous institutions and individuals are offering an array of educational resources, including summer schools, blogs, tutorials, paper reading groups, \emph{etc.}, as summarized in Table~\ref{tab:airs-edu-ai4science}. These resources cover a wide range of topics, from Fundamental concepts in AI to advanced techniques specific to scientific domains. Additionally, collaborations between academia and industry are enriching the educational landscape by providing real-world applications and case studies. However, despite the commendable efforts made to expand educational resources, challenges persist in establishing a systematic approach to learning AI for Science. The rapid pace of methodological advancements and the interdisciplinary nature of the field pose hurdles in curating a comprehensive curriculum. To address this, it becomes imperative to prioritize the development of structured educational programs that encompass the breadth and depth of AI for Science. Such programs should provide a well-rounded understanding of fundamental concepts, advanced methodologies, and their applications across scientific disciplines.

\subsection{Prospective and Proposed Actions}

\noindent{\emph{Authors: Yuanqi Du, Yaochen Xie, Xiner Li, Shurui Gui, Tianfan Fu, Jimeng Sun, Xiaofeng Qian, Shuiwang Ji}}\newline

Significant progresses have been made to develop resources for AI for Science in recent years (Table~\ref{tab:airs-edu-ai4science} and Table~\ref{tab:airs-edu-book}). However, these resources often operate independently, lacking a cohesive and systematic road map. As the field undergoes paradigm shifts, it is crucial to develop a unified road map and resources to fill the missing gap. By recognizing the need for comprehensive educational materials, collaborative community platforms, and effective knowledge collection methods, we can better equip researchers, practitioners, and students with the tools and insights necessary to navigate the evolving landscape of AI for Science. Building on top of them, individuals are encouraged to contribute their expertise to subareas of AI for Science. It is through collective efforts that we can fully leverage the potential of AI for Science.

\begin{figure}[t]
    \centering
    {\includegraphics[width = 0.65\textwidth]{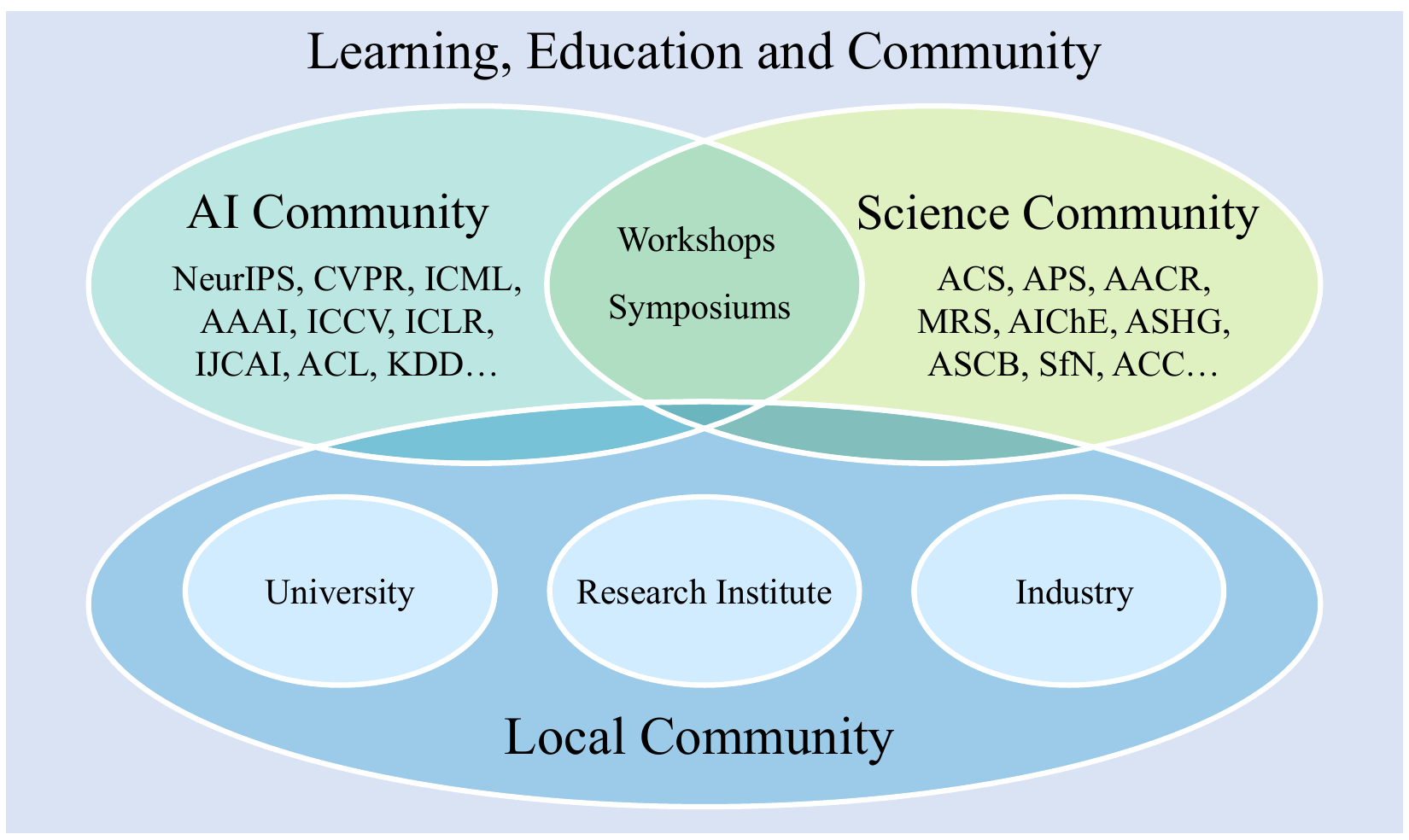}}
    \caption{Within the realm of AI and Science, there exists a vibrant global community encompassing researchers, experts, and enthusiasts from various backgrounds. This global network is complemented by local communities, including universities, research institutes, and industry partners. Through concerted efforts and collaboration, these communities form a powerful ecosystem, driving innovation, knowledge exchange, and transformative discoveries at both local and global scales.}
    \label{fig:edu2}
\end{figure}

\begin{table}[!t]\centering
\caption{Learning resources for fundamental AI or Science fields (open-source denotes that the code is publicly available and further development is permitted). Note that this table is by no means complete and only consists of a small set of available resources.}\label{tab:airs-edu-fundamental}
\fontsize{6}{8}\selectfont
\begin{tabular}{ p{1.5cm} | m{3cm} m{2cm} m{6cm} }
\toprule
&Fundamental AI/Science &Type &Description \\
\midrule
\multirow{8}{*}{\parbox{1.5cm}{Symposiums/ Conferences}} &\href{https://www.aps.org/}{APS} & Physics & American Physical Society\\
&\href{https://www.acs.org}{ACS} & Chemistry & American Chemical Society\\
&\href{https://www.mrs.org}{MRS} & Materials & Materials Research Society\\
&\href{https://www.aiche.org}{AIChE} & Chemistry & American Institute of Chemical Engineers\\
&\href{https://www.tms.org}{TMS} & Materials & The Minerals, Metals \& Materials Society\\
&\href{https://nips.cc/}{NeurIPS} & AI & Neural Information Processing System \\
&\href{https://iclr.cc/}{ICLR} & AI & Intl. Conf. on Learning Representations\\
&\href{https://icml.cc/}{ICML} &AI & Intl. Conf. on Machine Learning\\
&\href{https://aaai.org/aaai-conferences-and-symposia/}{AAAI} & AI & AAAI Conference on Artificial Intelligence\\
\hline\multirow{6}{*}{Courses} 
&\href{http://web.stanford.edu/class/cs279/}{Computational Biology} &Biology & -\\
&\href{https://ocw.mit.edu/courses/8-04-quantum-physics-i-spring-2013/}{Quantum Physics} & Physics & -\\
&\href{https://www.coursera.org/specializations/machine-learning-introduction?utm\_medium=sem\&utm\_source=gg\&utm\_campaign=B2C\_NAMER\_machine-learning-introduction\_stanford\_FTCOF\_specializations\_country-US-country-CA\&campaignid=685340575\&adgroupid=46849728719\&device=c\&keyword=andrew\%20ng\%20machine\%20learning\&matchtype=b\&network=g\&devicemodel=\&adposition=\&creativeid=606098666387\&hide\_mobile\_promo\&gclid=CjwKCAjwhJukBhBPEiwAniIcNWMch9sLCCvwsImakJ\_ky0em\_O34bX8d4vFYh9qb-ZRRDdBACjHS6BoC1WwQAvD\_BwE}{Machine Learning} & AI & -\\
&\href{https://www.google.com/search?q=coursera+deep+learning\&rlz=1C5CHFA\_enUS991US991\&oq=coursera+deep+learning\&aqs=chrome..69i57j0i512l9.3855j0j4\&sourceid=chrome\&ie=UTF-8}{Deep Learning} & AI & -\\
&\href{https://www.youtube.com/watch?v=5orzn-XA29M\&t=14s}{Theoretical Chemistry} & Chemistry & -\\
& \href{http://faculty.washington.edu/sbrunton/me565/}{Mechanical Engineering Analysis} & Engineering & -\\
\hline\multirow{45}{*}{Software \& Library} 
&\href{https://pyscf.org/}{PySCF}& Quantum Chemistry & Open-Source Quantum Chemistry Code\\
&\href{https://psicode.org}{PSI4}& Quantum Chemistry & Open-Source Quantum Chemistry Code\\
&\href{https://www.nwchem-sw.org}{NWChem}& Quantum Chemistry & Open-Source Quantum Chemistry Code\\
&\href{https://www.cp2k.org}{CP2K}& Quantum Chemistry &  Open-Source Quantum Chemistry Code\\
&\href{https://orcaforum.kofo.mpg.de/app.php/portal/}{ORCA}& Quantum Chemistry & Quantum Chemistry Code\\
&\href{https://gaussian.com/}{GAUSSIAN}& Quantum Chemistry & Quantum Chemistry Code\\
&\href{https://www.q-chem.com/}{Q-Chem}& Quantum Chemistry & Quantum Chemistry Code\\
&\href{https://www.quantum-espresso.org}{Quantum-ESPRESSO} & First-Principles & Open-Source Electronic Structure Code\\
&\href{https://www.abinit.org}{ABINIT} & First-Principles & Open-Source Electronic Structure Code\\
&\href{https://wiki.fysik.dtu.dk/gpaw/}{GPAW}& First-Principles & Open-Source Electronic Structure Code\\
&\href{https://berkeleygw.org}{BerkeleyGW}& First-Principles & Open-Source  Electronic Structure Code\\
&\href{http://www.west-code.org}{WEST}& First-Principles & Open-Source  Electronic Structure Code\\
&\href{https://www.octopus-code.org}{Octopus}& First-Principles & Open-Source  Electronic Structure Code\\
&\href{https://exciting-code.org}{exciting}& First-Principles & Open-Source  Electronic Structure Code\\
&\href{https://departments.icmab.es/leem/siesta/}{SIESTA}& First-Principles & Open-Source  Electronic Structure Code\\
&\href{https://www.openmx-square.org}{OpenMX}& First-Principles & Open-Source  Electronic Structure Code\\
&\href{http://abacus.ustc.edu.cn/main.htm}{ABACUS} & First-Principles & Open-Source Electronic Structure Code\\
&\href{https://wannier.org}{Wannier90}& First-Principles & Open-Source  Electronic Structure Code\\
&\href{https://epw-code.org}{EPW} & First-Principles &  Open-Source  Electronic Structure Code\\
&\href{http://www.wien2k.at}{WIEN2k} & First-Principles &  Electronic Structure Code\\
&\href{https://www.vasp.at}{VASP} & First-Principles &  Electronic Structure Code\\
&\href{https://fhi-aims.org}{FHI-aims} & First-Principles &  Electronic Structure Code\\
&\href{http://www.castep.org}{CASTEP} & First-Principles &  Electronic Structure Code\\
&\href{https://github.com/materialsproject/pymatgen}{pymatgen} & Materials & Open-Source Python Library for Materials Analysis\\
&\href{https://wiki.fysik.dtu.dk/ase/}{ASE}& Materials & Open-Source Python Library for Atomistic Simulations\\
& \href{https://jarvis-tools.readthedocs.io/en/master/index.html}{JARVIS-Tools} & Materials & Software Package for Atomistic Data-Driven Materials Design\\
& \href{https://www.aflowlib.org/src/paoflow/}{PAOFLOW} & Materials & Open-Source Code for Post-Processing First-Principles Calculations\\
& \href{http://xtalopt.github.io}{XtalOpt} & Materials & Open-Source Crystal Structure Search Code  \\
& \href{http://www.calypso.cn}{CALYPSO} & Materials & Crystal Structure Search Code\\
& \href{https://uspex-team.org/en/uspex/overview}{USPEX} & Materials & Crystal Structure Search Code  \\
& \href{https://jmol.sourceforge.net}{Jmol} & Atomistic & Open-Source Atomistic Visualization Software\\
& \href{http://li.mit.edu/Archive/Graphics/A/}{AtomEye} & Atomistic & Open-Source Atomistic Visualization Software\\
& \href{https://www.ovito.org}{OVITO} & Atomistic & Open-Source Atomistic Visualization Software\\
& \href{https://two.avogadro.cc}{Avogadro 2} & Atomistic & Open-Source Atomistic Visualization Software\\
& \href{https://jp-minerals.org/vesta/en/}{VESTA} & Atomistic & Atomistic Visualization Software\\
& \href{https://pymol.org/2/}{PyMOL} & Atomistic & Molecular Visualization Software\\
& \href{https://www.rdkit.org/}{RDKit} & Cheminformatics & Open-Source Cheminformatics Software \\ 
& \href{https://openbabel.org/wiki/Main_Page}{OpenBabel} & 
Cheminformatics & Open-Source Cheminformatics Software \\ 
& \href{https://vina.scripps.edu/}{AutoDock Vina} & Cheminformatics & Open-Source Molecular Docking \\ 
& \href{https://openmm.org/}{OpenMM} & Molecular Dynamics & Open-Source Molecular Simulation Package \\
& \href{https://www.gromacs.org/}{GROMACS} & Molecular Dynamics & Open-Source Molecular Simulation Package \\
& \href{https://ambermd.org/}{Amber} & Molecular Dynamics & Molecular Simulation Package \\
& \href{https://www.lammps.org/\#gsc.tab=0}{LAMMPS} & Molecular Dynamics & Open-Source Molecular Simulation Package \\
&\href{https://www.mdanalysis.org/}{MDAnalysis} & Molecular Dynamics & Open-Source Python Library for Molecular Dynamics Analysis\\
&\href{https://www.rosettacommons.org/software}{Rosetta} & Biology & Protein Structure Analysis \\ 
&\href{https://www.biotite-python.org/}{Biotite} & Biology & Open-Source Python Library for Computational Molecular Biology \\
&\href{https://biopython.org/}{Biopython} & Biology & Open-Source Python Library for Biological Computation \\ 
& \href{https://scanpy.readthedocs.io/en/stable/}{ScanPy} & Biology & Open-Source Python Library for Single-Cell Analysis \\ 
&\href{http://www.clawpack.org/}{PyClaw} & PDE & Open-Source Finite Volume Numerical Solvers for PDE in Python\\
\bottomrule
\end{tabular}
\end{table}

\begin{table}[!t]\centering
\caption{Learning resources for AI for Science. Note that this table is by no means complete and only consists of resources commonly used by the authors.}\label{tab:airs-edu-ai4science}
\fontsize{6}{8}\selectfont
\begin{tabularx}{\textwidth}{ l|>{\centering\arraybackslash}X c >{\centering\arraybackslash}X}\toprule[1pt]
& AI for Science &Type &Description \\\midrule
\multirow{8}{*}{Workshops} 
&\href{https://ai4sciencecommunity.github.io/}{AI4Science} &General & AI for Science\\
&\href{https://ml4physicalsciences.github.io/}{ML4PS} &General & Machine Learning for Physical Sciences\\
&\href{https://sites.google.com/umn.edu/nsfaiworkshop2023/home}{NSF AI4Science} &General & AI-Enabled Scientific Revolution \\
&\href{https://www.mlsb.io/}{MLSB} & Atomistic & Machine Learning for Structural Biology \\
&\href{https://moleculediscovery.github.io/workshop2022/}{ML4Molecules} & Atomistic & Machine Learning for Molecules\\
&\href{https://sites.google.com/view/ai4mat}{AI4Mat} & Atomistic & AI for Acc. Materials Design\\
&\href{https://jarvis.nist.gov/events/aims}{AIMS} & Atomistic & Artificial Intelligence for Materials Science\\
&\href{https://simdl.github.io/}{SimDL} &Continuum  & Deep Learning for Simulation\\
\midrule
\multirow{2}{*}{Symposiums/Conferences}
&\href{http://cogsys.org/symposium/discovery-2023/schedule.html}{AAAI Spring Symposium} &General & Comp. Approaches to Scientific Discovery\\
&\href{https://www.moml2023.m2d2.io/}{MoML} & Atomistic & Molecular ML Conference\\
\midrule\multirow{9}{*}{Research Institutes and Labs} 
&\href{http://www.ipam.ucla.edu/}{IPAM} &General & Institute for Pure \& Applied Math. at UCLA \\
&\href{https://science.ai.cornell.edu/}{CUAISci} &General & Cornell University AI for Science Institute\\
&\href{https://www.ai4science.caltech.edu/}{AI4Science} &General & AI for Science Initiative at Caltech\\
&\href{https://ai4science-amsterdam.github.io/}{AI4ScienceLab} &General & AI for Science Lab at UvA \\
&\href{https://a3d3.ai/}{A3D3} &General & Acc. AI Algo. for Data-Driven Discovery \\
&\href{https://iaifi.org/}{IAIFI} &General & Institute for AI and Fundam. Interactions \\
&\href{https://datascience.uchicago.edu/research/ai-science/}{AI \& Science} &General & AI \& Science Initiative at UChicago \\
&\href{https://moleculemaker.org/}{Molecule Maker Lab Institute} & Atomistic & AI Institute for Molecule Discovery and Synthesis \\
&\href{https://dynamicsai.org/}{AI Institute in Dynamic Systems} &Continuum  & - \\
\midrule\multirow{7}{*}{Tutorials \& Blogs} 
&\href{https://ai4science101.github.io/}{AI4Science101 Blog Series} &General & -\\
&\href{https://www.anl.gov/ai/reference/ai-for-science-tutorial-series}{AI4Science Tutorial Series} &General & -  \\&\href{http://wangleiphy.github.io/lectures/DL.pdf}{Deep Learning
and Quantum Many-Body Computation} &Quantum & -\\
&\href{https://arxiv.org/pdf/2101.11099.pdf}{Tutorial on Quantum Many-body problem} &Quantum & -\\
&\href{https://zongyi-li.github.io/blog/}{Neural Operator} &Continuum & - \\
&\href{https://benmoseley.blog/my-research/so-what-is-a-physics-informed-neural-network/}{Physics-Informed Neural Networks} & Continuum & - \\
\midrule\multirow{3}{*}{Reading Groups \& Seminars} &\href{https://www.cmu.edu/aced/sciML.html}{Scientific ML Webinar} &General & Scientific Machine Learning Webinar Series\\
&\href{https://psolsson.github.io/AI4ScienceSeminar}{AI4Science Seminar} &General & AI for Science Seminar at Chalmers\\
&\href{https://m2d2.io/talks/m2d2/about/}{M2D2 Reading Group} & Atomistic &  Molecular Modeling
\& Drug Discovery\\
\midrule
\multirow{5}{*}{Courses} 
&\href{https://www.youtube.com/@Eigensteve}{Data-driven Science and Engineering} &General & -\\
&\href{https://uvagedl.github.io/}{Group Equivariant Deep Learning} & General & -\\
&\href{https://symm4ml.mit.edu/symm4ml}{Symmetry and its application to ML} &General & -\\
&\href{https://datascience.uchicago.edu/events/ai-science-summer-school-2023/}{AI for Science Summer School} &General & AI for Science Summer School at UChicago \\
&\href{https://www.youtube.com/watch?v=KIGG-IA9awU\&t=14s}{Crash Course on Neural Operators} &Continuum  & - \\
\midrule\multirow{9}{*}{Software \& Libraries} 
&\href{https://e3nn.org/}{E3NN} & General & Machine Learning and Symmetry Library \\
&\href{https://github.com/divelab/DIG}{DIG} & General & Geometric Deep Learning Library \\
&\href{https://www.netket.org/}{NetKet} &Quantum & Machine Learning for Quantum Physics\\
&\href{https://deepchem.io/}{DeepChem} & Atomistic & Machine Learning for Molecules \\
&\href{https://tdcommons.ai/}{TDC} & Atomistic & Machine Learning for Therapeutic Molecules \\
&\href{https://github.com/deepmodeling/deepmd-kit}{DeePMD} & Atomistic & Deep Learning Interatomic Potential and Force Field\\&\href{https://github.com/yuanqidu/M2Hub}{M$^2$Hub} & Atomistic & Machine Learning for Materials Discovery \\
&\href{https://github.com/google/jax-cfd}{Jax CFD} &Continuum & Machine Learning for Computational Fluid Dynamics \\
&\href{https://github.com/tum-pbs/PhiFlow}{$\Phi_{\text{Flow}}$} & Continuum & Open-Source Python PDE Solver Compatible with Popular Deep Learning Frameworks  \\ 
\midrule\multirow{4}{*}{Competitions \& Benchmarks} &\href{https://opencatalystproject.org/}{Open Catalyst Project} & Atomistic & Discover New Catalyst \\
&\href{https://ogb.stanford.edu/}{Open Graph Benchmark} & Atomistic & Molecular Property Prediction \\
&\href{https://microsoft.github.io/pdearena/}{PDEArena} &Continuum & Operator Learning \\&\href{https://github.com/pdebench/PDEBench}{PDEBench} &Continuum & Operator Learning \\
\midrule\multirow{4}{*}{Review Papers} 
&\href{https://arxiv.org/pdf/1903.10563.pdf}{Machine Learning and Physical Sciences} &General & - \\
&\href{https://pubs.acs.org/doi/10.1021/acs.jpclett.9b03664}{Quantum Chemistry in the Age of Machine Learning} &Quantum & - \\
&\href{https://iopscience.iop.org/article/10.1088/2516-1075/ac572f/meta}{Roadmap on Machine learning in electronic structure} &Quantum & - \\
&\href{https://arxiv.org/abs/2107.01272}{Physics-Guided Deep Learning for Dynamical System} &Continuum & - \\
\bottomrule
\end{tabularx}
\end{table}

\begin{table}[!t]\centering
\caption{Recommended books for fundamental AI, Science, and AI for Science fields. Note that this table is by no means complete and only consists of resources commonly used by the authors.}\label{tab:airs-edu-book}
\fontsize{6}{8}\selectfont
\begin{tabular}{ p{4.0cm} p{3cm} p{1.9cm} p{3.6cm} }
\toprule[1pt]
Title &Author &Domain &Info \\
\midrule
 Deep Learning & Ian Goodfellow, Yoshua Bengio, and Aaron Courville & AI &  2016. MIT Press.~\cite{goodfellow2016deep}\\
 Pattern Recognition and Machine Learning & Christopher M. Bishop and Nasser M. Nasrabadi
 & AI & 2006. Vol. 4. Springer.~\cite{bishop2006pattern}\\
 Machine Learning: A Probabilistic Perspective & Kevin P. Murphy & AI & 2012. MIT Press.~\cite{murphy2012machine}\\
 Advanced Engineering Mathematics &  Erwin Kreyszig & Mathematics & 2011. John Wiley \& Sons.~\cite{kreyszig2007advanced}\\
 The Feynman Lectures on Physics: The New Millennium Edition &Richard Feynman, Robert Leighton, and Matthew Sands &Physics & 2011. Basic Books.~\cite{feynman2011}\\
 Group Theory in a Nutshell for Physicists &Anthony Zee &Group Theory & 2016. Vol.17. Princeton University Press.~\cite{zee2016group}\\
 Group Theory: Application to the Physics of Condensed Matter &Mildred S. Dresselhaus, Gene Dresselhaus, and Ado Jorio &Group Theory & 2007. Springer Berlin, Heidelberg.~\cite{dresselhaus2007group}\\
 Group Theory in Quantum Mechanics: An Introduction to Its Present Usage &Volker Heine &Group Theory & 2007. Courier Corporation.~\cite{heine2007group}\\
 An Introduction to Tensors and Group Theory for Physicists &Nadir Jeevanjee &Group Theory & 2011. Springer.~\cite{jeevanjee2011reprOp}\\
 Symmetry Principles in Solid State and Molecular Physics &Melvin Lax &Group Theory & 2001. Courier Corporation.~\cite{lax2001symmetry}\\
 Introduction to Quantum Mechanics &David J. Griffiths and Darrell F. Schroeter &Quantum Mechanics & 2018. Cambridge University Press.~\cite{griffiths2018introduction} \\
 Modern Quantum Mechanics & J. J. Sakurai and J. Napolitano &Quantum Mechanics & 2020. Cambridge University Press.~\cite{sakurai2020modern}\\
 Quantum Theory of Angular Momentum &D. A. Varshalovich, A. N. Moskalev, and V. K. Khersonskii &Quantum Mechanics & 1988. World Scientific.~\cite{varshalovich1987quantum} \\
 Fundamentals of Condensed Matter Physics &Marvin L. Cohen and Steven G. Louie &Quantum Theory & 2016. Cambridge University Press. ~\cite{CohenLouie2016Fundamentals} \\
 Quantum Theory of Materials &Efthimios Kaxiras and John D. Joannopoulos &Quantum Theory & 2019. Cambridge University Press. ~\cite{Kaxiras_Joannopoulos_2019} \\
  Electronic Structure: Basic Theory and Practical Methods &Richard M. Martin &Quantum Theory & 2020. Cambridge University Press.~\cite{martin2020electronic} \\
 Modern Quantum Chemistry: Introduction to Advanced Electronic Structure Theory &Attila Szabo and Neil S. Ostlund &Quantum Chemistry & 2012. Courier Corporation.~\cite{szabo2012modern} \\
Density-Functional Theory of Atoms and Molecules &Robert G. Parr and Weitao Yang &DFT & 1995. Oxford University Press.~\cite{ParrYang1995DFT} \\
 A Primer in Density Functional Theory &Carlos Fiolhais, Fernando Nogueira, and Miguel A. L. Marques &DFT & 2003. Springer Berlin, Heidelberg.~\cite{Fiolhais2003DFTPrimer} \\
 Density Functional Theory: An Advanced Course &Eberhard Engel and Reiner M. Dreizler &DFT & 2011. Springer Berlin, Heidelberg.~\cite{EngelDreizler2011DFT} \\
 Density Functional Theory: An Approach to the Quantum Many-Body Problem &Reiner M. Dreizler and Eberhard K. U. Gross &DFT & 2012. Springer Berlin, Heidelberg.~\cite{DreizlerGross2012DFT} \\
 Interacting Electrons: Theory and Computational Approaches & Richard M. Martin, Lucia Reining, and David M. Ceperley &DFT & 2016. Cambridge University Press.~\cite{martin2016Interacting} \\
 Density Functional Theory: A Practical Introduction &David S. Sholl and Janice A. Steckel &DFT & 2009. John Wiley \& Sons.~\cite{Sholl2009DFT} \\
 A Chemist’s Guide to Density Functional Theory &Wolfram Koch and Max C. Holthausen &DFT & 2001. John Wiley \& Sons.~\cite{koch2015chemist} \\
 Materials Modelling using Density Functional Theory &Feliciano Giustino &Materials Modeling & 2014. Oxford University Press.~\cite{Giustino2014modeling} \\
 Handbook of Materials Modeling &Sidney Yip &Materials Modeling & 2005. Springer Netherlands.~\cite{Yip2005Handbook} \\
 A Physical Introduction to Fluid Mechanics &Alexander J. Smits &Fluid Mechanics & 2000. John Wiley \& Sons Incorporated.~\cite{smits2000physical} \\
 Lectures in Fluid Mechanic & Alexander J. Smits &Fluid Mechanics & 2009. (MAE 553).~\cite{Smits:notes}\\
 Turbulent Flows &Stephen B. Pope &Fluid Mechanics & 2000. Cambridge University Press.~\cite{pope2000turbulent} \\
 Turbulence, Coherent Structures, Dynamical Systems and Symmetry & Philip Holmes, John L. Lumley, Gahl Berkooz, and Clarence W Rowley &Fluid Mechanics & 2012. Cambridge University Press.~\cite{holmes2012turbulence}\\
 Introduction to Partial Differential Equations &Peter J. Olver &PDE & 2014. Vol.1. Springer.~\cite{olver2014introduction} \\
 Partial Differential Equations & Lawrence C. Evans &PDE & 2022. Vol.19. American Mathematical Society.~\cite{evans2022partial} \\ 
 Geometric Deep Learning: Grids, Groups, Graphs, Geodesics, and Gauges & Michael M. Bronstein, Joan Bruna, Taco Cohen, and Petar Veličković & AI \& Geometry & arXiv preprint arXiv:2104.13478 (2021).~\cite{bronstein2021geometric}\\
 Data-driven Science \& Engineering: Machine learning, dynamical systems, and control & Steven L. Brunton and J. Nathan Kutz & AI \& Engineering & 2022. Cambridge University Press.~\cite{brunton2022data}\\
 Deep Learning for Molecules \& Materials & Andrew D. White & AI \& Atomistic & LiveCoMS 3, 1499 (2021).~\cite{white2021deep}\\
\bottomrule
\end{tabular}
\end{table}

\clearpage
\hypertarget{conclusion}{\section{Conclusion}}
\label{sec:conclusion}

\ifshowname\textcolor{red}{(Yi)}\else\fi Advances in deep learning have revolutionized many artificial intelligence (AI) fields. 
Recently, deep learning has started to advance natural sciences by improving, accelerating, and enabling our understanding of natural phenomena,
giving rise to a new area of research, known as AI for science. From our perspective and that of many others, AI for science opens a door for a new paradigm of scientific discovery and represents one of the most exciting areas of interdisciplinary research and innovation.
Generally speaking, some scientific processes are described with equations that could be too complicated to be solvable, while others are understood from observable data acquired via (expensive) experiments. The mission of AI is to solve such scientific problems accurately and efficiently, along with many other parameters, such as symmetry in AI models, interpretability, out-of-distribution generalization and causality, uncertainty quantification, etc.
In this work, we provide a technical and unified review of several research areas in AI for science that researchers have been working on during the past several years. We organize different areas of AI for science by the spatial and temporal scales at which the physical world is modeled.
In each area, we provide a precise problem setup and discuss the key challenges of using AI to solve such problems. We then provide a survey of major approaches that have been developed, along with datasets and benchmarks for evaluation. We further summarize the remaining challenges and point out several future directions for each area.
Particularly, as AI for science is an emerging field of research,  we have compiled categorized lists of resources in this work to facilitate learning and education.
We understand that given the evolving nature of this area, our work is by no means comprehensive or conclusive. Thus, we expect to continuously include more topics as the area develops and welcome any feedback and comments from the community.
 
\clearpage
\begin{acks}

\noindent
We would like to thank Steven L. Brunton, Petar Veličković, and Max Welling for their valuable discussions.

\vspace{0.1cm} A.A. is supported by the Bren named chair professorship at California Institute of Technology.

\vspace{0.1cm} K.A. and C.W.C. acknowledge support from the National Science Foundation under Grant No. CHE-2202693. K.A. acknowledges additional support from the National Science Foundation's Graduate Research Fellowship Program under Grant No. 2141064.

\vspace{0.1cm} A.F. and M.Z. gratefully acknowledge the support of NIH under No. R01HD108794, US Air Force under No. FA8702-15-D-0001, Kempner Institute for the Study of Natural and Artificial Intelligence, Harvard Data Science Initiative, ASAP Initiative, and awards from Amazon Research, Google Research Scholar Program, Bayer Science, AstraZeneca Research, Roche Alliance with Distinguished Scientists, Sanofi, and Pfizer. A.F. acknowledges the support of the Kempner Institute Graduate Student Fellowship.

\vspace{0.1cm} T.F. and J.S. acknowledge supports from NSF award SCH-2205289, SCH-2014438, IIS-1838042. T.F. acknowledge Wenhao Gao and Yuanqi Du for their help. 

\vspace{0.1cm} X.F. is supported by the MIT-GIST collaboration.

\vspace{0.1cm} N.G and S.G. are Funded by the Federal Ministry of Education and Research (BMBF) and the Free State of Bavaria under the Excellence Strategy of the Federal Government and the Länder.

\vspace{0.1cm} H.J. acknowledges support from the Molecule Maker Lab Institute: an AI research institute program supported by NSF under award No. 2019897 and No. 2034562. The views and conclusions contained herein are those of the authors and should not be interpreted as necessarily representing the official policies, either expressed or implied, of the U.S. Government. The U.S. Government is authorized to reproduce and distribute reprints for governmental purposes notwithstanding any copyright annotation therein.
H.J. is also supported by DOE Center for Advanced Bioenergy and Bioproducts Innovation U.S. Department of Energy, Office of Science, Office of Biological and Environmental Research under Award Number DE-SC0018420.

\vspace{0.1cm} C.K.J. was supported by the A*STAR Singapore National Science Scholarship (PhD). SVM was supported by the UKRI Centre for Doctoral Training in Application of Artificial Intelligence to the study of Environmental Risks (EP/S022961/1).

\vspace{0.1cm} S.J. acknowledges partial supports by the National Science Foundation under grants
IIS-2006861, IIS-1908220, IIS-1955189, and IIS-1908198,
National Institutes of Health under grant U01AG070112,
Cisco Research,
Presidential Impact Fellowship and institutional supports of Texas A\&M University,
and Texas Water Resources Institute.

\vspace{0.1cm} Y.L. is supported by the First Year Assistant Professor (FYAP) Award of Florida State University.

\vspace{0.1cm} A.S. and X.F.Q. gracefully acknowledge the partial supports by the National Science Foundation under grants DMR-1753054, DMR-2103842, OAC-1835690, and CMMI-2226908 and the Air Force Office of Scientific Research (AFOSR) under Grant No. FA9550-24-1-0207. This work was also partially supported by the donors of ACS Petroleum Research Fund under Grant \#65502-ND10.

\vspace{0.1cm} X.N.Q. acknowledges partial supports by the National Science Foundation under grants CCF-1553281, IIS-1812641, OAC-1835690, DMR-2119103, and IIS-2212419. 
Acknowledgment is also made to the Course Development program at Texas A\&M Institute of Data Science.

\vspace{0.1cm} H.S. and T.J. are supported by Machine Learning for Pharmaceutical Discovery and Synthesis (MLPDS) consortium, DARPA Accelerated Molecular Discovery program, and the MIT-GIST collaboration.

\vspace{0.1cm} R.W. and R.Y. are supported in part by the U.S. Department Of Energy ASCR, \#DE-SC0022255, U. S. Army Research Office under Grant W911NF-20-1-0334, Google Faculty Award, Amazon Research Award, Facebook Data Science Award, and NSF Grants 2134274, 2107256 and 2134178.

\vspace{0.1cm} M.X. and J.L. gratefully acknowledge the support of DARPA under Nos. HR00112190039 (TAMI), N660011924033 (MCS); ARO under Nos. W911NF-16-1-0342 (MURI), W911NF-16-1-0171 (DURIP); NSF under Nos. OAC-1835598 (CINES), OAC-1934578 (HDR), CCF-1918940 (Expeditions), NIH under No. 3U54HG010426-04S1 (HuBMAP), Stanford Data Science Initiative, Wu Tsai Neurosciences Institute, Amazon, Docomo, GSK, Hitachi, Intel, JPMorgan Chase, Juniper Networks, KDDI, NEC, and Toshiba.
M.X. and S.E. gratefully acknowledge the support of NSF (\#1651565), ARO (W911NF-21-1-0125), ONR (N00014-23-1-2159), CZ Biohub, Stanford HAI.
M.X. thanks the generous support of Sequoia Capital Stanford Graduate Fellowship.

\end{acks}

\clearpage
\appendix

\section{Classifying and Computing Irreducible Representations} \label{representation_theory}

\noindent{\emph{Authors: YuQing Xie, Tess Smidt}}\newline


Here we give a brief overview of the classification and computation of irreps for various types groups.
A key tool for classifying and computing the irreps of various groups is Schur's lemma.





\begin{theorem}[Schur's lemma]
    Let $\rho_X$ and $\rho_Y$ be irreps of group $G$ acting on vector spaces $X$ and $Y$ respectively. Let $Q:X\to Y$ be a linear map such that $Q\rho_X(g)=\rho_Y(g)Q$ for all $g\in G$. Then $Q$ must be zero or an isomorphism. If $\rho_X=\rho_Y$ and $X=Y$ is finite-dimensional over an algebraically closed field, then $Q$ must be a scalar multiple of the identity.
\end{theorem}

Using Schur's lemma, it is possible to algorithmically decompose any reducible representation into irreps. Suppose we have some reducible representation $\rho_X$ acting on space $X$. Consider the set of linear transformations $Q:X\to X$ such that $Q\rho_X(g)=\rho_X(g)Q$. Note that the condition $Q\rho_X(g)=\rho_X(g)Q$ can be rewritten as $Q\rho_X(g)-\rho_X(g)Q=0$. Since $Q\rho_X(g)-\rho_X(g)Q$ is a linear operation on $Q$, this is just a nullspace problem. In particular, using standard linear algebra techniques we can find a basis $Q_1,Q_2,\ldots,Q_m$ spanning this nullspace.

Suppose $V\subset X$ is a subspace where $\rho_X|_V$ is an irrep and there are no other subspaces isomorphic to this one. Then by Schur's lemma, if we restrict all the $Q_i$ to $V$, they must all either be a multiple of the identity or $0$. In particular, for any linear combination $Q=\sum_{i=1}^r c_iQ_i$, this means that $V$ must be an eigenspace of $Q$. The case where there are multiple copies of the same irrep is more complicated but a similar result holds. Hence, we can pick random coefficients $r_i$ and compute $Q=\sum_{i=1}^r r_iQ_i$. The eigenspaces of this $Q$ will with high probability give us a decomposition of $X$ into irreps.

The method described above is extremely powerful since it gives us a way to decompose a representation into irreps without knowing what the irreps were in the first place.

\vspace{0.1cm}\noindent\textbf{Classifying irreps for Finite Groups:}
While the method described above is great for computing irreps, it does not tell us how to classify them. In particular, we would like a way to identify isomorphic irreps as the same. This motivates the concept of characters and character theory.

\begin{definition}[Character]
    The character of a representation $\rho_X$ is defined as
    \begin{equation}
        \chi_{\rho_X}(g)=\mathrm{Tr}[\rho_X(g)].
    \end{equation}
\end{definition}

One can check that isomorphic representations share the same character. It turns out the converse is true as well, representations with the same character must be isomorphic. Hence, characters give a better way of labelling representations. Further, since conjugation leaves a trace invariant, the character is the same across all elements of $G$ in the same conjugacy class. So it is natural to list the character of a representation as a function of the conjugacy classes of the group. We refer the reader to a group theory textbook for more discussion on characters, such as \textit{Group Theory: Application to the physics of Condensed Matter} \cite{dresselhaus2007group}.

In the case of finite groups, it turns out we can construct a representation which contains all the irreps. This representation is known as the regular representation.

\begin{definition}[Regular representation]
    Let $G$ be a group and let $V$ be a vector space generated by the group (each element $g\in G$ is identified with a basis of $V$). The regular representation $\rho$ is defined by
    \begin{equation}
        \rho(g)h=gh.
    \end{equation}
\end{definition}

We can construct the regular representation of any group using its multiplication table. Decomposing this representation into irreps would give us all possible irreps of the group.

\vspace{0.1cm}\noindent\textbf{Classifying irreps for the Semisimple Lie Groups:}
Classifying the irreps of infinite groups is in general a very hard problem. Using characters fail since we need characters for every group element and there would be infinite many of them. However, for the case of semisimple Lie algebras the classification is well known. This also covers the majority of groups used in scientific applications such as $SO(3)$. Here we give a brief overview of the mathematics involved using $SO(3)$ as an example. We begin by defining Lie groups and Lie algebras.

\begin{definition}[Lie group]
    A Lie group is a group that is also a finite-dimensional smooth manifold. In particular, group multiplication and inversion are smooth maps.
\end{definition}

A simple example of a Lie group is $SO(2)$, the group of 2D rotations. The corresponding manifold for $SO(2)$ is the circle where we can map the polar angle of each point to a counterclockwise rotation by that angle. Another example is $SO(3)$, the group of 3D rotations. The manifold corresponding to $SO(3)$ is more complicated and is the real projective space $\mathbb{RP}^3$.

Because Lie groups are differentiable manifolds, one can instead study a local neighborhood rather than the entire manifold. One typically looks at the tangent space around the identity element in the group. The group multiplication induces a structure on this tangent space known as the Lie bracket. This tangent space along with the Lie bracket structure is known as a Lie algebra. Formally, a Lie algebra is defined as follows.
\begin{definition}[Lie algebra]
    A Lie algebra is a vector space $\mathfrak g$ (over some field $F$) together with a binary operation called the Lie bracket $[\cdot,\cdot]:\mathfrak g \times \mathfrak g \to \mathfrak g$ which satisfies
    \begin{enumerate}
        \item Bilinearity
        \[[ax+by,z]=a[x,z]+b[y,z] \qquad [z,ax+by]=a[z,x]+b[z,y]\]
        for scalars $a,b\in F$ and $x,y,z\in\mathfrak{g}$
        \item Alternativity
        \[[x,x]=0\]
        for $x\in\mathfrak g$
        \item Jacobi identity
        \[[x,[y,z]]+[y,[z,x]]+[z,[x,y]]=0\]
        for all $x,y,z\in\mathfrak g$
        \item Anticommutativity
        \[[x,y]=-[y,x]\]
        for all $x,y\in\mathfrak g.$
    \end{enumerate}
\end{definition}
For matrices, the Lie bracket is just the commutator $[x,y]=xy-yx$.

As an example, let us derive the Lie algebra $\mathfrak{so}(3)$ corresponding to $SO(3)$. Any rotation close to the identity can be written as a perturbation $I+\epsilon X$ where $X$ is in the tangent space around the identity (Note: the physics convention adds an extra factor of $i$ in front of $X$ but the math convention does not). Our main condition is orthogonality, so we must have
\[(I+\epsilon X)^\intercal(I+\epsilon X)=I+\epsilon(X^\intercal+X)+\epsilon^2X^\intercal X=I.\]
Since $\epsilon^2$ is small, the condition on $X$ is $X^\intercal+X=0$ or antisymmetry. One set of bases for this vector space is
\begin{equation}
    x=\begin{pmatrix}
        0 & 0 & 0\\
        0 & 0 & -1\\
        0 & 1 & 0
    \end{pmatrix} \quad
    y=\begin{pmatrix}
        0 & 0 & 1\\
        0 & 0 & 0\\
        -1 & 0 & 0
    \end{pmatrix} \quad
    z=\begin{pmatrix}
        0 & -1 & 0\\
        1 & 0 & 0\\
        0 & 0 & 0
    \end{pmatrix}. \label{adjoint_rep}
\end{equation}
One can check that the commutation relations are $[x,y]=z,[y,z]=x,[z,x]=y$. It is interesting to note that except for a factor of $i$ due to the physics convention, the commutation relations of $\mathfrak{so}(3)$ are exactly the commutation relations of the spin operators. This is not a coincidence, the group used in quantum mechanics to describe spin is $SU(2)$ and the corresponding Lie algebra $\mathfrak{su}(2)$ is the same as $\mathfrak{so}(3)$.

These commutation relations are what define the Lie algebra. A set of matrices such as the ones shown above which satisfy these relations are called representations of the Lie algebra.
\begin{definition}[Lie algebra representation]
    Consider a Lie algebra $\mathfrak g$ and vector space $X$. A representation of $\mathfrak g$ is a pair $(\rho_X,X)$ where
    \[\rho_X:\mathfrak g\to\mathfrak gl(V)\]
    is an algebra homomorphism from $\mathfrak g$ to the general linear algebra of $X$. In particular, that $\rho_X$ is a homomorphism means that
    \[\rho_X\big([A,B]_{\mathfrak{g}}\big) = \big[\rho_X(A),\rho_X(B)\big]_{\mathfrak{gl}(V)} = \rho_X(A)\rho_X(B) - \rho_X(B)\rho_X(A).\]
\end{definition}


Note the similarity to group representations. The definitions of reducible and irreducible representations are analogous. It turns out for a class of Lie algebras known as semisimple Lie algebras, all representations can be reduced to a sum of irreducible ones. This is known as Weyl's theorem on complete reducibility. Hence, classifying irreps of the semisimple Lie algebras classifies all representations. Further, there is also Schur's lemma for irreducible representations of Lie algebras so we can use the algorithm described earlier to also decompose arbitrary Lie algebra representations into irreducible ones.

To classify the representations of the semisimple Lie algebras, one must take a close look at the vector space the representation acts on. Given any matrix and a vector space, one can always split the vector space into eigenspaces of the matrix. However, if there are repeated eigenvalues then this does not completely split the vector space. But if we have a set of commuting matrices, then we can possibly split the vector space more finely. For the semisimple Lie algebras, it turns out there is a particular subalgebra where all elements commute and lets us split the vector space of any representation as finely as possible. This subalgebra is known as a Cartan subalgebra.
\begin{definition}[Cartan subalgebra]
    Suppose we have a Lie algebra $\mathfrak g$. A subalgebra $\mathfrak h$ of $\mathfrak g$ is a Cartan subalgebra if it is
    \begin{enumerate}
        \item Nilpotent. That is the following sequences terminates in the zero subalgebra
        \[\mathfrak h\geq[\mathfrak h,\mathfrak h]\geq[\mathfrak h,[\mathfrak h,\mathfrak h]]\geq[\mathfrak h,[\mathfrak h,[\mathfrak h,\mathfrak h]]]\geq\ldots\]

        \item Self normalizing. That is for all $g\in\mathfrak g$ such that $[g,\mathfrak h]\subset \mathfrak h$, we must have $g\in\mathfrak h$.
    \end{enumerate}
\end{definition}
For $SO(3)$, one choice of a Cartan subalgebra is that spanned by $z$ or $\mathfrak h=Fz$. We can check that $[z,z]=0$ so the subspace is nilpotent. Further, one can check that $[y,z]=x\notin\mathfrak h$ and $[x,z]=-y\notin\mathfrak h$. Note we could have chosen any 1 dimensional subspace such as $x$ or $y$, however the choice of $z$ matches up with the conventions used elsewhere and does not matter for purposes of classifying the representations.

With this subspace, we can then use the representations on this subspace to break up the vector space the representations act on.
\begin{definition}[Weights and weight spaces]
    Let $\mathfrak g$ be a semisimple Lie algebra and let $\mathfrak h$ be a Cartan subalgebra. Let $(\rho_X,X)$ be a representation. Let $\lambda$ be a linear functional $\lambda:\mathfrak h\to \mathbb C$. Then the weight space $V_\lambda$ is the subspace
    \[V_\lambda=\{v\in X:\forall h\in\mathfrak h,\ \rho(h)v=\lambda(h)v\}.\]
    The linear functionals with nonzero weight space are called weights.
\end{definition}
Essentially, the simultaneous eigenspaces are called weight spaces and the corresponding eigenvalues are called weights. Since $\lambda$ is linear on $\mathfrak h$, we can fully specify $\lambda$ with a list of its eigenvalues for a basis of $\mathfrak h$. In particular, if we have an orthonormal basis for $\mathfrak h$ say $b_1,b_2,\ldots,b_n$, then we can define an inner product on the dual space as $\braket{\lambda_1,\lambda_2}=\sum_{i=1}^n\lambda_1(b_i)\lambda_2(b_i)$ so we view it as just a vector of eigenvalues. It turns out there is a unique (up to rescaling) way to define an inner product on the algebra which lets us construct this orthonormal basis. This uses something called an adjoint representation which is similar to the regular representation for finite groups.
\begin{definition}[Adjoint representation, roots, Killing form]
    Let $\mathfrak g$ be a Lie algebra. The adjoint representation is the representation $(\mathrm{ad}_\mathfrak{g},\mathfrak g)$ such that
    \[\mathrm{ad}_\mathfrak{g}(A)B=[A,B]_\mathfrak{g}.\]

    The nonzero weights of the adjoint representation are called roots and the weight spaces are called root spaces. We typically denote the set of roots as $\Phi.$

    The Killing form is an symmetric bilinear form on $\mathfrak g$ defined by
    \[K(A,B)=\mathrm{Tr}[\mathrm{ad}_\mathfrak{g}(A)\circ\mathrm{ad}_\mathfrak{g}(B)]\]
    and can be used to define an inner product.
\end{definition}
One can check that the representation for $\mathfrak{so}(3)$ specified in \eqref{adjoint_rep} is in fact the adjoint representation. Using this, we can check that the eigenvalues of $\mathrm{ad}_{\mathfrak{so}(3)}(z)$ are $1,0,-1$ so $(1),(-1)$ are the roots (Note: we made a choice in our scaling of the eigenvalues. In principle we could have just as well picked $z/2$ and had $1/2,0,-1/2$ but our choice is easier to work with). While in general, the weights can be any vector, there is an important class of weights called integral weights.

\begin{definition}[Integral weight]
    Let $\mathfrak g$ be a Lie algebra and $\mathfrak h$ be a Cartan subalgebra. A weight $\lambda$ is called integral if for all roots $\alpha\in\Phi$,
    \[2\frac{\braket{\lambda,\alpha}}{\braket{\alpha,\alpha}}\]
    is an integer, where $\braket{\cdot,\cdot}$ is the inner product on the dual space as described above.
\end{definition}
In the case of $\mathfrak{so}(3)$, the roots are $(1),(-1)$. Hence $\braket{\alpha,\alpha}=1$ so the integral weights are half integer valued.

While any representation can have multiple weights, it turns out all representations can be uniquely identified by a highest weight. To do this in general, we must pick a set of positive roots $\Phi^+$ and use this to define a partial ordering on the weight space. We can then define a dominant weight as one which always has a nonnegative inner product with the positive roots. We refer the reader to a standard textbook on Lie algebras for more details on these concepts such as \textit{Lie Groups, Lie Algebras, and Representations: An Elementary Introduction} \cite{hall2013lie}. In the case of $\mathfrak{so}(3)$ however, this is easy. We can just pick the positive roots to be $(1)$. Since our vectors are 1-dimensional, we can order our weights just by numeric value. Further, the dominant weights are simply those with nonnegative numerical value.

We now present the main result known as the theorem of highest weight which classifies all irreducible representations of semisimple Lie algebras.
\begin{theorem}[Theorem of highest weight]
    Let $\mathfrak g$ be a finite-dimensional semisimple Lie algebra. Then
    \begin{enumerate}
        \item Let $(\rho_X,X)$ be an irreducible representation. Then $(\rho_X,X)$ has a unique highest weight and the highest weight is dominant and integral
        \item If two representations have the same highest weight, then they are isomorphic
        \item For every dominant integral weight $\lambda$, there is an irreducible representation with highest weight $\lambda$.
    \end{enumerate}
\end{theorem}
Hence, the dominant integral weights classify all the irreducible representations. In the case of $\mathfrak{so}(3)$, this means we can list the representations by nonnegative half integers, corresponding to spin representations and why we can label them by a single number $\ell$. This makes sense since $\mathfrak{so}(3)$ and $\mathfrak{su}(2)$ are equivalent. It turns out only the integer ones correspond to representations of $SO(3)$.
\clearpage
\bibliography{bib/OpenMat,bib/dive,bib/OpenDock,bib/intro,bib/OpenProt,bib/OpenQM,bib/OpenPDE,bib/OpenDFT,bib/Group,bib/OpenMol, bib/OpenSSL, bib/OpenOOD, bib/OpenXAI, bib/OpenUQ,bib/OpenEDU,bib/general-books,bib/tess_paperpile, bib/Aria, bib/tess, bib/michael}

\bibliographystyle{ACM-Reference-Format}






\end{document}